Université de Lille
École Doctorale Mathématiques, sciences du numérique et de leurs interactions

# Thèse de Doctorat

Spécialité Informatique

présentée par
**Patrick Saux**

---

# Mathematics of statistical sequential decision-making: concentration, risk-awareness and modelling in stochastic bandits, with applications to bariatric surgery

---

## Mathématiques de la prise de décision séquentielle statistique : concentration, aversion au risque et modélisation pour les bandits stochastiques, et applications à la chirurgie bariatrique

sous la direction d'**Odalric-Ambrym Maillard** et de **Philippe Preux**.

---

Soutenue publiquement à **Villeneuve d'Ascq**, le **30 janvier 2024** devant le jury composé de

| | | | |
|---|---|---|---|
| Mme. Myriam **Maumy-Bertrand** | MCF, HDR | Université de Technologies de Troyes | Rapporteuse |
| M. Adrien **Saumard** | MCF, HDR | ENSAI Rennes | Rapporteur |
| Mme. Julie **Josse** | DR, HDR | Inria Montpellier | Examinatrice |
| M. Peter **Grünwald** | PROF | CWI Amsterdam | Examinateur |
| M. Rodolphe **Thiébaut** | PU-PH | Université de Bordeaux | Examinateur |
| M. Marc **Tommasi** | PR | Université de Lille | Président du jury |
| M. Odalric-Ambrym **Maillard** | CR, HDR | Inria Lille | Directeur de thèse |
| M. Philippe **Preux** | PR | Université de Lille | Directeur de thèse |
| M. François **Pattou** | PU-PH | Université de Lille et CHU Lille | Invité |

---

Centre de Recherche en Informatique, Signal et Automatique de Lille (CRIStAL),
UMR 9189, Équipe Scool, 59650, Villeneuve d'Ascq, France

*What ?...*

SHAKESPEARE

Des essais ? – Allons donc, je n'ai pas essayé !
Étude ? – Fainéant je n'ai jamais pillé.
Volume ? – Trop broché pour être relié...
De la copie ? – Hélas non, ce n'est pas payé !

Un poëme ? – Merci, mais j'ai lavé ma lyre.
Un livre ? – ...Un livre, encor, est une chose à lire !...
Des papiers ? – Non, non, Dieu merci, c'est cousu !
Album ? – Ce n'est pas blanc, et c'est trop décousu.

Bouts-rimés ? – Par quel bout ?... Et ce n'est pas joli !
Un ouvrage ? – Ce n'est poli ni repoli.
Chansons ? – Je voudrais bien, ô ma petite Muse !...
Passe-temps ? – Vous croyez, alors, que ça m'amuse ?

– Vers ?... vous avez flué des vers... – Non, c'est heurté.
– Ah, vous avez couru l'Originalité ?...
– Non... c'est une drôlesse assez drôle, – de rue –
Qui court encor, sitôt qu'elle se sent courue.

– Du chic pur ? – Eh qui me donnera des ficelles !
– Du haut vol ? Du haut-mal ? – Pas de râle, ni d'ailes !
– Chose à mettre à la porte ? – ...Ou dans une maison
De tolérance. – Ou bien de correction ? – Mais non !

– Bon, ce n'est pas classique ? – À peine est-ce français !
– Amateur ? – Ai-je l'air d'un monsieur à succès ?
Est-ce vieux ? – Ça n'a pas quarante ans de service...
Est-ce jeune ? – Avec l'âge, on guérit de ce vice.

... ÇA c'est naïvement une impudente pose ;
C'est, ou ce n'est pas çà : rien ou quelque chose...
– Un chef-d'œuvre ? – Il se peut : je n'en ai jamais fait.
– Mais, est-ce du huron, du Gagne, ou du Musset ?

– C'est du... mais j'ai mis là mon humble nom d'auteur,
Et mon enfant n'a pas même un titre menteur.
C'est un coup de raccroc, juste ou faux, par hasard...
L'Art ne me connaît pas. Je ne connais pas l'Art.

Préfecture de police, 20 mai 1873.
— Tristan Corbière, "Ça", *Les Amours Jaunes*

## Acknowledgements

First and foremost, I would like to thank my directors Odalric-Ambrym and Philippe: for trusting me, all the way back to the rough times of 2020, and for your guidance, kindness and wisdom during all these years. Odalric, I will cherish the hours spent in your office chatting and drawing obscure inequalities on the board. You are a treasure trove of mathematical knowledge, and I am convinced many more people will follow the creed of *the more applied you go, the stronger theory you need*. Or said differently, *check out saxophone Odalric, he knows all the lemmas*. Philippe, your optimism, your cool and your unfaltering curiosity have been instrumental in every success of the SequeL and Scool family, and I feel so fortunate that our paths crossed back at the MVA. I would also like to thank my unofficial third director, François, who gleefully took me onboard the bariatric roller coaster.

I would also like to express my sincere gratitude to the members of the jury: Myriam and Adrien, for having read and reviewed this doorstopper of a thesis; Peter, from whom I fondly remember the SAVI workshop in Eindhoven and the dinner we shared; Julie and Rodolphe, who I know will contribute to cementing the implication of Inria in the landscape of numerical sciences in healthcare; Marc, for the impromptu and thesis-saving presidency.

Although ascetic at times, research is, for the most part, a team effort, and for this I am thankful to my all coauthors: Dorian, who truly was the catalyser of my research on stochastic bandits — *sickness for the thickness: a tale of tails*; Pierre, for the countless hours spent at the CHU; Julien, Maxence and Tomy; Romain, for introducing me to the world of agriculture; Sayak and Aditya; and all the bariatric gang: Violeta, Hélène, Robert, Iva, Céline and so many more. Here's to all the others who also participated, one way or another, at the lab or elsewhere: Émilie, Deb (I am still bitter we lost your watch that night), Rémy, Timothée, Alena, Riccardo, Nathan, Reda, Sarah, Mathieu, Marc, Matheus, Hector, Cyrille, Sumit, Hernan, Waris, Thomas, Clémence, Shubhada, Benoîte. Last but not least, a special thanks to my brother-in-arms, my companion of many coffees, angers and laughters, a *forban* of science, Fabien.

I would also like to thank my former colleagues and mentors for their lasting impact on me: Argyris, Nicolas, Andrey, Terry, Camille, Jérôme, Tom, Darren.

Dearest friends, some of whom crossed oceans to attend the defence, here's to you: Hugo (it has been twenty years!), Thibaut and Pauline, without whom I probably never would have taken the leap back to academia; Flochtr... Flora; Florian, the fastest man alive; Mister Артём; the notorious J.C.D.; Céline, Nicolas – *Harry ! Jack !*; Alexandra, Loïs; Clément; Ritavan; la baronne Marie, Maria, Yanne, *Chris* et tout le kibboutz; Kevin – *remonterais-je ma parole pour être en phase avec l'heure dite ?*; Ismail – حبيبى لا شكرة على واجب.

Dear family, Maman, Papa, for your unconditional support and love; Pascale, François, Mathilde, Élise, for your warm welcome and kindness in the past thirteen years – thank you.

Finally, a special thank you is due to the one with whom I shared most of my time during this thesis, Phika \-;——;^=. And of course, none of this would mean anything or even have been remotely possible if not for Cécile. Thank you. I love you.

# Abstract

This thesis aims to study some of the mathematical challenges that arise in the analysis of statistical sequential decision-making algorithms for postoperative patients follow-up. Stochastic bandits (multiarmed, contextual) model the learning of a sequence of actions (policy) by an agent in an uncertain environment in order to maximise observed rewards. To learn optimal policies, bandit algorithms have to balance the exploitation of current knowledge and the exploration of uncertain actions. Such algorithms have largely been studied and deployed in industrial applications with large datasets, low-risk decisions and clear modelling assumptions, such as clickthrough rate maximisation in online advertising. By contrast, digital health recommendations call for a whole new paradigm of small samples, risk-averse agents and complex, nonparametric modelling. To this end, we developed new safe, anytime-valid concentration bounds, (Bregman, empirical Chernoff), introduced a new framework for risk-aware contextual bandits (with elicitable risk measures) and analysed a novel class of nonparametric bandit algorithms under weak assumptions (Dirichlet sampling). In addition to the theoretical guarantees, these results are supported by in-depth empirical evidence. Finally, as a first step towards personalised postoperative follow-up recommendations, we developed with medical doctors and surgeons an interpretable machine learning model to predict the long-term weight trajectories of patients after bariatric surgery.

# Résumé

Cette thèse porte sur l'étude des défis mathématiques liés à l'analyse d'algorithmes de prise de décision séquentielle pour le suivi postopératoire de patients. Les bandits stochastiques (multi-bras, contextuels) modélisent l'apprentissage d'une suite d'actions (politique) par un agent évoluant dans un environnement incertain afin de maximiser les récompenses observées. Afin d'apprendre de telles politiques optimales, les algorithmes de bandits se confrontent au dilemme de l'exploitation de l'information déjà acquise et de l'exploration d'actions incertaines. De tels algorithmes ont été amplement étudiés et déployés dans le cadre d'applications industrielles faisant intervenir de grands jeux de données, un faible risque pour chaque décision, ainsi que des hypothèses de modélisation bien définies, comme par exemple la maximisation du taux de clic dans la publicité en ligne. À l'inverse, le problème des recommandations de santé requiert un tout nouveau paradigme de petits échantillons, d'aversion au risque et de modélisation complexe et non paramétrique. Dans cette optique, nous avons développé de nouvelles bornes de concentration uniforme en temps (Bregman, Chernoff empirique), introduit un nouveau cadre de bandits contextuels conscients des risques (au moyen de mesure de risque élicitables), et analyser une nouvelle classe d'algorithmes de bandits non paramétriques sous des hypothèses faibles (échantillonnage de Dirichlet). Outre des garanties théoriques, ces résultats sont étayés par des observations empiriques. Nous avons enfin développé, en coordination avec une équipe de médecins et chirurgiens dans le cadre d'un projet de suivi postopératoire, un modèle interprétable par apprentissage automatique de prédiction de la trajectoire de poids à long terme de patients après chirurgie bariatrique.

# Contents

# List of acronyms

**AGB** Adjustable gastric banding

**BCP** Boundary crossing probability

**BMI** Body mass index

**c.d.f.** Cumulative distribution function

**CART** Classification and regreesion tree

**CGF** Cumulant generating function

**DS** Dirichlet sampling

**EF** Exponential family

**EWL** Excess weight loss

**i.i.d.** Independent and identically distributed

**LASSO** Least absolute shrinkage and selection operator

**MAD** Median absolute deviation

**MAP** Maximum a posteriori

**MGF** Moment generating function

**MLE** Maximum likelihood estimator

**MV** Mean-variance

**OGD** Online gradient descent

**p.d.f.** Probability density function

**RCT** Randomised controlled trial

**RMSE** Root mean squared error

**RYGB** Roux-en-Y gastric bypass

**SDE** Stochastic differential equation

**SG** Sleeve gastrectomy

**SOSG** Second order sub-Gaussian

**SPEF** Single parameter exponential family

**SSOSG** Symmetrised second order sub-Gaussian

**T2D** Type 2 diabetes

**TS** Thompson sampling

**TWL** Total weight loss

**UCB** Upper confidence bound

# List of symbols

**Mathematical notations**

| | |
|---|---|
| $\mathbb{R}$ | Set of real numbers |
| $\mathbb{C}$ | Set of complex numbers |
| $\mathbb{N}$ | Set of integers |
| $\mathcal{X}_{\pm}$ | Set of nonnegative (respectively nonpositive) elements of a set $\mathcal{X} \subseteq \mathbb{R}$ |
| $\mathcal{X}^{\star}$ | Set of nonzero elements of a set $\mathcal{X} \subseteq \mathbb{C}^d$ |
| $\mathcal{X}^d$ | $d$-fold Cartesian product of a set $\mathcal{X}$ |
| $\mathfrak{P}(\mathcal{X})$ | Set of all subsets of a set $\mathcal{X}$ |
| $\lvert\mathcal{X}\rvert$ | Cardinal of a set $\mathcal{X}$ |
| $\mathcal{X}^{(\mathbb{N})}$ | Set of finite sequences of elements of $\mathcal{X}$ |
| $X \sqcup Y$ | Concatenation of two sequences $X$ and $Y$ |
| $\mathcal{M}_d(\mathcal{X})$ | Set of square matrices of size $d$ over a set $\mathcal{X}$ |
| $\mathcal{S}_d(\mathcal{X})$ | Set of symmetric square matrices of size $d$ over a set $\mathcal{X}$ |
| $\mathcal{S}_d^{+(+)}(\mathcal{X})$ | Set of symmetric positive (semi)definite square matrices of size $d$ over a set $\mathcal{X}$ |
| $[K]$ | Set of integers from $1$ to $K \in \mathbb{N}$ |
| $\mathfrak{D}^n$ | $n$-simplex, i.e. $\{w \in [0,1]^{n+1},\ \sum_{i=1}^{n+1} w_i = 1\}$ |
| $\mathfrak{S}_n$ | Symmetric group, i.e. group of permutations of $\{1, \ldots, n\}$ |
| $\langle x, y\rangle$ | Euclidean scalar product of vectors $x, y \in \mathbb{R}^d$, $\langle x, y\rangle = \sum_{i=1}^d x_i y_i$ |
| $\lVert x\rVert_2$ | Euclidean norm of a vector $x \in \mathbb{R}^d$, $\lVert x\rVert_2 = \sqrt{\sum_{i=1}^d x_i^2}$ |
| $\lVert x\rVert_\infty$ | Supremum norm of a vector $x \in \mathbb{R}^d$, $\lVert x\rVert_\infty = \max_{i=1,\ldots,d}\lvert x_i\rvert$ |
| $\langle x, y\rangle_A$ | Scalar product of vectors $x, y \in \mathbb{R}^d$ induced by a matrix $A \in \mathcal{S}_d^{++}(\mathbb{R})$ |

## List of symbols

| | |
|---|---|
| $\|x\|_A$ | Euclidean norm of a vector $x \in \mathbb{R}^d$ induced by a matrix $A \in \mathcal{S}_d^{++}(\mathbb{R})$ |
| $\mathbb{B}^d_{\|\cdot\|}(x, R)$ | Ball in $\mathbb{R}^d$ of radius $R$ centered on $x$ with respect to the norm $\|\cdot\|$ |
| $A \preccurlyeq B$ | Loewner order for matrices $A, B \in \mathcal{S}_d(\mathcal{X})$, i.e. $B - A \in \mathcal{S}_d^+(\mathcal{X})$ |
| $\lambda_{\min}(A)$ | Smallest eigenvalue of a matrix $A \in \mathcal{M}_d(\mathcal{X})$ |
| exp | Exponential function $\mathbb{C} \to \mathbb{C}^*$ |
| log | Natural logarithm $\mathbb{R}_+^* \to \mathbb{R}$ |
| $\mathrm{Tr} A$ | Trace of a matrix $A \in \mathcal{M}_d(\mathbb{C})$ |
| $\det A$ | Determinant of a matrix $A \in \mathcal{M}_d(\mathbb{C})$ |
| $f = \mathcal{O}(g)$ | $f$ is bounded above by $g$ asymptotically (up to a constant), i.e. $\limsup \frac{f}{g} < +\infty$ |
| $f = o(g)$ | $f$ is dominated by $g$ asymptotically, i.e. $\lim \frac{f}{g} = 0$ |
| $f = \Omega(g)$ | $f$ is bounded below by $g$ asymptotically (up to a constant), i.e. $\liminf \frac{f}{g} > 0$ |
| $f = \widetilde{\mathcal{O}}(g)$ | There exists $\beta \in \mathbb{R}$ such that $f = \mathcal{O}\left(g \log(g)^\beta\right)$ |
| $f = \Theta(g)$ | $f = \mathcal{O}(g)$ and $f = \Omega(g)$ |
| $f \sim g$ | $f$ is equivalent to $g$ asymptotically, i.e. $\lim \frac{f}{g} = 1$ |
| $x \vee y$ | Maximum of two elements $x$ and $y$ of an ordered set |
| $x \wedge y$ | Minimum of two elements $x$ and $y$ of an ordered set |
| $\mathrm{dom} f$ | Domain of a function $f$, $\mathrm{dom} f = \{x \in \mathcal{X}, f(x) < +\infty\}$ if $f : \mathcal{X} \to \mathbb{R} \cup \{\pm\infty\}$ convex |
| $f\|_A$ | Restriction of a function or a measure $f$ to a domain $A$ |
| $f^\star$ | Fenchel-Legendre conjugate of a function $f : \mathbb{R}^d \to \mathbb{R}$ |

### Probability and statistics

| | |
|---|---|
| $\mathbb{P}$ | Probability measure |
| $\mathbb{E}[X]$ | Expectation of a random variable $X$ |
| $\mathbb{V}[X]$ | Variance (or covariance matrix) of a random variable $X$ |
| $\mathbb{P}_\theta$ | Probability measure under a statistical model described by $\theta$ |
| $\mathbb{E}_\theta[X]$ | Expectation of a random variable $X$ under a statistical model described by $\theta$ |
| $\mathbb{V}_\theta[X]$ | Variance of a random variable $X$ under a statistical model described by $\theta$ |
| $X \sim \nu$ | A random variable $X$ follows a probability distribution $\nu$ |
| $X \overset{\mathcal{D}}{=} Y$ | Random variables $X$ and $Y$ are equal in distribution |
| $\mathcal{M}_1^+(\mathcal{X})$ | Set of probability measures on a measurable space $\mathcal{X}$ |
| $\mathbb{L}^p(\mathcal{X})$ | Set of probability measures $\nu \in \mathcal{M}_1^+((\mathcal{X}, \|\cdot\|))$ such that $\mathbb{E}_{X\sim\nu}[\|X\|^p] < +\infty$ |
| $\mathbb{L}^\rho(\mathcal{X})$ | Set of probability measures $\nu \in \mathcal{M}_1^+((\mathcal{X}, \|\cdot\|))$ such that $\mathbb{E}_{X\sim\nu}[\|\rho(X)\|] < +\infty$ |
| $\bar{A}$ | Complement of an event $A \subseteq \Omega$, i.e. $\bar{A} = \Omega \setminus A$ |
| $\mathrm{KL}(\nu \parallel \nu')$ | Kullback-Leibler divergence between distributions $\nu$ and $\nu'$ |
| $\delta_{\mathrm{TV}}(\nu, \nu')$ | Total variation distance between distributions $\nu$ and $\nu'$ |
| $\mathcal{U}(\mathcal{X})$ | Uniform distribution over a compact set $\mathcal{X}$ |
| $\mathcal{N}(\mu, \sigma^2)$ | Gaussian distribution with mean $\mu \in \mathbb{R}$ and standard deviation $\sigma \in \mathbb{R}_+^*$ |
| $\mathcal{E}(\lambda)$ | Exponential distribution with intensity $\lambda \in \mathbb{R}_+^*$ |
| $\mathcal{B}(p)$ | Bernoulli distribution with success probability $p \in [0, 1]$ |
| $\mathcal{P}(\lambda)$ | Poisson distribution with intensity $\lambda \in \mathbb{R}_+^*$ |
| $\chi^2(k)$ | Chi-square distribution with $k \in \mathbb{R}_+^*$ degrees of freedom |
| $\mathrm{Dir}(\alpha)$ | Dirichlet distribution on the $n$-simplex with weights $\alpha \in (\mathbb{R}_+^*)^{n+1}$ |
| $\delta_x$ | Dirac distribution on point $x$ |
| $\widehat{\nu}_{\mathcal{X}}$ | Empirical distribution of the finite set $\mathcal{X}$, i.e. $\widehat{\nu}_{\mathcal{X}} = \frac{1}{\lvert\mathcal{X}\rvert}\sum_{x\in\mathcal{X}}\delta_x$ |
| CVaR | Conditional Value at Risk, left tail |
| $\mathcal{F}_{\mathcal{G},R}$ | Family of $R$-sub-Gaussian distributions |
| $\mathcal{F}_\ell$ | Family of light tailed distributions |
| $\mathcal{F}_{M,\varepsilon}$ | Family of distributions with raw $1+\varepsilon$-th moment bounded by $M$ |
| $\mathcal{F}_{\kappa,\varepsilon}^{\textbf{centred}}$ | Family of distributions with central $1+\varepsilon$-th moment bounded by $\kappa$ |

**Stochastic bandits**

| | |
|---|---|
| $K$ | Number of arms |
| $\pi_t$ | Policy, i.e. arm played, at time $t$ |
| $N_t^k$ | Number of pulls to arm $k$ at time $t$ (under a given policy) |
| $\rho$ | Risk measure $\mathcal{M}_1^+(\mathbb{R}) \to \mathbb{R} \cup \{\pm\infty\}$ |
| $\mathcal{R}_T$ | Cumulative pseudo regret (under a given policy and risk measure) at time $T$ |
| $\rho^\star$ | Optimal risk measure across arms |
| $\mathcal{K}_{\inf}^{\mathcal{F}}$ | Extremal Kullback-Leibler operator over family $\mathcal{F}$ |

## List of symbols

| | |
|---|---|
| $X_t$ | Action in $\mathbb{R}^d$ played at time $t$ |
| $Y_t$ | Reward in $\mathbb{R}$ received at time $t$ |

### Exponential families

| | |
|---|---|
| $\Theta$ | Finite dimensional set of parameters, $\Theta \subset \mathbb{R}^d$ |
| $\mathcal{L}$ | Log-partition function $\Theta \to \mathbb{R}$ |
| $\mathcal{B}_{\mathcal{L}}(\theta', \theta)$ | Bregman divergence with potential function $\mathcal{L}$ between parameters $\theta$ and $\theta'$ |

### Empirical Chernoff

| | |
|---|---|
| $\rho$ | Ratio of first and second order sub-Gaussian parameters |
| $\mathcal{F}^2_{\mathcal{G},\rho}$ | Family of second order sub-Gaussian distributions |
| $\mathring{\mathcal{F}}^2_{\mathcal{G},\rho}$ | Family of symmetrised second order sub-Gaussian distributions |
| $W_k$ | $k$-th branch of the Lambert $W$ function $(k \in \{-1, 0\})$ |
| $U(b, c; z)$ | Tricomi's confluent hypergeometric function with parameters $b$, $c$, variable $z$ |
| $K(b, c; z)$ | Kummer's confluent hypergeometric function with parameters $b$, $c$, variable $z$ |

### Risk-aware bandits

| | |
|---|---|
| $\mathcal{L}$ | Strongly convex risk function $\mathbb{R} \times \mathbb{R}^p \to \mathbb{R}$ |
| $\partial^j \mathcal{L}$ | $j$-th derivative with respect to the second argument of $\mathcal{L}$ |
| $m$ | Minimum curvature of $L$, i.e. $m = \inf \partial^2 \mathcal{L}$ |
| $M$ | Maximum curvature of $L$, i.e. $M = \sup \partial^2 \mathcal{L}$ |
| $V_t^\alpha$ | Global Hessian at time $t$ with regularisation $\alpha$ |
| $H_t^\alpha$ | Local Hessian at time $t$ with regularisation $\alpha$ |
| $e_p$ | $p$-expectile |

### Signage

| | |
|---|---|
| | Original contribution within the introduction |
| | Probability and statistics content |
| | Bandits content |
| | Bariatric surgery content |
| | Content skippable at first reading, published elsewhere |

# Foreword

Ἀγεωμέτρητος μηδεὶς εἰσίτω

## Statistical sequential decision-making in healthcare

According to the American college of surgeons, “the surgeon is responsible for the preoperative diagnosis of the patient, for performing the operation, and for providing the patient with postoperative surgical care and treatment” (American college of surgeons, 2023). Regarding the last point, let us say, for the sake of simplicity, that the surgeon has two possible recommendations for postoperative follow-up. The first option is to schedule an appointment directly at the surgery department, where the patient can be examined by a large, specialised and possibly multidisciplinary team. The second option is to refer the patient to their general practitioner of choice. While the former ensures a priori the best care, it is also the most costly and likely the most impractical for the patient (longer waiting times, longer commute, etc.), and therefore the latter may be an appealing alternative for patients with no major postoperative complications. In addition, let us imagine that a third option has recently emerged in the form of remote electronic consultation, which is even less costly than a face-to-face visit to the general practitioner. Which of these three options should the surgeon recommend?

Ignoring the part of subjectivity that feeds into such a decision process [1], let us say that the surgeon has asked a team of statisticians to compute a *postoperative care score* to help inform their decision. This imaginary score[2] is a single number, normalised between 0 and 100, that reflects the cost-benefit analysis of each alternative and has been calibrated on medical records of recent patients of the surgery department. Naturally, every patient has responded differently to postoperative care, and thus the scores are modelled as random variables. We illustrate in Figure 1 the mean score (± standard deviation) observed for each follow-up option.

[1]This thesis is, after all, about quantitative sciences. This is not to say that this approach is right for medical recommendations. Social sciences, economics, geopolitics and many other fields have a lot to say too on this topic.

[2]We do not discuss either *how* such scores could be computed. Implementing these in practice would require a significant interdisciplinary effort and prospective testing to ensure they incentivise desirable behaviours.

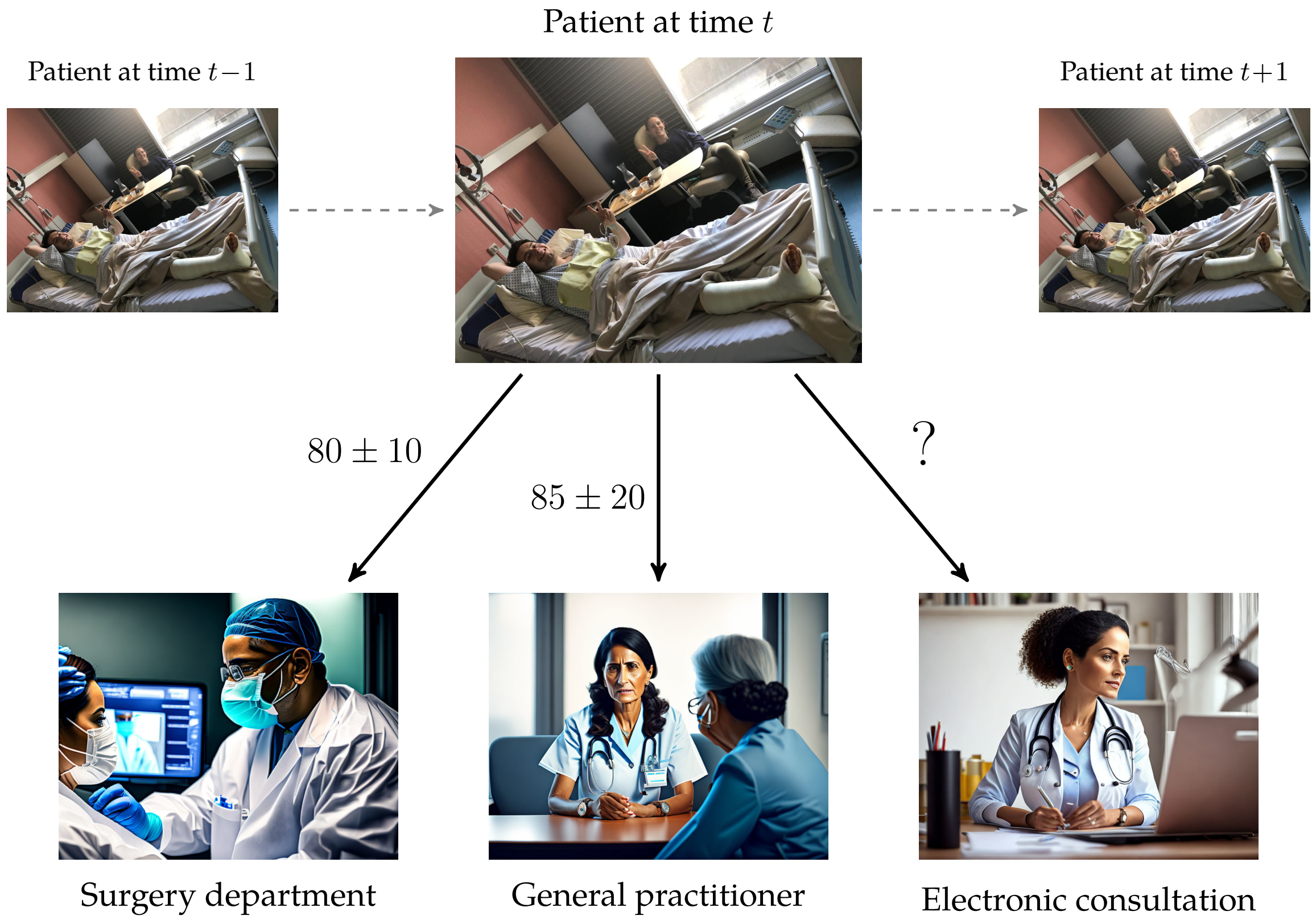

**Figure 1** – Should the surgeon recommend postoperative follow-up at the surgery department, the safest but most costly option? At the general practitioner, less costly but also less specialised? Using electronic consultation, which has seldom been tried?[3]

A simple, naive solution would be to send every patient to the general practitioner, which exhibited the highest postoperative score (85). However, due to the inherent randomness of patients' responses, and the potentially small sample size (only recent patients at this specific surgery department were accounted for in these statistics), it is still possible that another option would offer a better mean score. Furthermore, future patients, including those who were subjected to this recommender system, may be used to continuously update these statistics, and therefore the recommended option may change with additional data.

This idealised toy model is an example of a *stochastic bandit*. Informally, a bandit represents a game where an *agent* (the player) picks an *arm* among a set of possible alternatives (here, the three follow-up pathways) and receives a *reward* (here, the postoperative score). The goal of the player is to maximise (a certain statistic of) the total reward accumulated during the game.

[3]The top image is a direct contribution of the author. Credits: Cécile Gauthiez, Flora Tixier. The bottom images were generated from text prompts using https://deepdreamgenerator.com/generator.

More generally, bandits belong to a field of quantitative methods at the crossroads of mathematics, computer science and machine learning, known as

statistical sequential decision-making

- data-driven, provable probabilistic performance guarantees, uncertainty quantification;
- prospective design, online recommendations, continuous learning;
- actions are recommended to an agent interacting with an environment to maximise rewards.

Rather than a single model, stochastic bandits form a large class of methods. A simple branching decision pattern as described above is called a *multiarmed bandit*: patients are considered equal and the agent is looking for the best *one-size-fits-all* solution. Realistically, designing postoperative follow-up programmes calls for more sophisticated solutions. When the reward is a function of one or several contextual variables that change between rounds, this model becomes a *contextual bandit*. For instance, every round may represent a new patient with some clinical characteristics (sex, weight, height, age, etc.) that condition the postoperative scores (e.g. the score of the surgery department follow-up is higher for older patients as they are more at risk of developing complications and thus require more specialised care). Even more sophisticated approaches form the basis of *reinforcement learning*, which studies agents that are able to learn autonomously in complex environments. All these models are part of what is colloquially (and often somewhat abusively) known as artificial intelligence.

## A success story in online advertising

Industrial applications of statistical sequential learning soared with the widespread use of Internet starting in the early 2000s. Indeed, the core businesses of many technology companies revolve around providing accurate recommendations to customers: which ad to click on when using a search engine, which movie to watch on a streaming platform, which item to purchase on an online marketplace, etc. As an example of this growing interest, Netflix hosted from 2006 to 2009 an open competition called the *Netflix Prize* for predicting user ratings for films using collaborative filtering, a core component, alongside stochastic bandit methods, of modern recommender systems. In terms of revenues, Amazon reported a positive impact of the use of personalised recommendations of 35% through cross-sales (Jannach and Jugovac, 2019). In 2020, the global recommender system market size was valued at USD 1.8 billion and is expected to continue its fast expansion (Grand View Research, 2020).

Beyond the obvious business incentive, we posit that the success of recommender systems, and bandits in particular, in online marketing is also due to the conjunction of favourable modelling and data environments:

- data is available in huge quantities (e.g. millions of customers) and quality (cookies, metadata, purchase or viewing history, etc.);
- each decision is virtually risk-free: although maintaining engagement in the long run can be challenging, it is unlikely that a customer will stop interacting with the platform altogether if a single recommendation is unsatisfying;
- many recommender systems are essentially optimisers of *clickthrough rates*, i.e. the expected frequencies of a positive response from a customer exposed to a given ad, which makes for a clear, well-studied theoretical setting (Bernoulli trials, logistic models, etc.).

## Towards recommender systems in healthcare

Given the successes of sequential learning in online advertising, it is tempting to replicate this approach in other fields. In this thesis, we focus primarily on healthcare applications, where we seek to recommend efficient postoperative pathways (follow-up visits at the hospital versus at the general practitioner as seen above, but also *when* to schedule such visits). However, we identified several drawbacks that hinder the deployment of existing approaches and call for new theoretical and methodological groundwork.

- **Small samples**: clinical trials require careful preparation and usually concern at most thousands of individuals. Examining electronic registries or national health databases allows to scale up, albeit with typically much less precise or reliable data.
- **Risk**: even a single inappropriate decision can be detrimental to a patient (if for instance an algorithm suggested to skip the next follow-up visit that would have made it possible to detect a life-threatening complication). Healthcare is hardly a numbers game where we seek to optimise the *average* outcome.
- **Modelling**: medical datasets often comprise heterogeneous variables, which may not fit the scope of classical statistical models. Moreover, the constraint of small samples prevents any large numbers effect (normal distribution, central limit theorem). The main criterion to optimise is not necessarily a binary outcome (Bernoulli trial), but may be a continuous target ("what dose of a certain drug should a patient take"), a date ("when should a patient schedule a follow-up visit"), etc.

This thesis explores some mathematical aspects of these three challenges, and fits within a long-term collaboration between Inria (team Scool) and the bariatric surgery department at Lille University Hospital to develop such recommender systems for the treatment of chronic diseases, and in particular obesity.

These roadblocks towards practical recommender systems are however not specific to healthcare, and we hope this line of research fosters new developments in other critical areas. A complementary field of applications we consider in this manuscript is the recommendation of agricultural practices, which shares many of the technical challenges that we face in healthcare: small data (e.g. it takes a whole year to observe the effect of new crop management policies on the harvest yield), risk (e.g. crop mismanagement can threaten food security in developing countries) and modelling complexity. A major difference however is the availability of efficient, open-source simulators for crop yields supported by engineering knowledge and largely accessible data, allowing us to perform synthetic experiments on which to test our algorithms. This is in stark contrast with medical data, which are highly sensitive and protected, and also exceedingly complex to simulate accurately. For these reasons, we use agricultural data as a proof of concept for the methods we develop here, hoping for later use in healthcare.

# List of contributions

### Publications in peer-reviewed journals

- Patrick Saux, Pierre Bauvin, Violeta Raverdy, Julien Teigny, Hélène Verkindt, Tomy Soumphonphakdy, Maxence Debert, David Nocca, Carel W. Le Roux, Robert Caiazzo, Philippe Preux, François Pattou, et al. Development and validation of an interpretable machine learning based calculator for predicting 5 year-weight trajectories after bariatric surgery: a multinational retrospective cohort SOPHIA study. *The Lancet Digital Health*, 2023 (Chapter 8)

- Robert Caiazzo, Pierre Bauvin, Camille Marciniak, Patrick Saux, Geoffrey Jacqmin, Raymond Arnoux, Salomon Benchetrit, Jerome Dargent, Jean-Marc Chevallier, Vincent Frering, et al. Impact of robotic assistance on complications in bariatric surgery at expert laparoscopic surgery centers. a retrospective comparative study with propensity score. *Annals of Surgery*, pages 10–1097, 2023 (Chapter 8)

### Publications in peer-reviewed international conferences with proceedings

- Sayak Ray Chowdhury, Patrick Saux, Odalric-Ambrym Maillard, and Aditya Gopalan. Bregman deviations of generic exponential families. In *Conference On Learning Theory*, pages 394–449. PMLR, 2023 (Chapter 3)

- Patrick Saux and Odalric-Ambrym Maillard. Risk-aware linear bandits with convex loss. In *International Conference on Artificial Intelligence and Statistics*, pages 7723–7754. PMLR, 2023 (used in Chapter 5)

- Dorian Baudry, Patrick Saux, and Odalric-Ambrym Maillard. From optimality to robustness: Adaptive re-sampling strategies in stochastic bandits. *Advances in Neural Information Processing Systems*, 34:14029–14041, 2021c (Chapter 6)

## Workshop presentations in peer-reviewed international conferences

- Patrick Saux and Odalric-Ambrym Maillard. Risk-aware linear bandits with convex loss. In *European Workshop on Reinforcement Learning*, 2022 (Chapter 5)

## Software

- Patrick Saux, Pierre Bauvin, Julien Teigny, and Maxence Debert. Bariatric weight trajectory prediction. https://bariatric-weight-trajectory-prediction.univ-lille.fr, 2022 (Chapter 8)
  - Implementation of the ML pipeline and the smoothing of weight trajectories in Python (server), maintenance of the Javascript user interface (client).
  - Digital deposit IDDN.FR.001.370002.000.S.P.2022.000.31230, Agence pour la Protection des Programmes (APP), Inria, CHU Lille, Université de Lille.
  - As of October 2023, a live demonstrator is available at the *Interface* space, Inria Lille.
- https://github.com/rlberry-py/rlberry
  - Implementation of the multiarmed bandit agents and environments.
- https://pypi.org/project/concentration-lib (Chapter 1, Chapter 3, Chapter 4)

## Unpublished collaborations not presented in this thesis

- Romain Gautron, Patrick Saux, Odalric-Ambrym Maillard, Marc Corbeels, Chandra A. Madramootooe, and Nitin Joshi. Quantifying the uncertainty of crop management decisions based on crop model simulations. Unpublished
- Benoîte de Saporta, Odalric-Ambrym Maillard, and Patrick Saux. Piecewise deterministic Markov decision processes: deterministic flows meet stochastic actions. Unpublished

## Reviewing

- Conferences: AISTATS 2022 (top reviewer), AISTATS 2023 (top reviewer), ALT 2024 (upcoming), AISTATS 2024 (upcoming).
- Workshop: EWRL 2022.

# Part I

# Introduction

# Chapter 1

# Background material on statistical sequential decision-making models

*Two roads diverged in a wood, and I —*
*I took the one less traveled by,*
*And that has made all the difference.*
— Robert Frost, *The Road Not Taken*

In this introductory chapter, we present an overview of the main mathematical tools to analyse statistical sequential decision-making problems. As introduced in the foreword, the main models we study in this thesis are stochastic bandits, which form a subclass of the larger field of reinforcement learning that is especially suited for rigorous mathematical analysis. A critical assumption in any bandit problem is to specify a statistical model for the reward distributions, and we present standard examples such as sub-Gaussian, bounded, exponential families, and others. Finally, a pivotal concept in both the design and analysis of bandit algorithms is the concentration of measure phenomenon, for which we provide a detailed introduction, completed with original results.

**Contents**

## 1.1 Mathematical setting and notations

We consider a topological space $(\Omega, \mathcal{T})$ and further assume that it is separable (there exists a countable dense subset), metrisable (there exists a metric $d_{\mathcal{T}}$ such that $d_{\mathcal{T}}$ induces the topology $\mathcal{T}$) and that the metric space $(\Omega, d_{\mathcal{T}})$ is complete (every Cauchy sequence converges); in other words, $(\Omega, d_{\mathcal{T}})$ is a Polish space. We let $\mathcal{A}$ be the corresponding Borel $\sigma$-algebra, i.e. the smallest $\sigma$-algebra containing the open sets of $(\Omega, \mathcal{T})$, which we denote by $\mathcal{A} = \sigma(\mathcal{T})$, and $\mathbb{P}$ the corresponding Borel measure. We further assume that $\mathbb{P}$ is positive, locally finite and inner regular ($\forall A \in \mathcal{A},\ \mathbb{P}(A) = \sup\{\mathbb{P}(C),\ C \subset A \text{ compact}\}$) and that $\mathbb{P}(\Omega) = 1$; in other words, $\mathbb{P}$ is a (Radon) probability measure, and $(\Omega, \mathcal{A}, \mathcal{P})$ is a probability space. We refer to Rudin (1987) for background material on measure theory.[1] The **Borel-Cantelli lemma** states that if a sequence of events $(A_n) \in \mathcal{A}^{\mathbb{N}}$ satisfies $\sum_{n \in \mathbb{N}} \mathbb{P}(A_n) \leqslant +\infty$, then almost surely only a finite number of them occur, i.e. $\mathbb{P}(\limsup_{n \to +\infty} A_n) = 0$. Given a finite sequence of $n \in \mathbb{N}$ events $(A_i)_{i=1}^n \in \mathcal{A}^n$, the **union bound** (also known as the Bonferonni inequality) states that $\mathbb{P}(\bigcup_{i=1}^n A_i) \leqslant \sum_{i=1}^n \mathbb{P}(A_i)$. In particular, if all events $A_i$ occur with probability at most $\delta/n$ for $i \in \{1, \dots, n\}$, then the complement events $\bar{A}_i = \Omega \setminus A_i$ satisfy $\mathbb{P}(\bigcap_{i=1}^n \bar{A}_i) \geqslant 1 - \delta$.

We recall that a **random variable** on a measurable space $(\mathcal{X}, \mathcal{B})$ is a measurable function $(\Omega, \mathcal{A}) \to (\mathcal{X}, \mathcal{B})$; when the context is clear, we drop the reference to the $\sigma$-algebra $\mathcal{A}$, $\mathcal{B}$ or both. Unless stated otherwise, we use the notation $X$ for random variables on the Euclidean space $\mathcal{X} = \mathbb{R}^d$ for some $d \in \mathbb{N}$ and $Y$ for random variables on $\mathcal{X} = \mathbb{R}$, equipped with the Euclidean norm $\|\cdot\|$ and the Lebesgue measure. Alternatively, we sometimes consider *discrete* random variables on $\mathbb{N}$ equipped with the counting measure. The pushforward measure $\nu \colon \mathcal{B} \to [0,1]$ defined for all Borel sets $B$ of $\mathcal{X}$ as $\nu(B) = \mathbb{P}\left(\{\omega \in \Omega,\ X(\omega) \in B\}\right)$ is called the **distribution** of $X$. We denote by $\mathcal{M}_1^+(\mathcal{X})$ the set of all such distributions. We recall that $\mathbb{E}_{X \sim \nu}[X] = \int_{\mathcal{X}} x d\nu(x)$ is the expectation of $X$ assuming its distribution is $\nu$, and for $p \in \mathbb{R}_+^\star$, we denote by $\mathbb{L}^p(\mathcal{X}) = \{\nu \in \mathcal{M}_1^+(\mathcal{X}),\ \mathbb{E}_{X \sim \nu}[\|X\|^p] < +\infty\}$ the space of $p$-integrable measures. The **support** of a distribution $\nu$, denoted by $\operatorname{Supp} \nu$, is the smallest closed subset $C \subseteq \mathcal{X}$ such that $\mathbb{P}_{X \sim \nu}(X \in C) = 1$. If $\operatorname{Supp} \nu \subseteq \mathbb{R}_+$, the probability measure is related to its expectation by **Markov's inequality** $\lambda \mathbb{P}_{X \sim \nu}(X \geqslant \lambda) \leqslant \mathbb{E}_{X \sim \nu}[X]$. If $\mathcal{X} \subseteq \mathbb{R}$, the cumulative distribution function of $\nu$, or **c.d.f.**, is the function $x \in \mathbb{R} \mapsto \mathbb{P}_{X \sim \nu}(X \leqslant x)$. The set of discontinuity points of the c.d.f. is called the set of atoms of $\nu$ and correspond to points $x$ such that $\mathbb{P}_{X \sim \nu}(X = x) > 0$. The **total variation distance** between two distributions $\nu, \nu' \in \mathcal{M}_1^+(\mathcal{X})$ is defined as $\delta_{\mathrm{TV}}(\nu, \nu') = \sup_{B \in \mathcal{B}} |\nu(B) - \nu'(B)|$, which also satisfies the equality $\delta_{\mathrm{TV}}(\nu, \nu') = \sup_f |\mathbb{E}_{X \sim \nu}[f(X)] - \mathbb{E}_{X \sim \nu'}[f(X)]|$, where the supremum is taken over all measurable functions $f \colon \mathcal{X} \to [-1, 1]$.

A measure $\nu$ is said to be absolutely continuous with respect to another measure $\nu'$ defined on the same measurable space $(\mathcal{X}, \mathcal{B})$ if for all $B \in \mathcal{B}$, $\nu'(B) = 0$ implies $\nu(B) = 0$. In that case, there exists a measurable function $\frac{\mathrm{d}\nu}{\mathrm{d}\nu'} : \mathcal{X} \to \mathbb{R}_+$ called the Radon-Nikodym derivative, such that $\nu(B) = \int_B \frac{\mathrm{d}\nu}{\mathrm{d}\nu'}(x)d\nu'(x)$ for all $B \in \mathcal{B}$. If $\nu$ is a probability measure, $\frac{\mathrm{d}\nu}{\mathrm{d}\nu'}$ is called the **density** of $\nu$ with respect to $\nu'$. If both $\nu$ and $\nu'$ are probability measures, we define their **Kullback-Leibler divergence** as $\mathrm{KL}(\nu \parallel \nu') = \mathbb{E}_{X\sim\nu}[\log \frac{\mathrm{d}\nu}{\mathrm{d}\nu'}(X)]$. Moreover, the KL divergence and the total variation distance are related by **Pinsker's inequality**: $\delta_{\mathrm{TV}}(\nu, \nu') \leqslant \sqrt{\frac{1}{2}\mathrm{KL}(\nu \parallel \nu')}$.

[1] These assumptions will not be used directly in the rest of this thesis but form a convenient setting to apply standard results of probability and mathematical analysis.

For a set $\mathcal{I}$ and a sequence of probability spaces $(\mathcal{X}_i, \mathcal{B}_i, \nu_i)_{i\in\mathcal{I}}$, we define the product probability space $(\prod_{i\in\mathcal{I}} \mathcal{X}_i, \bigotimes_{i\in\mathcal{I}} \mathcal{B}_i, \bigotimes_{i\in\mathcal{I}} \nu_i)$ where $\prod_{i\in\mathcal{I}} \mathcal{X}_i$ is the Cartesian product, $\bigotimes_{i\in\mathcal{I}} \mathcal{B}_i$ is the $\sigma$-algebra generated by subsets of the form $\prod_{i\in\mathcal{I}} B_i$ where $(B_i)_{i\in\mathcal{I}} \in \prod_{i\in\mathcal{I}} \mathcal{B}_i$, and $\bigotimes_{i\in\mathcal{I}} \nu_i \in \mathcal{M}_1^+(\prod_{i\in\mathcal{I}} \mathcal{X}_i)$ is the **product measure** defined by $\bigotimes_{i\in\mathcal{I}} \nu_i((B_i)_{i\in\mathcal{I}}) = \prod_{i\in\mathcal{I}} \nu_i(B_i)$. For a sequence of subspaces $(\mathcal{H}_i)_{i\in\mathcal{I}} \subseteq \prod_{i\in\mathcal{I}} \mathcal{M}_1^+(\mathcal{X}_i)$, we define the space of product measures $\bigotimes_{i\in\mathcal{I}} \mathcal{H}_i = \{\boldsymbol{\nu} = \bigotimes_{i\in\mathcal{I}} \nu_i,\ (\nu_i)_{i\in\mathcal{I}} \in \prod_{i\in\mathcal{I}} \mathcal{H}_i\}$. A sequence of random variables $\mathbb{X} = (X_i)_{i\in\mathcal{I}}$ is called a **stochastic process**. If the distribution of $\mathbb{X}$ is $\bigotimes_{i\in\mathcal{I}} \nu_i$, we say the random variables are **independent**. If all these random variables are defined on the same measurable space $(\mathcal{X}, \mathcal{B})$ and if all their distributions are equal, we say they are **identically distributed**. In addition, if $\mathcal{I}$ is a totally ordered set, we define the **natural filtration** of the process $\mathbb{X}$ as the sequence of $\sigma$-algebra $(\mathcal{G}_i)_{i\in\mathcal{I}}$ such that for all $i \in \mathcal{I}$, $\mathcal{G}_i = \sigma(\{X_j^{-1}(B),\ j \in \mathcal{I}, j \leqslant i, B \in \mathcal{B}\})$. A stochastic process $(X'_i)_{i\in\mathcal{I}}$ is **adapted** to the filtration $(\mathcal{G}_i)_{i\in\mathcal{I}}$ if $X'_i$ is $(\mathcal{G}_i, \mathcal{B})$-measurable for all $i \in \mathcal{I}$. If $\mathcal{I} = \mathbb{N}$, $(X'_i)_{i\in\mathcal{I}}$ is **predictable** with respect to $(\mathcal{G}_i)_{i\in\mathcal{I}}$ if $(X'_{i+1})_{i\in\mathcal{I}}$ is adapted to $(\mathcal{G}_i)_{i\in\mathcal{I}}$. Finally, a random variable $\tau$ defined on the filtered probability space $(\Omega, \mathcal{A}, (\mathcal{G}_i)_{i\in\mathcal{I}}, \mathbb{P})$ with values in $\mathcal{I}$ is a **stopping time** if $\{\tau \leqslant i\} \in \mathcal{G}_i$ for all $i \in \mathcal{I}$. In practice, we often consider $\mathcal{I} = \mathbb{N}$ or $\mathcal{I} = \mathbb{R}_+^\star$, and interpret it as a set of discrete or continuous times.

☞ *In the rest of this chapter, we present an overview of stochastic bandits, concentration of measure theory and the interplay between them, which constitutes the theoretical foundations of this thesis. We provide proofs of these results for several reasons. First, while most of them are known, they are typically presented here at a higher level of generality than is commonly found in the literature (e.g. the lower bounds of Theorems 1.6 and 1.10, adapted to generic risk measures). Second, the rest of this thesis heavily builds on the techniques presented here (e.g. the method of mixture of time-uniform concentration, extended to various settings in Chapters 3, 4 and 5). Finally, to the best of our knowledge, some of these results are actually new (e.g. the mixture peeling technique of Proposition 1.29) or folk knowledge (e.g. the generic analysis of UCB of Theorem 1.33). We indicate this last category of results by the symbol (💡). For readability, we defer the proofs to Appendix A.*

## 1.2 How to play? An overview of stochastic bandits

The history of stochastic bandits begins with Thompson (1933), which studied statistical properties of two samples likelihood tests, motivated by the design of clinical trials, and introduced what is now known as *multiarmed bandits*. Their theoretical foundations were layed in the 1980s and 1990s (Gittins, 1979; Whittle, 1980; Lai et al., 1985; Agrawal, 1995; Burnetas and Katehakis, 1996), and they have since found industrial applications in recommender systems (Mary et al., 2015; Felício et al., 2017), dynamic pricing (Misra et al., 2019), network routing (Boldrini et al., 2018) and personalised online advertisement (Singh et al., 2022); see Bouneffouf et al. (2020) for a recent survey. Stochastic bandits are also a core component of several areas of theoretical statistics and machine learning, such as sequential testing (Lai, 2001; Bartroff et al., 2012), online optimisation (Shalev-Shwartz et al., 2012; Hazan et al., 2016), active learning and games (Cesa-Bianchi and Lugosi, 2006), and reinforcement learning (Puterman, 2014; Sutton and Barto, 2018; Szepesvári, 2022). In this section, we introduce notations, definitions and basic results that are used throughout this thesis. For a more in-depth introduction to stochastic bandits, we refer the reader to Slivkins (2019); Lattimore and Szepesvári (2020).

### Multiarmed bandits

We start by formalising the concept of multiarmed bandit model introduced in the Foreword. We adopt here the idiom of probability theory, which forms the theoretical foundations on which the analysis of such decision problems rely.

**Definition 1.2** (Bandit model). *Let $K \in \mathbb{N}^\star$. We denote by $[K]$ the* ***set of arms****. A* ***multiarmed stochastic bandit model*** *is a pair $((\boldsymbol{\nu}_t)_{t\in\mathbb{N}}, (\mathcal{F}_t)_{t\in\mathbb{N}})$ such that $(\boldsymbol{\nu}_t)_{t\in\mathbb{N}}$ is a sequence of product measures $(\boldsymbol{\nu}_t)_{t\in\mathbb{N}}$ and $(\mathcal{F}_t)_{t\in\mathbb{N}}$ is a sequence of subspaces of $\mathcal{M}_1^+(\mathbb{R})^K$ such that for all $t \in \mathbb{N}$, $\boldsymbol{\nu}_t = (\nu_t^k)_{k\in[K]} \in \mathcal{F}_t$. It is said to be* ***unstructured*** *if in addition, we have the factorisation*

$$\boldsymbol{\nu}_t = \bigotimes_{k\in[K]} \nu_t^k \in \mathcal{F}_t = \bigotimes_{k\in[K]} \mathcal{F}_t^k \,. \tag{1.1}$$

*Let $\pi = (\pi_t)_{t\in\mathbb{N}}$ be a stochastic process in $[K]$. The* ***reward*** *process associated with $\pi$ is defined as*

$$\mathbb{Y}^\pi = (Y_t^{\pi_t})_{t\in\mathbb{N}} \sim (\nu_t^{\pi_t})_{t\in\mathbb{N}} \,, \tag{1.2}$$

*and the corresponding* ***natural bandit filtration*** *is defined as*

$$(\mathcal{G}_t^\pi)_{t\in\mathbb{N}} = (\sigma(\{(\pi_s, Y_s^{\pi_s}),\ s \leqslant t\}))_{t\in\mathbb{N}} \,. \tag{1.3}$$

*The* ***number of pulls*** *associated with $\pi$ is the sequence*

$$\left(N_t^{k,\pi}\right)_{k\in[K],t\in\mathbb{N}} = \left(\sum_{s=1}^{t} \mathbb{1}_{\pi_s=k}\right)_{k\in[K],t\in\mathbb{N}} \,. \tag{1.4}$$

*Furthermore, $\pi$ is said to be a* ***bandit policy*** *if there exists a filtration $(\mathcal{G}_t^\perp)_{t\in\mathbb{N}}$ independent on $(\mathcal{G}_t^\pi)_{t\in\mathbb{N}}$ such that $\pi$ is predictable with respect to $(\sigma(\mathcal{G}_t^\pi \cup \mathcal{G}_t^\perp))_{t\in\mathbb{N}}$. Finally, a bandit model is said to be* ***stationary*** *if the sequence $((\boldsymbol{\nu}_t)_{t\in\mathbb{N}}, (\mathcal{F}_t)_{t\in\mathbb{N}})$ is constant.*

The natural bandit filtration $(\mathcal{G}_t^\pi)_{t\in\mathbb{N}}$ represents the information available to the agent right after making the decision $\pi_t$ and receiving the reward $Y_t^{\pi_t}$ at time $t \in \mathbb{N}$. Moreover, the definition of a bandit *policy* $\pi$ merely states that (i) $\pi$ depends on past arms and rewards, but not future observations (predictable with respect to the natural bandit filtration), and (ii) $\pi$ may also depend on an external source of randomness, unrelated to all the other random variables (encoded by the independent filtration $(\mathcal{G}_t^\perp)_{t\in\mathbb{N}}$). If $\pi$ is predictable with respect to the natural bandit filtration, it is said to be a *deterministic* policy, otherwise it is said to be a *stochastic* or *randomised* policy. When the policy is clear from the context, we drop the index $\pi$.

**Stationarity.** Unless stated otherwise, we consider stationary bandit in this thesis and therefore drop the dependency on $t$ in the definition of the bandit model and write $\boldsymbol{\nu} = (\nu_k)_{k\in[K]}$ and $(\mathcal{F}_k)_{k\in[K]}$. For background materials on nonstationary bandits, we refer to Garivier and

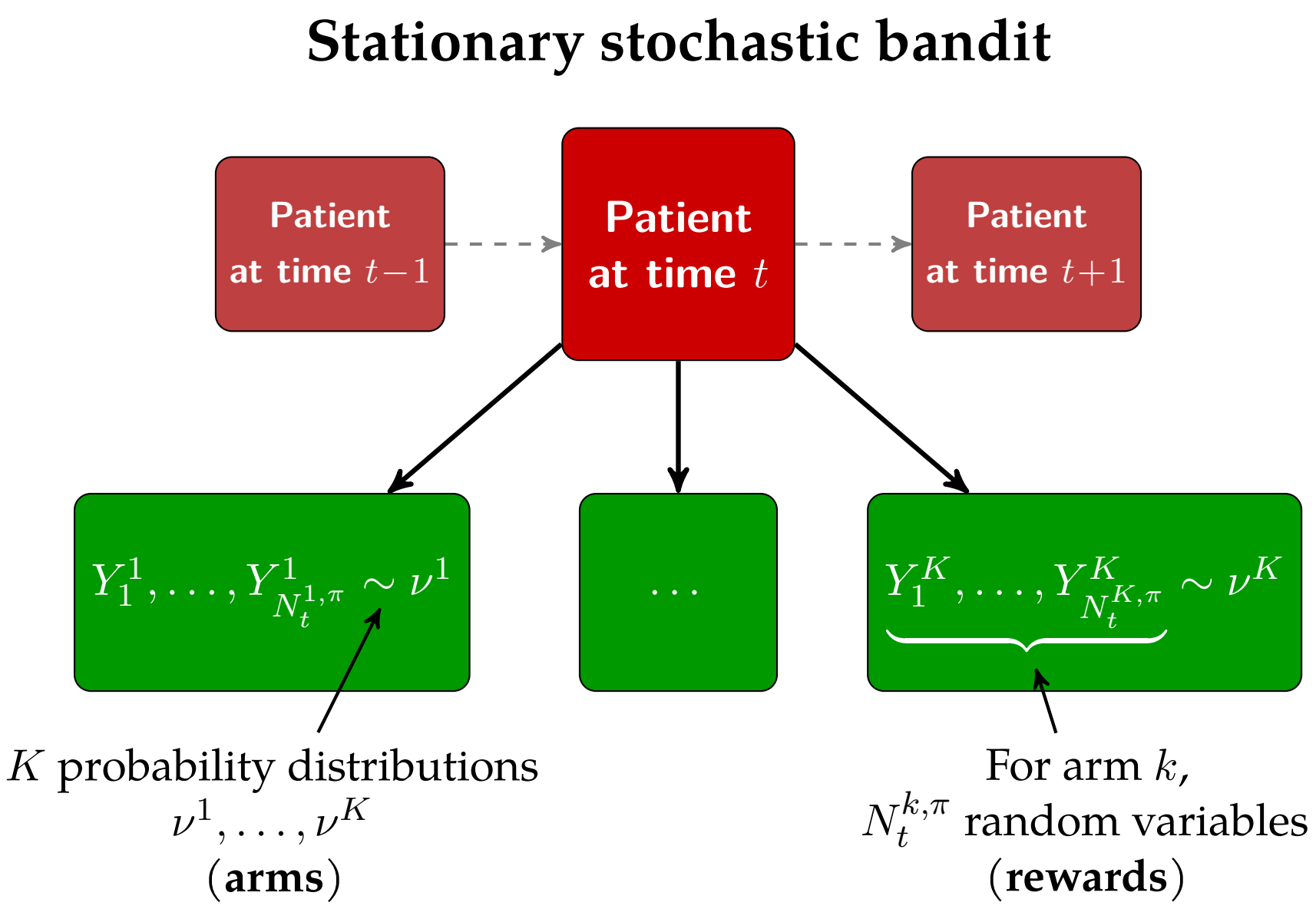

**Figure 1.1** – Graphical description of a stationary stochastic bandit for healthcare decision-making.

Moulines (2011); Allesiardo et al. (2017), as well as to Baudry et al. (2021b); Besson et al. (2022); Bhatt et al. (2023) for recent developments. An abstract graphical representation of a stationary stochastic bandit for healthcare decision-making is presented in Figure 1.1.

☞ *Many algorithms for nonstationary bandits rely on* change point detection *to identify when the underlying bandit distribution* $\boldsymbol{\nu}_t$ *significantly changes with respect* $t \in \mathbb{N}$. *Although we do not specifically consider nonstationary bandits in this thesis, we develop a new change point detection test in Chapter 3, in the context of generic exponential families.*

**Regret.** The learning goal of an agent facing a bandit problem is to find a policy $\pi$ in order to optimise a given criterion. Two main settings have been considered. The first one is called *best arm identification* and aims to identify an arm or a group of arms with certain properties, either by maximising the probability of correct identification given a fixed time horizon $T \in \mathbb{N}$ or by minimising the total number of pulls required to make this probability at least $1 - \delta$ for a fixed $\delta \in (0, 1)$. Notably, the agent does not seek to maximise the rewards accumulated during this learning phase in any way, which is why this problem is sometimes called *pure exploration*. We do not discuss this setting further and refer to Bubeck et al. (2009); Garivier and Kaufmann (2016); Jourdan et al. (2022, 2023) for recent advances on the topic. The second setting is precisely that of reward maximisation, or equivalently of *regret* minimisation.

**Definition 1.4** (Pseudo regret). *Let $((\boldsymbol{\nu}_t)_{t\in\mathbb{N}}, (\mathcal{F}_t)_{t\in\mathbb{N}})$ be a bandit model, $\pi$ a bandit policy and $\rho = (\rho_t)_{t\in\mathbb{N}}$ a sequence of mappings $\mathcal{M}_1^+(\mathbb{R}) \to \mathbb{R} \cup \{\pm\infty\}$ called the* ***risk measures****. For $t \in \mathbb{N}$, we define $\mathbb{L}^{\rho_t}(\mathbb{R}) = |\rho_t|^{-1}(\mathbb{R}) \subset \mathcal{M}_1^+(\mathbb{R})$ the set of probability measures that are $\rho_t$-integrable and assume that $\nu_t^k \in \mathbb{L}^{\rho_t}(\mathbb{R})$. The* ***cumulative pseudo regret*** *of policy $\pi$ with respect to the risk measures $(\rho_t)_{t\in\mathbb{N}}$ and bandit model $(\boldsymbol{\nu}_t)_{t\in\mathbb{N}}$ is the stochastic process*

$$(\mathcal{R}_T^{\pi,\rho,\boldsymbol{\nu}})_{T\in\mathbb{N}} = \left(\sum_{t=1}^{T} \rho_t^\star(\boldsymbol{\nu}_t) - \rho_t(\nu_t^{\pi_t})\right)_{T\in\mathbb{N}}, \tag{1.5}$$

*where $\rho_t^\star(\boldsymbol{\nu}_t) = \max_{k\in[K]} \rho_t(\nu_t^k)$. When the bandit model is stationary and $\rho$ is the expectation mapping, we use the notations $\mu_k = \rho(\nu_k) = \mathbb{E}_{Y\sim\nu_k}[Y]$, $\mu^\star = \max_{k\in[K]} \mu_k$ and $k^\star \in \operatorname{argmax}_{k\in[K]} \mu_k$.*

The $\rho$-integrability condition is implicitly assumed to hold for every bandit model we consider in this thesis; under the expectation risk measure, this simply means that the bandit measure for each arm $k \in [K]$ is integrable, i.e. $\nu_k \in \mathbb{L}^1(\mathbb{R})$. A policy $\pi$ that minimises the pseudo regret $\mathcal{R}_T^{\pi,\rho,\boldsymbol{\nu}}$ also maximises the sum of rewards, measured by the risk measures $(\rho_t)_{t=1}^T$, i.e. $\sum_{t=1}^T \rho_t(\nu_t^{\pi_t})$ up to horizon $T \in \mathbb{N}$. The pseudo regret is convenient as it normalises this sum of rewards against an oracle benchmark policy that always selects $\rho_t^\star(\boldsymbol{\nu}_t)$. A typical example of risk measure is the expectation; we discuss other risk measures extensively in Chapter 5. In the stationary case, a policy that misses the optimal arm a nonvanishing fraction of the time suffers *linear* pseudo regret $\mathcal{R}_T^{\pi,\rho,\boldsymbol{\nu}} = \Theta(T)$, in contrast with a policy that learns to pull the optimal arm increasingly often, resulting in *sublinear* pseudo regret $\mathcal{R}_T^{\pi,\rho,\boldsymbol{\nu}} = o(T)$. Note that even though $\rho$ is deterministic, $\mathcal{R}_T^{\pi,\rho,\boldsymbol{\nu}}$ is a random variable because the policy $\pi$ itself is stochastic, and it is nonnegative thanks to the definition of $\rho_t^\star(\boldsymbol{\nu}_t)$.

To analyse the pseudo regret, both in theory and in practice, we either (i) bound its expectation $\mathbb{E}_{\pi,\boldsymbol{\nu}_\pi^{\otimes T}}[\mathcal{R}_T^{\pi,\rho,\boldsymbol{\nu}}] \leqslant f(T)$, where $f\colon \mathbb{R}_+^\star \to \mathbb{R}_+^\star$, and report the mean regret across independent replicates during numerical experiments; or (ii) bound the regret with high probability, i.e. $\mathbb{P}_{\pi,\boldsymbol{\nu}_\pi^{\otimes T}}(\mathcal{R}_T^{\pi,\rho,\boldsymbol{\nu}} \leqslant g(\delta, T)) \geqslant 1 - \delta$, where $g\colon (0,1) \times \mathbb{R}_+^\star \to \mathbb{R}_+^\star$ and $\delta \in (0,1)$, and report empirical quantiles of the regret across independent replicates. Lemma A.1 in Appendix A.1 shows that both bounds are essentially equivalent. In this thesis, we focus on expected pseudo regret when dealing with multiarmed bandits. The main reason is that controlling this quantity amounts to bounding each expected number of pulls independently, as shown in the next proposition.

**Proposition 1.5** (Pseudo regret decomposition). *Let $(\boldsymbol{\nu}, \mathcal{F})$ be a stationary, unstructured bandit. For any risk measure $\rho$ and policy $\pi$, we have*

$$\mathcal{R}_T^{\pi,\rho,\boldsymbol{\nu}} = \sum_{k\in[K]\setminus\{k^\star\}} \Delta_k^{\rho,\boldsymbol{\nu}} N_T^{k,\pi} \,, \tag{1.6}$$

*where $\Delta_k^{\rho,\boldsymbol{\nu}} = \rho^\star(\boldsymbol{\nu}) - \rho(\nu_k)$ is the* ***suboptimality gap*** *of arm $k \in [K]$. In particular, we have*

$$\mathbb{E}_{\pi,\boldsymbol{\nu}_\pi^{\otimes T}}\left[\mathcal{R}_T^{\pi,\rho,\boldsymbol{\nu}}\right] = \sum_{k\in[K]\setminus\{k^\star\}} \Delta_k^{\rho,\boldsymbol{\nu}} \mathbb{E}_{\pi,\boldsymbol{\nu}_\pi^{\otimes T}}\left[N_T^{k,\pi}\right] . \tag{1.7}$$

**Instance-dependent lower bound for multiarmed bandits (💡).** The regret decomposition above allows to focus on each suboptimal arm separately. The following result establishes a *model-dependent* lower bound on the expected number of pulls to such arms, and thus on the expected pseudo regret. This lower bound represents the best possible regret achievable by a policy in terms of quantities that depend on the bandit model $\boldsymbol{\nu}, \mathcal{F})$, such as the optimal risk measure $\rho^\star(\boldsymbol{\nu})$. First introduced by Lai et al. (1985) for parametric families of distributions and then extended by Burnetas and Katehakis (1996) for more general families under the expectation risk measure, it was recently adapted for the conditional value at risk measure (Baudry et al., 2021a). Using similar generic change of measure arguments, we show here that it actually holds for virtually *any* reasonable risk measures (see Appendix A.1 for the proof).

**Theorem 1.6** (Instance-dependent regret lower bound). *Let* $(\boldsymbol{\nu}, \mathcal{F})$ *be a stationary, unstructured bandit,* $\rho$ *a risk measure and* $\pi$ *a policy. Assume that:*

(*i*) $\rho$ *is* confusable *over* $\mathcal{F}$*, i.e. for any measure* $\boldsymbol{\nu}' = \bigotimes_{k\in[K]} \nu'_k \in \mathcal{F}$*, there exists a measure* $\widetilde{\boldsymbol{\nu}}' \in \mathcal{F}$ *such that for any* $k, j \in [K]$,

$$\begin{cases} \widetilde{\nu}'_j = \nu'_k & \text{if } j \neq k, \\ \rho(\widetilde{\nu}'_k) > \rho^\star(\boldsymbol{\nu}') & \text{otherwise;} \end{cases} \tag{1.8}$$

(*ii*) $\pi$ *is* strongly consistent *over* $\mathcal{F}$*, i.e. for any measure* $\boldsymbol{\nu}' \in \mathcal{F}$ *and suboptimal arm* $k \in [K]$ *such that* $\rho(\nu'_k) < \rho^\star(\boldsymbol{\nu}')$*, we have, for any* $\varepsilon \in [0, 1)$*, when* $T \to +\infty$,

$$\mathbb{E}_{\pi,\boldsymbol{\nu}_\pi^{\otimes T}}\left[N_T^{k,\pi}\right] = o(T^{1-\varepsilon})\,. \tag{1.9}$$

*Then for any suboptimal arm* $k \in [K] \setminus \{k^\star\}$*, we have the asymptotic* ***lower bound***

$$\liminf_{T\to+\infty} \frac{\mathbb{E}_{\pi,\boldsymbol{\nu}_\pi^{\otimes T}}\left[N_T^{k,\pi}\right]}{\log T} \geqslant \frac{1}{\mathcal{K}_{\inf}^{\mathcal{F}_k,\rho}\left(\nu_k; \rho^\star(\boldsymbol{\nu})\right)}\,, \tag{1.10}$$

*where we define the* extremal Kullback-Leibler *operator over* $\mathcal{F}_k$ *as:*

$$\begin{aligned} \mathcal{K}_{\inf}^{\mathcal{F}_k,\rho} : \mathcal{F}_k \times \mathbb{R} &\longrightarrow \mathbb{R} \\ (\nu; \rho^\star) &\longmapsto \inf_{\nu'\in\mathcal{F}_k} \left\{\mathrm{KL}(\nu \,\|\, \nu'),\ \rho(\nu') > \rho^\star\right\}\,. \end{aligned} \tag{1.11}$$

*In particular, the expected pseudo regret satisfies the following lower bounded:*

$$\liminf_{T\to+\infty} \frac{\mathbb{E}_{\pi,\boldsymbol{\nu}_\pi^{\otimes T}}\left[\mathcal{R}_T^{\pi,\rho,\boldsymbol{\nu}}\right]}{\log T} \geqslant \sum_{k\in[K]\setminus\{k^\star\}} \frac{\Delta_k^{\rho,\boldsymbol{\nu}}}{\mathcal{K}_{\inf}^{\mathcal{F}_k,\rho}\left(\nu_k; \rho^\star(\boldsymbol{\nu})\right)}\,. \tag{1.12}$$

When the risk measure is the expectation, we simply write $\mathcal{K}_{\inf}^{\mathcal{F}}$ instead.

**Minimax lower bound.** As noted aboved, the regret lower bound of Theorem 1.6 is model-dependent, in the sense that it is specific to a given bandit model $(\boldsymbol{\nu}, \mathcal{F})$. Note that the right-hand side is independent of the policy $\pi$, so this lower bound actually holds for $\inf_\pi \mathbb{E}_{\pi,\boldsymbol{\nu}_\pi^{\otimes T}}[\mathcal{R}_T^{\pi,\rho,\boldsymbol{\nu}}]$. A complementary, model-agnostic lower bound can be found by studying the robustness of the policy with respect to the worst possible bandit model, i.e. $\inf_\pi \sup_{\boldsymbol{\nu}} \mathbb{E}_{\pi,\boldsymbol{\nu}_\pi^{\otimes T}}[\mathcal{R}_T^{\pi,\rho,\boldsymbol{\nu}}]$, which is also known as the *minimax* expected pseudo regret. Such worst case lower bounds typically scale as $\Omega(\sqrt{KT})$ (Audibert and Bubeck, 2009). Note that policies can be optimal with respect to the model-dependent regret but perform poorly in terms of minimax expected pseudo regret, and vice versa. Recently, *best of both worlds* policies (Zimmert and Seldin, 2021) have been designed to simultaneously match both types of lower bounds.

**Alternative definitions of regret.** The regret of Definition 1.4 is called the *dynamic* pseudo regret for nonstationary bandit models. Even in the stationary case, there exists several alternative notions of regret. In general, the regret compares the quality of a policy $\pi$ against an oracle benchmark policy $\pi^\star$. In our definition, quality is measured by $\rho$ and $\pi^\star$ selects at each time the arm $k$ with the highest risk measure. One may instead compare the rewards directly and let $\pi^\star$ selects the arm with the best total reward in hindsight at horizon $T$, i.e. $R_T^\pi = \max_{k\in[K]} \sum_{t=1}^T Y_t^k - Y_t^{\pi_t}$. However, this regret is less convenient to analyse and interpret due to the additional randomness of the rewards themselves. For instance, using the expectation criterion, if $\boldsymbol{\nu}$ is a product of $R$-sub-Gaussian bandit measures (see Section 1.3) such that all arms are centred ($\mu_k = 0$ for all $k \in [K]$), then $\mathbb{E}[\max_{k\in[K]} \sum_{t=1}^T Y_t^k] \leqslant R\sqrt{2T\log K}$ (Boucheron et al., 2013, Theorem 2.5), which is asymptotically sharp (up to a constant multiplicative factor) for many distributions (Gaussian, Bernoulli, etc.). Consequently, for these distributions, the regret scales as $\mathbb{E}[R_T^\pi] = \Theta\left(\sqrt{T\log K}\right)$ regardless of the policy, while the pseudo regret $\mathcal{R}_T$ is uniformly zero, which corresponds with the intuition that no option is better than another. An even harder benchmark would be to consider the policy $\pi^\star$ that selects the arm with the best reward at each time, i.e. $\sum_{t=1}^T \max_{k\in[K]} Y_t^k - Y_t^{\pi_t}$, which can scale linearly with $T$ on some instances, irrespective of the policy $\pi^\star$. Finally, an altogether different approach, pioneered in Cassel et al. (2018), is to consider *trajectories* of rewards, i.e. $\mathbb{Y}_T^\pi = (Y_t^{\pi_t})_{t=1}^T$, and compare policies using a risk measure $U\colon \mathbb{R}^{(\mathbb{N})} \to \mathbb{R}$, i.e. $\mathcal{R}_T^U = \mathbb{E}[U(\mathbb{Y}_T^{\pi^\star})] - \mathbb{E}[U(\mathbb{Y}_T^\pi)]$, where $\pi^\star = \operatorname{argmax}_\pi \mathbb{E}[U(\mathbb{Y}_T^\pi)]$.

**Structured bandits.** An unstructured bandit model simply means that the rewards generated by each arm are mutually independent. In particular, pulling a given arm does not give any information on the marginal distributions of the other arms. By contrast, *structured* bandits assume some dependency, potentially allowing for more efficient learning strategies by leveraging shared information between arms. Generic structures have been studied in the literature (Combes et al., 2017), as well as specific instance such as unimodal bandits ($k \in [K] \mapsto \mu_k$ is a unimodal function, see e.g. Combes and Proutiere (2014); Saber et al. (2021)), Lipschitz bandits ($k \in [K] \mapsto \mu_k$ is a Lipschitz function, see e.g. Bubeck et al. (2011); Magureanu et al. (2014)) and bandits with groups of similar arms (Pesquerel et al., 2021), among others. An important class of structured bandits is that of *contextual* bandits, where $k \in [K] \mapsto \mu_k(X_t)$ is allowed to depend on a vector of side information $X_t \in \mathbb{R}^d$ at time $t \in \mathbb{N}$. In the rest of this thesis, we focus on unstructured bandits and contextual bandits only.

**Toy example.** We now formalise the multiarmed bandit introduced in the Foreword (Figure 1). In this example, a surgeon (the agent) is faced with $K = 3$ follow-up options: at the surgery department ($k = 1$), at the general practitioner ($k = 2$) or remotely ($k = 3$). Say they adopted a policy $\pi$, so that the $t$-th patient is addressed to the postoperative pathway $\pi_t \in \{1, 2, 3\}$. Upon completion of the follow-up visit, the surgeon observes a score (reward) generated by an unknown distribution $\nu_{\pi_t}$. For simplicity, we assume here that the problem is stationary, i.e. only the policy may evolve with time, not the reward distributions. The scores are normalised between $0$ and $100$, so the bandit model is $((\nu_k)_{k\in[K]}, (\mathcal{F}_k)_{k\in[K]})$ and $\mathcal{F}_k$ is the family of distributions supported in the interval $[0, 100]$ for each $k$. We also assume for now that it is unstructured ($(\nu_k)_{k\in[K]}$ forms a product measure); for instance the scores generated by visiting the surgery department do not provide information on the scores of the other two alternatives. The goal of the surgeon is to adapt their policy $\pi$ in order to maximise the cumulative rewards $\sum_{t=1}^{T} \rho(\nu_{\pi_t})$ (i.e. minimise the pseudo regret), where $\rho$ is mapping from reward distributions to scalar values that encodes the risk preference of the surgeon.

## Contextual bandits

Building on the foundations laid by Definition 1.2, we extend the notion of multiarmed bandits to account for contextual information that influences the distribution of rewards.

**Definition 1.7** (Contextual bandit model). *Fix $d \in \mathbb{N}$. Let $(\mathcal{X}_t)_{t\in\mathbb{N}} \in \mathfrak{P}(\mathbb{R}^d)^{\mathbb{N}}$ be a sequence of subsets of $\mathbb{R}^d$ called the* ***action sets****. A* ***contextual bandit model*** *is a sequence of mappings $(\nu_t)_{t\in\mathbb{N}}$ such that $\nu_t \colon \mathcal{X}_t \to \mathcal{M}_1^+(\mathbb{R})$ for all $t \in \mathbb{N}$. Let $\mathbb{X} = (X_t)_{t\in\mathbb{N}}$ be a stochastic process such that $X_t$ is a $\mathcal{X}_t$-valued random variable for all $t \in \mathbb{N}$. The* ***reward*** *process associated with $\mathbb{X}$ is defined as*

$$\mathbb{Y}^{\mathbb{X}} = \left(Y_t^{X_t}\right)_{t\in\mathbb{N}} \sim \left(\nu_t\left(X_t\right)\right)_{t\in\mathbb{N}} , \tag{1.13}$$

*and the corresponding* ***natural contextual bandit filtration*** *is defined as*

$$\left(\mathcal{G}_t^{\mathbb{X}}\right)_{t\in\mathbb{N}} = \left(\sigma\left(\left\{\left(\mathcal{X}_u, X_s, Y_s^{X_s}\right),\ s \leqslant t, u \leqslant t+1\right\}\right)\right)_{t\in\mathbb{N}} . \tag{1.14}$$

*Furthermore, $\mathbb{X}$ is said to be a* ***contextual bandit policy*** *if there exists a filtration $(\mathcal{G}_t^{\perp})_{t\in\mathbb{N}}$ independent on $(\mathcal{G}_t^{\mathbb{X}})_{t\in\mathbb{N}}$ such that $\mathbb{X}$ is predictable with respect to $(\sigma(\mathcal{G}_t^{\mathbb{X}} \cup \mathcal{G}_t^{\perp}))_{t\in\mathbb{N}}$. A bandit model is said to be* ***linear*** *if there exists $p \in \mathbb{N}^\star$, a sequence $(\ell_t^\star)_{t\in\mathbb{N}}$ of linear mappings and a mapping $\varphi$ such that the following factorisation holds for all $t \in \mathbb{N}$:*

$$\mathcal{X}_t \xrightarrow{\ \ell_t^\star\ } \mathbb{R}^p \xrightarrow{\ \varphi_t\ } \mathcal{M}_1^+(\mathbb{R}) , \qquad \nu_t = \varphi \circ \ell_t^\star$$

*Finally, it is said to be* ***stationary*** *if the sequence $(\nu_t)_{t\in\mathbb{N}}$ (or $(\ell_t^\star, \varphi_t)_{t\in\mathbb{N}}$) is constant and equal to the mapping $\nu \colon \mathcal{X} \to \mathcal{M}_1^+(\mathbb{R})$, where $\mathcal{X} = \bigcup_{t\in\mathbb{N}} \mathcal{X}_t$.*

Again, the definition of a contextual bandit policy $\mathbb{X} = (X_t)_{t\in\mathbb{N}}$ allows $X_{t+1}$ to depend on (i) the set of available actions $\mathcal{X}_{t+1}$, (ii) the history of past decisions and rewards (measurable with respect to $\mathcal{G}_t^{\mathbb{X}}$) and (iii) an independent source of randomness. In other words, the natural contextual bandit filtration represents the information available to the agent after making the decision $X_t$, receiving the reward $Y_t^{X_t}$ and observing the next action set $\mathcal{X}_{t+1}$, but right before making the decision $X_{t+1}$. It is often convenient to also consider instead the *adapted* filtration

$$(\bar{\mathcal{G}}_t^{\mathbb{X}})_{t\in\mathbb{N}} = \left(\sigma\left(\mathcal{X}_u, X_u, Y_s^{X_s}\right),\ s \leqslant t-1, u \leqslant t\right)_{t\in\mathbb{N}} , \tag{1.15}$$

which represents the information available after making the decision at time $t$ but *before* receiving the corresponding reward (in particular, the policy $\mathbb{X}$ is adapted to this filtration).

A simple example of linear bandit is to assume a linear statistical model between rewards and actions, e.g. $Y_t = \langle \theta^\star, X_t \rangle + \eta_t$, where $\theta^\star \in \mathbb{R}^d$ and $(\eta_t)_{t\in\mathbb{N}}$ is an i.i.d. Gaussian process with mean zero and variance $\sigma^2 \in \mathbb{R}_+^\star$ (i.e. $p = 1$, $\ell^\star \colon x \in \mathbb{R}^d \mapsto \langle \theta^\star, x \rangle$ and $\varphi \colon y \in \mathbb{R} \mapsto \mathcal{N}(y, \sigma^2)$).

The formulation above allows for a slightly more general notion of linearity; for instance, we could also parametrise the variance $\sigma^2$ as a function of the action $X_t$.

*In Chapter 5, we explore settings where other characteristics of the reward distribution than the expectation are linearly parametrised.*

Similarly to multiarmed bandits, we do not focus on nonstationary contextual bandits in this thesis, and refer instead to Cheung et al. (2019); Russac et al. (2019). However, we point out that our definition of stationarity allows for time-varying action sets $(\mathcal{X}_t)_{t\in\mathbb{N}}$, i.e. only the mapping from actions to rewards is constant with time, *not* the actions available to the agent.

This generic definition of contextual bandits encompasses many well-studied settings.

- **Unstructured $K$-armed bandit models.** $(\bigotimes_{k\in[K]} \nu_t^k)_{t\in\mathbb{N}}$ is a special instance of linear bandit in dimension $d = K$ where, if we denote by $\mathcal{X} = (e_k)_{k\in[K]}$ the canonical basis of $\mathbb{R}^K$, we have $\theta^\star = (k)_{k\in[K]}$, $\ell^\star : x \in \mathbb{R}^K \mapsto \langle \theta^\star, x\rangle$ and $\varphi_t : k \in [K] \subset \mathbb{R} \mapsto \nu_t^k$ for $t \in \mathbb{N}$.

- **Unstructured $K$-armed bandit location models.** If the $K$-armed bandit model can further be written as $Y_t^k = \mu_t^k + \eta_t$ with random variables $Y_t^k \sim \nu_t^k$ and $\mu_t^k \in \mathbb{R}$ for all $k \in [K]$ and $t \in \mathbb{N}$ (e.g. $\nu_t^k = \mathcal{N}(\mu_t^k, \sigma^2)$), we also have the representation with $d = K$, $\theta_t^\star = (\mu_t^k)_{k\in[K]}$ and $\varphi_t : y \in \mathbb{R} \mapsto y + \eta_t$ (where we identify $\eta_t$ with its distribution).

- **Linear $K$-armed bandit models**. Assume further that the action sets at all time $t \in \mathbb{N}$ are of size $K$, i.e. $\mathcal{X}_t = (X_t^k)_{k\in[K]}$, and that the reward associated with arm $k \in [K]$ is $Y_t^k = \langle \theta_t^\star, X_t^k\rangle + \eta_t$. This is a linear bandit in dimension $d = K$.

- **Linear $K$-armed bandit models with shared contexts**. Alternatively, assume that the agent observes a single vector $x_t \in \mathbb{R}^m$ at time $t \in \mathbb{N}$ and that each arm $k \in [K]$ generates a different reward distrib¸ution with respect to $x_t$, e.g. $Y_t^k = \langle \theta_t^k, x_t\rangle + \eta_t$. This is a linear bandit in dimension $d = Km$ with $\theta_t^\star$ is the concatenation of $K$ vectors $\theta_t^1, \dots, \theta_t^K$ and $\mathcal{X}_t = (\mathrm{e}_k \otimes x_t)_{k\in[K]}$, where $\otimes$ is the Kronecker product on $\mathbb{R}^K \times \mathbb{R}^m$.

**Definition 1.9** (Pseudo regret). *Let $(\nu_t)_{t\in\mathbb{N}}$ a contextual bandit model and $\rho = (\rho_t)_{t\in\mathbb{N}}$ a sequence of mappings $\mathcal{M}_1^+(\mathbb{R}) \to \mathbb{R} \cup \{\pm\infty\}$ called the **risk measures**. We assume that $\nu_t(\mathcal{X}_t) \subset \mathbb{L}^{\rho_t}(\mathbb{R})$ for all $t \in \mathbb{R}$. The **cumulative pseudo regret** of policy $\pi$ with respect to the risk measures $(\rho_t)_{t\in\mathbb{N}}$ and contextual bandit model $(\nu_t)_{t\in\mathbb{N}}$ is the stochastic process*

$$\left(\mathcal{R}_T^{\mathbb{X},\rho,\nu}\right)_{T\in\mathbb{N}} = \left(\sum_{t=1}^{T} \rho_t\left(\nu_t(X_t^\star)\right) - \rho_t\left(\nu_t(X_t)\right)\right)_{T\in\mathbb{N}}, \tag{1.16}$$

*where $X_t^\star = \operatorname{argmax}_{x\in\mathcal{X}_t} \rho_t(\nu_t(x))$.*

**Minimax lower bound for contextual bandits. (💡).** Lower bounds have also been derived for linear bandits under the expectation risk measure and various assumptions on the geometry of the action sets $(\mathcal{X}_t)_{t\in\mathbb{N}}$. In the model-dependent setting, for finite action sets, $\Omega(\log T)$ bounds have been proven, see e.g. Graves and Lai (1997); Combes et al. (2017). In this thesis however, we will focus on minimax regret bounds for contextual bandits. In that setting, Chu et al. (2011) introduced a $\Omega(\sqrt{dT})$ lower bound assuming a finite number $K$ of actions at each round (as in the bandit model with shared contexts presented above). This result was recently strengthened to $\Omega(\sqrt{dT\log(K)\log(T/d)})$ (Li et al., 2019). For infinite action sets, Rusmevichientong and Tsitsiklis (2010) proved a $\Omega(d\sqrt{T})$ lower bound for actions in the Euclidean unit ball $\mathbb{B}^d_{\|\cdot\|_2}(0,1)$ and Gaussian rewards; their proofs was later simplified and extended to the hypercube $\mathbb{B}^d_{\|\cdot\|_\infty}(0,1)$ in Lattimore and Szepesvári (2020). Drawing inspiration from the latter, we slightly extend their results to handle (i) generic risk measures that are linearly parametrised by the actions, and (ii) non-Gaussian models, assuming instead a milder condition on the bandit log likelihood; we refer to Appendix A.1 for the proof.

**Theorem 1.10** (Minimax regret lower bound for contextual bandits). *Let $\nu_\theta$ be a stationary contextual linear bandit with constant action set $\mathcal{X}_t = \mathcal{X}$ for all $t \in \mathbb{N}$, and $\rho$ a risk measure such that $\rho(\nu_\theta(x)) = \langle \theta, x\rangle$ for all $x \in \mathcal{X}$. Assume that there exists a mapping $g\colon \mathbb{R} \to \mathbb{R}_+$ such that $\mathbb{E}_{\mathbb{X},\theta}[\log \frac{\mathrm{d}\nu_\theta(X_t)}{\mathrm{d}\nu_{\theta'}(X_t)}(Y_t) \mid X_t] \leqslant g(\langle \theta - \theta', X_t\rangle/\sigma)$ for all $t \in \mathbb{N}$ and some fixed $\sigma \in \mathbb{R}_+^\star$ such that $g(y) = \mathcal{O}(y^2)$ when $y \to 0$. If either $(i)$ $\mathcal{X} = \mathbb{B}^d_{\|\cdot\|_\infty}(0,1)$ or $(ii)$ $\mathcal{X} = \mathbb{B}^d_{\|\cdot\|_2}(0,1)$, we have, for $T$ large enough,*

$$\inf_{\mathbb{X}} \sup_{\theta\in\mathbb{R}^d} \mathbb{E}_{\mathbb{X},\theta}\left[\mathcal{R}_T^{\mathbb{X},\rho,\nu_\theta}\right] = \Omega(\sigma d\sqrt{T}). \tag{1.17}$$

*The likelihood condition of Theorem 1.10 is satisfied in particular in the linear Gaussian bandit models $\nu_\theta\colon x \in \mathbb{R}^d \mapsto \mathcal{N}(\langle\theta, x\rangle, \sigma^2)$ with $g\colon y \in \mathbb{R} \mapsto y^2/2$. In Chapters 5 and 7, we apply this result to non-Gaussian models (with piecewise quadratic log likelihood) and other risk measures than the expectation (namely the expectiles).*

Although this lower bound is stated on the expected pseudo regret, we favour high probability bounds when dealing with contextual bandits in this thesis, as we analyse algorithms based on concentration results that naturally yield such bounds. Of note, Lemma A.1 extends readily to contextual bandit regret, so that both bounds are essentially equivalent.

**Toy example.** Contextual bandits offer a richer class of models to analyse the example of postoperative recommendations recalled above. When addressing the $t$-th patient to one of the follow-up pathways, the surgeon has access to clinical informations, represented by a vector $x_t$; for instance, $x_t$ may contain $4$ variables: sex, weight, height and age. In terms of postoperative score, each of the $K = 3$ options interacts differently with these additional variables, which we model as a linear $K$-armed bandit with shared contexts, i.e. the random reward for choosing option $k$ at time $t$ is $Y_t^k = \langle \theta_k, x_t\rangle + \eta_t$, where $\eta_t$ represents noise and $(\theta_{k,j})_{j\in\{1,\dots,4\}}$ is a vector parameter for option $k$. As an example, $\theta_{1,4} > 0$ and $\theta_{2,4} < 0$, $\theta_{3,4} < 0$ encodes a preference towards addressing older patients (age is the fourth component of $x_t$) to the surgery department ($k = 1$), as it increases the associated mean postoperative score.

## 1.3 What is the reward? An overview of statistical models

In this section, we introduce standard statistical models for stochastic bandits, i.e. families of measures $\mathcal{F} \subset \mathcal{M}_1^+(\mathbb{R})$ to represent reward distributions. Following standard conventions in statistics, we distinguish two broad categories of statistical models. If $\mathcal{F}$ can be embedded in a finite dimensional space (i.e. distributions in $\mathcal{F}$ can be described with a vector parameter), it is called a *parametric* model, otherwise it is called a *nonparametric* model. The models we consider in this thesis are summarised in Table 1.1.

**Parametric models and exponential families.** A typical example of parametric model is the family of Gaussian distributions, parametrised by a vector $(\mu, \sigma) \in \mathbb{R} \times \mathbb{R}_+^\star$ corresponding to their means and standard deviations.

More generally, Gaussian distributions form an instance of *exponential family*. Given a set $\mathcal{Y}$, an exponential family is a set of measures $\mathcal{F}_{h,F,\mathcal{L}}^{\mathrm{EF}} = \{\nu_\theta\}_{\theta\in\Theta} \subset \mathcal{M}_1^+(\mathcal{Y})$, parametrised in some open set $\Theta \subseteq \mathbb{R}^d$, such that given a common base measure $\bar{\nu}$, for all $\theta \in \Theta$, the measure $\nu_\theta$ has a density $p_\theta = \frac{\mathrm{d}\nu_\theta}{\mathrm{d}\bar{\nu}}$ given by

$$\begin{aligned} p_\theta \colon \mathcal{Y} &\longrightarrow \mathbb{R}_+ \\ y &\longmapsto h(y)\exp(\langle \theta, F(y)\rangle - \mathcal{L}(\theta)) \,, \end{aligned} \tag{1.18}$$

where $F \colon \mathcal{Y} \to \mathbb{R}^d$ is the *feature function*, $h : \mathcal{Y} \to \mathbb{R}_+$ is the *base function*, and $\mathcal{L}$ represents the normalisation term (also known as the *log-partition function*) given by the convex mapping

$$\begin{aligned} \mathcal{L} \colon \Theta &\longrightarrow \mathbb{R} \\ \theta &\longmapsto \log \int_{\mathcal{Y}} h(y)\exp(\langle \theta, F(y)\rangle) dy \,. \end{aligned} \tag{1.19}$$

Note that a change of base measure $\bar{\nu}$ only affects the base function $h$, so we often omit it, and we resort to a slight abuse of notation and refer to the distribution $\nu_\theta$ by its density $p_\theta$. We denote by $\Theta_{\mathcal{D}} = \{\theta \in \mathbb{R}^d : \mathcal{L}(\theta) < \infty\}$ the domain of $\mathcal{L}$ and by $\Theta_I = \{\theta \in \Theta_{\mathcal{D}} : \det \nabla^2\mathcal{L}(\theta) > 0\}$ the set on which its Hessian is invertible. Throughout this thesis, we assume that $\Theta \subset \Theta_{\mathcal{D}} \cap \Theta_I$, which is tantamount to assuming that the family is minimal, and ensures we only consider nondegenerate distributions ($\nabla\mathcal{L}$ is one-to-one on its domain). We use the notations $\mathbb{P}_\theta$ and $\mathbb{E}_\theta$ to explicitly refer to the probability and expectation under the distribution $p_\theta \in \mathcal{F}_{h,F,\mathcal{L}}^{\mathrm{EF}}$. We also assume that $p_\theta \in \mathbb{L}^F(\mathcal{Y})$ for all $\theta \in \Theta$, i.e. $\mathbb{E}_\theta[F(Y)] < +\infty$.

A fundamental property of exponential families is the following form for the KL divergence between two distributions with parameters $\theta, \theta' \in \Theta_{\mathcal{D}}$:

$$\mathrm{KL}(p_\theta \,\|\, p_{\theta'}) = \langle \theta - \theta', \mathbb{E}_\theta[F(Y)]\rangle - \mathcal{L}(\theta) + \mathcal{L}(\theta') \,. \tag{1.20}$$

Here, $\mathbb{E}_\theta\left[F(Y)\right]$ is called the vector of *expectation parameters* (also known as *dual parameters*), and is equal to $\nabla\mathcal{L}(\theta)$. Hence, it holds that $\mathrm{KL}(p_\theta \parallel p_{\theta'}) = \mathcal{B}_\mathcal{L}(\theta',\theta)$, where $\mathcal{B}_\mathcal{L}$ is known as the Bregman divergence (Bregman, 1967) with potential function $\mathcal{L}$, defined as

$$\begin{aligned}\mathcal{B}_\mathcal{L} \colon \Theta_\mathcal{D} \times \Theta_\mathcal{D} &\longrightarrow \mathbb{R}_+ \\ (\theta',\theta) &\longmapsto \mathcal{L}(\theta') - \mathcal{L}(\theta) - \langle \theta' - \theta, \nabla\mathcal{L}(\theta)\rangle\,. \end{aligned} \tag{1.21}$$

A particular type of exponential families of interest are the *single parameter exponential families* (SPEF), i.e. $\Theta \subseteq \mathbb{R}$. For instance, for a fixed $\sigma \in \mathbb{R}_+^\star$, $\{\mathcal{N}(\mu,\sigma^2),\ \mu \in \mathbb{R}\}$ forms a SPEF. Such statistical models are especially appealing in stochastic bandits since the extremal Kullback-Leibler operator has an explicit expression for SPEF. Indeed, for $\mathcal{F}^{\mathrm{EF}}_{h,F,\mathcal{L}} = \{p_\theta,\ \theta \in \Theta\}$ and risk measure $\rho\colon p_\theta \in \mathcal{F} \mapsto \mathbb{E}_\theta[F(Y)]$, the properties of univariate optimisation shows that:

$$\forall \theta \in \Theta, \forall \rho^\star > \rho(p_\theta),\ \mathcal{K}_{\inf}^{\mathcal{F},\rho}(p_\theta;\rho^\star) = \mathrm{KL}(p_\theta \parallel p_{\theta^\star}) = \mathcal{B}_\mathcal{L}(\theta^\star,\theta)\,, \tag{1.22}$$

where $\theta^\star \in \Theta$ is such that $\mathbb{E}_{\theta^\star}[Y] = \rho^\star$ (assuming it exists).

**Nonparametric models.** A typical example of nonparametric model is given by the family of distributions over $\mathbb{R}$ with a given bounded support, i.e.

$$\mathcal{F}_{[\underline{B},\overline{B}]} = \{\nu \in \mathcal{M}_1^+(\mathbb{R}),\ \mathrm{Supp}\,\nu \subseteq [\underline{B},\overline{B}]\} \tag{1.23}$$

for some $\underline{B},\overline{B} \in \mathbb{R}$. Such families cannot be embedded in a finite-dimensional space: indeed, any continuous function $p\colon [\underline{B},\overline{B}] \to \mathbb{R}$ (which forms an infinite-dimensional vector space) generates a distribution in $\mathcal{F}_{[\underline{B},\overline{B}]}$ with density $|p|/\int_{\underline{B},\overline{B}}|p(x)|dx$. These statistical models are popular in stochastic bandits for a similar reason as the SPEF, i.e. their extremal Kullback-Leibler operator can be computed easily with the dual form (Honda and Takemura, 2015):

$$\forall \nu \in \mathcal{F}_{[\underline{B},\overline{B}]}, \forall \mu^\star > \mathbb{E}_{Y\sim\nu}[Y],\ \mathcal{K}_{\inf}^{\mathcal{F}_{[\underline{B},\overline{B}]}}(\nu;\mu^\star) = \max_{\lambda \in [0,\frac{1}{\overline{B}-\mu^\star}]} \mathbb{E}_{Y\sim\nu}\left[\log\left(1 - \lambda(Y-\mu^\star)\right)\right]\,. \tag{1.24}$$

Note that the right-hand side does not depend on $\underline{B}$, and in fact $\mathcal{K}_{\inf}^{\mathcal{F}_{[\underline{B},\overline{B}]}}(\nu;\mu^\star) = \mathcal{K}_{\inf}^{\mathcal{F}_{(-\infty,\overline{B}]}}(\nu;\mu^\star)$ (Honda and Takemura, 2015, Theorem 2).

A critical property of Gaussian distributions is the behaviour of their tails, characterised by their moment generating function (MGF) $\lambda \in \mathbb{R} \mapsto \mathbb{E}_{Y\sim\mathcal{N}(\mu,\sigma^2)}[\exp(\lambda(Y-\mu))] = \exp(\lambda^2\sigma^2/2)$ for $\mu \in \mathbb{R}$ and $\sigma \in \mathbb{R}_+^\star$. A natural extension is to consider *sub-Gaussian* distributions defined by

$$\mathcal{F}_{\mathcal{G},R} = \{\nu \in \mathcal{M}_1^+(\mathbb{R}),\ \forall \lambda \in \mathbb{R},\ \mathbb{E}_{Y\sim}[\exp(\lambda(Y - \mathbb{E}_{Y\sim\nu}))] \leqslant \exp(\lambda^2R^2/2)\} \tag{1.25}$$

for $R \in \mathbb{R}_+^\star$, i.e. their MGF are upper bounded by a Gaussian MGF. Bounded distributions form a subset of sub-Gaussian distributions (Hoeffding's lemma, see Hoeffding (1963)):

$$\mathcal{F}_{[\underline{B},\overline{B}]} \subset \mathcal{F}_{\mathcal{G},(\overline{B}-\underline{B})/2}\,. \tag{1.26}$$

For Gaussian distributions, the parameter $\sigma^2$ that controls the MGF also represents the variance; for sub-Gaussian distributions, it is only an upper bound, as shown by the following lemma.

**Lemma 1.12**. *Let $R \in \mathbb{R}_+^\star$ and $\nu \in \mathcal{F}_{\mathcal{G},R}$. Then $\mathbb{V}_{Y\sim\nu}[Y] \leqslant R^2$.*

In particular, the MGF of any distribution $\nu$, if it is defined in a neighbourhood of zero, is equivalent to $\lambda \mapsto 1 + \mathbb{V}_{Y\sim\nu}[Y]\lambda^2$. The sub-Gaussian condition simply states that this relation holds globally for all $\lambda \in \mathbb{R}$. A less stringent requirement is to ask that the MGF is locally defined, which gives rise to the notion of *light tailed* distributions

$$\mathcal{F}_\ell = \left\{\nu \in \mathcal{M}_1^+(\mathbb{R}),\ \exists\lambda_\nu \in \mathbb{R}_+^\star,\ \forall\lambda \in (-\lambda_\nu, \lambda_\nu),\ \mathbb{E}_{Y\sim\nu}\left[e^{\lambda Y}\right] < +\infty\right\}\,. \tag{1.27}$$

In particular, if the MGF exists locally, we may obtain the $k$-th moments by differentiating it $k \in \mathbb{N}$ times, and thus moments of all order exist. An even less stringent condition is thus to ask only for the existence of a certain (raw or central) moment, i.e.

$$\mathcal{F}_{M,\varepsilon} = \left\{\nu \in \mathcal{M}_1^+(\mathbb{R}),\ \mathbb{E}_{Y\sim\nu}\left[|Y|^{1+\varepsilon}\right] \leqslant M\right\} \text{ or} \tag{1.28}$$

$$\mathcal{F}_{\kappa,\varepsilon}^{\text{centred}} = \left\{\nu \in \mathcal{M}_1^+(\mathbb{R}),\ \mathbb{E}_{Y\sim\nu}\left[|Y - \mathbb{E}_{Y\sim\nu}[Y]|^{1+\varepsilon}\right] \leqslant \kappa\right\} \tag{1.29}$$

for some $\varepsilon, M, \kappa \in \mathbb{R}_+^\star$. These form a subset of the so-called *heavy tailed* distributions.

Nonparametric models capture a much broader range of distributions than parametric ones and are less prone to model misspecification. However, this comes at the cost of typically worse algorithmic performances. For bandits, achieving logarithmic regret may be impossible when the underlying statistical model is too large, as evidenced by the following lemma.

**Lemma 1.13** (Obstruction to logarithmic pseudo regret). *If* $\mathcal{F} = \cup_{\underline{B},\overline{B}\in\mathbb{R}}\mathcal{F}_{[\underline{B},\overline{B}]}$ *(all bounded distributions),* $\mathcal{F} = \cup_{R\in\mathbb{R}_+^\star}\mathcal{F}_{\mathcal{G},R}$ *(all sub-Gaussian distributions),* $\mathcal{F} = \mathcal{F}_\ell$ *(all light tailed distributions),* $\mathcal{F} = \cup_{M\in\mathbb{R}_+^\star}\mathcal{F}_{M,\varepsilon} = \cup_{\kappa\in\mathbb{R}_+^\star}\mathcal{F}_{\kappa,\varepsilon}^{centred}$ *for some* $\varepsilon \in \mathbb{R}_+^\star$ *(all distributions with finite* $1+\varepsilon$*-th moment), or* $\mathcal{F} = \mathcal{M}_1^+(\mathbb{R})$ *(all probability distributions), then*

$$\forall \nu \in \mathcal{F},\ \forall \mu^\star > \mathbb{E}_{Y\sim\nu}[Y],\ \mathcal{K}_{\inf}^{\mathcal{F}}(\nu;\mu^\star) = 0\,. \tag{1.30}$$

The counterexample used in the proof (see Appendix A.2) for bounded distributions is borrowed from Hadiji and Stoltz (2023, Theorem 3, Theorem 11); see also Ashutosh et al. (2021) and Agrawal et al. (2021) in the sub-Gaussian and finite moment settings respectively. Informally, such counterexamples show that the main obstruction to logarithmic regret is the possible "mass leakage at infinity" that results from a lack of compactness in the family $\mathcal{F}$ (in bandits idiom, we say that this allows arbitrarily close *confusing instances* to exist).

**Table 1.1** – Summary of classical statistical models for stochastic bandits.

| Name | Definition | Dimension |
|---|---|---|
| Gaussian fixed variance $\sigma^2 \in \mathbb{R}_+^\star$ | $\mathcal{F}_{\mathcal{N},\sigma} = \left\{ p_\mu \colon y \in \mathbb{R} \mapsto \frac{1}{\sqrt{2\pi}\sigma} e^{-\frac{(y-\mu)^2}{2\sigma^2}}, \mu \in \mathbb{R} \right\}$ | 1 |
| SPEF | $\mathcal{F}^{\text{SPEF}}_{h,F,\mathcal{L}} = \left\{ p_\theta \colon y \in \mathcal{Y} \mapsto h(y) e^{\theta F(y) - \mathcal{L}(\theta)}, \theta \in \Theta \subseteq \mathbb{R} \right\}$ | 1 |
| Gaussian | $\mathcal{F}_{\mathcal{N}} = \left\{ p_{\mu,\sigma} \colon y \in \mathbb{R} \mapsto \frac{1}{\sqrt{2\pi}\sigma} e^{-\frac{(y-\mu)^2}{2\sigma^2}}, (\mu,\sigma) \in \mathbb{R} \times \mathbb{R}_+^\star \right\}$ | 2 |
| EF | $\mathcal{F}^{\text{EF}}_{h,F,\mathcal{L}} = \left\{ p_\theta \colon y \in \mathcal{Y} \mapsto h(y) e^{\langle \theta, F(y) \rangle - \mathcal{L}(\theta)}, \theta \in \Theta \subseteq \mathbb{R}^d \right\}$ | $d$ |
| Bounded $\underline{B}, \overline{B} \in \mathbb{R}$ | $\mathcal{F}_{[\underline{B},\overline{B}]} = \left\{ \nu \in \mathcal{M}_1^+(\mathbb{R}),\ \text{Supp}\ \nu \subseteq [\underline{B}, \overline{B}] \right\}$ | $\infty$ |
| Semibounded $B \in \mathbb{R}$ | $\mathcal{F}_{(-\infty,B]} = \{ \nu \in \mathcal{M}_1^+(\mathbb{R}),\ \text{Supp}\ \nu \subseteq (-\infty, B] \} \cap \mathbb{L}^1(\mathbb{R})$ or $\mathcal{F}_{[B,\infty)} = \{ \nu \in \mathcal{M}_1^+(\mathbb{R}),\ \text{Supp}\ \nu \subseteq [B, \infty) \} \cap \mathbb{L}^1(\mathbb{R})$ | $\infty$ |
| Sub-Gaussian $R \in \mathbb{R}_+^\star$ | $\mathcal{F}_{\mathcal{G},R} = \left\{ \nu \in \mathcal{M}_1^+(\mathbb{R}),\ \forall \lambda \in \mathbb{R},\ \mathbb{E}_{Y\sim\nu}\left[ e^{\lambda(Y - \mathbb{E}_{Y\sim\nu}[Y])} \right] \leqslant e^{\frac{\lambda^2 R^2}{2}} \right\}$ | $\infty$ |
| Light tailed | $\mathcal{F}_\ell = \left\{ \nu \in \mathcal{M}_1^+(\mathbb{R}),\ \exists \lambda_\nu \in \mathbb{R}_+^\star,\ \forall \lambda \in (-\lambda_\nu, \lambda_\nu),\ \mathbb{E}_{Y\sim\nu}\left[ e^{\lambda Y} \right] < +\infty \right\}$ | $\infty$ |
| Finite moment $M \in \mathbb{R}_+^\star, \varepsilon \in \mathbb{R}_+^\star$ | $\mathcal{F}_{M,\varepsilon} = \left\{ \nu \in \mathcal{M}_1^+(\mathbb{R}),\ \mathbb{E}_{Y\sim\nu}\left[ \lvert Y \rvert^{1+\varepsilon} \right] \leqslant M \right\}$ | $\infty$ |
| Finite moment (centred) $\kappa \in \mathbb{R}_+^\star, \varepsilon \in \mathbb{R}_+^\star$ | $\mathcal{F}^{\text{centred}}_{\kappa,\varepsilon} = \left\{ \nu \in \mathcal{M}_1^+(\mathbb{R}),\ \mathbb{E}_{Y\sim\nu}\left[ \lvert Y - \mathbb{E}_{Y\sim\nu}[Y] \rvert^{1+\varepsilon} \right] \leqslant \kappa \right\}$ | $\infty$ |

## 1.4 How to play (provably) well? Concentration and martingales

In this section, we introduce fundamental results that are ubiquitous in statistics and probability theory, known as the *concentration of measure* phenomenon. Not only do they offer powerful tools to analysis bandit algorithms (and beyond the scope of this these, machine learning at large), they also directly dictate the design of many policies, known as UCB (see Theorem 1.33). We adopt a fairly generic formulation that unifies several known results as the methods presented here form the basis of several key results in Chapters 3, 4 and 5. We conclude with two original contributions: an order-optimal time-uniform mixture bound with mixing parameter time peeling, and a generic analysis of UCB with arbitrary confidence sequences.

### Fixed sample concentration and the Cramér-Chernoff method.

We recall here the standard definition of martingales and the classical Cramér-Chernoff bound.

**Definition 1.14** (Martingale). *Let $\mathcal{T} = \mathbb{N}$ or $\mathcal{T} = \mathbb{R}_+$, $(\mathcal{G}_t)_{t\in\mathcal{T}}$ a filtration and $(\mathcal{Y}, \|\cdot\|)$ a normed space. A stochastic process $(M_t)_{t\in\mathcal{T}}$ is said to be a $(\mathcal{G}_t)_{t\in\mathcal{T}}$-**supermartingale** if for all $t \in \mathcal{T}$, (i) $M_t$ is $\mathcal{G}_t$-measurable, (ii) $\mathbb{E}[\|M_t\|] < +\infty$, and (iii) $\mathbb{E}[M_{t+1} \mid \mathcal{G}_t] \leqslant M_t$. If instead $\mathbb{E}[M_{t+1} \mid \mathcal{G}_t] \geqslant M_t$ for all $t \in \mathcal{T}$, it is said to be a $(\mathcal{G}_t)_{t\in\mathcal{T}}$-**submartingale**. Finally, it said to be a $(\mathcal{G}_t)_{t\in\mathcal{T}}$-**martingale** if it is both a supermartingale and a submartingale.*

A typical example of a martingale is $(\sum_{s=1}^t Y_s - \mu)_{t\in\mathbb{N}}$ where $(Y_t)_{t\in\mathbb{N}}$ is an i.i.d. sequence of integrable random variables with expectation $\mu$. In this thesis, we will mostly consider stochastic processes over $\mathbb{R}$ or $\mathbb{R}^d$, in which case the norm $\|\cdot\|$ is implicitly set to be the corresponding Euclidean norm. Moreover, for a given a nondecreasing, right-continuous mapping $F\colon \mathbb{R} \to \mathbb{R}$, we define the *generalised inverse* (de La Fortelle, 2020) of $F$ as $F^{-1}\colon u \in \mathbb{R} \mapsto \inf\{y \in \mathbb{R},\ F(y) > u\}$, which satisfies in particular the order-preserving property that $F^{-1}(u) \leqslant y$ implies $u \leqslant F(y)$ for all $y \in \mathbb{R}$ and $u \in \mathbb{R}_+$ (a similar definition exist for nonincreasing mappings, which we skip to avoid cluttering). When $F$ is invertible, its generalised inverse is equal to its inverse, which we also denote by $F^{-1}$.

**Proposition 1.15** (Generic Cramér-Chernoff bound). *Let $t \in \mathbb{N}$, $(\mathcal{G}_s)_{s=1}^t$ a filtration, $(S_s)_{s=1}^t$ a $(\mathcal{G}_s)_{s=1}^t$-adapted process, an open interval $\mathcal{I} \subseteq \mathbb{R}$ with $0 \in \mathcal{I}$ and a function $\psi \colon \mathcal{I} \to \mathbb{R}_+$ such that for all $s \in \{1, \dots, t-1\}$ and $\lambda \in \mathcal{I}$, we have $\log \mathbb{E}[\exp(\lambda(S_{s+1} - S_s) \mid \mathcal{G}_s] \leqslant \psi(\lambda)$. We let $\mathcal{I}_+ = \mathcal{I} \cap \mathbb{R}_+$ and $\mathcal{I}_- = \mathcal{I} \cap \mathbb{R}_-$ and define the Fenchel-Legendre transforms:*

$$\psi^{\star,+} \colon u \in \mathbb{R}_+ \mapsto \sup_{\lambda \in \mathcal{I}_+} \lambda u - \psi(\lambda) \quad \textit{and} \quad \psi^{\star,-} \colon u \in \mathbb{R}_+ \mapsto \sup_{\lambda \in \mathcal{I}_-} \lambda u - \psi(\lambda) \,. \tag{1.31}$$

*Then for any $\delta \in (0,1)$, we have*

$$\mathbb{P}\left(S_t \geqslant t(\psi^{\star,+})^{-1}\left(\frac{1}{t}\log\frac{1}{\delta}\right)\right) \leqslant \delta \quad \textit{and} \quad \mathbb{P}\left(S_t \leqslant t(\psi^{\star,-})^{-1}\left(\frac{1}{t}\log\frac{1}{\delta}\right)\right) \leqslant \delta \,. \tag{1.32}$$

Note that this generic bound applies to martingales $(S_t)_{t\in\mathbb{N}}$, not just sums of i.i.d. random variables (sometimes called *weak dependency*), as long as the control by $\psi$ holds. In particular, this is the case for sub-Gaussian distributions with $\psi \colon \lambda \in \mathbb{R} \mapsto \lambda^2 R^2/2$ for some $\mathbb{R}_+^\star$.

**Corollary 1.16** (Cramér-Chernoff sub-Gaussian bound). *Let $R \in \mathbb{R}_+^\star$, $t \in \mathbb{N}$ and $(Y_s)_{s=1}^t$ a sequence of i.i.d. random variables drawn from $\nu \in \mathcal{F}_{\mathcal{G},R}$ with expectation $\mu \in \mathbb{R}$. For any $\delta \in (0,1)$, we consider $\widehat{\mu}_t = 1/t \sum_{s=1}^t Y_s$ and*

$$\widehat{\Theta}_t^\delta = \left[\widehat{\mu}_t \pm R\sqrt{\frac{2}{t}\log\frac{2}{\delta}}\right] , \tag{1.33}$$

*Then $\widehat{\Theta}_t^\delta$ is a **confidence set** at level $\delta$ for $\mu$, i.e. $\mathbb{P}(\mu \in \widehat{\Theta}_t) \geqslant 1 - \delta$.*

### Time-uniform concentration and the method of mixtures

The main limitation of the above confidence sets is that they are only valid for a *fixed* sample size $t$, i.e. we control $\forall t \in \mathbb{N}$, $\mathbb{P}(\mu \in \widehat{\Theta}_t^\delta)$ rather than the much stronger statement $\mathbb{P}(\forall t \in \mathbb{N},\ \mu \in \widehat{\Theta}_t^\delta)$. The latter is of utmost importance to handle sequential settings where data is actively collected by an agent, rather than studied retrospectively. Stochastic bandits offer a typical example of such settings: although each arm $k \in [K]$ generates i.i.d. samples, the number of observations $N_t^k$ at time $t \in \mathbb{N}$ is a (random) function of past observations and the policy of the agent. Similarly, model-based reinforcement learning (Jaksch et al., 2010) suggests exploration strategies in Markov decision processes from random numbers of visits to state-action pairs.

**Definition 1.17** (Time-uniform confidence sequence). *Let $\mathcal{F} \subseteq \mathcal{M}_1^+(\mathcal{Y})$, $\rho \colon \mathcal{F} \to \mathbb{R}^d$ a mapping and $\delta \in (0,1)$. A **time-uniform confidence sequence** at level $\delta$ for $\rho$ and $\mathcal{F}$ is a sequence of mappings $(\Theta_t^\delta)_{t\in\mathbb{N}}$ such that for $t \in \mathbb{N}$, $\widehat{\Theta}_t^\delta \colon \mathcal{Y}^t \to \mathfrak{P}(\mathbb{R}^d)$ and for all $\nu \in \mathcal{F}$, we have*

$$\mathbb{P}_{\mathbb{Y}\sim\nu^{\otimes\mathbb{N}}}\left(\forall t \in \mathbb{N}, \rho(\nu) \in \widehat{\Theta}_t^\delta\left((Y_s)_{s=1}^t\right)\right) \geqslant 1-\delta\,, \tag{1.34}$$

*where $\mathbb{Y} = (Y_t)_{t\in\mathbb{N}} \sim \nu^{\otimes\mathbb{N}}$ denotes an i.i.d. sequence of random variables drawn from $\nu$. When the context is clear, we identify the mapping $\widehat{\Theta}_t^\delta$ with the set $\widehat{\Theta}_t^\delta((Y_s)_{s=1}^t)$, i.e. $(\widehat{\Theta}_t^\delta)_{t\in\mathbb{N}}$ is a set-valued stochastic process adapted to the natural filtration of $\mathbb{Y}$.*

In this thesis, confidence sequences typically consist of compact, convex subsets of $\mathbb{R}^d$, e.g. for a given $t \in \mathbb{N}$, $\Theta_t^\delta$ is a closed segment $(d=1)$ or is (contained in) an ellipsoid $(d \geqslant 2)$. Moreover for any fixed $t \in \mathbb{N}$, $\delta \in (0,1) \mapsto \widehat{\Theta}_t^\delta$ is typically a nonincreasing mapping (for the set inclusion), i.e. lower $\delta$ means less room for errors and thus larger confidence sets.

**Remark 1.18** (Equivalence between time-uniform and random time confidence). *Howard et al. (2020, Lemma 3) shows that the time-uniform bound $\mathbb{P}(\forall t \in \mathbb{N},\ \rho(\nu) \in \widehat{\Theta}_t^\delta) \geqslant 1-\delta$ is equivalent to $\mathbb{P}(\rho(\nu) \in \widehat{\Theta}_\tau^\delta) \geqslant 1-\delta$ for any adapted (not necessarily stopping) random time $\tau$.*

**Of p-values and e-values.** We briefly highlight the connections between confidence sets and p-values, which are ubiquitous in statistical hypothesis testing.

**Definition 1.19** (P-values). *Let $\mathcal{F} \subset \mathcal{M}_1^+(\mathbb{R})$, $\rho\colon \mathcal{F} \to \mathbb{R}$ and $t \in \mathbb{N}$ such that for all $\delta \in (0,1)$, $\widehat{\Theta}_t^\delta$ is a confidence set at level $\delta$ for $\rho$ and $\mathcal{F}$. Fix $\rho_0 \in \mathbb{R}$. The **p-variable** associated to the confidence sets $(\widehat{\Theta}_t^\delta)_{\delta \in (0,1)}$ and null hypothesis $\mathcal{H}_0 = \{\nu \in \mathcal{F},\ \rho(\nu) = \rho_0\}$ is the random variable defined as*

$$P_t^{\mathcal{H}_0} = \inf\left\{\delta \in (0,1),\ \rho_0 \notin \widehat{\Theta}_t^\delta\right\} . \tag{1.35}$$

*In particular, we have that $\mathbb{P}_{\mathbb{Y}\sim\nu^{\otimes t}}(P_t^{\mathcal{H}_0} \leqslant \delta) \leqslant \delta$ for all $\delta \in (0,1)$ and $\nu \in \mathcal{H}_0$. If for all $\delta \in (0,1)$, $(\widehat{\Theta}_t^\delta)_{t\in\mathbb{N}}$ is a time-uniform confidence sequence for all $\delta \in (0,1)$, then the stochastic process $(P_t^{\mathcal{H}_0})_{t\in\mathbb{N}}$ is called a **time-uniform p-process**, which satisfies in particular $\mathbb{P}_{\mathbb{Y}\sim\nu^{\otimes\mathbb{N}}}(\exists t \in \mathbb{N},\ P_t^{\mathcal{H}_0} \leqslant \delta) \leqslant \delta)$ and $\mathbb{P}_{\mathbb{Y}\sim\nu^{\otimes\mathbb{N}}}(P_\tau^{\mathcal{H}_0} \leqslant \delta) \leqslant \delta)$ for any adapted $\mathbb{N}$-valued random time $\tau$. Realisations of p-variables and p-processes are called **p-values** and **time-uniform p-values** respectively.*

In the light of this definition, the confidence parameter $\delta$ is an upper bound on the false discovery rate, or *type I error*, i.e. the probability of rejecting the null ($\rho_0 \notin \widehat{\Theta}_t^\delta$) even though the observed samples ($\mathbb{Y}$) were drawn from $\nu \in \mathcal{H}_0$. A typical threshold for type I error is $\delta = 0.05$.

Recently, p-values have come under scrutiny due to their frequent misuse and erroneous interpretation (Wasserstein and Lazar, 2016), leading to the so-called replication crisis in empirical sciences (Pashler and Wagenmakers, 2012). A major flaw of classical p-variables built from fixed samples confidence sets is their lack of robustness to stochastic sampling, i.e. $\mathbb{P}_{\mathbb{Y}\sim\nu^{\otimes\mathbb{N}}}(P_\tau^{\mathcal{H}_0} \leqslant \delta) \leqslant \delta$ is not guaranteed if $\tau$ is random. In particular, if an experiment is designed with sample size $T$, such p-variables are invalid under optional stopping (concluding the experiments after $t$ samples if $P_t^{\mathcal{H}_0} < 0.05$), optional continuation (collecting additional data $t > T$ if $P_T^{\mathcal{H}_0} > 0.05$) or even data peeking (computing $P_t^{\mathcal{H}_0}$ before $P_T^{\mathcal{H}_0}$).

To mitigate this issue, Johari et al. (2022) introduced the notion of anytime-valid p-values, which correspond to our definition of time-uniform p-processes. A related topic, that has gained widespread attention in the recent years, is that of *e-values* and *e-processes*, which are nonnegative processes $(E_t)_{t\in\mathbb{N}}$ such that $\mathbb{E}_{\mathbb{Y}\sim\nu^{\otimes\mathbb{N}}}[E_\tau] \leqslant 1$ for every adapted stopping time $\tau$ and hypothesis $\nu \in \mathcal{H}_0$ (in particular, this property allows for seamless testing under optional stopping or continuation). Equivalently, e-processes are envelopes of so-called *test martingales*, i.e. there exists a family of nonnegative martingales $\{(M_t^\nu)_{t\in\mathbb{N}},\ \nu \in \mathcal{H}_0\}$ with $M_0 = 1$ such that $E_t \leqslant M_t^\nu$ for all $t \in \mathbb{N}$. While we do not focus on such processes in this thesis, we use and introduce several time-uniform concentration results derived from nonnegative (super)martingales (see below, and Chapters 3 and 4 in particular), which are thus examples of e-processes. We refer to Grünwald et al. (2020); Vovk and Wang (2021) for an introduction, Turner and Grunwald (2023); Turner and Grünwald (2023) for applications to contingency tables and count data, and Ramdas et al. (2022) for a recent overview of this rapidly growing field.

**Numerical experiments on p-values.** As a witty illustrative example of the importance of time-uniform statistics, let us consider three friends playing a card game on holidays. At the end of each round, players are ranked as winner, neutral and looser. One of them (who may be the author of this thesis) started in a streak of losing hands, and was prone to stop playing after finishing a few rounds as looser. After a while, this player wanted to test whether he was unlucky or plain bad at the game. To this end, he considered the statistical Bernoulli model $\mathcal{F} = \{\mathcal{B}(p),\ p \in (0,1)\}$ ($p$ represents the probability of loosing a round) and the null hypothesis $\mathcal{H}_0 = \{p \leqslant 1/3\}$. The three friends played a total of $T = 226$ rounds across $N = 29$ sessions. Denoting by $(\tau_i)_{i=1}^N$ the (random) length of these sessions, they collected a dataset $\mathbb{Y} = ((Y_t^i)_{t=1}^{\tau_i})_{i=1}^N$ where each $(Y_t^i)_{t=1}^{\tau_i}$ is an i.i.d. sequence of random variable drawn from a distribution $\nu \in \mathcal{F}$. We report in Figure 1.2 the fixed sample and time-uniform p-values computed after each completed round (using the classical binomial test and the time-uniform Bregman confidence sequence for the Bernoulli SPEF derived in Chapter 3 respectively) on synthetic data drawn from $\mathcal{B}(1/2)$ (in $N = 1$ session) and on real data (in $N = 29$ sessions).

In the synthetic experiments, both methods easily rejected the null (which is indeed false) at $\delta = 0.05$ after 226 rounds. Being more conservative in nature, the time-uniform p-process started rejecting after about 100 rounds, which was much later than the first rejection of the fixed sample p-variable (about 25 rounds). However, computing in hindsight this p-variable renders it statistically invalid (peeking at the data, i.e. *p-hacking*); only the time-uniform process allows to prematurely stop the experiment while preserving type I error guarantee. In the real-world experiment, the resulting p-values after playing 226 rounds were 0.06 and 0.93 respectively, thus failing to reject the null (i.e. that player was not so bad after all). However, we see that the p-variable was deceived and would have rejected the null had the experiment been stopped after about 60 to 210 rounds (but again, without guarantees on the type I error). By contrast, the time-uniform p-process never rejected the null, which we interpret as a sign of its robustness to stochastic sampling. Indeed, although the outcomes of each round are modelled as i.i.d. random variables, the first sessions were stopped due to the early streak of bad luck (and grumpiness) of one player, thus skewing the results towards more losses.

**Nonnegative supermartingales and method of mixtures.** We now introduce the fundamental result of time-uniform concentration, which is essentially a corollary of Doob's maximal inequality for nonnegative supermartingale, also known as Ville's inequality (Ville, 1939). We present it here in a slightly more general formulation, and derive it from first principles to highlight the sort of supermartingale constructions that we use throughout this thesis.

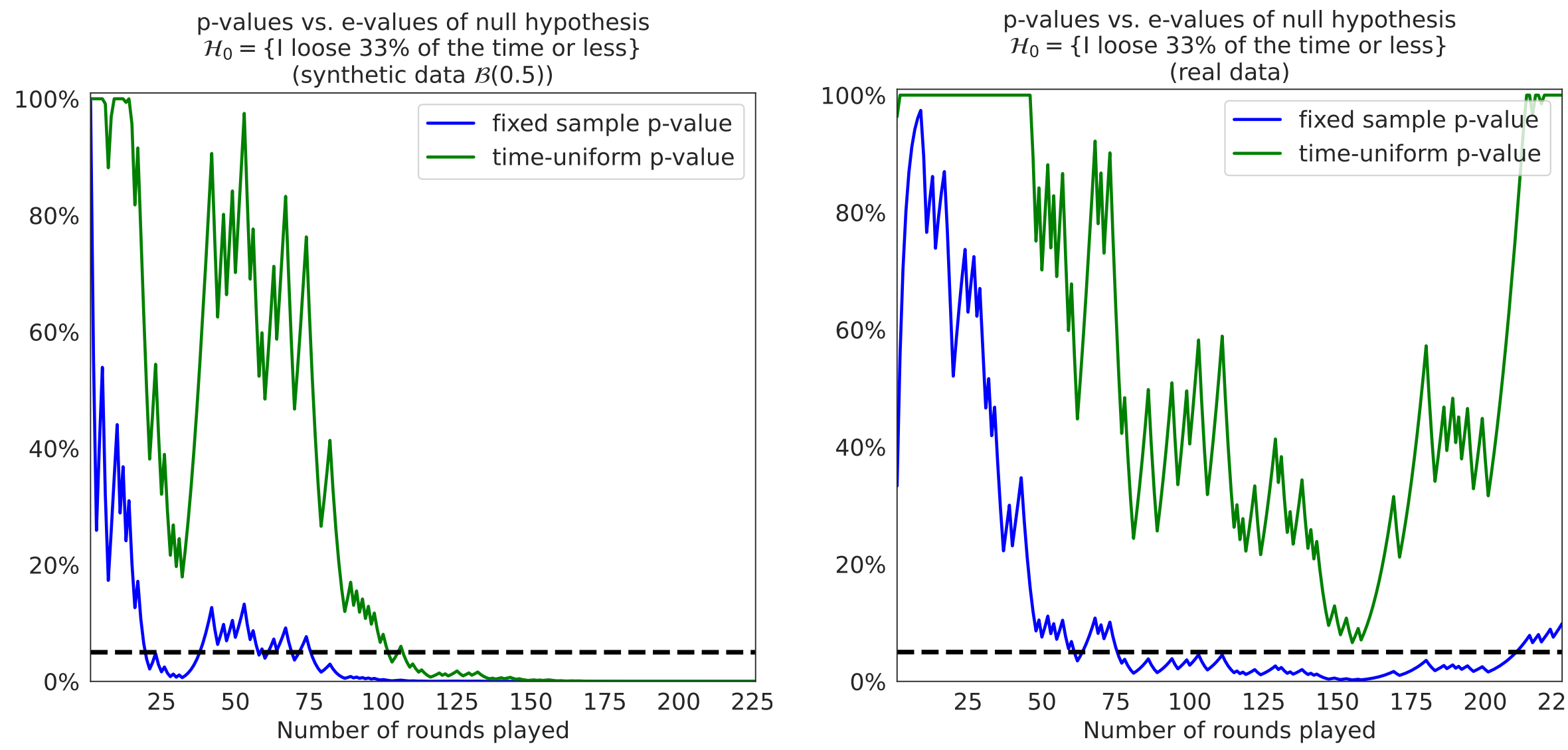

**Figure 1.2** – Realisations of p-variables (fixed sample) and p-processes (time-uniform) to test hypothesis on the skill of a card player. Left: synthetic experiment (226 i.i.d. realisations of $\mathcal{B}(1/2)$). Right: real data (226 rounds across 29 sessions). Black dashed line: significance level $\delta = 0.05$.

**Theorem 1.20** (Time-uniform concentration of nonnegative supermartingales). *Let $\mathcal{T} = \mathbb{N}$ or $\mathcal{T} = \mathbb{R}_+$, $(S_t)_{t\in\mathcal{T}}$ a real stochastic process adapted to a filtration $(\mathcal{G}_t)_{t\in\mathcal{T}}$, $(F_t)_{t\in\mathcal{T}}$ a family of continuous mappings $\mathbb{R} \to \mathbb{R}_+$ and $t_0 \in \mathcal{T}$ such that*

(i) $(M_t)_{t\geqslant t_0} = (F_t(S_t))_{t\geqslant t_0}$ *is a nonnegative $(\mathcal{G}_t)_{t\geqslant t_0}$-supermartingale,*

(ii) $\mathbb{E}[M_{t_0}] \leqslant 1$,

(iii) *All mappings $(F_t)_{t\geqslant t_0}$ are either nondecreasing or nonincreasing.*

*Then for any $\delta \in (0,1)$ and any $(\mathcal{G}_t)_{t\in\mathbb{N}}$-stopping time $\tau$ such that almost surely $\tau \geqslant t_0$, we have*

$$\mathbb{P}\left(S_\tau \geqslant F_\tau^{-1}\left(\frac{1}{\delta}\right)\right) \leqslant \delta \quad \text{or} \quad \mathbb{P}\left(S_\tau \leqslant F_\tau^{-1}\left(\frac{1}{\delta}\right)\right) \leqslant \delta\,, \tag{1.36}$$

*depending on whether the $(F_t)_{t\geqslant t_0}$ are nondecreasing or nonincreasing. Furthermore, we have*

$$\mathbb{P}\left(\forall t \geqslant t_0,\ S_t \geqslant F_t^{-1}\left(\frac{1}{\delta}\right)\right) \leqslant \delta \quad \text{or} \quad \mathbb{P}\left(\forall t \geqslant t_0,\ S_t \leqslant F_t^{-1}\left(\frac{1}{\delta}\right)\right) \leqslant \delta\,. \tag{1.37}$$

In the sub-Gaussian case, we couple the above generic bound with a specific supermartingale construction obtained by *mixing* the processes $(M_t^\lambda)_{t\in\mathbb{N}}$ over $\lambda \in \mathbb{R}$.

**Corollary 1.21** (Time-uniform sub-Gaussian concentration with the method of mixtures). *Let $R \in \mathbb{R}_+^\star$ and $(Y_t)_{t\in\mathbb{N}}$ be a sequence of i.i.d. random variables drawn from $\nu \in \mathcal{F}_{\mathcal{G},R}$ with expectation $\mu \in \mathbb{R}$. For any $\alpha \in \mathbb{R}_+^\star$ and $\delta \in (0,1)$, we consider $(\widehat{\mu}_t)_{t\in\mathbb{N}} = (1/t\sum_{s=1}^t Y_s)_{t\in\mathbb{N}}$ and*

$$\left(\widehat{\Theta}_{t,\alpha}^{\delta}\right)_{t\in\mathbb{N}} = \left(\left[\widehat{\mu}_t \pm R\sqrt{\frac{2}{t}\left(1+\frac{\alpha}{t}\right)\log\left(\frac{2}{\delta}\sqrt{1+\frac{t}{\alpha}}\right)}\right]\right)_{t\in\mathbb{N}} . \tag{1.38}$$

*Then $(\Theta_{t,\alpha}^{\delta})_{t\in\mathbb{N}}$ is a **time-uniform confidence sequence** at level $\delta$ for $\mu$, i.e.*

$$\mathbb{P}\left(\forall t \in \mathbb{N},\ \mu \in \widehat{\Theta}_{t,\alpha}^{\delta}\right) \geqslant 1-\delta\,, \tag{1.39}$$

*or equivalently for any adapted random time $\tau$ in $\mathbb{N}$,*

$$\mathbb{P}\left(\mu \in \widehat{\Theta}_{\tau,\alpha}^{\delta}\right) \geqslant 1-\delta\,. \tag{1.40}$$

**Remark 1.22** (Running intersection). *Another advantage of the time-uniform formulation is that if $(\widehat{\Theta}_t^{\delta})_{t\in\mathbb{N}}$ is a time-uniform confidence sequence at level $\delta$, then the running intersections $(\cap_{s\leqslant t}\widehat{\Theta}_s^{\delta})_{t\in\mathbb{N}}$, called the **confidence envelopes**, also form a time-uniform confidence sequence at the same level, which is at least as tight as the original sequence. Taking the running intersection is standard in sequential testing and was pioneered in Darling and Robbins (1967).*

We notice that the confidence sequence derived from the method of mixtures depends on a hyperparameter $\alpha$, that we call the *mixing parameter*. The following elementary lemma, which appeared in Howard et al. (2021, Proposition 3), suggests a natural tuning of $\alpha$.

**Lemma 1.23** (Tuning of the mixing parameter). *Let $t_0 \in \mathbb{N}$ and $\delta \in (0,1)$. Then we have $\gamma_\delta t_0 = \operatorname{argmin}_{\alpha\in\mathbb{R}_+^\star}\sqrt{2(t+\alpha)\log(\sqrt{1+t/\alpha}/\delta)}$, where $\gamma_\delta = -1/(1+W_{-1}(-\delta^2/e))$ and $W_{-1}$ is the first negative branch of the Lambert W function. In particular, in the R-sub-Gaussian case, we have $\gamma_{\delta/2}t_0 = \operatorname{argmin}_{\alpha\in\mathbb{R}_+^\star}|\widehat{\Theta}_{t_0,\alpha}^{\delta}|$.*

For $\delta = 0.05$, we have $\gamma_\delta \approx 0.12$ for the one-sided bound and $\gamma_{\delta/2} \approx 0.10$ for the two-sided confidence interval. Importantly, this tuning only works for a *fixed* sample size $t_0$. Note that fast numerical implementations of the Lambert $W$ function are readily available in many scientific libraries, e.g. *scipy.special.lambertw* in Python.

**Remark 1.24** (Mixtures beyond the sub-Gaussian case). *The main idea of the method of mixtures is to find a suitable mixing measure $\nu_\Lambda$ to compute the mixture supermartingale $(\int M_t^\lambda d\nu_\Lambda(\lambda))_{t\in\mathbb{N}}$. This is akin to finding a conjugate prior in Bayesian estimation to make such integral calculations tractable. Thanks to the natural Gaussian structure of the supermartingales $(M_t^\lambda)_{t\in\mathbb{N}}$ in the sub-Gaussian case, Gaussian measures make for suitable mixing measures in this setting. Under other assumptions, such as distributions with bounded supports, there is no such natural mixing measures. Of note, Waudby-Smith and Ramdas (2023) introduced a construction that replaces the mixture process with a so-called* hedged capital process *with predictable weights. We detail this construction, which we will use several times as a benchmark, in Appendix A.3.*

*The sub-Gaussian bound of Corollary 1.21 requires the knowledge of the sub-Gaussian parameter $R$. In fact, no time-uniform confidence sequences valid over* all *sub-Gaussian distributions may exhibit a similar $\mathcal{O}(\sqrt{\log(t/\delta)/t})$ concentration rate, as stated in Corollary 1.36 below. With this in mind, it is however tempting to replace $R$ with a data-dependent estimator, e.g. the sample standard deviation $\widehat{\sigma}_t = \sqrt{1/t\sum_{s=1}^t (Y_s - \widehat{\mu}_t)^2}$. Recently, Waudby-Smith et al. (2021) showed that this procedure induces an* asymptotic *time-uniform confidence sequence for the family of square-integrable distributions $\bigcup_{M\in\mathbb{R}_+^\star} \mathcal{F}_{M,1}$,*

$$\left(\widehat{\Theta}_{t,\alpha}^\delta\right)_{t\in\mathbb{N}} = \left(\left[\widehat{\mu}_t \pm \widehat{\sigma}_t\sqrt{\frac{2}{t}\left(1+\frac{\alpha}{t}\right)\log\left(\frac{2}{\delta}\sqrt{1+\frac{t}{\alpha}}\right)}\right]\right)_{t\in\mathbb{N}}, \tag{1.41}$$

*in the sense that there exists a true time-uniform confidence sequence $(\widetilde{\Theta}_t^\delta)_{t\in\mathbb{N}}$ (typically unknown) such that almost surely $\max\widetilde{\Theta}_t^\delta \sim \max\widehat{\Theta}_t^\delta$ and $\min\widetilde{\Theta}_t^\delta \sim \min\widehat{\Theta}_t^\delta$ when $t\to+\infty$ (although, much like the central limit theorem for fixed sample estimation, this does not provide information on the nonasymptotic regime). In Chapter 4, we consider a subset of the family of sub-Gaussian distributions satisfying a high order MGF condition that allows to construct a (nonasymptotic) confidence sequence for the sub-Gaussian parameter $R$.*

**Extension to multivariate distributions.** We have so far stated our results for the concentration of real-valued distributions. The same mixture construction may actually be used in the multivariate setting. The proposition below is taken from Abbasi-Yadkori et al. (2011, Theorem 1) and Lattimore and Szepesvári (2020, Theorem 20.4).

**Proposition 1.26** (Multivariate time-uniform sub-Gaussian concentration with the method of mixtures). *Let $d \in \mathbb{N}$, $R \in \mathbb{R}_+^\star$. We consider two stochastic processes, $(X_t)_{t\in\mathbb{N}}$ in $\mathbb{R}^d$ and $(\eta_t)_{t\in\mathbb{N}}$ in $\mathbb{R}$, and an adapted filtration $(\mathcal{G}_t)_{t\in\mathbb{N}}$ such that $(\eta_t)_{t\in\mathbb{N}}$ is conditionally $R$-sub-Gaussian, i.e. $\log \mathbb{E}[e^{\lambda\eta_{t+1}} \mid \mathcal{G}_t] \leqslant \lambda^2 R^2/2$ for all $\lambda \in \mathbb{R}$ and $t \in \mathbb{N}$. For any $\alpha \in \mathbb{R}_+^\star$, we consider the $\mathbb{R}^d$-valued process $(S_t)_{t\in\mathbb{N}}$ and the $\mathcal{S}_d^{++}(\mathbb{R})$-valued process $(V_t^\alpha)_{t\in\mathbb{N}}$ (positive definite matrices of size $d$) defined as*

$$(S_t)_{t\in\mathbb{N}} = \left(\sum_{s=1}^{t-1} \eta_s X_s\right)_{t\in\mathbb{N}} \quad \text{and} \quad (V_t^\alpha)_{t\in\mathbb{R}} = \left(\sum_{s=1}^{t-1} X_s X_s^\top + \alpha I_d\right)_{t\in\mathbb{R}} . \tag{1.42}$$

*Then for any $\delta \in (0,1)$ we have the following inequality (also valid for adapted random times):*

$$\mathbb{P}\left(\exists t \in \mathbb{N},\ \|S_t\|^2_{(V_t^\alpha)^{-1}} \geqslant R^2 \left(2\log\frac{1}{\delta} + \log\frac{\det V_t^\alpha}{\alpha^d}\right)\right) \leqslant \delta\,. \tag{1.43}$$

☞ *A typical use case of Proposition 1.26 is to construct time-uniform confidence sequences for $\theta^\star \in \mathbb{R}^d$ in the linear bandit setting $Y_t = \langle \theta^\star, X_t\rangle + \eta_t$, where $(\eta_t)_{t\in\mathbb{N}}$ is a centred noise process. In Chapter 5, we derive a similar multivariate time-uniform concentration bound with the method of mixtures for linear bandits with risk measures.*

**Optimal time-uniform concentration rate.** We have seen in various instances that the method of mixtures produces time-uniform confidence sequences of decreasing sizes $\mathcal{O}(R\sqrt{\log(t)/t})$ when the sample size $t$ goes to $+\infty$ and $R^2$ is some measure of variance (such as the sub-Gaussian parameter). A natural question is whether this rate can be improved. The answer is affirmative, which is a direct consequence of the celebrated law of iterated logarithms (see e.g. Darling and Robbins (1967)). For completeness, we provide a proof of a weak version, that also sheds light on the pivotal role of the exponential supermartingale.

**Proposition 1.28** (Law of iterated logarithms). *Fix $R \in \mathbb{R}_+^\star$. If (i) $\mathcal{T} = \mathbb{N}$ and $(S_t)_{t\in\mathcal{T}} = (\sum_{s=1}^t Y_s - \mu)_{t\in\mathcal{T}}$ where $(Y_t)_{t\in\mathcal{T}}$ is an i.i.d. sequence of random variables drawn from $\nu \in \mathcal{F}_{\mathcal{G},R}$ with expectation $\mu \in \mathbb{R}$, or (ii) $\mathcal{T} = \mathbb{R}_+$ and $(S_t)_{t\in\mathcal{T}}$ satisfies the SDE $dS_t = R dW_t$, then*

$$\mathbb{P}\left(\limsup_{t\to+\infty} \frac{S_t}{R\sqrt{2t\log\log t}} \leqslant 1,\ \liminf_{t\to+\infty} \frac{S_t}{R\sqrt{2t\log\log t}} \geqslant -1\right) = 1\,. \tag{1.44}$$

In simpler terms, for many centred Gaussian or $R$-sub-Gaussian processes $(S_t)_{t\in\mathcal{T}}$, almost surely, it occurs infinitely often that $S_t \leqslant \mathcal{O}(R\sqrt{t\log\log t})$, which imposes a lower bound $\Omega(R\sqrt{\log\log(t)/t})$ on the width of any time-uniform confidence sequences. Of note, in the Brownian motion case or if the variance of $(Y_t)_{t\in\mathbb{N}}$ is exactly $R$, the $\limsup$ and $\liminf$ are exactly $1$ and $-1$ respectively with probability $1$.

In particular, the method of mixtures appears to produce suboptimal confidence sequences as their concentration rate is $\mathcal{O}(\sqrt{\log(t)/t})$ (this is a universal feature of the method, even with other controls of the cumulant generating function by a function $\psi$, see Howard et al. (2021, Proposition 2)). Alternative constructions such as *time peeling* or discrete mixtures Howard et al. (2020, 2021) are able to achieve this asymptotic concentration rate. However, they typically do so at the cost of higher leading constant and wider confidence sets for small samples.

## Catching the optimal rate with mixtures (💡)

We now show that it is possible to force a $\mathcal{O}(\sqrt{\log\log(t)/t})$ concentration rate with the method of mixtures by combining it with a peeling argument, not on the process $(S_t)_{t\in\mathbb{N}}$ itself but on the mixing parameter $\alpha$. To the best of our knowledge, this result is new and offers a complementary view on mixing versus peeling. More precisely, we have seen in Lemma 1.23 that for a fixed time $t_0 \in \mathbb{N}$, the choice $\alpha \propto t_0$ provides the minimal width for the confidence sequence at time $t_0$. However, a time-dependent $\alpha$ is not supported by the theory as it would break the supermartingale property; in addition, it would make the time-uniform bound of Corollary 1.21 asymptotically bounded by $\mathcal{O}(1/\sqrt{t})$, in contradiction with Proposition 1.28. Nevertheless, we show that it is possible to make $\alpha$ partially time-dependent, i.e. piecewise constant over epochs of geometrically increasing length, resulting in an order-optimal concentration rate. We defer the proofs of the following two results to Appendix A.3.

**Proposition 1.29** (Asymptotically order-optimal time-uniform concentration with mixing parameter time peeling). *Let $(S_t)_{t\in\mathbb{N}}$ be a real-valued stochastic process adapted to a filtration $(\mathcal{G}_t)_{t\in\mathbb{N}}$ and $(F_t^\alpha)_{t\in\mathbb{N},\alpha>0}$ a sequence of mappings such that for each $\alpha \in \mathbb{R}_+^\star$,*

*(i) for each $t \in \mathbb{N}$, $x \mapsto F_t^\alpha(x)$ is nondecreasing,*

*(ii) for each $x \in \mathbb{R}$, $t \mapsto F_t^\alpha(x)$ is nonincreasing,*

*(iii) $(M_t^\alpha)_{t\in\mathbb{N}} = (F_t^\alpha(S_t))_{t\in\mathbb{N}}$ defines a nonnegative $(\mathcal{G}_t)_{t\in\mathbb{N}}$-supermartingale with $\mathbb{E}\left[M_0^\alpha\right] \leqslant 1$.*

*Fix $\Delta > 1$, $\eta > 0$, $\delta \in (0,1)$ and a sequence $(\alpha_n)_{n\in\mathbb{N}}$. For $t \in \mathbb{N}$, we define*

$$n_t = \left\lceil \frac{\log t}{\log \Delta} \right\rceil , \quad t_n = \lfloor \Delta^n \rfloor \quad \text{and} \quad u_t = \left(F_{t_n}^{\alpha_{n_t-1}}\right)^{-1} \left( \frac{\zeta(\eta) n_t^\eta}{\delta} \right) , \tag{1.45}$$

*where $\zeta(\eta) = \sum_{n=1}^{+\infty} n^{-\eta}$. Then for any $(\mathcal{G}_t)_{t\in\mathbb{N}}$-stopping time $\tau$ in $\mathbb{N}$, we have*

$$\mathbb{P}\left(S_\tau \geqslant u_\tau\right) \leqslant \delta\,. \tag{1.46}$$

In the sub-Gaussian case, this result translates to a mixture bound achieving the order-optimal asymptotic rate. To the best of our knowledge, this is the first time-uniform bound to achieve this while resorting only to explicit mixtures. In particular, it differs (albeit complementary) from the so-called discrete mixture bound introduced in Howard et al. (2021, Theorem 2) (see also Appendix A.3), which implements a peeling-like mixture, leading to a numerical optimisation problem rather than a closed-form inequality, in contrast to the following corollary.

**Corollary 1.30** (Asymptotically order-optimal time-uniform sub-Gaussian concentration with mixing parameter time peeling). *Let $R \in \mathbb{R}_+^\star$ and $(Y_t)_{t\in\mathbb{N}}$ be a sequence of i.i.d. random variables drawn from $\nu \in \mathcal{F}_{\mathcal{G},R}$ with expectation $\mu \in \mathbb{R}$. For $\delta \in (0,1)$, $\gamma_\delta = -1/(1 + W_{-1}(-\delta^2/e))$ and $t \in \mathbb{N}$, we consider*

$$\widehat{\Theta}_{t,\Delta,\eta}^\delta = \left( \left[ \widehat{\mu}_t \pm R \sqrt{\frac{2}{t^2}(1+\gamma_\delta) \lfloor \Delta^{\left\lceil \frac{\log t}{\log \Delta} \right\rceil} \rfloor \log\left( \frac{\zeta(\eta)}{\delta} \sqrt{1 + \frac{1}{\gamma_\delta}} \left\lceil \frac{\log t}{\log \Delta} \right\rceil^\eta \right)} \right] \right)_{t\in\mathbb{N}} . \tag{1.47}$$

*Then $(\widehat{\Theta}_{t,\Delta,\eta}^\delta)_{t\in\mathbb{N}}$ is a **time-uniform confidence sequence** at level $\delta$ for $\mu$ which satisfies the asymptotic relation $|\widehat{\Theta}_{t,\Delta,\eta}^\delta| = \mathcal{O}\left(R\sqrt{\log(\log(t)/\delta)/t}\right)$ when $t \to +\infty$.*

**Remark 1.31**. *The above result still holds if $(t_n)_{n\in\mathbb{N}}$ is replaced by any strictly increasing sequence of integers such that $t_0 = 0$ and $\lim_{n\to+\infty} t_n = +\infty$ and if $(\zeta(\eta)n^\eta)_{n\in\mathbb{N}}$ is replaced by any positive sequence $(c_n)_{n\in\mathbb{N}}$ such that $\sum_{n=1}^{\infty}\frac{1}{c_n} \leqslant 1$. The choice of geometric sequences for both $(t_n)_{n\in\mathbb{N}}$ and $(c_n)_{n\in\mathbb{N}}$ provides examples of such sequences and is standard in the literature on time peeling concentration (Garivier, 2013; Maillard, 2019b).*

In Figure 1.3, we compare these asymptotically optimal mixture peeling bounds with the standard sub-Gaussian mixture bound of Corollary 1.21 (on the right-hand figure, mixture peeling bounds are smoothed by upper bounding the various ceils and floors in their definition). We conclude that the improvement due to the $\mathcal{O}(\sqrt{\log\log(t)/t})$ concentration rate is only asymptotic in nature and leads to wider bounds in practice, even for fairly large samples. The take-home message is that virtually any mixture bound can be turned into an asymptotically optimal bound if theory calls for it, but doing so is often detrimental to the actual performances. Therefore, for the practitioner, we recommend using simple mixture bounds.

☞ *Mixture bounds in various settings beyond the sub-Gaussian case is the focus of Part II.*

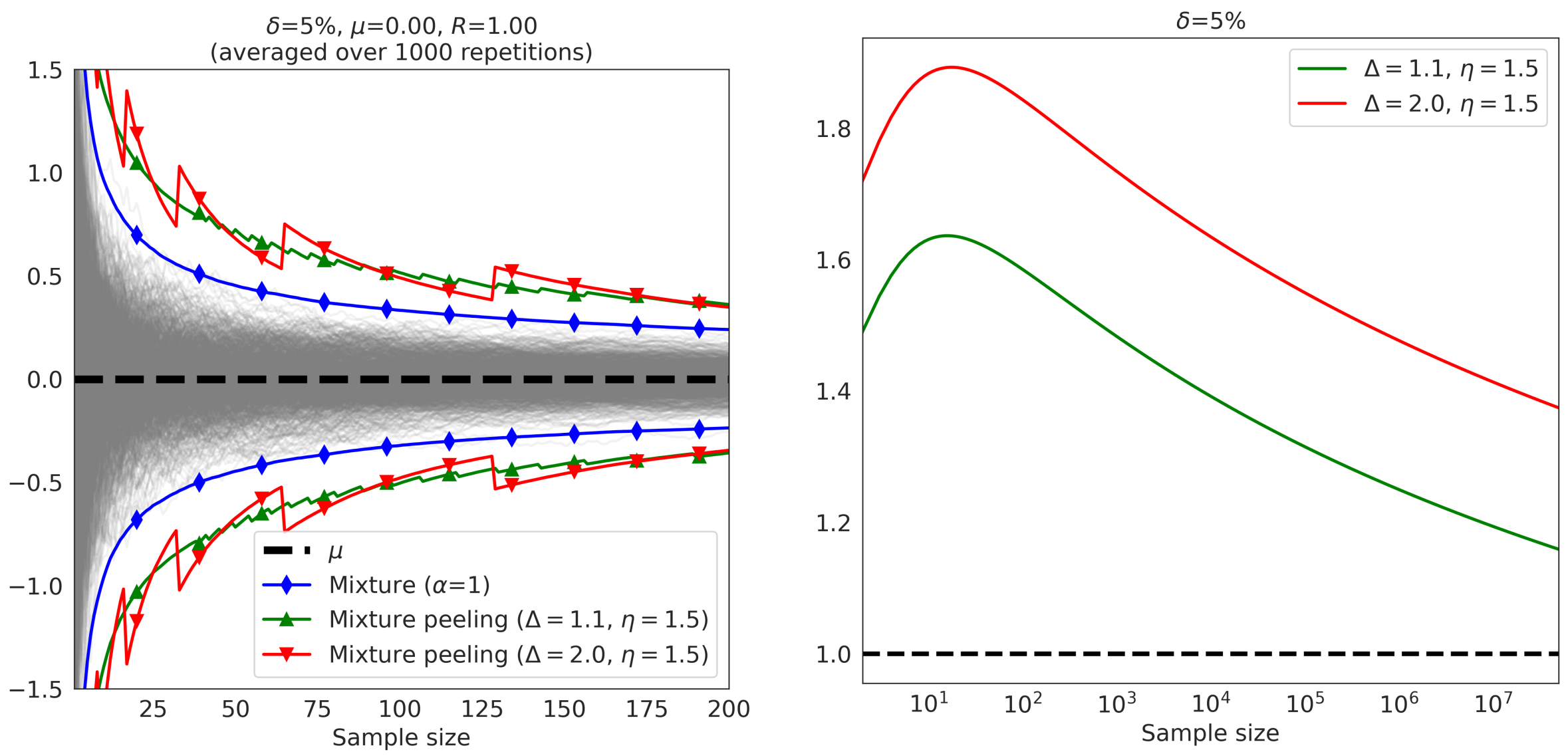

**Figure 1.3** – Left: Comparison of confidence envelopes around the mean for $\mathcal{N}(0,1)$, as a function of the sample size $t$, over 1000 independent replicates. Grey lines are trajectories of empirical means $\widehat{\mu}_t$. Right: ratio of upper mixture peeling bounds of Corollary 1.30 to the classical sub-Gaussian mixture bounds (greater than 1 means the classical mixture bound is tighter).

## Generic UCB algorithm for multiarmed bandits (💡)

We now show a direct application of time-uniform concentration to the design of deterministic $K$-armed bandit algorithms. We already introduced the upper confidence bound (UCB) policy in Section 1.3. In the original article (Auer et al., 2002), UCB was studied for bandit models $(\bigotimes_{k\in[K]} \nu_k, \bigotimes_{k\in[K]} \mathcal{F}_k)$ with $\mathcal{F}_k = \mathcal{F}_{\mathcal{N},\sigma}$ (Gaussian) or $\mathcal{F}_k = \mathcal{F}_{[\underline{B},\overline{B}]}$ (bounded) using sub-Gaussian concentration bounds. Here, we extend the analysis to *any* bandit models, provided there exist a confidence sequence for each arm. Although folk knowledge, this is, as far as we know, the first occurrence of such generic results. Precisely, we define for arm $k \in [K]$ and $\delta \in (0,1)$ the generalised UCB index $\mathrm{UCB}_k^\delta \colon \mathbb{R}^{(\mathbb{N})} \to \mathbb{R}$ such that if $\mathbb{Y}_t^k$ is the reward history of arm $k$ at time $t$, $\mathrm{UCB}_k^\delta(\mathbb{Y}_t^k)$ is the upper bound of the confidence sequence at level $\delta$ for $\mu_k = \mathbb{E}_{Y\sim\nu_k}[Y]$. To emphasise the dependency on the size $n = N_t^k$ of $\mathbb{Y}_t^k$, we also write $\mathrm{UCB}_{k,n}^\delta = \mathrm{UCB}_k^\delta(\mathbb{Y}_t^k)$; e.g. in the $R$-sub-Gaussian case with mixing parameter $\alpha \in \mathbb{R}_+^\star$, we have

$$\mathrm{UCB}_k^\delta(\mathbb{Y}_t^k) = R\sqrt{\frac{2}{N_t^k}\left(1+\frac{\alpha}{N_t^k}\right)\log\frac{2\sqrt{1+\frac{N_t^k}{\alpha}}}{\delta}}\,. \tag{1.48}$$

The main result of this section is the following theorem, that bounds the number of pulls to a suboptimal arm by the sample size required to statistically separate it from the optimal arm. We defer its proof, as well as those of the subsequent corollaries, to Appendix A.3.

**Algorithm 1** Generic UCB

**Input:** $K$ arms and upper confidence mapping $\mathbb{Y} \in (\mathbb{R})^{(\mathbb{N})} \mapsto \mathrm{UCB}_k^\delta(\mathbb{Y})$.
**Initialisation:** $\forall k \in [K]$, $\mathbb{Y}^k = \{\}$, $N^k = 0$.
**while** *continue* **do**
  $\pi \leftarrow \mathrm{argmax}_{k\in[K]} \mathrm{UCB}_k^\delta(\mathbb{Y}^k)$ ; ▷ `Best plausible arm`
  Observe $Y^\pi_{N^\pi}$, $\mathbb{Y}^\pi \leftarrow \mathbb{Y}^\pi \cup \{Y^\pi_{N^\pi}\}$, $N^\pi \leftarrow N^\pi + 1$ ; ▷ `Update`

**Theorem 1.33** (Generic UCB pseudo regret analysis). *Let $\delta \in (0,1)$ and $(\widehat{\Theta}^\delta_{k,n})_{n\in\mathbb{N}}$ be a time-uniform confidence sequence at level $\delta$ for $\rho_k = \rho(\nu_k)$ for each $k \in [K]$, where $\rho$ is a risk measure. We recall that $\Delta_k = \rho^\star - \rho_k$ and define*

$$\tau_k^\delta = \inf\{n \in \mathbb{N},\ \left|\widehat{\Theta}^\delta_{k,n}\right| < \Delta_k\}\,. \tag{1.49}$$

*Fix $T \in \mathbb{N}$. Then we have*

$$\mathbb{P}\left(\forall k \in [K],\ N_T^k \leqslant \tau_k^{\frac{\delta}{2K}}\right) \geqslant 1-\delta \quad \text{and} \quad \mathbb{E}_{\pi,\nu_\pi^{\otimes T}}[N_T^k] \leqslant \mathbb{E}_{\pi,\nu_\pi^{\otimes T}}[\tau_k^\delta] + 2\delta T\,. \tag{1.50}$$

*In particular, if $\delta = 1/T$, we have*

$$\mathbb{E}_{\pi,\nu_\pi^{\otimes T}}[N_T^k] \leqslant \mathbb{E}_{\pi,\nu_\pi^{\otimes T}}[\tau_k^{\frac{1}{T}}] + 2\,. \tag{1.51}$$

In particular, Algorithm 1 with upper confidence mapping $\mathrm{UCB}^\delta_{k,n} = \max \widehat{\Theta}^\delta_{k,n}$ is an instance of *no regret* algorithm: after playing each suboptimal arm $k \in [K] \setminus \{k^\star\}$ exactly $\tau_k^{\delta/(2K)}$ times, the generic UCB policy only selects the optimal arm $k^\star$ and stops accumulating pseudo regret with probability $1-\delta$. This can be translated to an upper bound on the expected number of pulls to arm $k$ by setting $\delta = 1/T$. The take-home message of this analysis is that the regret of Algorithm 1 is driven by how fast the confidence sequence shrinks for each arm, which is typically folk knowledge but, as far as we are aware, was never written in this form.

We now instantiate Theorem 1.33 to various statistical models under the expectation risk measure, for which time-uniform confidence sequences are known.

**Corollary 1.34** (Sub-Gaussian UCB). *Let $R \in \mathbb{R}_+^\star$ and assume that $\mathcal{F}_k = \mathcal{F}_{\mathcal{G},R}$. For $T \to +\infty$, Algorithm 1 with the time-uniform UCB of Corollary 1.21 satisfies*

$$\mathbb{E}_{\pi,\nu_\pi^{\otimes T}}\left[N_T^k\right] \leqslant \frac{2R^2}{\Delta_k^2}\log\left(\frac{4RT}{\Delta_k}\right) + \frac{4R^2}{\Delta_k^2}\log\log\left(\frac{4RT}{\Delta_k}\right) + o\left(\frac{R^2}{\Delta_k^2}\log\log T\right). \tag{1.52}$$

This matches (up to constants) the classical pseudo regret bound of UCB (see e.g. Lattimore and Szepesvári (2020, Theorem 7.1)). In particular, it proves that Algorithm 1 is asymptotically optimal (in the sense that it matches the lower bound of Theorem 1.6) in the Gaussian case ($\mathcal{K}_{\inf}^{\mathcal{F}^{\mathcal{N},R}}(\nu_k;\mu^\star) = \Delta_k^2/(2R^2)$) and order-optimal in the sub-Gaussian case ($\mathcal{O}(\log T)$ pseudo regret). Note that pseudo regret optimality is achieved even though the concentration rate of the confidence sequence is suboptimal ($\mathcal{O}(\sqrt{\log(t)/t}$ instead of $\mathcal{O}(\sqrt{\log\log(t)/t})$). In fact, we show the iterated logarithm rate only impacts the higher order terms of the regret bound.

**Corollary 1.35** (Iterated logarithm UCB). *For $k \in [K]$, let $(\widehat{\Theta}_{k,n}^\delta)_{n\in\mathbb{N}}$ a time-uniform confidence sequence at level $\delta$ for $\mu_k$ such that $|\widehat{\Theta}_{k,n}^\delta| \leqslant C\sqrt{\log(\log(n)/\delta)/n}$ for $n$ large enough and $C \in \mathbb{R}_+^\star$. For $T \to +\infty$, Algorithm 1 with these UCB satisfies*

$$\mathbb{E}_{\pi,\nu_\pi^{\otimes T}}\left[N_T^k\right] \leqslant \frac{C^2}{\Delta_k^2}\log T + \frac{C^2}{\Delta_k^2}\log\log\left(\frac{C^2}{\Delta_k^2}\log T\right) + o(\log\log\log T)\,. \tag{1.53}$$

As discussed in Section 1.4, the asymptotic order-optimal of confidence sequences achieving the iterated logarithm rate often comes at the cost of higher leading constant $C$. However, in the context of UCB algorithms, the constant $C$ directly impacts the first order term, while the additional $\log$ term only affects the higher order terms of the pseudo regret bound, which makes for another argument in favour of the method of mixtures.

Interestingly, the pseudo regret lower bound for stochastic bandits sheds light on the minimum time-uniform concentration rate that is achievable for a family of distributions $\mathcal{F}$.

**Corollary 1.36** (Obstruction to $\mathcal{O}(\sqrt{\log(n)/n})$ concentration rate). *Let $\mathcal{F} \subseteq \mathcal{M}_1^+(\mathbb{R})$ and $\rho \colon \mathcal{F} \to \mathbb{R}$ a risk measure. Assume that for all $\nu \in \mathcal{F}$ and $\rho^\star > \rho(\nu)$, we have $\mathcal{K}_{\inf}^{\mathcal{F},\rho}(\nu; \rho^\star) = 0$, and that $(\widehat{\Theta}_n^\delta)_{n\in\mathbb{N}}$ is a time-uniform confidence sequence at level $\delta \in (0,1)$ for $\rho$ and $\mathcal{F}$. Then for any $\delta \in (0,1)$ and $\alpha, \beta \in \mathbb{R}_+^\star$ such that $\alpha \geqslant \beta$, almost surely we have $|\widehat{\Theta}_n^\delta| n^\alpha \log(n/\delta)^{-\beta} \to +\infty$.*

In particular, following Lemma 1.13, it is hopeless to seek fast shrinking time-uniform confidence sequences that would hold for all probability distributions, or even distributions with finite $1+\varepsilon$-th moment (with no bound on said moment), light tailed distributions (with no explicit control of their MGF), sub-Gaussian distributions (with no bound on the sub-Gaussian parameter) or bounded distributions (with no prior knowledge of the support).

Going beyond the sub-Gaussian case, a consequence of Theorem 1.33 is that we may derive an efficient algorithm for the bandit model $(\bigotimes_{k\in[K]} \nu_k, \bigotimes_{k\in[K]} \mathcal{F}_k)$ as soon as we can build time-uniform confidence sequences that are valid for the families $\mathcal{F}_k$ with $k \in [K]$. We illustrate this principle on two more examples, starting with the family of heavy tailed distributions subject to a control of the $1+\varepsilon$-th moment.

**Corollary 1.37** (Heavy tailed UCB). *Let $M, \varepsilon \in \mathbb{R}_+^\star$ and assume that $\mathcal{F}_k = \mathcal{F}_{M,\varepsilon}$, i.e. $\mathbb{E}_{Y\sim\nu_k}[|Y|^{1+\varepsilon}] \leqslant M$. Let $(\widehat{\Theta}_{k,n}^\delta)_{n\in\mathbb{N}}$ a time-uniform confidence sequence at level $\delta$ for $\mu_k$ such that $|\widehat{\Theta}_{k,n}^\delta| \leqslant CM^{1/(1+\varepsilon)}(\log(n/\delta)/n)^{\varepsilon/(1+\varepsilon)}$ for $n$ large enough and $C \in \mathbb{R}_+^\star$. For $T \to +\infty$, Algorithm 1 with these UCB satisfies*

$$\mathbb{E}_{\pi,\nu_\pi^{\otimes T}}\left[N_T^k\right] \leqslant M^{\frac{1}{\varepsilon}} \left(\frac{C}{\Delta_k}\right)^{1+\frac{1}{\varepsilon}} \log\left(M^{\frac{1}{\varepsilon}} \left(\frac{C}{\Delta_k}\right)^{1+\frac{1}{\varepsilon}} T\right) + M^{\frac{1}{\varepsilon}} \left(\frac{C}{\Delta_k}\right)^{1+\frac{1}{\varepsilon}} \log\log\left(M^{\frac{1}{\varepsilon}} \left(\frac{C}{\Delta_k}\right)^{1+\frac{1}{\varepsilon}} T\right) + o\left(M^{\frac{1}{\varepsilon}} \left(\frac{C}{\Delta_k}\right)^{1+\frac{1}{\varepsilon}} \log\log T\right). \tag{1.54}$$

Such confidence sequences satisfying the concentration rate $\mathcal{O}(M^{1/(1+\varepsilon)}(\log(n/\delta)/n))$ can be derived from Bubeck et al. (2013, Lemma 1, Lemma 2) (using the truncated mean and median of means estimators respectively), and Catoni (2012, Proposition 2.4) (using Catoni's M-estimator) — those provide fixed sample confidence sets, that can be turned into time-uniform confidene sequences up to horizon $T$ by using $\delta/T$ instead of $\delta$ (union bound). Recently, Wang and Ramdas (2023a) provided a Catoni-style confidence sequence (based on the predictable weight construction detailed in Appendix A.3) that exhibits this concentration rate with high probability. Of note, the regret guarantee of Corollary 1.37 matches that of Bubeck et al. (2013).

Finally, we illustrate the flexibility of this generic UCB analysis by applying it to a recent heavy tailed bandit setting introduced by Basu et al. (2022), namely *bandits corrupted by Nature*. In this setting, when pulling arm measure $\nu_k \in \mathcal{F}_k$, the agent collects a reward drawn from a different distribution $\nu_k^\eta$ such that $\delta_{\mathrm{TV}}(\nu_k, \nu_k^\eta) \leqslant \eta \in [0,1)$ (typically $\nu_k^\eta = (1-\eta)\nu_k + \eta\nu_k'$ for some $\nu_k' \in \mathcal{M}_1^+(\mathbb{R})$, i.e. an outlier is generated with probability $\eta$). In the next corollary, we consider a corrupted bandit measure $\boldsymbol{\nu}^{\boldsymbol{\eta}} = \bigotimes_{k\in[K]} \nu_k^\eta$ and denote by $\mathbb{P}_{\pi,\boldsymbol{\nu}_\pi^{\eta\otimes T}}$ the probability induced by a bandit policy $\pi$ for $T$ rounds when observing rewards drawn from $\boldsymbol{\nu}^{\boldsymbol{\eta}}$.

**Corollary 1.38** (Heavy tailed UCB corrupted by Nature). *Let $\kappa, \varepsilon \in \mathbb{R}_+^\star$ and assume that $\mathcal{F}_k = \mathcal{F}_{\kappa,\varepsilon}^{centred}$, i.e. $\mathbb{E}_{Y\sim\nu_k}[|Y-\mu_k|^{1+\varepsilon}] \leqslant \kappa$. For $\delta \in (0,1)$ and $\eta \in [0,1)$, let $(\widehat{\Theta}_{k,n}^{\eta,\delta})_{n\in\mathbb{N}}$ a nonincreasing sequence of subsets of $\mathbb{R}$ such that*

*(i)* $\mathbb{P}_{\pi,\boldsymbol{\nu}_\pi^{\eta\otimes T}}(\forall n \in \mathbb{N},\ \mu_k \in \widehat{\Theta}_{k,n}^{\eta,\delta}) \geqslant 1-\delta$,

*(ii)* $\mathbb{P}_{\pi,\boldsymbol{\nu}_\pi^{\eta\otimes T}}(|\widehat{\Theta}_{k,n}^{\eta,\delta}| \leqslant C\kappa^{1/(1+\varepsilon)}\eta^{\varepsilon/(1+\varepsilon)}) \geqslant 1-\delta'$ *for any* $n \geqslant \log(4/(\delta\delta'))/\eta$, *where* $C \in \mathbb{R}_+^\star$ *and* $\delta' \in (0,1)$.

*Assume that $\Delta_k > C\kappa^{1/(1+\varepsilon)}\eta^{\varepsilon/(1+\varepsilon)}$. For $T \to +\infty$, Algorithm 1 with these UCB satisfies*

$$\mathbb{E}_{\pi,\boldsymbol{\nu}_\pi^{\eta\otimes T}}\left[N_T^k\right] \leqslant \frac{2}{\eta}\log(2T) + 5\,. \tag{1.55}$$

First, note that the assumption that the confidence sequences are nonincreasing is mild: not only are they expected to shrink with increasing sample sizes, but as we have seen in Section 1.4, time-uniform confidence sequences may be replaced at no further cost by their running intersections, which are nonincreasing by definition. Furthermore, note that $(\widehat{\Theta}_{k,n}^{\eta,\delta})_{n\in\mathbb{N}}$ is a *robust* time-uniform confidence sequence, in the sense that it captures the expectation $\mu_k$ of the true reward distribution $\nu_k$ from observations of the corrupted distributions $\nu_k^\eta$ (indicated by the use of the probability measure $\mathbb{P}_{\pi,\boldsymbol{\nu}_\pi^{\eta\otimes T}}$). An example of such robust confidence sequences is given in Wang and Ramdas (2023b) (the two conditions (i) and (ii) in the above corollary correspond to Theorem 4 and Theorem 5 in this reference, provided that the corruption parameter is small enough, i.e. $\eta \leqslant \varepsilon/(7(1+\varepsilon))$). Note that this robustness comes at the cost that the width of the confidence sequence may not shrink to zero; however, in this generic UCB analyis, we only need $|\widehat{\Theta}_{k,n}^{\eta,\delta}|$ to become less than $\Delta_k$. In other words, if the gaps are large enough to be "detectable" by the robust confidence sequences (as we assume in Corollary 1.38), the pseudo regret of Algorithm 1 scales as $\mathcal{O}(\log T)$. Of note, Basu et al. (2022) identified in the finite variance setting ($\varepsilon = 1$) a similar phase transition, with two separate regimes: easily distinguishable (low $\kappa$, high $\Delta_k$) versus hardly distinguishable (high $\kappa$, low $\Delta_k$), with detailed logarithmic guarantees in both cases.

**Remark 1.39** (Examples of *corruption by Nature*). *This bandit setting typically represents situations where rewards are subject to human errors. Basu et al. (2022) motivated this modelling approach with an application in agriculture (manual counting of fallen mites after applying a treatment). In digital health, potential use cases include remote consultations, self-assessments by patients, transcripts of hanwritten medical files, etc. In all these scenarios, rewards are extracted from noisy, potentially faulty, informations that are* corrupted *in a nonadversarial fashion.*

# Chapter 2

# Literature review and outline

*Mais, se disait Durtal, du moment que l'on patauge dans l'inconnu, [...] on peut aussi facilement admettre le « Credo quia absurdum » de saint Augustin et se répéter, avec Tertullien, que si le surnaturel était compréhensible, il ne serait pas le surnaturel et que c'est justement parce qu'il outrepasse les facultés de l'homme qu'il est divin. Ah ! et puis zut, à la fin du compte !*

— Joris-Karl Huysmans, *Là-bas*

In this second introductory chapter, we review existing algorithms for stochastic bandits and concentration bounds, highlighting their scope of applicability and limitations. We argue that the current state of the art in these fields is somewhat ill-adapted for critical applications under constraints of small samples, risk aversion and nonparametric model specifications, such as healthcare. Motivated by this opportunity to make progress towards filling the long-standing gap between theory and practice, we present our contributions, which we develop in the rest of this thesis.

**Contents**

## 2.1 How do others play? Overview of bandit algorithms

In this section, we review classical algorithms that implement bandit policies with sublinear regret under various statistical models for the rewards. Any efficient learning strategy in a setting with partial information must face the classical dilemma between **exploration and** **exploitation**; in bandits specifically, it needs to obtain enough information from arms that have been scarcely sampled (exploration) to avoid missing the optimal arm, while also focusing on arms that have already performed well in order to accumulate high rewards (exploitation). Many algorithms have been proposed for stochastic bandits (see Lattimore and Szepesvári (2020) for a recent textbook survey), and we propose here a (nonexhaustive) list of such methods (summarised in Table 2.1), with a particular focus on recent advances in this field.

We recall from Chapter 1 that the quality of a multiarmed bandit policy $\pi$ played on a bandit model $(\boldsymbol{\nu}, \mathcal{F})$ is measured by its pseudo regret $\mathcal{R}_T^{\pi,\boldsymbol{\nu}}$, which grows at least logarithmically in $T$, with optimal rate given by the extremal Kullback-Leibler operator $\mathcal{K}_{\text{inf}}^{\mathcal{F}}$ (Theorem 1.6) (unless stated otherwise, algorithms presented here consider the expectation risk measure). It is crucial to consider both the family $\mathcal{F}$ and the policy $\pi$ simultaneously since the latter is often specialised to operate under the specific setting defined by the former. In particular, we recall that if a family $\mathcal{F}$ is "too large", the $\mathcal{K}_{\text{inf}}^{\mathcal{F}}$ operator may vanish, which meaning that no logarithmic instance-dependent pseudo regret is achievable (Lemma 1.13).

A generic template to implement bandit policies is the *index policy* algorithm, which (i) computes for each arm an index equal to a given (possibly randomised) function of its reward history and (ii) consequently plays the arms with the highest such index (Algorithm 2).

**Algorithm 2** Index policy

**Input:** $K$ arms, index function $I\colon \mathbb{N} \times \mathbb{R}^{(\mathbb{N})} \to \mathcal{M}_1^+(\mathbb{R})$.
**Initialisation:** $t = 1, \forall k \in [K], \mathbb{Y}^k = \{\}, N^k = 0$.
**while** *continue* **do**

  $\pi \leftarrow \text{argmax}_{k\in[K]} I_t(\mathbb{Y}^k)$ ; ▷ `Greedy with respect to the index function`

  Observe $Y_{N^\pi}^\pi$, $\mathbb{Y}^\pi \leftarrow \mathbb{Y}^\pi \cup \{Y_{N^\pi}^\pi\}, N^\pi \leftarrow N^\pi + 1, t \leftarrow t+1$ ; ▷ `Update`

**Remark 2.1** (Initialisation and continuation). *In practice, it is common to start bandit algorithms with an arbitrary policy for a few rounds, e.g. play each of the $K$ arms once for $t = 1, \dots, K$. Moreover, most of our algorithms are* anytime, *i.e. they run without the knowledge of a time horizon $T$, we thus refer to the keyword* continue *to mean either $t \leqslant T$ or a user-specified stopping criterion.*

**Greedy policy.** A straightforward instance of such index policies is the *greedy* policy, for which $I_t(\mathbb{Y}_t^k) = \widehat{\mu}_t^k$ (empirical mean of rewards in $\mathbb{Y}_t^k$) for any $k \in [K]$ and $t \in \mathbb{N}$. This policy does *not* have sublinear regret guarantees. Indeed, imagine a simple two-armed uniform bandit model $\boldsymbol{\nu} = \mathcal{U}([1,2]) \otimes \mathcal{U}([0,4])$ with the expectation risk measure. Clearly, the second arm is optimal ($\mu_2 = 2$ vs. $\mu_1 = 3/2$). Let us assume that each arm is played once at initialisation, collecting reward $Y_1^1$ and $Y_1^2$ respectively. Under the event $\{Y_1^2 \in [0,1]\}$, which holds with probability $1/4$, the greedy policy will always play arm $1$ since $Y_t^1 \geqslant Y_1^2$ for all $t \geq 1$ almost surely, thus accumulating linear pseudo regret. This is typical case of failing to address the exploration-exploitation tradeoff: an efficient agent must keep the possibility of exploring seemingly suboptimal alternatives to compensate for the noisy bandit feedback.

**Deterministic index policies.** The simplest bandit policy to achieve sublinear regret is the *upper confidence bound* (UCB) policy (Auer et al., 2002), which computes $I_t(\mathbb{Y}_t^k) = \widehat{\mu}_t^k + R\sqrt{2\log(1/\delta_t)/N_t^k}$, where $R \in \mathbb{R}_+^\star$ and $(\delta_t)_{t\in\mathbb{N}}$ are chosen at initialisation. This is an instance of the *optimism in the face of uncertainty* principle: under an $R$-sub-Gaussian model, $I_t(\mathbb{Y}_t^k)$ corresponds to the upper bound of a $\delta_t$-confidence interval around $\mu_k$, i.e. the policy plays the arm that has a plausible chance to be the best, rather than the empirical best according to past observations. This policy was later extended to SPEF $\mathcal{F} = \{\nu_\theta, \theta \in \Theta\}$ (with identity feature function) with better regret guarantees ($\mathcal{K}_{\inf}$-optimal) by kl-UCB (Cappé et al., 2013) using $I_t(\mathbb{Y}_t^k) = \max\{\theta \in \Theta,\ N_t^k \mathcal{B}_{\mathcal{L}}(\widehat{\theta}_t^k; \theta) \leqslant f(t)\}$ where $\widehat{\theta}_t^k = (\nabla\mathcal{L})^{-1}(\widehat{\mu}_t^k)$ and $f$ is a slowly growing, nondecreasing function, corresponding to a concentration rate for the KL divergence. Of note, ISM-NORMAL (Cowan et al., 2017) circumvents the need for the prior knowledge of $R$ for Gaussian distributions by estimating the variance.

Other deterministic index policies were introduced for bounded distributions, such as UCB-V (Audibert et al., 2009) (using empirical Bernstein inequalities rather than sub-Gaussian concentration bounds) and empirical KL-UCB (Cappé et al., 2013) (exploiting the dual form of the extremal Kullback-Leibler operator and empirical likelihood), and for upper bounded light tailed distributions, IMED (Honda and Takemura, 2015), with a lower bound-inspired index $I_t(\mathbb{Y}_t^k) = -(N_t^k \mathcal{K}_{\inf}^{\mathcal{F}_{(-\infty,\overline{B}]}}(\widehat{\nu}_{\mathbb{Y}_t^k}; \widehat{\mu}_t^\star) + \log N_t^k)$, where $\widehat{\mu}_t^\star = \max_{k\in[K]} \widehat{\mu}_t^k$; these last two algorithms are $\mathcal{K}_{\inf}$-optimal. For heavy tailed distributions, we mention extensions of UCB (Bubeck et al., 2013) and KL-UCB (Agrawal et al., 2021) under a known moment condition.

Note that these algorithms implement bandit policies $\pi$ that are predictable with respect to the bandit filtration $(\mathcal{G}_t^\pi)_{t\in\mathbb{N}}$ (i.e. without introducing an independent filtration $(\mathcal{G}_t^\perp)_{t\in\mathbb{N}}$). In particular, $\pi_{t+1}$ is constant given $\mathcal{G}_t^\pi$, i.e. the same history of past decisions and rewards up to time $t$ would lead to a reproducible decision at time $t+1$.

**Randomised policies.** The policies described above ensure sufficient exploration by exploiting the concentration of deterministic estimators of the mean for each arm. An alternative way to induce exploration is to consider instead randomised estimators. The first example of such policies, going back as far as Thompson (1933), is *Thompson sampling* (TS), where $I_t(\mathbb{Y}_t^k) \sim p(\mu_k \mid \mathbb{Y}_t^k)$ is the posterior distribution of $\mu_k$ after observing $\mathbb{Y}_t^k$ (for a given prior $p(\mu_k)$). This policy was proven $\mathcal{K}_{\inf}$-optimal for SPEF in Korda et al. (2013) and logarithmic for bounded distributions using a reduction to the Bernoulli case called the *binarisation trick* (Agrawal and Goyal, 2012). Recently, it was further extended to multinomial and bounded distributions by nonparametric Thompson sampling (NPTS) (Riou and Honda, 2020), with $I_t(\mathbb{Y}_t^k) = \sum_{i=1}^{N_t^k} W_i^{k,t} Y_i^k + W_{N_t^k+1}^{k,t} \overline{B}$, where $(W_i^{k,t})_{i=1}^{N_t^k+1}$ follows the Dirichlet distribution $\mathcal{D}_{N_t^k+1}$, and recently to heavy tailed distributions (Baudry et al., 2023). Other algorithms using bootstrapping schemes have also been proposed (Osband and Roy, 2015; Kveton et al., 2019a,b; Wang et al., 2020): they share the idea of computing a noisy mean for empirical samples, enhanced by some exploration aid appropriately tuned to the statistical models they consider.

A different class of algorithms implement randomised policies that are not necessarily index policies but instead sample arms at random according to a carefully tuned distribution. Such algorithms are often designed primarily for adversarial bandits and draw inspiration from optimisation techniques in online learning, see for instance the family of EXP algorithms and its variants (Seldin et al., 2013; Seldin and Slivkins, 2014; Seldin and Lugosi, 2017). While robust to adversarial attacks (with formal minimax regret guarantees), these policies often exhibit poor performances in stochastic environments. Recently, Zimmert and Seldin (2021) combined online mirror descent with Tsallis entropy regularisation to obtain policies with *best-of-both worlds* guarantees for bounded rewards, i.e. $\mathcal{O}(\log T)$ and $\mathcal{O}(\sqrt{T})$ instance dependent and minimax pseudo regrets respectively. Furthermore, inspired by the early works of Maillard (2011); Honda and Takemura (2011), Bian and Jun (2022) used Boltzmann-like sampling probabilities in the context of sub-Gaussian rewards, an approach called *Maillard sampling*.

Finally, an altogether different approach based on subsampling was introduced in Baransi et al. (2014) and later extended in Chan (2020); Baudry et al. (2020); in particular, this last work established $\mathcal{K}_{\inf}$-optimality for all SPEF *without having to know which family precisely* (up to possible forced exploration). The common design principle to these algorithms, which do not implement index policies, it to simulate instead noisy estimators by artificially restricting the amount of information used by the seemingly optimal arm. However, these proofs heavily rely on tail properties of SPEF and are thus difficult to generalise beyond these parametric families.

Contrary to deterministic algorithms, these randomised policies $\pi$ are predictable with respect to the natural filtration induced by the bandit *and* the random variables used within the algorithms. For instance in NPTS, we let $\mathcal{G}_t^{\perp} = \sigma(W_i^{k,t},\ k \in [K], i \in \{1, \dots, N_t^k + 1\})$ for $t \in \mathbb{N}$ such that $\pi_{t+1} = \mathrm{argmax}_{k \in K} \sum_{i=1}^{N_t^k} W_i^{k,t} Y_i^k + W_{N_t^k+1}^{k,t} \overline{B}$ is measurable with respect to $\sigma(\mathcal{G}_t^{\pi} \cup \mathcal{G}_t^{\perp})$. In particular, given only $\mathcal{G}_t^{\pi}$, $\pi_{t+1}$ remains a nondeterministic random variable, i.e. the same history of decisions and rewards up to time $t$ could lead to a different decision at time $t+1$.

**Remark 2.2**. *Of note, several of these algorithms developed and rigorously analysed for multiarmed bandits have recently been extended to the framework of Markov decision processes for reinforcement learning, in particular UCB (Bourel et al., 2020) and IMED (Pesquerel and Maillard, 2022).*

**With so many existing algorithms, why look for more?** While many algorithms achieve optimal regret for bounded distributions with the sole knowledge of their supports (IMED, NPTS, etc.), the assumptions needed for algorithms working with unbounded distributions (e.g. SPEF, sub-Gaussian, subexponential) generally assume a known parametric model for the tails. While such assumptions entail convenient theoretical properties, they may be ill-suited for use by practitioners, for whom identifying the setting and parameters of interest may be tedious or infeasible. Still, we believe that actionable, theoretically grounded algorithms for generic unbounded distributions, or for bounded ones without precisely knowing the range, would make for a significant step towards a broader adoption of sequential decision-making methods. For instance, as seen in the foreword, a recommender system for health based on the body mass index (BMI) may consider this variable to be bounded. Indeed, while class III obesity is defined by a BMI exceeding 40 $\text{kg}/\text{m}^2$, extreme BMI have been recorded (up to 204 $\text{kg}/\text{m}^2$). An algorithm operating under the sole assumption that BMI lies in the range $[0, 204]$ would therefore likely be overly conservative; for instance, such an algorithm would be designed to be efficient even for an absolutely unrealistic BMI distribution $1/2\delta_0 + 1/2\delta_{204}$.

This raises the question of robustness with respect to model misspecification: on the one hand, deploying recommendation algorithms in such challenging real-world settings calls for efficient policies across a large spectrum of (possibly loosely defined) distributions; on the other hand, Lemma 1.13 suggests that optimality is fundamentally incompatible with such weak hypotheses. In this thesis, we argue that there exists a sweet spot between these two extremes (rigid parametrisation versus virtually no assumption at all), albeit a scarcely investigated one.

A first step in this direction is the robust R-UCB algorithm of Ashutosh et al. (2021), that trades off logarithmic regret for $\mathcal{O}(f(T) \log T)$, where $f$ is a slowly growing function replacing the fixed sub-Gaussian parameter of the standard UCB, essentially tracking the possible mass leakage at infinity. However, it inherits the typically suboptimal practical performances of UCB and thus hardly constitutes a viable alternative for practitioners.

**Linear bandit policies.** The main index policies (UCB, TS) were also extended to contextual linear bandits, resulting in LinUCB (Abbasi-Yadkori et al., 2011) and LinTS (Agrawal and Goyal, 2013; Abeille and Lazaric, 2017). Moreover, instance-dependent optimal algorithms have recently been developed for finite action sets, such as Degenne et al. (2020) for sub-Gaussian SPEF rewards (with identity feature function $F$), which is inspired by saddle-point problems in pure exploration (interestingly, its formulation is flexible enough to adapt to other structured bandit models); Tirinzoni et al. (2020) for sub-Gaussian rewards using a primal-dual approach to match the structured lower bound; also for sub-Gaussian rewards, Kirschner et al. (2021), which couples information-directed sampling with a similar primal-dual analysis.

**Risk-aware algorithms.** Another hindrance to the adoption of existing recommendation algorithms that real-world agents may have a preference for policies that do *not* maximise average rewards but rather a different, risk-adjusted criterion. Besides the motivating example in healthcare (see Foreword), we mention the recommendations of crop management practices in agriculture, with specific risk-aversions to price fluctuations (market-oriented farmers) or food insecurity (smallholder farmers in developing countries) (Gautron et al., 2022).

While mathematically convenient (linear, naturally expressed by Gaussian likelihood, etc.), the expectation risk measure is known to equally weight large positive and negative outcomes, possibly leading to risky policies unsuitable to critical applications, and is also sensitive to outliers. In contrast, *risk-aware* measures emphasise different characteristics of the observed reward distributions, e.g. by stressing out the impact of adverse outcomes (Dowd, 2007). Such measures include the mean-variance (Markowitz, 1952), conditional Value-at-Risk (Rockafellar and Uryasev, 2000), which is a special case of spectral risk measures (Acerbi, 2002), entropic risk (Ahmadi-Javid, 2012) and the expectiles (Newey and Powell, 1987). These risk measures, in particular the conditional value at risk ($\mathrm{CVaR}$), have been studied as alternatives to the expectation criterion in classical multiarmed bandits (Galichet et al., 2013; Gopalan et al., 2017; Cassel et al., 2018; Tamkin et al., 2019; Prashanth et al., 2020; Pandey et al., 2021; Baudry et al., 2021a); the latter in particular proposes an extension of NPTS that is optimal in the sense of Theorem 1.6 for $\rho = \mathrm{CVaR}$. We also refer to Maillard (2013) for the entropic risk. In distributional reinforcement learning, quantile regression has been studied for DQN (Dabney et al., 2018). However, the literature on risk aware *contextual* bandits is much sparser, see e.g. Wirth et al. (2022) with mean-variance and $\mathrm{CVaR}$ bandit optimisation in the context of vehicular communication. Despite promising empirical results, contributions outside of multiarmed bandits are largely devoid of theoretical regret guarantees.

**Table 2.1** – Summary of classical bandit algorithms for the expectation risk measure and various stationary bandit models. Elements listed as prior knowledge are used within the algorithm.

| Algorithm | Regret guarantees | Algorithm prior knowledge |
|---|---|---|
| UCB1<br>Auer et al. (2002) | $\mathcal{K}_{\mathrm{inf}}$-optimal on $\mathcal{F}_{\mathcal{N},\sigma}$<br>Logarithmic on $\mathcal{F}_{\mathcal{G},R}$ | $\sigma \in \mathbb{R}_+^\star$ or $R \in \mathbb{R}_+^\star$ |
| UCB-V<br>Audibert et al. (2009) | Logarithmic on $\mathcal{F}_{[\underline{B},\overline{B}]}$ | $\underline{B}, \overline{B}$ |
| kl-UCB<br>Cappé et al. (2013) | $\mathcal{K}_{\mathrm{inf}}$-optimal on $\mathcal{F}^{\mathrm{SPEF}}_{h,F,\mathcal{L}}$ | $(\theta, \theta') \in \Theta^2 \mapsto \mathcal{B}_{\mathcal{L}}(\theta', \theta)$ |
| Empirical KL-UCB<br>Cappé et al. (2013) | $\mathcal{K}_{\mathrm{inf}}$-optimal on $\mathcal{F}_{[\underline{B},\overline{B}]}$ | $\underline{B}, \overline{B}$ |
| IMED<br>Honda and Takemura (2015) | $\mathcal{K}_{\mathrm{inf}}$-optimal on $\mathcal{F}^{\ell} \cap \mathcal{F}_{(-\infty,\overline{B}]}$ | $\overline{B}$ |
| LB-SDA, RB-SDA<br>Baudry et al. (2020) | $\mathcal{K}_{\mathrm{inf}}$-optimal on $\mathcal{F}^{\mathrm{SPEF}}$ | None<br>(may require forced exploration) |
| TS<br>Thompson (1933)<br>Korda et al. (2013)<br>Agrawal and Goyal (2012) | $\mathcal{K}_{\mathrm{inf}}$-optimal on $\mathcal{F}^{\mathrm{SPEF}}$<br>Logarithmic on $\mathcal{F}_{[\underline{B},\overline{B}]}$ | Suitable SPEF conjugate prior<br>$\underline{B}, \overline{B}$ (binarised) |
| NPTS<br>Riou and Honda (2020) | $\mathcal{K}_{\mathrm{inf}}$-optimal on $\mathcal{F}_{[\underline{B},\overline{B}]}$ | $\underline{B}, \overline{B}$ |
| Robust UCB<br>Bubeck et al. (2013) | Logarithmic on $\mathcal{F}_{M,\varepsilon}$ | $M, \varepsilon$ |
| $\mathcal{K}_{\mathrm{inf}}$-UCB<br>Agrawal et al. (2021) | $\mathcal{K}_{\mathrm{inf}}$-optimal on $\mathcal{F}_{M,\varepsilon}$ | $M, \varepsilon$ |
| $h$-NPTS<br>Baudry et al. (2023) | $\mathcal{K}_{\mathrm{inf}}$-optimal ($\eta \in \mathbb{R}_+^\star$) on<br>$\mathcal{F}^{\mathrm{centred}}_{\kappa,\varepsilon} \bigcap \left( \bigcup_{M \in \mathbb{R}_+^\star} \mathcal{F}^{M,1+2\varepsilon+\eta} \right)$ | $\kappa, \varepsilon$ |
| R-UCB<br>Ashutosh et al. (2021) | $\mathcal{O}\left(\log(T) f(T)\right)$ on $\mathcal{F}_\ell$ | Slowly growing function $f$ |
| Generic UCB<br>Chapter 1 | Depends on the confidence bounds<br>Logarithmic on $\mathcal{F}_{\mathcal{G},R}$, $\mathcal{F}^{\mathrm{EF}}_{h,F,\mathcal{L}}$, $\mathcal{F}_{M,\varepsilon}$, etc. | Time-uniform confidence<br>sequence (see Table 2.3) |
| Dirichlet sampling<br>Chapter 6 | $\mathcal{K}_{\mathrm{inf}}$-optimal on new settings<br>$\mathcal{O}\left(\log(T) \log\log(T)\right)$ on $\mathcal{F}_\ell$ (RDS) | New settings<br>None for RDS |

## 2.2 How do others concentrate? Overview of confidence bounds

Many learning tasks, and specifically bandits, require to estimate some characteristics of an unknown distribution (typically the mean) solely based on observed samples. The theory of concentration provides a natural way to control the magnitude of the estimation error. Besides the background material of Chapter 1, we refer the interested reader to the monographs of Boucheron et al. (2013); Raginsky et al. (2013); Dembo and Zeitouni (2009).

**Fixed samples.** In applied statistics, confidence sets and corresponding p-values are often derived from Gaussian assumptions (e.g. the ubiquitous mean $\pm 1.96 \times$ standard deviation/$\sqrt{n}$ formula at 95% significance). However, these are typically only valid asymptotically (central limit theorem) or for Gaussian distributions only (Student $t$-test), which is incompatible with our aim towards small samples and nonparametric guarantees. Similarly, resampling methods (bootstrap, jacknife, see Efron (1992)), although very flexible, lack finite sample guarantees.

Many concentration inequalities have been developed over the years to provide nonasymptotic confidence sets with provably high probability, most notably[1] Chernoff (1952); Hoeffding (1963) for sub-Gaussian or bounded distributions and Bernstein (1924); Bennett (1962) for bounded distributions with specified variance, all derived from bounding the MGF (as in Proposition 1.15), and later improved by Bentkus (2004) using tighter polynomial bounds. Motivated by practical applications, the variance specification was also replaced by data-dependent estimators, giving rise to *empirical* bounds, see e.g. Maurer and Pontil (2009); Kuchibhotla and Zheng (2021). While mathematically convenient, these simple inequalities are often quite loose, resulting in much less statistical power compared to parametric tests. Of note, such concentration bounds are nonetheless useful in other areas of statistical learning such as PAC learning, with e.g. the recent unexpected Bernstein inequality of Mhammedi et al. (2019), extended to CVaR generalisation bounds in Mhammedi et al. (2020). Recently, an altogether different method for bounded distributions (Phan et al., 2021), building on Anderson's bound, provided numerically sharp confidence sets for small samples, which however lack closed-form expressions and require Monte Carlo estimation to be calculated.

Outside the scope of bounded distributions, a vast corpus of the concentration literature has studied *self-normalised sums*, drawing inspiration from Student's $t$-statistic in the Gaussian case (Bercu and Touati, 2008; Delyon, 2009; Peña et al., 2009). The latest developments in this field provide deviation bounds under the sole assumption of known and finite variance (Bercu and Touati, 2019, Theorem 2.6); to our knowledge, no empirical extension of this self-normalised bound has been derived, thus again limiting the potential for practical applications.

[1] Like many other areas of science, the literature on the concentration of measure is no stranger to Stigler's law of eponymy. Many standard bounds can be traced back to the seminal work of Sergei Bernstein in the 1920s and 1930s, and then rediscovered several decades later. For ease of reference, we use the most common names.

**Anytime-valid statistics** The challenge of bridging the gap between theory and practice in bandits is all the more arduous as it requires more sophisticated concentration tools than those presented above. Although any fixed sample confidence set can be made uniformly valid over a finite horizon (union bound, known as Bonferonni correction in multiple testing), this comes at the cost of dramatically decreasing the power of the associated statistical tests. A better approach is to combine supermartingale techniques with such union arguments over geometric time grids, a technique known as time peeling (or stitching) — see Bubeck (2010); Cappé et al. (2013) for early uses in bandits, Garivier (2013), and more recently Maillard (2019b); see also Howard et al. (2020, 2021) for a recent, complementary survey of the history of this field. These peeling-based confidence sequences are able to match the $\mathcal{O}(\sqrt{\log\log(t)/t})$ lower bound prescribed by the law of iterated logarithm (Proposition 1.28); however, as argued in Chapter 1, a subtlety of time-uniform statistics is that such asymptotically optimal rates are often associated with wide confidence sets for small samples, which is undesirable in practice.

Another approach, which we favour in the thesis, is the *method of mixtures*, initiated by Robbins and Pitman (1949); Robbins (1970) and popularised further in Peña et al. (2009). It constitutes a powerful alternative to peeling for developing anytime-valid confidence sequences with explicit expressions and tight practical performances. It was originally developed for (sub-)Gaussian families, and has seen widespread use in bandits after the seminal work of Abbasi-Yadkori et al. (2011). Still, generalising this method beyond the sub-Gaussian case, for which natural conjugate mixing measures exist, has been a challenging issue. Kaufmann and Koolen (2021) applied an ad hoc discrete mixture construction to SPEF, with explicit bounds restricted to Gaussian and Gamma distributions (known variance and shape).

Finally, a fairly different *capital process* construction, has been recently developed in Shafer and Vovk (2019), in relation to the notion of e-processes mentioned in Chapter 1. It has been further analysed for bounded distributions in Waudby-Smith and Ramdas (2023), leading to so-called hedged capital confidence sequences with predictable mixtures, with sharp concentration with the sole knowledge of the distribution support.

**With so many existing confidence bounds, why look for more?** First, given their widespread use in statistical modelling, in particular in bandits, it is perhaps surprising that no cromulent anytime-valid confidence sequence was developed for generic exponential families. Moreover, following our discussion on extending nonparametric bandit algorithms to more practitioner-friendly settings, we believe it is worthwhile to investigate concentration tools for alternative specifications, for instance unbounded distributions with no prior information on the variance nor the sub-Gaussian parameter (in a similar fashion to empirical Bernstein inequalities).

**Table 2.2** – Summary of nonasymptotic, fixed sample confidence sets for the mean. The dependency on the confidence level $\delta \in (0,1)$ is typically $\mathcal{O}(\sqrt{\log(1/\delta)})$.

| Confidence set | Concentration rate | Prior knowledge |
|---|---|---|
| Chernoff, Hoeffding<br>Chernoff (1952)<br>Hoeffding (1963) | $\mathcal{O}\left(\frac{R}{\sqrt{t}}\right)$ on $\mathcal{F}_{\mathcal{G},R}$ and $\mathcal{F}_{[\underline{B},\overline{B}]}$ | $R = \frac{\overline{B}-\underline{B}}{2}$ |
| Bernstein, Bennett<br>Bernstein (1924)<br>Bennett (1962) | $\mathcal{O}\left(\frac{\sigma}{\sqrt{t}} + \frac{\overline{B}-\underline{B}}{t}\right)$ on $\mathcal{F}_{[\underline{B},\overline{B}]} \cap \mathcal{F}_{\sigma,1}^{\text{centred}}$ | $\sigma, \underline{B}, \overline{B}$ |
| Bentkus, Pinelis<br>Bentkus (2004)<br>Pinelis (2006) | $\mathcal{O}\left(\frac{\sigma}{\sqrt{t}} + \frac{\overline{B}-\underline{B}}{t}\right)$ on $\mathcal{F}_{[\underline{B},\overline{B}]} \cap \mathcal{F}_{\sigma,\varepsilon}^{\text{centred}}$<br>(tighter) | $\sigma, \varepsilon \geqslant 1, \underline{B}, \overline{B}$ |
| Bercu-Touati<br>Bercu and Touati (2019) | Self-normalised, $\mathcal{F}_{\sigma,\varepsilon}^{\text{centred}}$ | $\sigma, \varepsilon \geqslant 1$ |
| Student<br>Student (1908) | $\mathcal{O}\left(\frac{\widehat{\sigma}_t}{\sqrt{t}}\right)$ on $\mathcal{F}_{\mathcal{N}}$ | Gaussian |
| Empirical Bernstein<br>Maurer and Pontil (2009) | $\mathcal{O}\left(\frac{\widehat{\sigma}_t}{\sqrt{t}} + \frac{\overline{B}-\underline{B}}{t}\right)$ on $\mathcal{F}_{[\underline{B},\overline{B}]}$ | $\underline{B}, \overline{B}$ |
| Empirical Bentkus<br>Kuchibhotla and Zheng (2021) | $\mathcal{O}\left(\frac{\widehat{\sigma}_t}{\sqrt{t}} + \frac{\overline{B}-\underline{B}}{t}\right)$ on $\mathcal{F}_{[\underline{B},\overline{B}]}$<br>(tighter) | $\underline{B}, \overline{B}$ |
| Anderson-Phan<br>Phan et al. (2021) | Simulated, $\mathcal{F}_{[\underline{B},\overline{B}]}$<br>(very tight for small samples) | $\underline{B}, \overline{B}$ |
| Hedged capital<br>Waudby-Smith and Ramdas (2023) | $\mathcal{O}\left(\frac{\overline{B}-\underline{B}}{\sqrt{t}}\right)$ on $\mathcal{F}_{[\underline{B},\overline{B}]}$<br>(implicit bound, quite tight) | $\underline{B}, \overline{B}$ |
| Bregman deviations<br>Chapter 3 | $\mathcal{O}\left(\frac{1}{\sqrt{t}}\right)$ on $\mathcal{F}_{h,F,\mathcal{L}}^{\text{EF}}$ | Generic EF |
| Empirical Chernoff<br>Chapter 4 | $\approx \mathcal{O}\left(\frac{\widehat{\sigma}_t}{\rho\sqrt{t}}\right)$<br>(second order sub-Gaussian) | $\rho \in (0,1]$ |

**Table 2.3** – Summary of nonasymptotic, time-uniform confidence sequences for the mean. The dependency on the confidence level $\delta \in (0,1)$ is typically $\mathcal{O}(\sqrt{\log(1/\delta)})$.

| Confidence sequence | Concentration rate | Prior knowledge |
|---|---|---|
| Union bound over any $T$ fixed sample bounds | $\mathcal{O}\left(\sqrt{\frac{\log T}{t}}\right)$ at most (usually very slow) | Depends on the bound Horizon $T$ |
| Geometric time-peeling | $\mathcal{O}\left(R\sqrt{\frac{\log\log t}{t}}\right)$ (usually slow for small samples) | Depends on the bound |
| Method of mixtures (sub-Gaussian) | $\mathcal{O}\left(R\sqrt{\frac{\log\sqrt{t}}{t}}\right)$ on $\mathcal{F}_{\mathcal{G},R}$ | $R$ |
| Inverted stitching Howard et al. (2021) | $\mathcal{O}\left(R\sqrt{\frac{\log\log t}{t}}\right)$ on $\mathcal{F}_{\mathcal{G},R}$ | $R$ |
| Kaufmann-Koolen Kaufmann and Koolen (2021) | $\mathcal{O}\left(\sqrt{\frac{\log\log t}{t}}\right)$ on $\mathcal{F}^{\text{SPEF}}_{h,F,\mathcal{L}}$ | Gaussian (known variance) Gamma (known shape) |
| Hedged capital Waudby-Smith and Ramdas (2023) | $\mathcal{O}\left(\left(\overline{B}-\underline{B}\right)\sqrt{\frac{\log\log t}{t}}\right)$ on $\mathcal{F}_{[\underline{B},\overline{B}]}$ (implicit bound, quite tight) | $\underline{B}, \overline{B}$ |
| Bregman deviations Chapter 3 | $\mathcal{O}\left(\sqrt{\frac{\log\sqrt{t}}{t}}\right)$ on $\mathcal{F}^{\text{EF}}_{h,F,\mathcal{L}}$ | Generic EF |
| Empirical Chernoff Chapter 4 | $\approx\mathcal{O}\left(\frac{\widehat{\sigma}_t}{\rho}\sqrt{\frac{\log\sqrt{t}}{t}}\right)$ (second order sub-Gaussian) | $\rho \in (0,1]$ |

## 2.3 Outline and contributions

In these two preliminary chapters, we have introduced a statistical toolbox (stochastic bandits, concentration) to formally analyse statistical sequential decision-making problems. The main goal of this thesis is to adapt these well-known tools to the specific context of healthcare recommendations, which raises several new and difficult challenges compared to current use cases of such methods.

First, although we live in the era of big data, heralded with the large-scale use of internet, many fields of applications, among which healthcare is a prominent example, work with small, highly specialised datasets. To put it simply, what constitutes *big* data for a medical professional that has carefully monitored a cohort of patients over several years for a clinical trial is orders of magnitude lower than what is available for prediction and recommendation in an online marketplace, streaming platforms, advertisement company, etc.

- In Chapter 3, we adapt the method of mixtures, primarily designed for (sub)-Gaussian distributions, to the broader setting of generic exponential families. While these are parametric models, we believe these new concentration inequalities fill an important gap in the literature and provide practitioners with more extensive modelling tools. With applicability in mind, we instantiate our novel bound to many classical families and illustrate a use case in change point detection.

- In Chapter 4, we introduce the notion of second order sub-Gaussian distributions, a refinement on the classical notion of sub-Gaussian tail control that allows for empirical concentration inequalities without the need for a prior bound on the variance or on the support. We derive implementable confidence sets using special functions (Lambert $W$, confluent hypergeometric functions), both in fixed samples and time-uniform. We believe this new class of nonparametric distributions offer an appealing modelling alternative for practitioners, more flexible than rigid parametric specifications à la Gaussian, but sharper than many other nonparametric families (e.g. bounded distributions).

Second, it is noteworthy than the majority of the sequential decision-making literature considers only the expectation risk measure. Besides the fact that it makes for a convenient mathematical setting, we believe this is also driven by the current industrial applications of such methods, since most online recommender systems aim to maximise *clickthrough rate*, i.e. the expected frequency of visits to given ads, movies, purchasable items, etc. Although some risk-aware algorithms have been analysed for the simple multiarmed bandit problem, more sophisticated models such as contextual bandits have been almost exclusively studied in risk-neutral settings, barring a few mostly empirical studies.

- In Chapter 5, we revisit the LinUCB algorithm for linear bandits to optimise certain risk measures of the reward distributions, generalising beyond the standard mean criterion. We consider risk measures that are *elicitable* by convex loss functions, extending the classical sequential ridge regression to a broader class of empirical risk minimisation problems. Under a bounded curvature assumption of the loss, we are able to derive pseudo regret upper bounds based on a variant of the multivariate method of mixtures, that naturally extend the guarantees of the standard LinUCB.

Third, we study an alternative approach to deterministic UCB-style algorithms based on randomisation, continuing our effort towards more flexible, nonparametric models.

- In Chapter 6, we introduce *Dirichlet sampling*, a resampling scheme for multiarmed bandits with a generic exploration incentive structure that can be adapted to different, original settings. In particular, we obtain optimal guarantees for two new families of distributions, namely bounded under a detectability assumption alone (without knowing the bounds), and semibounded distributions with a quantile condition. We also derive near-logarithmic pseudo regret for the class of *all* light tailed distributions, with competitive performances on a vast range of experiments.

- In Chapter 7, we extend the Dirichlet sampling scheme to contextual linear bandits (in the generic elicitable risk framework of Chapter 5) using weighted bootstrap regression. We propose a heuristic regret analysis, backed by the theory of strong approximation of empirical processes, revealing a connection between Dirichlet sampling for nonparametric statistical models and the linear Gaussian Thompson sampling. We also show substantial improvements on numerical experiments compared to optimistic algorithms.

These first three parts offer different approaches to improve the current state of sequential recommendation algorithms and help bridge the gap between theory and practice. However, we believe that in order to truly contribute to applications beyond the scope of machine learning and mathematics, here in healthcare, it is essential to collaborate conjointly with medical practitioners and work with real patient data on clinically relevant issues. The last part of this thesis illustrates this commitment to such an interdisciplinary approach.

As mentioned earlier, this thesis is part of a broader, long-term collaboration between machine learning researchers at Inria Scool and medical surgeons and researchers at Lille University Hospital, aiming to develop data-driven recommendations for bariatric postoperative follow-up. We identified early on that a first step towards this ambitious goal was to predict with reasonable accuracy the expected weight trajectories of patients living with severe obesity. Indeed, long-term weight loss is the primary outcome of this type of surgical procedures, and irregular weight trajectories are often associated with complications, the early detection and prevention of which being the main concern of postoperative follow-up. In this part, we focus on more traditional machine learning methods than bandits, however we hope this paves a way to build further towards personalised prospective recommendations for patients.

- In Chapter 8, we report the development of a white box prediction model for weight trajectories, and its validation on a large international cohort. Starting on a rich dataset of patients who underwent surgery in Lille, we extract a subset of predictive covariates using LASSO feature selection, which we further use to train a set of decision trees to predict weight loss. This simple method is highly interpretable even by non-specialists, such as primary care providers. The prediction model, as well as a graphical overlay to display smooth weight trajectories, is embedded in a freely available website.

Finally, we acknowledge that this thesis is rather long. To help readers navigate it, we suggest a reading order, indicated by the following symbols.

- For readers with a background in statistics and probability interested in concentration bounds: start with Chapters 3 and Chapters 4.
- For readers interested in stochastic bandits: start with Chapters 5, 6 and 7.
- For readers with a medical background interested in bariatric surgery, start with Chapter 8.

Furthermore, readers may find in appendix technical results and complementary discussions. Appendix sections that we believe may be skipped at first reading and that are available in a similar form in published articles are indicated by the symbol ↝.

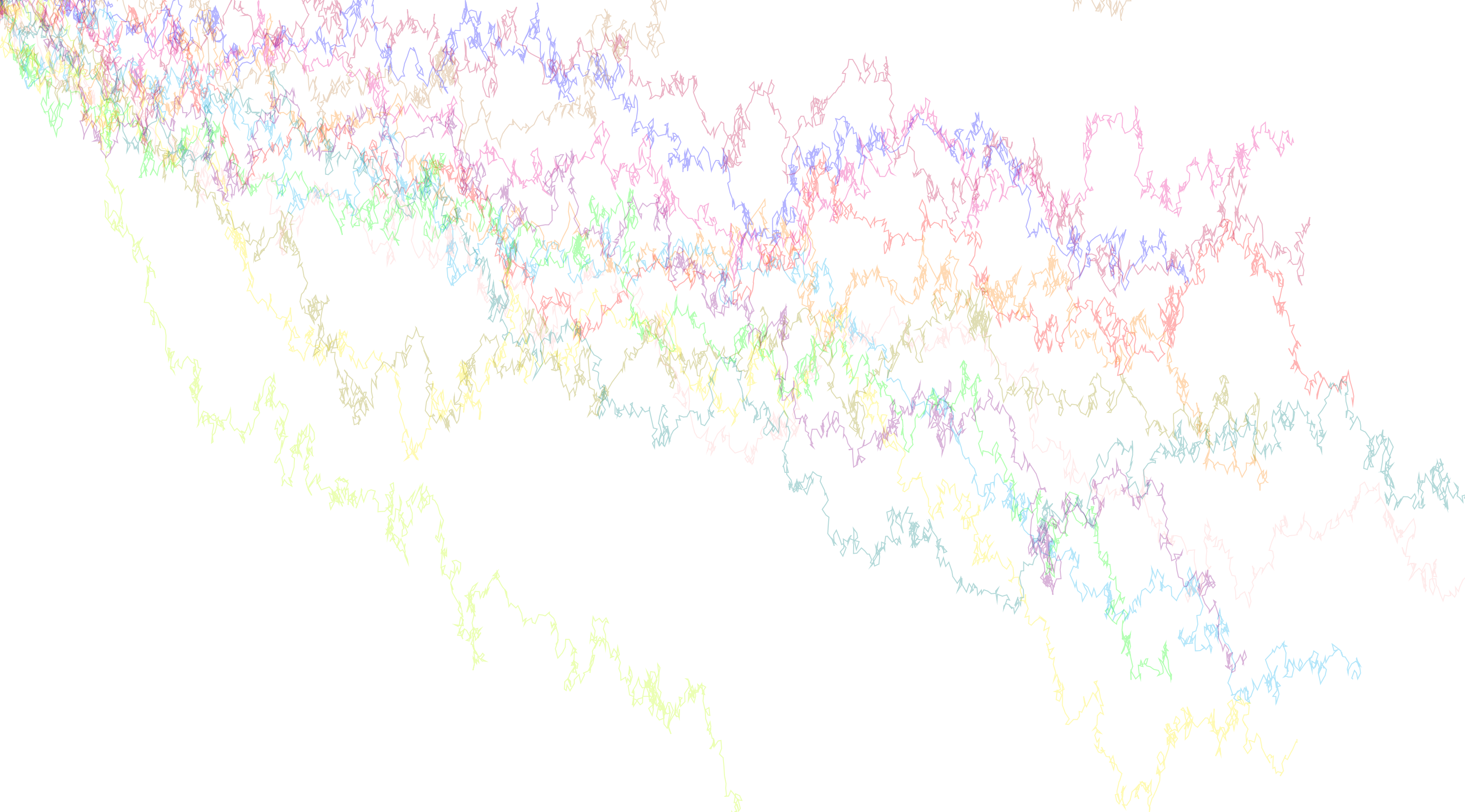

# Part II

# Safe, anytime-valid and efficient concentration

*or how to play smart with few data*

# Chapter 3

# Bregman deviations of generic exponential families ( )

*I mention his talk about angles because it suggests something Wilcox had told me of his awful dreams. He had said that the geometry of the dream-place he saw was abnormal, non-Euclidean, and loathsomely redolent of spheres and dimensions apart from ours.*

— H.P. Lovecraft, *The Call of Cthulhu*

The content of this chapter was published at the 2023 *Conference On Learning Theory* (COLT) (Chowdhury et al., 2023), and was presented at the 2022 *Probability PhD Day* at the University of Lille, the 2022 *Safe, Anytime-Valid Inference* (SAVI) workshop, and the 2023 *Machine Learning and Massive Data Analysis* (MLMDA) seminar at ENS Paris-Saclay. Compared to the article, we report here in the main text (i) the proofs of the time-uniform (Theorem 3.3) and doubly time-uniform (Theorem 3.8) confidence sequences, highlighting connections with the classical method of mixtures presented in Chapter 1, and (ii) additional numerical experiments, especially for the GLR change point test. The instantiation of our novel bound to classical families is summarised in Table 3.1 and is available off-the-shelf in the concentration-lib Python module. To avoid cluttering, the complete derivation of these bounds, which amounts to simple but tedious algebra, is deferred to Appendix B .

## Contents

## Outline and contributions

In this chapter, we revisit the method of mixtures for parametric exponential families of arbitrary dimension, expressing deviations in the natural (Bregman) divergence of the family. The setting of exponential families is convenient for integration in a Bayesian setup, thanks to the notion of conjugate prior that enables us to reduce computation of tedious integrals to simple parameter updates. Here, we exploit this property to obtain explicit mixtures of martingales, following the principle outlined in Theorem 1.20. Exponential families are largely used in modern machine learning, yet concentration tools available to the practitioner are comparatively scarce — the only previous work on exponential families (Kaufmann and Koolen, 2021) used a discrete mixture construction limited to SPEF, with numerical implementation only for Gaussian (known variance) and Gamma (known shape) distributions. To help close this gap, we obtain both sharp and computationally tractable confidence sets, especially in the small sample regime.

In Section 3.1, we first recall some background material on exponential families and their associated Bregman divergences. Section 3.2 states our main result (Theorem 3.3): a *time-uniform* concentration inequality for exponential families. Specifically, we control the Bregman deviations associated with the log-partition function of the family using a novel information-theoretic quantity, the *Bregman information gain*. On a high level, this quantifies the information gain about the parameter of a distribution in a given family after observing i.i.d. samples from it, which is measured in terms of the natural Bregman divergence of the family. To illustrate the utility of this general result, we detail in Section 3.3 how the Bregman information gain and deviation inequalities specialise for well-known exponential families, resulting in fully explicit confidence sets (Table 3.1). To the best of our knowledge, this is the first explicit time-uniform inequality for two-parameter Gaussian (i.e. both mean and variance are unknown), Chi-square, Weibull, Pareto, and Poisson distributions. In Section 3.4, we numerically evaluate the high-probability confidence sets built from our method for classical families, and achieve state-of-the-art time-uniform bounds. Finally, in Section 3.5, we generalise Theorem 3.3 to obtain a *doubly time-uniform* concentration inequality for generic exponential families (Theorem 3.8). We present an application of both these results in controlling the type I error of the generalised likelihood ratio (GLR) test, used for change point detection in exponential family models.

## 3.1 Preliminary: Bregman tail and duality properties

We start by introducing classical properties of Bregman divergences which are central to our confidence sequence construction. We consider an exponential family $\mathcal{F}^{\mathrm{EF}}_{h,F,\mathcal{L}} = \{p_\theta,\ \theta \in \Theta\}$ over a set $\mathcal{Y}$, parametrised by an open set $\Theta \subseteq \mathbb{R}^d$ (see Section 1.3 in Chapter 1). We recall the expression of the density and the Bregman divergence associated with the (finite, invertible) convex mapping $\mathcal{L} : \Theta \to \mathbb{R}$:

$$\begin{aligned} p_\theta \colon \mathcal{Y} &\longrightarrow \mathbb{R}_+ \\ y &\longmapsto h(y)e^{\langle \theta, F(y)\rangle - \mathcal{L}(\theta)} \end{aligned} \quad \text{and} \quad \begin{aligned} \mathcal{B}_\mathcal{L} \colon \Theta \times \Theta &\longrightarrow \mathbb{R}_+ \\ (\theta', \theta) &\longmapsto \mathcal{L}(\theta') - \mathcal{L}(\theta) - \langle \theta' - \theta, \nabla \mathcal{L}(\theta)\rangle \,. \end{aligned} \tag{3.1}$$

The first property links the Bregman divergence $\mathcal{B}_\mathcal{L}$ to the moment generating function of the random variable $F(Y)$ where $Y \sim p_\theta$ and $\theta \in \Theta$, which makes Bregman divergences well suited to control the tail behaviour of random variables appearing in concentration inequalities. The second one highlights duality properties that enables convenient algebraic manipulations. To this end, we let $\Lambda_\theta = \{\lambda \in \mathbb{R}^d,\ \lambda + \theta \in \Theta\}$ and define the function

$$\begin{aligned} \mathcal{B}_{\mathcal{L},\theta} \colon \Lambda_\theta &\longrightarrow \mathbb{R} \\ \lambda &\longmapsto \mathcal{B}_\mathcal{L}(\theta + \lambda, \theta) = \mathcal{L}(\theta + \lambda) - \mathcal{L}(\theta) - \langle \lambda, \nabla \mathcal{L}(\theta)\rangle \,. \end{aligned} \tag{3.2}$$

We also introduce its Legendre-Fenchel dual $\mathcal{B}^\star_{\mathcal{L},\theta} \colon x \mapsto \sup_{\lambda \in \Lambda_\theta} \langle \lambda, x\rangle - G(\lambda)$.

**Lemma 3.1** (Properties of Bregman divergences). *Let $\theta \in \Theta$ and $\lambda \in \Lambda_\theta$. The following equality holds:*

$$\log \mathbb{E}_\theta \left[\exp\left(\langle \lambda, F(Y) - \mathbb{E}_\theta[F(Y)]\rangle\right)\right] = \mathcal{B}_{\mathcal{L},\theta}(\lambda) \,. \tag{3.3}$$

*Furthermore, if $\nabla \mathcal{L}$ is one-to-one, the following Bregman duality relations hold for any parameters $\theta, \theta' \in \Theta$:*

$$\mathcal{B}_\mathcal{L}(\theta', \theta) = \mathcal{B}^\star_{\mathcal{L},\theta'}(\nabla \mathcal{L}(\theta) - \nabla \mathcal{L}(\theta')) = \mathcal{B}_{\mathcal{L}^\star}(\nabla \mathcal{L}(\theta), \nabla \mathcal{L}(\theta')) \,. \tag{3.4}$$

*More generally, the following holds for any $\alpha \in [0, 1]$ :*

$$\mathcal{B}^\star_{\mathcal{L},\theta'}\left(\alpha(\nabla \mathcal{L}(\theta) - \nabla \mathcal{L}(\theta'))\right) = \mathcal{B}_\mathcal{L}(\theta', \theta_\alpha) \,, \tag{3.5}$$

*where*

$$\theta_\alpha = \nabla \mathcal{L}^{-1}(\alpha \nabla \mathcal{L}(\theta) + (1 - \alpha)\nabla \mathcal{L}(\theta')) \,. \tag{3.6}$$

The second half of this technical lemma is essentially a change of variable formula to move back and forth between two representations of an exponential family: in natural parameters (measured by the Bregman divergence between $\theta'$ and $\theta$) and in expectation parametrisation (measured by the dual Bregman divergence between $\nabla\mathcal{L}(\theta) = \mathbb{E}_\theta[F(Y)]$ and $\nabla\mathcal{L}(\theta') = \mathbb{E}_{\theta'}[F(Y)]$ ). For more background on this, which forms the basis of the *information geometry* field, we refer to Amari (2016, Sections 2.1 and 2.7). This result is at the root of the martingale construction behind Theorem 3.3 in the next section.

*Proof of Lemma 3.1.* The first equality is immediate, since $\mathbb{E}_\theta[F(Y)] = \nabla\mathcal{L}(\theta)$ and

$$\log \mathbb{E}_\theta\left[\exp(\langle\lambda, F(Y)\rangle)\right] = \log \int \exp(\langle\lambda, F(y)\rangle + \langle\theta, F(y)\rangle - \mathcal{L}(\theta))dy \tag{3.7}$$

$$= \mathcal{L}(\lambda+\theta) - \mathcal{L}(\theta)\,. \tag{3.8}$$

We now turn to the duality formula. Using the definition of each terms, we have

$$\mathcal{B}^\star_{\mathcal{L},\theta'}(\nabla\mathcal{L}(\theta) - \nabla\mathcal{L}(\theta')) = \sup_{\lambda\in\Lambda_\theta} \langle\lambda, \nabla\mathcal{L}(\theta) - \nabla\mathcal{L}(\theta')\rangle - \left[\mathcal{L}(\theta'+\lambda) - \mathcal{L}(\theta') - \langle\lambda, \nabla\mathcal{L}(\theta')\rangle\right] \tag{3.9}$$

$$= \sup_{\lambda\in\Lambda_\theta} \langle\lambda, \nabla\mathcal{L}(\theta)\rangle - \mathcal{L}(\theta'+\lambda) + \mathcal{L}(\theta')\,. \tag{3.10}$$

An optimal $\lambda$ must satisfy $\nabla\mathcal{L}(\theta) = \nabla\mathcal{L}(\theta'+\lambda)$. Hence, provided that $\nabla\mathcal{L}$ is invertible, this means $\lambda = \theta - \theta'$. Plugin-in this value, we obtain

$$\mathcal{B}^\star_{\mathcal{L},\theta'}(\nabla\mathcal{L}(\theta) - \nabla\mathcal{L}(\theta')) = \langle\theta-\theta', \nabla\mathcal{L}(\theta)\rangle - \mathcal{L}(\theta) + \mathcal{L}(\theta') = \mathcal{B}_{\mathcal{L}}(\theta', \theta)\,. \tag{3.11}$$

The remaining equality $\mathcal{B}_{\mathcal{L}}(\theta',\theta) = \mathcal{B}_{\mathcal{L}^\star}(\nabla\mathcal{L}(\theta), \nabla\mathcal{L}(\theta'))$ is a standard result. Finally, regarding the generalisation, we note that

$$\mathcal{B}^\star_{\mathcal{L},\theta'}\left(\alpha(\nabla\mathcal{L}(\theta) - \nabla\mathcal{L}(\theta'))\right) = \sup_{\lambda\in\Lambda_\theta} \langle\lambda, \alpha\nabla\mathcal{L}(\theta) + (1-\alpha)\nabla\mathcal{L}(\theta')\rangle - \mathcal{L}(\theta'+\lambda) + \mathcal{L}(\theta')\,. \tag{3.12}$$

Hence, an optimal $\lambda$ must now satisfy $\alpha\nabla\mathcal{L}(\theta) + (1-\alpha)\nabla\mathcal{L}(\theta') = \nabla\mathcal{L}(\theta'+\lambda)$, that is $\lambda = \theta_\alpha - \theta'$. This further yields

$$\mathcal{B}^\star_{\mathcal{L},\theta'}\left(\alpha(\nabla\mathcal{L}(\theta) - \nabla\mathcal{L}(\theta'))\right) = \langle\theta_\alpha - \theta', \nabla\mathcal{L}(\theta_\alpha)\rangle - \mathcal{L}(\theta_\alpha) + \mathcal{L}(\theta') = \mathcal{B}_{\mathcal{L}}(\theta', \theta_\alpha)\,. \tag{3.13}$$

■

## 3.2 Time-uniform Bregman concentration

In this section, we are interested in controlling the deviation between a fixed parameter $\theta^\star \in \Theta$ and its estimate $\widehat{\theta}_t$ built from $t$ observations from distribution $p_{\theta^\star}$. We naturally measure this deviation in terms of the canonical Bregman divergence of the family. As in Section 1.4 in Chapter 1, we would like to control this deviation not only for a single sample size $t$, but *simultaneously for all* $t \in \mathbb{N}$. Namely, we would like to upper bound quantities of the form $\mathbb{P}\left(\exists t \in \mathbb{N} : \mathcal{B}_{\mathcal{L}}(\theta^\star, \widehat{\theta}_t) \geqslant \dots\right)$. We recall that such controls are paramount when observations are gathered sequentially (either actively or otherwise), especially when the number of observations is unknown beforehand, as in stochastic bandits or reinforcement learning.

### Bregman information gain

We first introduce a quantity that measures a form of *information gain* about $\theta^\star$ after observing $t$ samples, but expressed in terms of the natural Bregman divergence.

**Definition 3.2** (Bregman information gain). *Let $(Y_s)_{s=1}^t$ be i.i.d. samples drawn from $p_{\theta^\star}$, where $\theta^\star \in \Theta \subset \Theta_I$, and let $\theta_0 \in \Theta$ be a reference parameter. For any constant $c > 0$, the* ***Bregman information gain*** *about $\theta^\star$ from $\theta_0$ after observing $(Y_s)_{s=1}^t$ is defined as*

$$\gamma_{t,c}(\theta_0) = \log\left(\frac{\int_\Theta \exp\left(-c\mathcal{B}_{\mathcal{L}}(\theta', \theta_0)\right) d\theta'}{\int_\Theta \exp\left(-(t+c)\mathcal{B}_{\mathcal{L}}(\theta', \widehat{\theta}_{t,c}(\theta_0))\right) d\theta'}\right), \tag{3.14}$$

*where we defined the regularised parameter estimate of $\theta^\star$ by*

$$\widehat{\theta}_{t,c}(\theta_0) = (\nabla\mathcal{L})^{-1}\left(\frac{\sum_{s=1}^t F(Y_s) + c\nabla\mathcal{L}(\theta_0)}{t+c}\right). \tag{3.15}$$

**Parameter estimate.** To gain intuition, we observe that $\widehat{\theta}_{t,c}(\theta_0)$ is essentially a maximum a posteriori (MAP) estimate. Indeed, the likelihood of a sample $(Y_s)_{s=1}^t$ drawn from $p_\theta$ is given by $p\left((Y_s)_{s=1}^t \mid \theta\right) \propto \exp\left(\langle \theta, \sum_{s=1}^t F(Y_s)\rangle - t\mathcal{L}(\theta)\right)$. Now, given a reference point $\theta_0 \in \Theta$, consider the prior $p\left(\theta\right) \propto \exp\left(\langle \theta, \nabla\mathcal{L}(\theta_0) - c\mathcal{L}(\theta)\right)$, that involves the log-partition of the family at hand. Then the posterior over $\theta$ given the sample $(Y_s)_{s=1}^t$ takes the form

$$p\left(\theta \mid (Y_s)_{s=1}^t\right) \propto \exp\left(\left\langle \theta, \sum_{s=1}^t F(Y_s) + c\nabla L(\theta_0)\right\rangle - (t+c)\mathcal{L}(\theta)\right). \tag{3.16}$$

In particular, $p(\theta)$ is a conjugate prior in the Bayesian sense as the posterior distribution takes the same form as the prior distribution, with updated parameters. The MAP estimator $\widehat{\theta}_{t,c}^{\text{MAP}}(\theta_0) =$ $\operatorname{argmax}_{\theta\in\Theta} p\left(\theta \mid (Y_s)_{s=1}^t\right)$ can be derived by differentiating the exponentiated term (assuming mild regularity conditions as noted above in Section 1.3), resulting in $\widehat{\theta}_{t,c}^{\mathrm{MAP}}(\theta_0) = \widehat{\theta}_{t,c}(\theta_0)$.

To better interpret $c$, we examine the specific case of Bernoulli distributions, i.e. $\Theta = (0,1)$, $p_\theta(y) = \theta^y(1-\theta)^{1-y}$ for $y \in \{0,1\}$ (density with respect to the counting measure). The conjugate prior to $p_\theta$ is given by the $\mathrm{Beta}(a,b)$ distribution with $a, b \in \mathbb{R}_+^\star$, leading to the posterior distribution $\mathrm{Beta}(a + S_t, b + (t - S_t))$ with $S_t = \sum_{s=1}^t Y_t$. The MAP estimator is then $\widehat{\theta}_{t,a,b} = (S_t + a)/(a + b + t)$. For integer values of $a$ and $b$, this corresponds to the maximum likelihood estimator (MLE) after observing $a$ success and $b$ failures in addition to the $t$ Bernoulli trials $(Y_s)_{s=1}^t$. Anticipating on the sequel (see Appendix B.2), the regularised parameter estimate of Definition 3.2 with reference point $\theta_0 \in \Theta$ takes the form $\widehat{\theta}_{t,c}(\theta_0) = (S_t + c\theta_0)/(t+c)$, which corresponds to the MAP with parameters $a = c\theta_0$ and $b = c(1-\theta_0)$. Therefore, $c$ can be interpreted as a virtual number of prior samples drawn from the reference distribution $p_{\theta_0}$.

**Dependence on $\theta_0$ and example.** The acute reader can note that the considered parameter estimate $\widehat{\theta}_{t,c}(\theta_0)$ and Bregman information gain $\gamma_{t,c}(\theta_0)$ involve a reference parameter $\theta_0$. It makes sense to have such a local reference point since the Bregman divergence is typically linked to metrics with *local* (non-constant) curvature. Hence, the (information) geometry seen from the perspective of different points $\theta_0$ may vary, unlike in the Gaussian case. Specifically, for a Gaussian $\mathcal{N}(\mu, \sigma^2)$ with known variance $\sigma^2$, the Bregman information gain with respect to a reference point $\mu_0$ reads

$$\gamma_{t,c}^{\mathcal{N}}(\mu_0) = \frac{1}{2}\log\frac{2\pi\sigma^2}{c} - \frac{1}{2}\log\frac{2\pi\sigma^2}{t+c} = \frac{1}{2}\log\frac{t+c}{c}\,. \tag{3.17}$$

It is independent of the reference parameter $\mu_0$, which is a consequence of the fact that the Bregman divergence in this case is proportional to the squared Euclidean distance; in other words, Gaussian distributions with known variance exhibit invariant geometry. However, for other exponential families, the geometries are inherently different, and hence Bregman information gain depend explicitly on local reference points (see Section 3.3 for details). For reference, the classical Gaussian information gain, i.e. the mutual information between a prior $\mu \sim \mathcal{N}(\mu_0, \sigma^2)$ and the average of an i.i.d. sample $(Y_s)_{s=1}^t$ drawn from $\mathcal{N}(\mu, c\sigma^2)$ is $\frac{1}{2}\log((t+c)/c)$, which matches the Bregman information gain.

### Method of mixtures for generic exponential families

We now present the main result of this chapter — a time uniform confidence bound connecting the Bregman geometry of the exponential family with its Bregman information gain.

**Theorem 3.3** (Method of mixtures for generic exponential families). *Let $\delta \in (0,1]$ and $t \in \mathbb{N}$. Under the assumptions of Definition 3.2 and assuming the Bregman information gain is well-defined, consider the set*

$$\widehat{\Theta}^{\delta}_{t,c} = \left\{ \theta_0 \in \Theta : (t+c)\mathcal{B}_{\mathcal{L}}\left(\theta_0, \widehat{\theta}_{t,c}(\theta_0)\right) \leqslant \log\frac{1}{\delta} + \gamma_{t,c}(\theta_0) \right\}. \tag{3.18}$$

*Then $\left(\widehat{\Theta}^{\delta}_{t,c}\right)_{t\in\mathbb{N}}$ is a **time-uniform confidence sequence** at level $\delta$ for $\theta^\star$, i.e.*

$$\mathbb{P}_{\theta^\star}\left(\forall t \in \mathbb{N},\ \theta^\star \in \widehat{\Theta}^{\delta}_{t,c}\right) \geqslant 1-\delta\,, \tag{3.19}$$

*or equivalently, for any random time $\tau$ in $\mathbb{N}$,*

$$\mathbb{P}_{\theta^\star}\left(\theta^\star \in \widehat{\Theta}^{\delta}_{\tau,c}\right) \geqslant 1-\delta\,. \tag{3.20}$$

Note the implicit definition of the confidence set $\widehat{\Theta}^{\delta}_{t,c}$, where the parameter of interest $\theta_0$ appears in both arguments of the Bregman divergence, and also in the Bregman information gain $\gamma_{t,c}(\theta_0)$. Because of this, *computing* this confidence set from the equation in Theorem 3.3 may seem nontrivial at first glance. However, we show in Section 3.3 how these sets simplify for many classical families, revealing how the computation can be made efficiently. Moreover, we observe that these confidence sets are actually tighter than those of prior work, and as such are especially well suited to be used when $t$ is small (for large $t$, most methods produce essentially equivalent confidence sets).

*Proof of Theorem 3.3.* We adapt here the method of mixtures, leveraging properties of Bregman divergences recalled in Lemma 3.1 to build suitable exponential martingales and mixing measures. Let $(Y_t)_{t\in\mathbb{N}}$ denote an i.i.d. sequence drawn from $p_{\theta^\star} \in \mathcal{F}^{\mathrm{EF}}_{h,F,\mathcal{L}}$.

**Martingale and mixture martingale construction.** Let us note that $\mathbb{E}_{\theta^\star}[F(Y_1)] = \nabla\mathcal{L}(\theta^\star)$ and $\log \mathbb{E}_{\theta^\star}\left[\exp\langle\lambda, F(Y_1)\rangle\right] = \mathcal{L}(\theta^\star+\lambda) - \mathcal{L}(\theta^\star)$. Hence, we deduce that

$$\log \mathbb{E}_{\theta^\star}\left[\exp\langle\lambda, F(Y_1) - \mathbb{E}_{\theta^\star}[F(Y_1)]\rangle\right] = \mathcal{B}_{\mathcal{L},\theta^\star}(\lambda)\mathcal{L}(\theta^\star+\lambda) - \mathcal{L}(\theta^\star) - \langle\lambda, \nabla\mathcal{L}(\theta^\star)\rangle\,. \tag{3.21}$$

Now, for $t \in \mathbb{N}$, we let $\widehat{\mu}_t = \frac{1}{t}\sum_{s=1}^t F(Y_s)$ and $\mu = \mathbb{E}_{\theta^\star}[F(Y_1)]$. For any $\lambda \in \mathbb{R}^d$, the following quantity

$$M_t^\lambda = \exp\left(\langle \lambda, t(\widehat{\mu}_t - \mu)\rangle - t\mathcal{B}_{\mathcal{L},\theta^\star}(\lambda)\right) \tag{3.22}$$

thus defines a nonnegative martingale such that $\mathbb{E}_{\theta^\star}[M_t^\lambda] = 1$.

We now introduce the distribution $q(\theta|\alpha,\beta) = H(\alpha,\beta)\exp(\langle\theta,\alpha\rangle - \beta\mathcal{L}(\theta))$ over $\theta \in \Theta$, where $H$ is the normalisation term. We further introduce the following quantity

$$M_t = \int_{\Lambda_{\theta^\star}} M_t^\lambda q(\theta^\star + \lambda|\alpha,\beta)d\lambda\,, \tag{3.23}$$

where we recall that $\Lambda_{\theta^\star} = \{\lambda \in \mathbb{R}^d,\ \theta^\star + \lambda \in \Theta\}$. This also satisfies $\mathbb{E}_{\theta^\star}[M_t] = 1$. Further, we have the rewriting

$$M_t = H(\alpha,\beta)\int_{\Lambda_{\theta^\star}} \exp\left(\langle \lambda, t(\widehat{\mu}_t - \mu)\rangle - t\mathcal{B}_{\mathcal{L},\theta^\star}(\lambda) + \langle \lambda + \theta^\star, \alpha\rangle - \beta\mathcal{L}(\theta^\star + \lambda)\right)d\lambda\,. \tag{3.24}$$

**Choice of parameters and duality properties.** Setting $\alpha = c\nabla\mathcal{L}(\theta^\star)$ and $\beta = c$, we obtain

$$\begin{aligned} M_t &= H(c\nabla\mathcal{L}(\theta^\star), c)\int_{\Lambda_{\theta^\star}} \exp\left(\langle \lambda, t(\widehat{\mu}_t - \mu)\rangle - t\mathcal{B}_{\mathcal{L},\theta^\star}(\lambda) + c\langle \lambda + \theta^\star, \nabla\mathcal{L}(\theta^\star)\rangle - c\mathcal{L}(\theta^\star + \lambda)\right)d\lambda \\ &= G(\theta^\star, c)\int_{\Lambda_{\theta^\star}} \exp\left(\langle \lambda, t(\widehat{\mu}_t - \mu)\rangle - (t+c)\mathcal{B}_{\mathcal{L},\theta^\star}(\lambda)\right)d\lambda\,, \end{aligned} \tag{3.25}$$

where we introduce the following quantity

$$\begin{aligned} G(\theta^\star, c) = H(c\nabla\mathcal{L}(\theta^\star), c)\exp\left(c\langle\theta, \nabla\mathcal{L}(\theta^\star)\rangle - c\mathcal{L}(\theta^\star)\right) &= \frac{\exp\left(c\langle\theta^\star, \nabla\mathcal{L}(\theta^\star)\rangle - c\mathcal{L}(\theta^\star)\right)}{\int_\Theta \exp\left(c\langle\theta', \nabla\mathcal{L}(\theta^\star)\rangle - c\mathcal{L}(\theta')\right)d\theta'} \\ &= \left[\int_\Theta \exp\left(-c\mathcal{B}_{\mathcal{L}}(\theta', \theta^\star)\right)d\theta'\right]^{-1}. \end{aligned} \tag{3.26}$$

At this point, let us introduce $x = \frac{t}{t+c}(\widehat{\mu}_t - \mu)$. We also consider the Legendre-Fenchel dual function $\mathcal{B}^\star_{\mathcal{L},\theta}(x) = \max_\lambda \langle\lambda, x\rangle - \mathcal{B}_{\mathcal{L},\theta^\star}(\lambda)$ and denote $\lambda_x^\star$ its maximal point. Now, we note that $x + \nabla\mathcal{L}(\theta^\star) = \nabla\mathcal{L}(\theta^\star + \lambda_x^\star)$ and thus

$$\begin{aligned} \langle\lambda, x\rangle - \mathcal{B}_{\mathcal{L},\theta^\star}(\lambda) - \langle\lambda_x^\star, x\rangle + \mathcal{B}_{\mathcal{L},\theta^\star}(\lambda_x^\star) &= \langle\lambda - \lambda_x^\star, x + \nabla\mathcal{L}(\theta^\star)\rangle + \mathcal{L}(\theta^\star + \lambda_x^\star) - \mathcal{L}(\theta^\star + \lambda) \\ &= \langle\lambda - \lambda_x^\star, \nabla\mathcal{L}(\theta^\star + \lambda_x^\star)\rangle + \mathcal{L}(\theta^\star + \lambda_x^\star) - \mathcal{L}(\theta^\star + \lambda)\,. \end{aligned} \tag{3.27}$$

Also, since we have the equality $x = \frac{t}{t+c}(\widehat{\mu}_t - \nabla\mathcal{L}(\theta^\star))$, the quantity $x + \nabla\mathcal{L}(\theta^\star) = \nabla\mathcal{L}(\theta^\star + \lambda_x^\star)$ rewrites $\frac{t}{t+c}\widehat{\mu}_t + \frac{c}{t+c}\nabla\mathcal{L}(\theta^\star) = \nabla\mathcal{L}(\theta^\star + \lambda_x^\star)$, which justifies to introduce the following regularised

parameter estimate:

$$\widehat{\theta}_{t,c}(\theta^\star) = (\nabla\mathcal{L})^{-1}\left(\frac{t}{t+c}\widehat{\mu}_t + \frac{c}{t+c}\nabla\mathcal{L}(\theta^\star)\right). \tag{3.28}$$

**Martingale rewriting and conclusion.** Using this property, we note that

$$\begin{aligned}
M_t &= \exp\left((t+c)\mathcal{B}^\star_{\mathcal{L},\theta^\star}(x)\right) G(\theta^\star, c) \\
&\quad\times \int_{\Lambda_{\theta^\star}} \exp\left((t+c)\left[\langle \lambda - \lambda^\star_x, \nabla\mathcal{L}(\theta^\star+\lambda^\star_x)\rangle - \mathcal{L}(\theta^\star+\lambda) + \mathcal{L}(\theta^\star+\lambda^\star_x)\right]\right) d\lambda \\
&= \exp\left((t+c)\mathcal{B}^\star_{\mathcal{L},\theta^\star}(x)\right) G(\theta^\star, c) \\
&\quad\times \int_{\Lambda_{\theta^\star}} \exp\left(\langle \lambda + \theta^\star, (t+c)\nabla\mathcal{L}(\theta^\star+\lambda^\star_x)\rangle - (t+c)\mathcal{L}(\theta^\star+\lambda)\right) d\lambda \\
&\quad\times \exp\left(-(t+c)\langle \lambda^\star_x + \theta^\star, \nabla\mathcal{L}(\theta^\star+\lambda^\star_x) + (t+c)\mathcal{L}(\theta^\star+\lambda^\star_x)\rangle\right) \\
&= \exp\left((t+c)\mathcal{B}^\star_{\mathcal{L},\theta^\star}(x)\right)\frac{G(\theta^\star, c)}{G(\theta^\star+\lambda^\star_x, t+c)} = \exp\left((t+c)\mathcal{B}^\star_{\mathcal{L},\theta^\star}(x)\right)\frac{G(\theta^\star, c)}{G(\widehat{\theta}_{t,c}(\theta^\star), n+c)}.
\end{aligned} \tag{3.29}$$

Using Theorem 1.20, we obtain the following inequality for any random time $\tau$ in $\mathbb{N}$:

$$\forall \delta \in (0,1],\ \mathbb{P}_{\theta^\star}\left(\mathcal{B}^\star_{\mathcal{L},\theta^\star}\left(\frac{\tau}{\tau+c}(\widehat{\mu}_\tau - \mu)\right) \geqslant \frac{1}{\tau+c}\log\left(\frac{G(\widehat{\theta}_{\tau,c}(\theta^\star), \tau+c)}{G(\theta^\star, c)}\frac{1}{\delta}\right)\right) \leqslant \delta. \tag{3.30}$$

To conclude, we use the duality property of the Bregman divergence (Lemma 3.1), considering that $\nabla\mathcal{L}$ is invertible. Indeed, letting $\alpha = \tau/(\tau+c)$, and $\widehat{\theta}_\tau = (\nabla\mathcal{L})^{-1}(\widehat{\mu}_\tau)$, we obtain

$$\mathcal{B}^\star_{\mathcal{L},\theta^\star}\left(\alpha(\widehat{\mu}_\tau - \mu)\right) = \mathcal{B}_{\mathcal{L}}\left(\theta^\star, \theta_\alpha\right) \tag{3.31}$$

where $\theta_\alpha$ satisfies the equation

$$\theta_\alpha = \nabla\mathcal{L}^{-1}(\alpha\nabla\mathcal{L}(\theta_\tau) + (1-\alpha)\nabla\mathcal{L}(\theta^\star)) = \widehat{\theta}_{\tau,c}(\theta^\star). \tag{3.32}$$

We conclude by the equivalence between time-uniform, stopping times and random times concentration (Remark 1.18). ■

Interestingly, this proof shows that the parameter $c$, introduced as a local regularisation term in Definition 3.2 and interpreted as a virtual number of prior samples, is also a mixing parameter in the method of mixtures, in the same spirit as $\alpha$ in the sub-Gaussian setting (Corollary 1.21).

**Remark 3.4**. *We provide in Appendix B.1 a complementary result (Corollary B.2) using a Legendre function $\mathcal{L}_0$ instead of the mixing parameter $c$, which eschews the use of a local reference $\theta_0$. However, we argue that such a global regularisation is actually less convenient to use except for univariate and multivariate Gaussian distributions that anyway exhibit invariant geometry. Furthermore, in Theorem B.1, we prove a more general result that handles the case of a sequence $(Y_t)_{t\in\mathbb{N}}$ of random variables that are not independent, and having possibly different distributions from each other, which we apply to build confidence sets in linear bandits (see Appendix B.1).*

**Comparison with prior work.** Similar to Theorem 3.3, Kaufmann and Koolen (2021) extend the method of mixtures technique to derive time-uniform concentration bounds for exponential families. However, their proof technique is fairly different, relying instead on discrete mixtures and stitching, which only works for single parameter families, and involves case-specific calculations that are difficult to generalise beyond Gaussian (with known variance) and Gamma (with known shape). In contrast, our method applies to generic exponential families, including distributions with more than one parameter such as Gaussian when both mean and variance are unknown. Moreover, their discrete prior construction leads to technical constants seemingly unrelated to the exponential family model. Our prior is naturally induced by the exponential family, leveraging key properties of Bregman divergences, yields more intrinsic quantities (Bregman information gain) and perhaps a more elegant and shorter proof. Hence, our results are not only more general, but also of fundamental interest.

**Asymptotic behavior.** The asymptotic width of $\widehat{\Theta}^{\delta}_{t,c}$ depends on the behavior of $\gamma_{t,c}(\theta_0)$ as $t \to +\infty$. Standard arguments show that $\widehat{\theta}_{t,c}(\theta_0) \to \theta^\star$ and the Taylor expansion of the mapping $\theta' \mapsto \mathcal{B}_{\mathcal{L}}(\theta', \theta^\star)$ around $\theta' = \theta^\star$ is $\mathcal{B}_{\mathcal{L}}(\theta', \theta^\star) = \frac{1}{2}(\theta' - \theta^\star)^\top \nabla^2 \mathcal{L}(\theta^\star)(\theta' - \theta^\star) + o\left(\|\theta' - \theta^\star\|^2\right)$ ($\nabla^2\mathcal{L}(\theta^\star)$ is positive definite by assumption). Laplace's method for integrals (chapter 20 in Lattimore and Szepesvári (2020), Shun and McCullagh (1995)) gives the following estimate:

$$\int_\Theta \exp\left(-(t+c)\mathcal{B}_{\mathcal{L}}(\theta', \widehat{\theta}_{t,c}(\theta_0))\right) d\theta' \approx \int_{\mathbb{R}^d} \exp\left(-\frac{t+c}{2}(\theta^\star - \theta')^\top \nabla^2 \mathcal{L}(\theta^\star)(\theta^\star - \theta')\right) d\theta' \,, \tag{3.33}$$

which is a simple Gaussian integral, and thus $\gamma_{t,c}(\theta_0) = \frac{d}{2}\log(1 + t/c) + \mathcal{O}(1)$. This asymptotic scaling is worse than the $\log\log t$ rate prescribed by the law of iterated logarithm. However, this is a standard feature of the method of mixtures (see Section 1.4 in Chapter 1), which is compensated by its improved nonasymptotic sharpness, as evidenced in Section 3.4.

**Dependency on the parameter $c$.** In the Gaussian case, the parameter $c$ corresponds to the mixing parameter $\alpha$ of the sub-Gaussian method of mixtures (Corollary 1.21), and as such can be tuned to $c = \gamma_\delta t_0$ to tighten the confidence bound around $t = t_0$, with e.g. $\gamma_\delta \approx 0.12$ for $\delta = 0.05$ (Lemma 1.23). It is however insightful to study its influence for other families. We provide in Section 3.4 a detailed study of this parameter, concluding that the confidence bounds are not significantly altered over a large range of values, and that $c = \gamma_\delta t_0$ remains a valid heuristic for non-Gaussian families. Note that similarly to the sub-Gaussian method of mixtures, choosing a variable $c$ changing with $t$ is not allowed by the theory, as this would break the martingale property used in the proof and would be incompatible with the time-uniform lower bound $|\widehat{\Theta}^\delta_{t,c}| = \Omega(\sqrt{\log\log t / t})$ provided by the law of iterated logarithm (see Section 1.4 in Chapter 1). Indeed, the Bregman information gain with $c_t \propto t$ would be asymptotically $\gamma_{t,c_t}(\theta_0) = \frac{d}{2}\log(1 + \mathcal{O}(1)) + \mathcal{O}(1) = \mathcal{O}(1)$, leading to $|\widehat{\Theta}^\delta_{t,c}| = \mathcal{O}(1/\sqrt{t})$ when $t \to +\infty$.

### Application to bandits

One can apply the technique developed in Theorem 3.3 to build confidence sets in standard multiarmed bandit problems. Indeed, consider the bandit model $\boldsymbol{\nu} = \bigotimes_{k\in[K]} p_{\theta^\star_k} \in \bigotimes_{k\in[K]} \mathcal{F}_k$, where $\mathcal{F}_k$ is a generic exponential family (e.g. Gaussian) for each arm $k \in [K]$ and $(\theta^\star_k)_{k\in[K]} \in \prod_{k\in[K]} \Theta_k \subseteq \mathbb{R}^K$. As seen in Chapter 1, this setting is standard to analyse optimal bandit algorithms (Cappé et al., 2013; Korda et al., 2013; Baudry et al., 2020). We recall that, at each time $t \in \mathbb{N}$, we choose an arm $\pi_t \in [K]$ based on past observations, and draw a sample reward $Y^{\pi_t}_t$ from its distribution $p_{\theta^\star_{\pi_t}}$. The samples are then used to update knowledge about the parameters $(\theta^\star_k)_{k\in[K]}$ by building confidence sets for each of them. Let $N^k_t = \sum_{s=1}^t \mathbb{1}_{\pi_s = k}$ be the number of pulls to arm $k$ up to time $t$, and $\widehat{\theta}_{k,N^k_t,c}(\theta_0)$ and $\gamma_{k,N^k_t,c}(\theta_0)$ be the parameter estimate and Bregman information gain for arm $k$, respectively (similarly to Definition 3.2 with $t$ replaced by $N^k_t$ and samples taken from $p_{\theta^\star_k}$ — we recall that this construction is well-defined thanks to Doob's optional skipping). We construct the confidence set for arm $k$ at time $t$ as

$$\widehat{\Theta}^\delta_{k,N^k_t,c} = \left\{ \theta_0 \in \Theta_k,\ (N^k_t + c)\mathcal{B}_\mathcal{L}\left(\theta_0, \widehat{\theta}_{k,N^k_t,c}(\theta_0)\right) \leqslant \log\frac{1}{\delta} + \gamma_{k,N^k_t,c}(\theta_0) \right\}. \tag{3.34}$$

Then, similarly to Theorem 3.3, the true parameter $\theta^\star_k$ lies in the set $\widehat{\Theta}^\delta_{k,N^k_t,c}$ for all time steps $t \in \mathbb{N}$ with probability at least $1-\delta$. Finally, we take a union bound over $k \in [K]$ to obtain confidence sets for all arms (with widths inflated by an additive $\log K$ factor). Such construction is standard and is used in Abbasi-Yadkori et al. (2011) for (sub)-Gaussian families. Possible applications include UCB algorithms for regret minimisation (see Section 1.2 in Chapter 1) and designing GLR stopping rules for tracking algorithms in pure exploration (Garivier and Kaufmann, 2016) in the context of generic exponential families (see also Section 3.5 for another application of GLR tests using the parameter estimate $\widehat{\theta}_{t,c}(\theta_0)$). For regret minimisation specifically, we extend the generic UCB framework of Theorem 1.33 to generic exponential families.

**Proposition 3.5** (EF-Bregman-UCB). *Let $\Theta \subseteq \mathbb{R}^d$, $K \in \mathbb{N}$, $(\theta_k)_{k\in[K]} \in \Theta^K$ and a bandit model $(\bigotimes_{k\in[K]} p_{\theta_k}, (\mathcal{F}^{EF}_{h,F,\mathcal{L}})^{\otimes K})$. For a given mapping $\varphi\colon \mathbb{R}^d \to \mathbb{R}$, we consider the risk measure on $\mathcal{F}^{EF}_{h,F,\mathcal{L}}$ defined as $\rho\colon \theta \in \Theta \mapsto \varphi(\mathbb{E}_\theta[F(Y)]) = \varphi(\nabla\mathcal{L}(\theta))$. Assume that*

*(i) the curvature of the log-partition function is uniformly bounded, i.e. there exists $m, M \in \mathbb{R}^\star_+$ such that $mI_d \preccurlyeq \nabla^2\mathcal{L}(\theta) \preccurlyeq MI_d$ for all $\theta \in \Theta$;*

*(ii) $\varphi$ is L-Lipschitz with respect to the Euclidean norm $\|\cdot\|$.*

*Let $c \in \mathbb{R}^\star_+$. Then for $T \to +\infty$, Algorithm 1 with the risk measure $\rho$ and UCB given by $(\rho(\widehat{\Theta}^\delta_{k,n,c}))_{n\in\mathbb{N}}$ satisfies, for all suboptimal arm $k \in [K] \setminus \{k^\star\}$,*

$$\mathbb{E}_{\pi,\nu_\pi^{\otimes T}}\left[N_T^k\right] \leqslant \frac{2L^2M^2}{m\Delta_k^2}\log\left(\frac{4LMT}{\sqrt{m}\Delta_k}\right) + o\left(\frac{L^2M^2}{m\Delta_k^2}\log T\right), \tag{3.35}$$

*where $\Delta_k$ is the suboptimality gap for arm $k$.*

This shows that Algorithm 1 provides an example of an order-optimal deterministic algorithm for generic exponential families. The proof is deferred to Appendix B.4.

For the family $\mathcal{F}^{\mathcal{N},\sigma}$ of Gaussian with known variance $\sigma^2 \in \mathbb{R}^\star_+$, the log-partition function is quadratic, i.e. $\mathcal{L}(\theta) = \theta^2/(2\sigma^2)$, and thus has a constant curvature $\nabla^2\mathcal{L}(\theta) = 1/\sigma^2$ for all $\theta \in \mathbb{R}$. The expectation risk measure corresponds to $\varphi\colon \eta \in \mathbb{R} \mapsto 2\sigma^2\eta$, and therefore the proposition above applies with $m = M = 1/\sigma^2$ and $L = 2\sigma^2$. Consequently, we recover the classical (sub-)Gaussian UCB regret. Beyond the SPEF case, this paves a way to analyse Gaussian bandits with unknown variance under risk measures that depend on both mean and variance.

**Remark 3.6**. *The regret bound of Proposition 3.5 is by no mean optimal and is intended to illustrate applications of the time-uniform confidence sequence of Theorem 3.3. We anticipate that a more refined algorithm and analysis in the spirit of kl-UCB (Cappé et al., 2013) could yield asymptotically optimal UCB algorithms for generic exponential families.*

☞ *In Chapter 5, we analyse a more general framework for contextual bandits under a vast class of risk measures, with similar curvature assumptions as in Proposition 3.5.*

## 3.3 Specification to classical families

In this section, we specify the result of Theorem 3.3 to some classical exponential families. Interestingly, the literature on time-uniform concentration bounds outside of random variables that are bounded, gamma with fixed shape or Gaussian with known variance is significantly scarce, even though many more distributions are commonly used in machine learning models. We derive below explicit confidence sets for a range of distributions, which we believe will be of interest for the wider machine learning and statistics community. For instance, consider active learning in bandit problems (Carpentier et al., 2011), where one targets upper confidence bounds on the variance (rather than the mean); in the Gaussian case, this can be achieved with Chi-square concentration. Hao et al. (2019) studies the classical UCB algorithm for bandits under a weaker assumption that sub-Gaussianity, involving the Weibull concentration. In differential privacy, concentration of Laplace distribution (symmetrised exponential) is often used to study the utility of differentially private mechanisms (Dwork et al., 2014). Finally, heavy-tailed distributions such as Pareto have recently been of interest to study risk-averse or corruption in bandit problems (Holland and Haress, 2021; Basu et al., 2022).

We now make explicit the Bregman information gains and confidence sets for some illustrative families (more examples and full derivations are provided in Appendix B.2).

**Gaussian (unknown mean and variance).** Let $Y \sim \mathcal{N}(\mu, \sigma^2)$. Given samples $(Y_s)_{s=1}^t$, we define $S_t = \sum_{s=1}^t Y_s$ and $\widehat{\mu}_t = S_t/t$. Further, for $\mu, \sigma \in \mathbb{R} \times \mathbb{R}^+$, we define the normalised sum of squares $Z_t(\mu, \sigma) = \frac{1}{\sigma^2}\sum_{s=1}^t (Y_s - \mu)^2$. Then, for reference parameters $\mu_0 \in \mathbb{R}, \sigma_0 \in \mathbb{R}^+, c > 0$, the Bregman information gain reads

$$\gamma_{t,c}^{\mathcal{N}}(\mu_0, \sigma_0) = \frac{3}{2}\log\left(\frac{t}{t+c} Z_t(\widehat{\mu}_t, \sigma_0) + \frac{c}{t+c} Z_t(\mu_0, \sigma_0) + c\right) + f_{t,c}\,, \tag{3.36}$$

where $f_{t,c} = \left(\frac{t+c+1}{2}\right)\log(t+c) - \left(\frac{c}{2}+2\right)\log c - \frac{t}{2}(1+\log 2) + \log\Gamma\left(\frac{c+3}{2}\right) - \log\Gamma\left(\frac{t+c+3}{2}\right)$.

**Bernoulli.** Let $Y \sim \mathcal{B}(\mu)$, with unknown mean $\mu \in [0,1]$. Define, for reference parameter $\mu_0 \in \mathbb{R}$, $c > 0$, the estimate $\mu_{t,c}(\mu_0) = \frac{S_t + c\mu_0}{t+c}$. Then, the Bregman information gain is given by

$$\gamma_{t,c}^{\mathcal{B}}(\mu_0) = c\,\mathbb{H}^{\mathcal{B}}(\mu_0) - (t+c)\mathbb{H}^{\mathcal{B}}(\mu_{t,c}(\mu_0)) + \log\frac{\mathbf{B}(c\mu_0, c(1-\mu_0))}{\mathbf{B}((t+c)\mu_{t,c}(\mu_0), (t+c)(1-\mu_{t,c}(\mu_0))}, \tag{3.37}$$

where $\mathbb{H}^{\mathcal{B}}(\mu) = -(1-\mu)\log(1-\mu) - \mu\log\mu$ denotes the Bernoulli entropy function and $\mathbf{B}(\alpha, \beta) = \int_0^1 u^{\alpha-1}(1-u)^{\beta-1}du$ is the Beta function.

**Exponential.** Let $Y \sim \mathcal{E}(1/\mu)$, with unknown mean $\mu > 0$. For reference parameters $\mu_0, c > 0$, the Bregman information gain reads

$$\gamma^{\mathcal{E}}_{t,c}(\mu_0) = \log\left(S_t/\mu_0 + c\right) + \log\frac{\Gamma(c)}{\Gamma(t+c)} + (t+c-1)\log(t+c) - c\log c - t\,. \tag{3.38}$$

**Pareto.** Let $Y \sim \text{Pareto}\,(\alpha)$, with unknown shape $\alpha > 0$. Define $L_t = \sum_{s=1}^{t}\log Y_s$. Then, for reference parameters $\alpha_0, c > 0$, the Bregman information gain is given by

$$\gamma^{\text{Pareto}}_{t,c}(\alpha_0) = \log\left(\alpha_0 L_t + c\right) + \log\frac{\Gamma(c)}{\Gamma(t+c)} + (t+c-1)\log(t+c) - c\log c - t\,. \tag{3.39}$$

**Poisson.** Let $Y \sim \mathcal{P}(\lambda)$, where $\lambda > 0$. For reference parameters $\lambda_0, c > 0$, the Bregman information gain can be expressed as

$$\begin{aligned}\gamma^{\mathcal{P}}_{t,c}(\lambda_0) &= (S_t + c\lambda_0)\log\frac{S_t + c\lambda_0}{t+c} - S_t - c\lambda_0\log\lambda_0 \\ &\quad + \log I\left(c, c\lambda_0\right) - \log I\left(t+c, S_t + c\lambda_0\right),\end{aligned} \tag{3.40}$$

where $I(a,b) = \int_{-\infty}^{+\infty} e^{-ae^{\theta}+b\theta}d\theta$. Although, to the best of our knowledge, the integral $I(a,b)$ does not have a closed-form expression, it can be numerically estimated up to arbitrary precision.

**Chi-square.** Let $Y \sim \chi^2(k)$, where $k \in \mathbb{N}$ is unknown. Define $K_t = \sum_{s=1}^{t}\log\frac{Y_s}{2}$. For reference points $k_0 \in \mathbb{N}, c > 0$, introduce an estimate $k_{t,c}(k_0)$ satisfying $\psi_0\left(\frac{k_{t,c}(k_0)}{2}\right) = \frac{K_t + c\,\psi_0\left(\frac{k_0}{2}\right)}{t+c}$. In this case, the Bregman information gain is given by

$$\begin{aligned}\gamma^{\chi^2}_{t,c}(k_0) &= \frac{k_{t,c}(k_0)}{2}\left(K_t + c\psi_0\left(\frac{k_0}{2}\right)\right) - (t+c)\log\Gamma\left(\frac{k_{t,c}(k_0)}{2}\right) + c\log\Gamma\left(\frac{k_0}{2}\right) - c\,\frac{k_0}{2}\psi_0\left(\frac{k_0}{2}\right) \\ &\quad + \log\frac{J(c, c\psi_0(\frac{k_0}{2}))}{J(t+c, K_t + c\psi_0(\frac{k_0}{2}))}\,.\end{aligned} \tag{3.41}$$

where $J(a,b) = \sum_{k'=1}^{\infty}\exp\left(-a\log\Gamma\left(\frac{k'}{2}\right) + b\frac{k'}{2}\right)$. This integral can be estimated using numerical methods (see Section 3.4). This paves a way to build high-probability confidence sets for chi-square distributions which have not been adequately captured in prior work. The sums over $k'$ above derive from the martingale construction of Theorem 3.3 with discrete mixture (i.e. with respect to the counting measure). A continuous (i.e. with respect to the Lebesgue measure) mixture is also possible, as detailed in Remark B.3 in the appendix.

**Explicit confidence sets.** We now illustrate the confidence sets for the exponential families above, obtained by specifying the generic form and simplifying the resulting expression, in

Table 3.1. The technical details of the derivation of the specific forms for each illustrative family is postponed to Appendix B.2. In particular, we note that the confidence sequence with regularisation parameter $c$ for the SPEF of Gaussian distributions with fixed variance $\sigma^2$ is equivalent to the time-uniform bound of Corollary 1.21 for $R$-sub-Gaussian distributions (with $R = \sigma$ and $\alpha = c$).

**Table 3.1** – Summary of Bregman confidence sets given by Theorem 3.3 for representative families. Throughout, the following notations are used:
$S_t = \sum_{s=1}^{t} Y_s$, $\widehat{\mu}_t = \frac{S_t}{t}$, $Q_t(\mu) = \sum_{s=1}^{t} (Y_s - \mu)^2$, $Z_t(\mu,\sigma) = \frac{Q_t(\mu)}{\sigma^2}$,
$S_t^{(k)} = \sum_{s=1}^{t} Y_s^k$, $L_t = \sum_{s=1}^{t} \log Y_s$, $K_t = \sum_{s=1}^{t} \log \frac{Y_s}{2}$,
$I(a,b) = \int_{-\infty}^{+\infty} e^{-ae^\theta + b\theta} d\theta$,
$J(a,b) = \int_0^\infty \exp\left(-a \log\Gamma\left(\frac{k}{2}\right) + \frac{bk}{2}\right) dk$ ($dk$ is the Lebesgue measure if $k \in \mathbb{R}_+$ and the counting measure if $k \in \mathbb{N}$).

| Name | Parameters | Formula |
|---|---|---|
| Gaussian | $\mu \in \mathbb{R}$ | $\frac{1}{t+c}\frac{(S_t - t\mu)^2}{2\sigma^2} \leqslant \log\frac{1}{\delta} + \frac{1}{2}\log\frac{t+c}{c}$ |
| Gaussian | $\sigma \in \mathbb{R}_+$ | $\frac{Q_t(\mu)}{2\sigma^2} - \left(\frac{t+c}{2}+1\right)\log\left(\frac{Q_t(\mu)+c}{2\sigma^2}\right)$ $\leqslant \log\frac{1}{\delta} + \log\frac{\Gamma\left(\frac{c}{2}+2\right)}{\Gamma\left(\frac{t+c}{2}+2\right)} - \frac{t}{2}\log 2 - \left(\frac{c}{2}+1\right)\log c + \left(\frac{t+c}{2}+1\right)\log(t+c)$ |
| Gaussian | $\mu \in \mathbb{R}$<br>$\sigma \in \mathbb{R}_+$ | $\frac{1}{2}Z_t(\mu,\sigma) - \frac{t+c+3}{2}\log\left(\frac{t}{t+c}Z_t(\widehat{\mu}_t,\sigma) + \frac{c}{t+c}Z_t(\mu,\sigma) + c\right)$ $\leqslant \log\frac{1}{\delta} - \frac{t}{2}\log 2 - \left(\frac{c}{2}+2\right)\log c + \frac{1}{2}\log(t+c)$ $+ \log\Gamma\left(\frac{c+3}{2}\right) - \log\Gamma\left(\frac{t+c+3}{2}\right)$ |
| Bernoulli | $\mu \in [0,1]$ | $S_t \log\frac{1}{\mu} + (t - S_t)\log\frac{1}{1-\mu} + \log\frac{\Gamma(S_t + c\mu)\Gamma(t - S_t + c(1-\mu))}{\Gamma(c\mu)\Gamma(c(1-\mu))}$ $\leqslant \log\frac{1}{\delta} + \log\frac{\Gamma(t+c)}{\Gamma(c)}$ |
| Exponential | $\mu \in \mathbb{R}_+$ | $\frac{S_t}{\mu} - (t+c+1)\log\left(\frac{S_t}{\mu} + c\right)$ $\leqslant \log\frac{1}{\delta} + \log\frac{\Gamma(c)}{\Gamma(t+c)} - \log(t+c) - c\log c$ |
| Gamma | $\lambda \in \mathbb{R}_+$ | $\frac{S_t}{\lambda} - ((t+c)k+1)\log\left(\frac{S_t}{\lambda} + ck\right)$ $\leqslant \log\frac{1}{\delta} + \log\frac{\Gamma(ck)}{\Gamma((t+c)k)} - \log((t+c)k) - ck\log ck$ |
| Weibull | $\lambda \in \mathbb{R}_+$ | $\frac{S_t^{(k)}}{\lambda^k} - (t+c+1)\log\left(\frac{S_t^{(k)}}{\lambda^k} + c\right)$ $\leqslant \log\frac{1}{\delta} + \log\frac{\Gamma(c)}{\Gamma(t+c)} - \log(t+c) - c\log c$ |
| Pareto | $\alpha \in \mathbb{R}$ | $\alpha L_n - (t+c+1)\log(\alpha L_t + c)$ $\leqslant \log\frac{1}{\delta} + \log\frac{\Gamma(c)}{\Gamma(t+c)} - \log(t+c) - c\log c$ |
| Poisson | $\lambda \in \mathbb{R}_+$ | $t\lambda - S_t\log\lambda$ $\leqslant \log\frac{1}{\delta} + \log I(c, c\lambda) - \log I(t+c, S_t + c\lambda)$ |
| Chi-square | $k \in \mathbb{N}$<br>or $k \in \mathbb{R}_+$ | $t\log\Gamma\left(\frac{k}{2}\right) - \frac{k}{2}K_t - \log J\left(c, c\psi_0\left(\frac{k}{2}\right)\right)$ $+ \log J\left(t+c, K_t t + c\psi_0\left(\frac{k}{2}\right)\right) \leqslant \log\frac{1}{\delta}$ |

We now provide a set of illustrative numerical experiments to display the confidence envelopes resulting from Theorem 3.3 in the case of classical exponential families. We plot for a given value of the confidence level $\delta$ and regularisation parameter $c$ the convex sets $t \mapsto \bigcap_{s \leqslant t} \widehat{\Theta}^{\delta}_{s,c}$ (running intersection, see Section 1.4 in Chapter 1). In dimension $d = 1$, we report the extremal points of these intervals, which we call the upper and lower envelopes respectively (taking the running intersection ensures that these are respectively nonincreasing and nondecreasing, as explained in Remark 1.22). We refer to Figure 3.1 for Pareto, Chi-square and Gaussian (with unknown $\mu, \sigma$) and Appendix B.3 for many other families. Because we exploit the Bregman geometry, our bounds capture a larger setting than typical mean estimation; for instance, we are able to concentrate around the exponent $\alpha$ of a Pareto distribution even when the distribution is not integrable ($\alpha < 1$). Furthermore, for two-parameter Gaussian, apart form being *anytime*, our confidence sets are convex and bounded in contrast to the one based on Chi-square quantiles with a crude union bound (see Appendix B.3).

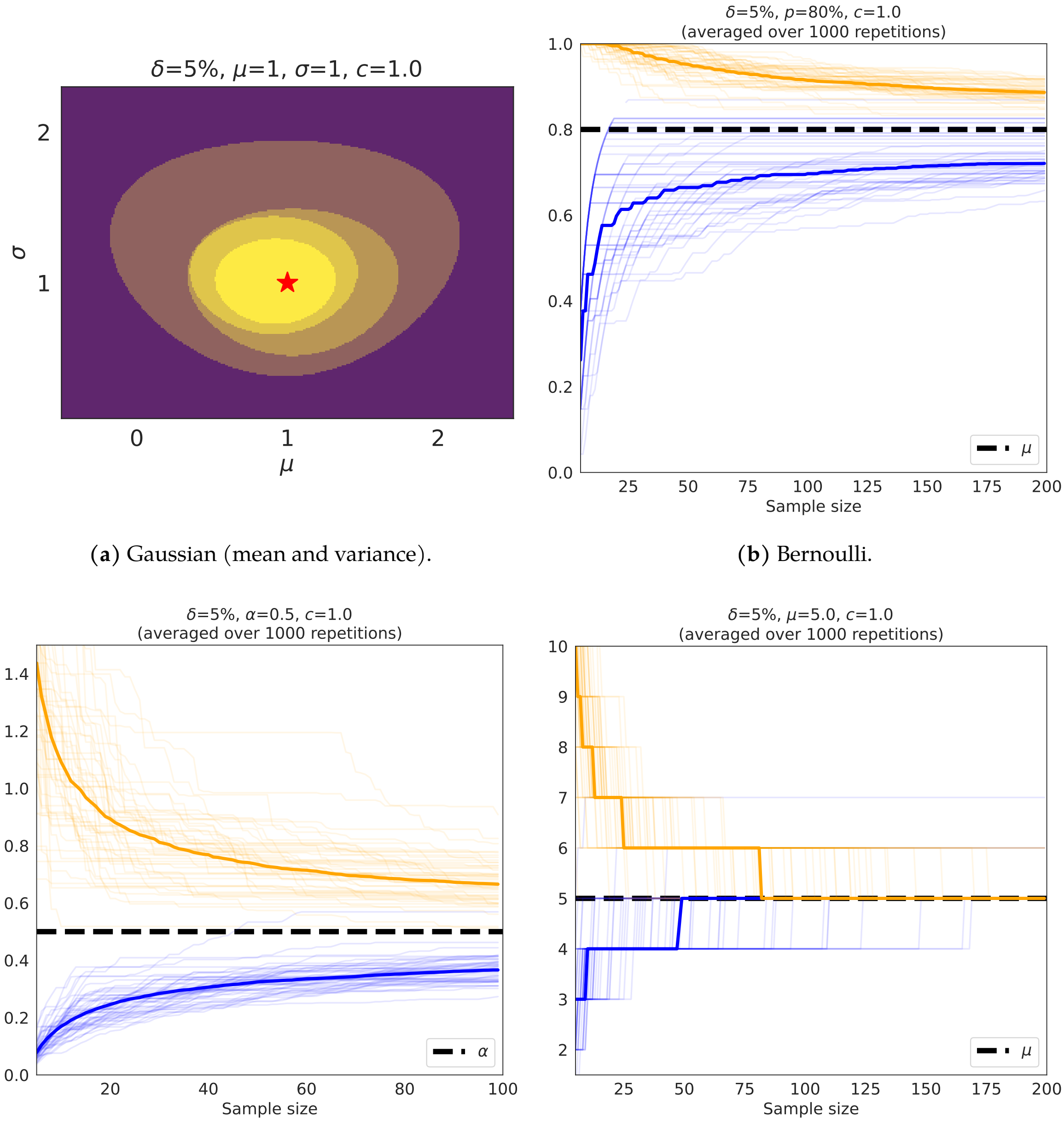

(**a**) Gaussian (mean and variance).

(**b**) Bernoulli.

(**c**) Pareto.

(**d**) Chi-square.

**Figure 3.1 –** Example of Gaussian time-uniform joint confidence sets for $(\mu, \sigma) = (1, 1)$ with sample sizes $t \in \{10, 25, 50, 100\}$ observations (smaller confidence sets correspond to larger sample sizes), and examples of time-uniform confidence envelopes for $\mathcal{B}(0.8)$, Pareto$(0.5)$ and $\chi^2(5)$ on several realisations as a function of the number of observations $t$. Thick lines indicate median curves over 1000 replicates.

## 3.4 Numerical experiments

In this section, we provide illustrative numerical results to show the resulting confidence bands built from Theorem 3.3 are promising alternatives to existing competitors (detailed in Appendix B.3). All the confidence sets presented here are implemented in the open source *concentration-lib* Python package. [1]

**Comparison.** We plot our confidence envelopes for Gaussian, exponential, Bernoulli and Chi-square distributions in Figure 3.2 and compare them to state-of-the-art time-uniform bounds, except for the Chi-square distribution, for which no prior benchmark existed; we report instead the confidence envelopes under two different mixture constructions (see Remark B.3 in Appendix B.2 for more details). Numerical experiments for other, less common distributions, with a similar scarcity of existing time-uniform bounds, are deferred to Appendix B.3. We fix $\delta = 0.05$ (note that all bounds exhibit the typical $\log 1/\delta$ dependency) and $c = 1$.

Interestingly, our upper and lower envelopes are not necessarily symmetric, and in the Bernoulli case, they fit within the supports of the distributions without clipping, contrary to most other methods, thus adapting to the local geometry of the family. Our bounds are comparable to those of Kaufmann and Koolen (2021), if slightly tighter; however, we emphasise again that our scope is wider and captures more distributions.

[1] https://pypi.org/project/concentration-lib.

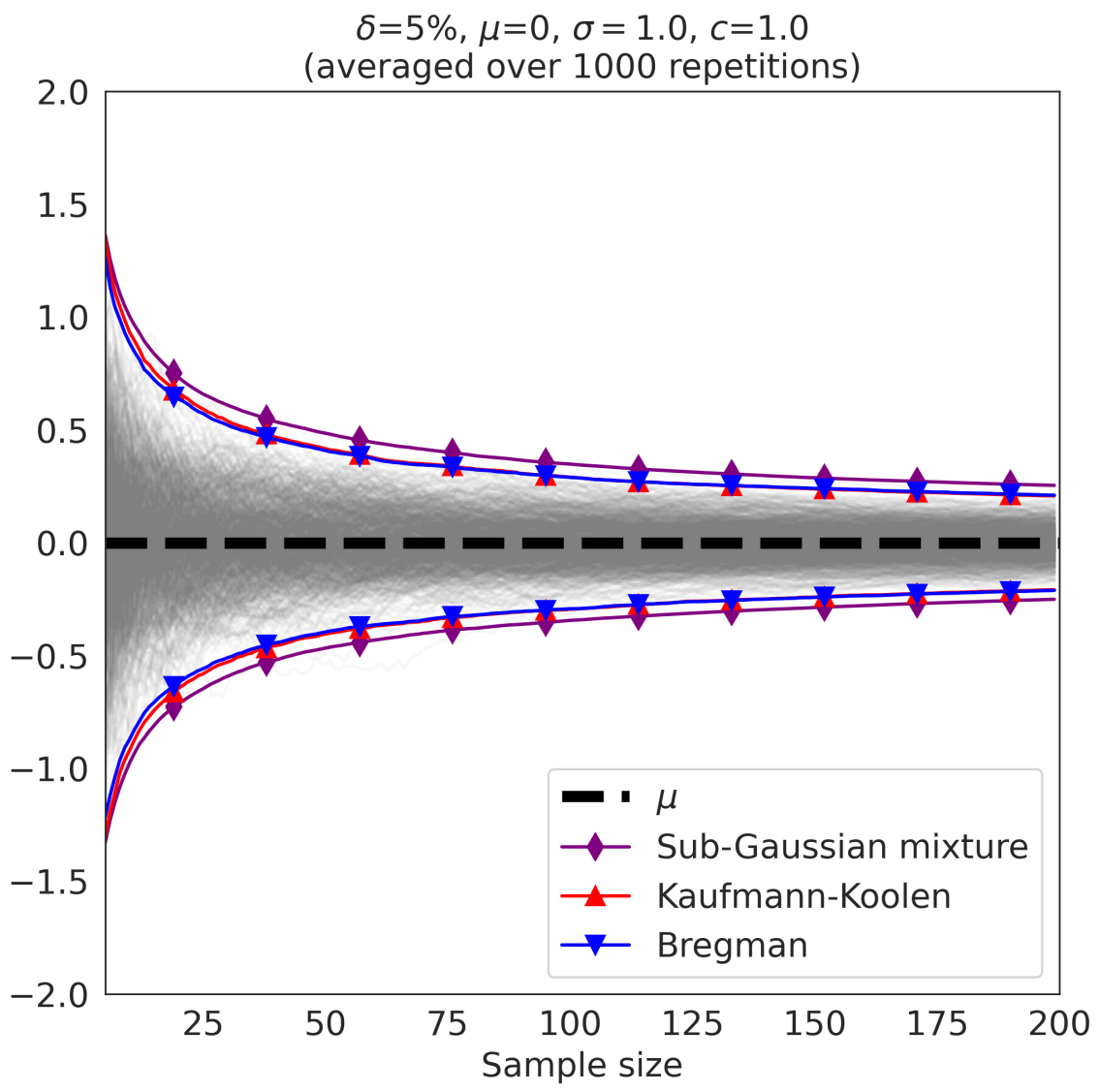

(**a**) Gaussian.

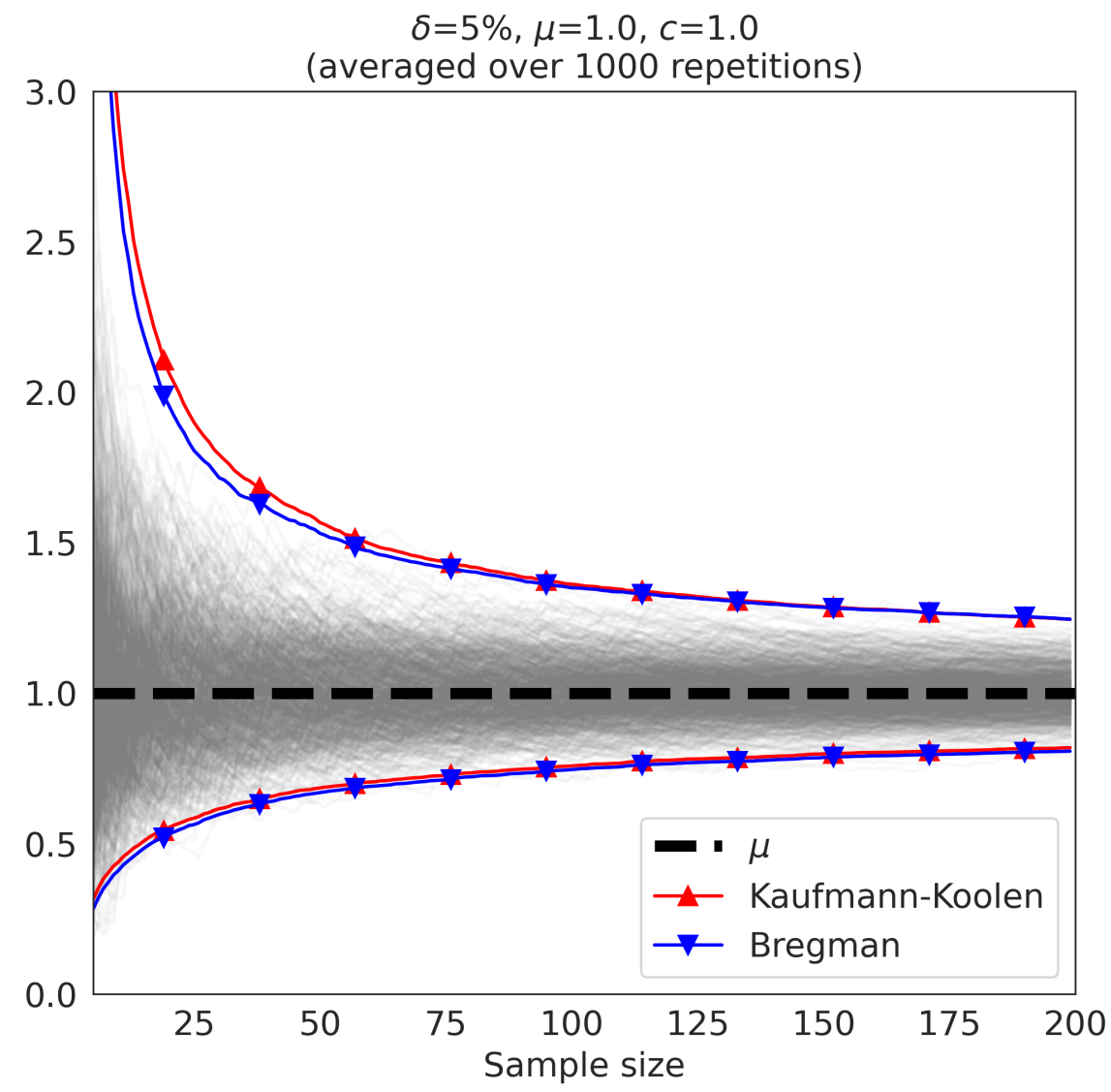

(**b**) Exponential.

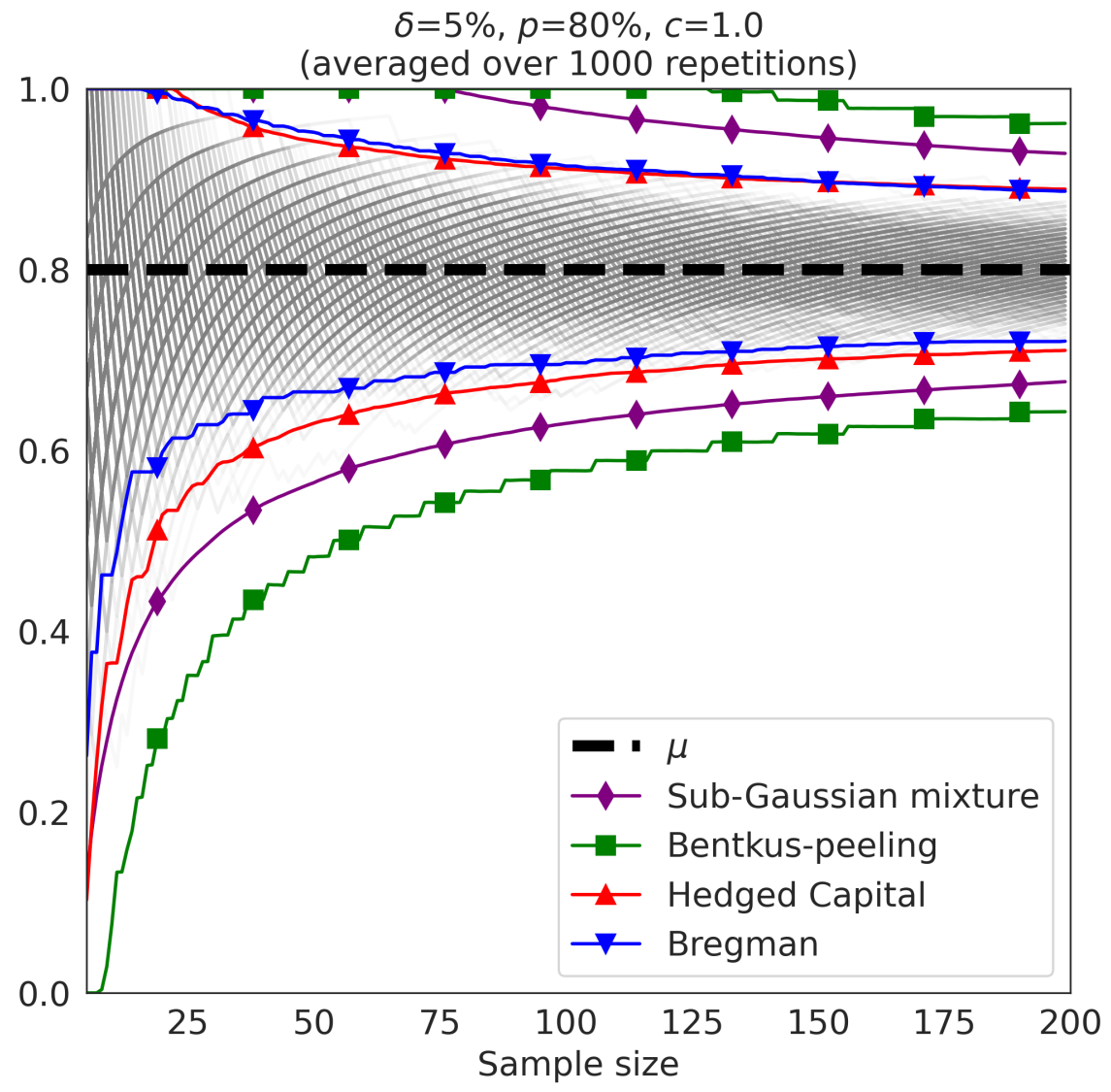

(**c**) Bernoulli.

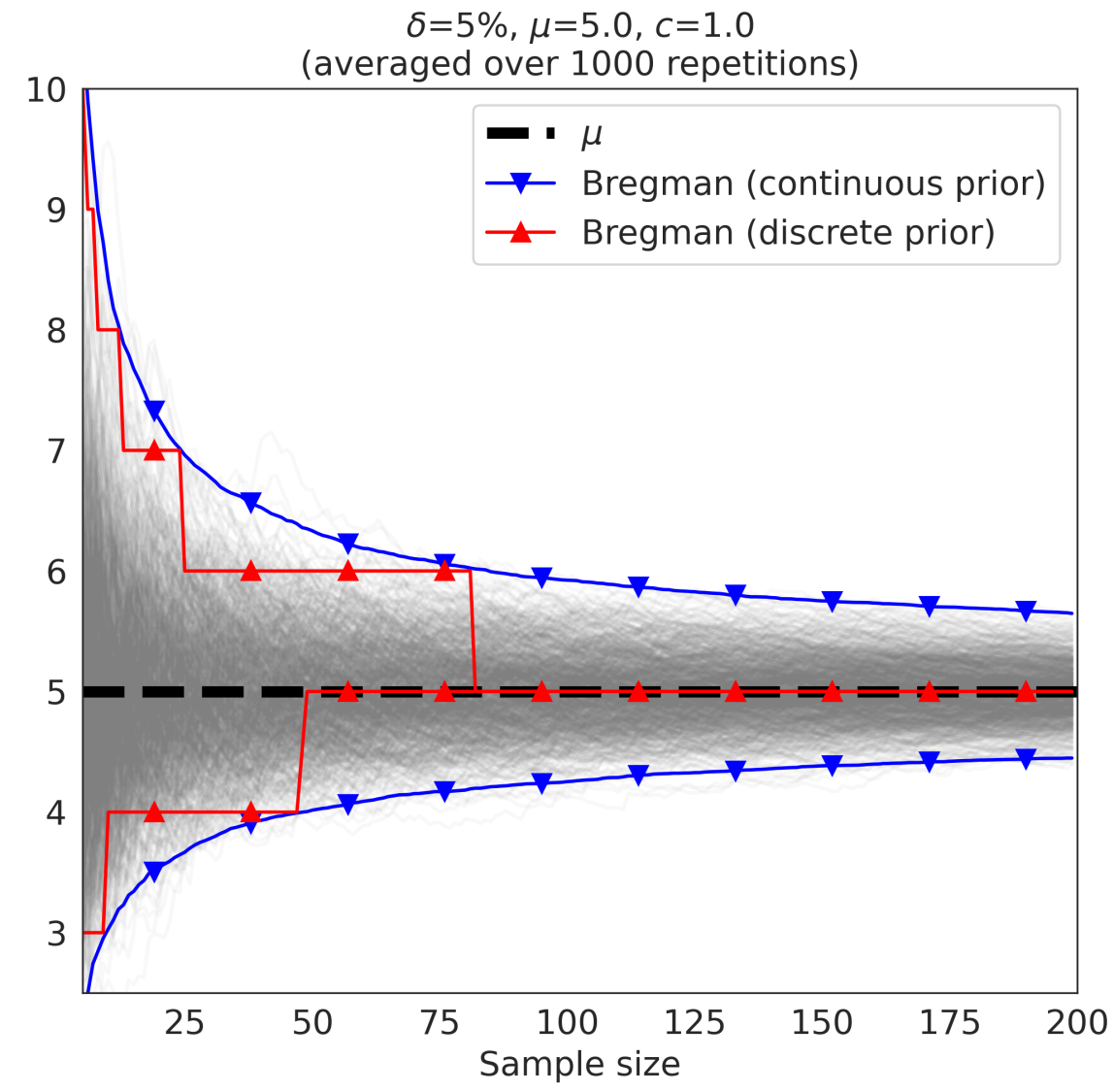

(**d**) Chi-square.

**Figure 3.2** – Comparison of median confidence envelopes around the mean for $\mathcal{N}(0,1)$, $\mathcal{E}(1)$, $\mathcal{B}(0.8)$ and $\chi^2(5)$, as a function of the sample size $t$, over 1000 independent replicates. Grey lines are trajectories of empirical means $\widehat{\mu}_t$.

**Numerical complexity.** Some of our formulas (Poisson, Chi-square) require the evaluation of integrals for which no closed-form expression exist. However, these can be estimated up to arbitrary precision by numerical routines, see Remarks B.4 and B.5 for details. Similarly, the use of special functions (digamma, ratio of Gamma) may lead to numerical instabilities or overflows for large $t$; we recommend instead to use the log-Gamma function and an efficient implementation of the log-sum-exp operator. We report results for $t \leqslant 200$ as we believe the small sample regime is where our bounds shine (most reasonable methods produce similar confidence sets for large $t$). Finally, most confidence sets reported in Figure 3.2 are implicitly defined (after reorganizing terms) as level sets $\{F(\theta) \leqslant 0\}$ for some function $F$. For one-dimensional families, we use a root search routine for a fast estimation of the boundaries of such sets; in higher dimension, we evaluate $F$ on a uniform grid (e.g. in Figure 3.1a, we use a $1024 \times 1024$ grid over $[-2, 4] \times [0.1, 4]$).

**Influence of the local regularisation parameter $c$.** We also performed additional experiments to get more insights on the tuning of $c$. Specifically, for varying values of $t_0 \in \mathbb{N}$, $c$ was chosen to minimise the width of the confidence interval at sample size $t = t_0$, i.e. $c_{t_0} = \min_{c>0} |\widehat{\Theta}_{t_0,c}(\delta)|$. We report in Figure 3.3 the resulting plots in the case of Bernoulli, Gaussian, exponential and Chi-square distributions. The resulting optimal choice $c_t^\star = \operatorname{argmin}_c |\widehat{\Theta}_{t,c}(\delta)|$ exhibits a linear trend $c_t^\star \approx 0.12t$, which seems consistent across the tested distributions. Of note, the constant $0.12$ is reminiscent of the optimal tuning of the sub-Gaussian method of mixtures (Lemma 1.23). This is in line with the observation in Section 3.2 that the confidence width $|\widehat{\Theta}_{t,c}(\delta)|$ is equivalent to the Gaussian confidence width when $t \to +\infty$, which implies that asymptotic tuning of $c_t^\star$ should be similar for all distributions. The experiments in Figure 3.3 thus confirms that this tuning is also near optimal for nonasymptotic, small sample sizes.

Furthermore, we report in Figure 3.4 the confidence envelopes for $c$ varying between $0.1$ and $30$. We find that in practice, for small samples, $c$ has a rather limited influence on the width of the confidence interval. We also report the heuristic $c_t \approx 0.12t$, which indeed provides the tightest envelope. However, we restate that such a sample size-dependent regularisation parameter $c$ is *not* supported by the theory (as it would violate the martingale construction and the law of iterated logarithms). With this in mind, it is however possible to choose a specific horizon of interest $t_0$ and set $t = 0.12t_0$ to promote sharpness around $t = t_0$.

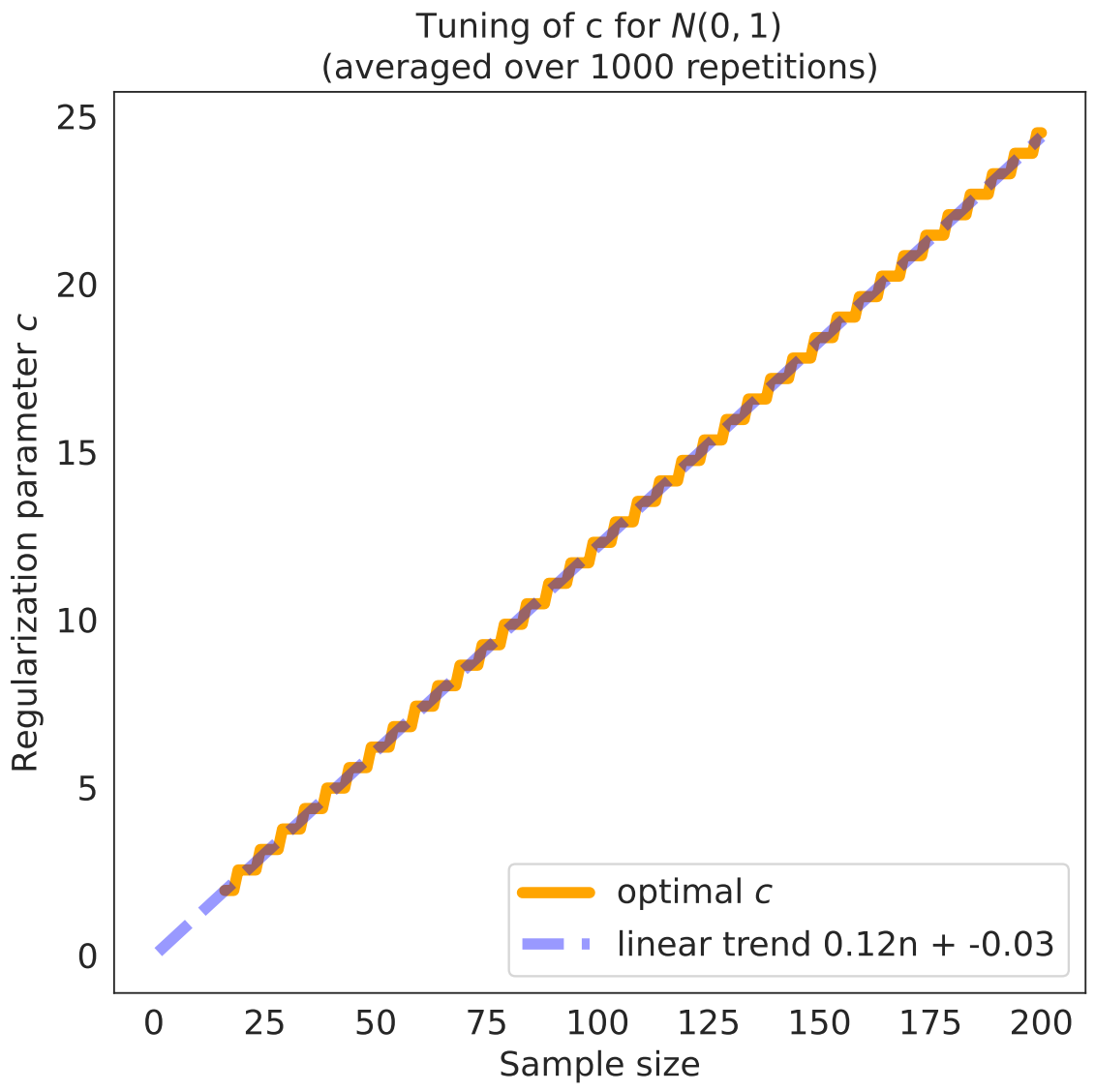

**(a)** Gaussian.

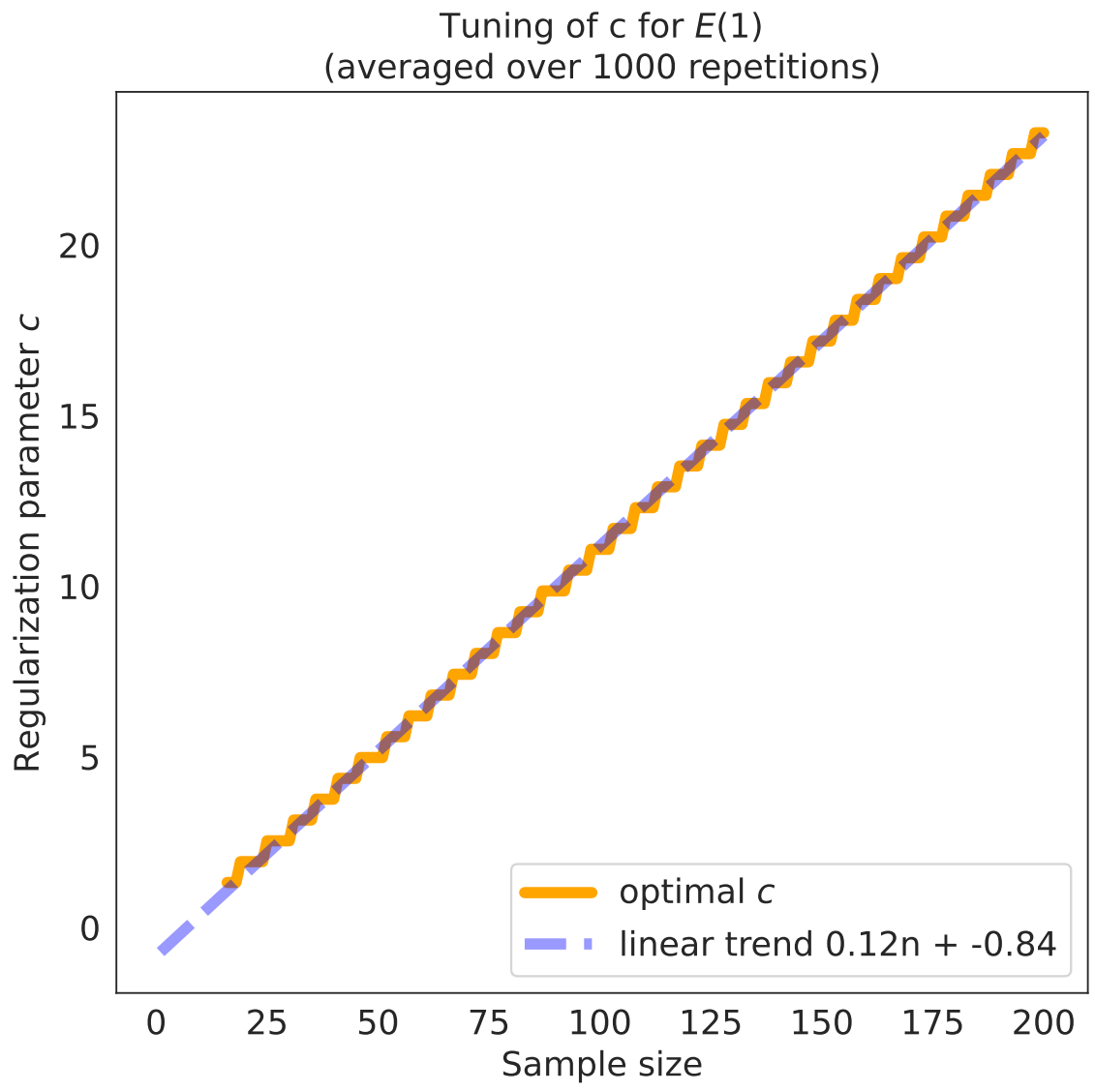

**(b)** Exponential.

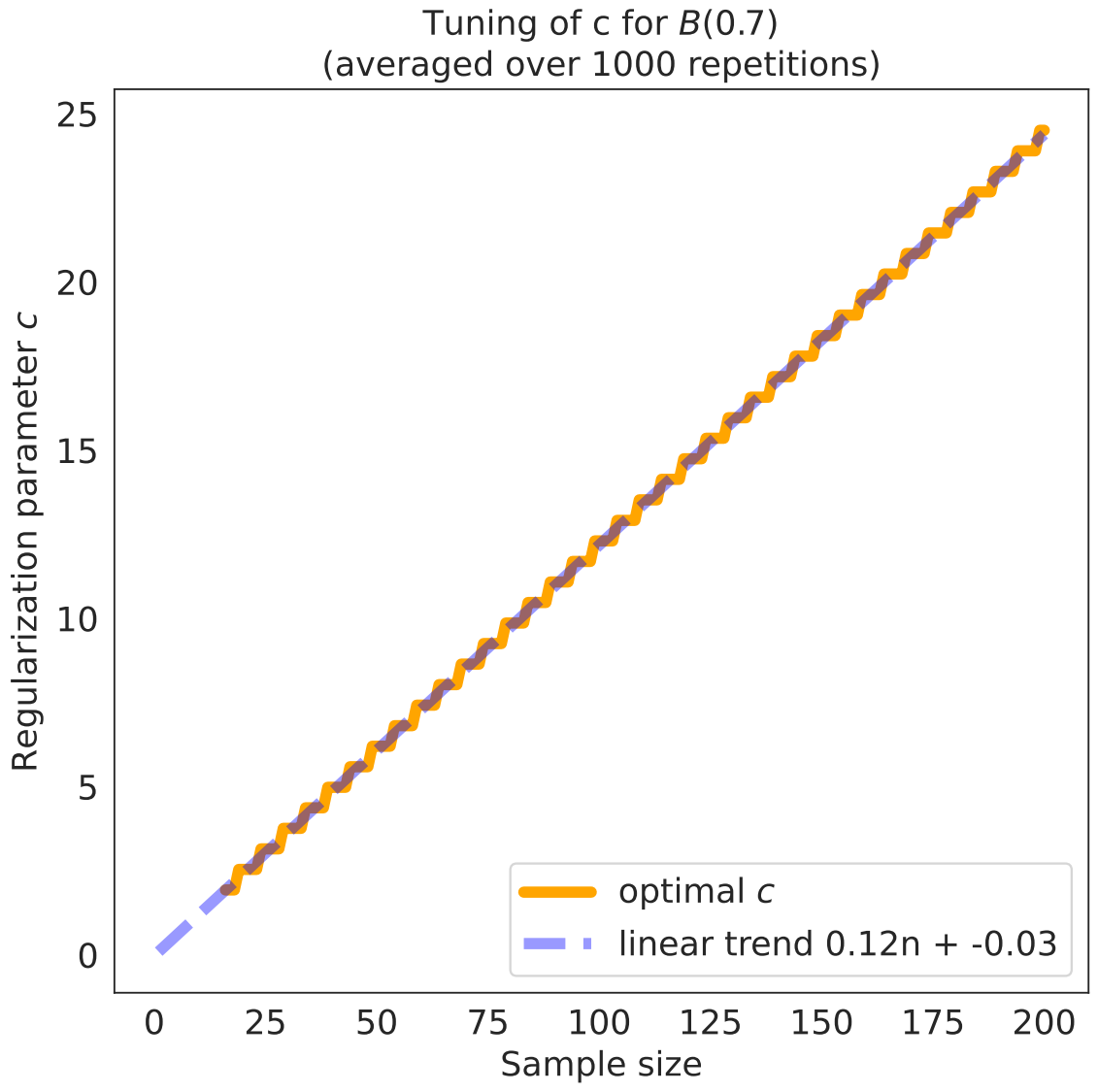

**(c)** Bernoulli.

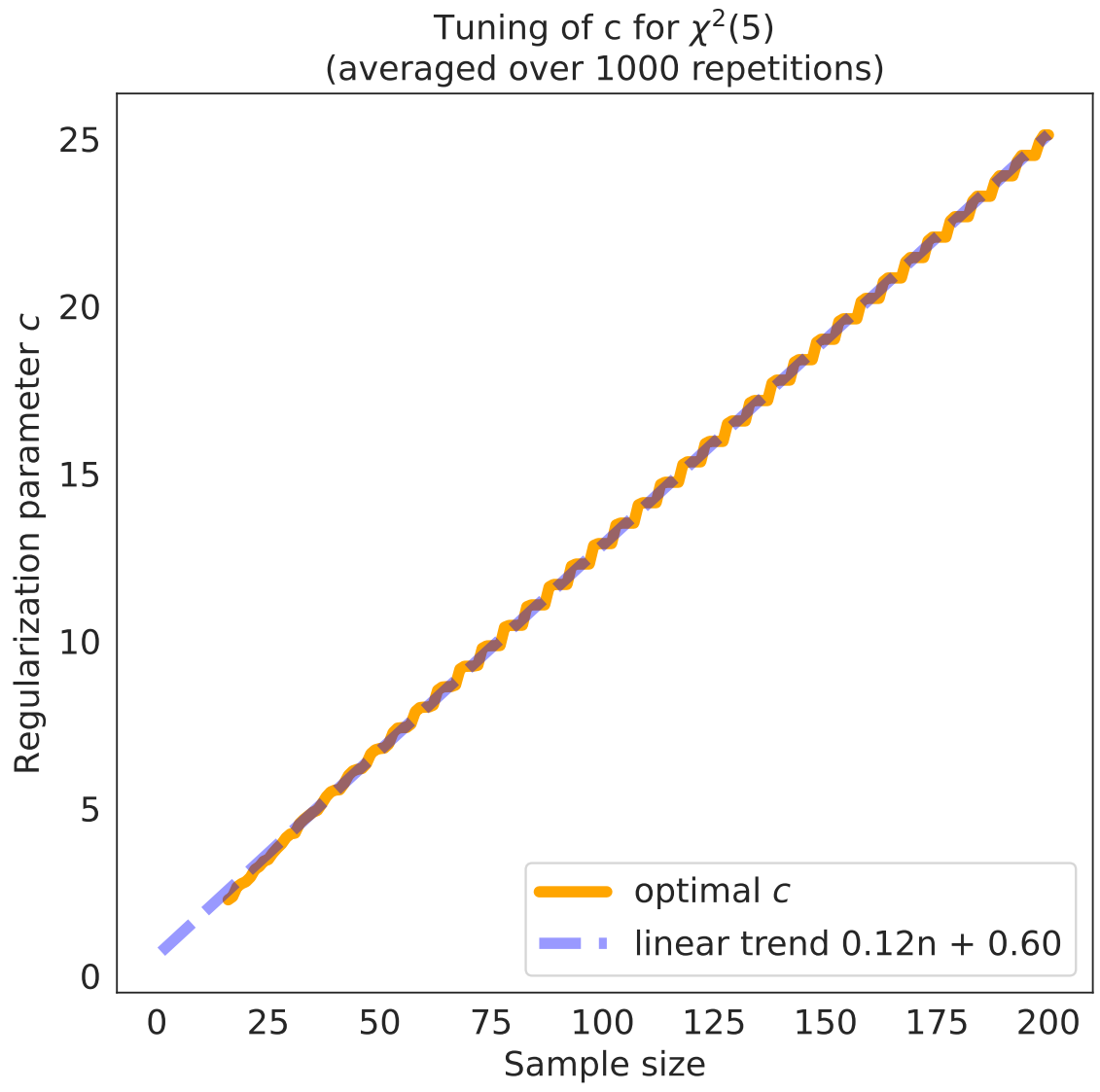

**(d)** Chi-square.

**Figure 3.3** – Optimal (smallest confidence interval diameter) local regularisation parameter $c_t^\star = \text{argmin}_c |\widehat{\Theta}_{t,c}(\delta)|$ with $\delta = 5\%$ for $\mathcal{N}(0,1)$, $\mathcal{E}(1)$, $\mathcal{B}(0.8)$ and $\chi^2(5)$. Note the apparent common linear trend $c_t^\star \approx 0.12t$. Results are averaged over 1000 independent simulations.

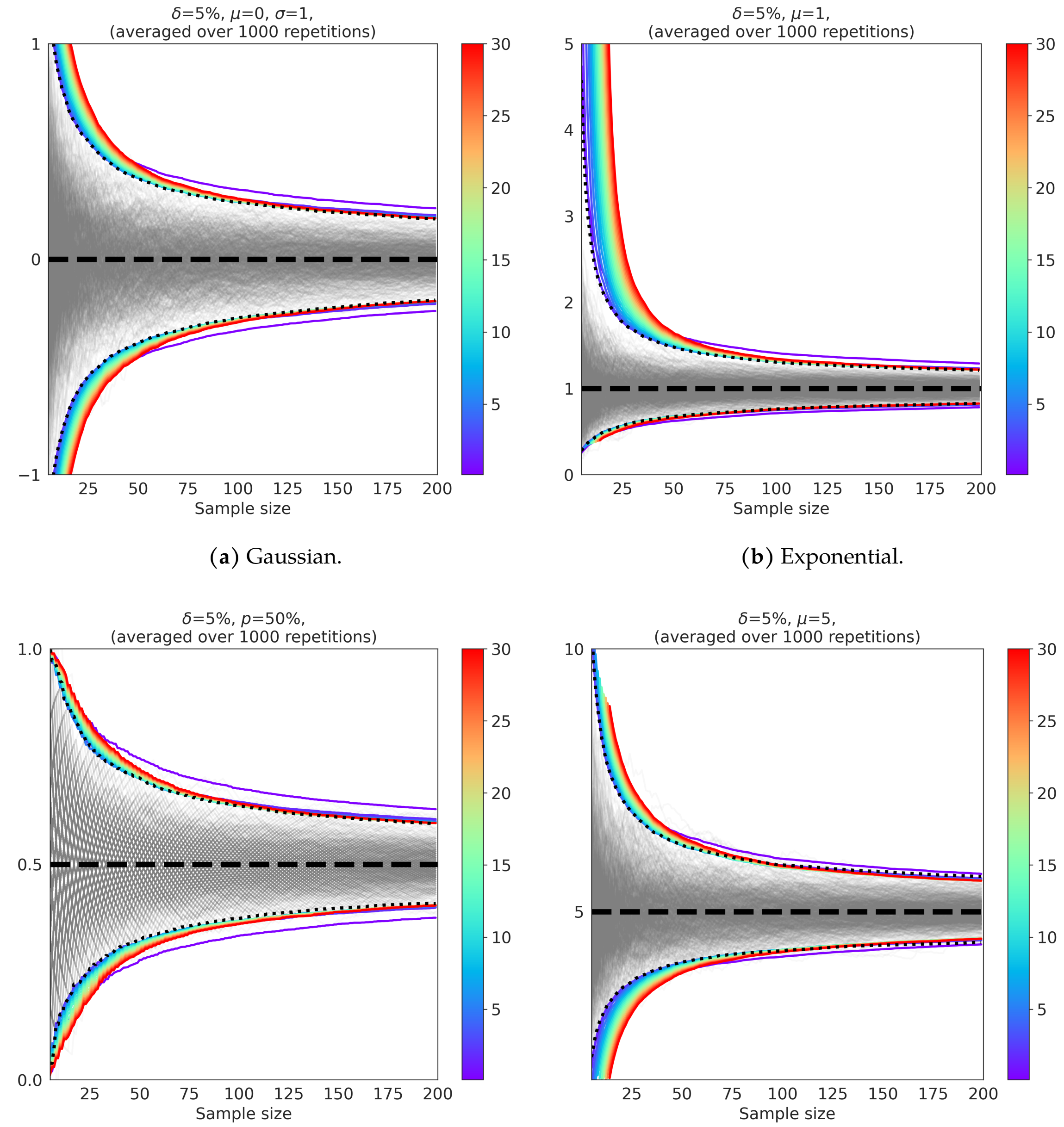

(**a**) Gaussian.

(**b**) Exponential.

(**c**) Bernoulli.

(**d**) Chi-square.

**Figure 3.4 –** Comparison of the time-uniform Bregman confidence envelopes for $\mathcal{N}(0,1)$, $\mathcal{E}(1)$, $\mathcal{B}(0.5)$ and $\chi^2(5)$, and varying mixing parameter $c \in [0.1, 20]$, as a function of the sample size $t$, over 1000 independent replicates. Grey lines are trajectories of empirical means $\widehat{\mu}_t$. Thick black dashed line: $\mu$. Dotted black line: heuristic $c_t \approx 0.12t$.

## 3.5 GLR test in exponential families and change point detection

In this section, we apply our result to control the false alarm rate (type I error) of generalised likelihood ratio (GLR) tests when detecting a change of measure from a sequence of observa-tions. More precisely, we consider a sequence of observations $(Y_t)_{t\in\mathbb{N}}$ such that for all $t \in \mathbb{N}$, $Y_t \sim p_{\theta^\star(t)}$ and $\theta^\star(t) \in \Theta$ is a (possibly time-varying) parameter describing an exponential family $\mathcal{F}$. We consider the composite null hypothesis that this sequence is stationary, i.e.

$$\mathcal{H}_0 = \exists \theta^\star \in \Theta,\ \forall t \in \mathbb{N},\ \theta^\star(t) = \theta^\star\,, \tag{3.42}$$

and we aim to test it against the composite alternative that there is a change point at time $t^\star$ in the distribution of the sequence, i.e.

$$\mathcal{H}_1 = \exists t^\star \in \mathbb{N},\ \theta_0^\star, \theta_1^\star \in \Theta,\ \forall t \in \mathbb{N},\ \theta^\star(t) = \theta_0^\star \mathbb{1}_{t \leqslant t^\star} + \theta_1^\star \mathbb{1}_{t > t^\star}\,. \tag{3.43}$$

The GLR test in the exponential family model $\mathcal{F}$ is defined, for a threshold $\alpha > 0$, as

$$\tau(\alpha; \mathcal{F}) = \min\{t \in \mathbb{N},\ \max_{s \in [0,t)} G^{\mathcal{F}}_{1:s:t} \geqslant \alpha\}, \tag{3.44}$$

where the GLR is defined to be $G^{\mathcal{F}}_{1:s:t} = \inf_{\theta'} \sup_{\theta_0,\theta_1} \log\left(\frac{\prod_{r=1}^{s} p_{\theta_0}(Y_r) \prod_{r=s+1}^{t} p_{\theta_1}(Y_r)}{\prod_{r=1}^{t} p_{\theta'}(Y_r)}\right)$. Then, the false alarm rate of the GLR test (i.e. the probability that a change is detected under the null, also known as the type I error) can be bounded using

$$\mathbb{P}_{\theta^\star}\left(\tau(\alpha;\mathcal{F}) < \infty\right) = \mathbb{P}_{\theta^\star}\left(\exists t \in \mathbb{N},\ \exists s < t,\ G^{\mathcal{F}}_{1:s:t} \geqslant \alpha\right). \tag{3.45}$$

Observe that solving for the inner supremums in the expression of GLR, we obtain

$$G^{\mathcal{F}}_{1:s:t} = \inf_{\theta'} s\mathcal{B}_{\mathcal{L}}(\theta', \widehat{\theta}_s) + (t-s)\mathcal{B}_{\mathcal{L}}(\theta', \widehat{\theta}_{s+1:t}), \tag{3.46}$$

where we introduce parameter estimates $\widehat{\theta}_s$ and $\widehat{\theta}_{s+1:t}$ such that $\nabla\mathcal{L}(\widehat{\theta}_s) = \frac{1}{s}\sum_{r=1}^{s} F(Y_r)$ and $\nabla\mathcal{L}(\widehat{\theta}_{s+1:t}) = \frac{1}{t-s}\sum_{r=s+1}^{t} F(Y_r)$, respectively. Hence, the false alarm rate can be controlled as

$$\begin{aligned}\mathbb{P}_{\theta^\star}(\tau(\alpha;\mathcal{F}) < \infty) \leqslant\ & \mathbb{P}_{\theta^\star}\left(\exists s \in \mathbb{N},\ s\mathcal{B}_{\mathcal{L}}(\theta^\star, \widehat{\theta}_s) \geqslant \alpha_1\right) \\ & + \mathbb{P}_{\theta^\star}\left(\exists t \in \mathbb{N},\ \exists s < t,\ (t-s)\mathcal{B}_{\mathcal{L}}(\theta^\star, \widehat{\theta}_{s+1:t}) \geqslant \alpha_2\right)\end{aligned} \tag{3.47}$$

for appropriate terms $\alpha_1, \alpha_2$ such that $\alpha = \alpha_1 + \alpha_2$. The control of the first term comes naturally from our time-uniform deviation result (Theorem 3.3) using a regularised version of the estimate $\widehat{\theta}_s$. For the second term, we need to study the concentration of $\mathcal{B}_{\mathcal{L}}(\theta, \widehat{\theta}_{s+1:t})$ uniformly over $s$ and $t$.

**Doubly time-uniform concentration.** For any $s < t$, $c > 0$ and reference parameter $\theta_0 \in \Theta$, we define the regularised estimate

$$\widehat{\theta}_{s+1:t,c}(\theta_0) = (\nabla\mathcal{L})^{-1}\left(\frac{1}{t-s+c}\left(\sum_{r=s+1}^{t} F(Y_r) + c\nabla\mathcal{L}(\theta_0)\right)\right), \tag{3.48}$$

built from $(t-s)$ observations $(Y_r)_{r=s+1}^t$. Similarly, we introduce the corresponding Bregman information gain $\gamma_{s+1:t,c}(\theta_0)$ as in Definition 3.2. The following result extends the time-uniform bound of Theorem 3.3 to control the Bregman deviation of $\widehat{\theta}_{s+1:t,c}(\theta_0)$ around $\theta_0$.

**Theorem 3.8** (Doubly time-uniform concentration). *Let $\delta \in (0,1]$ and $g\colon \mathbb{N} \to \mathbb{R}_+^*$ such that $\sum_{t=1}^{\infty} 1/g(t) \leqslant 1$ (e.g. $g(t) = \kappa(1+t)\log^2(1+t)$ where $\kappa = 2.10975$). Under the assumptions of Definition 3.2, consider the set*

$$\widehat{\Theta}^{\delta}_{s+1:t,c} = \left\{\theta_0 \in \Theta : (t-s+c)\mathcal{B}_{\mathcal{L}}(\theta_0, \widehat{\theta}_{s+1:t,c}(\theta_0)) \leqslant \log\frac{g(t)}{\delta} + \gamma_{s+1:t,c}(\theta_0)\right\}. \tag{3.49}$$

*Then $\left(\widehat{\Theta}^{\delta}_{s+1:t,c}\right)_{t\in\mathbb{N},s<t}$ is a **doubly time-uniform confidence sequence** at level $\delta$ for $\theta^\star$, i.e.*

$$\mathbb{P}_{\theta^\star}\left(\forall t \in \mathbb{N}, \forall s < t,\ \theta^\star \in \widehat{\Theta}^{\delta}_{s+1:t,c}\right) \geqslant 1-\delta\,. \tag{3.50}$$

**Remark 3.9**. *Maillard (2019a) proves a doubly uniform concentration inequality for means of (sub)-Gaussian random variables and conjectures that it could be extended to other types of changes, such as changes of variance in a Gaussian family. Theorem 3.8 can be seen as a generalisation of this result to generic exponential families. Again. this extension naturally comes with additional challenges due to the intrinsic local geometry of exponential families.*

*Proof of Theorem 3.8.* The proof follows a similar strategy based on exponential martingales as in Theorem 3.3.

Assume that the null hypothesis holds, i.e. that all the $(Y_t)_{t\in\mathbb{N}}$ come from a single distribution belonging to an exponential family model $\mathcal{F}$ with parameter $\theta^\star \in \mathbb{N}$ and let $t \in \mathbb{N}$ and $s \in \{0,\ldots,t-1\}$. We introduce the scan mean $\widehat{\mu}_{s+1:t} = \frac{1}{t-s}\sum_{r=s+1}^{t} F(Y_r)$ and its expectation, i.e.

the true mean $\mu = \nabla\mathcal{L}(\theta^\star)$. We define as in the proof of Theorem 3.3, for each $\lambda$ the martingale

$$M^{\lambda}_{s+1:t} = \exp\left(\langle \lambda, (t-s)(\widehat{\mu}_{s+1:t} - \mu)\rangle - (t-s)\mathcal{B}_{\mathcal{L},\theta^\star}(\lambda)\right) . \quad (3.51)$$

Then, applying the method of mixtures and replacing each term with its scan version from $s+1$ to $t$, we introduce the quantity

$$M_{s+1,t} = \exp\Big((t-s+c)\mathcal{B}^\star_{\mathcal{L},\theta^\star}(x)\Big) \frac{G(\theta^\star, c)}{G(\widehat{\theta}_{s+1:t,c}(\theta^\star), t-s+c)} \quad (3.52)$$

where $x = \frac{t-s}{t-s+c}(\mu_{s+1:t} - \nabla\mathcal{L}(\theta^\star))$ and $\widehat{\theta}_{s+1:t,c}(\theta^\star)$ is the regularised estimate from the scan samples

$$\widehat{\theta}_{s+1:t,c}(\theta^\star) = (\nabla\mathcal{L})^{-1}\left(\frac{\sum\limits_{r=s+1}^{t} F(Y_r) + c\nabla\mathcal{L}(\theta^\star)}{t-s+c}\right) . \quad (3.53)$$

Remarking that $\mathbb{E}_{\theta^\star}[M_{t,t}] \leqslant \mathbb{E}_{\theta^\star}[M_{t+1,t}] = 1$ and that $(M_{s+1,t})_s$ is a nonnegative supermartingale, we can now control its doubly-time uniform deviations following the proof of Maillard (2019b)[Theorem 3.2]. More precisely, for any non-decreasing function $g$, we show below that the following peeling-like result holds:

$$\mathbb{P}_\theta\left(\exists t \in \mathbb{N}, \exists s < t : M_{s+1,t} \geqslant \frac{g(t)}{\delta}\right) \leqslant \delta\mathbb{E}_{\theta^\star}\left[\max_{t\in\mathbb{N}} \max_{s<t} \frac{M_{s+1:t}}{g(t)}\right] \leqslant \delta \sum_{t=1}^{\infty} \frac{1}{g(t)} . \quad (3.54)$$

Rewriting the terms thanks to the duality formulas, yields

$$\mathbb{P}_{\theta^\star}\left(\exists t \in \mathbb{N}, \exists s < t : (t-s-c)\mathcal{B}_{\mathcal{L}}(\theta^\star, \widehat{\theta}_{s+1:t,c}(\theta^\star)) \geqslant \log\left(\frac{g(t)}{\delta}\right) + \gamma_{s+1:t,c}(\theta^\star)\right) \leqslant \delta \sum_{t=1}^{\infty} \frac{1}{g(t)} . \quad (3.55)$$

**Doubly time-uniform control of supermartingale sequences.** For completeness, we reproduce below the derivation from Maillard (2019b)[Theorem 3.2], applied to our setup. First, let us note that

$$\begin{aligned}\mathbb{P}_{\theta^\star}\left(\exists t \in \mathbb{N},\ \exists s < t,\ M_{s+1:t} \geqslant \frac{g(t)}{\delta}\right) &= \mathbb{P}_{\theta^\star}\left(\max_{t\in\mathbb{N}} \max_{s<t} \frac{M_{s+1:t}}{g(t)} \geqslant \frac{1}{\delta}\right) \\ &\leqslant \delta\mathbb{E}_{\theta^\star}\left[\max_{t\in\mathbb{N}} \max_{s<t} \frac{M_{s+1:t}}{g(t)}\right] . \end{aligned} \quad (3.56)$$

Let us also denote $\tau$ the random stopping time corresponding to the first occurrence $t$ of the event $\max_{s\in[0,t-1]} \frac{M_{s+1:t}}{g(t)} \geqslant \frac{1}{\delta}$. It is convenient to introduce the quantity $\overline{M}_t = \frac{1}{g(t)} \sum_{s\in\{0,\ldots,t-1\}} M_{s+1,t}$ for each $t \in \mathbb{N}$. Since each $M_{s+1,t}$ and $g(t)$ is nonnegative, we first get that for every random stopping time $\tau \in \mathbb{N}$, the following inequality holds:

$$\mathbb{E}_{\theta^\star}\left[\frac{\max_{s<\tau} M_{s+1,\tau}}{g(\tau)}\right] \leqslant \mathbb{E}_{\theta^\star}\left[\overline{M}_\tau\right] = \mathbb{E}_{\theta^\star}\left[\overline{M}_1 + \sum_{t=1}^{\infty}(\overline{M}_{t+1} - \overline{M}_t)\mathbb{1}_{\tau>t}\right]. \tag{3.57}$$

Furthermore, we note that, conveniently,

$$\overline{M}_{t+1} - \overline{M}_t = \frac{M_{t+1,t+1}}{g(t+1)} + \sum_{s=0}^{t-1}\left(\frac{M_{s+1,t+1}}{g(t+1)} - \frac{M_{s+1,t}}{g(t)}\right). \tag{3.58}$$

Next, by assumption, we note that $\mathbb{E}_{\theta^\star}[M_{s+1,t+1} \mid \mathcal{F}_t] \leqslant M_{s+1,t}$. Thus, since $\mathbb{1}_{\tau>t} \in \mathcal{F}_t$, we deduce that

$$\begin{aligned}
&\mathbb{E}_{\theta^\star}\left[\frac{\max_{s<\tau} M_{s+1,\tau}}{g(\tau)}\right] \\
&\leqslant \mathbb{E}_{\theta^\star}\left[\overline{M}_1\right] + \sum_{t=1}^{\infty}\frac{\mathbb{E}_{\theta^\star}[M_{t+1,t+1}]}{g(t+1)} + \sum_{t=1}^{\infty}\sum_{s<t}\mathbb{E}_{\theta^\star}\left[\left(\frac{1}{g(t+1)} - \frac{1}{g(t)}\right)M_{s+1,t}\mathbb{1}_{\tau>t}\right] \\
&= \mathbb{E}_{\theta^\star}\left[\overline{M}_1\right] + \sum_{t=1}^{\infty}\frac{\mathbb{E}_{\theta^\star}[M_{t+1,t+1}]}{g(t+1)} + \sum_{t=1}^{\infty}\sum_{s<t}\left(\frac{1}{g(t+1)} - \frac{1}{g(t)}\right)\underbrace{\mathbb{E}_{\theta^\star}\left[M_{s+1,t}\mathbb{1}_{\tau>t}\right]}_{\geqslant 0}.
\end{aligned} \tag{3.59}$$

Hence, the assumption that $g$ is non-decreasing ensures that the last sum is upper bounded by 0. Since on the other hand $\mathbb{E}_{\theta^\star}[M_{t+1,t+1}] \leqslant 1$ holds for all $t$ (and thus $\mathbb{E}_{\theta^\star}\left[\overline{M}_1\right] \leqslant 1/g(1)$), we deduce that the following inequality holds:

$$\mathbb{E}_{\theta^\star}\left[\frac{\max_{s<\tau} M_{s+1,\tau}}{g(\tau)}\right] \leqslant \frac{1}{g(1)} + \sum_{t=1}^{\infty}\frac{1}{g(t+1)} = \sum_{t=1}^{\infty}\frac{1}{g(t)} \leqslant 1. \tag{3.60}$$

■

**A regularised GLR test.** For a given false alarm (i.e. false positive) probability $\delta \in (0,1]$ and parameter $c > 0$, we define the regularised GLR test $\tau_c^\delta(\mathcal{F})$ as follows:

$$\widehat{\Theta}_{s,c}^\delta = \left\{\theta_0 \in \Theta : (s+c)\mathcal{B}_{\mathcal{L}}(\theta_0, \widehat{\theta}_{s,c}(\theta_0)) \leqslant \underbrace{\log\frac{1}{\delta} + \gamma_{s,c}(\theta_0)}_{\alpha_1}\right\}, \tag{3.61}$$

$$\widehat{\Theta}_{s+1:t,c}^\delta = \left\{\theta_0 \in \Theta : (t-s+c)\mathcal{B}_{\mathcal{L}}(\theta_0, \widehat{\theta}_{s+1:t,c}(\theta_0)) \leqslant \underbrace{\log\frac{g(t)}{\delta} + \gamma_{s+1:t,c}(\theta_0)}_{\alpha_2}\right\}, \tag{3.62}$$

$$\tau_c^\delta(\mathcal{F}) = \min\left\{t \in \mathbb{N} : \exists s < t, \widehat{\Theta}_{s,c}^{\delta/2} \cap \widehat{\Theta}_{s+1:t,c}^{\delta/2} = \emptyset\right\}, \tag{3.63}$$

Note that, under the measure $\mathbb{P}_{\theta^\star}$, the event $\theta^\star \in \widehat{\Theta}_{s,c}^{\delta/2}$ holds with probability at least $1-\delta/2$ by time-uniform concentration over $s$ (Theorem 3.3), and $\theta \in \widehat{\Theta}_{s+1:t,c}^{\delta/2}$ holds with probability at least $1-\delta/2$ by doubly time-uniform concentration over both $s$ and $t$ (Theorem 3.8). Then, by a union bound, this test is guaranteed to have a false alarm probability controlled by $\delta$. The factor $g(t)$ is essentially a function growing slightly faster than linearly that inflates the width of the confidence set in the doubly time-uniform bound to allow for the peeling-like construction above. It can be tuned to satisfy $\sum_{t=1}^{+\infty} 1/g(t) \leqslant 1$, thus controlling the above deviation probability by at most $\delta$ (ideally, $\sum_{t=1}^{+\infty} 1/g(t)$ should be close to 1 to avoid overinflating the confidence width, which would decrease the statistical power of change point test). A natural choice for this is $g(t) = \kappa(1+t)\log^2(1+t)$ where $\kappa = 2.10974 \geqslant \sum_{t=1}^{\infty} \frac{1}{(1+t)\log^2(1+t)}$ (for completeness, we derive in Appendix B.4 an elementary way to compute suitable values for $\kappa$); other common choices include $g(t) \propto t^\eta$ for some $\eta > 1$, though it leads to a looser bound due to the faster growth. Furthermore, if one is interested in detecting changes up to a known horizon $T$, one can replace the function $g(t)$ by the slightly tighter factor $g(t) = \kappa(1+t)\log^2(1+t)$ with $\kappa = \sum_{t=1}^{T} 1/g(t)$ and still guarantee a false alarm rate under $\delta$.

**Experiments on change point detection for Gaussian with unknown variance.** We illustrate the change point GLR test above in a numerical experiment as follows. We consider the one-dimensional exponential family of centered Gaussian distributions with unknown variance $\mathcal{F} = \{\mathcal{N}(0,\sigma^2), \sigma > 0\}$. A sequence of independent random variables $(Y_t)_{t\in\mathbb{N}}$ is drawn from $\mathcal{N}(0,\sigma_0^2)$ if $t \leqslant t^\star$ and from $Y_t \sim \mathcal{N}(0,\sigma_1^2)$ if $t > t^\star$, where $t^\star = 50$ and $\sigma_0 = 1$ and $\sigma_1 \in [1,2,3,4]$. Of note, this setting corresponds to an open question in Maillard (2019a), which studies GLR tests in sub-Gaussian families, which are adapted to the detection of changing means but not variances. Detection times are reported as $\min(\tau_c^\delta(\mathcal{F}), T)$ with $T = 100$, and $\tau_c^\delta(\mathcal{F}) > T$ is interpreted as no change being detected. For the doubly time-uniform confidence set in the definition of the regularised GLR test, we use the factor $g(t) = \kappa(1+t)\log^2(1+t)$ with $\kappa = 2.10974$.

An example of practical motivation for this setting is the design of remote health monitoring devices. In the case of diabetes for instance, for which there exists such wearable devices, it is important to control not only the average level of blood glucose but also its intraday variability, to ensure the metabolism adequately responds to the antidiabetic medication; if not, the concentration of glucose in the haemoglobin may exhibit a steady average but an increased variance, which should be detected as early as possible.

In Figure 3.5, we report the histograms of detection times across 1000 independent simulations for increasingly abrupt changes of variance ($\sigma_1 \in \{1,2,3,4\}$). As expected, the distribution of detection times shifts closer to the actual change point $t^\star$ when $\sigma_1$ increases, i.e. more obvious changes are detected earlier. This is confirmed in Figure 3.6, where we report the median and interquartile range of detection times for $\sigma_1 \in [1,4]$.

We empirically validate in Figure 3.5a the control of the false positive probability by at most $\delta = 5\%$. However, the actual false positive rate appears much lower $(0.2\%)$, which is a consequence of the looseness of the union bound involving the factor $g(t)$ in the doubly time-uniform confidence set. Following Maillard (2019a), a sharper approach would involve a direct concentration result on the pair $\left(\widehat{\theta}_{s,c}, \widehat{\theta}_{s+1:t,c}\right)$ (i.e. in the terminology of Maillard (2019a), a joint bound, as opposed to the current disjoint one). This is a nontrivial result which we leave for future work. Finally, note that the high false negative rate of Figure 3.5b (27.6%) is an artifact of thresholding the detection time at $T = 100$; increasing $T$ would enable later detections, thus reducing the number of observed false negatives, at the cost of increasing the average detection delay (as well as the computational burden of the experiment).

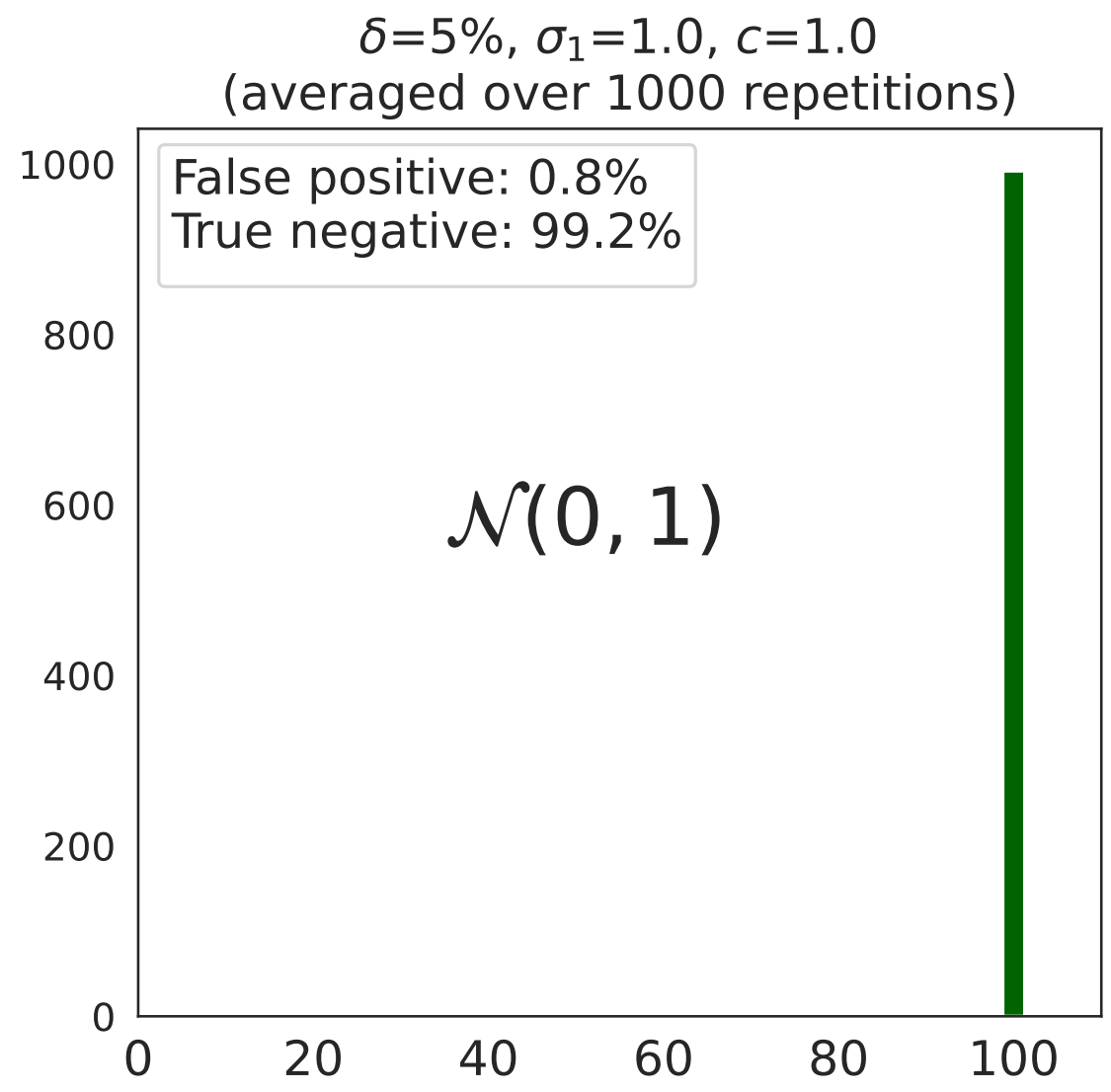

**(a)** $\sigma_0 = \sigma_1 = 1$ (no change).

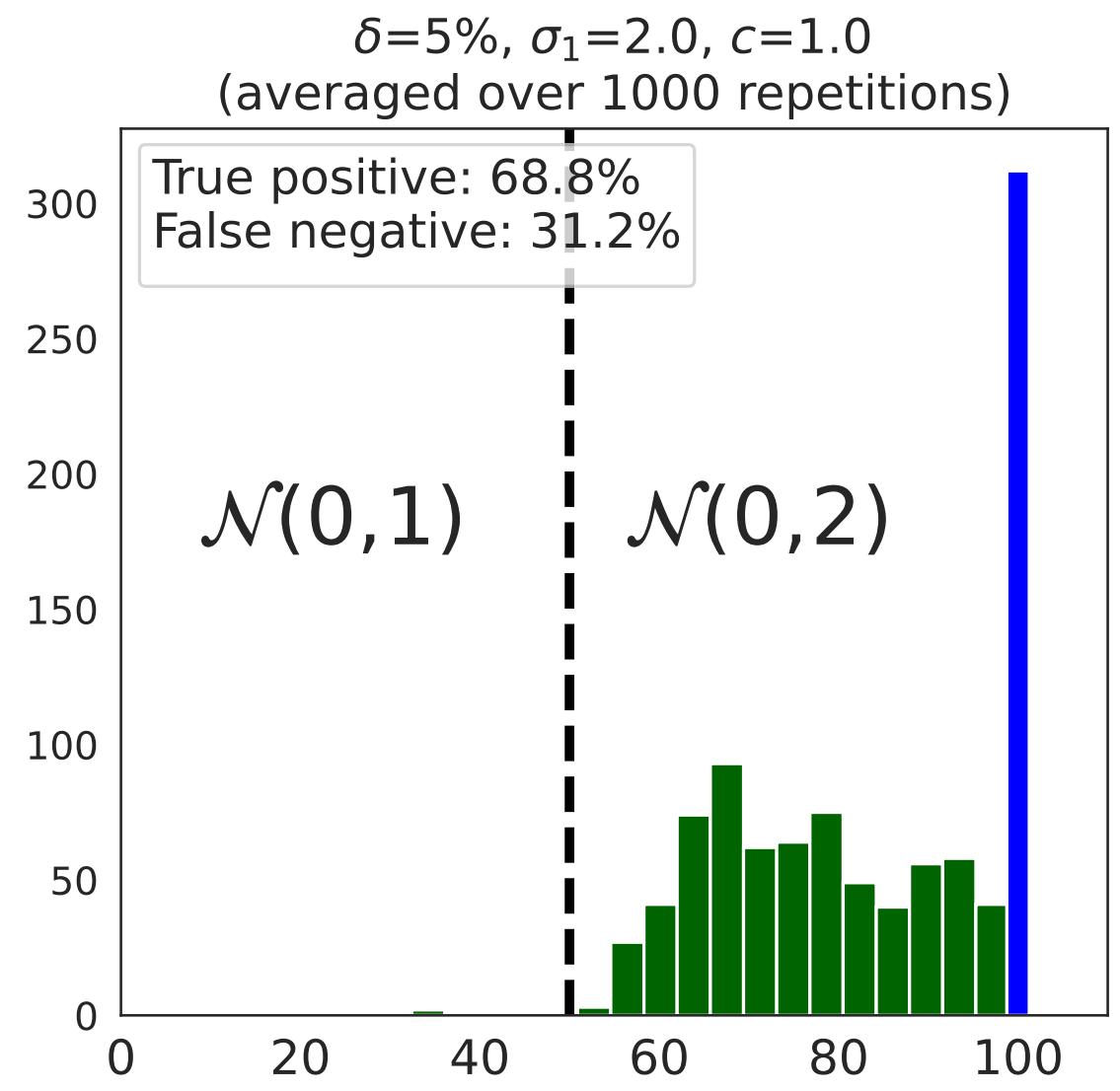

**(b)** $\sigma_0 = 1, \sigma_1 = 2$.

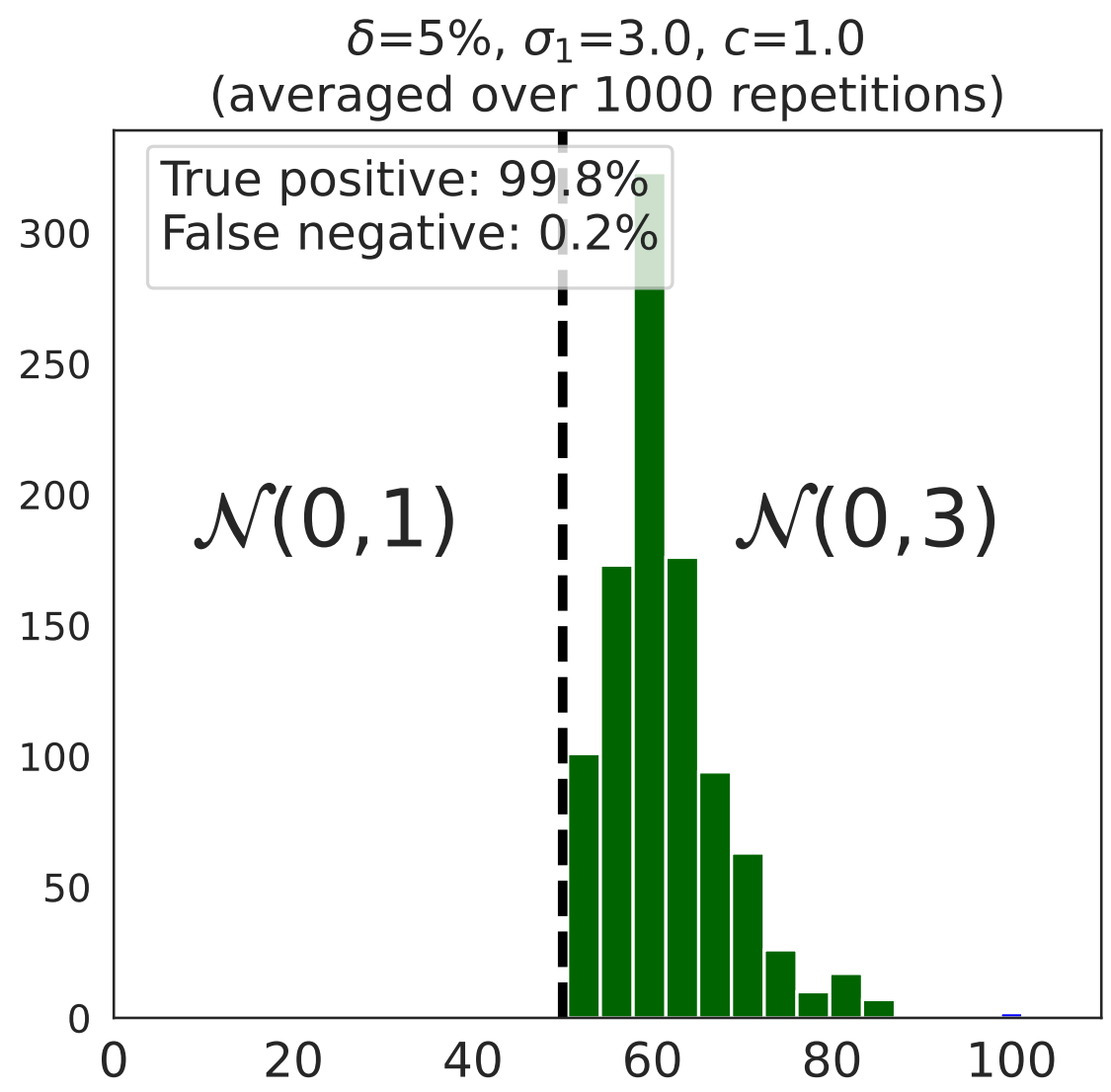

**(c)** $\sigma_0 = \sigma_1 = 3$.

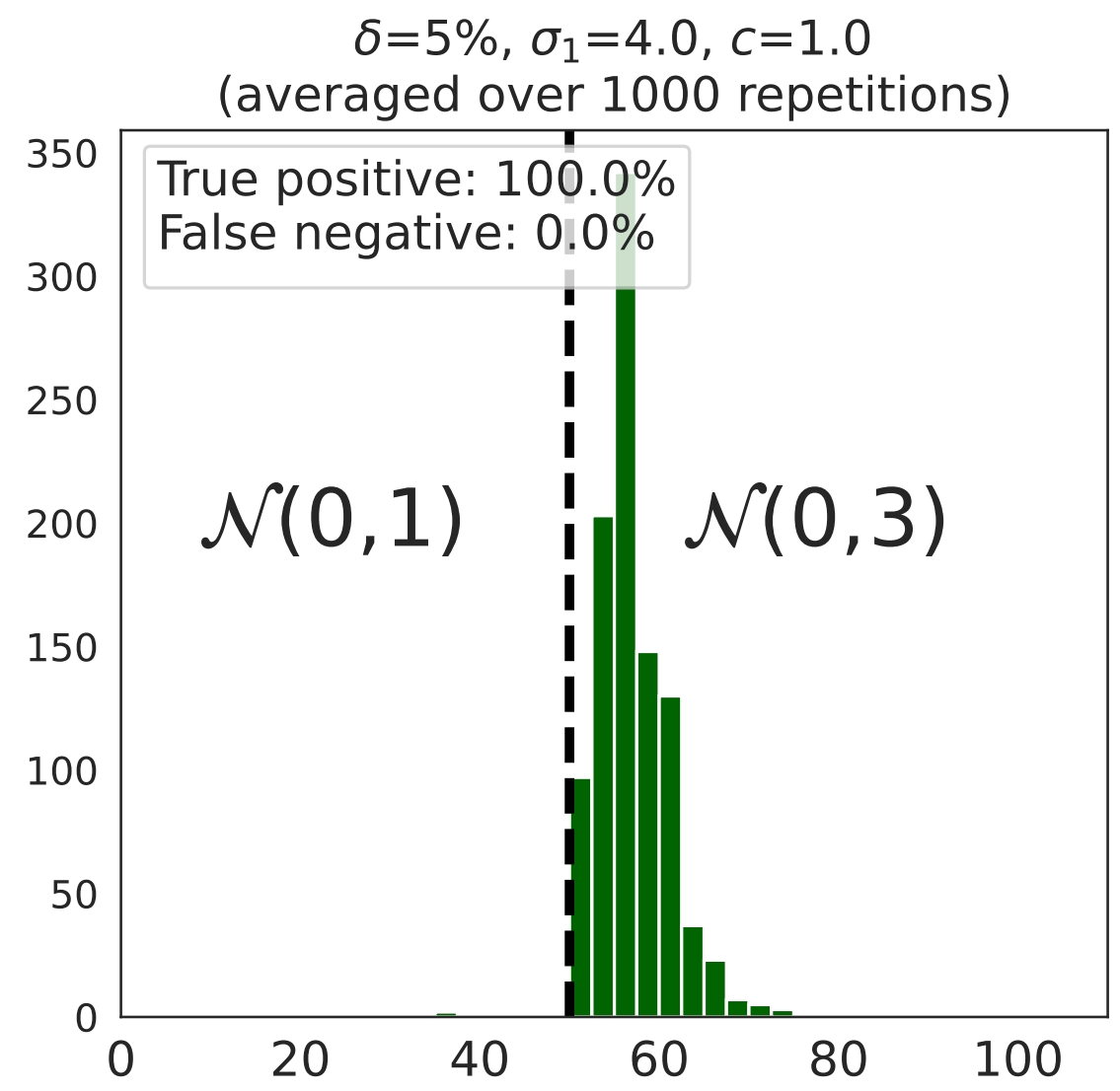

**(d)** $\sigma_0 = 1, \sigma_1 = 4$.

**Figure 3.5 –** Histograms of detection times $\tau_c^\delta(\mathcal{F})$ in the Gaussian setting described in Section 3.5. Green: correct detection (negative in (a), positive in (b), (c) and (d)). Blue: incorrect detection (positive in (a), negative in (b), (c) and (d)). Black dashed line: change point $t^\star$.

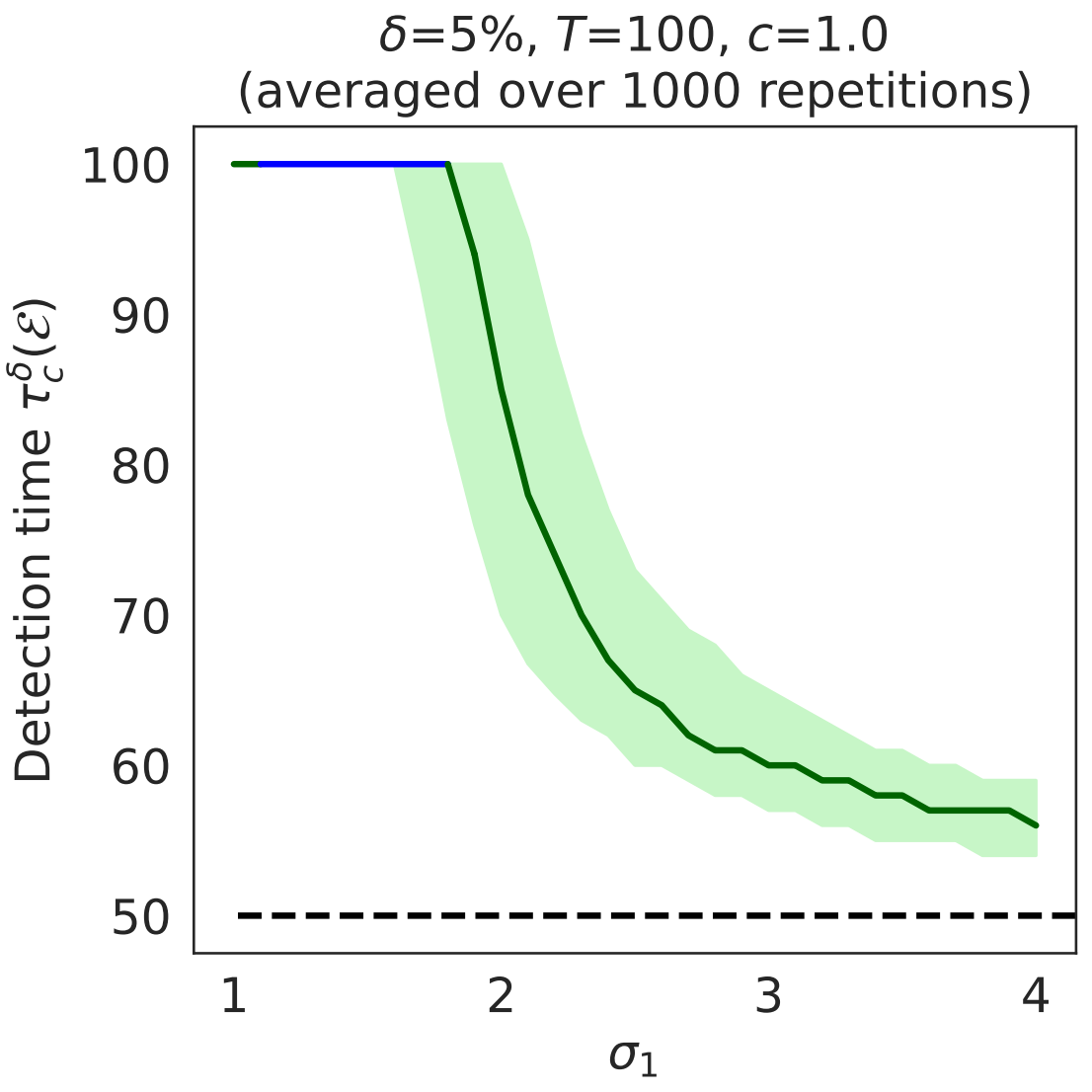

**Figure 3.6 –** Detection time $\tau_c^\delta(\mathcal{F})$ in the Gaussian setting described in Section 3.5. Solid line: median detection time. Shaded area: interquartile range of detection times (25th and 75th percentiles). Black dashed line: change point $t^\star$.

# Conclusion

We adapted the method of mixtures to derive a time-uniform deviation inequality for generic parametric exponential families expressed in terms of their Bregman divergences, highlighting the role of a quantity, the *Bregman information gain*, that is related to the geometry of the family. We specialised this general result to build confidence sets for classical examples. Our method compared favourably to the state-of-the art for Gaussian, Bernoulli and exponential distributions, and extends to other families like Chi-square, Pareto and two-parameter Gaussian, where no known time-uniform bound existed. Our method is also general enough to design GLR tests for change detection. An interesting direction for future work would be to consider the case where the log-partition function (and thus the Bregman divergence) is misspecified or unknown and has to be learned alongside the parameter.

# Chapter 4

# Empirical Chernoff concentration: beyond bounded distributions (🔀)

*Des crépuscules blancs tiédissent sous mon crâne*
*Qu'un cercle de fer serre ainsi qu'un vieux tombeau*
*Et triste, j'erre après un rêve vague et beau,*
*Par les champs où la sève immense se pavane.*
— Stéphane Mallarmé, *Renouveau*

The content of this chapter is new and has not been previously published. This work was initially inspired by modelling questions in agriculture, a domain that presents similar modelling challenges to healthcare, in particular small data samples and the need for adaptive nonparametric methods. As a tribute to this, we illustrate our novel concentration bounds on real and simulated agricultural data, showcasing the new second order sub-Gaussian condition as an alternative modelling option for practitioners. We defer to Appendix C.1 proofs and numerical evidence that certain classical distributions satisfy this condition, which amounts to checking simple but tedious inequalities.

## Contents

## Outline and contributions

In this chapter, we introduce a subclass of the family of sub-Gaussian distributions, called *second order sub-Gaussian*, that satisfies an additional cumulant condition, extending the standard Chi-square setting, and spans a vast range of nonparametric families, including Gaussian, Bernoulli, uniform, triangular and (presumably) Beta. This alternative model specification enables to substitute the prior knowledge of the sub-Gaussian parameter in the standard Cramér-Chernoff bound (Corollary 1.16) and method of mixtures (Corollary 1.21) with a data-dependent estimate, eschewing the need for a known bounded support as in previous empirical approaches. We detail the new concentration inequalities in Section 4.3, both in the fixed sample (Theorem 4.8) and time-uniform (Theorem 4.10) settings. We further instantiate these results in Section 4.4 to construct implicitly defined confidence sets for the mean, which can be easily implemented with a root search routine (Corollary 4.14). Building on an alternative symmetrised cumulant control, we also provide fully explicit confidence intervals (Corollary 4.17). We conclude by conducting a range of numerical experiments to (i) illustrate properties of the new mixture measure used to derive the time-uniform bounds, and (ii) compare these new bounds to state-of-the-art competitors on several illustrative examples, including a real-world field experiment in Canada, highlighting their merits in the small sample regime.

## 4.1 Concentration of the mean without knowing the variance

We consider the same setting as in Section 1.4 in Chapter 1, i.e. we observe an i.i.d. sequence $(Y_t)_{t\in\mathbb{N}}$ drawn from a distribution $\nu \in \mathbb{L}^1(\mathbb{R})$ and we seek to construct a sequence $(\Theta_t^\delta)_{t\in\mathbb{N}}$ of (random) subsets of $\mathbb{R}$, such that either $\mathbb{P}(\mu \in \Theta_t^\delta) \geqslant 1-\delta$ for all $t \in \mathbb{N}$ (fixed sample) or $\mathbb{P}(\forall t \in \mathbb{N},\ \mu \in \Theta_t^\delta) \geqslant 1-\delta$ (time-uniform) for any $\delta \in (0,1)$.

In some cases where $\nu$ belongs to a known parametric family, such confidence sets may be computed exactly. For instance, in the fixed sample setting, if $\nu$ is a Gaussian measure, the statistic $T_t$ defined as

$$T_t = \sqrt{t}\left(\widehat{\mu}_t - \mu\right)/\widehat{\sigma}_t, \quad \text{with} \quad \widehat{\mu}_t = \frac{1}{t}\sum_{s=1}^{t} Y_s \quad \text{and} \quad \widehat{\sigma}_t^2 = \frac{1}{t(t-1)}\sum_{1\leqslant s<s'\leqslant t} (Y_s - Y_{s'})^2\,, \tag{4.1}$$

follows the Student distribution with $(t-1)$ degrees of freedom, and therefore

$$\widehat{\Theta}_t^\delta = \left[\widehat{\mu}_t \pm \frac{\widehat{\sigma}_t}{\sqrt{t}} q_{\mathrm{Student}(t-1)}\left(\frac{1}{2} \pm \frac{1-\delta}{2}\right)\right] \tag{4.2}$$

is a confidence set at level $\delta$ and sample size $t$ for $\mu$, where $q_{\text{Student}(t-1)}$ denotes the quantile function of the distribution of $T_t$. In practical settings however, this Gaussian assumption is often an imperfect idealisation of the observed distributions. Moreover, even though such confidence sets are valid in an asymptotic sense as long as the underlying distribution is square integrable (thanks to the central limit theorem), they are not theorically valid for small samples.

A weaker assumption, at the heart of the Cramér-Chernoff method (Proposition 1.15), is to control the cumulant generating function (CGF) of $\nu$, defined as $\lambda \mapsto \log \mathbb{E}_{Y\sim\nu}\left[e^{\lambda(Y-\mu)}\right]$, typically by bounding it from above by the cumulant generating function of a reference distribution such as $\mathcal{N}\left(0, R^2\right)$ (these distributions form the family of $R$-sub-Gaussian distributions $\mathcal{F}_{\mathcal{G},R}$).

Knowing the parameter $R$ is of utmost importance in order to build actionable confidence intervals, as larger $R$ correspond to larger intervals and thus more uncertainty around the estimation of $\mu$. Assuming the support of $\nu$ is bounded, i.e. $\nu \in \mathcal{F}_{[\underline{B},\overline{B}]}$, Hoeffding's lemma (Hoeffding, 1963) suggests $R = \left(B - \underline{B}\right)/2$ as a valid choice for the sub-Gaussian parameter. However, this value is obtained through a worst case analysis that is only tight for Bernoulli-like distributions, and therefore likely to be rather loose for most other distributions that share the same support. Using a subexponential control under the same boundedness assumption and a bound on the variance paves the way for Bernstein inequalities (Hoeffding, 1963); for instance,

$$\widehat{\Theta}_t^\delta = \left[\widehat{\mu}_t \pm \left(R\sqrt{\frac{2\log\frac{2}{\delta}}{t}} + \frac{(\overline{B}-\underline{B})\log\frac{2}{\delta}}{3t}\right)\right] \tag{4.3}$$

is a confidence set at level $\delta$ and sample size $t$ for $\mu$ and the family $\mathcal{F}_{[\underline{B},\overline{B}]} \cap \mathcal{F}_{R^2,1}^{\text{centred}}$. In this example, the dependency on $R$ may be lifted, giving rise to the so-called *empirical Bernstein bounds*; for instance, Maurer and Pontil (2009, Theorem 4) shows that

$$\widehat{\Theta}_t^\delta = \left[\widehat{\mu}_t \pm \left(\widehat{\sigma}_t\sqrt{\frac{2\log\frac{3}{\delta}}{t}} + \frac{7(\overline{B}-\underline{B})\log\frac{3}{\delta}}{3(t-1)}\right)\right] \tag{4.4}$$

is a confidence set at level $\delta$ and sample size $t$ for $\mu$ and the family $\bigcup_{R\in\mathbb{R}_+^\star} \mathcal{F}_{[\underline{B},\overline{B}]} \cap \mathcal{F}_{R^2,1}^{\text{centred}} = \mathcal{F}_{[\underline{B},\overline{B}]}$.

To motivate our setting and the need to move beyond boundedness, let us consider the toy example of a $\mathrm{Beta}\left(\alpha/\varepsilon, \alpha/\varepsilon\right)$ distribution with $\alpha \in \mathbb{R}_+^\star$ and $\varepsilon \in (0,1)$ a small parameter. Intuitively, the confidence interval (4.2) should nearly apply since this distribution is close to being Gaussian (precisely, $\varepsilon^{-1/2}(\mathrm{Beta}\left(\alpha/\varepsilon, \alpha/\varepsilon\right) - 1/2) \to \mathcal{N}(0, 1/(8\alpha))$ in distribution). However, if the only information at hand is the support, one would pay an extra $\mathcal{O}\left(1/t\right)$ in the Maurer-Pontil bound, and although this term decays faster with $t$ than the variance contribution $\mathcal{O}\left(\sigma/\sqrt{t}\right)$, it may considerably widen the confidence interval in the small sample regime.

Anticipating on the rest of this chapter, we illustrate this phenomenon in Figure 4.1, where we provide an empirical comparison of several bounds, including other confidence intervals for bounded distributions (we refer to Table 2.2 for an overview), and the novel *empirical Chernoff* confidence sets we introduce in Section 4.4. We observe that our bounds appear to adapt to the small variance of the $\mathrm{Beta}$ distribution, thus outperforming other methods for bounded distributions, even the state-of-the-art hedged capital bounds (Proposition A.5). We also report Student's confidence interval (4.2), which we recall is not a priori valid for non-Gaussian distributions; indeed, at level $\delta = 5\%$, the boundary crossing rate (i.e. the frequency of the event $\mu \notin \widehat{\Theta}_t^\delta$) for small samples is around $8\%$ for this last bound, whereas a valid confidence set would cap this rate to at most $\delta = 5\%$.

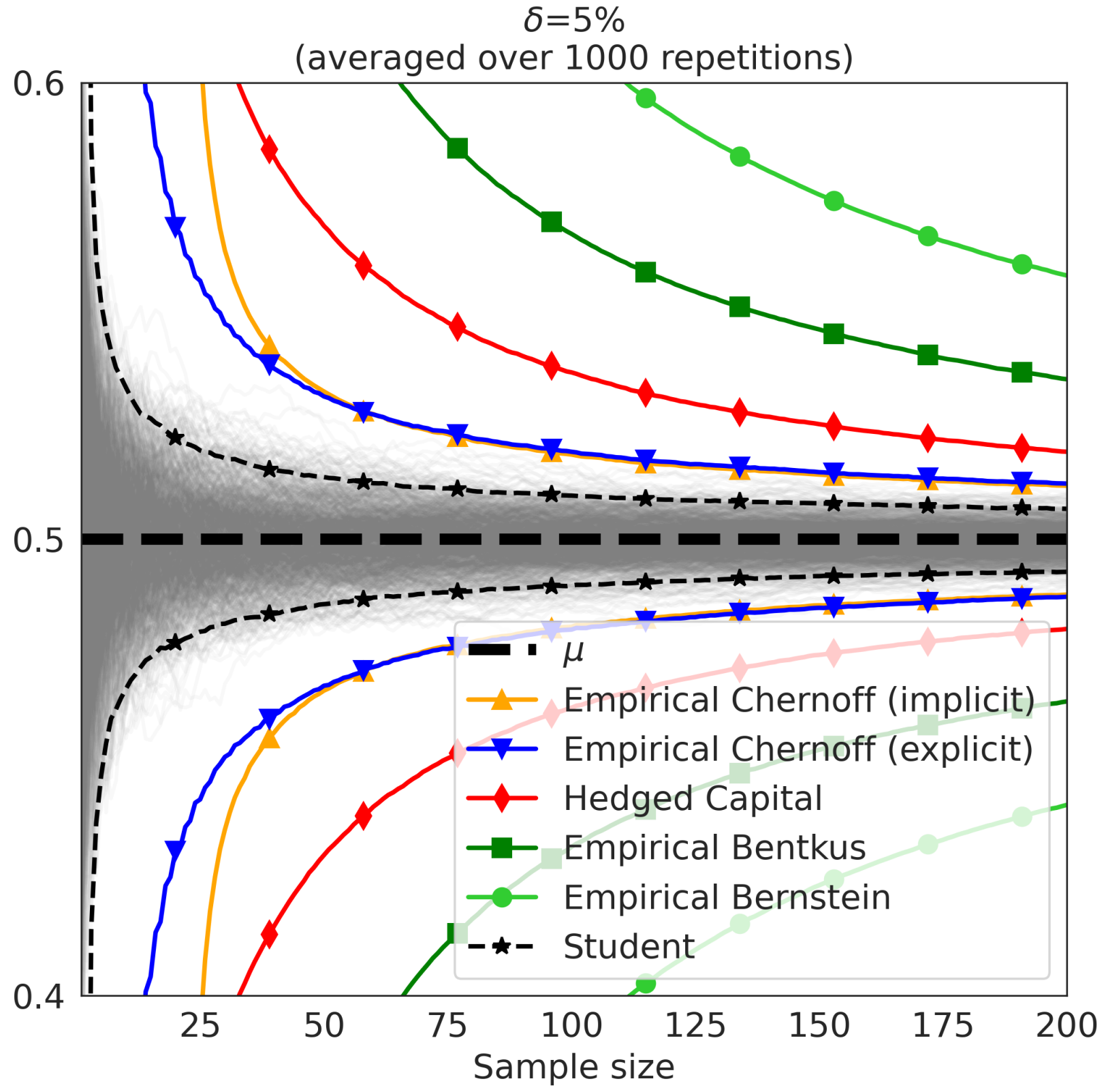

**Figure 4.1** – Comparison of median fixed sample confidence bounds around the mean for a $\mathrm{Beta}(50, 50)$ distribution, as a function of the sample size $t$, over $1000$ independent replicates. Grey lines are trajectories of empirical means $\widehat{\mu}_t$.

## 4.2 Second order sub-Gaussian distributions

**Motivating example.** The goal of *empirical* concentration is to replace the prior knowledge of a parameter $R$ controlling the size of the resulting confidence sets by a data-dependent estimate.

When this parameter is related to the variance of a distribution $\nu \in \mathcal{F}_{[\underline{B},\overline{B}]}$, the empirical variance process $(\widehat{\sigma}_t^2)_{t\in\mathbb{N}}$ itself is bounded, which allows to similarly control its deviations around the true variance. Of course, doing so bluntly would require the knowledge of a parameter related to the variance of $\widehat{\sigma}_t^2$ itself, typically a bound on the kurtosis of $\nu$. Interestingly, Lattimore (2017) analysed such concentration in the context of multiarmed bandits; however for empirical concentration, this merely shifts the required prior knowledge to higher moments, which may not be convenient in practice. Alternatively, we may exploit solely the boundedness to control all further moments (Maurer and Pontil, 2009; Kuchibhotla and Zheng, 2021).

In the unbounded case however, additional assumptions are required to derive a confidence set for $R$. To build intuition, let us consider a Gaussian random variable $Y \sim \mathcal{N}(\mu, R^2)$ for which we seek to establish a deviation bound around $\mu \in \mathbb{R}$ independent of $R \in \mathbb{R}_+^\star$. As seen in equation (4.2), Student's statistic provides a way to achieve this goal, however this approach is specific to the Gaussian and hardly generalisable; we apply instead the Cramér-Chernoff method (Proposition 1.15). The cumulant generating function (CGF) of the distribution of $Y$ is $\mathcal{K}\colon \lambda \in \mathbb{R} \mapsto \log \mathbb{E}_{Y\sim\nu}[\exp(\lambda(Y-\mu))] = R^2\lambda^2/2$, leading to a deviation bound of $R\sqrt{2t\log(1/\delta)}$ (Corollary 1.16). On the other hand, $(Y-\mu)^2/R^2$ follows a Chi-square distribution with a single degree of freedom, for which we denote the CGF as $\mathcal{K}_2\colon \lambda \mapsto -1/2\log(1-2\lambda)$, defined for $\lambda < 1/2$. Applying the Cramér-Chernoff method a second time with $\mathcal{K}_2(\lambda)$ for $\lambda < 0$, we establish after some simple algebra that:

$$\mathbb{P}\left( R \leqslant \sqrt{\frac{\frac{1}{t}\sum_{s=1}^t (Y_s-\mu)^2}{(\mathcal{K}_2^{\star,-})^{-1}\left(\frac{1}{t}\log\frac{1}{\delta}\right)}} \right) \geqslant 1-\delta\,, \tag{4.5}$$

where $\mathcal{K}_2^{\star,-}\colon u \in \mathbb{R}_+ \mapsto \sup_{\lambda\in\mathbb{R}_-} \lambda u - \mathcal{K}_2(\lambda)$ is the Fenchel-Legendre transform of the second order CGF $\mathcal{K}_2$. Finally, combining with a union bound the above equation with the sub-Gaussian bound obtained earlier enables to factor out $R$, resulting in a deviation bound on $\sum_{s=1}^t (Y_s-\mu)$ controlled only by $\sum_{s=1}^t (Y_s-\mu)^2$, $\delta$ and $t$, i.e.

$$\mathbb{P}\left( \sum_{s=1}^t Y_s - \mu \geqslant \sqrt{\frac{2\log\frac{2}{\delta}}{(\mathcal{K}_2^{\star,-})^{-1}\left(\frac{1}{t}\log\frac{2}{\delta}\right)}\sum_{s=1}^t (X_s-\mu)^2} \right) \leqslant \delta\,. \tag{4.6}$$

**Second order sub-Gaussian distributions.** Drawing inspiration from this simple example, we generalise this approach to distributions other than Gaussian by bounding their second order cumulant by that of a Chi-square distribution.

**Definition 4.1** (Second order sub-Gaussian). *Let $\nu$ be a distribution on $\mathbb{R}$ with expectation $\mu$. We define the second order (centred) cumulant generating function (CGF) of $\nu$ as*

$$\begin{aligned} \mathcal{K}_2 \colon \mathbb{R} &\longrightarrow \mathbb{R} \cup \{+\infty\} \\ \lambda &\longmapsto \log \mathbb{E}_{Y\sim\nu}\left[e^{\lambda(Y-\mu)^2}\right] . \end{aligned} \tag{4.7}$$

*Let $\rho, R \in \mathbb{R}_+^\star$. We define the family of $(\rho, R)$-**second order sub-Gaussian** distributions as*

$$\mathcal{F}^2_{\mathcal{G},\rho,R} = \left\{\forall\lambda \in \left(-\infty, \frac{1}{2R^2}\right),\ \mathcal{K}_2(\lambda) \leqslant -\frac{1}{2}\log\left(1 - 2R^2\left((\rho-1)\,\mathbb{1}_{\lambda<0} + 1\right)^2\lambda\right)\right\} . \tag{4.8}$$

*Moreover, we define the family of $\rho$-**second order sub-Gaussian** (SOSG) as*

$$\mathcal{F}^2_{\mathcal{G},\rho} = \bigcup_{R\in\mathbb{R}_+^\star} \mathcal{F}^2_{\mathcal{G},\rho,R} . \tag{4.9}$$

Let us clarify this definition. The second order CGF of a $(\rho, R)$-second order sub-Gaussian distribution $\nu$ is controlled on its right tail ($\lambda \in [0, 1/(2R^2))$) by the second order CGF of the square of a Gaussian random variable with variance $R^2$, and on its left tail ($\lambda \in \mathbb{R}_-^\star$) by that of Gaussian random variable with variance $(\rho R)^2$, i.e.

$$\mathcal{K}_2(\lambda) \leqslant \begin{cases} -\frac{1}{2}\log\left(1 - 2R^2\lambda\right) & \text{if } \lambda \in [0, \frac{1}{2R^2}) \\ -\frac{1}{2}\log\left(1 - 2(\rho R)^2\lambda\right) & \text{if } \lambda \in \mathbb{R}_-^\star \end{cases} \tag{4.10}$$

### Properties of second order sub-Gaussian distributions

Interestingly, the right tail control shows that this definition is a special case of the classical notion of $R$-sub-Gaussian distributions, which we summarise in the lemma below.

**Lemma 4.2** (Second order sub-Gaussian distributions and other families). *Let $\rho, R \in \mathbb{R}^\star$. Then we have*

$$\mathcal{F}^2_{\mathcal{G},\rho,R} \subset \mathcal{F}_{\mathcal{G},R} \quad \text{and} \quad \mathcal{F}_{\mathcal{N},R} \subset \mathcal{F}^2_{\mathcal{G},1,R} \,. \tag{4.11}$$

*Proof of Lemma 4.2.* The first inclusion is a direct consequence of Wainwright (2019, Theorem 2.6), which shows that the (first order) $R$-sub-Gaussian control $\mathbb{E}_{Y\sim\nu}[e^{\lambda(Y-\mu)}] \leqslant e^{\lambda^2 R^2/2}$ for $\lambda \in \mathbb{R}$ is equivalent to the second order control $\mathcal{K}(\lambda) \leqslant -1/2\log(1-2R^2\lambda)$ for $\lambda \in [0, 1/(2R^2))$. The second follows immediately from the expression of the CGF of the Chi-square distribution (in particular, the CGF control is tight for Gaussian distributions). ■

More generally, the case $\rho = 1$ means that both tails of the CGF $\mathcal{K}_2$ are controlled by the same scale parameter $R$. In general however, they may exhibit different scaling, which is why we explicitly introduce $\rho$ as the ratio of the left and right tail scaling parameters. The case $\rho = 1$ is actually a limiting case, as shown in the next lemma.

**Lemma 4.3** (Lower bound on the variance). *Let $\rho, R \in \mathbb{R}^\star_+$ and $\nu \in \mathcal{F}^2_{\mathcal{G},\rho,R}$. Then we have the inequalities $(\rho R)^2 \leqslant \mathbb{V}_{Y\sim\nu}[Y] \leqslant R^2$. In particular, $\rho \in (0, 1]$, and if $\rho = 1$ then $R^2 = \mathbb{V}_{Y\sim\nu}[Y]$.*

*Proof.* Since $\nu \in \mathcal{F}_{\mathcal{G},R}$ (Lemma 4.2), the upper bound $\mathbb{V}_{Y\sim\nu}[Y] \leqslant R^2$ follows from Lemma 1.12. The reverse bound is obtained from the left tail CGF control of Definition 4.1. Indeed, let $\mu \in \mathbb{R}$ and $\gamma\colon \lambda \in \mathbb{R}^\star_+ \mapsto \mathbb{E}_{Y\sim\nu}[\exp(-\lambda(Y-\mu)^2)] - (1+2(\rho R)^2\lambda)^{-1/2}$. By definition, $\gamma(\lambda) \leqslant 0$ for $\lambda \in \mathbb{R}^\star_+$ and a first order expansion around $\lambda = 0$ yields $\gamma(\lambda) = \lambda((\rho R)^2 - \mathbb{E}_{Y\sim\nu}[(Y-\mu)^2]) + o(\lambda)$. Since the dominating term must be nonpositive (and $\lambda > 0$), we obtain $\mathbb{V}_{Y\sim\nu}[Y] \geqslant (\rho R)^2$. ■

Arbel et al. (2020) introduced a unifying framework to study *strictly* sub-Gaussian distributions, for which the sub-Gaussian parameter $R^2$ is exactly the variance (rather than just an upper bound). In particular, they show that for some classical families (uniform, triangular, Bernoulli, Beta, etc.), only the symmetric distributions are strictly sub-Gaussian (e.g. $\mathcal{B}(1/2)$). A consequence of Lemma 4.3 is that SOSG distributions with $\rho = 1$ are special instances of strictly sub-Gaussian distributions. We extend their approach in the next proposition, which provides conditions for a sub-Gaussian distribution ($\nu \in \mathcal{F}_{\mathcal{G},R}$) to be SOSG ($\nu \in \mathcal{F}^2_{\mathcal{G},\rho}$).

**Proposition 4.4.** *Let $R \in \mathbb{R}_+^\star$ and $\rho \in (0,1]$. For a distribution $\nu \in \mathbb{L}^2(\mathbb{R})$ with expectation $\mu$ and variance $\sigma^2$, we recall that $\mathcal{K}\colon \lambda \in \mathbb{R} \mapsto \log \mathbb{E}_{Y\sim\nu}[\exp(\lambda(Y-\mu))]$ is the first order CGF of $\nu$ and we define the following mappings:*

$$\begin{aligned} \mathcal{H}\colon \mathbb{R}^\star &\longrightarrow \mathbb{R} & \text{and} \quad \mathcal{H}_2\colon \mathbb{R}_-^\star &\longrightarrow \mathbb{R} \\ \lambda &\longmapsto \frac{2\mathcal{K}(\lambda)}{\lambda^2} & \lambda &\longmapsto \frac{1-e^{-2\mathcal{K}_2(\lambda)}}{2\lambda}, \end{aligned} \tag{4.12}$$

*which we extend by continuity as $\mathcal{H}(0) = \mathcal{H}_2(0) = \sigma^2$. Then the following results hold.*

*(i) $\nu \in \mathcal{F}_{\mathcal{G},R}$ if and only if $\sup_{\lambda\in\mathbb{R}} \mathcal{H}(\lambda) \leqslant R^2$, and if so, $\nu \in \mathcal{F}^2_{\mathcal{G},\rho,R}$ if and only if $\inf_{\lambda\in\mathbb{R}_-} \mathcal{H}_2(\lambda) \geqslant (\rho R)^2$.*

*(ii) $\nu \in \mathcal{F}^2_{\mathcal{G},1}$ if and only if $\sup_{\lambda\in\mathbb{R}} \mathcal{H}(\lambda) = \inf_{\lambda\in\mathbb{R}_-} \mathcal{H}_2(\lambda)$.*

*(iii) If $\nu \in \mathcal{F}_{\mathcal{G},\sigma}$ and $(1 + 2\lambda \mathrm{d}\mathcal{K}_2/\mathrm{d}\lambda(\lambda)) \exp(-2\mathcal{K}_2(\lambda)) \leqslant 1$ for all $\lambda \in \mathbb{R}_-^\star$, then $\nu \in \mathcal{F}^2_{\mathcal{G},1}$.*

*Proof of Proposition 4.4.* The first result is simply a rewriting of the CGF control of sub-Gaussian and second order sub-Gaussian distributions with the mappings $\mathcal{H}$ and $\mathcal{H}_2$, i.e.

$$\forall \lambda \in \mathbb{R},\ \mathcal{K}(\lambda) \leqslant \sup_{\mathbb{R}} \mathcal{H} \frac{\lambda^2}{2} \quad \text{and} \quad \forall \lambda \in \mathbb{R}_-,\ \mathcal{K}_2(\lambda) \leqslant -\frac{1}{2} \log\left(1 - 2 \inf_{\mathbb{R}_-} \mathcal{H}_2 \lambda\right). \tag{4.13}$$

The second result is also a straightforward consequence of the above expressions as long as $\sup_{\lambda\in\mathbb{R}} \mathcal{H}(\lambda)$ and $\inf_{\lambda\in\mathbb{R}_-} \mathcal{H}_2(\lambda)$ are respectively finite and strictly positive. If these quantities are equal, we have that $\sup_{\lambda\in\mathbb{R}} \mathcal{H}(\lambda) \leqslant \mathcal{H}_2(0) = \sigma^2 < +\infty$ and $\inf_{\lambda\in\mathbb{R}_-} \mathcal{H}_2(\lambda) \geqslant \mathcal{H}(0) = \sigma^2 > 0$, which shows that $\nu$ is SOSG with the same scale parameter $R = \sigma$ on both tails, i.e. $\rho = 1$.

Finally, for $\lambda \in \mathbb{R}_-^\star$, we have

$$\frac{\mathrm{d}\mathcal{H}_2}{\mathrm{d}\lambda}(\lambda) = \frac{-1 + \left(1 + 2\lambda \frac{\mathrm{d}\mathcal{K}_2}{\mathrm{d}\lambda}(\lambda)\right) e^{-2\mathcal{K}(\lambda)}}{2\lambda^2}. \tag{4.14}$$

Therefore, the condition (iii) shows that $\mathcal{H}_2$ is nonincreasing, and therefore we have the equalities $\sigma^2 = \mathcal{H}(0) = \inf_{\lambda\in\mathbb{R}_-} \mathcal{H}_2(\lambda)$. By assumption, $\nu \in \mathcal{F}_{\mathcal{G},\sigma}$, hence $\sup_{\lambda\in\mathbb{R}} \mathcal{H}(\lambda) = \sigma^2$, so it follows from (ii) that $\nu \in \mathcal{F}^2_{\mathcal{G},1}$. ■

Using this characterisation, we now provide several examples of classical distributions that are second order sub-Gaussian (the proof, although elementary, involves quite lengthy calculations, which we defer to Appendix C.1).

**Proposition 4.5** (Examples). *The following distributions are second order sub-Gaussian.*

(*i*) ***Gaussian.*** *For* $\mu \in \mathbb{R}$, $\sigma \in \mathbb{R}_+^\star$, *we have* $\mathcal{N}(\mu, \sigma^2) \in \mathcal{F}^2_{\mathcal{G},1,\sigma}$.

(*ii*) ***Bernoulli.*** *For* $\underline{B}, \overline{B} \in \mathbb{R}$, $p \in [0,1]$ *and* $q = 1-p$, *we have* $p\delta_{\underline{B}} + q\delta_{\overline{B}} \in \mathcal{F}^2_{\mathcal{G},\rho_p,R_p}$ *with*

$$R_p = \left(\overline{B} - \underline{B}\right)\left(\sqrt{\frac{\frac{1}{2} - p \wedge q}{\log\left(\frac{p\vee q}{p\wedge q}\right)}}\mathbb{1}_{p\neq\frac{1}{2}} + \frac{1}{2}\mathbb{1}_{p=\frac{1}{2}}\right) \quad \text{and} \quad \rho_p = p \wedge q\sqrt{\frac{\log\left(\frac{p\vee q}{p\wedge q}\right)}{\frac{1}{2} - p\wedge q}}\mathbb{1}_{p\neq\frac{1}{2}} + \mathbb{1}_{p=\frac{1}{2}}\,. \tag{4.15}$$

*In particular,* $\mathcal{B}(p)\mathcal{F}^2_{\mathcal{G},1}$ *if and only if* $p = 1/2$.

(*iii*) ***Uniform.*** *For* $\underline{B}, \overline{B} \in \mathbb{R}$, *we have* $\mathcal{U}([\underline{B}, \overline{B}]) \in \mathcal{F}^2_{\mathcal{G},1,\sigma}$ *with* $\sigma^2 = (\overline{B} - \underline{B})^2/12$.

(*iv*) ***Symmetric triangular.*** *For* $\underline{B}, \overline{B} \in \mathbb{R}, \mathcal{U}([\underline{B}, \overline{B}]) \in \mathcal{F}^2_{\mathcal{G},1}$, *the distribution with p.d.f.* $p\colon y \in \mathbb{R} \mapsto 4/(\overline{B} - \underline{B})^2((y - \underline{B})\mathbb{1}_{y\in[\underline{B},(\underline{B}+\overline{B})/2]} + (\overline{B} - y)\mathbb{1}_{y\in((\underline{B}+\overline{B})/2,\overline{B}]})$ *is in* $\mathcal{F}^2_{\mathcal{G},\rho,\sigma}$ *with* $\sigma^2 = (\overline{B} - \underline{B})^2/24$ *and* $\rho = \sqrt{3/\pi} \approx 0.977$.

(*v*) ***Gaussian mixture.*** *For* $\mu_0, \mu_1 \in \mathbb{R}$ *and* $\sigma_0, \sigma_1 \in \mathbb{R}_+^\star$, $p \in (0,1)$ *and* $q = 1-p$, *the mixture of* $\mathcal{N}(\mu_0, \sigma_0^2)$ *and* $\mathcal{N}(\mu_1, \sigma_1^2)$ *with weights* $p$ *and* $q$ *is* $\mathcal{F}^2_{\mathcal{G},\rho_p,R_p}$ *with*

$$c_p = \frac{q-p}{4(\log q - \log p)}\mathbb{1}_{p\neq\frac{1}{2}} + \frac{1}{8}\mathbb{1}_{p=\frac{1}{2}}\,, \tag{4.16}$$

$$R_p = \sqrt{\sigma_0^2 \vee \sigma_1^2 + 2c_p} \quad \text{and} \quad \rho_p = \frac{\sigma_0 \wedge \sigma_1}{\sqrt{\sigma_0^2 \vee \sigma_1^2 + 2c_p}}\,. \tag{4.17}$$

Rather than providing an exhaustive list, this last proposition intents to show that many common distributions are indeed SOSG. For all examples except the Gaussian mixtures, the ratio $\rho$ is tight, as it results from a direct calculation. In this last case, what we provide is only a lower bound on the optimal ratio $\rho$, which is likely not tight. This is not directly due to a limitation of our approach but it is rather a consequence of the challenging problem of finding sharp sub-Gaussian control for general mixture distributions. In particular, the bound reported in Chafai and Malrieu (2010) and used in the proof, which to the best of our knowledge is the sharpest in the literature, depends only on the maximum variance of the mixture components, regardless of their respective weights in the mixture.

We acknowledge that computing the exact ratio $\rho$ often involves intricate formal calculations. For symmetric Beta distributions, we only conjecture that the optimal ratio is $\rho = 1$, which is consistent with their strict sub-Gaussianity. Nonetheless, this claim is supported with accurate empirical evidence, which we provide in Appendix C.1.

**Conjecture 4.6** (Symmetric Beta distributions are second order sub-Gaussian). *Let* $\alpha \in \mathbb{R}_+^\star$. *Then* $\mathrm{Beta}\,(\alpha, \alpha) \in \mathcal{F}^2_{\mathcal{G},1,\sigma}$ *with* $\sigma = 1/(4(2\alpha + 1))$.

In the light of these examples, it is tempting to assume that the case $\rho = 1$, is related to the symmetry of the distributions at hand. Arbel et al. (2020) disproved this for strictly sub-Gaussian distributions, by exhibiting examples of sub-Gaussian distributions that were respectively symmetric but not strictly sub-Gaussian, and asymmetric and yet strictly sub-Gaussian. However, these counterexamples involved rather specific combinations of Dirac distributions, so we believe our notion of second order sub-Gaussian distributions capture a vast, nonparametric range of natural distributions.

**Remark 4.7** (Link with log-concavity).*An alternative nonparametric specification for statistical estimation is that of* shape-constrained *distributions, which imposes e.g. log-concave density functions.* Hillion et al. (2019) *showed that the only strongly log-concave distribution with strong log-concavity parameter equal to its variance is the corresponding Gaussian measure. A direct consequence of this is that the only SOSG distributions with* $\rho = 1$ *and strongly log-concave densities with respect to the Lebesgue measure on* $\mathbb{R}$ *are the Gaussian distributions. However, as illustrated above, the SOSG definition covers many other distributions, in particular ones with compact supports, which is incompatible with the strong log-concavity.*

In the next section, we derive closed-form concentration bounds for the family of $(\rho, R)$-second order sub-Gaussian but also for the larger family $\mathcal{F}^2_{\mathcal{G},\rho}$, that is independently of the actual value of the scale parameter $R$ (the same way empirical Bernstein inequalities extend the classical Bernstein inequalities for bounded distributions). In this sense, $\rho$ can be seen as a measure of how much the concentration of $Y^2$ "leaks" to the concentration of $Y$.

## 4.3 Empirical Chernoff concentration

We first state concentration bounds for fixed samples, which extend those of Corollary 1.16, as we believe the novel framework of second order sub-Gaussian distributions may offer appealing modelling alternatives even for classical statistics. We then derive time-uniform bounds, which extends those of Corollary 1.21. We also detail the resulting confidence sets, which are described by implicit inequalities. Finally, we introduce a variant of the second order sub-Gaussian control that allows for fully explicit confidence sets.

### Fixed sample concentration

As seen in the Gaussian example in Section 4.2, the Cramér-Chernoff method in the context of the second order CGF control of Definition 4.1 involves computing the inverse of the Fenchel-Legendre transform of the CGF of a Chi-square distribution. This can be obtained in closed form using the Lambert $W$ function, which we have already encountered in the tuning of the mixing parameter for the sub-Gaussian method of mixtures (Lemma 1.23) and the analysis of generic UCB algorithms (Corollary 1.34). It is a multivalued function over the complex plane defined implicitly as the solutions in variable $\omega \in \mathbb{C}$ of the equation $\omega e^{\omega} = z$ for a given $z \in \mathbb{C}$. For our use case, we only consider the principal branch $W_0$ and the first negative branch $W_{-1}$, which give the unique two real solutions for $z \in (-\frac{1}{e}, 0)$, with $W_0(z) > -1$ and $W_{-1}(z) < -1$. In addition, both branches are joining at $z = -1/e$, i.e. $\lim_{z \to -1/e} W_0(z) = \lim_{z \to -1/e} W_{-1}(z) = 1$.

**Theorem 4.8** (Empirical Chernoff concentration). *Let $\rho \in (0,1]$, $R \in \mathbb{R}_+^\star$, $t \in \mathbb{N}$ and $(Y_s)_{s=1}^t$ a sequence of i.i.d. random variables drawn from $\nu$ with expectation $\mu \in \mathbb{R}$. Let $\delta \in (0,1)$ and*

$$S_t = \sum_{s=1}^t Y_s - \mu \quad \textit{and} \quad Q_t = \sum_{s=1}^t (Y_s - \mu)^2 \,. \tag{4.18}$$

$(i)$ ***Second order bounds.*** *If $\nu \in \mathcal{F}^2_{\mathcal{G},\rho,R}$ then*

$$\mathbb{P}\left(Q_t \leqslant -W_0\left(\frac{-\delta^{\frac{2}{t}}}{e}\right)(\rho R)^2 t\right) \leqslant \delta \quad \textit{and} \quad \mathbb{P}\left(Q_t \geqslant -W_{-1}\left(\frac{-\delta^{\frac{2}{t}}}{e}\right) R^2 t\right) \leqslant \delta \,, \,. \tag{4.19}$$

$(ii)$ ***Self-normalised bound.*** *If $\nu \in \mathcal{F}^2_{\mathcal{G},\rho}$ then*

$$\mathbb{P}\left(\frac{S_t}{\sqrt{Q_t}} \geqslant \frac{1}{\rho}\sqrt{\frac{2\log\frac{2}{\delta}}{-W_0\left(\frac{-(\delta/2)^{\frac{2}{t}}}{e}\right)}}\right) \leqslant \delta \,. \tag{4.20}$$

*Proof of Theorem 4.8.* Let $\nu \in \mathcal{F}^2_{\mathcal{G},\rho,R}$. The second order sub-Gaussian condition satisfied by $\nu$ coincides with the CGF control of the generic Cramér-Chernoff bound of Proposition 1.15 applied to (i) the process $(Q_s/(\rho R)^2)_{s=1}^t$ with $\psi(\lambda) = -1/2\log(1-2\lambda)$ if $\lambda \in \mathcal{I}_- = \mathbb{R}^\star_-$ and (ii) the process $(Q_s/R^2)_{s=1}^t$ with $\psi(\lambda) = -1/2\log(1-2R^2\lambda)$ if $\lambda \in \mathcal{I}_+ = (0, 1/2)$. Hence, to prove the second order bounds, we need to compute the inverses of the Fenchel-Legendre transforms $\psi^{\star,-}(u) = \sup_{\lambda\in\mathcal{I}_-} \lambda u - \psi(\lambda)$ and $\psi^{\star,+}(u) = \sup_{\lambda\in\mathcal{I}_+} \lambda u - \psi(\lambda)$ for some $u \in \mathbb{R}$.

**Left tail control.** For $\lambda \in \mathcal{I}_-$, let $g(\lambda) = \lambda u + 1/2\log(1-2\lambda)$, which defines a differentiable function satisfying $\mathrm{d}g/\mathrm{d}\lambda(\lambda) = u - 1/(1-2\lambda)$. Setting this derivative to zero yields $\lambda^\star = 1/2(1-1/u)$, which is indeed in $\mathcal{I}_-$ provided that $u \in (0,1)$ and thus $\psi^{\star,-}(u) = 1/2(u-1) - 1/2\log(u)$. Solving $\psi^{\star,-}(u) = y$ in $y \in \mathbb{R}^\star_+$ yields, after simple algebra $-ue^{-u} = -e^{-(2y+1)} \in (-1/e, 0)$. Since we have $-uu > -1$, the solution is given by the principal branch of the Lambert $W$ function, i.e. $u = -W_0(-e^{-(2y+1)})$. We conclude by letting $y = 1/t\log(1/\delta)$ for $\delta \in (0,1)$.

**Right tail control.** For $\lambda \in \mathcal{I}_+$, we define $g(\lambda) = \lambda u + 1/2\log(1-2\lambda)$. Similarly to the left tail control, setting the derivative of $g$ to zero yields $\lambda^\star = 1/2(1-1/u)$, which belongs to $\mathcal{I}_+$ provided that $u > 1$. Solving $\psi^{\star,+}(u) = y$ yields a similar equation $-ue^{-u} = -e^{-(2y+1)}$. However, this time we have $-u < -1$, and hence $u = -W_{-1}(-e^{-(2y+1)})$.

**Self-normalised bound.** Thanks to Lemma 4.2, $\nu$ is also $R$-sub-Gaussian, and therefore the event $\mathcal{E}^\delta_1 = \{S_t \geqslant R\sqrt{2t\log(1/\delta)}\}$ holds with probability at most $\delta$ (Corollary 1.16). We lety

$$\widehat{R}^\delta_t = \frac{1}{\rho}\sqrt{\frac{Q_t}{-tW_0\left(\frac{-(\delta)^{\frac{2}{t}}}{e}\right)}} \quad \text{and} \quad \mathcal{E}^\delta_2 = \left\{R \geqslant \widehat{R}^\delta_t\right\}. \tag{4.21}$$

The left tail control implies that the event $\mathcal{E}^\delta_2$ holds with probability at most $\delta$. By a simple union bound argument, the intersection of the complement events $\overline{\mathcal{E}^{\delta/2}_1} \cap \overline{\mathcal{E}^{\delta/2}_2}$ holds with probability at least $1-\delta$. We conclude by noting that $\overline{\mathcal{E}^{\delta/2}_1} \cap \overline{\mathcal{E}^{\delta/2}_2} \subset \{S_t \leqslant \widehat{R}^\delta_t\sqrt{2t\log(1/\delta)}\}$. ■

The second order bounds are reminiscent of Bernstein-like inequalities in Birgé and Massart (1998, Lemma 8), Laurent and Massart (2000, Lemma 1) and Maillard (2019b, Lemma 2.5). However, these bounds relied on relaxations of the Chi-square CGF to obtain easier expressions instead of the tighter control by the Lambert $W$ function. In the limit of large samples $t \to +\infty$, $\delta^{2/t}$ goes to 1 and thus $-W_k(-\delta^{2/t}/e) \to 1$ for $k \in \{-1, 0\}$. Furthermore, Corless et al. (1996) gives the following asymptotic expansion $W_k(z) = -1 + \varepsilon_k\sqrt{2(ez+1)} + o(\sqrt{z+\frac{1}{e}})$ when $z \to -\frac{1}{e}$ (from above) and $\varepsilon_0 = 1, \varepsilon_{-1} = -1$. Using the Taylor expansion $\delta^{2/t} = \exp(\frac{2}{t}\log\delta) =$

$1 - 2/t \log(1/\delta) + o(1/t)$, we obtain the following asymptotic bounds:

$$\mathbb{P}\left(Q_t \leqslant (\rho R)^2\, t\left(1 - 2\sqrt{\frac{1}{t}\log\frac{1}{\delta}}\right) + o(1)\right) \leqslant \delta \tag{4.22}$$

$$\mathbb{P}\left(Q_t \geqslant R^2 t\left(1 + 2\sqrt{\frac{1}{t}\log\frac{1}{\delta}}\right) + o(1)\right) \leqslant \delta\,. \tag{4.23}$$

The self-normalised bound does not depend on $R$, which is instead replaced by a data-dependent estimator $\widehat{R}_t^\delta$ proportional to $\sqrt{Q_t/t}$, and thus represents an empirical extension of the classical Cramér-Chernoff bound for SOSG distributions. The price to pay for dropping the dependency on $R$ comes in the form of $1/\rho$, which measures the tightness of the estimation of $R$ using the quadratic sum $Q_t$.

### Time-uniform concentration

We now turn to the extension of Corollary 1.21 to the family of SOSG distributions. Similarly to Theorem 4.8, we provide two second order bounds and a self-normalised bound that combines the first and second order sub-Gaussian properties. Before we state this result, we recall the definition of two special functions known as the *confluent hypergeometric functions* (Magnus et al. (1953), Abramowitz and Stegun (1968, Chapter 13)).

**Definition 4.9** (Confluent hypergeometric functions). *Consider the half space of $\mathbb{R}^2$ defined by $\mathcal{H} = \{(b, c) \in \mathbb{R}^2,\ b < c\}$. We define **Tricomi's confluent hypergeometric function** as*

$$\begin{aligned} U\colon \mathcal{H} \times \mathbb{R}_+^\star &\longrightarrow \mathbb{R}_+^\star \\ (b, c; z) &\longmapsto \frac{1}{\Gamma(b)} \int_0^{+\infty} u^{b-1}(1+u)^{c-b-1} e^{-zu} du\,, \end{aligned} \tag{4.24}$$

*and **Kummer's confluent hypergeometric function** as*

$$\begin{aligned} M\colon \mathcal{H} \times \mathbb{R}_+^\star &\longrightarrow \mathbb{R}_+^\star \\ (b, c; z) &\longmapsto \frac{\Gamma(c)}{\Gamma(b)\Gamma(c-b)} \int_0^{1} u^{b-1}(1-u)^{c-b-1} e^{zu} du\,. \end{aligned} \tag{4.25}$$

Fast numerical approximations of these special functions are implemented in many scientific libraries, e.g. in Python *scipy.special.hyperu* (Tricomi) and *scipy.special.hyp1f1* (Kummer).

For $\beta, \gamma, \zeta \in \mathbb{R}_+^\star$ such that $\beta < \gamma$, we define the auxiliary functions

$$G^{\mathrm{T}}_{\beta,\gamma,\zeta,t} \colon \mathbb{R}_+^\star \longrightarrow \mathbb{R}_+^\star \quad z \longmapsto \frac{U\left(\beta, \gamma + \frac{t}{2}; \zeta + \frac{z}{2}\right)}{U(\beta,\gamma;\zeta)} \qquad \text{and} \qquad G^{\mathrm{K}}_{\beta,\gamma,\zeta,t} \colon \mathbb{R}_+^\star \longrightarrow \mathbb{R}_+^\star \quad z \longmapsto \frac{M\left(\beta, \gamma + \frac{t}{2}; \zeta + \frac{z}{2}\right)}{M(\beta,\gamma;\zeta)} \,. \tag{4.26}$$

Note that $G^{\mathrm{T}}_{\beta,\gamma,\zeta,t}$ and $G^{\mathrm{K}}_{\beta,\gamma,\zeta,t}$ are respectively decreasing and increasing, and for a fixed $z \in \mathbb{R}_+^\star$, $t \mapsto G^{\mathrm{T}}_{\beta,\gamma,\zeta,t}(z)$ and $t \mapsto G^{\mathrm{K}}_{\beta,\gamma,\zeta,t}(z)$ are respectively increasing and decreasing. Moreover, we observe the following boundary conditions:

$$G^{\mathrm{T}}_{\beta,\gamma,\zeta,0}(0) = 1\,, \qquad \lim_{z\to+\infty} G^{\mathrm{T}}_{\beta,\gamma,\zeta,t}(z) = 0 \quad \text{and} \quad \lim_{t\to+\infty} G^{\mathrm{T}}_{\beta,\gamma,\zeta,t}(z) = +\infty\,, \tag{4.27}$$

$$G^{\mathrm{K}}_{\beta,\gamma,\zeta,0}(0) = 1\,, \qquad \lim_{z\to+\infty} G^{\mathrm{K}}_{\beta,\gamma,\zeta,t}(z) = +\infty \quad \text{and} \quad \lim_{t\to+\infty} G^{\mathrm{K}}_{\beta,\gamma,\zeta,t}(z) = 0\,. \tag{4.28}$$

**Theorem 4.10** (Empirical Chernoff time-uniform concentration). *Let $\rho \in (0,1]$, $R \in \mathbb{R}_+^\star$ and $(Y_t)_{t\in\mathbb{N}}$ be a sequence of i.i.d. random variables drawn from $\nu$ with expectation $\mu \in \mathbb{R}$. Let $\alpha, \beta, \gamma, \zeta \in \mathbb{R}_+^\star$ such that $\beta < \gamma$, $t_0 \in \mathbb{N}$, $\delta \in (0,1)$ and*

$$(S_t)_{t\in\mathbb{N}} = \left(\sum_{s=1}^{t} Y_s - \mu\right)_{t\in\mathbb{N}} \quad \text{and} \quad (Q_t)_{t\in\mathbb{N}} = \left(\sum_{s=1}^{t} (Y_s - \mu)^2\right)_{t\in\mathbb{N}} \,. \tag{4.29}$$

(*i*) ***Second order bounds.*** *If $\nu \in \mathcal{F}^2_{\mathcal{G},\rho,R}$ and $G^{\mathrm{T}}_{\beta,\gamma,\zeta,t_0}(0) > 1/\delta$ then*

$$\mathbb{P}\left(\exists t \geqslant t_0,\ Q_t \leqslant (\rho R)^2 \left(G^{\mathrm{T}}_{\beta,\gamma,\zeta,t}\right)^{-1}\left(\frac{1}{\delta}\right)\right) \leqslant \delta\,, \tag{4.30}$$

*and if $G^{\mathrm{K}}_{\beta,\gamma,\zeta,t_0}(0) < 1/\delta$ then*

$$\mathbb{P}\left(\exists t \geqslant t_0,\ Q_t \geqslant R^2 \left(G^{\mathrm{K}}_{\beta,\gamma,\zeta,t}\right)^{-1}\left(\frac{1}{\delta}\right)\right) \leqslant \delta\,. \tag{4.31}$$

(*ii*) ***Self-normalised bound.*** *If $\nu \in \mathcal{F}^2_{\mathcal{G},\rho}$ and $G^{\mathrm{T}}_{\beta,\gamma,\zeta,t_0}(0) > 2/\delta$ then*

$$\mathbb{P}\left(\exists t \geqslant t_0,\ \frac{S_t}{\sqrt{Q_t}} \geqslant \frac{1}{\rho}\sqrt{\frac{2(t+\alpha)\log\left(\frac{2}{\delta}\sqrt{1+\frac{t}{\alpha}}\right)}{\left(G^{\mathrm{T}}_{\beta,\gamma,\zeta,t}\right)^{-1}\left(\frac{2}{\delta}\right)}}\right) \leqslant \delta\,. \tag{4.32}$$

*Proof of Theorem 4.10.* Let $\nu \in \mathcal{F}^2_{\mathcal{G},\rho,R}$. We will use the CGF control of Definition 4.1 to construct suitable nonnegative supermartingale via the method of mixtures.

**Left tail control.** For $\lambda \in \mathbb{R}_-^\star$ and $t \in \mathbb{N}$, we let $M_t^\lambda = (1-2\lambda)^{t/2}\exp(\lambda Q_t)$ (the general case with scaling parameter $\rho R$ can be deduced by considering $Q_t/(\rho R)^2$ instead). The process $(Q_t)_{t\in\mathbb{N}}$ is measurable with respect to the natural filtration of $(Y_t)_{t\in\mathbb{N}}$, denoted by $(\mathcal{G}_t)_{t\in\mathbb{N}}$, integrable and

$$\mathbb{E}\left[M_{t+1}^\lambda \,\middle|\, \mathcal{G}_t\right] = \mathbb{E}\left[\underbrace{(1-2\lambda)^{\frac{t}{2}}\, e^{\lambda Q_t}}_{\mathcal{G}_t\text{-measurable}}\, \underbrace{\sqrt{1-2\lambda}e^{\lambda Y_{t+1}^2}}_{\mathcal{G}_t\text{-independent}} \,\middle|\, \mathcal{G}_t\right] = M_t^\lambda\, \underbrace{\mathbb{E}\left[\sqrt{1-2\lambda}e^{\lambda Y_{t+1}^2}\right]}_{\leqslant 1} \leqslant M_t^\lambda\,. \tag{4.33}$$

Therefore $(M_t^\lambda)_{t\in\mathbb{N}}$ is a nonnegative supermartingale. It is convenient to use the change of variable $u = -2\lambda \in \mathbb{R}_+^\star$ before the mixing construction, i.e. we let $M_t^u = (1+u)^{t/2}\exp(-u/2Q_t)$. For $\beta, \gamma, \zeta \in \mathbb{R}_+^\star$, we consider the distribution with the following density (with respect to the Lebesgue measure):

$$\begin{aligned} p\colon \mathbb{R}_+^\star &\longrightarrow \mathbb{R}_+^\star \\ u &\longmapsto \frac{u^{\beta-1}\,(1+u)^{\gamma-\beta-1}\,e^{-\zeta u}}{\Gamma(\beta)U\,(\beta,\gamma;\zeta)}\,, \end{aligned} \tag{4.34}$$

which satisfies $\int_0^{+\infty} p(u)du = 1$ (Definition 4.9). As in the proof of Corollary 1.21, the mixture process $(M_t)_{t\in\mathbb{N}} = (\int_0^{+\infty} M_t^u p(u)du)_{t\in\mathbb{N}}$ remains a nonnegative supermartingale and is related to the quadratic process $(Q_t)_{t\in\mathbb{N}}$ by the identity $M_t = G^{\mathrm{T}}_{\beta,\gamma,\zeta,t}(Q_t)$ for all $t \in \mathbb{N}$. Theorem 1.20 shows that we may obtain a time-uniform bound on $(Q_t)_{t\in\mathbb{N}}$ by inverting $G^{\mathrm{T}}_{\beta,\gamma,\zeta,t}$. Since $(t,z) \mapsto G^{\mathrm{T}}_{\beta,\gamma,\zeta,t}(z)$ is decreasing in $z$ (to $0$) and increasing in $t$ (to $+\infty$), it is enough to find $t_0 \in \mathbb{N}$ such that $G^{\mathrm{T}}_{\beta,\gamma,\zeta,t_0}(0) > 1/\delta$ to guarantee the existence of $(G^{\mathrm{T}}_{\beta,\gamma,\zeta,t})^{-1}(1/\delta)$ for all $t \geqslant t_0$.

**Right tail control.** For $\lambda \in (0, 1/2)$ and $t \in \mathbb{N}$, we let $M_t^\lambda = (1-2\lambda)^{t/2}\exp(\lambda Q_t)$ (again, the general case with scaling parameter $R$ can be deduced by considering $Q_t/R^2$ instead). The process $(M_t^\lambda)_{t\in\mathbb{N}}$ still defines a nonnegative supermartingale. However we now consider the change of variable $u = 2\lambda \in (0,1)$, $M_t^u = (1-u)^{t/2}\exp(u/2Q_t)$ and the density defined as

$$\begin{aligned} p\colon (0,1) &\longrightarrow \mathbb{R}_+^\star \\ u &\longmapsto \frac{\Gamma(\gamma)u^{\beta-1}\,(1-u)^{\gamma-\beta-1}\,e^{\zeta u}}{\Gamma(\beta)\Gamma(\gamma-\beta)M\,(\beta,\gamma;\zeta)}\,, \end{aligned} \tag{4.35}$$

with $\int_0^1 p(u)du = 1$ by definition of Kummer's function $M$. Again, the mixture process $(M_t)_{t\in\mathbb{N}} = (\int_0^1 M_t^u p(u)du)_{t\in\mathbb{N}}$ is still a nonnegative supermartingale and satisfies the identity $M_t = G^{\mathrm{K}}_{\beta,\gamma,\zeta,t}(Q_t)$. Since $(t,z) \mapsto G^{\mathrm{K}}_{\beta,\gamma,\zeta,t}(z)$ is increasing in $z$ (to $+\infty$) and decreasing in $t$ (to $0$), it is enough to find $t_0 \in \mathbb{N}$ such that $G^{\mathrm{K}}_{\beta,\gamma,\zeta,t_0}(0) < 1/\delta$ to guarantee the existence of $(G^{\mathrm{K}}_{\beta,\gamma,\zeta,t})^{-1}(1/\delta)$ for all $t \geqslant t_0$.

**Self-normalised bound.** Let $\alpha \in \mathbb{R}_+^\star$. By Lemma 4.2, $\nu$ is also $R$-sub-Gaussian, and therefore the event $\mathcal{E}_1^\delta = \{\exists t \geqslant t_0,\ S_t \geqslant R\sqrt{2(t+\alpha)\log(\sqrt{1+t/\alpha}/\delta)}\}$ holds with probability at most $\delta$ (Corollary 1.21). We let

$$\widehat{R}_t^\delta = \frac{1}{\rho}\sqrt{\frac{Q_t}{\left(G^{\mathrm{T}}_{\beta,\gamma,\zeta,t}\right)^{-1}\left(\frac{1}{\delta}\right)}} \quad \text{and} \quad \mathcal{E}_2^\delta = \left\{\exists t \geqslant t_0,\ R \geqslant \widehat{R}_t^\delta\right\}. \tag{4.36}$$

The left tail control implies that the event $\mathcal{E}_2^\delta$ holds with probability at most $\delta$. By a simple union bound argument, the intersection of the complement events $\overline{\mathcal{E}_1^{\delta/2}} \cap \overline{\mathcal{E}_2^{\delta/2}}$ holds with probability at least $1-\delta$. Finally, we conclude by noting that the following inclusion holds: $\overline{\mathcal{E}_1^{\delta/2}} \cap \overline{\mathcal{E}_2^{\delta/2}} \subset \{\exists t \geqslant t_0,\ S_t \leqslant \widehat{R}_t^\delta\sqrt{2(t+\alpha)\log(\sqrt{1+t/\alpha}/\delta)}\}$. ■

**Remark 4.11**. *Note the similarity with the fixed sample case (Theorem 4.8), where the lower and upper second order bounds involve different branches of the Lambert $W$ function; here, Tricomi's $U$ and Kummer's $M$ functions are two linearly independent solutions of a second order differential equation known as Kummer's equation. In fact, the $W$ function and the hypergeometric functions are related (Rathie and Ozelim, 2022), suggesting a deeper connection between the two bounds.*

**Mixing parameters tuning.** Compared to the first order sub-Gaussian mixture, our second order construction depends on three mixing parameters $\beta$, $\gamma$ and $\zeta$, for which we now suggest a natural tuning. We recall that the mixing measure for the first order sub-Gaussian bound is $\Lambda \sim \mathcal{N}(0, 1/\alpha)$ for some $\alpha \in \mathbb{R}_+^\star$, and thus $\Lambda^2$ follows a $\mathrm{Gamma}(1/2, 2/\alpha)$ distribution, the density of which is $p\colon u \in \mathbb{R}_+^\star \mapsto \sqrt{\alpha/(2\pi)}u^{-1/2}e^{-\alpha u/2}$. This is precisely the mixing density used in for the second order mixture with $\beta = 1/2$, $\gamma = 3/2$ and $\zeta = \alpha/2$. We suggest these particular values, with either the standard default choice $\alpha = 1$ or $\alpha \approx 0.12 t_0$ for some $t_0 \in \mathbb{N}$ as in the tuning of the first order bound. In Section 4.5, we further support this heuristic with empirical evidence.

**Initial condition $t_0$.** A practical difference with the first order mixture bound is that the results of Theorem 4.10 are valid uniformly for all $t$ larger than an initial condition $t_0 \in \mathbb{N}$ such that $G^{\mathrm{T}}_{\beta,\gamma,\zeta,t_0}(0) > 1/\delta$, rather than for all $t \in \mathbb{N}$. In particular, this prevents the choice $t_0 = 0$ since $G^{\mathrm{T}}_{\beta,\gamma,\zeta,0}(0) = 1$. However, we argue that this restriction is quite mild in practice. First, notice that $\lim_{t_0\to+\infty} G^{\mathrm{T}}_{\beta,\gamma,\zeta,t_0}(0) = +\infty$, so a large enough $t_0$ is bound to satisfy the desired inequality. Moreover, for $z \to 0^+$, we have $U(b,c;z) \sim \Gamma(c-1)/\Gamma(b) z^{1-c}$ for $c > 1$ and $b < c$ (Abramowitz and Stegun, 1968, Chapter 13.5), and thus $G^{\mathrm{T}}_{\beta,\gamma,\zeta,t_0}(0) \sim \Gamma(\gamma + t_0/2 - 1)/\Gamma(\gamma-1)\zeta^{-t_0/2}$, which is rapidly growing with $t_0$ (a similar argument holds for the upper bound and the condition $G^{\mathrm{K}}_{\beta,\gamma,\zeta,t_0}(0) < 1/\delta$). In practice, only small values of $t_0$ (typically less than $5$) are necessary, therefore this additional requirement does not constitute an impediment.

**Asymptotic behaviour of $G^{\mathrm{T}}_{\beta,\gamma,\zeta,t}$.** Finally, we highlight that the large sample behaviour of the second order term $(G^{\mathrm{T}}_{\beta,\gamma,\zeta,t})^{-1}$, despite its rather intricate definition, is actually quite simple.

**Lemma 4.12**. *Let $\beta,\gamma,\zeta \in \mathbb{R}_+^\star$ such that $\beta < \gamma$. Then $(G^{\mathrm{T}}_{\beta,\gamma,\zeta,t})^{-1}(1/\delta) \sim t + 2(\delta/U(\beta,\gamma;\zeta))^2$ when $t \to +\infty$ and $\delta \to 0$. In particular if $\delta = 1/t$, we have $(G^{\mathrm{T}}_{\beta,\gamma,\zeta,t})^{-1}(t) \sim t$.*

The proof is quite direct but amounts to asymptotic expansions of yet more special functions, which we defer to Appendix C.1 to avoid cluttering. As an elegant interpretation when $\delta$ is small, we may think of the time-uniform empirical Chernoff bound as

$$\mathbb{P}\left(\exists t \geqslant t_0, S_t \gtrsim \frac{\widehat{\sigma}_t(\mu)}{\rho}\sqrt{2\,(t+\alpha)\log\left(\frac{2}{\delta}\sqrt{1+\frac{t}{\alpha}}\right)}\right) \leqslant \delta\,, \tag{4.37}$$

which is clearly reminiscent of the classical sub-Gaussian bound where the sub-Gaussian parameter $R$ is swapped for $\widehat{\sigma}_t(\mu)/\rho$, with $\widehat{\sigma}^2_t(\mu) = 1/t\sum_{s=1}^t (Y_s-\mu)^2$.

**Extension to multivariate distributions.** Although this chapter focuses on one-dimensional estimation, we briefly highlight that the second order sub-Gaussian technique we introduced here easily extends to concentration bounds obtained by vector-valued supermartingales, mirroring Proposition 1.26. Again, the added value compared to classical sub-Gaussian bounds is to eschew the prior knowledge of the variance-related parameter $R$.

**Proposition 4.13** (Multivariate empirical Chernoff time-uniform concentration). *Let $d \in \mathbb{N}$, $\rho \in (0, 1]$. We consider two stochastic processes, $(X_t)_{t\in\mathbb{N}}$ in $\mathbb{R}^d$ and $(\eta_t)_{t\in\mathbb{N}}$ in $\mathbb{R}$, and an adapted filtration $(\mathcal{G}_t)_{t\in\mathbb{N}}$ such that $(\eta_t)_{t\in\mathbb{N}}$ is a conditionally $\rho$-SOSG, i.e. there exists $R \in \mathbb{R}_+^\star$ such that for all $\lambda \in (-\infty, 1/(2R^2))$ and $t \in \mathbb{N}$, we have $\log \mathbb{E}[\exp(\lambda\eta_{t+1}^2) \mid \mathcal{G}_t] \leqslant -1/2 \log(1 - 2R^2((\rho-1)\mathbb{1}_{\lambda<0} + 1)^2\lambda)$. For any $(\alpha, \beta, \gamma, \zeta) \in (\mathbb{R}_+^\star)^4$ such that $\gamma > \beta$, we consider the $\mathbb{R}^d$-valued process $(S_t)_{t\in\mathbb{N}}$, the $\mathbb{R}_+$-valued process $(Q_t)_{t\in\mathbb{N}}$ and the $\mathcal{S}_d^{++}(\mathbb{R})$-valued process $(V_t^\alpha)_{t\in\mathbb{N}}$ defined as*

$$(S_t)_{t\in\mathbb{N}} = \left(\sum_{s=1}^{t-1} \eta_s X_s\right)_{t\in\mathbb{N}}, (Q_t)_{t\in\mathbb{N}} = \left(\sum_{s=1}^{t-1} \eta_s^2\right)_{t\in\mathbb{N}}, \text{ and } (V_t^\alpha)_{t\in\mathbb{N}} = \left(\sum_{s=1}^{t-1} X_s X_s^\top + \alpha I_d\right)_{t\in\mathbb{N}}. \tag{4.38}$$

*Then for any $\delta \in (0, 1)$, if $G^{\mathrm{T}}_{\beta,\gamma,\zeta,t_0}(0) > 2/\delta$, we have the following inequality:*

$$\mathbb{P}\left(\exists t \geqslant t_0,\ \frac{1}{Q_t}\|S_t\|^2_{(V_t^\alpha)^{-1}} \geqslant \frac{2\log\frac{2}{\delta} + \log\frac{\det V_t^\alpha}{\alpha^d}}{\rho^2\left(G^{\mathrm{T}}_{\beta,\gamma,\zeta,t}\right)^{-1}\left(\frac{2}{\delta}\right)}\right). \tag{4.39}$$

*Proof of Proposition 4.13.* This follows from the classical multivariate bound of Proposition 1.26, i.e. $\mathbb{P}(\exists t \in \mathbb{N},\ \|S_t\|^2_{(V_t^\alpha)^{-1}} \geqslant R^2(2\log 1/\delta + \log(\det V_t^\alpha/\alpha^d)) \leqslant \delta$ for some $R \in \mathbb{R}_+^\star$, combined with the high probability time-uniform lower bound on $R$ derived from Theorem 4.10 above (in particular, even though $(\eta_t)_{t\in\mathbb{N}}$ may not be a sum of i.i.d. squared random variables, the assumption that it is conditionally SOSG allows for the same supermartingale construction). ■

## 4.4 Implicit and explicit empirical Chernoff confidence sets

We now instantiate the deviation bounds of the previous section to build confidence sets for the mean of SOSG distributions. First, a straightforward application yields *implicit* sets, which are slightly computationally cumbersome. At the cost of extending the definition of second order sub-Gaussian control, we also present explicit confidence intervals.

### Implicit empirical confidence intervals

The first result is a direct application of the bounds derived in the previous section.

**Corollary 4.14** (Empirical Chernoff implicit confidence sets). *Let $\rho \in (0,1]$ and $(Y_t)_{t\in\mathbb{N}}$ be a sequence of i.i.d. random variables drawn from $\nu$ with expectation $\mu \in \mathbb{R}$. Assume that $\nu \in \mathcal{F}^2_{\mathcal{G},\rho}$. For $t \in \mathbb{N}$, we consider the empirical mean estimator and the empirical variance mapping defined as*

$$\widehat{\mu}_t = \frac{1}{t}\sum_{s=1}^{t} Y_s \quad \textit{and} \quad \widehat{\sigma}_t^2 \colon \mathbb{R} \longrightarrow \mathbb{R}_+$$
$$m \longmapsto \frac{1}{t}\sum_{s=1}^{t}(Y_s - m)^2 \,. \tag{4.40}$$

(*i*) ***Fixed sample confidence set.***

$$\widehat{\Theta}^{\delta}_{t,\rho} = \left\{ m \in \mathbb{R},\ m \in \left[\widehat{\mu}_t \pm \frac{1}{\rho}\sqrt{\frac{2\widehat{\sigma}_t^2(m)\log\frac{3}{\delta}}{-tW_0\left(\frac{-(\delta/3)^{\frac{2}{t}}}{e}\right)}}\right]\right\} \tag{4.41}$$

*is a* ***confidence set*** *at level $\delta$ for $\mu$, i.e. $\mathbb{P}(\mu \in \widehat{\Theta}^{\delta}_{t,\rho}) \geqslant 1-\delta$.*

(*ii*) ***Time-uniform confidence sequence.*** *Let $c = (\alpha,\beta,\gamma,\zeta) \in (\mathbb{R}_+^\star)^4$ such that $\beta < \gamma$ and $t_0 \in \mathbb{N}$ such that $G^{\mathrm{T}}_{\beta,\gamma,\zeta,t_0}(0) > 3/\delta$.*

$$\left(\widehat{\Theta}^{\delta}_{t,\rho,c}\right)_{t\geqslant t_0} = \left(\left\{ m \in \mathbb{R},\ m \in \left[\widehat{\mu}_t \pm \frac{1}{\rho}\sqrt{\frac{2\widehat{\sigma}_t^2(m)(1+\frac{\alpha}{t})\log\left(\frac{3}{\delta}\sqrt{1+\frac{t}{\alpha}}\right)}{\left(G^{\mathrm{T}}_{\beta,\gamma,\zeta,t}\right)^{-1}\left(\frac{2}{\delta}\right)}}\right]\right\}\right)_{t\in\mathbb{N}} \tag{4.42}$$

*is a* ***time-uniform confidence sequence*** *at level $\delta$ for $\mu$, i.e.*

$$\mathbb{P}\left(\forall t \geqslant t_0,\ \mu \in \widehat{\Theta}^{\delta}_{t,\rho,c}\right) \geqslant 1-\delta\,, \tag{4.43}$$

*or equivalently for any adapted random time $\tau$ in $\mathbb{N}$ such that $\tau \geqslant t_0$ almost surely,*

$$\mathbb{P}\left(\mu \in \widehat{\Theta}^{\delta}_{\tau,\rho,c}\right) \geqslant 1-\delta\,. \tag{4.44}$$

*Proof of Corollary 4.14.* This is a straightforward consequence of the self-normalised bounds of Theorems 4.8 and 4.10 respectively. In both cases, we invoke a union argument over three events (the upper and lower first order deviation bounds coming from the sub-Gaussian control and the upper bound on $R$ derived from the second order left tail control), hence the factor $3/\delta$. ■

**Numerical implementation.** The confidence sets of equations (4.41) and (4.42) are implicitly defined as level sets of functions of $m$ (i.e. $m$ is constrained by inequalities that also depend on $m$). An important consideration for practical implementation is whether these are convex subsets of $\mathbb{R}$, i.e. intervals, in which case upper and lower boundaries can be found with the help of a simple root search. We address this issue in the following technical lemma.

**Lemma 4.15** (Convexity of implicit confidence sets). *Let $t \in \mathbb{N}$, $(\xi_s)_{s=1}^t \in \mathbb{R}^t$ a deterministic sequence and $\omega \in (0,1)$. Then the set*

$$\Theta_t = \left\{ m \in \mathbb{R},\ m \in \left[ \frac{1}{t}\sum_{s=1}^t \xi_s \pm \omega \sqrt{\frac{1}{t}\sum_{s=1}^t (\xi_s - m)^2} \right] \right\} \tag{4.45}$$

*is a convex subset of $\mathbb{R}$, i.e. an interval.*

*Proof of Lemma 4.15.* This set can be written as the intersection of the level sets of the two following mappings, where we use the shorthand $\widehat{\xi}_t = 1/t \sum_{s=1}^t \xi_s$:

$$f\colon \mathbb{R} \longrightarrow \mathbb{R} \qquad \text{and} \qquad g\colon \mathbb{R} \longrightarrow \mathbb{R}$$

$$m \longmapsto -m + \widehat{\xi}_t - \omega \sqrt{\frac{1}{t}\sum_{s=1}^t (\xi_s - m)^2} \qquad m \longmapsto m - \widehat{\xi}_t - \omega \sqrt{\frac{1}{t}\sum_{s=1}^t (\xi_s - m)^2}\,, \tag{4.46}$$

i.e. we have $\Theta_t = f^{-1}(\mathbb{R}_-) \cap g^{-1}(\mathbb{R}_-)$. We prove that when $\omega \leqslant 1$, $f$ and $g$ have the monotonicity of their linear terms $-m$ and $+m$ respectively. Indeed, these functions are differentiable and for $m \in \mathbb{R}$ we have $\mathrm{d}f/\mathrm{d}m(m) = -1 + \omega\varepsilon(m)$ and $\mathrm{d}g/\mathrm{d}m(m) = 1 + \omega\varepsilon(m)$ with

$$\varepsilon(m) = \frac{\sum_{s=1}^t \xi_s - m}{\sqrt{t \sum_{s=1}^t (\xi_s - m)^2}}\,. \tag{4.47}$$

We deduce from the Cauchy-Schwarz inequality that $\varepsilon(m) \in [-1, 1]$ for all $m \in \mathbb{R}$, and hence $f$ and $g$ are respectively nonincreasing and nondecreasing if $\omega \leqslant 1$. In particular, $f^{-1}(\mathbb{R}_-)$ and $g^{-1}(\mathbb{R}_-)$ are convex sets (semibounded intervals). Finally, convexity is stable under intersection, hence $\Theta_t$ is convex. ∎

In particular, applying this lemma to the sequence $(Y_s)_{s=1}^t$, we see that the self-normalised fixed sample and time-uniform confidence sets are indeed convex as soon as

$$\frac{1}{\rho}\sqrt{\frac{2\log\frac{3}{\delta}}{-tW_0\left(\frac{-(\delta/3)^{2/t}}{e}\right)}} \leqslant 1 \quad \text{and} \quad \frac{1}{\rho}\sqrt{\frac{2(1+\frac{\alpha}{t})\log\left(\frac{3}{\delta}\sqrt{1+\frac{t}{\alpha}}\right)}{G_{\beta,t}^{-1}(3/\delta)}} \leqslant 1 \tag{4.48}$$

respectively. Note that the denominators of the left-hand sides typically grow with $t$, so these confidence sets are guaranteed to be intervals after collecting enough samples. These two conditions are straightforward to check as they involve only quantities known to the modeller; for instance with $\rho = 1$ and $\delta = 0.05$, the fixed sample set is convex after $t = 23$ samples, and with the standard tuning $(\alpha, \beta, \gamma, \zeta) = (1, 1/2, 3/2, 1/2)$, the time-uniform set is convex after 32 samples. In our opinion, this problem of convexity has often been overlooked in the self-normalised concentration literature, which may have hindered their adoption by statistical practitioners. Interestingly, several recent works in this field have started to study similar considerations, see e.g. Waudby-Smith and Ramdas (2023) for bounded distributions and Wang and Ramdas (2023a) for heavy tailed distributions.

### Explicit empirical confidence intervals

Although the above confidence sets are provably convex, we would ideally prefer to lift the above sample size requirement and directly derive explicit confidence intervals. To circumvent this issue, it is tempting to replace the quadratic term $\widehat{\sigma}_t^2(m)$ by the empirical variance $\widehat{\sigma}_t^2 = \frac{1}{t(t-1)}\sum_{1\leqslant s<s'\leqslant t}(X_s - X_{s'})^2$, which substitutes the mean estimate $m$ with symmetric differences between independent random variables. In order to formalise this, we introduce an alternative, complementary definition of second order sub-Gaussianity.

**Definition 4.16** (Symmetrised second order sub-Gaussian). *Let $\nu$ be a distribution on $\mathbb{R}$ with expectation $\mu$. We define the symmetrised second order cumulant generating function of $\nu$ as*

$$\begin{aligned}\mathring{\mathcal{K}}_2\colon \mathbb{R} &\longrightarrow \mathbb{R} \cup \{+\infty\} \\ \lambda &\longmapsto \log \mathbb{E}_{Y \perp\!\!\!\perp Y'}\left[e^{\frac{\lambda}{2}(Y-Y')^2}\right] ,\end{aligned} \tag{4.49}$$

*where $Y \perp\!\!\!\perp Y' \sim \nu$ denotes two i.i.d. random variables drawn from $\nu$. Let $\rho, R \in \mathbb{R}_+^\star$. We define the family of $(\rho, R)$-**symmetrised second order sub-Gaussian** distributions as*

$$\mathring{\mathcal{F}}^2_{\mathcal{G},\rho,R} = \mathcal{F}_{\mathcal{G},R} \cap \left\{\forall \lambda \in \mathbb{R}_-^\star,\ \mathring{\mathcal{K}}_2(\lambda) \leqslant -\frac{1}{2}\log\left(1 - 2\left(\rho R\right)^2\right)\right\} . \tag{4.50}$$

*Moreover, we define the family of $\rho$-**symmetrised second order sub-Gaussian** (SSOSG) as*

$$\mathcal{F}^2_{\mathcal{G},\rho} = \bigcup_{R \in \mathbb{R}_+^\star} \mathcal{F}^2_{\mathcal{G},\rho,R} . \tag{4.51}$$

Examples of SSOSG distributions can be easily deduced from examples of SOSG distributions. First, if $Y \perp\!\!\!\perp Y'$ are $R$-sub-Gaussian, then $Z = (Y - Y')/2$ is also $R$-sub-Gaussian. The symmetrised second order condition is thus equivalent to requiring the distribution of $Z$ to be $\rho$-SOSG. Examples of such distributions include (i) the Gaussian distributions $\mathcal{N}(\mu, \sigma^2)$ ($Z \sim \mathcal{N}(0, \sigma^2)$) and (ii) the uniform distributions ($Z$ follows a symmetric triangular distribution around zero). Again, this list is not meant to be exhaustive but rather suggests that the SSOSG offers a natural nonparametric alternative to other classical statistical models.

**Technical detour: U-statistics.** In order to study the deviations of $\widehat{\sigma}_t^2$, as we did earlier for the quadratic sum $Q_t$, it is natural to introduce the random variable $\exp(\lambda \widehat{\sigma}_t^2)$ for $\lambda \in \mathbb{R}_-^\star$. However, the definition above does not immediately provide a control on this quantity because $\widehat{\sigma}_t^2$ is *not* a sum of i.i.d. random variables (each of the $Y_1, \dots, Y_t$ appears multiple times in the sum that defines $\widehat{\sigma}_t^2$. It is however possible to disentangle these multiple dependencies and decompose the empirical variance into a sum of **U-statistics** (Hoeffding, 1992), which are themselves sums of independent random variables. Indeed, for $t \in \mathbb{N}$ and $\pi \in \mathfrak{S}_t$ a permutation of $\{1, \dots, t\}$, we define the U-statistics as

$$V_t^\pi = \frac{1}{\lfloor t/2 \rfloor} \sum_{s=1}^{\lfloor t/2 \rfloor} \left(Z_s^\pi\right)^2 \text{ , where for } s \in \{1, \dots, \lfloor t/2 \rfloor\},\ Z_s^\pi = \frac{Y_{\pi(2s)} - Y_{\pi(2s-1)}}{\sqrt{2}} . \tag{4.52}$$

By construction, the sequence $(Z_s^\pi)_{s=1,\dots,\lfloor t/2 \rfloor}$ is i.i.d., and the $V_t^\pi$ are identically distributed for $\pi \in \mathfrak{S}_t$. In particular, if we denote by $I$ the identity permutation, $V_t^I = \frac{1}{\lfloor t/2 \rfloor} \sum_{s=1}^{\lfloor t/2 \rfloor} (X_{2s} - X_{2s-1})^2/2$ defines an adapted process to the natural filtration defined by $\mathcal{G}_t = \sigma(Y_s, s \leqslant t)$. Moreover, the empirical variance can be recovered by averaging over all U-statistics, i.e. we have

$$\widehat{\sigma}_t^2 = \frac{1}{t!} \sum_{\pi \in \mathfrak{S}_t} V_t^\pi \,. \tag{4.53}$$

This allows to derive the following explicit confidence sets, mirroring those of Corollary 4.14.

**Corollary 4.17** (Empirical Chernoff explicit confidence sets). *Let $\rho \in (0,1]$ and $(Y_t)_{t\in\mathbb{N}}$ be a sequence of i.i.d. random variables drawn from $\nu$ with expectation $\mu \in \mathbb{R}$. Assume that $\nu \in \mathring{\mathcal{F}}^2_{\mathcal{G},\rho}$.*

(*i*) ***Fixed sample confidence set.***

$$\widehat{\Theta}^\delta_{t,\rho} = \left[ \widehat{\mu}_t \pm \frac{\widehat{\sigma}_t}{\rho} \sqrt{\frac{2\log\frac{3}{\delta}}{-tW_0\left(\frac{-(\delta/3)^{\frac{2}{\lfloor t/2 \rfloor}}}{e}\right)}} \right] \tag{4.54}$$

*is a **confidence set** at level $\delta$ for $\mu$, i.e. $\mathbb{P}(\mu \in \widehat{\Theta}^\delta_{t,\rho}) \geqslant 1-\delta$.*

(*ii*) ***Time-uniform confidence sequence.*** *Let $c = (\alpha,\beta,\gamma,\zeta) \in (\mathbb{R}_+^\star)^4$ such that $\beta < \gamma$ and $t_0 \in \mathbb{N}$ such that $G^{\mathrm{T}}_{\beta,\gamma,\zeta,\lfloor t_0/2 \rfloor}(0) > 3/\delta$.*

$$\left(\widehat{\Theta}^\delta_{t,\rho,c}\right)_{t \geqslant t_0} = \left( \left[ \widehat{\mu}_t \pm \frac{1}{\rho} \sqrt{\frac{2\lfloor t/2 \rfloor V^I_{\lfloor t/2 \rfloor}(1+\frac{\alpha}{t}) \log\left(\frac{3}{\delta}\sqrt{1+\frac{t}{\alpha}}\right)}{t\left(G^{\mathrm{T}}_{\beta,\gamma,\zeta,\lfloor t/2 \rfloor}\right)^{-1}\left(\frac{3}{\delta}\right)}} \right] \right)_{t\in\mathbb{N}} \tag{4.55}$$

*is a **time-uniform confidence sequence** at level $\delta$ for $\mu$, i.e.*

$$\mathbb{P}\left(\forall t \geqslant t_0,\ \mu \in \widehat{\Theta}^\delta_{t,\rho,c}\right) \geqslant 1-\delta\,, \tag{4.56}$$

*or equivalently for any adapted random time $\tau$ in $\mathbb{N}$ such that $\tau \geqslant t_0$ almost surely,*

$$\mathbb{P}\left(\mu \in \widehat{\Theta}^\delta_{\tau,\rho,c}\right) \geqslant 1-\delta\,. \tag{4.57}$$

*Proof of Corollary 4.17.* The general strategy here is to derive high probability upper bounds for $R$ in the new definition of SSOSG distributions (the first order sub-Gaussian bounds are unchanged). From there, a simple union argument as in the proof of Corollary 4.14 yields the expression of the confidence sets. We now prove these two deviation bounds.

**Fixed sample.** Let $u \in \mathbb{R}_+^\star$ and $\lambda \in \mathbb{R}_-^\star$. The function $z \in \mathbb{R} \mapsto \exp(\lambda z)$ is nonincreasing. Combining this with Markov's inequality yields

$$\mathbb{P}\left(\widehat{\sigma}_t^2 \leqslant u\right) = \mathbb{P}\left(e^{\lambda \widehat{\sigma}_t^2} \geqslant e^{\lambda u}\right) \leqslant e^{-\lambda u}\mathbb{E}\left[e^{\lambda \widehat{\sigma}_t^2}\right]. \tag{4.58}$$

Moreover, this function is also convex and satisfies the inequality $\mathbb{1}_{z>0} \leqslant \exp(\lambda z)$ for all $z \in \mathbb{R}$, so by Jensen's inequality we have

$$\begin{aligned}\mathbb{P}\left(\widehat{\sigma}_t^2 \leqslant u\right) = \mathbb{E}\left[\mathbb{1}_{\widehat{\sigma}_t^2 - u > 0}\right] \leqslant \mathbb{E}\left[e^{\lambda(\widehat{\sigma}_t^2 - u)}\right] \leqslant e^{-\lambda u}\mathbb{E}\left[\frac{1}{t!}\sum_{\pi \in \mathfrak{S}_t} e^{\lambda V_t^\pi}\right] &= \frac{e^{-\lambda u}}{t!}\sum_{\pi \in \mathfrak{S}_t}\mathbb{E}\left[e^{\lambda V_t^\pi}\right] \\ &= e^{-\lambda u}\mathbb{E}\left[e^{\lambda V_t^I}\right],\end{aligned} \tag{4.59}$$

where the last line comes from the fact that all $V_t^\pi$ are identically distributed. Now we may simply apply the Cramér-Chernoff method on the right-hand side by noting that $\lfloor t/2 \rfloor V_t^I$ is a sum of $\lfloor t/2 \rfloor$ i.i.d. random variables $(Z_s^I)_{s=1}^{\lfloor t/2 \rfloor}$ satisfying the second order CGF control $\log \mathbb{E}[\exp(\lambda Z_s^I)] \leqslant -1/2 \log(1 - 2(\rho R)^2 \lambda)$ thanks to the SSOSG condition of Definition 4.16. Therefore, we obtain the inequality

$$\mathbb{P}\left(\widehat{\sigma}_t^2 \leqslant -W_0\left(\frac{-\delta^{\frac{2}{\lfloor t/2 \rfloor}}}{e}\right)(\rho R)^2\right) \leqslant \delta. \tag{4.60}$$

**Time-uniform.** We consider the adapted process defined as

$$\left(M_t^\lambda\right) = \left((1-2\lambda)^{\frac{\lfloor t/2 \rfloor}{2}} e^{\lambda \frac{\lfloor t/2 \rfloor V_t^I}{(\rho R)^2}}\right)_{t \in \mathbb{N}}. \tag{4.61}$$

By Definition 4.16, this is a nonnegative supermartingale. By using the same mixture construction as in the proof of Theorem 4.10, $if G^{\mathrm{T}}_{\beta,\gamma,\zeta,\lfloor t_0/2 \rfloor}(0) > 1/\delta$, we obtain

$$\mathbb{P}\left(\exists t \geqslant t_0,\ \lfloor t/2 \rfloor V_t^I \leqslant (\rho R)^2 \left(G^{\mathrm{T}}_{\beta,\gamma,\zeta,,\lfloor t/2 \rfloor}\right)^{-1}\left(\frac{1}{\delta}\right)\right) \leqslant \delta. \tag{4.62}$$

■

**Remark 4.18**. *The supermartingale construction behind the time-uniform bound prevents the use of the process $(\widehat{\sigma}_t^2)_{t\in\mathbb{N}}$ as it is* not *adapted to the natural filtration, which is why we used the (less efficient) estimator $(V_t^I)_{t\in\mathbb{N}}$ instead. Note that it is also possible to fix any permutation $\pi$ of $\mathbb{N}$ and consider $(V_t^\pi)_{t\in\mathbb{N}}$; however, this introduces a dependency on the ordering of the observations $(Y_t)_{t\in\mathbb{N}}$, a phenomenon also observed in Waudby-Smith and Ramdas (2023) (for different reasons).*

We conclude the theoretical part of this chapter by mentioning a direct consequence for multiarmed bandits, following the generic UCB analysis developed in Chapter 1.

**Corollary 4.19** (Empirical Chernoff UCB). *Let $\rho \in (0,1]$, $K \in \mathbb{N}$, $(\rho_k)_{k\in[K]} \in (0,1]^K$ and a bandit model $(\bigotimes_{k\in[K]} \nu_k, \bigotimes_{k\in[K]} \mathcal{F}^2_{\mathcal{G},\rho_k})$ or $(\bigotimes_{k\in[K]} \nu_k, \bigotimes_{k\in[K]} \mathring{\mathcal{F}}^2_{\mathcal{G},\rho_k})$. Let $c = (\alpha,\beta,\gamma,\zeta) \in (\mathbb{R}_+^\star)^4$ such that $\beta < \gamma$, and for any $\delta \in (0,1)$, we consider the $K$ time-uniform empirical Chernoff confidence sequences $(\widehat{\Theta}^\delta_{k,n,\rho_k,c})_{n\in\mathbb{N}}$ built from either Corollary 4.14 or Corollary 4.17 and i.i.d. observations from $\nu_k$. Then for $T \to +\infty$, Algorithm 1 with the expectation risk measure and UCB given by $(\widehat{\Theta}^\delta_{k,n,\rho_k,c})_{n\in\mathbb{N}}$ satisfies, for all suboptimal arm $k \in [K] \setminus \{k^\star\}$,*

$$\mathbb{E}_{\pi,\boldsymbol{\nu}_{pi}^{\otimes T}}\left[N_T^k\right] \leqslant \frac{2\sigma_k^2}{\rho_k^2\Delta_k^2}\log\left(\frac{4\sigma_k T}{\rho_k\Delta_k}\right) + o\left(\frac{\sigma_k^2}{\rho_k^2\Delta_k^2}\log T\right), \tag{4.63}$$

*where $\sigma_k^2$ is the variance of $\nu_k$ and $\Delta_k$ its suboptimality gap.*

The proof is essentially that of the $R$-sub-Gaussian UCB of Corollary 1.34 coupled with the asymptotic expansion of $G^{\mathrm{T}}_{\beta,\gamma,\zeta,t}$ developed in Lemma 4.12; we postpone the details to Appendix C.1. This shows that logarithmic pseudo regret is achievable for the (symmetrised) second order sub-Gaussian distributions. Note that the variances $(\sigma_k^2)_{k\in[K]\setminus\{k^\star\}}$ appear in the pseudo regret bound but *not* in algorithm itself. In particular, this provides a logarithmic bandit policy for the class of all Gaussian distributions (without knowing their variances) and all uniform distributions (without knowing their supports) simultaneously.

## 4.5 Numerical experiments

We now report experiments to support the theoretical insights of the previous sections and compare the empirical Chernoff confidence sets with standard benchmarks. All the confidence sets presented here are implemented in the open source *concentration-lib* Python package. [1]

**Sensitivity of the second order lower bound on $R$.** We start by investigating the influence of the Tricomi mixing parameters $\beta, \gamma, \zeta$ in the second order bound of Theorem 4.10. In Figure 4.2a, we report confidence envelopes for the parameter $R$, i.e. running intersections of the intervals

$$\left[\sqrt{\frac{Q_t}{\left(G^{\mathrm{K}}_{\beta,\gamma,\zeta,t}\right)^{-1}\left(\frac{2}{\delta}\right)}}, \sqrt{\frac{Q_t}{\left(G^{\mathrm{T}}_{\beta,\gamma,\zeta,t}\right)^{-1}\left(\frac{2}{\delta}\right)}}\right] \quad \text{with } Q_t = \sum_{s=1}^{t}(Y_s-\mu)^2 \quad \text{and } t \in \mathbb{N}\,. \tag{4.64}$$

We consider samples from the standard Gaussian distribution $\mathcal{N}(\mu, R)$ with $\mu = 0$ and $R = 1$ (results are similar for other distributions). We focus on the upper bound (i.e. the Tricomi rather than the Kummer mixtures) as it is the one of interest for empirical Chernoff confidence sequences. Results are displayed in Figures 4.2b, 4.2c and 4.2d. We conclude that these parameters have a moderate influence on the actual width of the confidence envelopes, and that the suggested tuning $\beta = 1/2$, $\gamma = 3/2$ and $\zeta = \alpha/2$ is a reasonable default choice.

[1] https://pypi.org/project/concentration-lib.

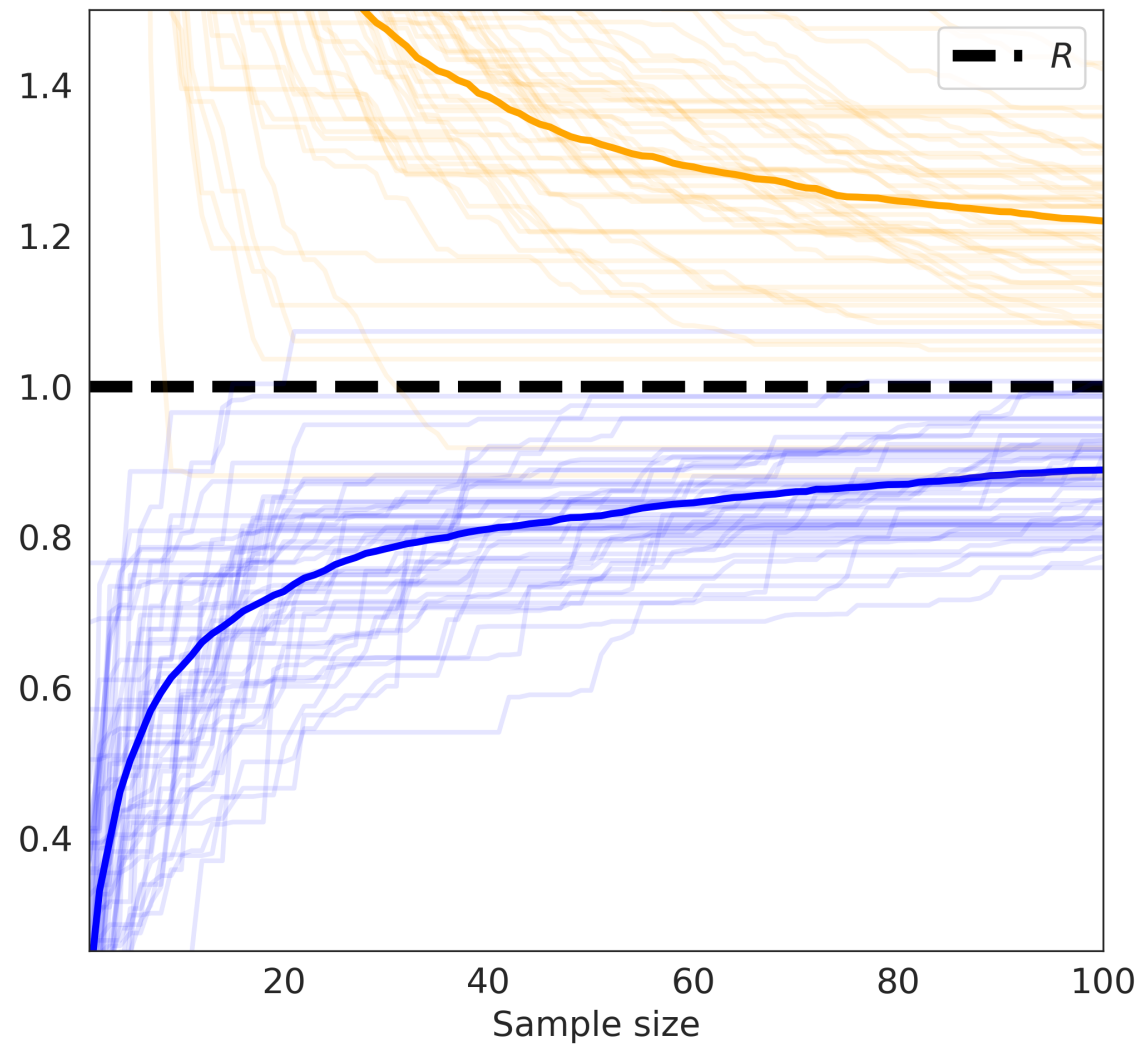

**(a)** $\beta = 1/2, \gamma = 3/2, \zeta = 1/2$.

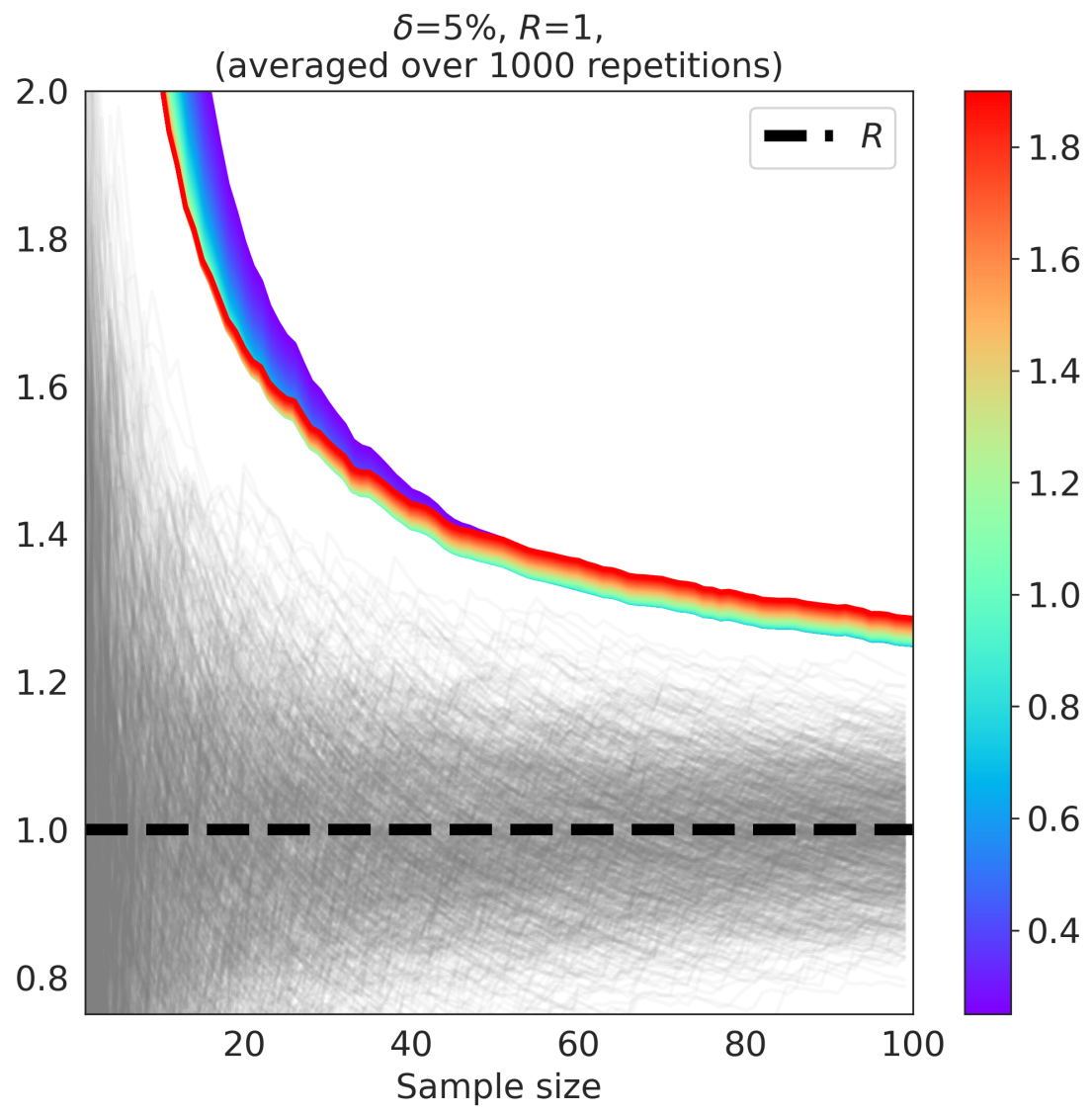

**(b)** Varying $\beta$, $\gamma = \beta + 1, \zeta = 1/2$.

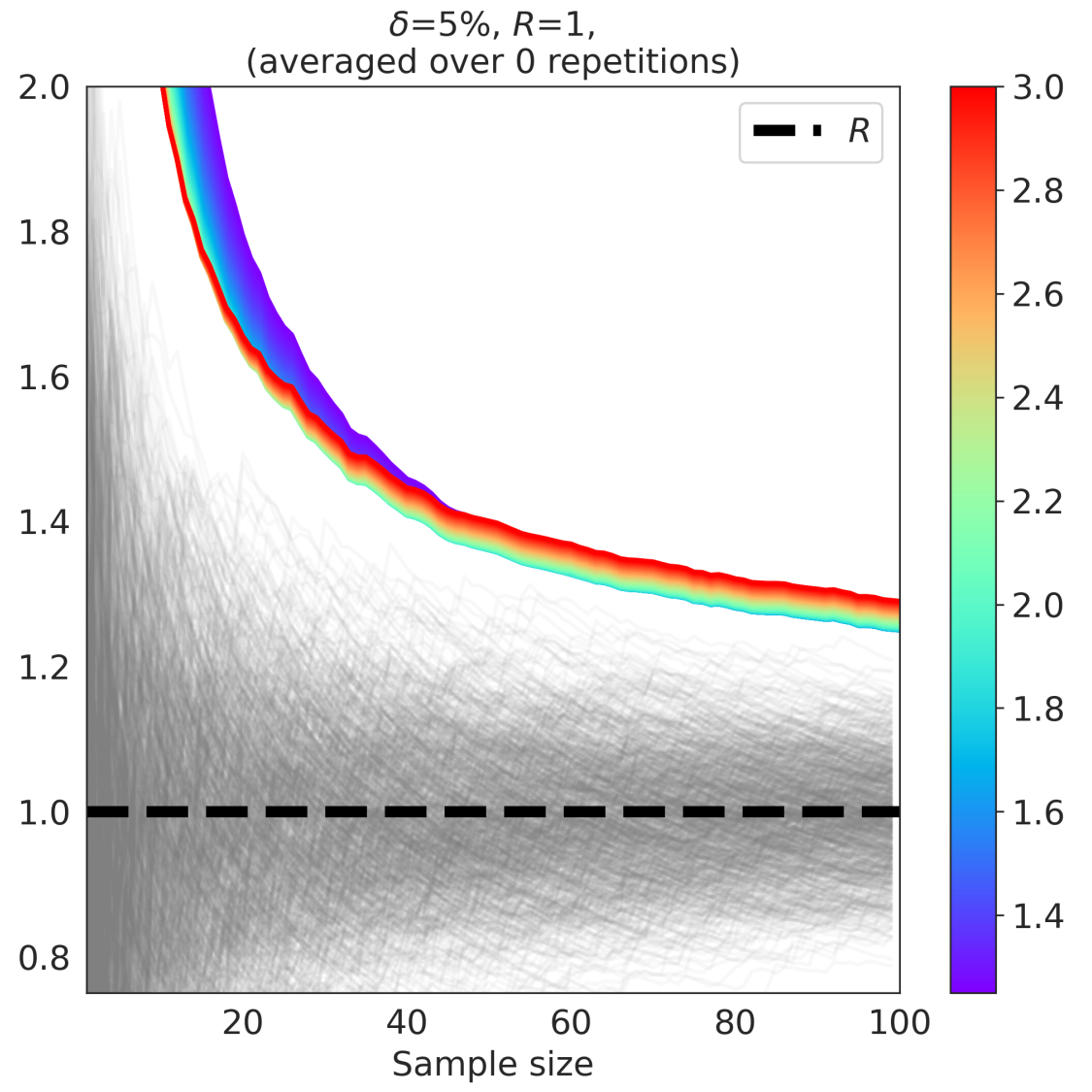

**(c)** $\beta = \gamma - 1$, varying $\gamma$, $\zeta = 1/2$.

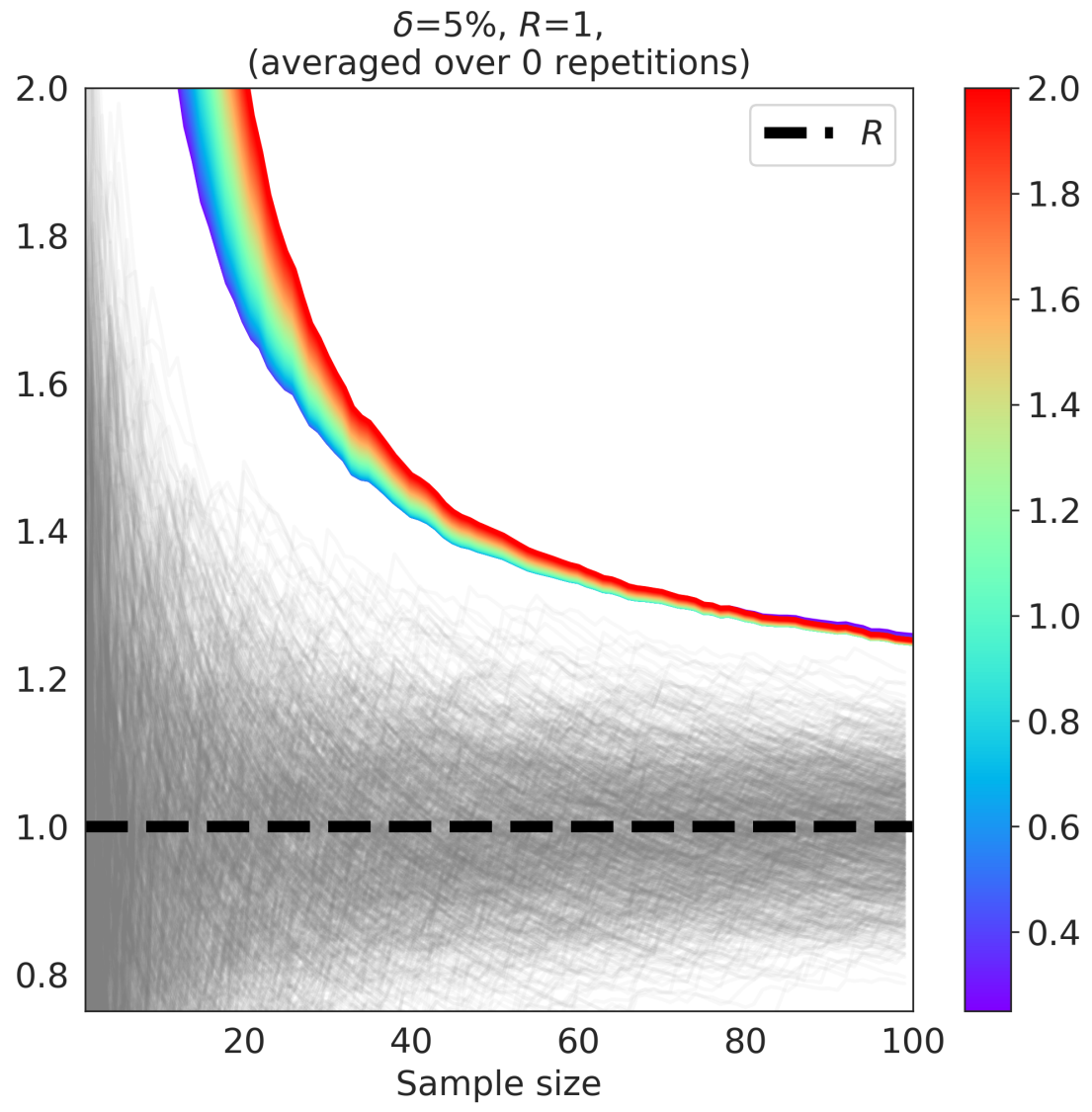

**(d)** $\beta = 1/2, \gamma = 3/2$, varying $\zeta$.

**Figure 4.2 –** Top left: examples of time-uniform confidence envelopes for $R$ on several realisations of $\mathcal{N}(0, R)$ with $R = 1$, $\rho = 1$ as a function of the number of observations $t$. Thick lines indicate median curves over 1000 replicates. Other three figures: tuning of the three Tricomi mixing parameters $\beta < \gamma$ and $\zeta$, also over 1000 independent replicates. Grey lines are trajectories of empirical second moments $\sqrt{Q_t/t}$, drawn for $\mathcal{N}(0, R)$ with $R = 1$, $\rho = 1$. Thick black dashed line: $R$.

**Estimation of the asymptotic behaviour of $(G^{\mathrm{T}}_{\beta,\gamma,\zeta,t})^{-1}$.** To further support the asymptotic analysis of Lemma 4.12, that showed that $(G^{\mathrm{T}}_{\beta,\gamma,\zeta,t})^{-1}(1/\delta) \sim t$ when $t \to +\infty$, we compute this inverse mapping (using a simple Brent root search) for increasing values of $t$ and report the results in Figure 4.3. First, for both ranges $t \in [1, 200]$ and $t \in [1, 10^5]$, we see that a simple straight line provides a good fit to the curve $t \mapsto (G^{\mathrm{T}}_{\beta,\gamma,\zeta,t})^{-1}(t)$. Furthermore, we notice that the fitted slope increases when we consider a longer range, and seemingly converges to 1. We interpret this as the cost of small sample concentration: if $(G^{\mathrm{T}}_{\beta,\gamma,\zeta,t})^{-1}(1/\delta) \approx c_t t$ with $c_t \in (0, 1]$, the empirical Chernoff time-uniform confidence sequence has the approximate width

$$\frac{2\widehat{\sigma}_t(\mu)}{\rho}\sqrt{\frac{2}{c_t t}\left(1+\frac{\alpha}{t}\right)\log\left(\frac{3}{\delta}\sqrt{1+\frac{t}{\alpha}}\right)}\,. \tag{4.65}$$

For large $t$, $c_t \approx 1$ so we essentially recover the $R$-sub-Gaussian width with $R \approx \widehat{\sigma}_t(\mu)/\rho$. However, for smaller sample sizes, the term $1/\sqrt{c_t} > 1$ inflates the confidence set, representing the cost of learning an estimator for $R$ jointly with the confidence sequence for $\mu$.

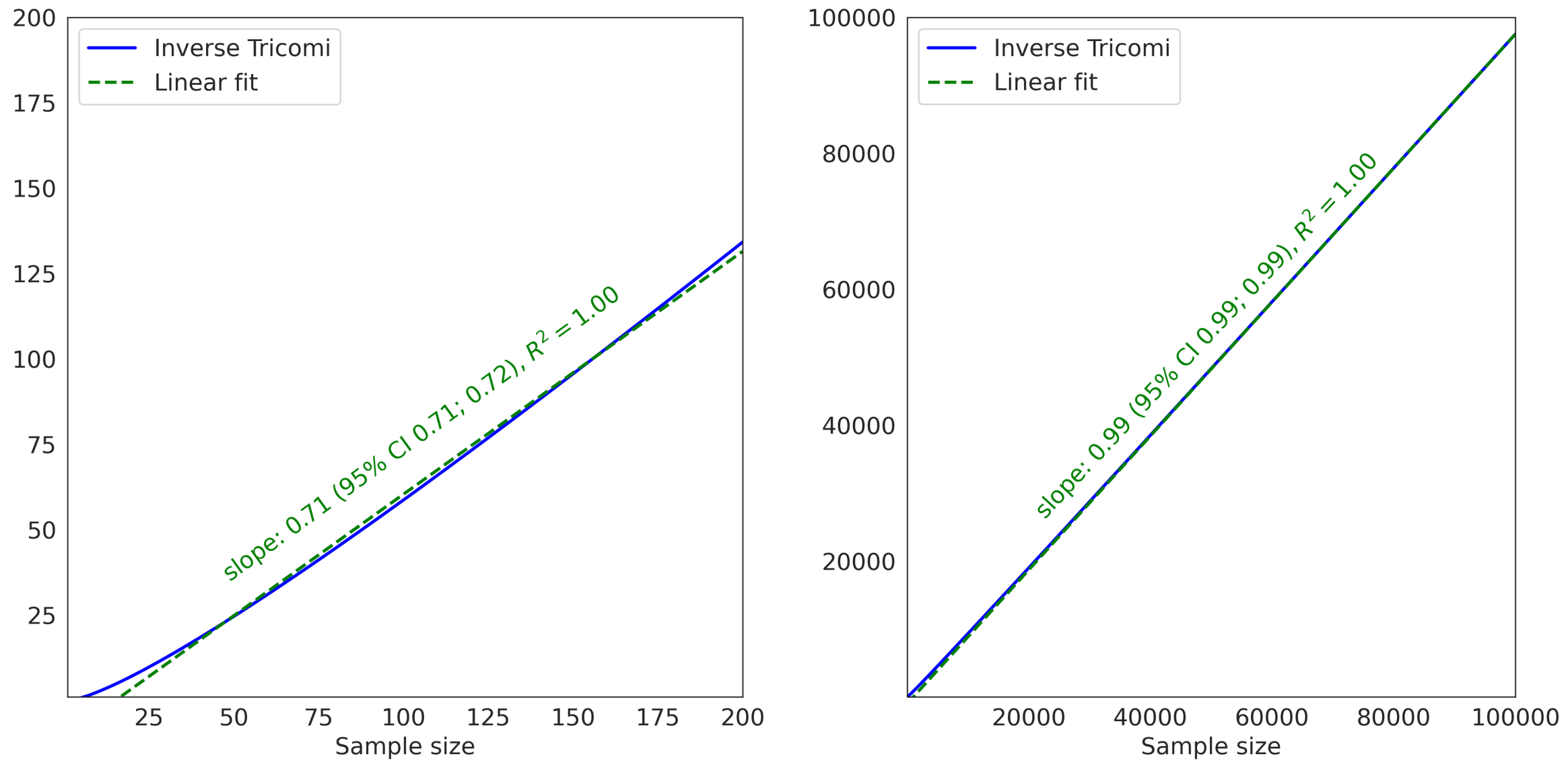

**Figure 4.3** – Inverse Tricomi mapping $t \mapsto (G^{\mathrm{T}}_{\beta,\gamma,\zeta,t})^{-1}(1/\delta)$ for $\delta = 1/t$, $\beta = 1/2$, $\gamma = 3/2$ and $\zeta = 1/2$ (blue) and linear fit (green). Left: $t \in [1, 200]$. Right: $t \in [1, 10^5]$. The linear fit is performed on the last $30\%$ of the range for $t$ to account for the asymptotic effect.

**Fixed sample variance estimation on a real-world small sample.** We now show the merits of the SOSG condition for practical modelling. We consider a 23-year maize field experiment on the research farm of McGill University in Sainte-Anne-de-Bellevue, Québec, Canada (Joshi et al., 2017).[2] Every year of the experiment, after harvest, the actual crop yield (i.e. the weight of harvested dry grain per hectare) was measured and compared to the yield predicted by a physical model taking as inputs the environment (precipitation, average temperatures throughout the year, etc.) and the crop management policy (fertilisation, planting dates, etc.). Leaving aside two years with missing values, this resulted in a set of 21 data points $\{e_1, \ldots, e_{21}\}$, which we interpret as realisations of a random variable $E$. We call these measurements *calibration errors* as they are used downstream to calibrate a stochastic crop yield simulator (DSSAT, which we also mention below and in Chapter 6). Because the physical model builds on years of agricultural engineering expertise and in vivo validation, we assume that $\mathbb{E}[E] = 0$, i.e. the prediction model is unbiased. However, given the limited sample size, this calibration error introduces an additional *incompressible* uncertainty in the downstream simulation, in the sense that collecting more data to improve calibration is not feasible (the modeller would have to wait for another harvest). The goal here is to estimate a fixed sample, nonasymptotic confidence interval for the standard deviation $\sigma = \sqrt{\mathbb{V}[E]}$.

[2]We thank Chandra A. Madramootoe, Nitin Joshi, and especially Romain Gautron for their help collecting and interpreting this dataset.

Common assumptions in the literature available to the modeller are (i) $E$ is Gaussian, in which case a confidence interval for $\sigma$ follows from the quantiles of the Chi-square distribution, and (ii) $E$ is bounded. For the latter, expert knowledge suggests that the *yield potential* (i.e. the maximum attainable yield) for this farm is 20,000 kg/ha, i.e. $E \in [0, 2 \times 10^4]$. In this case, confidence intervals may be deduced from either Maurer-Pontil (Maurer and Pontil, 2009) or Bentkus (Kuchibhotla and Zheng, 2021) variance bound (note that these are the heart of respectively empirical Bernstein and Bentkus bounds, as recalled in Table 2.2). Alternatively, we may use the second order bound of Theorem 4.8 to obtain a confidence interval for $\sigma$ under the assumption that $E$ is $(\rho, R)$-SOSG and strictly sub-Gaussian (i.e. $R = \sigma$). As seen in Proposition 4.5, this model specification encompasses Gaussian and many standard bounded distributions, and is thus a compromise between the two previous modelling assumptions.

We report the results in Figure 4.4 with $\rho = 1$ for the Empirical Chernoff bound. Both Maurer-Pontil and Bentkus bounds provide very loose estimations (with the former yielding only a trivial negative lower bound even after 21 years). Of all bounds, the Chi-square one is the tightest but also the one that relies on the most stringent assumptions. By contrast, the empirical Chernoff method appears as a relaxation of the parametric Gaussian hypothesis without significantly widening the resulting confidence intervals. We hope this example may inspire practitioners to consider SOSG as an alternative to standard modelling assumptions.

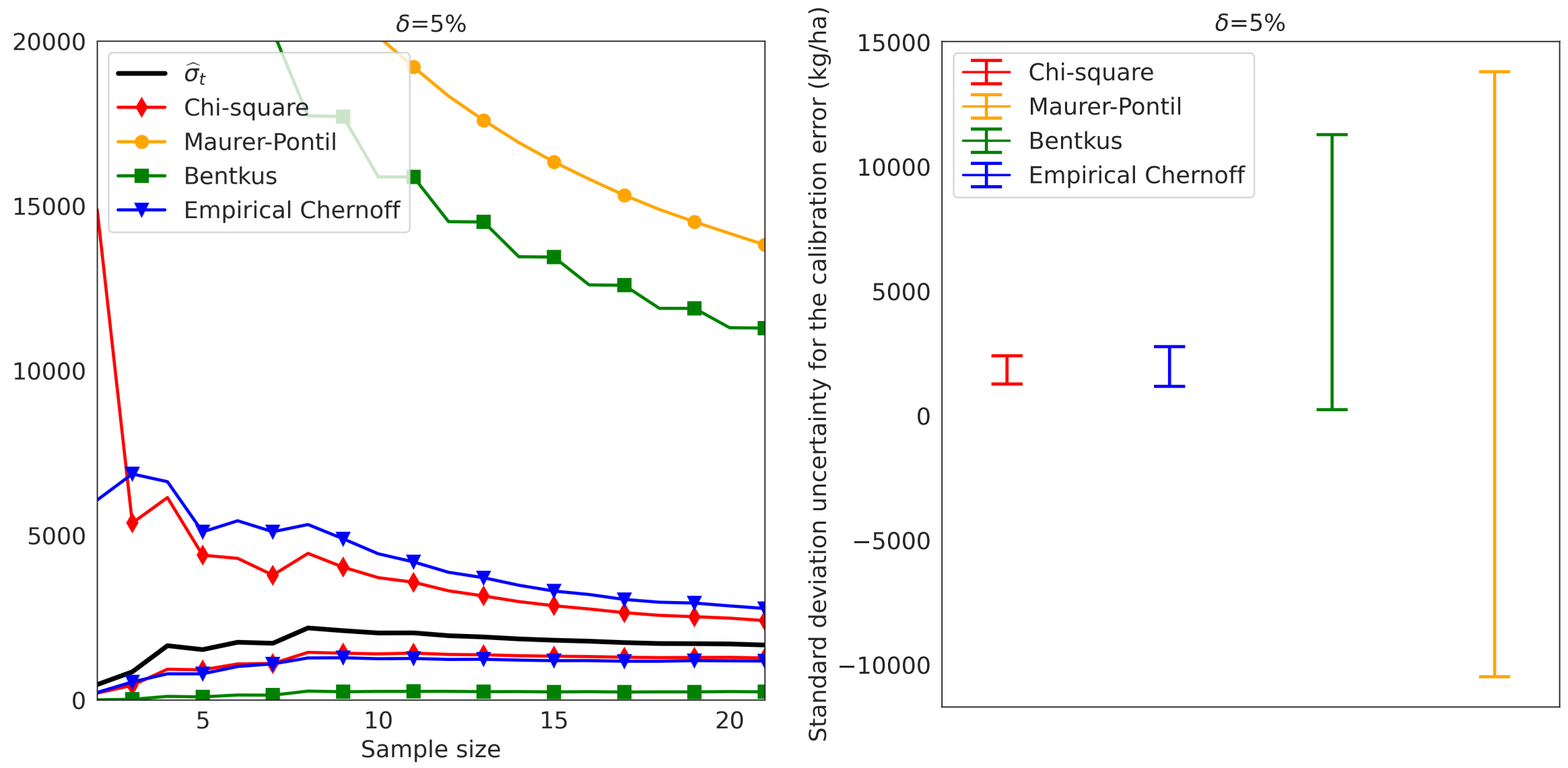

**Figure 4.4** – Estimation of the standard deviation $\sigma$ from 23 yearly measurements of crop yield prediction errors in the maize field experiment in Sainte-Anne-de-Bellevue, Québec, Canada. Left: confidence bounds as a function the sample size (number of years). Right: confidence bars after 21 years, ranked by increasing width.

**Comparison of time-uniform confidence sequences.** Finally, we report the implicit and explicit confidence sequences of Corollaries 4.14 and 4.17 on example distributions and compare them with state-of-the-art benchmarks (note that we consider time-uniform bounds here, as opposed to the fixed sample introductory example of Figure 4.1).

In Figure 4.5a, we compare our bounds with the sub-Gaussian method of mixtures on $\mathcal{N}(0,1)$. With $R = 1$, which corresponds to having prior knowledge of the variance, the classical mixture bound is the tightest (note that in the Gaussian specifically, the supermartingale construction of the method of mixtures is actually a martingale), although the gap with the empirical Chernoff envelopes is quickly vanishing. However, assuming an imperfect, conservative prior knowledge, materialised here by $R = 2$, the sub-Gaussian bound is significantly outperformed by our bounds, showcasing their ability to adapt to the actual variance. Similarly, they are able to adapt to the support and variance of uniform distributions, although the benefit is less clear here as we compare ourselves to hedged capital, which yields very sharp bounds in practice (and is also somewhat adaptive to the true variance thanks to the definition of its predictable weights, see Proposition A.5).

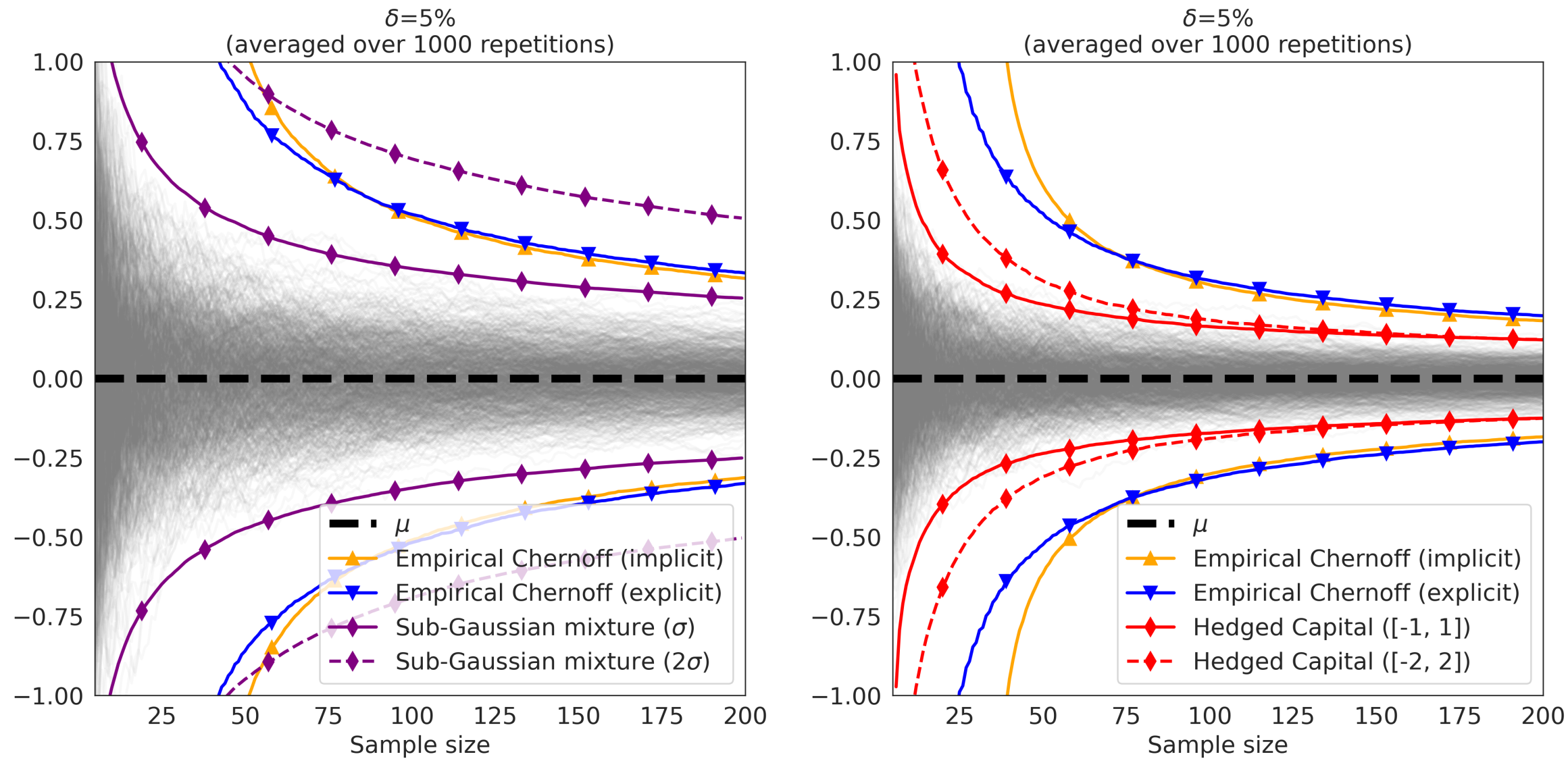

(**a**) Gaussian. (**b**) Uniform.

**Figure 4.5** – Comparison of median confidence envelopes around the mean for $\mathcal{N}(0,1)$ $(\rho = 1)$ and $\mathcal{U}([-1,1])$ $(\rho = 1$ and $\rho = \sqrt{3/\pi}$ for the implicit and explicit bounds respectively), as a function of the sample size $t$, over $1000$ independent replicates. Grey lines are trajectories of empirical means $\widehat{\mu}_t$.

Building on this initial experiment, we test our confidence bound on a crop yield distribution simulated by DSSAT (Hoogenboom et al., 2019), represented in Figure 4.6a. Samples are generated by specifying a crop management policy (e.g. a planting date) and drawing environmental conditions (e.g. weather) from a stochastic model. We estimate the true mean crop yield from $10^6$ samples, resulting in $\mu = 4217\text{kg}/\text{ha}$. We defer to Section 6.4 in Chapter 6 a more in-depth description of the DSSAT experimental setting. Note that this particular instance of the simulator has been calibrated using the field experiment described above for the variance estimation problem. Since the maize crop yields are bounded (in this case by a yield potential of 20,000 kg/ha), we compare our bounds to hedged capital, see Figure 4.6b.

Contrary to the uniform case, we notice that the yield distribution is localised on a narrow region of the support. The empirical Chernoff approach is able to discover and exploit this information to produce sharp confidence sequences, while the hedged capital ones is hindered by the sole prior knowledge of the support.

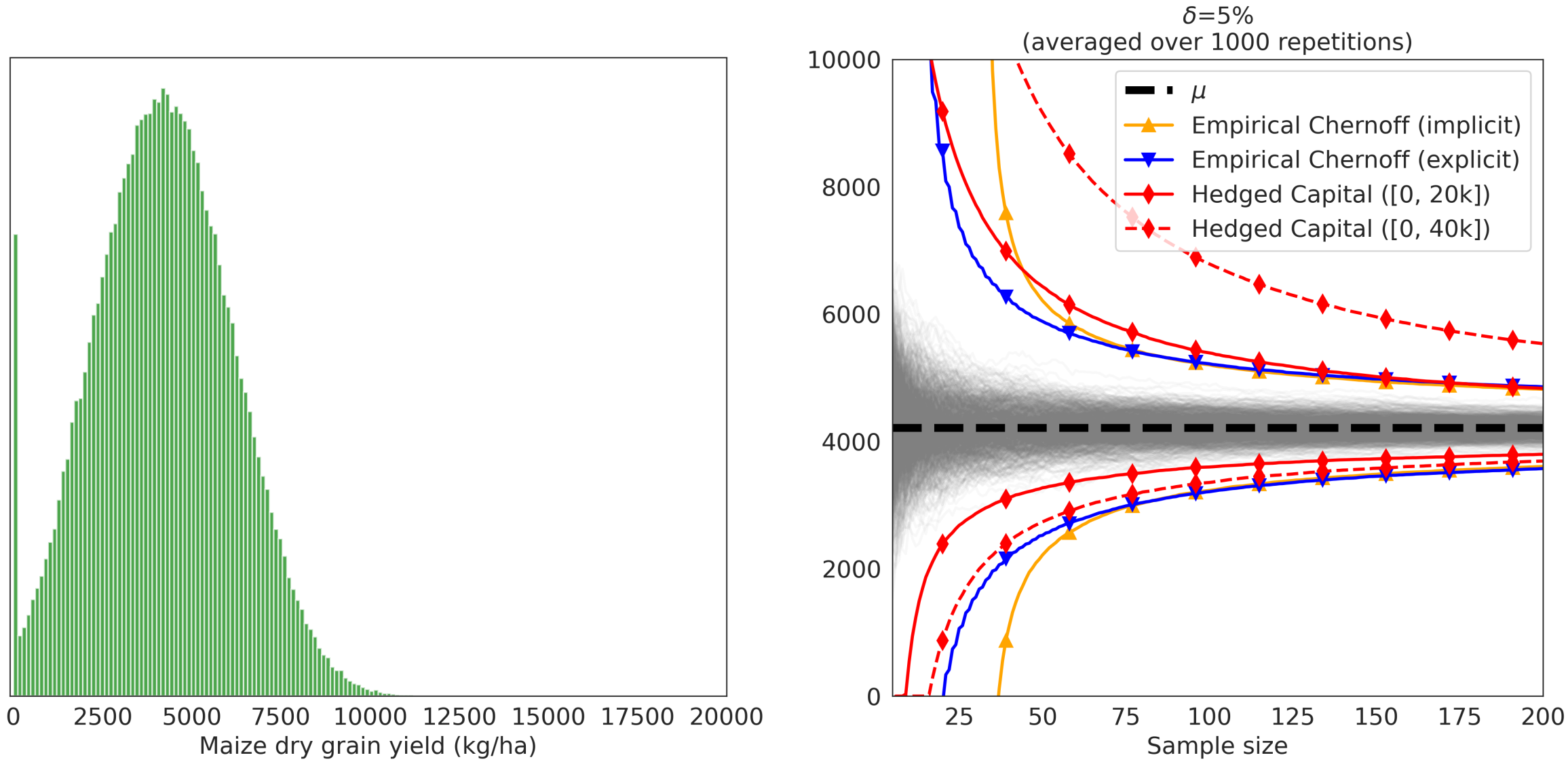

(**a**) DSSAT-simulated maize yield distribution.

(**b**) Time-uniform confidence sequences.

**Figure 4.6** – Comparison of median confidence envelopes around the mean for a DSSAT-simulated distribution, as a function of the sample size $t$, over $1000$ independent replicates. We used $\rho = 1$ for the empirical Chernoff bounds. Grey lines are trajectories of empirical means $\widehat{\mu}_t$.

# Conclusion

We have introduced a refinement of the sub-Gaussian condition that substitutes the knowledge of the sub-Gaussian parameter with a data-dependent estimator, at the cost of an alternative parameter $\rho$. We believe an interesting direction for future research would be to better understand the role of $\rho$ and have a finer characterisation of second order sub-Gaussian distributions.

**Take-home message:**

**☞ A vast range of both parametric and non-parametric model specifications are available to study anytime-valid statistics.**

We have extended the existing literature on concentration bounds to cover:

- **generic exponential families** (Gaussian with unknown variance, Chi-square, Poisson, Pareto, etc.);

- **second order sub-Gaussian** distributions (nonparametric specification, variance-adaptive without prior knowledge of a bounded support).

These confidence sequences are sharp in practice even for in the regime of small data and are available either in closed form or via a simple root search.

Beyond their theoretical interest, these bounds were designed to appeal to practitioners, especially for critical applications concerned with safety with respect to sampling (to avoid false discovery or p-hacking) and constrained by small samples.

As a side result, they also provide deterministic UCB algorithms with logarithmic pseudo regret guarantees for new settings.

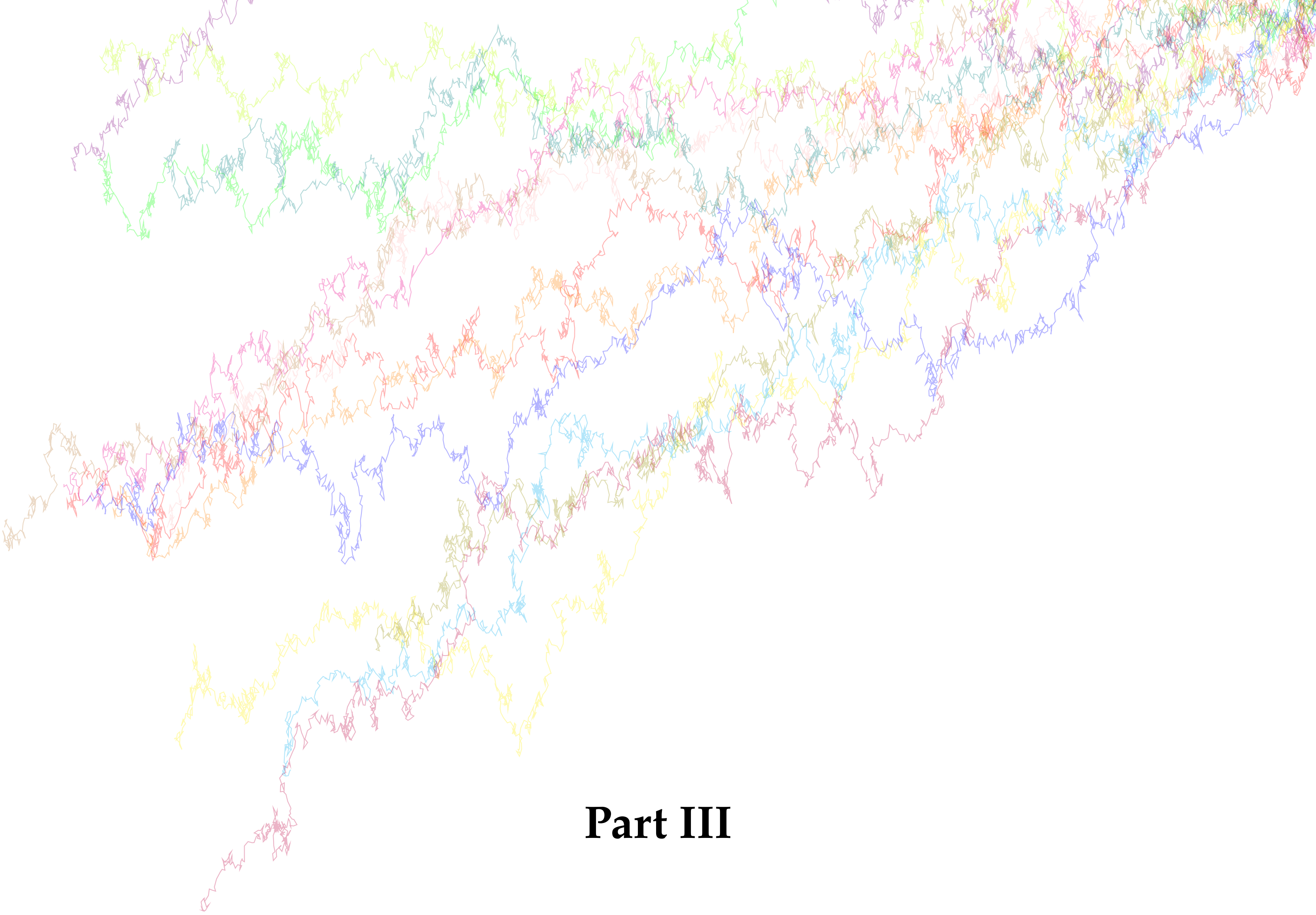

# Part III

# Risk awareness in contextual bandits

*or how to play safe*

# Chapter 5

# Risk-aware linear bandits with elicitable risk measures ( )

*I take risks, sometimes patients die. But not taking risks causes more patients to die.*
— "Detox", *House, M.D.*

The content of this chapter was presented at the 2022 *European Workshop on Reinforcement Learning* (EWRL) (Saux and Maillard, 2022) and published at the 2023 *International Conference on Artificial Intelligence and Statistics* (AISTATS) (Saux and Maillard, 2023). Compared to that article, we here report in the main text (i) the proofs the multivariate method of mixtures for empirical risk minimisation (Proposition 5.12), generalising the standard bound of Proposition 1.26, and the regret analysis of LinUCB-CR (Theorem 5.18), and (ii) a new minimax lower bound for expectiles (within at most polylogarithmic terms of our upper bound for LinUCB-CR). Alternative regret analyses (stochastic action sets, online gradient descent) and an overview of common risk measures are deferred to Appendix D.

## Contents

## Outline and contributions

We recall that contextual bandits are sequential decision-making models where at each time step an agent observes a set of possible actions, or contexts, picks one and receives a stochastic reward, the distribution of which is a function of the selected action. The goal of the agent is to learn a reward-maximising policy, facing the classical exploitation-exploration dilemma. A prominent example of such models is the linear bandit, which assumes a linear relationship actions and the *expected* rewards. In this setting, a standard learning strategy consists in estimating the reward model by ridge regression coupled with an appropriate exploration scheme, e.g. optimism (Abbasi-Yadkori et al., 2011), Thompson (Agrawal and Goyal, 2013; Abeille and Lazaric, 2017) or information-directed sampling (Russo and Van Roy, 2014; Kirschner et al., 2021).

As discussed in Chapter 3, an obstacle to the adoption by practitioners of such sequential models for critical recommendations (agricultural practices concerned with food security, medical treatment optimisation, etc.) is the lack of efficient and theoretically sound algorithms beyond the simplistic expectation risk measure. In this chapter, we investigate an extension of the linear bandit where a given risk measure, rather than the mean, is linearly parametrised by the chosen actions. Specifically, we consider the case of convex risk measures which can be *elicited* as minimisers of certain loss functions, which naturally extends the standard ridge regression. This definition covers expectiles and entropic risk, and may be extended to mean-variance and conditional value-at-risk using multivariate risk measures. To our knowledge, this setting is new, although related to existing approaches, such as bandits with regression oracles (Foster and Rakhlin, 2020) and generalised linear bandits (GLB) (Filippi et al., 2010; Li et al., 2017; Faury et al., 2020), that extend beyond linearity while still optimising mean rewards.

We introduce a generalisation of LinUCB to these elicitable risk measures. Except for the standard mean-linear bandit setting, learning the linear mapping between actions and risk measures cannot be performed sequentially, which presents theoretical and numerical challenges similar to GLB. We derive *time-uniform* confidence sets (Proposition 5.12) based on the method of mixtures and introduce a geometric condition (Lemma 5.15) to ensure sublinear pseudo regret (with high probability, uniformly in time) in this new setting (Theorem 5.18). Using recent developments on time-uniform matrix concentration, we further strengthen the regret bound in the case of stochastic actions with a known covariance lower bound (Theorem 5.23). To mitigate the numerical burden, we introduce an episodic version of LinUCB with online gradient descent approximation (Theorem 5.26), inspired by previous works on online regression (Korda et al., 2015) and the recent literature on approximate Thompson sampling for GLB (Ding et al., 2021).

## 5.1 Contextual bandits with risk

We consider the stationary contextual bandit setting of Section 1.2 where an agent sequentially observes at time $t \in \mathbb{N}$ a subset $\mathcal{X}_t \subseteq \mathbb{R}^d$, then chooses an action $X_t \in \mathcal{X}_t$ and receives a stochastic reward $Y_t$, the distribution of which is dependent on $X_t$, and is denoted by $\nu(X_t)$ (we drop the superscript $\mathbb{X}$ to reference the dependency on the policy $\mathbb{X} = (X_t)_{t\in\mathbb{N}}$ to avoid cluttering). We recall the definition of the adapted contextual bandit filtration $(\bar{\mathcal{G}}_t)_{t\in\mathbb{N}} = (\sigma(\mathcal{X}_1, X_1, Y_1, \dots, \mathcal{X}_{t-1}, X_{t-1}, Y_{t-1}, \mathcal{X}_t, X_t)_{t\in\mathbb{N}}$, which encodes the information available to the agent after observing the decision set and picking an action at time $t \in \mathbb{N}$ but before collecting the corresponding reward.

Informally, the goal of the agent is to learn a representation of the mapping $\nu$ in order to select actions that induce high rewards. As hinted in Section 1.2, a standard inductive bias in this context is to assume a linear statistical model between rewards and contexts, typically of the form $Y_t = \langle \theta^\star, X_t \rangle + \eta_t$, where $(\eta_t)_{t\in\mathbb{N}}$ is some centred stochastic process. We recall that our definition of linear bandits extends beyond this setting by assuming the following factorisation

$$\mathcal{X} \xrightarrow{\ \ell^\star\ } \mathbb{R}^p \xrightarrow{\ \varphi\ } \mathcal{M}_1^+(\mathbb{R})\,, \qquad \nu = \varphi \circ \ell^\star : \mathcal{X} \to \mathcal{M}_1^+(\mathbb{R})$$

where $\ell^\star\colon \mathcal{X} \to \mathbb{R}^p$ denotes a linear map. In other words, the reward distribution (but not necessarily its mean) is linearly parametrised by the chosen action. We denote such a linear bandit by $(\varphi, \ell^\star)$. When the distribution depends on a single parameter ($p = 1$), we represent the linear form by $\ell^\star\colon x \in \mathcal{X} \mapsto \langle \theta^\star, x\rangle$, where $\theta^\star \in \mathbb{R}^d$, and we use the notation $(\varphi, \theta^\star)$, or equivalently we say that it is represented by $Y \sim \varphi(\langle \theta^\star, X\rangle)$. In the following, we also denote by $\Theta \subseteq \mathbb{R}^d$ parameter space. We use the notation $\mathbb{P}$ and $\mathbb{E}$ the refer to the probability and expectation induced by this contextual bandit model.

As an example, let us consider the following Gaussian mapping

$$\begin{aligned} \nu\colon \mathcal{X} &\longrightarrow \mathcal{M}_1^+(\mathbb{R}) \\ x &\longmapsto \mu(x) + \sigma(x)\mathcal{N}(0,1)\,. \end{aligned} \tag{5.1}$$

If $\sigma$ is constant and $\mu(x) = \langle \theta_\mu^\star, x\rangle$ for some $\theta_\mu^\star \in \Theta$, we recover a standard linear bandit model, in which the goal is to maximise the cumulative average rewards $\sum_{t=1}^T \mu(X_t)$ (or equivalently, minimise the pseudo regret associated with the expectation risk measure). However, in many applications, the agent may be averse to high reward volatility, which can be encoded by $\mu(x) - \lambda\sigma(x)$ for some $\lambda > 0$. We detail in Appendix D.1 how many standard risk measures (entropic, $p$-expectile) realise this mean-variance tradeoff.

## Overview of risk measures

**Convex loss.** In the bandit setting, the agent faces the classical dilemma between exploitation (playing the most promising actions) and exploration (playing other actions to gain information). In most algorithms, the exploitation takes the form of a supervised estimation that consists in learning the mapping $\varphi$ at time $t$ from the past observations $\{(X_s, Y_s),\ 1 \leqslant s \leqslant t-1\}$. When the expected reward is parametrised by $Y_t = \langle \theta^\star, X_t \rangle + \eta_t$ with $\mathbb{E}[\eta_t \mid \bar{\mathcal{G}}_t] = 0$, a standard strategy consists in estimating $\theta^\star$ by ridge regression, i.e. as the unique solution to the strongly convex minimisation problem

$$\widehat{\theta}_t = \min_{\theta \in \Theta} \sum_{s=1}^{t-1} (Y_s - \langle \theta, X_s \rangle)^2 + \frac{\alpha}{2} \|\theta\|_2^2 , \tag{5.2}$$

where $\alpha \geqslant 0$ is a regularisation parameter. Assuming for now the solution is in the interior of $\Theta$, the solution can be written as

$$\widehat{\theta}_t = (V_t^\alpha)^{-1} \sum_{s=1}^{t-1} Y_s X_s \in \mathbb{R}^d , \tag{5.3}$$

where we define the $d \times d$ positive (semi)definite matrix

$$V_t^\alpha = \sum_{s=1}^{t-1} X_s X_s^\top + \alpha I_d \in \mathcal{S}_d^+(\mathbb{R}) . \tag{5.4}$$

This approach presents several advantages: it can be computed efficiently via sequential matrix inversion (with complexity $\mathcal{O}(d^2)$ at each step thanks to the Sherman-Morrison formula for the rank-one update $V_{t+1}^\alpha = V_t^\alpha + X_t X_t^\top$ starting from $V_0^\alpha = \alpha I_d$) and explicit confidence ellipsoids for $\theta^\star$ can be constructed analytically around $\widehat{\theta}_t$ to tune exploration (Abbasi-Yadkori et al., 2011). The implicit limitation of this procedure is that it can only estimate the expectation $\mathbb{E}[Y]$, which is the solution of the quadratic loss minimisation problem $\operatorname{argmin}_{\theta \in \Theta} \mathbb{E}[(Y - \langle \theta, X \rangle)^2]$. We call this standard setting the *mean-linear* bandit.

**Elicitable risk measure.** Drawing inspiration from this simple example, we aim to estimate other characteristics of the reward distribution than the mean, using other convex losses than the quadratic loss.

**Definition 5.1** (Elicitable risk measure). *Let $p \in \mathbb{N}$ and $\mathcal{L} \colon \mathbb{R} \times \mathbb{R}^p \to \mathbb{R}$ be a strongly convex loss function. The **risk measure** elicited by the loss $\mathcal{L}$ is defined as*

$$\begin{aligned} \rho_{\mathcal{L}} \colon \mathcal{M}_1^+(\mathbb{R}) &\longrightarrow \mathbb{R}^p \cup \{\infty\} \\ \nu &\longmapsto \arg\min_{\xi \in \mathbb{R}^p} \mathbb{E}_{Y \sim \nu}\left[\mathcal{L}(Y, \xi)\right] . \end{aligned} \tag{5.5}$$

*Similarly, we define the conditional risk measure elicited by the loss $\mathcal{L}$ with respect to the adapted bandit filtration for all $t \in \mathbb{N}$ by*

$$\begin{aligned} \rho_{\mathcal{L}}(\cdot \mid \bar{\mathcal{G}}_t) \colon \mathcal{M}_1^+(\mathbb{R}) &\longrightarrow \mathbb{R}^p \cup \{+\infty\} \\ \nu &\longmapsto \arg\min_{\xi \in \mathbb{R}^p} \mathbb{E}_{Y \sim \nu}\left[\mathcal{L}(Y, \xi) \mid \bar{\mathcal{G}}_t\right] . \end{aligned} \tag{5.6}$$

*Finally, we recall the definition of $\rho_{\mathcal{L}}$-integrable probability measures as $\mathbb{L}^{\rho_{\mathcal{L}}}(\mathbb{R}^p) = \rho_{\mathcal{L}}^{-1}(\mathbb{R}^p)$.*

Note that the assumption that $\mathcal{L}$ is strongly convex ensures that $\rho_{\mathcal{L}}$ is well-defined (it could be relaxed to simply assuming that the solution of the minimisation problem that defines $\rho_{\mathcal{L}}(\nu)$ is unique for all distributions $\nu \in \mathbb{L}^{\rho}_{\mathcal{L}}(\mathbb{R}^p)$). With this definition, $\rho_{\mathcal{L}}(\nu)$ is a vector in $\mathbb{R}^p$; when $p = 1$, we call these *scalar*, or *first order* elicitable risk measures (Ziegel, 2016). Examples of such risk measures include the mean, the median, and more generally any quantile and expectile, which we further discuss below as special cases of risk measures associated with convex potentials. Other examples are any generalised moments $\rho(\nu) = \mathbb{E}_{Y\sim\nu}[T(Y)]$, where $T \colon \mathbb{R} \to \mathbb{R}$ is a $\nu$-integrable mapping, and the entropic risk defined by $\rho_{\mathcal{L}}(\nu) = \frac{1}{\gamma} \log \mathbb{E}_{Y\sim\nu}[e^{\gamma Y}]$ (Maillard, 2013). Unfortunately, not all risk measures commonly encountered in the risk literature are first order elicitable. In particular, neither the variance nor the CVaR can be expressed as scalar risk measures with respect to a convex loss (Fissler et al., 2016; Fissler and Ziegel, 2016). They are however are second order elicitable, in the sense that the pairs (mean, variance) and (VaR, CVaR) are jointly elicitable. We refer to Appendix D.1 for a summary and further interpretations of elicitable risk measures. In the rest of this chapter, we only consider scalar risk measure and leave the extension to measures like CVaR for further work. Moreover, we implicitly assume that all considered measures $\nu$ have finite risk measure, i.e. $\nu \in \mathbb{L}^{\rho}_{\mathcal{L}}(\mathbb{R})$.

We now formalise the link between reward distributions of linear bandits and convex losses. The goal here is to study linear bandits where generic elicitable risk measures of the reward distributions are linear functions of the actions played by the agent, the same way expected rewards are for mean-linear bandits.

**Definition 5.2** (Loss adapted to a bandit). *A strongly convex loss $\mathcal{L}\colon \mathbb{R} \times \mathbb{R}^p \to \mathbb{R}$ is said to be adapted to a linear bandit $(\varphi, \ell^\star)$ if*

$$\forall x \in \mathcal{X},\ \ell^\star \in \underset{\ell\colon \mathcal{X}\to\mathbb{R}^p \text{ linear}}{\operatorname{argmin}} \mathbb{E}_{Y\sim\varphi\circ\ell(x)} \left[\mathcal{L}\left(Y, \ell(x)\right)\right] . \tag{5.7}$$

*In other words, whenever action $x \in \mathcal{X}$ is played, the risk measure $\rho_{\mathcal{L}}$ of the reward distribution generated by the bandit is a linear function of $x$.*

**Convex potential.** A special case of interest is when the convex loss $\mathcal{L} = \mathcal{L}_\psi$ derives from a potential $\psi$, that is when $\mathcal{L}_\psi\colon (y, \xi) \in \mathbb{R} \times \mathbb{R} \mapsto \psi(y - \xi)$. We denote by $\rho_\psi$ the corresponding risk measure. This setting covers the ordinary least square potential associated to the mean, but also quantiles and expectiles. We assume the reader to be familiar with the former, but perhaps less so with the latter. Following Newey and Powell (1987), we define the $p$-expectile of $\nu \in \mathbb{L}^{\rho_\psi}(\mathbb{R}) = \mathbb{L}^2(\mathbb{R})$ for $p \in (0, 1)$ as

$$e_p(\nu) = \underset{\xi\in\mathbb{R}}{\operatorname{argmin}}\, \mathbb{E}_{Y\sim\nu}[|p - \mathbb{1}_{Y<\xi}|(Y - \xi)^2]\,, \tag{5.8}$$

which is uniquely defined by strong convexity. Expectiles have been studied in particular in the context of risk management (Bellini and Di Bernardino, 2017) and risk-aware Bayesian optimisation (Picheny et al., 2022). Furthermore, under some symmetry conditions, quantiles and expectiles are known to coincide (Abdous and Remillard, 1995), and thus expectiles can be seen as a smooth (in particular differentiable) generalisation of quantiles (see Philipps (2022) for further interpretation of the notion of expectiles). We refer the reader to Table 5.1 for a summary of risk measures elicited by convex potentials.

**Table 5.1 –** Example of Risk Measures Elicited by Convex Potentials.

| Name | Potential $\psi(z)$ | Risk measure $\rho_\psi$ |
|---|---|---|
| Mean | $z^2/2$ | $\rho_\psi = \int y\nu(dy)$ |
| Quantile $p \in (0, 1)$ | $(p - \mathbb{1}_{z<0})z$ | $\int_{-\infty}^{\rho_\psi} \nu(dy) = p$ |
| Expectile $p \in (0, 1)$ | $\lvert p - \mathbb{1}_{z<0}\rvert z^2$ | $(1-p) \int_{-\infty}^{\rho_\psi} \lvert y-\rho_\psi\rvert \nu(dy)$ $= p \int_{\rho_\psi}^{\infty} \lvert y - \rho_\psi\rvert \nu(dy)$ |

In the terminology defined above, the ordinary least square potential is adapted to the mean-linear reward model $Y_t = \langle \theta^\star, X_t \rangle + \eta_t$ with $\mathbb{E}[\eta_t \mid \bar{\mathcal{G}}_t] = 0$. More generally, such additive decompositions exist for losses derived from potentials, as evidenced by the following lemma.

**Lemma 5.3.** *Assume that the loss $\mathcal{L}_\psi$ derived from a strongly convex and differentiable potential $\psi$ is adapted to the linear bandit $(\varphi, \theta^\star)$. Then there exists a stochastic process $(\eta_t)_{t\in\mathbb{N}}$ such that the bandit is represented at time $t \in \mathbb{N}$ by $Y_t \sim \langle \theta^\star, X_t \rangle + \eta_t$ with $\rho_\psi(\eta_t \mid \bar{\mathcal{G}}_t) = 0$.*

*Proof of Lemma 5.3.* This result is a straightforward application of the following technical lemma.

**Lemma 5.4** (Risk Measures $\rho_\psi$ are additive). *Let $\psi\colon \mathbb{R} \to \mathbb{R}$ be a strongly convex, differentiable function, and $\nu \in \mathbb{L}^{\mathcal{L}_\psi}(\mathbb{R})$. Then $\rho_\psi(\nu + c) = \rho_\psi(\nu) + c$.*

*Proof of Lemma 5.4.* For the sake of simplicity, we assume $\nu$ admits a density $p$ (with respect to e.g. the Lebesgue measure) and that $\psi$ and $p$ are regular enough to allow for differentiation under the following integral. Then the risk measure associated with $\mathcal{L}_\psi$ reads

$$\rho_\psi(\nu) = \operatorname*{argmin}_{\xi\in\mathbb{R}} \int \psi\,(y - \xi)\, p(y) dy \tag{5.9}$$

and the first order condition gives

$$\int \psi'(y - \rho_\psi(\nu))p(y)dy = 0\,. \tag{5.10}$$

Similarly, for any $c \in \mathbb{R}$, we have

$$\int \psi'(y - \rho_\psi(\nu + c))p(y - c)dy = 0\,, \tag{5.11}$$

since the density of $\nu + c$ is given by $y \mapsto p(y - c)$. We now deduce from a simple change of variable $z = y - c$ that $\int \psi'(z + c - \rho_\psi(\nu + c))p(z)dz = 0$, which shows that $\rho_\psi(\nu + c) - c$ is also a minimiser of $\xi \mapsto \int \psi\,(y - \xi)\, p(y)dy$. By uniqueness ($\psi$ is strongly convex), we deduce that $\rho_\psi(\nu + c) = \rho_\psi(\nu) + c$. ∎

Define the process $(\eta_t)_{t\in\mathbb{N}}$ by $\eta_t = Y_t - \langle \theta^\star, X_t \rangle$ for $t \in \mathbb{N}$. To compute $\rho_\psi(\nu_t \mid \bar{\mathcal{G}}_t)$, note that $X_t$ is measurable with respect to $\bar{\mathcal{G}}_t$, therefore by Lemma 5.4 and the properties of conditional expec-tation, we have that $\rho_\psi(\eta_t \mid \bar{\mathcal{G}}_t) = \rho_\psi\left(Y_t \mid \bar{\mathcal{G}}_t\right) - \langle\theta^\star, X_t\rangle = \rho_\psi\left(\varphi\left(\langle\theta^\star, X_t\rangle\right) \mid \bar{\mathcal{G}}_t\right) - \langle\theta^\star, X_t\rangle = 0$ by definition of $\mathcal{L}_\psi$ being adapted to the bandit $(\varphi, \theta^\star)$. ■

**Non-unicity of adapted loss.** In general, a risk measure can be described by multiple different adapted losses. First, the set of losses that elicit a given risk measure is a cone invariant by scalar translation, i.e. $\rho_{\alpha\mathcal{L}+\beta} = \rho_{\mathcal{L}}$ for all $\alpha > 0$ and $\beta \in \mathbb{R}$. Other less trivial examples of non-unicity arise even for the simple mean criterion. Theorem 1 in Banerjee et al. (2005) shows that $\mathbb{E}_{Y\sim\nu}[Y] = \rho_{\mathcal{B}_\psi}(\nu)$ where $\psi$ is any strictly convex, differentiable function and $\mathcal{B}_\psi \colon (y, \xi) \mapsto \psi(y) - \psi(\xi) - \psi'(\xi)(y - \xi)$ is the Bregman divergence induced by $\psi$, which generalises the quadratic potential. In fact, every continuously differentiable loss that elicits the mean has this form (Theorem 3 and 4 in Banerjee et al. (2005)). In particular, this shows that the converse of Lemma 5.3 cannot hold: except for the quadratic potential, $\mathcal{B}_\psi$ does not derive from a convex potential, even though it is an adapted loss to the mean-linear bandit, which admits an additive decomposition $Y_t \sim \langle\theta^\star, X_t\rangle + \eta_t$ with $\mathbb{E}[\eta_t \mid \bar{\mathcal{G}}_t] = 0$. Similarly, the pairs (mean, variance) and $(\mathrm{VaR}, \mathrm{CVaR})$ can be elicited by families indexed by differentiable, strictly convex functions (Table D.1, Appendix D.1).

### Contextual bandits with elicitable risk measures

We recall the notion of pseudo regret for contextual bandits (Definition 1.9) and specify it to the setting of linear bandits with elicitable risk measures.

**Definition 5.5** (Risk-aware pseudo-regret). *For a linear bandit $(\varphi, \theta^\star)$, we define the pseudo-regret associated to a risk measure $\rho_{\mathcal{L}}$ and a sequence of actions $(X_t)_{t\in\mathbb{N}}$ as the stochastic process*

$$(\mathcal{R}_T)_{T\in\mathbb{N}} = \left(\sum_{t=1}^{T} \rho_{\mathcal{L}}\left(\varphi(\langle\theta^\star, X_t^\star\rangle)\right) - \rho_{\mathcal{L}}\left(\varphi(\langle\theta^\star, X_t\rangle)\right)\right)_{T\in\mathbb{N}}, \tag{5.12}$$

*where $X_t^\star = \operatorname{argmax}_{x\in\mathcal{X}_t} \rho_{\mathcal{L}}\left(\varphi(\langle\theta^\star, x\rangle)\right)$ is the optimal action with respect to $\rho_{\mathcal{L}}$ at time $t \in \mathbb{N}$.*

By definition, if the loss $\mathcal{L}$ is adapted to the linear bandit, this notion of pseudo regret reduces to $\mathcal{R}_T = \sum_{t=1}^{T} \langle\theta^\star, X_t^\star\rangle - \langle\theta^\star, X_t\rangle$, which is formally the same as the standard pseudo regret for mean-linear bandits. What differs though is the meaning of $\langle\theta^\star, X_t\rangle$, which now represents an elicitable risk measure for the reward distribution. As an example, this paves a way for expectile-linear bandit of the form $Y_t = \langle\theta^\star, X_t\rangle + \eta_t$ where the conditional expectile of $\eta_t$ is zero and expectile rewards are measured as linear forms of the actions $\langle\theta^\star, X_t\rangle$.

**Supervised estimation of $\theta^\star$.** Similarly to how ridge regression provides natural estimators of the mean, we define

$$\widehat{\theta}_t = \operatorname*{argmin}_{\theta \in \Theta} \sum_{s=1}^{t-1} \mathcal{L}(Y_s, \langle \theta, X_s \rangle) + \frac{\alpha}{2} \|\theta\|_2^2 \,, \tag{5.13}$$

which corresponds to the empirical risk minimisation estimator, also known as an M-estimator Huber (2004), associated to loss $\mathcal{L}$, with $\mathbb{L}^2$ regularisation parameter $\alpha > 0$ (again, $\widehat{\theta}_t$ is well-defined by strong convexity). Assuming that $\mathcal{L}$ is differentiable and that $\widehat{\theta}_t$ is in the interior of $\Theta$, it is characterised by the first order condition, also known as Z-estimation (Kosorok, 2008),

$$\alpha \widehat{\theta}_t = - \sum_{s=1}^{t-1} \partial \mathcal{L}(Y_s, \langle \widehat{\theta}_t, X_s \rangle) X_s \,, \tag{5.14}$$

where $\partial \mathcal{L}(y, \xi)$ stands for the derivative of $\xi \mapsto \mathcal{L}(y, \xi)$. When $\widehat{\theta}_t$ is not in the interior of $\Theta$, an additional projection onto $\Theta$ is necessary, which we denote by the operator $\Pi$ (such an operator is detailed in Section 5.2). We also define the mapping

$$\begin{aligned} H_t^\alpha \colon \Theta &\longrightarrow \mathcal{S}_d^+(\mathbb{R}) \\ \theta &\longmapsto \sum_{s=1}^{t-1} \partial^2 \mathcal{L}\left(Y_s, \langle \theta, X_s \rangle\right) X_s X_s^\top + \alpha I_d \,, \end{aligned} \tag{5.15}$$

which represents the Hessian of the empirical loss associated with the minimisation problem, where $\partial^2$ stands for the second order derivative with respect to the second coordinate.

Of note, when the loss $\mathcal{L}$ derives from the quadratic potential $\psi \colon \xi \in \mathbb{R} \mapsto \xi^2/2$, it holds that $H_t^\alpha(\theta) = V_t^\alpha$ for all $\theta \in \Theta$ and we thus fall back to the mean-linear case. For all other choices of the loss function $\mathcal{L}$, the Hessian matrix $H_t^\alpha(\theta)$ depends on $\theta$, and in particular no closed-form expression of $\widehat{\theta}_t$ in terms of the inverse of $H_t^\alpha$ is available. As we detail in the next sections, this introduces technical challenges to the analysis of linear bandit algorithms and forces the use of convex programming algorithms to numerically evaluate $\widehat{\theta}_t$.

**Remark 5.6** . *Similar complications arise in the case of generalised linear bandits (GLB) $Y_t = \mu(\langle\theta^\star, X_t\rangle) + \eta_t$, with $\mathbb{E}[\eta_t \mid \bar{\mathcal{G}}_t] = 0$ and $\mu$ a nonlinear link function. Under parametric assumptions on $Y_t$ (typically one-dimensional exponential family), GLB can be seen as a special case of the risk-aware setting with $\mathcal{L}$ the negative log-likelihood loss, with the analogy $\mu \leftrightarrow \partial\mathcal{L}$. Despite this formal similarity, GLB is designed solely to optimise the mean criterion. Another difference with our setting is that pseudo regret for GLB is commonly defined as $\sum_{t=1}^T \mu(\langle\theta^\star, X_t^\star\rangle) - \mu(\langle\theta^\star, X_t\rangle)$, which is smaller than $\mathcal{R}_T$ when $\mu$ is contracting.*

**Remark 5.7** . *At selection time $t$, i.e. after observing the action set $\mathcal{X}_t$ but before playing $X_t$, the agent has access to all possible actions $x \in \mathcal{X}_t$ (measurable with respect to the natural $\sigma$-algebra $\mathcal{G}_t$). Still, the regression scheme of equation (5.13) only involves $(X_s, Y_s)_{s=1}^{t-1}$. Instead of an isotropic ridge regularisation term $\|\theta\|^2$, we thus consider the directional penalisation $\langle\theta, x\rangle$ and build confidence sets for the corresponding estimators in order to assess the candidate action $x \in \mathcal{X}_t$. This was recently studied in Ouhamma et al. (2021), leading to explicit UCB policies. Of note, this allows to lift the boundedness requirements on $\theta$ and $\mathcal{X}_t$, corresponding to our Assumptions 5.16 and 5.17. This improvement being orthogonal to the risk focus of this chapter, we only leave it as a remark.*

**Extension of LinUCB to convex losses.** The main benefit of the formulation of risk-awareness in terms of convex losses is that it suggests a transparent generalisation of the standard LinUCB algorithm (OFUL in Abbasi-Yadkori et al. (2011), Ch.19 in Lattimore and Szepesvári (2020)), essentially substituting the least-squares estimate with the empirical risk minimiser associated with $\mathcal{L}$. The general idea of such optimistic algorithms is to play at time $t$ the action $x \in \mathcal{X}_t$ with the highest plausible reward. In the mean-linear case with ridge regression, this highest plausible reward takes the form of $\langle\widehat{\theta}_t, x\rangle + \gamma_t(x)$, where $\gamma_t(x)$ is a certain action-dependent quantity also known as the *exploration bonus*. We write the general structure of our extension of LinUCB (CR for Convex Risk) in Algorithm 3.

## 5.2 Regret analysis of LinUCB-CR

The goal of this section is to derive an exploration bonus sequence $(\gamma_t)_{t\in\mathbb{N}}$ and a projection operator $\Pi$ that ensure sub-linear pseudo regret of the corresponding LinUCB instance. To this end, we introduce the following control on the curvature of the adapted loss $\mathcal{L}$.

**Algorithm 3** LinUCB-CR

**Input:** regularisation parameter $\alpha$, projection operator $\Pi$,
exploration bonus sequence $(\gamma_t)_{t\in\mathbb{N}}$.
**Initialisation:** Observe $\mathcal{X}_1$, $t = 1$.
**while** *continue* **do**

$\widehat{\theta} \leftarrow \operatorname*{argmin}_{\theta\in\mathbb{R}^d} \sum_{s=1}^{t-1} \mathcal{L}(Y_s, \langle\theta, X_s\rangle) + \frac{\alpha}{2}\|\theta\|_2^2$ ; ▷ `Empirical risk minimisation`

$\bar{\theta} \leftarrow \Pi(\widehat{\theta})$ ; ▷ `Projection`

$X_t \leftarrow \operatorname*{argmax}_{x\in\mathcal{X}_t} \langle\bar{\theta}, x\rangle + \gamma_t(x)$ ; ▷ `Play`

Observe $Y_t$ and $\mathcal{X}_{t+1}$; ▷ `Observe reward and next action set`

$t \leftarrow t + 1$ .

**Assumption 5.8** (Bounded Loss Curvature). *There exists $m$ and $M$ such that*

$$\forall y, \xi \in \mathbb{R},\ m \leqslant \partial^2\mathcal{L}(y, \xi) \leqslant M\,. \tag{5.16}$$

*We call the parameter $\kappa = \frac{M}{m}$ the **conditioning** of $\mathcal{L}$.*

**Remark 5.9**. *This assumption is reminiscent of the standard lower bound on the derivative of the link function $\mu'$ commonly encountered in the GLB literature.*

### Martingale property and concentration

A key property for the analysis of mean-linear bandits is that the sum process $\sum_{s=1}^{t-1} \eta_s X_s$ naturally defines a vector-valued martingale in $\mathbb{R}^d$ with respect to the filtration $\left(\bar{\mathcal{G}}_t\right)_{t\in\mathbb{N}}$ (Abbasi-Yadkori et al., 2011). This is not the case in general for bandits associated with generic convex losses. Instead, for a given loss $\mathcal{L}$, we know that $\theta^\star = \operatorname{argmin}_\theta \mathbb{E}[\mathcal{L}(Y_t, \langle\theta, X_t\rangle) \mid \bar{\mathcal{G}}_t]$. Assuming $\mathcal{L}$ is differentiable and using the shorthand $\partial^j\mathcal{L}_t^\star = \partial^j\mathcal{L}(Y_t, \langle\theta^\star, X_t\rangle)$ for $j \in \mathbb{N}$ and $t \in \mathbb{N}$, this implies that $\mathbb{E}[\partial^1\mathcal{L}_t^\star \mid \bar{\mathcal{G}}_t] = 0$ since $X_t$ is measurable with respect to $\bar{\mathcal{G}}_t$. A direct consequence of this property of $(\partial^1\mathcal{L}_t^\star)_{t\in\mathbb{N}}$ is that

$$S_t = \sum_{s=1}^{t-1} \partial^1\mathcal{L}_s^\star X_s \in \mathbb{R}^d \tag{5.17}$$

defines a $\left(\bar{\mathcal{G}}_t\right)_{t\in\mathbb{N}}$-martingale. This process is at the heart of the next proposition, which establishes multivariate confidence bounds using the method of mixtures. To this end, we detail below a helpful transformation of the sum process $(S_t)_{t\in\mathbb{N}}$ into a nonnegative supermartingales (Lemma 5.11) under a standard sub-Gaussian assumption (Assumption 5.10) and state the high-probability uniform deviation bound we obtain (Proposition 5.12).

**Assumption 5.10** (Sub-Gaussian). *$\partial^1\mathcal{L}^\star$ is a conditionally sub-Gaussian process with respect to the filtration $(\bar{\mathcal{G}}_t)_{t\in\mathbb{N}}$, i.e. there exists $R > 0$ such that*

$$\forall t \in \mathbb{N},\ \forall \lambda \in \mathbb{R},\ \log \mathbb{E}\left[\exp\left(\lambda \partial^1 \mathcal{L}_t^\star\right) \,\middle|\, \bar{\mathcal{G}}_t\right] \leqslant \frac{\lambda^2 R^2}{2}\,. \tag{5.18}$$

**Lemma 5.11** (Supermartingale control). *Under Assumptions 5.8 and 5.10, there exists $\sigma > 0$ such that for any $t \in \mathbb{N}$ and $\lambda \in \mathbb{R}^d$, the following inequality holds:*

$$\mathbb{E}\left[\exp\left(\langle\lambda, X_t\rangle\partial^1\mathcal{L}_t^\star - \frac{\sigma^2}{2}\langle\lambda, X_t\rangle^2\partial^2\mathcal{L}_t^\star\right) \,\middle|\, \bar{\mathcal{G}}_t\right] \leqslant 1\,. \tag{5.19}$$

*Proof of Lemma 5.11.* Assumption 5.8 implies that $\partial^2\mathcal{L}_t^* \geqslant m$, therefore it is sufficient to show that there exists $\sigma' > 0$ such that

$$\mathbb{E}\left[\exp\left(\langle\lambda, X_t\rangle\partial^1\mathcal{L}_t^* - \frac{m\sigma'^2}{2}\langle\lambda, X_t\rangle^2\right) \,\middle|\, \bar{\mathcal{G}}_t\right] \leqslant 1\,. \tag{5.20}$$

Since $X_t$ is $\bar{\mathcal{G}}_t$-measurable, this is equivalent to

$$\mathbb{E}\left[\exp\left(\langle\lambda, X_t\rangle\partial^1\mathcal{L}_t^* \middle| \bar{\mathcal{G}}_t\right]\right) \leqslant \exp\left(\frac{m\sigma'^2}{2}\langle\lambda, X_t\rangle^2\right)\,, \tag{5.21}$$

which follows from the sub-Gaussian property of the process $\partial^1\mathcal{L}^*$ with parameter $\sigma = \sqrt{m}\sigma'$ (Assumption 5.10). ∎

**Proposition 5.12** (Method of mixtures with convex loss). *Let $\beta > 0$ and $\delta \in (0,1)$. Under Assumptions 5.8 and 5.10, we have the following inequality:*

$$\mathbb{P}\left(\exists t \in \mathbb{N},\ \|S_t\|^2_{H_t^\beta(\theta^\star)^{-1}} \geqslant \sigma^2\left(2\log\frac{1}{\delta} + \log\frac{\det H_t^\beta(\theta^\star)}{\det \beta I_d}\right)\right) \leqslant \delta\,. \tag{5.22}$$

*Proof of Proposition 5.12.* The proof is similar to that of Proposition 1.26 and follows the method of mixture. To this end, we first construct a $(\bar{\mathcal{G}}_t)_{t\in\mathbb{N}}$-nonnegative supermartingale as follows: for $\lambda \in \mathbb{R}^d$, we define the process for $t \in \mathbb{N}$ by

$$M_t^\lambda = \exp\left(\lambda^\top S_t - \frac{\sigma^2}{2}\|\lambda\|^2_{H_t^0(\theta^\star)}\right)\,. \tag{5.23}$$

We recall the expression of the Hessian $H_t^0(\theta) = \sum_{s=1}^{t-1} \partial^2\mathcal{L}\left(Y_s, \langle\theta, X_s\rangle\right) X_s X_s^\top$ and that in particular $\|\lambda\|^2_{H_t^0(\theta^\star)} = \sum_{s=1}^{t-1} \partial^2\mathcal{L}\left(Y_s, \langle\theta, X_s\rangle\right)\left(\lambda^\top X_s\right)^2$. This process is $(\bar{\mathcal{G}}_{tt\in\mathbb{N}})$-adapted, integrable, nonnegative and is indeed a supermartingale since

$$\begin{aligned}
\mathbb{E}\left[M_{t+1}^\lambda \,\middle|\, \bar{\mathcal{G}}_t\right] &= \mathbb{E}\left[\exp\left(\lambda^\top S_{t+1} - \frac{\sigma^2}{2}\|\lambda\|^2_{H_{t+1}^0(\theta^\star)}\right) \,\middle|\, \bar{\mathcal{G}}_t\right] \\
&= \mathbb{E}\left[\exp\left(\lambda^\top S_t - \frac{\sigma^2}{2}\|\lambda\|^2_{H_t^0(\theta^\star)} + \partial\mathcal{L}\left(Y_t, \langle\theta^\star, X_t\rangle\right)\lambda^\top X_t - \frac{\sigma^2}{2}\partial^2\mathcal{L}\left(Y_t, \langle\theta^\star, X_t\rangle\right)\left(\lambda^\top X_t\right)^2\right) \,\middle|\, \bar{\mathcal{G}}_t\right] \\
&= e^{\lambda^\top S_t - \frac{\sigma^2}{2}\|\lambda\|^2_{H_t^0(\theta^\star)}} \underbrace{\mathbb{E}\left[\exp\left(\partial\mathcal{L}\left(Y_t, \langle\theta^\star, X_t\rangle\right)\lambda^\top X_t - \frac{\sigma^2}{2}\partial^2\mathcal{L}\left(Y_t, \langle\theta^\star, X_t\rangle\right)\left(\lambda^\top X_t\right)^2\right) \,\middle|\, \bar{\mathcal{G}}_t\right]}_{\leqslant 1} \\
&\leqslant e^{\lambda^\top S_t - \frac{\sigma^2}{2}\|\lambda\|^2_{H_t^0(\theta^\star)}} \qquad \text{(Lemma 5.11)} \\
&= M_t^\lambda\,. \qquad (5.24)
\end{aligned}$$

Now we construct a new supermartingale by mixing the $(M_t^\lambda)_{t\in\mathbb{N}}$. Formally, let $\Lambda$ a $\mathbb{R}^d$-valued random variable independent of the rest and $M_t = \mathbb{E}[M_t^\Lambda \mid \bar{\mathcal{G}}_\infty]$ where $\bar{\mathcal{G}}_\infty = \sigma(\bigcup_{t\in\mathbb{N}}\bar{\mathcal{G}}_t)$. If $\Lambda$ admits a density $p$ with respect to the Lebesgue measure, we have $M_t = \int_{\mathbb{R}^d} M_t^\lambda p(\lambda)d\lambda$. For the choice $\Lambda \sim \mathcal{N}(0, \frac{1}{\beta\sigma^2}I_d)$ with $\beta > 0$, we have, by completing the square in the exponential,

$$M_t = \frac{(\beta\sigma^2)^{d/2}}{(2\pi)^{d/2}}\int_{\mathbb{R}^d}\exp\left(-\lambda^\top S_t + \frac{\sigma^2}{2}\left(\lambda^\top\left(H_t^0(\theta^\star) + \beta I_d\right)\lambda\right)\right)d\lambda$$

$$
= \frac{(\beta\sigma^2)^{d/2}}{(2\pi)^{d/2}} \exp\left(\frac{\sigma^2}{2}\bar{\lambda}^\top H_t^\beta(\theta^\star)\bar{\lambda}\right) \int_{\mathbb{R}^d} \exp\left(-\frac{\sigma^2}{2}\left(\lambda - \bar{\lambda}\right)^\top H_t^\beta(\theta^\star)\left(\lambda - \bar{\lambda}\right)\right) d\lambda
$$

$$
= \left(\frac{\beta^d}{\det H_t^\beta(\theta^\star)}\right)^{\frac{1}{2}} \exp\left(\frac{\sigma^2}{2}\bar{\lambda}^\top H_t^\beta(\theta^\star)\bar{\lambda}\right), \tag{5.25}
$$

where $\bar{\lambda} = \frac{1}{\sigma^2} H_t^\beta(\theta^\star)^{-1} S_t$ and $H_t^\beta(\theta) = H_t^0(\theta) + \beta I_d$ is the regularised Hessian, which is positive definite and hence invertible. This expression further simplifies to

$$
M_t = \left(\frac{\det \beta I_d}{\det H_t^\beta(\theta^\star)}\right)^{\frac{1}{2}} \exp\left(\frac{1}{2\sigma^2}\|S_t\|^2_{H_t^\beta(\theta^\star)^{-1}}\right). \tag{5.26}
$$

From there, the argument is standard and follows the framework of Theorem 1.20. In particular, the choice of $\tau = \inf\left\{t \in \mathbb{N},\ \|S_t\|^2_{H_t^\beta(\theta^\star)^{-1}} \geqslant \sigma^2\left(2\log\frac{1}{\delta} + \log\frac{\det H_t^\beta(\theta^\star)}{\det \beta I_d}\right)\right\}$ and a straightforward application of Markov's inequality reveals that

$$
\mathbb{P}(\tau < \infty) = \mathbb{P}\left(\exists t \in \mathbb{N},\ M_t \geqslant \frac{1}{\delta}\right) \leqslant \mathbb{E}[M_\tau]\delta \leqslant \delta, \tag{5.27}
$$

which is exactly the expected result. ■

**Discussion on Lemma 5.11 and Assumption 5.10.** As shown in the proof, Lemma 5.11 alone implies Proposition 5.12. While this lemma may be valid in more general settings, it is conveniently implied by Assumption 5.8 and the sub-Gaussian control of Assumption 5.10. In the rest of the paper, in particular in the regret bounds of Theorem 5.18, 5.23 and 5.26, $\sigma$ will refer to the parameter that appears in the supermartingale control of Lemma 5.11.

Regarding Assumption 5.10, note that for a mean-linear bandit $Y_t = \langle\theta^\star, X_t\rangle + \eta_t$ with adapted loss $\mathcal{L}(y,\xi) = \frac{1}{2}(y-\xi)^2$, we have that $\partial\mathcal{L}_t^\star = \eta_t$, which is classically assumed to be sub-Gaussian. For other bandits, and thus other adapted losses, it may be more convenient to make assumptions on the distribution of observable quantities such as $X_t$ and $Y_t$ rather than directly on $\partial\mathcal{L}_t^\star$. Formally, this raises the question of how the sub-Gaussian property of a random variable $Z$ transfers to $f(Z)$ for a given mapping $f$. While to our knowledge no complete answer is available, several partial results are available in the concentration literature.

(i) If $Z$ is Gaussian with variance $\sigma^2$ and $f$ is $M$-Lipschitz, the Tsirelson-Ibragimov-Sudakov inequality (Boucheron et al., 2013, Theorem 5.5) shows that $f(Z)$ is $M\sigma$-sub-Gaussian. In particular, the Lipschitz assumption holds for $\partial\mathcal{L}$ if the loss curvature is bounded from above by $M$. More generally, if $Z$ can be written as a $\sigma$-Lipschitz function of a $\mathcal{N}(0,1)$, then $f(Z)$ is $M\sigma$-sub-Gaussian.

(ii) If the density of $Z$ is strongly logconcave, then $f(Z)$ is sub-Gaussian (with parameter related to the largest eigenvalue of the Hessian of the logdensity, see Vershynin (2018, Theorem 5.2.15)).

(iii) If $Z$ is bounded (i.e. actions and rewards are bounded) and $f$ is Lipschitz and separately convex, then $f(Z)$ is sub-Gaussian (application of the entropy method, see e.g. Boucheron et al. (2013, Theorem 6.10)). The boundedness assumption can be lifted at the cost of a slightly more stringent condition than the sub-Gaussianity of $Z$, see Adamczak (2005, Theorem 3).

In short, Assumption 5.10 holds in a variety of settings, under rather mild assumptions on either $X_t$ and $Y_t$ or the loss $\mathcal{L}$. Also of note, $\partial^1\mathcal{L}$ is $M$-Lipschitz under Assumption 5.8.

**Confidence sequence for $\theta^\star$.** To help write the above confidence set in terms of $\theta^\star$ and the empirical estimator $\widehat{\theta}_t$, we introduce the function

$$\begin{aligned} F_t^\alpha \colon \Theta &\longrightarrow \mathbb{R}^d \\ \theta &\longmapsto \sum_{s=1}^{t-1} \partial\mathcal{L}\left(Y_s, \langle\theta, X_s\rangle\right) X_s + \alpha\theta\,, \end{aligned} \tag{5.28}$$

As seen above, $F_t^\alpha(\widehat{\theta}_t) = 0$ and $F_t^\alpha(\theta^\star) = S_t + \alpha\theta^\star$. Noticing that

$$\|F_t^\alpha(\theta^\star) - F_t^\alpha(\widehat{\theta}_t)\|^2_{H_t^\beta(\theta^\star)^{-1}} = \|S_t + \alpha\theta^\star\|_{H_t^\beta(\theta^\star)^{-1}} \leqslant \|S_t\|_{H_t^\beta(\theta^\star)^{-1}} + \alpha\|\theta^\star\|_{H_t^\beta(\theta^\star)^{-1}}\,, \tag{5.29}$$

we immediately derive the following result (note the use of a priori different regularisation parameters $\alpha$ and $\beta$, which we exploit later in Lemma 5.15).

**Corollary 5.13** (Time-uniform confidence sequence for $\theta^\star$). *Let $t \in \mathbb{N}$, $\delta \in (0,1)$, $\alpha, \beta > 0$, and define the set*

$$\widehat{\Theta}_t^\delta = \left\{ \theta \in \Theta,\ \|F_t^\alpha(\theta) - F_t^\alpha(\widehat{\theta}_t)\|_{H_t^\beta(\theta)^{-1}} \leqslant \sigma \sqrt{2\log\frac{1}{\delta} + \log\frac{\det H_t^\beta(\theta)}{\det \beta I_d}} + \alpha\|\theta\|_{H_t^\beta(\theta)^{-1}} \right\}. \tag{5.30}$$

*Then under Assumptions 5.8 and 5.10, $\left(\widehat{\Theta}_t^\delta\right)_{t\in\mathbb{N}}$ is a* ***time-uniform confidence sequence*** *at level $\delta$ for $\theta^\star$, i.e.*

$$\mathbb{P}\left(\forall t \in \mathbb{N},\ \theta^\star \in \widehat{\Theta}_t^\delta\right) \geqslant 1 - \delta\,, \tag{5.31}$$

*or equivalently, for any random time $\tau$ in $\mathbb{N}$,*

$$\mathbb{P}\left(\theta \in \widehat{\Theta}_\tau^\delta\right) \geqslant 1 - \delta\,. \tag{5.32}$$

We constantly use this result in the following, in particular to construct the projection operator $\Pi$. Indeed, we define

$$\bar{\theta}_t = \Pi(\widehat{\theta}_t) = \operatorname*{argmin}_{\theta\in\Theta} \|F_t^\alpha(\theta) - F_t^\alpha(\widehat{\theta}_t)\|_{H_t^\beta(\theta)^{-1}}\,, \tag{5.33}$$

such that, in particular, we have the property that $\Pi(\widehat{\theta}_t) \in \widehat{\Theta}_t^\delta$ with high probability.

**Remark 5.14**. *Although we formulated the bounded curvature condition (Assumption 5.8) globally, we note that we only require it to hold in a convex neighborhood of $\theta^\star$ containing $\bar{\theta}_t$, and Corollary 5.13 shows that with high probability, $\|\theta^\star - \bar{\theta}_t\|_2$ is bounded (going from the $H_t^\beta(\theta^\star)^{-1}$ norm to the Euclidean norm can be done by simple positive definite matrix inequalities). Therefore, one could instead assume a local curvature control on $\partial\mathcal{L}(y, \langle\theta, x\rangle)$ for $x \in \mathcal{X}_t$ and $\theta$ in a ball around $\theta^\star$, in the same spirit as Assumption 1 in Li et al. (2017) for GLB.*

### Optimism and local metrics

We recall here the principle of optimism in the face of uncertainty and adapt it to the framework of elicitable risk measures. We denote by $r_t = \langle \theta^\star, X_t^\star \rangle - \langle \theta^\star, X_t \rangle$ the instantaneous pseudo regret, where $\langle \theta^\star, X_t^\star \rangle = \max_{x \in \mathcal{X}_t} \langle \theta^\star, x \rangle$ is the optimal risk measure associated with $\mathcal{L}$ at time for the actions available at time $t$. Then, simple algebra shows that

$$\begin{aligned} r_t &= \langle \theta^\star - \bar{\theta}_t, X_t^\star \rangle - \langle \theta^\star - \bar{\theta}_t, X_t \rangle + \langle \bar{\theta}_t, X_t^\star - X_t \rangle \\ &= \Delta(X_t^\star, \bar{\theta}_t) + \Delta(X_t, \bar{\theta}_t) + \langle \bar{\theta}_t, X_t^\star - X_t \rangle \,, \end{aligned} \tag{5.34}$$

where we define for $x \in \mathcal{X}$ and $\theta \in \Theta$, $\Delta(x, \theta) = |\langle \theta^\star - \theta, x \rangle|$ the absolute error made by $\theta$ with respect to the true parameter of the linear bandit $\theta^\star$ in the direction of $x$. If we know a sequence of functions $\gamma_t \colon \mathcal{X} \to \mathbb{R}_+$ such that with high probability, for all $t \in \mathbb{N}$ and $x \in \mathcal{X}_t$, $\Delta(x, \bar{\theta}_t) \leqslant \gamma_t(x)$, then the principle of optimism recommends the action $X_t \in \operatorname{argmax}_{x \in \mathcal{X}_t} \langle \bar{\theta}_t, x \rangle + \gamma_t(x)$, i.e. the one leading to the best plausible reward with respect to the confidence on the prediction error of $\bar{\theta}_t$. In this case, $r_t \leqslant \Delta(X_t^\star, \bar{\theta}_t) + \Delta(X_t, \bar{\theta}_t) + \gamma_t(X_t) - \gamma(X_t^\star) \leqslant 2\gamma_t(X_t)$ with high probability, and hence $\mathcal{R}_T \leqslant 2 \sum_{t=1}^T \gamma_t(X_t)$. We detail below how Corollary 5.13 coupled with standard assumptions provides such a bound.

**Bound on the prediction error.** We follow the standard strategy of decoupling the dependency on $\bar{\theta}_t$ and $x$ in $\Delta(x, \bar{\theta}_t)$. By the Cauchy-Schwarz inequality, we have, for some positive definite matrix $P$ to be determined later,

$$\Delta(x, \bar{\theta}_t) = |\langle P^{\frac{1}{2}}(\theta^\star - \bar{\theta}_t), P^{-\frac{1}{2}} x \rangle| \leqslant \|\theta^\star - \bar{\theta}_t\|_P \|x\|_{P^{-1}} \,. \tag{5.35}$$

As we see below, a natural choice for $P$ is the (average) Hessian of the empirical risk minimisation problem, and therefore the term $\|x\|_{P^{-1}}$ can be handled by the elliptical potential lemma (Lemma 11 in Abbasi-Yadkori et al. (2011)). To control the remainder term in $\theta^\star - \bar{\theta}_t$, we borrow technical tools from the classical approach developed for generalised linear bandits (Filippi et al., 2010; Faury et al., 2020) and note that

$$F_t^\alpha(\theta^\star) - F_t^\alpha(\bar{\theta}_t) = \bar{H}_t^\alpha(\theta^\star, \bar{\theta}_t)(\theta^\star - \bar{\theta}_t) \,, \tag{5.36}$$

where $\bar{H}_t^\alpha(\theta^\star, \bar{\theta}_t) = \int_0^1 H_t^\alpha(u\theta^\star + (1-u)\bar{\theta}_t) du$ is the average of the Hessian matrices along the segment $[\bar{\theta}_t, \theta^\star]$[1] (this follows from the observation that the differential of $F_t^\alpha$ is $H_t^\alpha$). Therefore the choice $P = \bar{H}_t^\alpha(\theta^\star, \bar{\theta}_t)$ yields

$$\|\theta^\star - \bar{\theta}_t\|_P = \|F_t^\alpha(\theta^\star) - F_t^\alpha(\bar{\theta}_t)\|_{\bar{H}_t^\alpha(\theta^\star, \bar{\theta}_t)^{-1}}$$

[1] By convexity of $\Theta$, the segment $[\bar{\theta}_t, \theta^\star]$ lies in $\Theta$. However, we could also use average the Hessian matrices along a curve, i.e. $\bar{H}_t^\alpha(\theta^\star, \bar{\theta}_t) = \int_0^1 H_t^\alpha(\gamma_u) du$ where $\gamma \colon [0,1] \to \Theta$ is a smooth, unit speed path connecting $\bar{\theta}_t$ and $\theta^\star$.

$$\leqslant \|F_t^\alpha(\theta^\star) - F_t^\alpha(\widehat{\theta}_t)\|_{\bar{H}_t^\alpha(\theta^\star,\bar{\theta}_t)^{-1}} + \|F_t^\alpha(\bar{\theta}_t) - F_t^\alpha(\widehat{\theta}_t)\|_{\bar{H}_t^\alpha(\theta^\star,\bar{\theta}_t)^{-1}} . \tag{5.37}$$

To conclude, we need to find a way to relate the local metric defined by $\bar{H}_t^\alpha(\theta^\star, \bar{\theta}_t)^{-1}$ to those defined by $H_t^\beta(\theta^\star)^{-1}$ and $H_t^\beta(\bar{\theta}_t)^{-1}$, for which we have high confidence bounds. This motivates the following assumption.

**Lemma 5.15** (Transportation of Local Metrics). *Under Assumption 5.8, for $\alpha > 0$, there exists $\kappa > 0, \beta > 0$ such that*

$$\bar{H}_t^\alpha(\theta^\star, \bar{\theta}_t) \succcurlyeq \frac{1}{\kappa} H_t^\beta(\theta^\star) \quad \textit{and} \quad \bar{H}_t^\alpha(\theta^\star, \bar{\theta}_t) \succcurlyeq \frac{1}{\kappa} H_t^\beta(\bar{\theta}_t) . \tag{5.38}$$

*Proof of Lemma 5.15.* The result follows from simple calculations and the bounds provided by Assumption 5.8:

$$\begin{aligned}
\bar{H}_t^\alpha(\theta^\star, \bar{\theta}_t) &= \sum_{s=1}^{t-1} \int_0^1 \partial^2 \mathcal{L}(Y_s, \langle u\theta^\star + (1-u)\bar{\theta}_t, X_s\rangle) du X_s X_s^\top + \alpha I_d \\
&= \sum_{s=1}^{t-1} \int_0^1 \partial^2 \mathcal{L}(Y_s, \langle \theta^\star, X_s\rangle) \frac{\partial^2 \mathcal{L}(Y_s, \langle u\theta^\star + (1-u)\bar{\theta}_t, X_s\rangle)}{\partial^2 \mathcal{L}(Y_s, \langle \theta^\star, X_s\rangle)} du X_s X_s^\top + \alpha I_d \\
&\succcurlyeq \frac{m}{M} \sum_{s=1}^{t-1} \partial^2 \mathcal{L}(Y_s, \langle \theta^\star, X_s\rangle) X_s X_s^\top + \alpha I_d \qquad \text{(Assumption 5.8)} \\
&= \frac{1}{\kappa} \left( \sum_{s=1}^{t-1} \partial^2 \mathcal{L}(Y_s, \langle \theta^\star, X_s\rangle) X_s X_s^\top + \kappa\alpha I_d \right) \\
&= \frac{1}{\kappa} H_t^{\kappa\alpha}(\theta^\star) ,
\end{aligned} \tag{5.39}$$

which is the desired result with $\beta = \kappa\alpha$ and $\kappa = \frac{M}{m}$ the conditioning of the loss $\mathcal{L}$. The other inequality with $\bar{H}_t^\beta(\bar{\theta}_t)$ is derived similarly. ■

Again, we keep the formulation fairly generic as Lemma 5.15 may hold beyond losses with bounded curvature. For instance, in a special case of GLB, namely the logistic bandit, it is shown in Faury et al. (2020) that this lemma holds thanks to self-concordance properties of the sigmoid link function.

**Examples.** We conclude this section by discussing examples of standard losses and whether they satisfy the above conditions.

- **Expectiles.** The expectile loss is derived from the strongly convex potential $\psi_2(z) = |p - \mathbb{1}_{z<0}|z^2$, the second derivative of which is $\psi_2''(z) = 2|p - \mathbb{1}_{z<0}|$. Thus, Assumption 5.8 holds with $m = 2\min(p, 1-p)$ and $M = 2\max(p, 1-p)$ (see Section 5.4 for more details).

- **Quantiles.** The quantile loss is derived from the potential $\psi_1(z) = (p - \mathbb{1}_{z<0})z$, which is piecewise linear. In particular, it is not strongly convex and thus does not satisfy Assumption 5.8, i.e. bandits with quantile regression are outside the scope of this work.

### Regret analysis

We make two additional standard assumptions that prior bounds are known on $\theta^\star$ and on the actions $\mathcal{X} = \bigcup_{t\in\mathbb{N}} \mathcal{X}_t$, which is standard in the existing literature on linear bandits.

**Assumption 5.16** (Prior bound on parameters). *All parameters are bounded by a constant $S > 0$, i.e. $\Theta \subseteq \mathbb{B}^d_{\|\cdot\|_2}(0, S)$. In particular, this implies that $\|\theta^\star\|_{H_t^\beta(\theta^\star)^{-1}} \leqslant \frac{S}{\sqrt{\beta}}$ for any $\beta > 0$.*

**Assumption 5.17** (Prior bound on actions). *All actions are bounded by a constant $L > 0$, i.e. $\mathcal{X} \subseteq \mathbb{B}^d_{\|\cdot\|_2}(0, L)$.*

We now obtain a high probability upper bound on the pseudo regret incurred by Algorithm 3 for an explicit choice of exploration bonus sequence $(\gamma_t)_{t\in\mathbb{N}}$ and projection $\Pi$. As is standard for contextual bandits with possibly infinite action sets, this bound is *minimax* (worst case) as it does not depend explicitly on the optimality gaps $\langle\theta^\star, X_t^\star\rangle - \langle\theta^\star, X_t\rangle$.

**Theorem 5.18** (Pseudo regret upper bound for LinUCB-CR - 1). *Let $\delta \in (0,1)$, $\alpha \geqslant \max(1, L^2)$ and define for $t \in \mathbb{N}$ the exploration bonus*

$$\begin{aligned} \gamma_t \colon \mathcal{X}_t &\longrightarrow \mathbb{R}_+ \\ x &\longmapsto c_t^\delta \|x\|_{H_t^{\kappa\alpha}(\bar{\theta}_t)^{-1}} \,, \end{aligned} \tag{5.40}$$

$$c_t^\delta = 2\kappa \left( \sigma \sqrt{2 \log \frac{2}{\delta} + d \log \frac{m}{\alpha} + \log \det V_t^{\frac{\alpha}{m}}} + \sqrt{\frac{\alpha}{\kappa}} S \right) \tag{5.41}$$

*and the projection operator*

$$\begin{aligned} \Pi \colon \mathbb{R}^d &\longrightarrow \Theta \\ \theta &\longmapsto \underset{\theta' \in \Theta}{\operatorname{argmin}} \|F_t^\alpha(\theta') - F_t^\alpha(\theta)\|_{H_t^{\kappa\alpha}(\theta')^{-1}} \,. \end{aligned} \tag{5.42}$$

*Under Assumptions 5.8, 5.10, 5.16 and 5.17, with probability at least $1-\delta$, for all $T \in \mathbb{N}$, the pseudo regret of Algorithm 3 is upper bounded by*

$$\mathcal{R}_T \leqslant 2 c_T^\delta \max\left( \frac{1}{\sqrt{m}}, \frac{L}{\sqrt{\kappa\alpha}} \right) \sqrt{2Td \log\left(1 + \frac{mTL^2}{d\kappa\alpha}\right)} \,, \tag{5.43}$$

*and in particular, we have*

$$\mathcal{R}_T = \mathcal{O}\left( \frac{\kappa\sigma d}{\sqrt{m}} \sqrt{T} \log \frac{TL^2}{d} \right) . \tag{5.44}$$

*Proof of Theorem 5.18.* The proof of this result follows the standard regret analysis of LinUCB, up to the modifications detailed in the previous sections. First, we justify the choice of exploration sequence $(\gamma_t)_{t\in\mathbb{N}}$, which naturally derives from the optimistic principle and the analysis of local metrics. Then, we use a somewhat crude bound on the Hessian to simplify the analysis and reduce it to the so-called elliptic potential lemma.

Indeed, as established in Section 5.2, with probability at least $1-\delta$, the cumulative pseudo regret $\mathcal{R}_T$ is upper bounded for all $T \in \mathbb{N}$ by $2\sum_{t=1}^T \gamma_t(X_t)$ provided that

$$\mathbb{P}\left( \forall t \in \mathbb{N},\ \Delta(X_t, \bar{\theta}_t) \leqslant \gamma_t(X_t) \right) \geqslant 1 - \delta \,, \tag{5.45}$$

$$\text{where} \quad \Delta(X_t, \theta) = |\langle \theta^\star - \theta, X_t \rangle| \leqslant \|\theta^\star - \bar{\theta}_t\|_{\bar{H}_t^\alpha(\theta^\star, \bar{\theta}_t)} \|X_t\|_{\bar{H}_t^\alpha(\theta^\star, \bar{\theta}_t)^{-1}} \,. \tag{5.46}$$

**Tuning of the Exploration Bonus Sequence.** The transportation of local metrics (Lemma 5.15, implied by the curvature bound of Assumption 5.8) reveals that

$$\begin{aligned}\|\theta^\star - \bar\theta_t\|_{\bar H_t^\alpha(\theta^\star,\bar\theta_t)} &\leqslant \|F_t^\alpha(\theta^\star) - F_t^\alpha(\widehat\theta_t)\|_{\bar H_t^\alpha(\theta^\star,\bar\theta_t)^{-1}} + \|F_t^\alpha(\bar\theta_t) - F_t^\alpha(\widehat\theta)\|_{\bar H_t^\alpha(\theta^\star,\bar\theta_t)^{-1}} \\ &\leqslant \sqrt{\kappa}\left(\|F_t^\alpha(\theta^\star) - F_t^\alpha(\widehat\theta_t)\|_{H_t^\beta(\theta^\star)^{-1}} + \|F_t^\alpha(\bar\theta_t) - F_t^\alpha(\widehat\theta)\|_{H_t^\beta(\bar\theta_t)^{-1}}\right). \end{aligned} \tag{5.47}$$

Thanks to the supermartingale control of Lemma 5.11, we deduce from Corollary 5.13 that with probability at least $1-\delta$, the following inequalities hold for all $t \in \mathbb{N}$:

$$\|F_t^\alpha(\theta^\star) - F_t^\alpha(\widehat\theta_t)\|_{H_t^\beta(\theta^\star)^{-1}} \leqslant \sigma\sqrt{2\log\frac{1}{\delta} + \log\frac{\det H_t^\beta(\theta^\star)}{\det \beta I_d}} + \alpha\|\theta^\star\|_{H_t^\beta(\theta^\star)^{-1}}, \tag{5.48}$$

$$\|F_t^\alpha(\bar\theta_t) - F_t^\alpha(\widehat\theta_t)\|_{H_t^\beta(\bar\theta_t)^{-1}} \leqslant \sigma\sqrt{2\log\frac{1}{\delta} + \log\frac{\det H_t^\beta(\bar\theta_t)}{\det \beta I_d}} + \alpha\|\bar\theta_t\|_{H_t^\beta(\bar\theta_t)^{-1}}. \tag{5.49}$$

The prior bound on parameters (Assumption 5.16) implies that $\|\theta\|_{H_t^\beta(\theta)^{-1}} \leqslant S/\sqrt{\beta}$ for $\theta \in \{\theta^\star, \bar\theta_t\}$. Furthermore, the curvature bound (Assumption 5.8) implies that $H_t^\beta(\theta) \preccurlyeq M V_t^{\beta/M}$, and therefore $\det H_t^\beta(\theta) \leqslant M^d \det V_t^{\beta/M}$ for $\theta \in \{\theta^\star, \bar\theta_t\}$. Combining these together with a time-uniform union bound over the two events of equations 5.48 and 5.49, and substituting the expression of $\beta = \kappa\alpha$, where $\kappa = \frac{M}{m}$ is the conditioning of the convex loss $\mathcal{L}$, we obtain:

$$\|\theta^\star - \bar\theta_t\|_{\bar H_t^\alpha(\theta^\star,\bar\theta_t)} \leqslant 2\sqrt{\kappa}\left(\sigma\sqrt{2\log\frac{2}{\delta} + d\log\frac{m}{\alpha} + \log\det V_t^{\frac{\alpha}{m}}} + \sqrt{\frac{\alpha}{\kappa}}S\right). \tag{5.50}$$

By the same arguments, it holds that $\bar H_t^\alpha(\theta^\star, \bar\theta_t)^{-1} \preccurlyeq \kappa H_t^{\kappa\alpha}(\bar\theta_t)^{-1}$ and therefore

$$\|X_t\|_{\bar H_t^\alpha(\theta^\star,\bar\theta_t)^{-1}} \leqslant \sqrt{\kappa}\|X_t\|_{H_t^{\kappa\alpha}(\bar\theta_t)^{-1}}. \tag{5.51}$$

This shows that

$$\begin{aligned}\gamma_t\colon \mathcal{X}_t &\longrightarrow \mathbb{R}_+ \\ x &\longmapsto \underbrace{2\kappa\left(\sigma\sqrt{2\log\frac{2}{\delta} + d\log\frac{m}{\alpha} + \log\det V_t^{\frac{\alpha}{m}}} + \sqrt{\frac{\alpha}{\kappa}}S\right)}_{c_t^\delta}\|x\|_{H_t^{\kappa\alpha}(\bar\theta_t)^{-1}}\end{aligned} \tag{5.52}$$

is a valid choice of exploration sequence.

**Bounding the pseudo regret.** Going back to the pseudo regret $\mathcal{R}_T$, we notice that $(c_t^\delta)_{t=1}^T$ is a positive, nondecreasing sequence. Hence, with probability at least $1-\delta$, for all $T \in \mathbb{N}$, we have

$$\mathcal{R}_T \leqslant 2\sum_{t=1}^T \gamma_t(X_t) \leqslant 2c_T^\delta \sum_{t=1}^T \|X_t\|_{H_t^{\kappa\alpha}(\bar{\theta}_t)^{-1}} . \tag{5.53}$$

A priori, the direct analysis of the right-hand side is tedious due to the dependency on $\bar{\theta}_t$ in the local metric. However, we notice that the curvature bound (Assumption 5.8) also implies the weaker control $H_t^{\kappa\alpha}(\bar{\theta}_t)^{-1} \preccurlyeq \frac{1}{m}(V_t^{\frac{\kappa\alpha}{m}})^{-1}$, which translates to

$$\|X_t\|_{H_t^{\kappa\alpha}(\bar{\theta}_t)^{-1}} \leqslant \frac{1}{\sqrt{m}}\|X_t\|_{(V_t^{\frac{\kappa\alpha}{m}})^{-1}} . \tag{5.54}$$

This bound is less informative as it looses the local information carried by $\bar{\theta}_t$, but still sufficient to obtain sublinear pseudo regret growth. We recall the following result, which is a direct consequence of the deterministic elliptic potential lemma (Lemma 11, Abbasi-Yadkori et al. (2011)) and the Cauchy-Schwarz inequality.

**Lemma 5.19** (Deterministic elliptic potential). *Let $(x_t)_{t\in\mathbb{N}}$ be an arbitrary sequence of vectors in $\mathbb{B}^d_{\|\cdot\|}(0, L)$, $\gamma > 0$ and $v_t = \sum_{s=1}^{t-1} x_s x_s^\top + \gamma I_d \in \mathcal{S}_d^{++}(\mathbb{R})$ for $t \in \mathbb{N}$. Then*

$$\sum_{s=1}^t \|x_s\|_{v_s^{-1}} \leqslant \max(1, \frac{L}{\sqrt{\gamma}})\sqrt{2td\log\left(1+\frac{tL^2}{d\gamma}\right)} . \tag{5.55}$$

Note that this result holds in our case (with $\gamma = \frac{\kappa\alpha}{m}$) thanks to the prior bound on actions (Assumption 5.17).

**Conclusion of the proof.** With high probability, uniformly in $T \in \mathbb{N}$, the pseudo regret of LinUCB-CR is upper bounded by

$$\mathcal{R}_T \leqslant 2\sum_{t=1}^T \gamma_t(X_t) \leqslant 2c_T^\delta \max\left(\frac{1}{\sqrt{m}}, \frac{L}{\sqrt{\kappa\alpha}}\right)\sqrt{2Td\log\left(1+\frac{mTL^2}{d\kappa\alpha}\right)} . \tag{5.56}$$

Going back to the expression of $c_T^\delta$, it follows from simple algebra (see e.g. Lattimore and Szepesvári (2020, proof of Lemma 19.4)) that $\det V_t^{\frac{\alpha}{m}} \leqslant \left(\frac{\alpha}{m} + \frac{TL^2}{d}\right)^d$, and thus we have the asymptotic control $c_T^\delta = \mathcal{O}\left(\kappa\sigma\sqrt{d\log\frac{TL^2}{d}}\right)$ when $T \to +\infty$, which implies

$$\mathcal{R}_T = \mathcal{O}\left(\frac{\kappa\sigma d}{\sqrt{m}}\sqrt{T}\log\frac{TL^2}{d}\right). \tag{5.57}$$

■

**Remark 5.20**. *This regret bound scales with $\kappa m^{-1/2}$, where $m$ is the minimum curvature of the loss $\mathcal{L}$ and $\kappa$ the coefficient of transportation of local metrics. Under Assumption 5.8, this scales as $m^{-3/2}$. In the limit of flattening loss $m \to 0$, learning with this strategy becomes impossible. We show in Appendix D.2 that a small modification of the exploration sequence reduces this dependency to $\kappa^{1/2}m^{-1/2}$, at the cost of loosing local information carried by $H_t^{\kappa\alpha}(\bar{\theta}_t)$. An analogous dependency on $m^{-1}$ was observed for GLB in Filippi et al. (2010). In the special case of logistic bandit, Faury et al. (2020) obtained a $\kappa$ independent of $m$ using self-concordance, and even pushed the dependency on $m^{-1/2}$ to higher order terms in $T$ using a more intricate algorithmic design. We conjecture that a similar construction could apply here but leave this open for future work.*

**Remark 5.21**. *In the mean-linear case, $m = M = \kappa = 1$ and $H_t^\alpha = V_t^\alpha$, thus this result matches the standard LinUCB upper bound (Abbasi-Yadkori et al., 2011) and is compatible with the minimax lower bound $\mathcal{O}(d\sqrt{T})$ for actions in $\mathbb{B}^d_{\|\cdot\|_2}(0,1)$ (Theorem 1.10). In Section 5.4, we show that the lower bound also applies to expectile bandit models (Corollary 5.29).*

### Regret analysis under stochactic action sets

Theorem 5.18 holds for arbitrary (potentially adversarial) sequence of action sets $(\mathcal{X}_t)_{t\in\mathbb{N}}$. If these are instead stochastically generated, the regret bound can be further tightened.

**Assumption 5.22** (Stochastic action sets). *Let $\nu_{\mathcal{X}} \in \mathcal{M}_1^+\left(\mathfrak{P}\left(\mathbb{B}^d_{\|\cdot\|_2}(0, L)\right)\right)$, i.e. $\nu_{\mathcal{X}}$ is a probability measure such that samples drawn from it are sets of vectors of norm at most $L$.*

(*i*) *For $t \in \mathbb{N}$, $\mathcal{X}_t \sim \nu_{\mathcal{X}}$ defines an i.i.d. sequence of random action sets.*

(*ii*) *Recall that $X_t \in \mathcal{X}_t$ denotes the action selected by the agent at time $t \in \mathbb{N}$. Then $\mathbb{E}\left[X_t X_t^\top \mid \bar{\mathcal{G}}_{t-1}\right] \succcurlyeq \rho_{\mathcal{X}} L^2 I_d > 0$.*

The lower bound on conditional covariance of actions of Assumption 5.22 is new, although related to more standard settings. In the case of finite action sets $\mathcal{X}_t = \{X_{k,t}, k \in [K]\}$, Li et al. (2017); Kim et al. (2023) considered a lower bound on the unconditional average covariance across arms $\mathbb{E}[1/K \sum_{k\in[K]} X_{k,t} X_{k,t}^\top]$. We argue that this assumption is quite mild in the sense that for non-degenerate $\nu_{\mathcal{X}}$, the conditioning is essentially irrelevant. At time $t$, $\mathcal{X}_t$ is drawn independently of $\bar{\mathcal{G}}_{t-1}$, and $X_t \in \mathcal{X}_t$ is selected in a $\bar{\mathcal{G}}_{t-1}$-measurable fashion. To violate the covariance inequality, there should exist a fixed strict subspace $\mathcal{V} \subset \mathbb{R}^d$ such that with some probability $\mathcal{X}_t \cap \mathcal{V} \neq \emptyset$ (when randomising over the action set $\mathcal{X}_t$) and $X_t$ should be one of the vectors in $\mathcal{V}$; however, if e.g. $\nu_{\mathcal{X}}$ spans an open set, this almost surely cannot happen.

The following result shows that under this condition of diversity of the action sets, the pseudo regret guarantee of optimistic strategies can be slightly improved by a factor $\mathcal{O}(\sqrt{\log T})$.

**Theorem 5.23** (Pseudo regret upper bound for LinUCB-CR - 2). *Let $\delta \in (0,1)$, $\alpha > 0$ and define $t_0 = \lceil \frac{8}{\rho_{\mathcal{X}}^2} \log \frac{2}{\delta} - \frac{2\kappa\alpha}{m\rho_{\mathcal{X}} L^2} \rceil$. Under Assumptions 5.8, 5.10, 5.16, 5.17 and 5.22, for $T \geqslant t_0$, with probability at least $1 - 2\delta$, the pseudo regret of Algorithm 3 is bounded by*

$$\mathcal{R}_T \leqslant 4c_T^\delta \sqrt{\frac{2T}{m\rho_{\mathcal{X}}}} \left(1 + \frac{C}{\sqrt{T}}\right), \tag{5.58}$$

*where*

$$C = \frac{1}{L^2}\sqrt{\frac{\kappa\alpha - 4\frac{mL^2}{\rho_{\mathcal{X}}}\log\frac{2}{\delta}}{2m\rho_{\mathcal{X}}}} - \frac{1}{2L}\sqrt{t_0 - 1 + \frac{2(\kappa\alpha - \frac{4mL^2}{\rho_{\mathcal{X}}}\log\frac{2}{\delta})}{m\rho_{\mathcal{X}}L^2}} + \frac{1}{2}\max\left(1, L\sqrt{\frac{m}{\kappa\alpha}}\right)\sqrt{\rho_{\mathcal{X}} d t_0 \log\left(1 + \frac{mL^2 t_0}{d\kappa\alpha}\right)}. \tag{5.59}$$

*In particular, we have*

$$\mathcal{R}_T = \mathcal{O}\left(\kappa\sigma\sqrt{\frac{dT}{m\rho_{\mathcal{X}}}\log\frac{TL^2}{d}}\right). \tag{5.60}$$

The reduced logarithmic dependency in $T$ cannot be significantly improved further. Indeed, Li et al. (2019) proved a $\Omega(\sqrt{dT\log(K)\log(T/d)})$ lower bound on the expected pseudo regret, although in the context of action sets containing at most $K$ vectors. Note that Assumption 5.22 and $\rho_{\mathcal{X}}$ do not influence the design of Algorithm 1, only the $\mathcal{O}\left(\sqrt{\log T}\right)$ term in its regret bound. Compared to Kim et al. (2023), our proof relies on line crossing arguments developed in Howard et al. (2020) rather than on a crude union bound, leading to improved constants and higher order terms (even in the mean-linear case), and preserves the time-uniform property of the regret bound. We defer these technical details and the proof of Theorem 5.23 to Appendix D.2.

**Remark 5.24** (Dependency of $\rho_{\mathcal{X}}$ on $d$). *Following the argument in Kim et al. (2021) in the case of unconditional covariance control, we obtain the following bound on $\rho_{\mathcal{X}}$:*

$$\begin{aligned} d\rho_{\mathcal{X}} L^2 &\leqslant d\lambda_{\min}\left(\mathbb{E}\left[X_t X_t^\top \mid \bar{\mathcal{G}}_{t-1}\right]\right) \\ &\leqslant \sum_{\lambda \in \mathcal{S}_t} \lambda = \mathrm{Tr}\left(\mathbb{E}\left[X_t X_t^\top \mid \bar{\mathcal{G}}_{t-1}\right]\right) = \mathbb{E}\left[\mathrm{Tr}\left(X_t X_t^\top\right) \mid \bar{\mathcal{G}}_{t-1}\right] \leqslant L^2 , \end{aligned} \tag{5.61}$$

*where $\mathcal{S}_t$ denotes the spectrum of the symmetric matrix $\mathbb{E}\left[X_t X_t^\top \mid \bar{\mathcal{G}}_{t-1}\right]$. Therefore $\rho_{\mathcal{X}}^{-1} \geqslant d$. Moreover, Bastani et al. (2021); Kim et al. (2021, 2023) identified two families of examples where $\rho_{\mathcal{X}} = \mathcal{O}(d^{-1})$ in the unconditional case:*

(*i*) *If the distribution of $X \in \mathcal{X}_t$ (marginal distribution of a each action) admits a density $p$ with respect to the Lebesgue measure supported in $\mathbb{B}^d_{\|\cdot\|_2}(0, L)$ and such that $p(x) \geqslant p_{\min} > 0$ for all $x \in \mathbb{B}^d_{\|\cdot\|_2}(0, L)$, then $\rho_{\mathcal{X}} = \frac{p_{\min}}{(d+2)}\mathrm{vol}\left(\mathbb{B}^d_{\|\cdot\|_2}(0,1)\right)$ is a suitable choice (Kim et al., 2023, Lemma C.1). In general, the volume of the Euclidean unit ball in $\mathbb{R}^d$ is $\mathrm{vol}\left(\mathbb{B}^d_{\|\cdot\|_2}(0,1)\right) = \frac{\pi^{d/2}}{\Gamma(\frac{d}{2}+1)} \sim \frac{1}{\sqrt{d\pi}}\left(\frac{2\pi e}{d}\right)^{\frac{d}{2}}$, which goes to $0$ when $d \to +\infty$. In certain cases though, such as the uniform and truncated Gaussian distributions, $p_{\min}$ is proportional to $\mathrm{vol}\left(\mathbb{B}^d_{\|\cdot\|_2}(0,1)\right)$, thus leading to $\rho_{\mathcal{X}} = \mathcal{O}(d^{-1})$.*

(*ii*) *If the covariance matrix $\mathbb{E}\left[XX^\top\right]$ exhibits a certain structure, for instance AR(1), tridiagonal or block diagonal, then $\rho_{\mathcal{X}} = \mathcal{O}(d^{-1})$, regardless of the marginal distributions.*

*In particular, the regret upper bound of Theorem 5.23 scales linearly with $d$, which is consistent with the lower bound discussed in Remark 5.21.*

## 5.3 Approximate strategy with online gradient descent

So far, we have shown that the standard LinUCB principle can be extended to the convex loss setting with similar pseudo regret guarantees under some curvature assumption. However, this comes at the cost of a significant computational overhead since the estimator $\widehat{\theta}_t$ needs to be calculated from scratch at each step as $\mathrm{argmin}_{\theta \in \mathbb{R}^d} \sum_{s=1}^{t-1} \mathcal{L}(Y_s, \langle \theta, X_s \rangle) + \frac{\alpha}{2}\|\theta\|_2^2$. As a reminder, in the standard mean-linear case, this estimator has an analytical expression that amounts to incrementally inverting the matrix $V_t^\alpha$, which can be done efficiently from the knowledge of the inverse of $V_{t-1}^\alpha$ via the Sherman-Morrison formula.

We propose an alternative algorithm that exploits online gradient descent (OGD) to compute a fast approximation of the empirical risk minimiser $\widehat{\theta}_t$. This may be of practical interest to deploy risk-aware linear bandits in time-sensitive environments, such as in real-time online recommendation systems. Moreover, it can also be relevant in the mean-linear setting with high dimensional action sets, where computing gradients may be more tractable than inverting a large $d \times d$ matrix. For $n, h \in \mathbb{N}$, we define the aggregated gradient of the empirical loss as

$$\begin{aligned} \nabla^{\alpha}_{n,h} \colon \mathbb{R}^d &\longrightarrow \mathbb{R}^d \\ \widehat{\theta} &\longmapsto \sum_{k=1}^{h} \partial\mathcal{L}(Y_{(n-1)h+k}, \langle \widehat{\theta}, X_{(n-1)h+k} \rangle) + \alpha\theta\,. \end{aligned} \tag{5.62}$$

**Algorithm 4** LinUCB-OGD-CR

**Input:** horizon $T$, regularisation parameter $\alpha$, projection operator $\Pi$, exploration bonus sequence $(\gamma^{\text{OGD}}_{t,T})_{t \leqslant T}$, step sequence $(\varepsilon_t)_{t\in\mathbb{N}}$, episode length $h > 0$.
**Initialisation:** Observe $\mathcal{X}_1$, set $\widehat{\theta}^{\text{OGD}}_0$, $n = 1$.
**for** $t = 1, \ldots, T$ **do**
  **if** $t = nh + 1$ **then**
    $\widehat{\theta}^{\text{OGD}}_n \leftarrow \widehat{\theta}^{\text{OGD}}_{n-1} - \varepsilon_{n-1}\nabla^{\alpha}_{n,h}(\widehat{\theta}^{\text{OGD}}_{n-1})$ ; ▷ `OGD`
    $\bar{\theta}^{\text{OGD}} \leftarrow \frac{1}{n}\sum_{j=1}^{n} \Pi(\widehat{\theta}^{\text{OGD}}_j)$ ; ▷ `Average`
    $n \leftarrow n + 1$ ;
  $X_t \leftarrow \operatorname{argmax}_{x\in\mathcal{X}_t} \langle \bar{\theta}^{\text{OGD}}, x \rangle + \gamma^{\text{OGD}}_{t,T}(x)$ ; ▷ `Play with same parameter for` $h$ `steps`
  Observe $Y_t$ and $\mathcal{X}_{t+1}$; ▷ `Observe reward and next action set`
  $t \leftarrow t + 1$ .

The intuition behind Algorithm 4 is that at time $t = nh+1$, the approximation error between the OGD estimate $\bar{\theta}^{\text{OGD}}_n$ and the exact minimiser of the empirical risk $\widehat{\theta}_t$ induces additional exploration, which translates to an increased regret compared to LinUCB. In other words, LinUCB-OGD trades off accuracy for computational efficiency. The episodic structure is borrowed from Ding et al. (2021) and is key to ensure sufficient convexity of the aggregate loss $\nabla^{\alpha}_{n,h}(\widehat{\theta})$. This allows to leverage the strong approximation guarantees of OGD, which we extend in the following proposition by relaxing the standard boundedness requirement of the gradient (Theorem 3.3, Hazan et al. (2016)) to a weaker sub-Gaussian control at a given parameter. We prove in Appendix D.3 an extension of the following proposition, with an explicit bound on the OGD regret, which we report below in the $\mathcal{O}$ notation for the sake of concision.

**Proposition 5.25** (OGD regret, sub-Gaussian gradients). *Let $\mathcal{C}$ a convex subset of $\mathbb{R}^d$ and $\Pi$ the projection operator onto $\mathcal{C}$. For $j = 1, \dots, N$, let $\ell_j : \mathcal{C} \longrightarrow \mathbb{R}_+$ a twice differentiable convex function and $a, A > 0$ such that $aI_d \preccurlyeq \nabla^2 \ell_j(z) \preccurlyeq AI_d$ for all $z \in \mathcal{C}$. Define the OGD update at step $j$ by $z_j = \Pi(z_{j-1} - \varepsilon_{j-1} \nabla \ell_j(z_{j-1}))$ and $\bar{z}_n = \arg\min_{z \in \mathcal{C}} \sum_{j=1}^n \ell_j(z)$. Assume that there exists $z^* \in \mathcal{C}$ such that $\nabla \ell_j(z^*) = g_j + \frac{\alpha}{n} z^*$ with $\alpha \geqslant 0$ and $g$ a centered, $\mathbb{R}^d$-valued $\sigma$-sub-Gaussian process, and also that $\mathcal{C}$ is bounded, i.e diam$(\mathcal{C}) = \sup_{z,z' \in \mathcal{C}} \|z - z'\| < \infty$. Then with probability at least $1-\delta$, the OGD regret with step size $\varepsilon_j = \frac{3}{aj}$ is bounded for all $n \leqslant N$ by*

$$\sum_{j=1}^{n} \ell_j(z_j) - \ell_j(\bar{z}_n) \leqslant \frac{9}{2a} \left( 2d\sigma^2 \log \frac{2dN}{\delta} + A^2 \mathit{diam}(\mathcal{C})^2 + \frac{\alpha^2}{n^2} \|z^*\|^2 \right) (1 + \log n) \ . \tag{5.63}$$

*In particular, we have*

$$\sum_{j=1}^{N} \ell_j(z_j) - \ell_j(\bar{z}_N) = \mathcal{O}\left( \frac{d\sigma^2}{a} \log^2 N \right) \tag{5.64}$$

*when $N \to +\infty$. In addition, if $g$ is uniformly bounded by a constant $G > 0$, the regret with step size $\varepsilon_s = \frac{1}{aj}$ can be reduced to the almost sure bound:*

$$\sum_{j=1}^{n} \ell_j(z_j) - \ell_j(\bar{z}_n) \leqslant \frac{G^2}{2a} (1 + \log n) \ . \tag{5.65}$$

Finally, we show that the approximation error of OGD induces at most a polylog correction in the pseudo regret bound of LinUCB-OGD-CR, which we prove in Appendix D.3.

**Theorem 5.26** (Pseudo regret of LinUCB-OGD-CR). *Let* $\varepsilon_h > 0$ *and* $h = \lceil \frac{2\varepsilon_h}{\rho_{\mathcal{X}} L^2} + \frac{8}{\rho_{\mathcal{X}}^2} \log \frac{2}{\delta} \rceil$. *Assume that* $\partial\mathcal{L}(Y_t, \langle \theta^\star, X_t \rangle)$ *is* $\sqrt{m}\sigma$*-sub-Gaussian for all* $t \leqslant T$. *Under Assumptions 5.8, 5.10, 5.16, 5.17 and 5.22, there exists* $C' > 0$ *such that with probability at least* $1 - (1 + T/h)\delta$ *the pseudo regret of Algorithm 4 with the projection operator of Theorem 5.18, the exploration bonus sequence*

$$\begin{aligned} \gamma_{t,T}^{OGD} : \mathcal{X}_t &\longrightarrow \mathbb{R}_+ \\ x &\longmapsto (c_t^\delta + c_{t,T}^{OGD,\delta}) \|x\|_{H_t^{\kappa\alpha}(\bar{\theta}^{OGD}_{\lfloor \frac{t-1}{h} \rfloor})^{-1}} , \end{aligned} \tag{5.66}$$

$$c_{t,T}^{OGD,\delta} = \sqrt{\left(L^2 + \frac{\alpha}{mMt}\right)\left(\frac{2\kappa C' dh^2\sigma^2}{\varepsilon_h^2} \log\left(\frac{2dT}{h\delta}\right) \log\left(\frac{t}{h}\right)\right)} , \tag{5.67}$$

*and the OGD step sequence of Proposition 5.25 satisfies*

$$\mathcal{R}_T = \mathcal{O}\left(\sigma\sqrt{\frac{\kappa dT}{m\rho_{\mathcal{X}}}}\left(\sqrt{\kappa \log\left(\frac{TL^2}{d}\right)} + h \log(dT)\right)\right) . \tag{5.68}$$

The episode length $h$ scales as $\mathcal{O}(\rho_{\mathcal{X}}^{-2})$, which grows at least as fast as $\mathcal{O}(d^2)$ in the action dimension $d$. This is sufficient to bound with high probability the smallest eigenvalue of the Hessian of the aggregate losses $\nabla^\alpha_{n,h}$ and thus ensure their strong convexity. However, longer episodes also means less frequent updates of $\bar{\theta}_n^{\mathrm{OGD}}$, i.e. less learning, which is materialised by the additional dependency on $h$ in the regret bound. In Lemma D.7, we deduce a tighter, more intricate expression for $h$, although still scaling as $\mathcal{O}(\rho_{\mathcal{X}}^{-2})$. We only report the simpler expression here to avoid cluttering.

**Remark 5.27**. *The union bound used in Proposition 5.25 imposes the knowledge of the horizon* $T$ *at runtime (in the definition of* $\gamma_{t,T}$*), thus making Algorithm 4 not anytime.*

*In Chapter 7, we propose another randomised algorithm for linear bandits with elicitable convex losses, where the effect of randomisation also adds (heuristically) a polylog correction in $T$ to the pseudo regret bound.*

## 5.4 Applications and numerical experiments

We conducted three numerical experiments to illustrate the merits of both risk-aware algorithms, under expectiles and entropic risk criteria. As far as we are aware, no algorithm exists for the expectile criterion; for entropic risk, Maillard (2013) analyses a variant of KL-UCB but only for the non-contextual multiarmed bandit problem (and without numerical evidences). Instead, we designed these numerical experiments so that the optimal arms were different depending on the criterion of interest (mean versus risk-aware), and we benchmarked the risk-aware algorithms against the classical mean-linear LinUCB (Abbasi-Yadkori et al., 2011).

**Computing expectiles.** We detail two cases of distributions for which expectiles are known. For $p \in (0,1)$, we denote by $e_p(\nu)$ the $p$-expectile of distribution $\nu$.

- If $\nu = \mathcal{N}(0,1)$, then, letting $\phi$ and $\Phi$ be the p.d.f. and c.d.f. of $\nu$ respectively, we obtain after simple calculus and the identity $\phi'(y) = -y\phi(y)$ the following fixed point equation:

$$e_p(\nu) = \frac{2p\phi(e_p(\nu)) - 1}{(1-2p)\Phi(e_p(\nu)) + p}\,, \tag{5.69}$$

  from which one can estimate the value of $e_p(\nu)$ using a fast iterative scheme. The general Gaussian case $\nu = \mathcal{N}(\mu, \sigma^2)$ is then easily deduced from the relation $e_p(\nu) = \mu + \sigma e_p(\mathcal{N}(0,1))$. Expectile calculations for a few other classical distributions are covered in Philipps (2022).

- If $\nu$ is the so-called expectile based distribution (Picheny et al., 2022; Arbel et al., 2023) with asymmetric density (with respect to the Lebesgue measure) given by

$$f_{\mu,\sigma,p}(y) = \frac{c_p}{\sigma} \exp\left(-\frac{|p - \mathbb{1}_{y<\mu}|(y-\mu)^2}{2\sigma^2}\right)\,, \tag{5.70}$$

  where $c_p = \frac{\sqrt{2p(1-p)}}{\sqrt{\pi}(\sqrt{p}+\sqrt{1-p})}$, then $e_p(\nu) = \mu$. In other words, these distributions offer a family parametrised directly by their expectile, generalizing the family of Gaussian distributions parametrised by their mean (for a given variance).

We recall that the $p$-expectile can be elicited by the convex potential $\psi(z) = |p - \mathbb{1}_{z<0}|z^2$. The second derivative of this potential is given by $\psi''(z) = 2(1-p)\mathbb{1}_{z<0} + 2p\mathbb{1}_{z>0}$, which is bounded between $2p$ and $2(1-p)$. In particular, Assumption 5.8 holds with conditioning $\kappa = \frac{M}{m} = \frac{1-p}{p}$ if $p \leqslant \frac{1}{2}$ and $\kappa = \frac{M}{m} = \frac{p}{1-p}$ otherwise. Note that the two classes of distributions considered above are Gaussian or strongly logconcave, which fits the scope of the supermartingale control of Lemma 5.11. In addition, the following minimax lower bound holds.

**Corollary 5.29** (Minimax lower bound for expectile bandits). *Let $\sigma > 0$ and $p \in (0,1)$. Assume that $\mathcal{X}_t = \mathcal{X}$ for all $t \in \mathbb{N}$. If either (i) $\mathcal{X} = \mathbb{B}^d_{\|\cdot\|_\infty}(0,1)$ or (ii) $\mathcal{X} = \mathbb{B}^d_{\|\cdot\|_2}(0,1)$, the expected regret of any policy on the bandit model $\nu_\theta(x)$ with density $f_{\langle\theta,x\rangle,\sigma,p}$ for all $x \in \mathcal{X}, \theta \in \mathbb{R}^d$ and $\rho$ the $p$-expectile risk measure is at least, for $T$ large enough,*

$$\sup_{\theta\in\mathbb{R}^d} \mathbb{E}_\theta\left[\mathcal{R}_T\right] = \Omega(\sigma d\sqrt{T}). \tag{5.71}$$

*Proof of Corollary 5.29.* Let us assume that $\sigma = 1$, the general case follows from a simple scaling argument, just as in the classical Gaussian case. For $\theta, \theta' \in \mathbb{R}^d$ and $x \in \mathcal{X}$, let $\mu = \langle\theta, x\rangle$ and $\mu' = \langle\theta', x\rangle$. We have the following likelihood identity:

$$\mathbb{E}_{Y\sim\nu_\theta(x)}\left[\log\frac{\mathrm{d}\nu_\theta(x)}{\mathrm{d}\nu_{\theta'}(x)}(Y)\right] = \frac{1}{2\sigma^2}\mathbb{E}_{Y\sim\nu_\theta(x)}\left[|p - \mathbb{1}_{Y<\mu'}|\left(Y-\mu'\right)^2 - |p-\mathbb{1}_{Y<\mu}|\left(Y-\mu\right)^2\right]. \tag{5.72}$$

Fix $\varepsilon \in \mathbb{R}$ and $\mu' = \mu + \varepsilon$. In order to apply Theorem 1.10, we need to prove that the right-hand side is $\mathcal{O}(\varepsilon^2)$ when $\varepsilon \to 0$. Assume that $\varepsilon > 0$ (the reverse case is obtained similarly). The Taylor expansion $(Y-\mu-\varepsilon)^2 = (Y-\mu)^2 + 2(Y-\mu)\varepsilon + \mathcal{O}(\varepsilon^2)$ allows to simplify the above expectation as

$$\begin{aligned}
&\mathbb{E}_{Y\sim\nu_\theta}\left[|p-\mathbb{1}_{Y<\mu+\varepsilon}|\left(Y-\mu\right)^2 + 2|p-\mathbb{1}_{Y<\mu+\varepsilon}|(Y-\mu)\varepsilon - |p-\mathbb{1}_{Y<\mu}|\left(Y-\mu\right)^2\right] + \mathcal{O}(\varepsilon^2)\\
&= \mathbb{E}_{Y\sim\nu_\theta}\left[\left(|p-\mathbb{1}_{Y<\mu+\varepsilon}| - |p-\mathbb{1}_{Y<\mu}|\right)\left(Y-\mu\right)^2\right] + 2\varepsilon\mathbb{E}_{Y\sim\nu_\theta}\left[|p-\mathbb{1}_{Y<\mu+\varepsilon}|(Y-\mu)\right] + \mathcal{O}(\varepsilon^2)\\
&= (2p-1)\mathbb{E}_{Y\sim\nu_\theta}\left[\mathbb{1}_{Y\in[\mu,\mu+\varepsilon]}\left(Y-\mu\right)^2\right] + 2\varepsilon\mathbb{E}_{Y\sim\nu_\theta}\left[|p-\mathbb{1}_{Y<\mu+\varepsilon}|(Y-\mu)\right] + \mathcal{O}(\varepsilon^2).
\end{aligned} \tag{5.73}$$

The first expectation can be rewritten as $\int_0^\varepsilon y^2 f_{0,\sigma,p}(y)dy$, and since the integrand $y \mapsto y^2 f_{0,\sigma,p}(y)$ is equivalent to $y \mapsto c_p/\sigma y^2$ in the neighbourhood of zero, this integral contributes to $\mathcal{O}(\varepsilon^3)$.

For the second term, we also write the integral and split it in three parts:

$$
\begin{aligned}
&\mathbb{E}_{Y\sim\nu_\theta}\left[|p-\mathbb{1}_{Y<\mu+\varepsilon}|(Y-\mu)\right] \\
&\quad = \frac{c_p}{\sigma}\Bigg(\int_{-\infty}^{\mu}(1-p)(y-\mu)e^{-(1-p)\frac{(y-\mu)}{2\sigma^2}}dy + \int_{\mu}^{\mu+\varepsilon}(1-p)(y-\mu)e^{-p\frac{(y-\mu)^2}{2\sigma^2}}dy \\
&\qquad + \int_{\mu+\varepsilon}^{+\infty}p(y-\mu)e^{-p\frac{(y-\mu)^2}{2\sigma^2}}\Bigg)\,. \qquad (5.74)
\end{aligned}
$$

Now, we note that $\mathbb{E}_{Y\sim\nu_\theta}[|p-\mathbb{1}_{Y<\mu(x)}|(Y-\mu(x))] = 0$ by definition of the $p$-expectile $\mu$ (first order condition). Hence, adding and substracting $c_p/\sigma\int_{\mu}^{+\infty}p(y-\mu)\exp(-p(y-\mu)^2/(2\sigma^2))dy$ to this expression yields

$$
\mathbb{E}_{Y\sim\nu_\theta}\left[|p-\mathbb{1}_{Y<\mu+\varepsilon}|(Y-\mu)\right] = \frac{c_p}{\sigma}\int_{\mu}^{\mu+\varepsilon}(1-2p)(y-\mu)e^{-p\frac{(y-\mu)^2}{2\sigma^2}}dy = \mathcal{O}(\varepsilon)\,, \qquad (5.75)
$$

since the integrand is bounded on $[\mu,\mu+\varepsilon]$. Combining all these estimates results in the bound $\mathcal{O}(\varepsilon^2)$, which concludes the proof by Theorem 1.10. ■

We recall that although this lower bound is on the expected pseudo regret, Lemma A.1 shows that it translates to a high probability bound as in Theorems 5.18, 5.23 and 5.26.

**Computing entropic risk.** For a distribution $\nu$, the entropic risk at level $\gamma > 0$ takes the form $\rho_\gamma(\nu) = \frac{1}{\gamma}\log\mathbb{E}_{Y\sim\nu}\left[e^{\gamma Y}\right]$ and corresponds to the loss $\mathcal{L}\colon (y,\xi)\mapsto \xi+\frac{1}{\gamma}(e^{\gamma(y-\xi)}-1)$. Derivatives of this loss satisfy the following identities, where $\partial$ represents the differentiation operator with respect to the second coordinate $\xi$:

$$
\begin{aligned}
\partial\mathcal{L}(y,\xi) &= 1-e^{\gamma(y-\xi)}, \\
\partial^2\mathcal{L}(y,\xi) &= \gamma e^{\gamma(y-\xi)},
\end{aligned}
$$

and is thus in particular strictly convex.

For a Bernoulli-like distribution $\nu = p\delta_a + (1-p)\delta_b$, with $p\in(0,1)$, $a,b\in\mathbb{R}$, the entropic risk takes the simple form $\rho_\gamma(\nu) = \frac{1}{\gamma}\log\left(pe^{\gamma a}+(1-p)e^{\gamma b}\right)$. If $\nu$ has a bounded support with diameter $\mathcal{D}$, then it is clear that the Hessian of the loss is controlled by $m = \gamma e^{-\gamma\mathcal{D}} \leqslant \partial^2\mathcal{L} \leqslant \gamma e^{\gamma\mathcal{D}} = M$, and therefore the conditioning number of the loss $\kappa$ can be bounded by $e^{2\gamma\mathcal{D}}$. Finally, $\nu$ being bounded also fits the scope of the supermartingale control of Lemma 5.11.

**Remark 5.30** (Computing generic risk measures). *If a density $p$ and a loss function $\mathcal{L}$ are known, one may resort to numerical integration to approximate the following quantity up to arbitrary precision:*

$$\mathbb{E}_{Y\sim\nu}\left[\mathcal{L}(Y,\xi)\right] = \int \mathcal{L}(y,\xi)p(y)dy \approx \sum_i w_i \mathcal{L}(y_i,\xi)p(y_i),$$

*where the weights $(w_i)$ and knots $(y_i)$ depend on the approximation routine. Then, one may simply run a minimisation algorithm on the function $\xi \mapsto \sum_i w_i \mathcal{L}(y_i,\xi)p(y_i)$ to estimate $\rho_{\mathcal{L}}(\nu)$.*

**Experiment 1: multiarmed Gaussian bandit with expectile noise.** We considered $K = 2$ Gaussian arms with expectiles at level $p = 10\%$ equal to 1 and 0 respectively. This bandit can be represented by constant orthonormal actions $\mathcal{X}_t = \{[1\ 0]^\top,\ [0\ 1]^\top\}$, parameter $\theta^\star = [1\ 0]^\top$ and noise distributions $\mathcal{N}\left(\mu_k, \sigma_k^2\right)$, with $\mu_k$ and $\sigma_k$ chosen such that the expectile of the corresponding noise is zero for $k \in \{1, 2\}$. This can be achieved with e.g. $\mu_1 \approx 0.44$, $\sigma_1 = 0.5$ and $\mu_2 \approx 2.62$, $\sigma_2 = 3$, which was the setup for this experiment. Note that for a given expectile level $p \in (0, 1)$ and standard deviation $\sigma$, finding the unique mean $\mu$ such that $\mathcal{N}\left(\mu, \sigma^2\right)$ has zero $p$-expectile can be easily done via a numerical root search, using the formula for Gaussian expectiles described above.

The optimal arm with respect to the expectile criterion is the first one by definition. However, the expectations of these arms are in reversed order, making the second one optimal with respect to the mean criterion.

As recalled in Example 5.2, the corresponding expectile loss satisfies Assumption 5.8 with $m = 2\min(p, 1-p)$ and $M = 2\max(1-p, p)$. Note that the more risk-averse $(p \to 0)$, the flatter the loss $(m \to 0)$ and thus the harder it is to learn. This matches the intuition on risk-aware measures: by focusing on the more extreme events, they require more samples to reach the same statistical accuracy.

**Experiment 2: linear bandit with expectile asymmetric noise.** We considered a second example with non-Gaussian noise and non-orthogonal features. We defined the action set at time $t$ by $\mathcal{X}_t = \{X_t^1, X_t^2\} \subset \mathbb{R}^3$ where:

- $X_t^1 = \frac{Z_t^1}{\|Z_t^1\|_2}$ with $Z_t^1 \sim \mathcal{N}([1\ 0\ 0]^\top, \sigma_x I_3)$,
- $X_t^2 = \frac{Z_t^2}{\|Z_t^2\|_2}$ with $Z_t^1 \sim \mathcal{N}([0\ 1\ 0]^\top, \sigma_x I_3)$,
- We set the action noise to an arbitrary value $\sigma_x = 0.1$.

- $(Z_t^1, Z_t^2)_{t\in\mathbb{N}}$ are all independent random variables.

This construction results in bounded, anisotropic actions. We chose $\theta^\star = [0.9\ 0\ 1]^\top$, so that $\langle\theta^\star, X_t^1\rangle$ is likely higher than $\langle\theta^\star, X_t^2\rangle$, thus favoring $X_t = X_t^1$ in the expectile model $Y_t = \langle\theta^\star, X_t\rangle + \eta_t$. To model the zero $p$-expectile noise $\eta_t$ with $p = 10\%$, we used the expectile based distribution presented above with $\mu_1 = \mu_2 = 0$ and $\sigma_1 = 0.5$ if action $X_t^1$ is played, and $\sigma_2 = 1.5$ otherwise, resulting in different mean noise $\mathbb{E}[\eta_t \mid \bar{\mathcal{G}}_t] \approx 1.8$ and $\mathbb{E}[\eta_t \mid \bar{\mathcal{G}}_t] \approx 3.3$ respectively. Note that this distribution is also strongly logconcave, thus fitting the scope of the supermartingale control of Lemma 5.11. As in the previous example, this setting was designed to deceive the mean criterion by inverting the order of optimal actions.

**Experiment 3: multiarmed Bernoulli bandit with entropic risk noise.** The last experiment consisted of $K = 2$ Bernoulli-like arms $\nu_1 = \frac{1}{2}\delta_1 + \frac{1}{2}\delta_{-1}$ and $\nu_2 = \frac{1}{4}\delta_2 + \frac{3}{4}\delta_{-2}$, which corresponds to means $\mu_1 = 0$, $\mu_2 = -1$ and entropic risk $\rho_\gamma(\nu_1) \approx 0.43$ and $\rho_\gamma(\nu_2) \approx 0.67$ at level $\gamma = 1$. Again, this setting was designed so that the best optimal arm is different under the mean and entropic risk criteria.

**Results.** In each of the three settings, we ran an instance of Algorithm 3, i.e. LinUCB (convex risk), and Algorithm 4, i.e. LinUCB-OGD (convex risk). We also ran a standard LinUCB algorithm for the mean criterion (Abbasi-Yadkori et al., 2011). Hyperparameters $m, M$ and $\kappa$ were tuned according to the analysis above. Regularisation was fixed at $\lambda = 0.1$. As is customary in bandit experiments, the parameter $\sigma$, which in the formal analysis is derived from the supermartingale control of the noise, was considered a degree of freedom to control the amount of exploration; we arbitrarily fixed it at $\sigma = 0.1$ in experiments 1 and 2 and at $\sigma = 1$ in experiment 3. For the LinUCB-OGD variant, the step size for the OGD scheme was set to $\varepsilon_n = 0.1/n$, following the linear decay suggested by Proposition 5.25, and the frequency of OGD update to $h = 5$. In addition, all algorithms went through an initial warmup phase where each arm was played 5 times, in order to ensure better stability of the initial estimations of $\theta$.

In all three examples, the mean criterion algorithm was deceived and accumulated linear expectile and entropic risk pseudo regret, while both risk-aware algorithms exhibited sublinear trends. Interestingly, the LinUCB-OGD variant showed higher regrets due to the approximate minimisation of the loss criterion by OGD, but remained below the mean criterion LinUCB benchmark. Figure 5.1 reproduces the results of each experiments across 500 independent replications. Finally, average runtimes for each algorithm are reported in Table 5.2. Calculations were performed on a distributed infrastructure comprised of 80 CPUs. While the values themselves are not indicative, as they would vary on a different system, their relative magnitudes illustrate the computational gain of the OGD scheme over solving the empirical risk minimisation problem at each step as required in LinUCB (convex risk). Note also that the standard LinUCB with mean criterion is faster due to the sequential nature of the ridge regression estimator. Indeed, this procedure involves inverting at each step a $d \times d$ matrix subject to rank one updates, which can be calculated efficiently via the Sherman-Morrison formula. By contrast, other convex losses than the one derived from the quadratic potential loose this sequential form and require solving the corresponding regression problem from scratch at each time step.

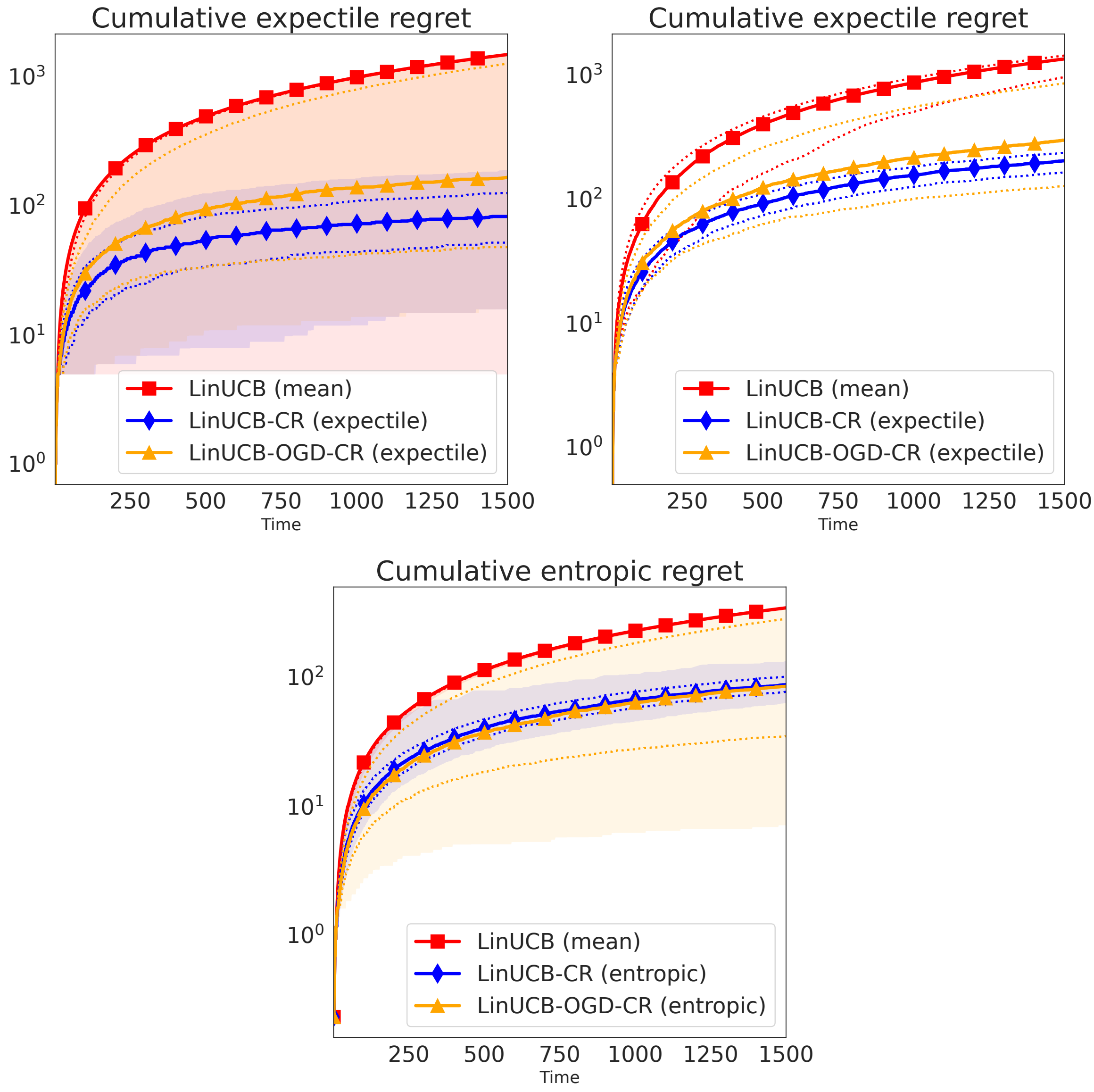

**Figure 5.1** – Left: two-armed Gaussian expectile bandit. Center: two-armed linear expectile bandit with $\mathbb{R}^3$ contexts and expectile-based asymmetric noises. Right: two-armed Bernoulli entropic risk bandit. Thick lines denote median cumulative pseudo regret over 500 independent replicates. Dotted lines denote the 25 and 75 pseudo regret percentiles. Shaded areas denote the 5 and 95 percentiles.

**Table 5.2** – Runtimes for the classical LinUCB and Algorithms 3 (LinUCB for convex risk) and 4 (LinUCB-OGD for convex risk) in each experiments. Runtimes are reported in seconds as mean $\pm$ standard deviation, estimated across 500 independent replications with time horizon $T = 1500$.

| Algorithm | Experiment 1 | Experiment 2 | Experiment 3 |
|---|---|---|---|
| LinUCB (mean) | $0.4 \pm 0.0$ | $37.2 \pm 4.9$ | $0.6 \pm 0.0$ |
| LinUCB-CR (convex risk) | $231.0 \pm 21.7$ | $814.8 \pm 88.3$ | $519.1 \pm 33.3$ |
| LinUCB-OGD-CR (convex risk) | $20.4 \pm 3.9$ | $60.2 \pm 12.0$ | $25.7 \pm 4.9$ |

## Conclusion

We have introduced a new setting for contextual bandits, building on the recent interest for risk-awareness in multiarmed bandits. We reviewed the literature on risk measures, in particular the notion of elicitability, that allows to extend the risk minimisation framework of ridge regression beyond standard mean-linear bandits. To lift the regret analysis of optimistic algorithms to the setting of scalar risk measures $\rho_{\mathcal{L}}$ elicited by a convex loss $\mathcal{L}$, we showed that uniformly bounding the curvature of the loss (Assumption 5.8) is sufficient to maintain near optimal minimax regret ($\mathcal{O}(d\sqrt{T})$, as evidenced in the novel lower bound of Corollary 5.29), up to polylog terms (Theorem 5.18 and 5.23). More precisely, we identified two key conditions, namely a supermartingale control (Lemma 5.11) and a transportation inequality (Lemma 5.15), that guarantee sublinear regret; while these are direct consequences of the bounded curvature assumption, they may hold in different settings, as was recently discovered in GLB.

Going further, we believe it would be interesting to extend the linear model between actions and risk measures to generalised linear models ($\rho_{\mathcal{L}}(Y_t) = \mu(\langle\theta, X_t\rangle)$ for some link function $\mu\colon \mathbb{R} \to \mathbb{R}$), kernelised bandits ($\rho_{\mathcal{L}}(Y_t) = f(X_t)$ where $f$ belongs to some RKHS) or neural bandits ($\rho_{\mathcal{L}}(Y_t) = f_\theta(X_t)$ where $f_\theta$ is a neural network with weights $\theta$). Moreover, capturing well-established risk measures such as mean-variance, conditional value-at-risk or quantiles would require to adapt the theory to high-order elicitable measures and to non-smooth losses. Finally, we believe the technical results developed here, can pave a way for the design and the analysis of randomised strategies in the contextual risk-aware setting, such as Thompson sampling. We discuss a heuristic example of such extensions in Chapter 7.

**Take-home message:**

☞ **It is possible to sequentially optimise other risk measures than the expectation with contextual information for little additional complexity.**

We have extended the classical linear bandit framework to handle sequential optimisation of other distributional properties of the rewards than their means. To this end, we considered *elicitable risk measures*, defined as minimisers of expected convex losses, including:

- the mean criterion, a special case (symmetric quadratic loss);
- the **expectile risk measures**, which correspond to asymmetric quadratic losses and encode a risk aversion (negative and positive outcomes are weighted differently).

Up to mild assumptions on the risk measures, this new formalism is not significantly more difficult. In particular, we have obtained minimax logarithmic pseudo regret guarantees for a natural extension of LinUCB in this setting.

# Part IV

# Nonparametric modelling in bandits

*or how to play well without knowing much*

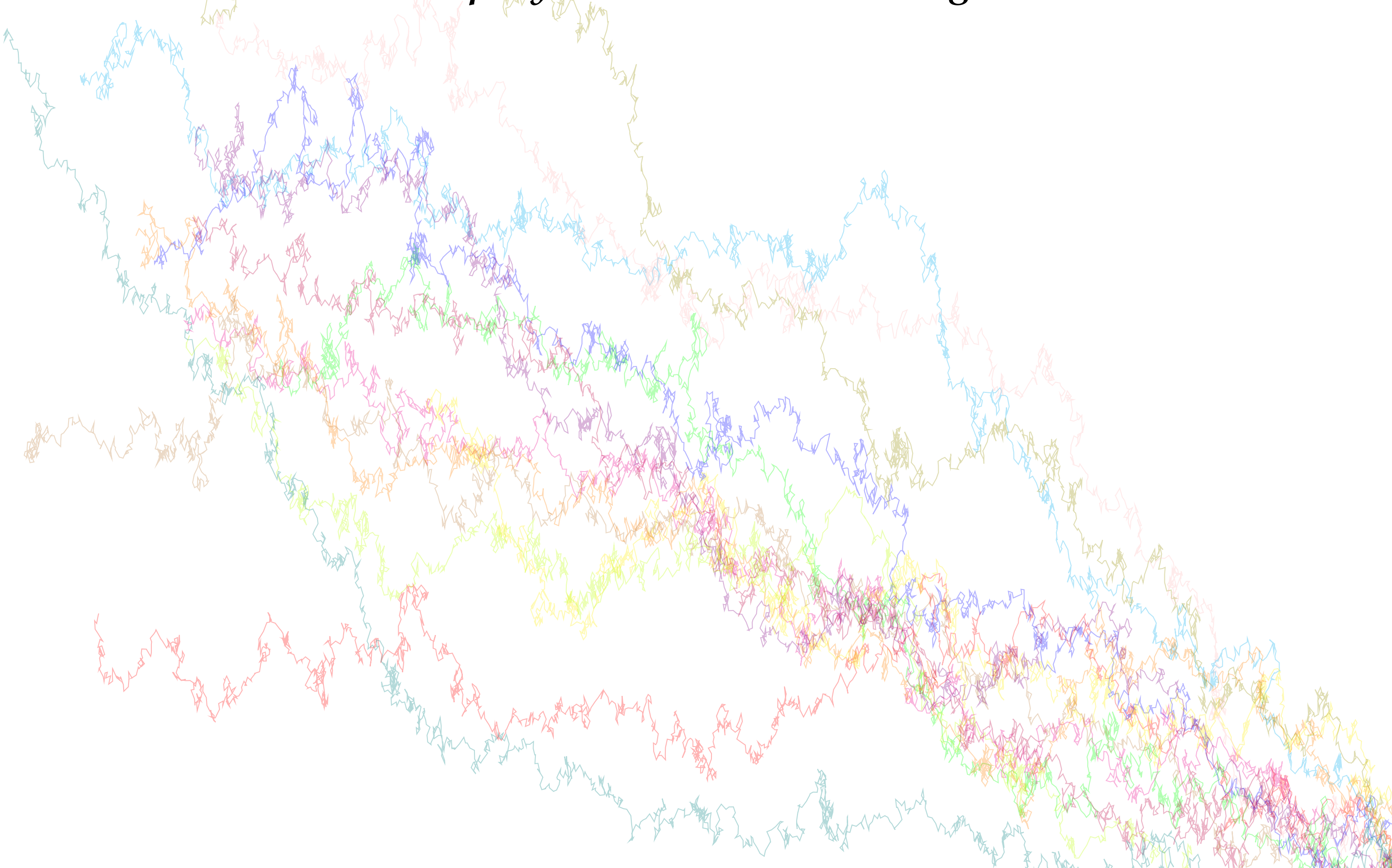

# Chapter 6

# From optimality to robustness: Dirichlet sampling strategies in stochastic bandits ( )

> [...] *va devant toi, au hasard, sans boussole. Si tu vois une marque faite sur le sable, prends la fuite et dis au sable du Désert : Je te veux intact ; dis-moi où nul pied ne t'a touché.*
>
> — Ernest Hello, *Paroles de Dieu*

The content of this chapter was published at the 2021 *Conference on Neural Information Processing Systems* (NeurIPS) (Baudry et al., 2021c) and presented at the 2021 *Inria Scool* seminar. The notations and presentation have been heavily reworked compared to that article. To avoid cluttering, the proofs of the regret analysis, which involve careful manipulations of the round-bases structure and fine properties of Dirichlet distributions, are deferred to Appendix E. We instead stress out (i) the problem of boundary crossing probabilities (in particular, we correct a detail in the proof of Lemma 6.8), highlighting connections with concentration and the study of the extremal Kullback-Leibler operator, and (ii) the extensive numerical experiments. Of note, building on similar posterior sampling principles, several works followed-up on ours: Tiapkin et al. (2022a,b) in reinforcement learning, and Jourdan et al. (2022) in pure exploration.

## Contents

## Outline and contributions

We recall from Section 1.2 in Chapter 1 that a $K$-armed stochastic bandit is a model for decision making problem in which a learner sequentially picks actions among $K$ arms and collects random rewards according to the product measure $\boldsymbol{\nu} = (\nu_k)_{k\in[K]} \in \bigotimes_{k\in[K]} \mathcal{F}_k$ (unstructured bandit), where each family is assumed to be integrable ($\mathcal{F}_k \subseteq \mathbb{L}^1(\mathbb{R})$), with mean $\mu_k$. The objective of the learner is to adapt their policy $(\pi_t)_{t\in[T]}$ to minimise the expected pseudo regret

$$\mathbb{E}[\mathcal{R}_T] = \mathbb{E}\left[\sum_{t=1}^{T} \mu^\star - \mu_{\pi_t}\right] = \sum_{k=1}^{K} \Delta_k \mathbb{E}\left[N_T^k\right] , \tag{6.1}$$

where $N_T^k = \sum_{t=1}^T \mathbb{1}_{\pi_t = k}$ is the number of pulls to arm $k$ after $T$ steps, $\mu^\star = \max_{k\in[K]} \mu_k$ is the maximum expected reward and $\Delta_k = \mu^\star - \mu_k$ is the *suboptimality gap* of arm $k \in [K]$. Optimal policies are constrained by the asymptotic lower bound of Theorem 1.6, i.e.

$$\liminf_{T\to\infty} \frac{\mathbb{E}[\mathcal{R}_T]}{\log T} \geqslant \sum_{k:\Delta_k>0} \frac{\Delta_k}{\mathcal{K}_{\inf}^{\mathcal{F}_k}(\nu_k;\mu^\star)} , \tag{6.2}$$

where the extremal Kullback-Leibler operator over a family of distributions $\mathcal{F} \subseteq \mathbb{L}^1(\mathbb{R})$ is:

$$\begin{aligned} \mathcal{K}_{\inf}^{\mathcal{F}} : \mathcal{F} \times \mathbb{R} &\longrightarrow \mathbb{R} \\ (\nu;\mu^\star) &\longmapsto \inf_{\nu'\in\mathcal{F}} \left\{ \mathrm{KL}(\nu \,\|\, \nu'),\ \mathbb{E}_{Y\sim\nu'}[Y] > \mu^\star \right\} . \end{aligned} \tag{6.3}$$

For integrable distributions supported in $(-\infty, \overline{B})$, we use the notation $\mathcal{K}_{\inf}^{\overline{B}} = \mathcal{K}_{\inf}^{\mathcal{F}_{(-\infty,\overline{B})}}$.

As discussed in Chapter 2, an outstanding issue in multiarmed bandits is to design efficient algorithms under weak, practitioner-friendly hypotheses. In this chapter, we propose simple alternative settings allowing for unspecified tail shapes while still avoiding "mass leakage" at infinity (Lemma 1.13). We introduce a flexible template for randomised index policies, **Dirichlet sampling** (DS), by generalising the index of NPTS (Riou and Honda, 2020) with data-dependent exploration incentives, implemented within a round-based structure inspired by Chan (2020); Baudry et al. (2020) (Section 6.1). We then introduce in Section 6.2 a first regret decomposition of DS algorithms under general assumptions, and the technical results that allow to fine-tune the algorithm for different families (Section 6.2). We provide three instances of DS algorithms and their respective settings and regret guarantees in Section 6.3: (i) *bounded Dirichlet sampling* (BDS) tackles bounded distributions with possibly unknown upper bounds; (ii) *quantile Dirichlet sampling* (QDS) substitutes common tail assumptions for a mild quantile condition while preserving logarithmic regret for unbounded distributions; last, *robust Dirichlet sampling* (RDS) exhibit slightly larger than logarithmic regret for any unspecified *light-tailed* distributions, making it a competitor to the robust R-UCB algorithm of Ashutosh et al. (2021).

We study in Section 6.4 a use case in agriculture using the DSSAT simulator (Hoogenboom et al., 2019), which naturally faces the questions of robustness and model specification that motivated this work and shows the merit of DS over state-of-the-art methods for this problem. We leave possible extensions (e.g. heavy tailed distributions, with tools such as truncated or median of means estimators, see Bubeck et al. (2013)) for future work.

## 6.1 Dirichlet sampling algorithms

We recall the definition of NPTS and motivate the introduction of a round-based structure to analyse data-dependent exploration bonuses for Dirichlet sampling.

**Background.** NPTS is an index strategy where the index of each arm is a *random reweighting* of their past rewards, augmented by an *exploration bonus*. The weights are drawn from the Dirichlet distribution $\mathcal{D}_{n+1} = \text{Dir}((1, \dots, 1))$ for samples of size $n+1$, which is the uniform distribution on the $n$-simplex $\mathfrak{D}^n = \{w \in [0,1]^{n+1}, \sum_{i=1}^{n+1} w_i = 1\}$ and matches the Bayesian posterior (i.e. Thompson Sampling) for multinomial arms. The exploration bonus is simply the known upper bound of the support, and avoids underexploration of potentially unlucky good arm. We provide further explanations on the Dirichlet distribution in Appendix E.2.

The simplicity of NPTS and its strong theoretical guarantees are appealing for further generalisation. As we fully depart from the Bayesian approach, considering other exploration bonuses, we derive a new family of algorithms under the name of *Dirichlet Sampling*. We keep the two principles of reweighting the observations using a Dirichlet distribution and the exploration aid, and explore how to apply them to more general (e.g. unbounded) distributions. In particular, we allow in DS some pre-processing of the observations before reweighting (see section 6.2 and 6.3) and motivate in Section 6.2 the use of a *data-dependent* bonus, that use information from several arms. The additional complexity in the analysis requires a change of algorithm structure, dropping the index policy for a *leader vs challenger* approach (Chan, 2020).

**Round-based algorithm.** We define a round as a step of the algorithm at the end of which a set of (possibly several) arms are selected to be pulled. Let $\mathcal{A}_r \subseteq [K]$ be the subset of the arms pulled at round $r \in \mathbb{N}$. The (essentially unchanged) definitions of the number of pulls and $T$-round cumulative expected pseudo regret in this context are

$$N_T^k = \sum_{r=1}^{T} \mathbb{1}_{k \in \mathcal{A}_r} , \tag{6.4}$$

$$\mathbb{E}[\mathcal{R}_T] = \mathbb{E}\left[\sum_{r=1}^{T} \sum_{k \in [K]} \Delta_k \mathbb{1}_{k \in \mathcal{A}_r}\right] = \sum_{k \in [K]} \Delta_k \mathbb{E}[N_T^k] . \tag{6.5}$$

We consider the $T$-round regret for simplicity, as it is a simple upper bound of the regret after $T$ pulls (moreover, as we discuss below, we expect $\mathcal{A}_r$ to be often reduced to a single, optimal arm when $r$ is large enough). At the beginning of each round we define a reference arm (leader), and then organise pairwise comparisons called *duels* between this arm and the other arms (challengers). The leader is chosen as the arm with largest sample size,

$$\ell_r \in \underset{k\in[K]}{\text{argmax}}\, N_r^k\,, \tag{6.6}$$

where ties are broken first in favour of the best arm, then with a random choice. A major motivation for structuring the algorithm in such a way is that the leader will have a sample size that is *linear* in the number of rounds, as at least one arm is chosen at each round. This ensures strong statistical properties that we will exploit to design the exploration bonus of DS strategies. Randomising the index of the leader is also unnecessary: it competes against each challenger with its *empirical mean*. We also dismiss all the arms $k$ that satisfy $N_r^k = N_r^{\ell_r}$ with the same argument. These design choices have a practical interest as they avoid the computation time of drawing the largest weight vectors. We believe this can be an alternative of independent interest for computationally intensive index policies.

**Challenger's index.** Following notations of Chapter 1, for a given pair of challenger and leader $k, \ell \in [K]$, we denote by $\mathbb{Y}_r^k = (Y_i^k)_{i=1}^{N_r^k}$ and $\mathbb{Y}_r^\ell = (Y_i^\ell)_{i=1}^{N_r^\ell}$ their respective reward histories, and let $\widehat{\mu}_r^k$ and $\widehat{\mu}_r^\ell$ be their respective empirical average rewards at round $r$. The pairwise comparison between $k$ and $\ell$ is then performed by calculating an *index*, that can depend on both reward histories $\mathbb{Y}_r^k$ and $\mathbb{Y}_r^\ell$, but not directly on the current round $r$. For now, we denote by $\widetilde{\mu}\colon \mathbb{R}^{(\mathbb{N})} \times \mathbb{R}^{(\mathbb{N})} \to \mathbb{R}$ a generic function to represent such an index.

The duel between $k$ and $\ell$ includes two steps, and the challenger is declared the winner whenever one of the following two conditions holds:

(i) $\widehat{\mu}_r^k \geqslant \widehat{\mu}_r^\ell$ (first attempt with empirical mean), or

(ii) $\widetilde{\mu}\left(\mathbb{Y}_r^k, \mathbb{Y}_r^\ell\right) \geqslant \widehat{\mu}_r^\ell$,

in this order. We summarise in Algorithm 5 the steps of the DS algorithm for the generic index $\widetilde{\mu}$. The algorithm is part of the DS family if this index includes Dirichlet reweighting of the observations augmented by an exploration bonus. However, the algorithm structure in Algorithm 5 may in principle be combined with any randomised index, which is of independent interest as we will see in Section 6.2. In this work, we are primarily interested in the following canonical example of Dirichlet sampling index, which directly generalises the NPTS index with a data-dependent (instead of fixed) exploration bonus.

**Definition 6.1** (DS index). *Let $B\colon \mathbb{R}^{(\mathbb{N})} \times \mathbb{R}^{(\mathbb{N})} \to \mathbb{R}$. The* ***DS index*** *associated with $B$ is the mapping defined as*

$$\begin{aligned} \widetilde{\mu}_W : \mathbb{R}^{(\mathbb{N})} \times \mathbb{R}^{(\mathbb{N})} &\longrightarrow \mathcal{M}_1^+(\mathbb{R}) \\ (\mathbb{Y}, \mathbb{Y}') &\longmapsto \sum_{i=1}^{n} W_i y_i + W_{n+1} B\left(\mathbb{Y}, \mathbb{Y}'\right) , \end{aligned} \tag{6.7}$$

*where $\mathbb{Y} = (y_i)_{i=1}^n \in \mathbb{R}^{(\mathbb{N})}$, $\mathbb{Y}' \in \mathbb{R}^{(\mathbb{N})}$ and $W = (W_i)_{i=1}^{n+1} \sim \mathcal{D}_{n+1}$.*

To avoid cluttering, we resort to a slight abuse of notation as we identify the distribution $\widetilde{\mu}_W\left(\mathbb{Y}, \mathbb{Y}'\right)$ and its instanciation as a random variable. In the rest of this chapter, by *drawing a DS index* at round $r + 1$ for a challenger $k$ and leader $\ell$, we mean to generate a realisation of a uniform Dirichlet weight vector $W$ of size $N_r^k + 1$ (independent of the rest) while keeping elements of $\mathbb{Y}_r^k$ and $\mathbb{Y}_r^\ell$ fixed, and returning the weighted sum defined in $\widetilde{\mu}_W(\mathbb{Y}_r^k, \mathbb{Y}_r^\ell)$.

In the next section we study theoretical properties of Dirichlet sampling algorithms, and discuss the choice of the index $\widetilde{\mu}_W$ for different families of distributions.

**Algorithm 5** Generic round-based randomised index bandit algorithm

**Input:** $K$ arms, index function $\widetilde{\mu}$

**Initialisation:** $t = 1, r = 1, \forall k \in [K], \mathbb{Y}^k = \{\}, N^k = 0, S^k = 0, \widehat{\mu}^k = 0.$

**while** *continue* **do**

  $\mathcal{A} = \{\}$ ; ▷ Set of arms to pull at the end of the round

  **if** $r = 1$ **then**

    $\mathcal{A} = [K]$ ; ▷ All arms are pulled at the first round

  **else**

    // Pick the leader

    $\mathcal{L} = \operatorname{argmax}_{k \in [K]} N^k$ ; ▷ Arm(s) with the largest number of samples

    $\bar{\mathcal{L}} = \operatorname{argmax}_{k \in \mathcal{L}} \widehat{\mu}^k$ ; ▷ Keep best arm(s) in $\mathcal{L}$

    $\ell \sim \mathcal{U}(\bar{\mathcal{L}})$ ; ▷ Random choice if several candidates

    // Duels

    **for** $k \in [K] \setminus \mathcal{L}$ ; ▷ Challengers in $\mathcal{L}$ are eliminated

    **do**

      **if** $\widehat{\mu}^k \geqslant \widehat{\mu}^\ell$ **then**

        $\mathcal{A} \leftarrow \mathcal{A} \cup \{k\}$ ; ▷ First duel with empirical mean

      **else**

        Draw index $\widetilde{\mu}(\mathbb{Y}^k, \mathbb{Y}^\ell)$

        **if** $\widetilde{\mu}(\mathbb{Y}^k, \mathbb{Y}^\ell) \geqslant \widehat{\mu}^\ell$ **then**

          $\mathcal{A} \leftarrow \mathcal{A} \cup \{k\}$ ; ▷ Second duel with DS index

  // Collect reward and update quantities

  **if** $|\mathcal{A}| = 0$ **then**

    $\mathcal{A} \leftarrow \{\ell\}$ ; ▷ If no winning challenger $\ell$ is pulled

  Shuffle $\mathcal{A}$

  **for** $k \in \mathcal{A}$ **do**

    Observe $Y^k_{N^k}$, $\mathbb{Y}^k \leftarrow \mathbb{Y}^k \cup \{Y^k_{N^k}\}$, $N^k \leftarrow N^k + 1$ ; ▷ Update

    $S^k \leftarrow S^k + Y^k_{N^k}$, $\widehat{\mu}^k \leftarrow S^k / N^k$, $t \leftarrow t + 1$ ;

  $r = r + 1$ .

## 6.2 Regret analysis and technical results

In this section, we analyse the regret of DS algorithms. We first derive a general regret decomposition for any randomised index $\widetilde{\mu}$ that holds thanks to the round-based structure. We then introduce several properties of Dirichlet sampling, that theoretically guide proper tuning of a DS index. We finally instantiate DS for three different problems and provide regret bounds in these settings. Starting with the regret decomposition, we exhibit general sufficient conditions to prove regret guarantees arbitrary indices. The first one concerns the finite sample concentration of the mean of each distribution.

**Assumption 6.2** (Concentration of empirical means). *For each $k \in [K]$ and $\nu_k \in \mathcal{F}_k$, there exists a* good rate function *$I_k : \mathcal{I}_k \to \mathbb{R}_+$ with $\mathcal{I}_k \subseteq \mathbb{R}$ an open set, such that*

*(i)* $\forall k, \ell \in [K],\ \mu_k \in \mathcal{I}_\ell$,

*(ii)* $\forall x \in \mathcal{I}_k \setminus \{\mu_k\},\ I_k(x) > 0$,

*(iii) for all i.i.d. sequence $(Y_i)_{i\in\mathbb{N}}$ drawn from $\nu_k$, $n \in \mathbb{N}$ and $\varepsilon > 0$ such that $[\mu_k - \varepsilon, \mu_k + \varepsilon] \subset \mathcal{I}_k$ and $\widehat{\mu}_n = 1/n \sum_{i=1}^n Y_i$, we have:*

$$\mathbb{P}\left(\widehat{\mu}_n \geqslant \mu_k + \varepsilon\right) \leqslant e^{-nI_k(\mu_k+\varepsilon)} \quad \text{and} \quad \mathbb{P}\left(\widehat{\mu}_n \leqslant \mu_k - \varepsilon\right) \leqslant e^{-nI_k(\mu_k-\varepsilon)} . \tag{6.8}$$

This hypothesis is standard in the bandit literature, and is for instance satisfied by any *light tailed* distributions, with good rate function given by the Fenchel-Legendre transform of their CGF (Cramér's theorem, Proposition 1.15); in particular, if $\nu_k$ is $R$-sub-Gaussian, $I_k : x \in \mathbb{R} \mapsto \frac{(x-\mu_k)^2}{2R^2}$ is a valid choice. We refer to the monograph Dembo and Zeitouni (2009) for general techniques to derive such good rate functions. In particular, this shows that the empirical mean of a suboptimal arm is exponentially unlikely to exceed $\mu^\star$ (with rate $I_k(\mu^\star)$).

We now provide an upper bound on the round-regret presented in Section 6.1 for Algorithm 5. This upper bound exhibits two non-constant terms, that represent (i) the expected number of pulls of sub-optimal arms when the best arm is the leader, and (ii) the cost of underexploration of the best arm. These terms are, to some extent, present in the regret bounds of most randomised algorithms. Our result is in this sense similar to Kveton et al. (2019b, Theorem 1). Theorem 6.3 generalises it for an index that may depend on the history of two arms instead of one, which is a result of independent interest (indeed, at this step of the analysis, the index has not been instanciated to a specific form yet).

**Notations.** For simplicity, we now consider indices that only depends on the rewards of the leader through their (empirical) mean, which we denote by $\widetilde{\mu} : \mathbb{R}^{(\mathbb{N})} \times \mathbb{R} \to \mathbb{R}$. Theorem 6.3 below holds for any index using statistics on the leader's history that have concentration properties similar to Assumption 6.2 (e.g. possibly quantiles, variance, etc.) with slight adaptations of the proof. For instance, in Algorithm 5, the index of challenger $k$ against leader $\ell$ at round $r+1$ becomes $\widetilde{\mu}(\mathbb{Y}^k_r; \widehat{\mu}^\ell_r)$. We recall the notations $\mathbb{Y}^\star_{(n)} = (Y^\star_i)_{i=1}^n \sim \nu_{k^\star}^{\otimes n}$ for the list of the first $n \in \mathbb{N}$ random rewards of the optimal arm $k^*$ and $\widehat{\mu}^{k^\star}_{(n)} = \frac{1}{n}\sum_{i=1}^n Y^\star_i$ for their empirical mean.

**Theorem 6.3** (Generic regret decomposition of round-based randomised algorithms). *Consider a bandit measure $\boldsymbol{\nu} = (\nu_k)_{k\in[K]}$. Under Assumption 6.2, for any DS index $\widetilde{\mu}$ and $\varepsilon \in [0, \Delta_k)$, the expected number of pulls of each suboptimal arm $k \in [K] \setminus \{k^\star\}$ is upper bounded by*

$$\mathbb{E}\left[N_T^k\right] \leqslant n_k(T) + B_{T,\varepsilon} + C_{\boldsymbol{\nu},\varepsilon}\,, \tag{6.9}$$

*where $n_k(T) = \mathbb{E}\left[\sum_{r=1}^{T-1} \mathbb{1}_{k\in\mathcal{A}_{r+1}, \ell_r = k^\star}\right]$, $C_{\boldsymbol{\nu},\varepsilon}$ is a constant dependent on the bandit measure $\boldsymbol{\nu}$ and $\varepsilon$ but not $T$, and*

$$B_{T,\varepsilon} = \sum_{j\in[K]\setminus\{k^\star\}} \sum_{n=1}^{\lceil 2\log(T)/I_{k^*}(\mu_j+\varepsilon)\rceil} \sup_{\mu\in[\mu_j-\varepsilon, \mu_j+\varepsilon]} \mathbb{E}_{\mathbb{Y}^\star_{(n)}\sim\nu_{k^\star}^{\otimes n}}\left[\frac{\mathbb{1}_{\widehat{\mu}^{k^\star}_{(n)}\leqslant\mu}}{\mathbb{P}\left(\widetilde{\mu}(\mathbb{Y}^\star_{(n)};\mu) \geqslant \mu \;\middle|\; \mathbb{Y}^\star_{(n)}\right)}\right]. \tag{6.10}$$

The proof, which we defer to Appendix E.1, follows the general outline of Chan (2020), and makes all the components of $C_{\boldsymbol{\nu},\varepsilon}$ explicit. This term is related to the deviations of empirical means for all arms and is typically bounded by a (problem-dependent) constant under light tailed concentration (Assumption 6.2) so it does not depend on the index $\widetilde{\mu}$ but only on the rate functions and the means of each arm. The other two terms reflect the exploration strategy: $n_k(T)$ is the expected number of pulls of arm $k$ when the best arm is the leader — we interpret it as the sample size required to statistically distinguish both arms at horizon $T$; on the other hand, $B_{T,\varepsilon}$ measures the capacity of the best arm to recover from first few low rewards.

Theorem 6.3 is formulated in a rather generic fashion and can be regarded as a counterpart of Kveton et al. (2019b, Theorem 1). We will later analyse instances of Dirichlet sampling where the first order term of the regret is driven entirely by $n_k(T)$. We therefore introduce the following condition to control the contribution of $B_{T,\varepsilon}$ to the regret.

**Assumption 6.4** (Sufficient exploration of optimal arm). *For any $\varepsilon > 0$ and any sequence $(n^\star(T))_{T\in\mathbb{N}}$ such that $n^\star(T) = \mathcal{O}(\log T)$, we have*

$$\sum_{n=1}^{n^\star(T)} \mathbb{E}_{\mathbb{Y}^\star_{(n)}\sim\nu_{k^\star}^{\otimes n}}\left[\frac{\mathbb{1}_{\widehat{\mu}^{k^\star}_{(n)}\leqslant\mu^\star-\varepsilon}}{\mathbb{P}\left(\widetilde{\mu}(\mathbb{Y}^\star_{(n)};\mu^\star-\varepsilon)\geqslant\mu^\star-\varepsilon \mid \mathbb{Y}^\star_{(n)}\right)}\right] = o(\log T)\,. \tag{6.11}$$

The left-hand side represents the expected cost in terms of regret of underestimating the optimal arm; intuitively, it measures the expected number of losing rounds before finally winning one when starting with low rewards. This is a classic decomposition in bandit analysis, and a counterpart of Assumption 6.4 holds for most index policies with provable regret guarantees, e.g. Theorem 1 in Kveton et al. (2019b, Theorem 1) or Agrawal and Goyal (2012, Lemma 4)). Notably, this regret decomposition depends only on the distribution of the best arm and its randomised Dirichlet sampling index when it is a challenger.

**Corollary 6.5** (Conditions for controlled expected pseudo regret). *Under Assumption 6.2 and 6.4, the $T$-round cumulative expected pseudo regret of Algorithm 5 satisfies*

$$\mathbb{E}[\mathcal{R}_T] \leqslant \sum_{k\in[K]\setminus\{k^\star\}} \Delta_k n_k(T) + o(\log T)\,. \tag{6.12}$$

*Proof of Corollary 6.5.* The bound $\mathbb{E}[N_k(T)] \leqslant n_k(T) + o(\log T)$ for all $k \in [K]\setminus\{k^\star\}$ follows directly from Theorem 6.3 and Assumption 6.4, and we conclude by invoking the standard regret decomposition of equation 6.5. ■

Up to this point this result is rather abstract, but this standardised analysis allows us to instantiate the Dirichlet sampling algorithm on different classes of problems and calibrate the DS index in order to ensure Assumption 6.4 holds and to make $n_k(T)$ explicit. In particular if $n_k(T) = \mathcal{O}(\log T)$, we recover the classical logarithmic regret. In the next section, we present technical results to justify calibrations of the DS index for several kind of families.

**Technical tool: boundary crossing probability.** We highlight key properties of sums of random variables reweighted by Dirichlet weight vectors that help us suggest a sound tuning of the exploration bonus $B$ used in Definition 6.1 for different kind of families.

**Definition 6.6** (Boundary crossing probability (BCP)). *We define the **boundary crossing probability** (BCP) as*

$$\begin{aligned}[BCP]\colon \mathbb{R}^{(\mathbb{N})} \times \mathbb{R} &\longrightarrow [0,1] \\ (\mathbb{Y};\mu) &\longmapsto \mathbb{P}_{W\sim\mathcal{D}_{n+1}}\left(\sum_{i=1}^{n+1} W_i y_i \geqslant \mu\right), \end{aligned} \tag{6.13}$$

*where $\mathbb{Y} = (y_i)^{n+1}$ and $\mathcal{D}_{n+1}$ is the Dirichlet distribution with parameter $(1,\dots,1)$ of size $n+1 \in \mathbb{N}^\star$, i.e. the uniform distribution on the $n$-simplex $\mathfrak{D}^n$.*

We emphasise that here that the sequence $\mathbb{Y}$ is considered fixed, and the only source of randomness comes from the Dirichlet weights $W$. For this reason, we denote the elements of $\mathbb{Y}$ by the lower case letters $(y_i)_{i=1}^{n+1}$, as opposed to the upper case letters used for random rewards. In the context of DS algorithms, at round $r+1$, for challenger $k$ and leader $\ell$, the first $n = N_k(r)$ elements correspond to random rewards $(Y_i^k)_{i=1}^n$ from a challenger arm $k$, and the last one corresponds to the added exploration bonus $B(\mathbb{Y}_r^k, \mathbb{Y}_r^\ell)$, so that $[BCP]$ correspond to the conditional probability

$$[BCP]\left(\mathbb{Y}_r^k \sqcup \left(B(\mathbb{Y}_r^k, \mathbb{Y}_r^\ell)\right); \mu\right) = \mathbb{P}\left(\widetilde{\mu}_W(\mathbb{Y}_r^k, \mathbb{Y}_r^\ell) \geqslant \mu \;\middle|\; \mathbb{Y}_r^k, \mathbb{Y}_r^\ell\right). \tag{6.14}$$

This quantity is of much interest as both the growth of $n_k(T)$ and Assumption 6.4 can be derived from respectively upper and lower bounds on $[BCP]$. When all observations are pairwise distinct, a closed-form expression has been derived[1] in Cho and Cho (2001) using geometric arguments on the intersection of the $n$-simplex by hyperplanes:

$$[BCP]\left(\mathbb{Y}_{n+1};\mu\right) = \sum_{i=1}^{n+1} \frac{(y_i-\mu)_+^n}{\prod_{\substack{j=1\\ j\neq i}}^{n+1} y_i - y_j}. \tag{6.15}$$

Unfortunately, bounding this expression is hardly feasible since the denominators are of alternating signs. Instead, Riou and Honda (2020, Lemma 14 and 15) provide lower and upper bounds on the probability directly, resorting to classical properties of the Dirichlet distribution that we recall in Appendix E.2, and complete with additional technical results.

[1] To the best of our knowledge, such a closed formula only exist when the random Dirichlet weights follow $\mathrm{Dir}(\alpha)$ with $\alpha_i \leqslant 1$ for all $i = 1,\dots,n+1$. Here, we only use the case $\alpha_i = 1$ for all $i = 1,\dots,n+1$.

Building on these previous results, we propose two novel bounds on $[BCP]$. The first is a direct adaptation of Riou and Honda (2020, Lemma 15), using the empirical maximum instead of a the known support upper bound.

**Lemma 6.7** (Upper bound on $[BCP]$). *Let $n \in \mathbb{N}$ and $\mathbb{Y}_{n+1} = (y_i)_{i=1}^{n+1} \in \mathbb{R}^{n+1}$. For any $\mu < \max \mathbb{Y}_{n+1}$, we have*

$$[BCP]\left(\mathbb{Y}_{n+1};\mu\right) \leqslant \exp\left(-\max_{\lambda\in[0,1)}\sum_{i=1}^{n+1}\log\left(1-\lambda\frac{y_i-\mu}{\max \mathbb{Y}_{n+1}-\mu}\right)\right)$$
$$\leqslant \exp\left(-(n+1)\mathcal{K}_{\inf}^{\max \mathbb{Y}_{n+1}}(\widehat{\nu}_{\mathbb{Y}_{n+1}};\mu)\right), \tag{6.16}$$

*where we recall that $\widehat{\nu}_{\mathbb{Y}_{n+1}}$ denotes the empirical measure of the finite sequence $\mathbb{Y}_{n+1}$.*

*Proof of Lemma 6.7.* We follow a similar proof to that of Riou and Honda (2020, Lemma 15), which consists in using the Chernoff method to upper bound $[BCP]$ and writing the uniform Dirichlet weights with exponential variables. Indeed, for $n \in \mathbb{N}$, $W \sim \mathcal{D}_{n+1}$ has the same distribution as $\left(R_i/\sum_{j=1}^{n+1} R_j\right)_{i=1}^{n+1}$, where $\boldsymbol{R} = (R_j)_{j=1}^{n+1}$ is an i.i.d. sample of exponential random variables drawn from $\mathcal{E}(1)$. Then for any $\lambda \in [0,1)$, we have that

$$\begin{aligned}
[BCP](\mathbb{Y}_{n+1};\mu) &= \mathbb{P}_{\boldsymbol{R}\sim\mathcal{E}(1)^{\otimes(n+1)}}\left(\sum_{i=1}^{n+1} R_i y_i \geqslant \mu\sum_{j=1}^{n+1} R_j\right)\\
&= \mathbb{P}_{\boldsymbol{R}\sim\mathcal{E}(1)^{\otimes(n+1)}}\left(\sum_{i=1}^{n+1} R_i\left(y_i-\mu\right) \geqslant 0\right)\\
&= \mathbb{P}_{\boldsymbol{R}\sim\mathcal{E}(1)^{\otimes(n+1)}}\left(\sum_{i=1}^{n+1} R_i\frac{y_i-\mu}{\max \mathbb{Y}_{n+1}-\mu} \geqslant 0\right) \qquad (\max \mathbb{Y}_{n+1}-\mu>0)\\
&= \mathbb{P}_{\boldsymbol{R}\sim\mathcal{E}(1)^{\otimes(n+1)}}\mathbb{P}\left(\exp\left(\lambda\sum_{i=1}^{n+1} R_i\frac{y_i-\mu}{\max \mathbb{Y}_{n+1}-\mu}\right)\geqslant 1\right)\\
&\leqslant \prod_{i=1}^{n+1}\mathbb{E}_{R_i\sim\mathcal{E}(1)}\left[\exp\left(\lambda R_i\frac{y_i-\mu}{\max \mathbb{Y}_{n+1}-\mu}\right)\right],
\end{aligned} \tag{6.17}$$

where the last line comes from Markov's inequality. Since $\lambda \in [0,1)$, each term is well-defined and has an explicit formula (moment generating function of $\mathcal{E}(1)$), hence

$$
\begin{aligned}
[BCP](\mathbb{Y}_{n+1};\mu) &\leqslant \prod_{i=1}^{n+1} \frac{1}{1-\lambda\frac{y_i-\mu}{\max \mathbb{Y}_{n+1}-\mu}} \\
&= \exp\left(-\sum_{i=1}^{n+1} \log\left(1-\lambda \frac{y_i-\mu}{\max \mathbb{Y}_{n+1}-\mu}\right)\right). \qquad (6.18)
\end{aligned}
$$

We obtain the first inequality of the lemma by choosing the maximum over all possible values of $\lambda$, and the second inequality is direct when writing the dual problem associated with $\mathcal{K}_{\inf}^{\max \mathbb{Y}_{n+1}}(\widehat{\nu}_{\mathbb{Y}_{n+1}};\mu)$ (e.g. Honda and Takemura (2010, 2015)). ■

If $\mathbb{Y}_{n+1}$ is bounded by $\overline{B} \in \mathbb{R}$, replacing $\max \mathbb{Y}_{n+1}$ by $\overline{B}$ effectively makes the $\mathcal{K}_{\inf}$ term data-independent. This formulation allows us to consider sets $\mathbb{Y}_{n+1}$ generated as samples of unbounded distributions. The second result introduces a lower bound on $[BCP]$ that precisely depends on the growth rate of the maximum of $\mathbb{Y}_{n+1}$.

**Lemma 6.8** (Lower bound on $[BCP]$). *Let $n \in \mathbb{N}$ and $\mathbb{Y}_{n+1} = (y_i)_{i=1}^{n+1} \in \mathbb{R}^{n+1}$ such that $\max \mathbb{Y}_{n+1}$ is unique. For any $\mu < \max \mathbb{Y}_{n+1}$, we have*

$$
[BCP](\mathbb{Y}_{n+1};\mu) \geqslant \exp\left(-n\frac{\widehat{\Delta}^{+}(\mathbb{Y}_{n+1};\mu)}{\max \mathbb{Y}_{n+1}-\mu}\right), \qquad (6.19)
$$

*where we defined the* empirical excess gap *mapping as*

$$
\begin{aligned}
\widehat{\Delta}^{+} : \mathbb{R}^{(\mathbb{N})} \times \mathbb{N} &\longrightarrow \mathbb{R} \\
(\mathbb{Y};\mu) &\longmapsto \frac{1}{|\mathbb{Y}|} \sum_{\substack{y\in\mathbb{Y} \\ y<\max \mathbb{Y}}} (\mu-y)_{+}. \qquad (6.20)
\end{aligned}
$$

*Proof of Lemma 6.8.* The idea here is to truncate all the observations that are larger than the threshold $\mu$ except the maximum of $\max \mathbb{Y}_{n+1}$, in order to make all but a single term vanish in the closed-form expression of $[BCP]$ (equation 6.15). The difficulty however is that this formula only holds if all elements of $\mathbb{Y}_{n+1}$ are pairwise distinct, which is impossible if at least two of them are truncated to $\mu$.[2] To circumvent this, let us assume (without loss of generality) that $\max \mathbb{Y}_{n+1} = y_{n+1}$ and define $\widetilde{y}_i^q = y_i \wedge \mu - \varepsilon_i^q$ for $i \in \{1, \ldots, n\}$, $q \geqslant 1$ and $\varepsilon_i^q = \frac{1}{iq} \min_{j \neq i} |y_i \wedge \mu - y_j \wedge \mu|$. These new variables are pairwise distinct, as otherwise if we had $\widetilde{y}_i^q = \widetilde{y}_j^q$ for some $i \neq j$, then $|y_i \wedge \mu - y_j \wedge \mu| = |\varepsilon_i^q - \varepsilon_i^q| < \frac{1}{q} |\frac{1}{i} - \frac{1}{j}| \min_{j \neq i} |y_i \wedge \mu - y_j \wedge \mu|$, which would be contradictory since $\frac{1}{q} |\frac{1}{i} - \frac{1}{j}| < 1$. Going back to $[BCP]$, because of the truncation, the boundary crossing event with $\widetilde{\mathbb{Y}}_{n+1} = (\widetilde{y}_i^q)_{i=1}^n \sqcup (y_{n+1})$ is less likely than with $\mathbb{Y}_{n+1}$, and thus the following inequalities hold:

$$
\begin{aligned}
[BCP] \left( \mathbb{Y}_{n+1}; \mu \right) &\geqslant [BCP] \left( \widetilde{\mathbb{Y}}_{n+1}; \mu \right) \\
&= \frac{(y_{n+1} - \mu)^n}{\prod_{i=1}^{n} y_{n+1} - \widetilde{y}_i^q} \\
&= \exp \left( - \sum_{i=1}^{n} \log \left( \frac{y_{n+1} - \widetilde{y}_i^q}{y_{n+1} - \mu} \right) \right) \\
&= \exp \left( - \sum_{i=1}^{n} \log \left( 1 + \frac{\mu - \widetilde{y}_i^q}{y_{n+1} - \mu} \right) \right) \\
&\geqslant \exp \left( - \sum_{i=1}^{n} \frac{\mu - \widetilde{y}_i^q}{y_{n+1} - \mu} \right) \qquad (\forall x \in \mathbb{R}_+^\star,\ \log(1+x) < x) \\
&= \exp \left( - \sum_{i=1}^{n} \frac{(\mu - \widetilde{y}_i^q)_+}{y_{n+1} - \mu} \right) .
\end{aligned} \tag{6.21}
$$

[2] This technical detail was overlooked in the original paper, we provide the full derivation here.

Letting $q \to +\infty$ results in $\widetilde{y}_i^q \to y_i$ for all $i \in \{1, \ldots, n\}$, which yields the result. ∎

In particular, we see in this expression that the growth rate of the maximum sample may hinder the exponentially decaying lower bound on $[BCP]$. As it turns out, this exponential decay is a crucial condition for the optimality of DS algorithms, as it allows to upper bound the sum in Assumption 6.4, where $[BCP]$ appears in the denominator.

**Remark 6.9** .*When $\mathbb{Y}_{n+1}$ is a realisation of a sequence of i.i.d. random variables, the growth rate of $\max \mathbb{Y}_{n+1}$ is directly related to the tail behaviour of their distribution. If it is upper bounded by $\overline{B}$, $\max \mathbb{Y}_{n+1}$ is upper bounded independently of $n$, and we thus recover the lower bound of Riou and Honda (2020). If it is light tailed, we recall from Boucheron et al. (2013) that $\max \mathbb{Y}_{n+1} = \mathcal{O}(\log n)$ with high probability. By contrast, heavy tailed distributions such as the Cauchy distribution, which are outside the scope of this chapter, exhibit a growth rate $\Omega(n)$, altogether preventing the above lower bound to decay to zero.*

**Asymptotic behaviour of the empirical $\mathcal{K}_{\inf}$.** The analysis of $[BCP]$, in particular the upper bound of Lemma 6.7, makes appear by the empirical extremal Kullback-Leibler operator $\mathcal{K}_{\inf}^{\max \mathbb{Y}_{n+1}}$, which we recall is calculated over distributions upper bounded by $\max \mathbb{Y}_{n+1}$ for a given set of observations. However, the regret lower bound of Burnetas and Katehakis (1996) involves $\mathcal{K}_{\inf}^{\mathcal{F}}$, calculated over distributions belonging to a base family $\mathcal{F} \subset \mathbb{L}^1(\mathbb{R})$. The analysis of NPTS (Riou and Honda, 2020) considers only distributions with bounded support, for which both operators can be made arbitrarily close in the relevant Lévy topology. We show that this is essentially the only favourable setting where such manipulations are possible.

**Lemma 6.10** (Asymptotic behaviour of $\mathcal{K}_{\inf}^{\max \mathbb{Y}_{n+1}}$). *Let $(Y_n)_{n\in\mathbb{N}} \in \mathbb{R}^{\mathbb{N}}$ an i.i.d. sequence of random variables drawn from $\nu \in \mathbb{L}^1(\mathbb{R})$ and $\mathbb{Y}_{n+1} = (Y_i)_{i=1}^{n+1}$ for all $n \in \mathbb{N}$. Then for all $\mu \in \mathbb{R}$,*

$$\mathcal{K}_{\inf}^{\max \mathbb{Y}_{n+1}}\left(\widehat{\nu}_{\mathbb{Y}_{n+1}}; \mu\right) = \mathcal{O}_{a.s.}\left(\frac{1}{\max \mathbb{Y}_{n+1}}\right). \tag{6.22}$$

*Proof of Lemma 6.10.* Combining Lemmas 6.7 and 6.8, we have

$$\exp\left(-n\frac{\widehat{\Delta}^+\left(\mathbb{Y}_{n+1};\mu\right)}{\max \mathbb{Y}_{n+1} - \mu}\right) \leqslant [BCP]\left(\mathbb{Y}_{n+1};\mu\right) \leqslant \exp\left(-(n+1)\mathcal{K}_{\inf}^{\max \mathbb{Y}_{n+1}}(\widehat{\nu}_{\mathbb{Y}_{n+1}};\mu)\right). \tag{6.23}$$

Taking the $\log$ on both sides yields

$$\mathcal{K}_{\inf}^{\max \mathbb{Y}_{n+1}}(\widehat{\nu}_{\mathbb{Y}_{n+1}};\mu) \leqslant \frac{n}{n+1}\frac{\widehat{\Delta}^+\left(\mathbb{Y}_{n+1};\mu\right)}{\max \mathbb{Y}_{n+1} - \mu}. \tag{6.24}$$

To conclude, note that the empirical excess gap converges almost surely when $n \to +\infty$ since the measure $\nu$ is integrable (law of large numbers). ∎

In particular, if the distribution $\nu$ is unbounded, then $\max \mathbb{Y}_{n+1} \to +\infty$ (the sequence $(\max \mathbb{Y}_{n+1})_{n\in\mathbb{N}}$ is almost surely nondecreasing and cannot converge, otherwise $\nu$ would be bounded), and thus, $\mathcal{K}_{\inf}^{\max \mathbb{Y}_{n+1}}\left(\widehat{\nu}_{\mathbb{Y}_{n+1}};\mu\right) \to 0$ almost surely. Therefore, the $\mathcal{K}_{\inf}$ operator cannot be continuous with respect to the family over which it is defined in the sense that the following assertions are mutually exclusive:

(i) $\mathcal{F}$ contains an unbounded distribution $\nu$ and $\mathcal{K}_{\inf}^{\mathcal{F}}(\nu;\mu) > 0$,

(ii) $\mathcal{K}_{\inf}^{\max \mathbb{Y}_{n+1}}\left(\widehat{\nu}_{\mathbb{Y}_{n+1}};\mu\right) \longrightarrow \mathcal{K}_{\inf}^{\mathcal{F}}(\nu;\mu)$.

A direct consequence is that a direct generalisation of NPTS to unbounded distributions is impossible while preserving logarithmic regret, forcing us to either let go of logarithmic guarantees or alter the measure $\nu$ to recover the continuity between the empirical $\mathcal{K}_{\text{inf}}^{\max \mathbb{Y}_{n+1}}$ and $\mathcal{K}_{\text{inf}}^{\mathcal{F}}$. We show examples of these approaches in the next section.

**Empirical validation.** Note that the above result does not give any information on the tightness of the upper bound $\mathcal{K}_{\text{inf}}^{\max \mathbb{Y}_{n+1}}(\widehat{\nu}_{\mathbb{Y}_{n+1}}; \mu) = \mathcal{O}_{\text{a.s.}}(1/\max \mathbb{Y}_{n+1})$. To investigate this, we conducted a numerical experiment where (i) we calculated $\mathcal{K}_{\text{inf}}^{\mathcal{F}}$ for different choices of $\mathcal{F}$ among classical SPEF (Gaussian with fixed variance, exponential and Bernoulli), using the explicit formula for their one-dimensional KL divergence (see Chapter 1), and (ii) simulated $\mathcal{K}_{\text{inf}}^{\max \mathbb{Y}_{n+1}}$ for various sample sizes $n$, using the dual formulation of the optimisation problem for upper bounded families (equation (1.24) in Chapter 1). Precisely, we generated sequences $(Y_i^m)_{i=1}^{n+1}$ for $m \in \{1, \dots, M\}$ with $M = 10^4$, and then calculated $\mathcal{K}_{\text{inf}}^{\max \mathbb{Y}_{n+1}^m}(\widehat{\nu}_{\mathbb{Y}_{n+1}^m}; \mu)$ for standard values of $\nu$ and $\mu$, where $\mathbb{Y}_{n+1}^m = (Y_i^m)_{i=1}^{n+1}$. We report in Figure 6.1 the densities fitted to the $M$ realisations of the empirical $\mathcal{K}_{\text{inf}}$ operator, as well as their mean $1/M \sum_{m=1}^{M} \mathcal{K}_{\text{inf}}^{\max \mathbb{Y}_{n+1}^m}(\widehat{\nu}_{\mathbb{Y}_{n+1}^m}; \mu)$.

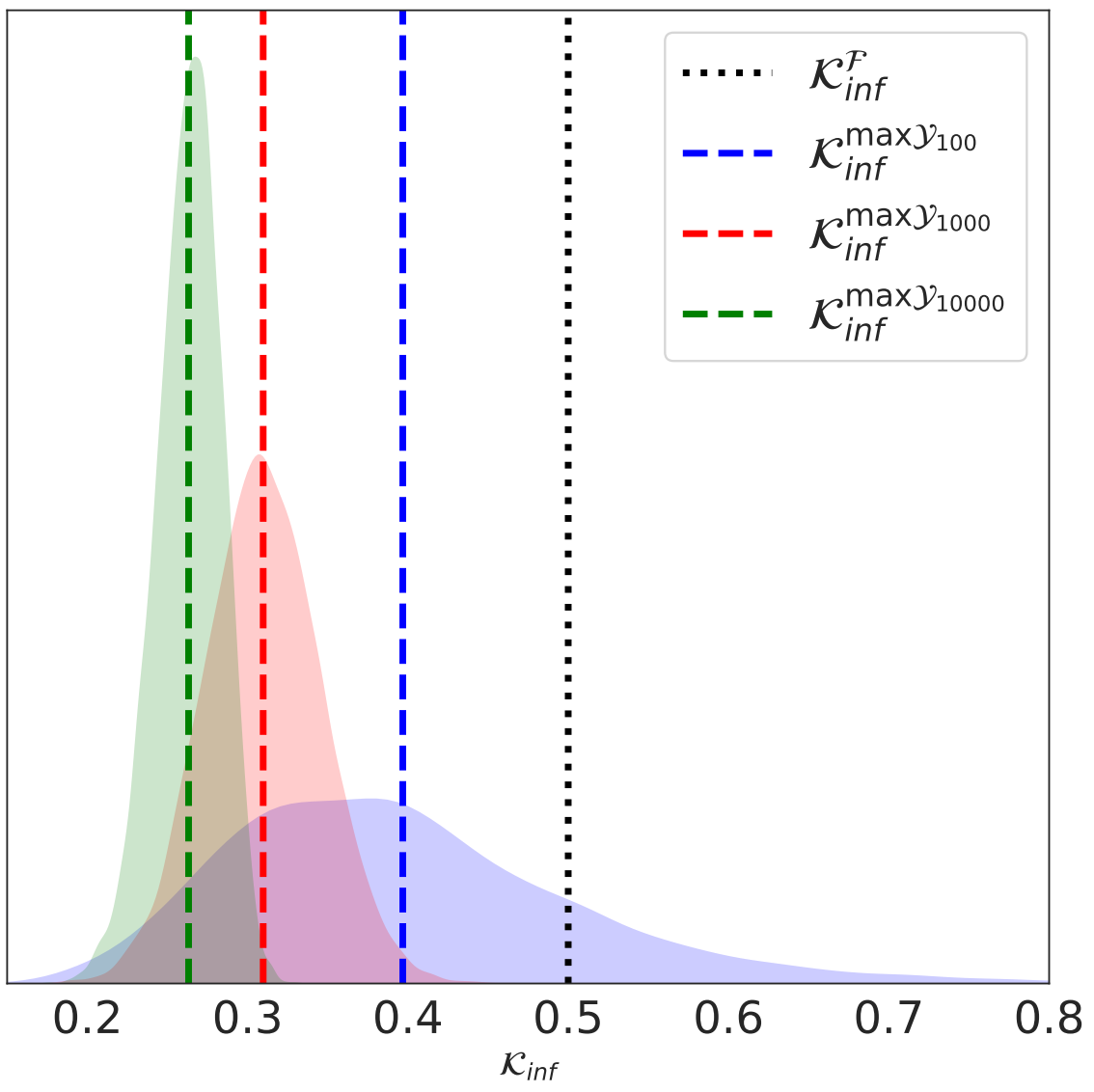

(**a**) Gaussian.

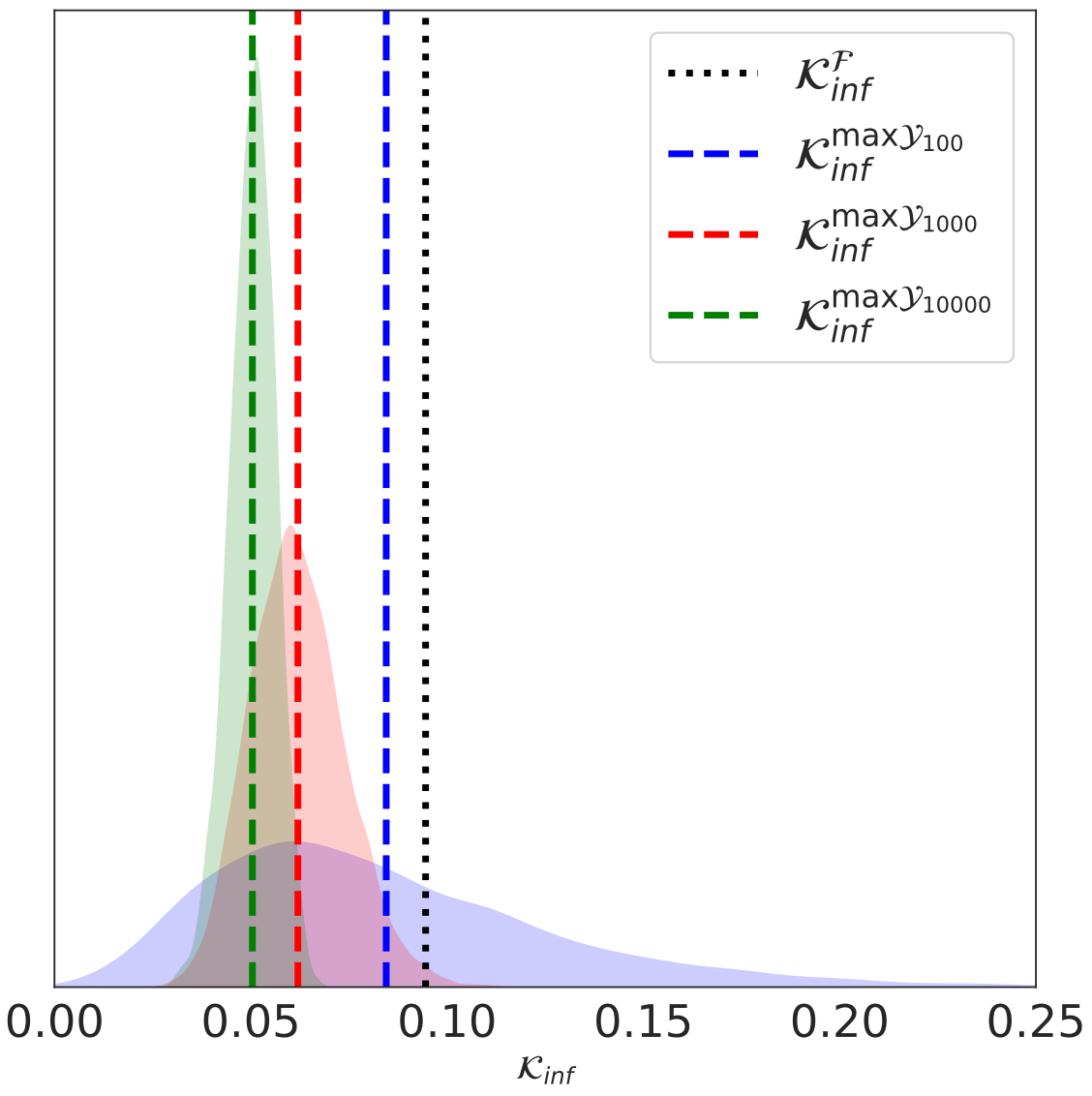

(**b**) Exponential.

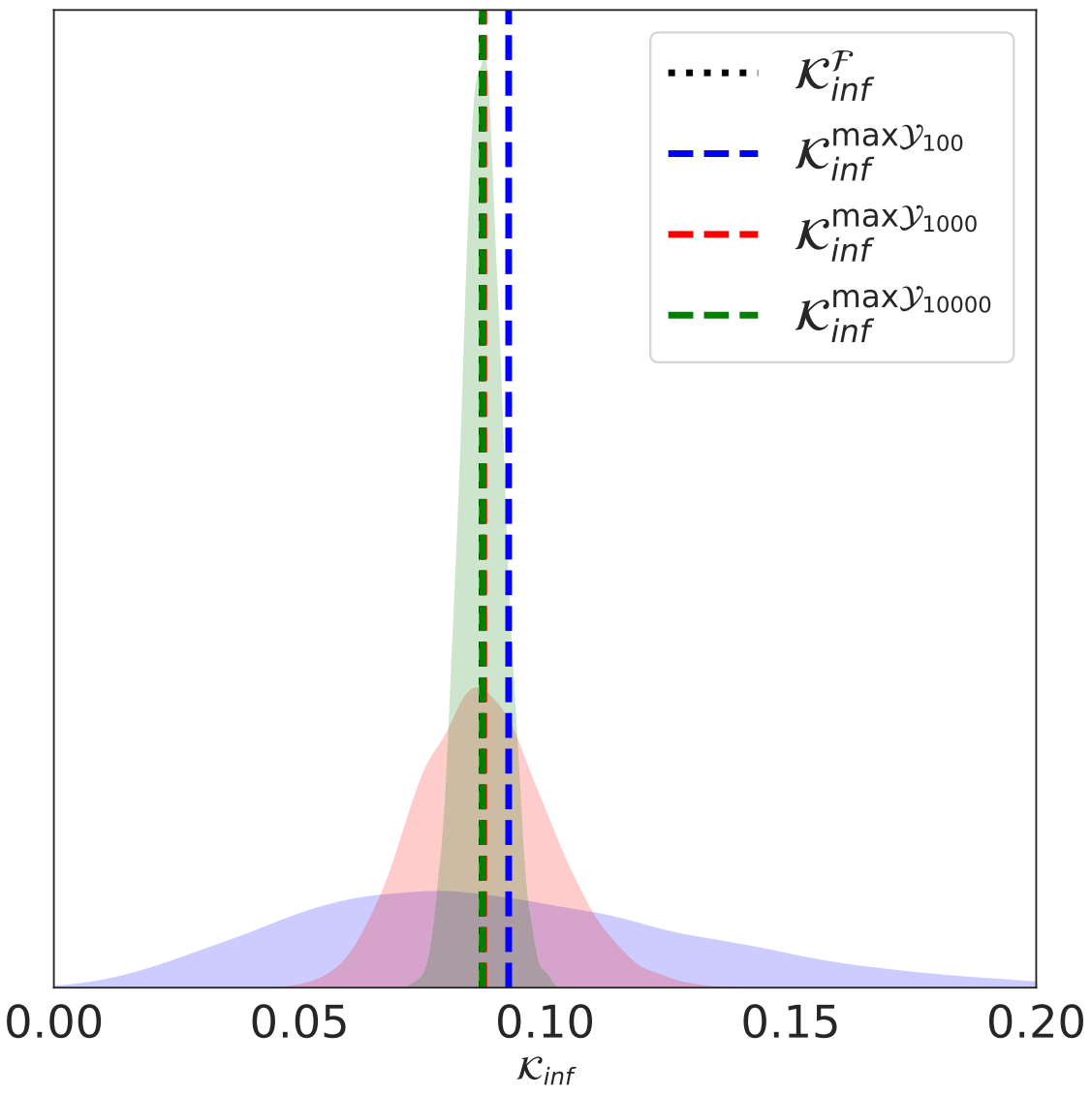

(**c**) Bernoulli.

**Figure 6.1** – Dotted line: $\mathcal{K}_{\text{inf}}^{\mathcal{F}}(\nu;\mu)$ with respect to a given SPEF. Dashed lines: empirical $\mathcal{K}_{\text{inf}}^{\max \mathbb{Y}_{n+1}}(\widehat{\nu}_{\mathbb{Y}_{n+1}};\mu)$ for sample size $n = 10^2, 10^3, 10^4$, averaged over $10^4$ independent replicates (fitted densities are shown in the background). Top left: Gaussian with fixed variance, $\nu = \mathcal{N}(2,1)$, $\mu = 3$. Top right: exponential, $\mathcal{E}(\frac{1}{2})$, $\mu = 3$. Bottom: Bernoulli, $\nu = \mathcal{B}(\frac{1}{2})$, $\mu = 0.7$.

For the unbounded SPEF (Figures 6.1a and 6.1b), we observed that the empirical $\mathcal{K}_{\text{inf}}$ decreases away from $\mathcal{K}_{\text{inf}}^{\mathcal{F}}(\nu;\mu)$ with increasing sample sizes $n \in \mathbb{N}$, which is consistent with the bound $\mathcal{O}(1/\max_{\mathbb{Y}_{n+1}})$. By contrast, the Bernoulli distribution (Figure 6.1c) showed a concentration pattern around $\mathcal{K}_{\text{inf}}^{\mathcal{F}}(\nu;\mu)$.

Furthermore, we empirically validated the relation between $\mathbb{E}\left[\mathcal{K}_{\text{inf}}^{\max \mathbb{Y}_{n+1}}\right]$ (estimated by $1/M \sum_{m=1}^{M} \mathcal{K}_{\text{inf}}^{\max \mathbb{Y}_{n+1}^m}(\widehat{\nu}_{\mathbb{Y}_{n+1}^m}; \mu))$ and the running maximum $\max \mathbb{Y}_{n+1}$ as a function of $n \in \mathbb{N}$. Indeed, we have $\max \mathbb{Y}_{n+1} = \Theta(\log n)$ for exponential distributions and $\max \mathbb{Y}_{n+1} = \Theta(\sqrt{\log n})$ for Gaussian distributions, and therefore we expect $\log \mathcal{K}_{\text{inf}}^{\max \mathbb{Y}_{n+1}}(\widehat{\nu}_{\mathbb{Y}_{n+1}}; \mu) \approx -\log\log n$ and $\log \mathcal{K}_{\text{inf}}^{\max \mathbb{Y}_{n+1}}(\widehat{\nu}_{\mathbb{Y}_{n+1}}; \mu) \approx -\frac{1}{2}\log\log n$ respectively. Figure 6.2 shows the outcome of a linear model (with intercept) for the regression of $\log \mathcal{K}_{\text{inf}}^{\max \mathbb{Y}_{n+1}}(\widehat{\nu}_{\mathbb{Y}_{n+1}}; \mu)$ on $\log\log n$, which confirms the linear trend ($R^2 = 1.00$ for both Gaussian and exponential SPEF) and approximately recovers the expected slopes (although $-1/2$ and $-1$ lie both outside the estimated $95\%$ confidence intervals for the slopes of the Gauss and exponential SPEF respectively, indicating that there may be an additional factor at play). Again, the case of the Bernoulli SPEF shows no significant dependency on $n$ as $\max \mathbb{Y}_{n+1} \to 1$.

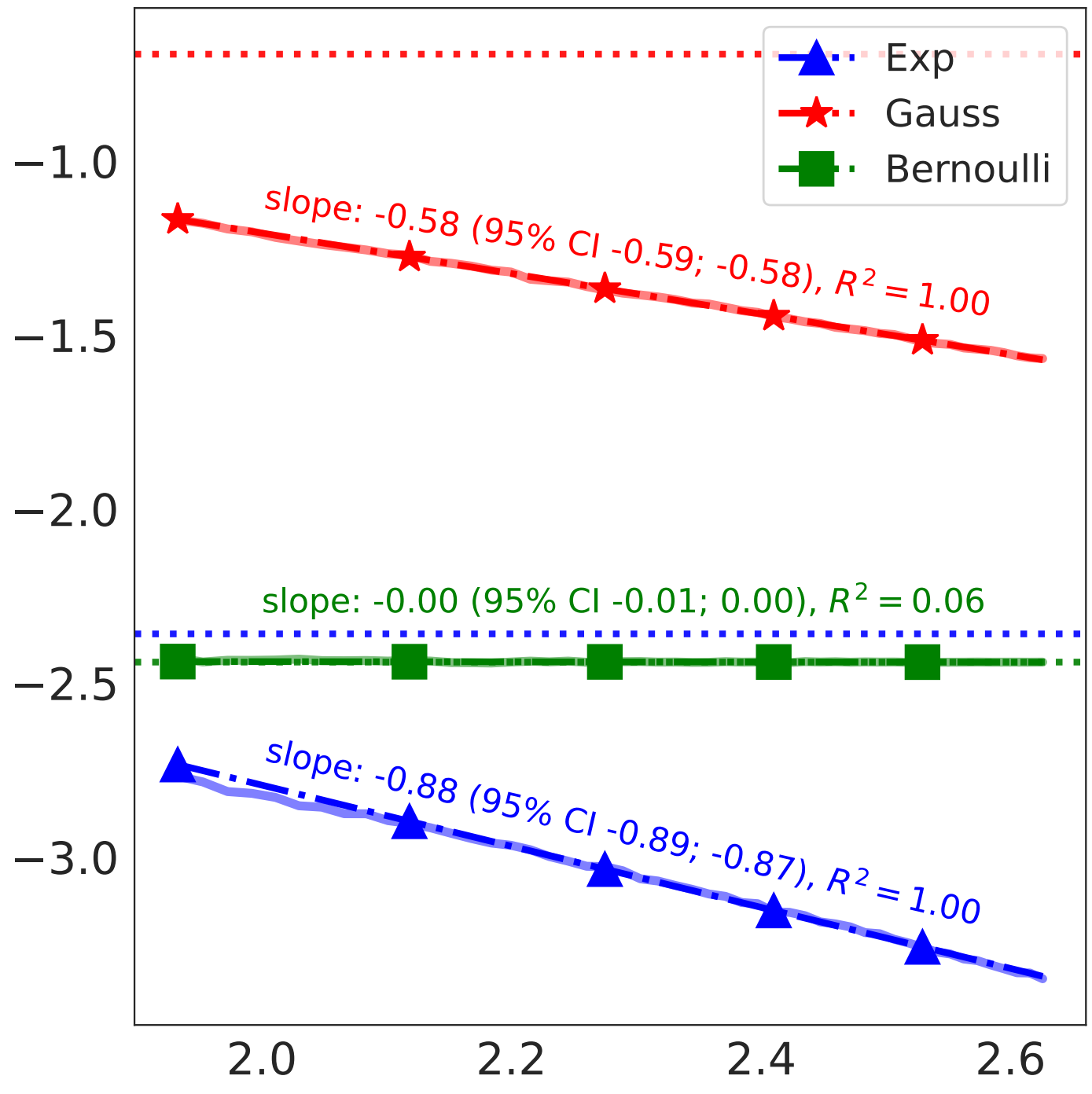

**Figure 6.2** – Linear fit (dashed) of $\log \mathcal{K}_{\text{inf}}^{\max \mathbb{Y}_{n+1}}(\widehat{\nu}_{\mathbb{Y}_{n+1}}; \mu)$ (solid) on $\log\log n$ and resulting slopes. Dotted lines correspond to $\log \mathcal{K}_{\text{inf}}^{\mathcal{F}}(\nu; \mu)$ for the corresponding SPEF $\mathcal{F}$. X-axis: $\log\log n$. Y-axis: $\log \mathcal{K}_{\text{inf}}$.

**Canonical exploration bonus.** The lower bounds suggest a natural tuning of the bonus. Indeed, inspecting the exponential term in the lower bound of Lemma 6.8 reveals that if $\max \mathbb{Y}_{n+1} - \mu$ is of the order of the empirical excess gap, then the exponential decay is preserved. We formalise this intuition in Corollary 6.12 below.

**Definition 6.11** (Canonical exploration bonus). *We define the following bonus mapping:*

$$\begin{aligned} B\colon \mathbb{R}^{(\mathbb{N})} \times \mathbb{R} \times \mathbb{R}_+^\star &\longrightarrow \mathbb{R} \\ (\mathbb{Y}; \mu, \rho) &\longmapsto \mu + \frac{\rho}{|\mathbb{Y}|} \sum_{y \in \mathbb{Y}} (\mu - y)_+ \,. \end{aligned} \tag{6.25}$$

We interpret $\rho$ as the *leverage* of the empirical excess gap. It is a free parameter that balances the importance of the exploration bonus and the decay of the $[BCP]$ lower bound.

**Corollary 6.12** (Lower bound on $[BCP]$ with canonical bonus). *Let $n \in \mathbb{N}$, $\mathbb{Y}_n \in \mathbb{R}^n$ and $\rho_n > 0$. Define $Y_{n+1} = \max\left(B(\mathbb{Y}_n; \mu, \rho_n), \max \mathbb{Y}_n\right)$, and $\mathbb{Y}_{n+1} = \mathbb{Y}_n \sqcup (Y_{n+1})$. For all $\mu \in \mathbb{N}$, we have*

$$[BCP]\,(\mathbb{Y}_{n+1}; \mu) \geqslant \exp\left(-\frac{n}{\rho_n}\right) \,. \tag{6.26}$$

*Proof of Corollary 6.12.* By construction of $B$, we have, for $n \in \mathbb{N}$ and $\mathbb{Y}_n = (y_i)_{i=1}^n$,

$$\max \mathbb{Y}_{n+1} \geqslant \mu + \frac{\rho_n}{n} \sum_{i=1}^n (\mu - y_i)_+ = \mu + \rho_n \widehat{\Delta}_n^+(\mathbb{Y}_{n+1}; \mu) \,, \tag{6.27}$$

and thus in particular,

$$\exp\left(-\frac{\widehat{\Delta}_n^+(\mathbb{Y}_{n+1}; \mu)}{\max \mathbb{Y}_{n+1} - \mu}\right) \geqslant e^{-\frac{1}{\rho_n}} \,. \tag{6.28}$$

We conclude by plugging this inequality in the expression of Lemma 6.8. ∎

**Remark 6.13**.*In practice for DS algorithms, at round $r+1$ and for challenger $k$ and leader $\ell$, we set the $(N_r^k+1)$-th sample to $B\left(\mathbb{Y}_r^k; \widehat{\mu}_r^\ell, \rho_{N_r^k}\right)$, without the additional maximum over $\mathbb{Y}_r^k$. As we will see for the* RDS *algorithm, we can choose a sequence $(\rho_n)_{n\in\mathbb{N}}$ that goes to $+\infty$ at an appropriate rate to ensure that, with high probability, the exploration bonus is indeed larger than all previous rewards $\mathbb{Y}_r^k$. For other algorithms, choosing a constant sequence $\rho_n = \rho$ for all $n \in \mathbb{N}$ will be sufficient.*

The analysis of DS algorithms then comes down to tuning the leverage sequence $(\rho_n)_{n\in\mathbb{N}}$, assuming weak distributional hypotheses for a given class of distribution. In the next section we discuss three examples of such DS algorithms and their theoretical regret guarantees.

## 6.3 Theoretical guarantees of three DS instances

Building on the results from previous the section, we now instantiate the DS algorithms for three bandits problems. We first prove that optimal guarantees can be derived for DS with bounded distributions under a nonstandard definition of the problem, motivated by practical considerations. Then, we consider a natural extension to unbounded distributions using a simple truncation mechanism, ensuring logarithmic regret under assumptions on some quantile of the distributions. Finally, we consider a simple DS algorithm securing slightly larger than logarithmic regret for the entire family of *light tailed distributions*. In the rest of this chapter, we use the DS index of Definition 6.1 (except for QDS, where we slightly amend this definition), namely, at round $r+1$ with challenger $k$ and leader $\ell$, conditional on $\mathbb{Y}_r^k = (Y_i^k)_{i=1}^{N_r^k}$ and $\mathbb{Y}_r^\ell$,

$$\widetilde{\mu}_W\left(\mathbb{Y}_r^k; \widehat{\mu}_r^\ell\right) = \sum_{i=1}^{N_r^k} W_i Y_i^k + W_{N_r^k+1} B\left(\mathbb{Y}_r^k; \widehat{\mu}_r^\ell, \rho_{N_r^k}^k\right) , \tag{6.29}$$

where $W \sim \mathcal{D}_{N_r^k+1}$ and $B$ is the canonical bonus of Definition 6.11 with leverages $(\rho_n^k)_{n\in\mathbb{N}}$, i.e.

$$B\left(\mathbb{Y}_r^k; \widehat{\mu}_r^\ell, \rho_{N_r^k}^k\right) = \widehat{\mu}_r^\ell + \frac{\rho_{N_r^k}^k}{N_r^k} \sum_{i=1}^{N_r^k} \left(\widehat{\mu}_r^\ell - Y_i^k\right)_+ . \tag{6.30}$$

We detail below each DS instance and defer the proofs of the three regret theorems in Appendix E.3. In all cases, the proof consists in showing that Assumptions 6.2 and 6.4 hold for each proposed algorithms in their respective settings and deriving an expression for $n_k(T)$.

**Optimality for bounded distributions.** We start by defining subsets of interests of the set of bounded distributions over $\mathbb{R}$ and instantiate the canonical bonus function $B$ of Definition 6.11 to these families.

**Definition 6.14** (Bounded Dirichlet sampling). *Let $\underline{B}, \overline{B} \in \mathbb{R}$ such that $\underline{B} < \overline{B}$, $p \in [0,1]$ and $\gamma \in [0, \overline{B} - \underline{B}]$. We recall that $\mathcal{F}_{[\underline{B},\overline{B}]}$ is the set of bounded distributions supported in $[\underline{B}, \overline{B}]$ and define the subset of such distributions with at least $p$ mass in a neighbourhood of $\overline{B}$ of size $\gamma$:*

$$\mathcal{F}^{\gamma,p}_{[\underline{B},\overline{B}]} = \left\{ \nu \in \mathcal{F}_{[\underline{B},\overline{B}]},\ \mathbb{P}_\nu\left( [\overline{B} - \gamma, \overline{B}] \right) \geqslant p \right\} . \tag{6.31}$$

*Then we define the* ***bounded Dirichlet sampling*** (BDS) *bonus associated with a distribution $\nu$ as*

$$\begin{aligned} B^{\mathrm{BDS}} \colon \mathbb{R}^{(\mathbb{N})} \times \mathbb{R} \times \mathbb{R}_+ &\longrightarrow \mathbb{R} \\ (\mathbb{Y}; \mu, \rho) &\longmapsto \begin{cases} \overline{B} & \text{if } \nu \in \mathcal{F}_{[\underline{B},\overline{B}]} \text{ with known } \overline{B}\,, \\ (\max \mathbb{Y} + \gamma) \vee (B(\mathbb{Y}; \mu, \rho)) & \text{if } \nu \in \mathcal{F}^{\gamma,p}_{[\underline{B},\overline{B}]} \text{ with unknown } \overline{B} \\ & \text{but known } p, \gamma\,. \end{cases} \end{aligned} \tag{6.32}$$

Plugging the BDS index for $\mathcal{F}_{[\underline{B},\overline{B}]}$ in Algorithm 5 makes it essentially equivalent to NPTS (Riou and Honda, 2020) (up to the round-based structure, which is not a fundamental change since this particular index is constant for all pairs of challenger and leader). Crucially, the support upper bound $\overline{B}$ (but not $\underline{B}$) appears in the definition of the index, and therefore must be known to the agent beforehand. The BDS index for $\mathcal{F}^{\gamma,p}_{[\underline{B},\overline{B}]}$ covers a novel setting where neither bounds $\underline{B}$ and $\overline{B}$ are known precisely but $\overline{B}$ is *detectable*, in the sense that observing samples in its neighbourhood occurs with sufficient probability. Circling back to the intuitive notion of "mass leakage at infinity" discussed in the introduction, this condition excludes for instance distributions of the form $(1-\varepsilon)\nu + \varepsilon\delta_{y_\varepsilon}$ with $\nu \in \mathcal{F}_{[\underline{B},\overline{B}-\gamma]}$, $\varepsilon \in [0,1]$ arbitrarily small and $y_\varepsilon \in \mathbb{R}$ arbitrarily large. We provide in Section 6.4 examples of applications of this detectable setting.

Consider a $K$-armed bandit measure $\boldsymbol{\nu} = (\nu_k)_{k\in[K]}$, and $(\underline{B}_k)_{k\in[K]} \in \mathbb{R}^K$ and $(\overline{B}_k)_{k\in[K]} \in \mathbb{R}^K$. We refer to the bandit model where the rewards of arm $k \in [K]$ are bounded between (a possibly unknown) $\underline{B}_k$ and a known $\overline{B}_k$ as

$$\textbf{Bandit model (B1):} \quad \boldsymbol{\nu} \in \bigotimes_{k\in[K]} \mathcal{F}_{[\underline{B}_k,\overline{B}_k]} \,. \tag{6.33}$$

Additionally, for $(p_k)_{k\in[K]} \in [0,1]^K$ and $(\gamma_k)_{k\in[K]} \in \prod_{k\in[K]} [0, \overline{B}_k - \underline{B}_k]$, we refer to the bandit model where the bounds $(\overline{B}_k)_{k\in[K]} \in \mathbb{R}^K$ are unknown but detectable as

$$\textbf{Bandit model (B2):} \quad \boldsymbol{\nu} \in \bigotimes_{k\in[K]} \mathcal{F}^{\gamma_k, p_k}_{[\underline{B}_k, \overline{B}_k]} . \tag{6.34}$$

**Theorem 6.15** (Optimality of BDS). *Algorithm 5 with* BDS *index satisfies:*

(*i*) *in bandit model (B1),*

$$\mathbb{E}[N_T^k] \leqslant \frac{\log T}{\mathcal{K}_{\inf}^{\overline{B}_k}(\nu_k; \mu^\star)} + O(1)\,; \tag{6.35}$$

(*ii*) *in bandit model (B2), for any suboptimal arm* $k \in [K] \setminus \{k^\star\}$ *and* $\rho^k \geqslant -1/\log(1-p_k)$,

$$\mathbb{E}[N_T^k] \leqslant \frac{\log T}{\mathcal{K}_{\inf}^{\mathfrak{B}_k}(\nu_k; \mu^\star)} + O(1)\,, \tag{6.36}$$

*where* $\mathfrak{B}_k = \Big(\overline{B}_k + \gamma_k\Big) \vee \Big(\mu^\star + \rho^k \mathbb{E}_{Y\sim\nu_k}[(\mu^\star - Y)_+]\Big)$.

**Unbounded distributions: truncating the upper tail.** Let consider the family $\mathcal{F}_{[\underline{B},+\infty)}$ for some unknown $\underline{B} \in \mathbb{R}$. A natural way to extend algorithms designed for a distribution in $\mathcal{F}_{[\underline{B},\overline{B}]}$ (where $\overline{B} \in \mathbb{R}$) is to truncate its right tail and summarise it by a single Dirac mass.

**Definition 6.16** (Quantile Dirichlet sampling). *Let* $\alpha \in (0,1)$, $\underline{B} \in \mathbb{R}$ *and* $q_{1-\alpha}$ *the quantile operator at level* $1-\alpha$. *We define the right tail conditional expectation*

$$\begin{aligned} C_\alpha \colon \mathcal{F}_{[\underline{B},+\infty)} &\longrightarrow \mathbb{R} \\ \nu &\longmapsto \mathbb{E}_{Y\sim\nu}\left[Y \mid Y > q_{1-\alpha}(\nu)\right], \end{aligned} \tag{6.37}$$

*and the truncation operator*

$$\begin{aligned} \mathcal{T}_\alpha \colon \mathcal{F}_{[\underline{B},+\infty)} &\longrightarrow \mathcal{F}_{[\underline{B},+\infty)} \\ \nu &\longmapsto \nu|_{[\underline{B},q_{1-\alpha}(\nu)]} + \alpha\delta_{C_\alpha(\nu)}\,. \end{aligned} \tag{6.38}$$

*Then we define the* ***quantile Dirichlet sampling*** (QDS) *bonus as*

$$\begin{aligned} B^{\mathrm{QDS}} \colon \mathbb{R}^{(\mathbb{N})} \times \mathbb{R} \times \mathbb{R}_+ &\longrightarrow \mathbb{R} \\ (\mathbb{Y};\mu,\rho) &\longmapsto B(\operatorname{Supp} \mathcal{T}_\alpha(\widehat{\nu}_{\mathbb{Y}});\mu,\rho)\,. \end{aligned} \tag{6.39}$$

In other words, for $\nu \in \mathcal{F}_{[\underline{B},+\infty)}$, the truncation operator $\mathcal{T}_\alpha$ (i) does not change $\nu$ below its $1-\alpha$-quantile, and (ii) "summarises" its $\alpha$-right tail by concentrating it into a single point $C_\alpha(\nu)$, chosen so that the truncation preserves the total expectation (when $\nu$ is atomless, $C_\alpha(\nu)$ coincides with the *conditional value at risk* (CVaR) — although, contrary to the definition in Appendix D.1, CVaR is defined here on the right tail rather than the left tail).

Consider a $K$-armed bandit measure $\boldsymbol{\nu} = (\nu_k)_{k\in[K]}$, $(\underline{B}_k)_{k\in[K]} \in \mathbb{R}^K$, $(\alpha_k)_{k\in[K]} \in (0,1)^K$ and $(\rho^k)_{k\in[K]} \in \mathbb{R}_+^K$. We define the following setting to analyse the regret of the QDS instance:

$$\textbf{Bandit model (Q):}\quad \boldsymbol{\nu} \in \bigotimes_{k\in[K]} \mathcal{F}_{\underline{B}_k}^{\alpha_k}\,, \tag{6.40}$$

where

$$\mathcal{F}_{\underline{B}_k}^{\alpha_k} = \left\{\nu \in \mathcal{F}_{[\underline{B}_k,+\infty)},\ \forall \mu > \mathbb{E}_{Y\sim\nu_k}[Y],\ \mathcal{K}_{\inf}^{\mathcal{F}_{[\underline{B}_k,+\infty)}}(\nu_k;\mu) \geqslant \mathcal{K}_{\inf}^{\mathfrak{M}_q^k}(\mathcal{T}_{\alpha_k}(\nu_k);\mu)\right\}, \tag{6.41}$$

$$\mathfrak{M}_q^k = (q_{1-\alpha_k}(\nu_k)) \vee \left(\mu^\star + \rho^k \mathbb{E}_{Y\sim\nu_k}[(\mu^\star - Y)_+]\right)\,. \tag{6.42}$$

Although quite technical, this condition essentially states that the bandit problem with arm $k \in [K]$ taken on the complete family of distributions with support bounded from below by $\underline{B}_k$ is no harder than an alternative bandit problem with truncated distributions and an upper bounded family, with an upper bound depending on the $1-\alpha_k$ quantile and the leverage $\rho^k$ of the exploration bonus.

The meaning of the QDS index is the following. At round $r+1$ with challenger $k$ and leader $\ell$, let $Y_{(1)}^k \leqslant \ldots \leqslant Y_{(N_r^k)}^k$ be the order statistics of the reward history $\mathbb{Y}_r^k = (Y_i^k)_{i=1}^{N_r^k}$, as well as $N_r^{k,\alpha_k} = \lceil N_r^k \alpha_k \rceil / N_r^k$. Truncating the top $\alpha_k$-fraction of the samples in $\mathbb{Y}_r^k$ and averaging them results in $C_{\alpha_k}(\widehat{\nu}_r^k) = \frac{1}{N_r^k - N_r^{k,\alpha_k}+1} \sum_{i=N_r^{k,\alpha_k}}^{N_r^k} Y_{(i)}^k$, where we recall that $\widehat{\nu}_r^k = \widehat{\nu}_{\mathbb{Y}_r^k}$ is the empirical measure of $\mathbb{Y}_r^k$. We then slightly amend the definition of the DS index to account for the occurrences of the truncated rewards, i.e.:

$$\widetilde{\mu}_W^{\mathrm{QDS}}\left(\mathbb{Y}_r^k; \widehat{\mu}_r^\ell\right) = \sum_{i=1}^{N_r^{k,\alpha_k}-1} W_i Y_{(i)}^k + W_{N_r^{k,\alpha_k}} C_{\alpha_k}(\widehat{\nu}_r^k) + W_{N_r^{k,\alpha_k}+1} B^{\mathrm{QDS}}\left(\mathbb{Y}_r^k; \widehat{\mu}_r^\ell, \rho^k\right) , \qquad (6.43)$$

where $W \sim \mathrm{Dir}\left((1, \ldots, 1, N_r^{k,\alpha_k}, 1)\right)$. Thanks to the aggregation properties of Dirichlet distributions (see Appendix E.2), this is equivalent to using uniform Dirichlet weights $\mathcal{D}_{N_r^k+1}$ on the sequence $(\underbrace{Y_{(1)}^k, \ldots, Y_{(N_r^{k,\alpha_k})}^k}_{N_r^{k,\alpha_k} \text{ times}}, \underbrace{C_{\alpha_k}(\widehat{\nu}_r^k), \ldots, C_{\alpha_k}(\widehat{\nu}_r^k)}_{N_r^k - N_r^{k,\alpha_k}+1 \text{ times}})$.

**Theorem 6.17** (Logarithmic expected pseudo regret of QDS). *For any $\varepsilon > 0$ small enough, Algorithm 5 with* QDS *index satisfies, in bandit model* (Q)*, for any suboptimal arm $k \in [K] \setminus \{k^\star\}$, $\alpha' < \alpha_k$ and $\rho^k \geqslant (1+\alpha')/\alpha'^2$,*

$$\mathbb{E}[N_T^k] \leqslant \frac{\log T}{\mathcal{K}_{\inf}^{\mathfrak{M}_k^C}(\mathcal{T}_{\alpha_k}(\nu_k), \mu^\star) - \varepsilon} + \mathcal{O}(1) , \qquad (6.44)$$

*with $\mathfrak{M}_C^k = C_{\alpha_k}(\nu_k) \vee \left(\mu^\star + \rho^k \mathbb{E}_{Y\sim\nu_k}[(\mu^\star - Y)_+]\right)$.*

This result is of particular interest as it captures the continuum between bounded and light tailed distributions. In our opinion, it sheds new light on the interpretation of impossibility results of e.g. Ashutosh et al. (2021): logarithmic regret can be achieved *without fully specifying the tail with precise parameters*, as a simple quantile condition is sufficient to avoid pathological distributions. The family $\mathcal{F}_{\underline{B}}^\alpha$ for some $\alpha \in (0,1)$ and $\underline{B} \in \mathbb{R}$ trades off the mathematical convenience of parametric model specification for the ability to capture perhaps more realistic distributions. We further discuss this condition in Appendix E.4 and show that virtually any lower bounded bandit model can satisfy the quantile condition of bandit model (Q), provided the leverages $(\rho^k)_{k\in[K]}$ are large enough (although the exact bound on the leverage is model-dependent and thus of limited practical intersest). We also detail examples of classical families for which this quantile condition holds (exponential, Gaussian).

**Remark 6.18**. *The restriction to the semibounded case* $\underline{B} > -\infty$ *is due to our proof technique, based on a discretisation of the support of the truncated distribution (see Appendix E.3). Note that the actual value of* $\underline{B}$ *does not need to be known at the algorithm runtime, which is intuitive since* $\mathcal{K}_{\inf}^{\mathcal{F}_{(-\infty,\overline{B}]}} = \mathcal{K}_{\inf}^{\mathcal{F}_{[\underline{B},\overline{B}]}}$ *for all* $\underline{B}, \overline{B} \in \mathbb{R}$ *(Honda and Takemura, 2015, Theorem 2). Different theoretical tools, such as quantisation of probability measures (Graf and Luschgy, 2007; Montes, 2020) may unlock the proof of a logarithmic regret for* QDS *in the doubly unbounded case.*

One may wonder whether the quantile condition and the truncation technique are necessary to achieve theoretical regret guarantees. Our last algorithm investigates this issue.

**Robust regret for light tailed distributions** The third and final instance of DS algorithms requires little setup as it effectively ensures sublinear regret under virtually no assumption other than reward distributions being in the light tailed family $\mathcal{F}_\ell$, which we recall from Section 1.3 in Chapter 1 only imposes the existence of the moment generating function around zero.

Consider a $K$-armed bandit measure $\boldsymbol{\nu} = (\nu_k)_{k\in[K]}$. We define the following bandit model:

$$\textbf{Bandit model (R):} \quad \boldsymbol{\nu} \in \mathcal{F}_\ell^{\otimes K} . \tag{6.45}$$

We call *robust Dirichlet sampling* (RDS) bonus the canonical bonus of Definition 6.11 coupled with nondecreasing sequences $(\rho_n^k)_{n\in\mathbb{N}} \in \mathbb{R}_+^{\mathbb{N}}$ for each $k \in [K]$, as in equation 6.30. For an appropriate choice of such sequences, the resulting DS algorithm has sublinear regret for *any* light tailed bandit model, which is why we call it *robust*.

**Theorem 6.19** (Expected pseudo regret bound for RDS). *Algorithm 5 with* RDS *index satisfies, in bandit model* $(R)$*, for any suboptimal arm* $k \in [K] \setminus \{k^\star\}$*,* $T$ *large enough,* $\eta \in (0, 1]$ *and* $\varepsilon > 0$*,*

$$\mathbb{E}[N_T^k] \leqslant n_k^{\eta,\varepsilon}(T) + \mathcal{O}(1)\,, \tag{6.46}$$

*where* $n_k^{\eta,\varepsilon}(T)$ *is a solution to the fixed point equation*

$$n_k^{\eta,\varepsilon}(T) = \frac{\log T}{\eta(\Delta_k - \varepsilon)} (M^k_{n_k^{\eta,\varepsilon}(T)} - \mu^\star)\,, \tag{6.47}$$

*and, for all* $n \in \mathbb{N}$*,*

$$M_n^k = \rho_n^k \vee F_k^{-1}\left(\exp\left(-\frac{1}{n^2(\log n)^2}\right)\right)\,, \tag{6.48}$$

*where* $F_k^{-1}$ *is the (pseudo)inverse of the c.d.f. of* $\nu_k$*. In particular, if* $\rho_n^k = \mathcal{O}(\log n)$ *then*

$$\mathbb{E}[N_T^k] = \mathcal{O}(\log(T) \log\log(T))\,. \tag{6.49}$$

The term $M_{k,n}$ derives from a high probability upper bound on the maximum of $n \in \mathbb{N}$ rewards from arm $k \in [K]$, which we discuss further in Appendix E.3. For light tailed distributions, it holds that $M_n^k = \mathcal{O}(\log n)$ (using Jensen's inequality as in the proof of Boucheron et al. (2013, Theorem 2.5)). Hence, choosing $\rho_n^k = \mathcal{O}(\log n)$ we can further simplify the regret upper bound to $\mathcal{O}(\log(T) \log\log(T))$. Of note, this slightly larger than logarithmic rate is a direct consequence of the lower bound on $[BCP]$ stated in Lemma 6.8. In our opinion this is a small cost compared to the adaptive power of RDS. We recommend setting the sequence of leverages to $\rho_n^k = \rho_n = \sqrt{\log(1+n)}$ (identical for all arms), which corresponds (up to a constant multiplicative factor) to the growth rate of the maximum of sub-Gaussian samples and is empirically validated (see Section 6.4). We emphasise that in this case, RDS eschews any hyperparameter tuning, a desirable feature for the practitioner with little information on the problem at hand. Furthermore, in the next section we show that this algorithm performs very well in practice despite its nonlogarithmic asymptotic guarantees.

## 6.4 Numerical experiments and application to a crop farming bandit

We present in this section various experimental results for the DS algorithms. First, we introduce a multiarmed bandit inspired by a decision-making problem in agriculture. In particular, we show that each of the three settings we introduced, namely bounded with unknown but detectable upper bound (BDS), unbounded with a quantile condition (QDS) and robust (RDS), are eligible assumptions for simulated maize grain yields, providing new modelling tools for the practitioner. Going further, we detail another setting for which QDS and RDS apply but that is outside the scope of SPEF algorithms: the family of Gaussian mixtures, which can be of practical relevance as these distributions are used for nonparametric estimation of arbitrary probability densities via *kernel density estimation*. We also test the sensitivity of the DS algorithms to their exploration bonus. For BDS, we consider a toy bandit problem with uniform arms and provide heuristics to tune the hyperparameter $\rho$ that complete the theoretical recommendations. For RDS, we follow the experimental setting of the robust UCB for light-tailed distributions presented in Ashutosh et al. (2021) and show the superiority of the Dirichlet ampling approach for different leverage sequences $(\rho_n)_{n\in\mathbb{N}}$.

### DSSAT bandit

We consider a practical decision-making problem using the DSSAT[3] simulator (Hoogenboom et al., 2019). Harnessing more than 30 years of expert knowledge, this simulator is calibrated on historical field data (soil measurements, genetics, planting date, etc.) and generates realistic crop yields. Such simulations are used to explore crop management policies *in silico* before implementing them in the real world, where their actual effect may take months or years to manifest themselves. More specifically, we model the problem of selecting a planting date for maize grains among $K = 7$ possible options, all else being equal, as a seven-armed bandit with reward distributions $\bigotimes_{k\in[K]} \nu_k$ corresponding to crop yields. The resulting distributions incorporate historical variability and exogenous randomness coming from a stochastic weather model; we illustrate their histograms in Figure 6.3.

[3] *Decision Support System for Agrotechnology Transfer* is an open-source project maintained by the DSSAT Foundation, see https://dssat.net/.

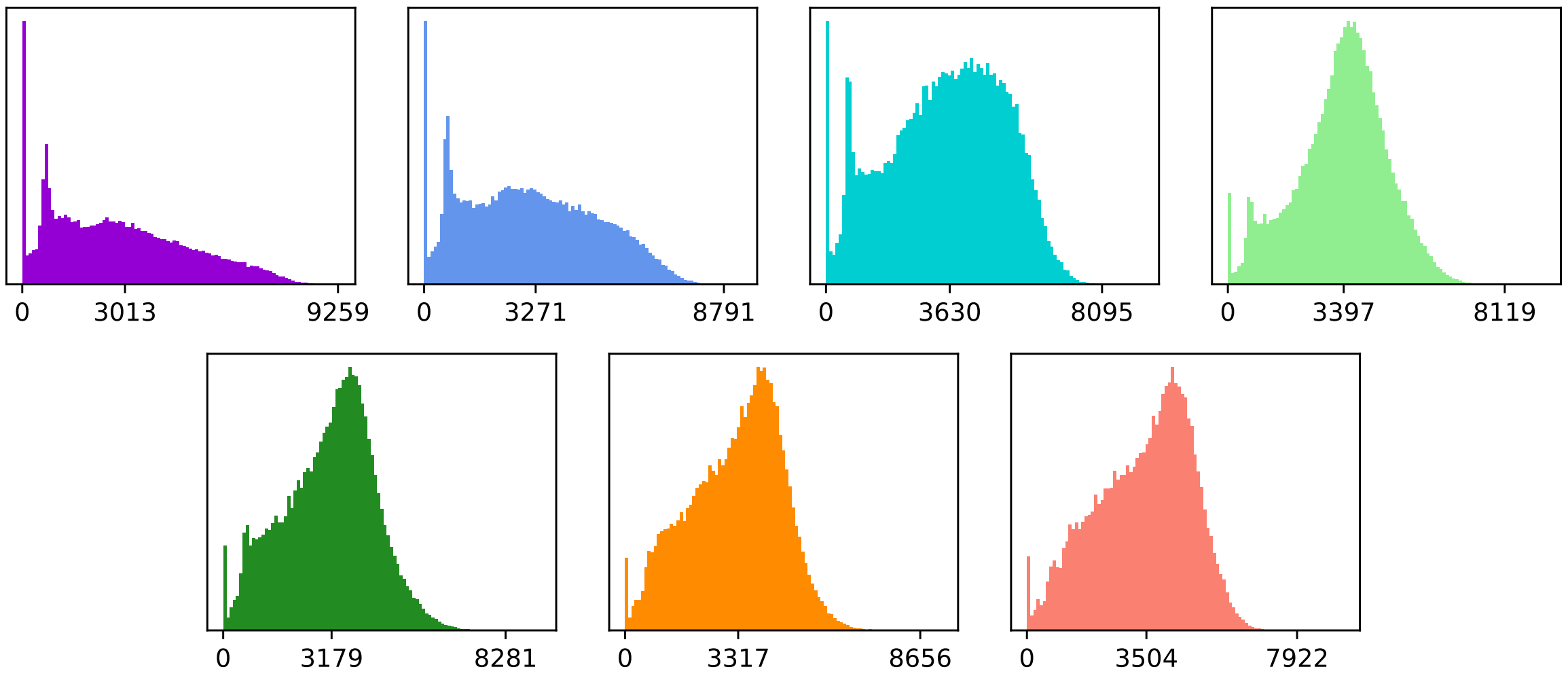

**Figure 6.3** – Distribution of simulated maize dry grain yield (kg/ha) for seven different planting dates, computed on $10^6$ samples. Reported on the x-axis are the distribution minimum, mean and maximum values. The optimal arm is the third one (mean 3630 kg/ha).

**Modelling** A natural choice for the learner would be to use algorithms adapted for bounded distributions with known support. Indeed, one could argue that crop yields are fundamentally bounded by some large value, called the *yield potential*, that can be provided with some expert knowledge. However, this approach is limited when the upper bound is not known accurately (few data, new environment, etc.), as overestimating it has a direct impact on the theoretical regret guarantees (cf. Lemma 6.10). Moreover, yield distributions are typically right skewed, multimodal and exhibit a peak at zero corresponding to years of zero harvest, hence they hardly fit a convenient parametric model (e.g. SPEF) or a stringent tail control (e.g. sub-Gaussian).

We believe that the novel Dirichlet sampling algorithms we introduced in Section 6.3 provide promising alternative modelling choices for this problem. In the DSSAT setting in particular, the three exploration bonuses detailed in Theorems 6.15, 6.17 and 6.19 encapsulate relevant assumptions: BDS keeps the bounded support hypothesis but introduces a possible uncertainty on the upperbound, QDS posits that the conditional value at risk is a meaningful summary of tail statistics, and RDS capture realistic distributions under weak, light tailed assumptions.

**Benchmark algorithms.** We present the benchmark algorithms used to compare against the DS algorithms on the DSSAT bandit problem, along with their necessary assumptions (extracted from Table 2.1).

Assuming yield distributions are bounded, one can use classical algorithms such as UCB1 (Auer et al., 2002) or Thompson sampling with $\mathrm{Beta}$ prior using the binarisation trick introduced in Agrawal and Goyal (2012). These algorithms enjoy logarithmic regret without the optimal $\mathcal{K}_{\text{inf}}$ constant of Burnetas and Katehakis (1996) for nonparametric models.

Other algorithms for bounded distributions include empirical IMED (Honda and Takemura, 2015) and NPTS (Riou and Honda, 2020). The former is based on the calculation of $\mathcal{K}_{\text{inf}}$ indices directly inspired by the Burnetas-Katehakis lower bound. We distinguish IMED, which relies on a SPEF assumption to explicitly compute the $\mathcal{K}_{\text{inf}}$ and therefore falls short of the scope of DSSAT, from empirical IMED, which solves the (dual) convex optimisation problem defined by the $\mathcal{K}_{\text{inf}}$ of the empirical distribution of each arm and requires boundedness.

As discussed above, these algorithms require the explicit knowledge of upper bounds $(\overline{B}_k)_{k\in[K]}$ such that Supp $\nu_k \subset [0, \overline{B}_k]$ for all $k \in [K]$. To represent the fact that tight bounds are sometimes unknown to the practitioner (uncertain environment, possibility of events unobserved so far, etc.), we run two variants of the above algorithms: (i) with a rather tight bound of $\overline{B}_k = 10^4$ kg/ha for all arms, with closely matches the maximum yields over all arms on simulated data and thus represent a strong prior information, and (ii) with the same bound inflated by $50\%$, i.e. $\overline{B}_k = 1.5 \times 10^4$ kg/ha, which we deem a conservative estimate.

Finally, RB-SDA (Baudry et al., 2020) is a recent subsampling algorithm using a similar round-based structure as our DS algorithms. Its optimality is only established under tail conditions satisfied by some SPEF; in particular, it enjoys none of the theoretical guarantees of the previous algorithms and is shown for empirical comparison only. Note that although it has been analysed under strong parametric assumptions, the algorithm itself is nonparametric, and in particular is agnostic to the choice of the upper bound.

**Tuning** For BDS we chose the parameters $\rho = 4, \gamma = 3500$, corresponding to $p \approx 20\%$ in the hypothesis of Theorem 6.15, which is conservative in our example. For QDS, we set $\rho = 4$ to be able to compare with BDS and a quantile $1 - \alpha$ with $\alpha = 5\%$. Finally for RDS, we chose $\rho_n = \sqrt{\log(1+n)}$, which fits the theoretical framework of Theorem 6.19 and ensures that regret is bounded by $\mathcal{O}(\log(T)\log\log(T))$.

**Results** We report statistics of the cumulative regret of the DS algorithms and the benchmarks in Figure 6.4 and Table 6.1. As expected, UCB1 and binarised TS perform poorly on the DSSAT problem, hinting that this particular bandit instance is not easy and requires more sophisticated methods. RB-SDA achieves small regret but exhibits larger dispersion than other methods ($95\%$ quantile of empirical total regret is $0.99 \times 10^6$, standard deviation is $0.26 \times 10^6$), meaning that some runs in the Monte Carlo simulations suffer high regret; we interpret this as evidence that RB-SDA operates outside its theoretical scope here and is therefore not backed by strong regret guarantees. The DS algorithms achieve similar or even slightly better regret than empirical IMED and NPTS using the prior knowledge of a tight upper bound. However the latter two suffer from using the conservative bound instead. Note that contrary to RB-SDA, empirical IMED and NPTS remain theoretically sound in both the prior knowledge and conservative cases, but the larger bound drives the exploration-exploitation balance towards more exploration than is optimal. The overall winner is RDS, which outperformed all others on both mean and median total regret. Finally, the tuning of the leverage $\rho$ in the DS algorithms is done using plausible heuristics (see Figure 6.6) and have not been optimised to suit this particular problem.

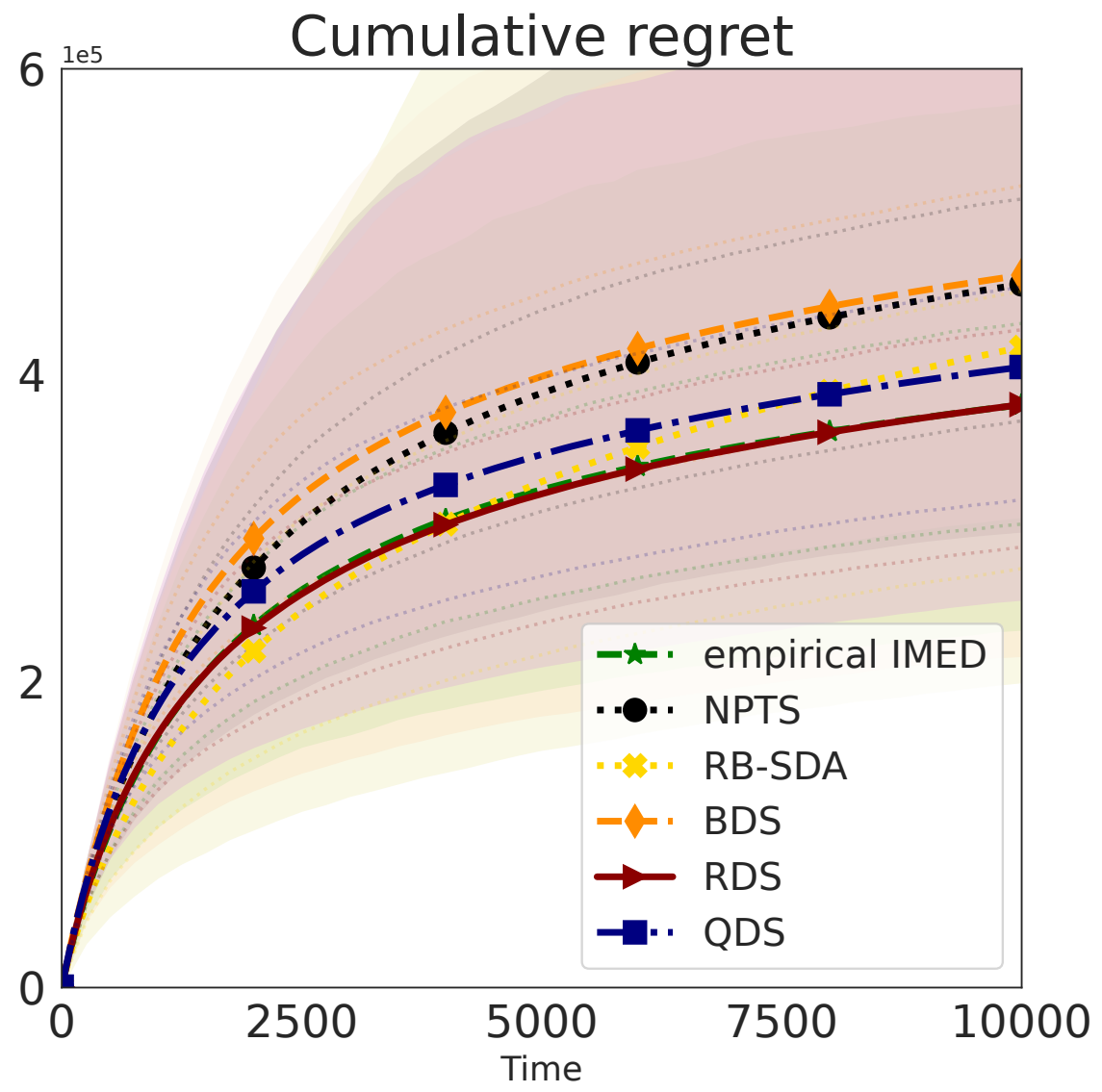

(**a**) Optimal algorithms, tight upper bound.

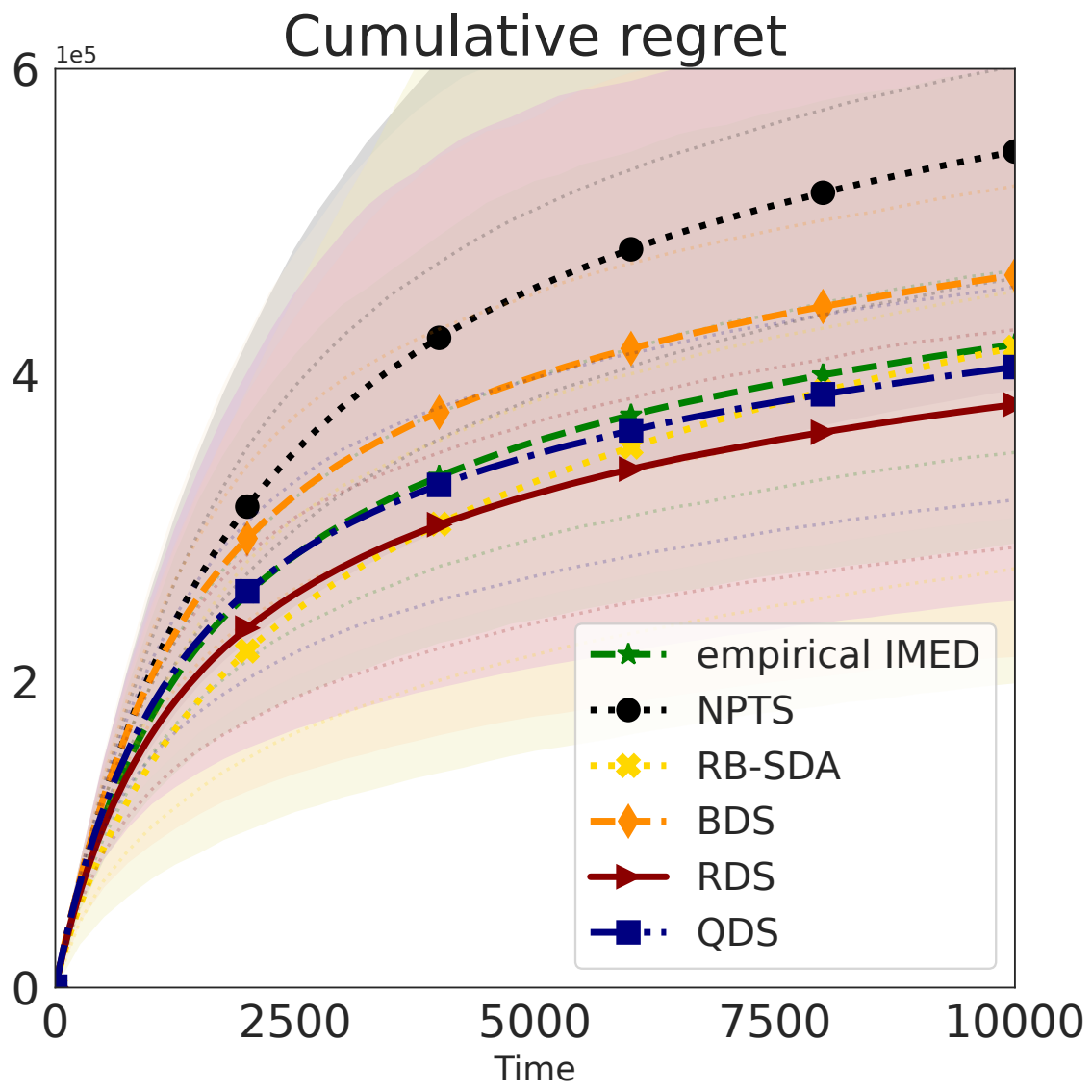

(**b**) Optimal algorithms, conservative upper bound.

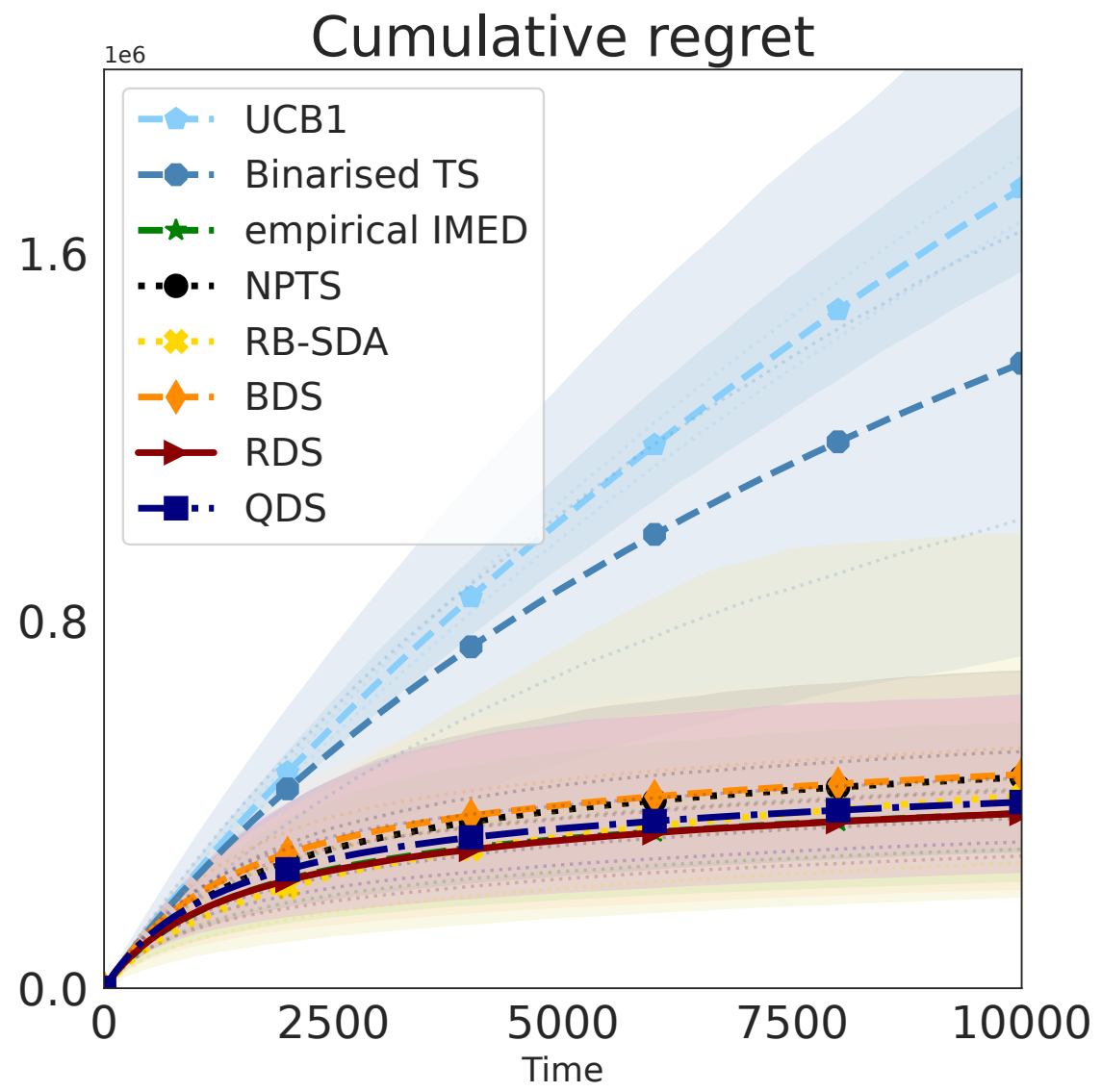

(**c**) All algorithms, tight upper bound.

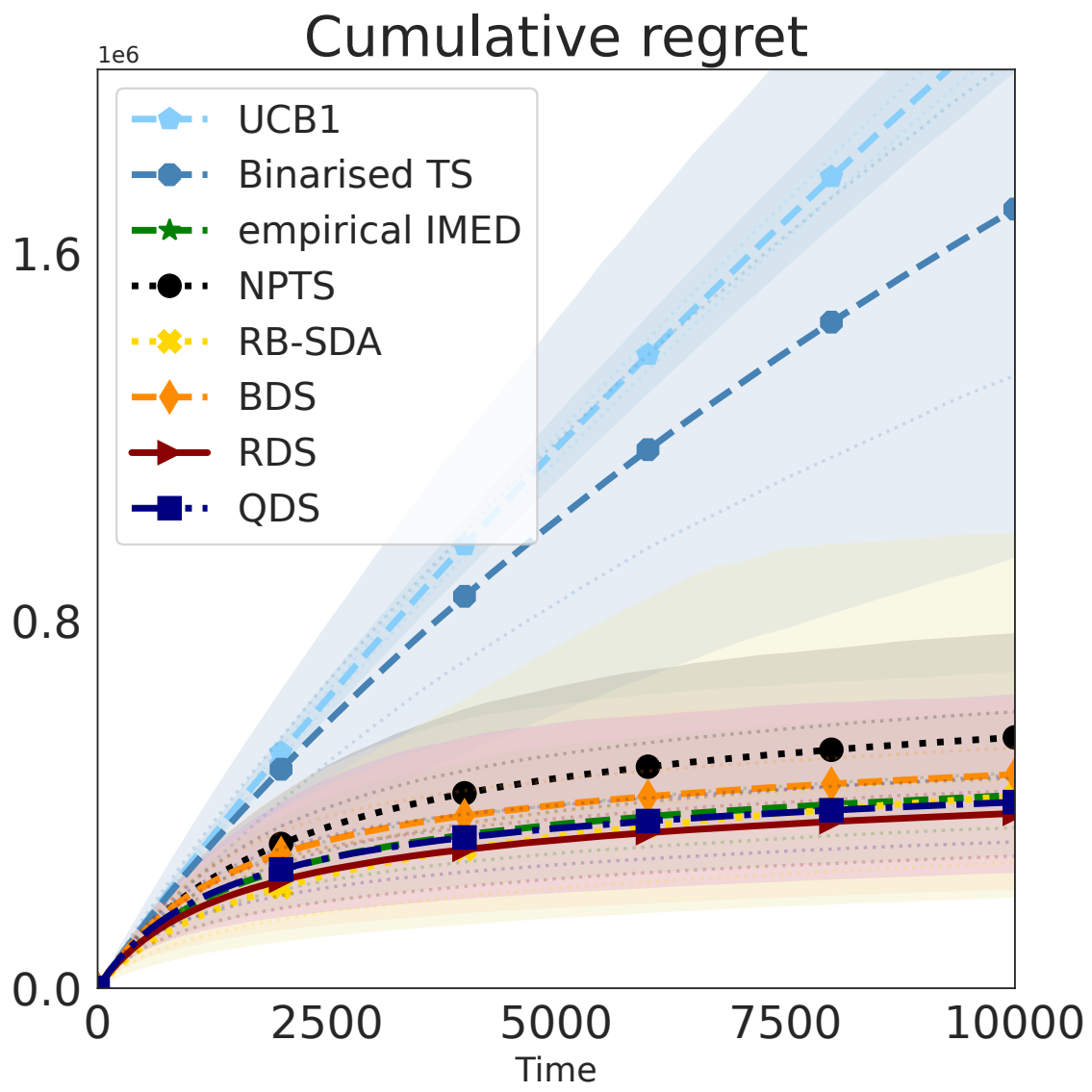

(**d**) All algorithms, conservative upper bound.

**Figure 6.4** – Comparison of DS algorithms against benchmarks on a seven-armed DSSAT bandit, with parameters $\rho = 4$ (BDS), $\rho = 4, \alpha = 5\%$ (QDS) and $\rho_n = \sqrt{\log(1+n)}$ (RDS). From left to right: tight ($10^4$ kg/ha) or conservative ($1.5 \times 10^4$ kg/ha) reward upper bound. From top to bottom: theoretically optimal benchmarks only (empirical IMED, NPTS and RB-SDA), or all benchmarks (UCB1, and binarised Thompson sampling). Thick lines denote mean cumulative regret up to horizon $T = 10^4$ over 5000 independent replicates. Dotted lines denote to respectively 25th and 75th regret percentiles. Shaded areas denote the 5th and 95th percentiles.

**Table 6.1** – Regret on DSSAT bandit at $T = 10^4$, over $N = 5000$ independent replicates. Scale = $10^6$ kg/ha. SD: standard deviation. IQR: interquartile range.

| Algorithm | 5% quantile | Mean ± SD | Median ± IQR | 95% quantile |
|---|---|---|---|---|
| UCB1 | 1.56 | $1.74 \pm 0.11$ | $1.74 \pm 0.14$ | 1.92 |
| UCB1 (conservative) | 2.00 | $2.13 \pm 0.08$ | $2.13 \pm 0.11$ | 2.26 |
| Binarised TS | 0.72 | $1.36 \pm 0.46$ | $1.29 \pm 0.63$ | 2.20 |
| Binarised TS (conservative) | 0.94 | $1.70 \pm 0.50$ | $1.66 \pm 0.69$ | 2.57 |
| Empirical IMED | 0.23 | **0.38** $\pm 0.13$ | $0.36 \pm 0.13$ | **0.58** |
| Empirical IMED (conservative) | 0.29 | $0.42 \pm 0.10$ | $0.4 \pm 0.12$ | 0.60 |
| NPTS | 0.30 | $0.46 \pm 0.14$ | $0.43 \pm 0.14$ | 0.69 |
| NPTS (conservative) | 0.39 | $0.55 \pm 0.13$ | $0.53 \pm 0.18$ | 0.77 |
| RB-SDA | **0.20** | $0.42 \pm 0.26$ | **0.35** $\pm 0.18$ | 0.99 |
| BDS | 0.31 | $0.47 \pm 0.13$ | $0.44 \pm 0.14$ | 0.68 |
| QDS | 0.25 | $0.41 \pm 0.14$ | $0.38 \pm 0.14$ | 0.64 |
| RDS | 0.22 | **0.38** $\pm 0.16$ | **0.35** $\pm 0.14$ | 0.63 |

## Gaussian mixture

Many real world situations (loss profile of a portfolio of financial assets, crop yields, statistics of heterogeneous populations, etc.) exhibit multimodal distributions. The Gaussian mixture model is perhaps the simplest example of such distributions and is ubiquitous in many areas of machine learning and engineering (speech recognition, clustering, etc.), in particular as a nonparametric model for kernel density estimation. Still, to the best of our knowledge, it escapes the scope of current optimal bandit methods as it is neither bounded nor SPEF. Thanks to the different sets of assumptions in which they operate, both RDS and QDS are eligible algorithms to tackle the problem of sequential decision-making in a Gaussian mixture environment, at the cost of slightly larger than logarithmic regret and lower $\mathcal{K}_{\inf}$ rate respectively.

We consider two arms distributed as independent mixture of Gaussian distributions:

$$\nu_1 = \frac{1}{2}\mathcal{N}(-0.3, 1) + \frac{1}{2}\mathcal{N}(1.3, 1)\,, \tag{6.50}$$

$$\nu_2 = \frac{1}{10}\mathcal{N}(-1.5, 1) + \frac{4}{5}\mathcal{N}(0.6, 1) + \frac{1}{10}\mathcal{N}(2.5, 1)\,. \tag{6.51}$$

Note that both mixtures have total variance equal to $1$, but different means ($\mu_1 = 0.5$ vs. $\mu_2 = 0.58$). Due to the lack of theoretically grounded benchmark, we run three SPEF algorithms (`kl-UCB`, IMED and Thompson sampling) assuming the arms belong to the SPEF of Gaussian distributions with fixed variance $1$. We also run UCB1, calibrated for the family of $1$-sub-Gaussian distributions. This is an example of *model misspecification*.

We run RDS with $\rho_n = \sqrt{\log(1+n)}$, which matches the asymptotic growth rate of the maximum of i.i.d. Gaussian samples, and QDS with $\alpha = 5\%$, $\rho = 4$; we recall that Appendix E.4 shows empirical evidence that the quantile condition required by QDS holds for a large variety of $\alpha$ and $\rho$ in the case of Gaussian tails. Note that the use of QDS in this context is technically out of scope of Theorem 6.17 since Gaussian mixtures are not lower bounded; we believe however that this is an artifact of our proof technique that could lifted with a finer analysis.

Results are reported in Figure 6.5. Both RDS and QDS outperform other existing methods; in particular, among the misspecified SPEF algorithms, only IMED exhibit comparable regret growth, albeit with higher overall regret. As this bandit problem is complicated (small optimality gap, non SPEF distributions), all algorithms have a relatively large variance.

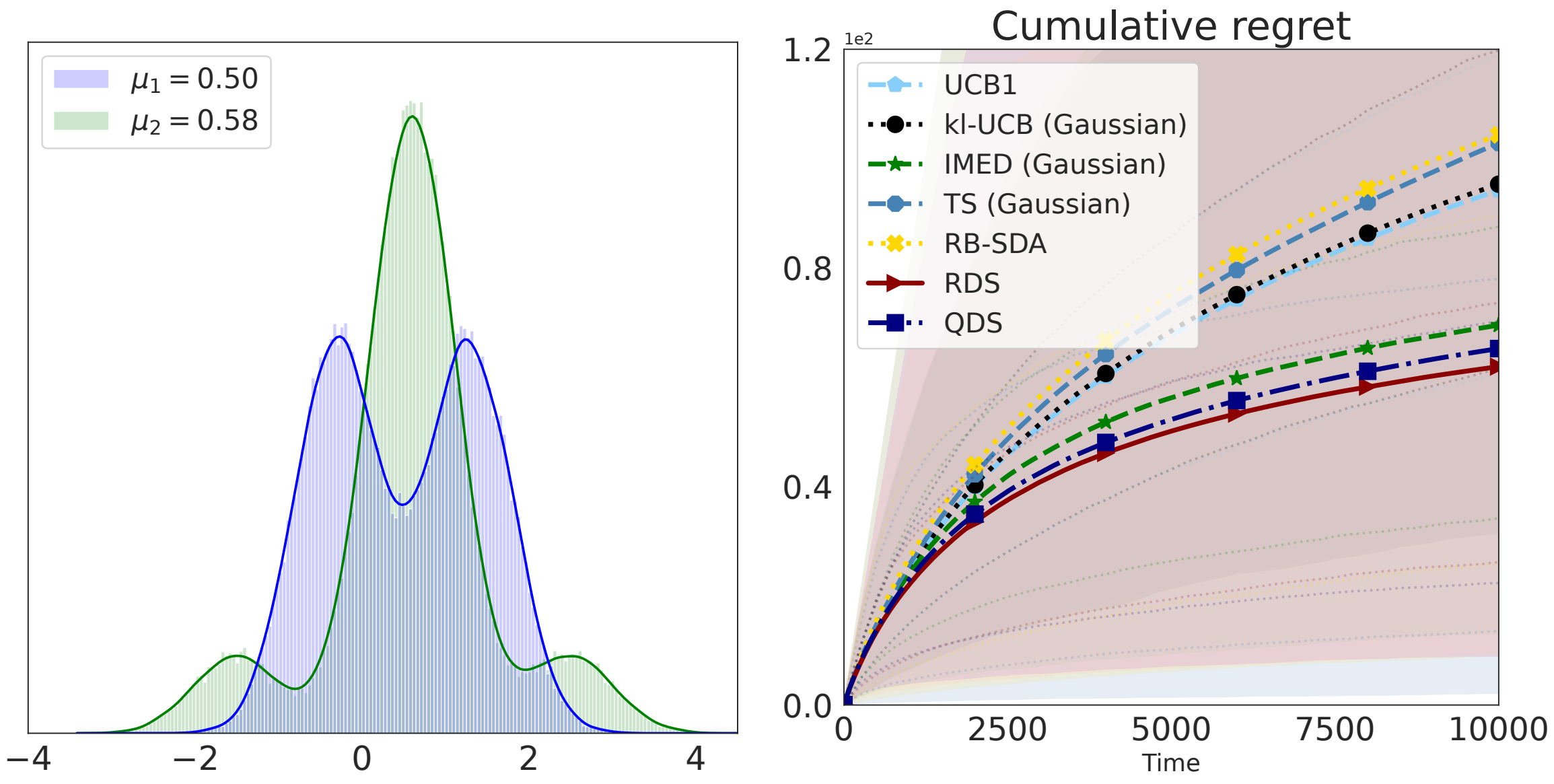

**Figure 6.5 –** Left: Gaussian mixture arms. Right: comparison of QDS and RDS against benchmarks on a two-armed bandits with these arms. UCB1, IMED, `kl-UCB` and TS are run assuming (sub-)Gaussian arms with same variance as the mixtures. RDS: $\rho_n = \sqrt{\log(1+n)}$. QDS: $\rho = 4, \alpha = 5\%$. Thick lines denote mean cumulative regret up to horizon $T = 10^4$ over 5000 independent replicates. Dotted lines denote to respectively 25th and 75th regret percentiles. Shaded areas denote the 5th and 95th percentiles.

### BDS parameters sensitivity.

We now study the impact of the leverage parameter $\rho$ on the regret of BDS. Theorem 6.15 suggests to scale it as $\rho = -1/\log(1-p)$, where $p \in (0,1)$ is one of the parameter of BDS. We believe this bonus to be rather conservative when $p$ is small and the distributions considered exhibit little skewness; as an example, if a distribution is such that at most $25\%$ of its mass is located to the right of the optimal mean reward $\mu^\star$, $\rho \approx 4$ should be a suitable tuning.

To investigate this, we consider a toy bandit instance with two arms following uniform distributions on $[0,1]$ and $[0.2, 0.9]$ respectively (note that the upper bound is different for each arm yet the distribution of mass near their respective bounds is the same, thus fitting the setting (B2) of BDS). These distributions are shown in Figure 6.6, and in particular their means are $0.5$ and $0.55$ respectively. For $\gamma = 0.1$, we compute the expected regret of BDS obtained with the theoretical tuning $\rho = -1/\log(1-p) \approx 9.5$, and compare it with other choices of $\rho$. Figure 6.6 shows that only the most extreme tuning $\rho = 50$ exhibits significant, albeit still sublinear, regret. Small deviations from the theoretical tuning yields similar regret, the heuristic $\rho = 4$ discussed above being slightly better, which tends to confirm our belief that the analysis of Theorem 6.15 could be sharpened. Note that the exploration incentive given by $\rho$ is necessary since smaller values (e.g. $\rho = 0.1$) tends to accumulate more regret.

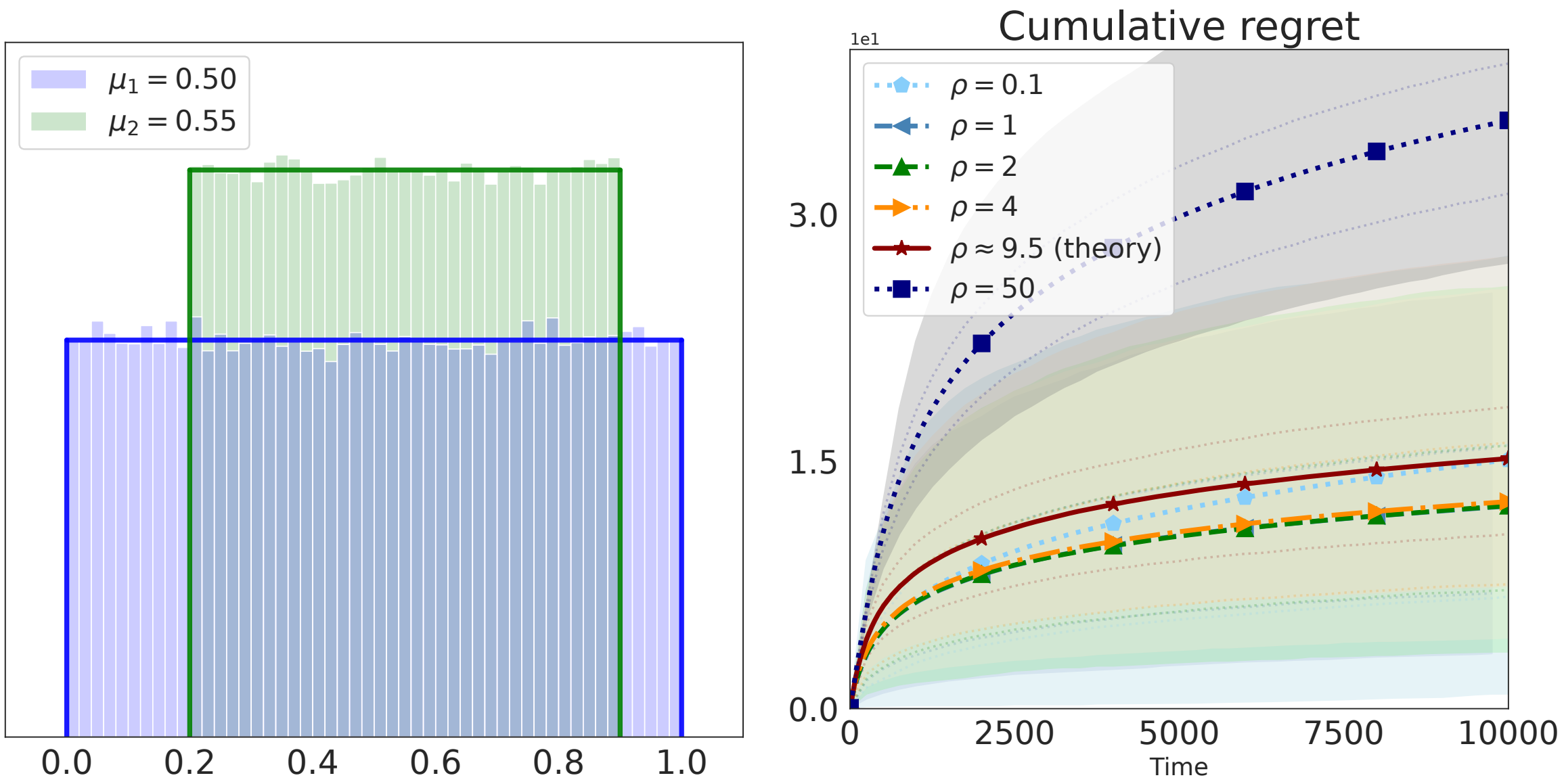

**Figure 6.6 –** Left: uniform arms $\mathcal{U}([0,1])$ and $\mathcal{U}([0.2,0.9])$. Right: comparison of different instances of leverage $\rho$ for BDS on a two-armed bandits with these arms. Thick lines denote mean cumulative regret up to horizon $T = 10^4$ over 5000 independent replicates. Dotted lines denote to respectively 25th and 75th regret percentiles. Shaded areas denote the 5th and 95th percentiles.

### Robustness for light-tailed bandits: comparison with R-UCB-LT.

The study of statistically robust bandit algorithms is fairly recent, and as such is yet to have well-established benchmarks. Ashutosh et al. (2021) introduce R-UCB-LT, an adaptation of the standard sub-Gaussian UCB to enforce robustness with respect to light tailed distributions (in the sense of Definition **??**). We reproduce the setting of their experiment, namely two Gaussian arms $\mathcal{N}(1,1)$ and $\mathcal{N}(2,3)$, and compare several variants of both R-UCB-LT and RDS against a misspecified UCB1 (the misspecification takes the form of an overly optimistic 1-sub-Gaussian assumption, while the second arm is only $\sqrt{3}$-sub-Gaussian). Both R-UCB-LT and RDS rely on a slowly growing exploration bonus, denoted respectively by $f$ and $\rho$; we run both algorithms with $f$ and $\rho$ equal to $\log^2$, $\log$ and $\sqrt{\log}$.

Results are reported in Figure 6.7. As expected, the misspecified UCB1 exhibits much faster regret growth than the robust algorithms. However, RDS outperforms R-UCB-LT, the best average regret being achieved by RDS with $\rho_n = \sqrt{\log(1+n)}$ and $\rho_n = \log(1+n)$ for $n \in \mathbb{N}$. Furthermore, the regret to RDS appears to be somewhat monotonic (slightly increasing) with respect to the growth rate of $(\rho_n)_{n\in\mathbb{N}}$, and the best results are achieved by the one matching the asymptotic growth rate of the maximum of i.i.d. Gaussian samples, as recommended by Theorem 6.19. On the other hand, the best version of R-UCB-LT is obtained with $f = \log$ (for which we do not find a theoretical intuition) and the performance gap is significant when other bonuses are considered. In light of these results, RDS seems less sensitive to its hyperparameter than R-UCB-LT, which is another sort of robustness guarantee.

Of note, the exploration bonus of R-UCB-LT for an arm $k \in [K]$ scales with the global running time $t \in \mathbb{N}$ (or round $r$ in our round-based structure) instead of the number of pulls $N_t^k$ as with RDS, which is perhaps a more natural, intrinsic way to tune exploration. We also tested R-UCB-LT and RDS with powers of $\log\log$ to account for the expected scaling $N_t^k \approx \log t$ for suboptimal arm $k \in [K] \setminus \{k^\star\}$. This resulted in similarly poor performances for R-UCB-LT; we do not report these curves to avoid cluttering.

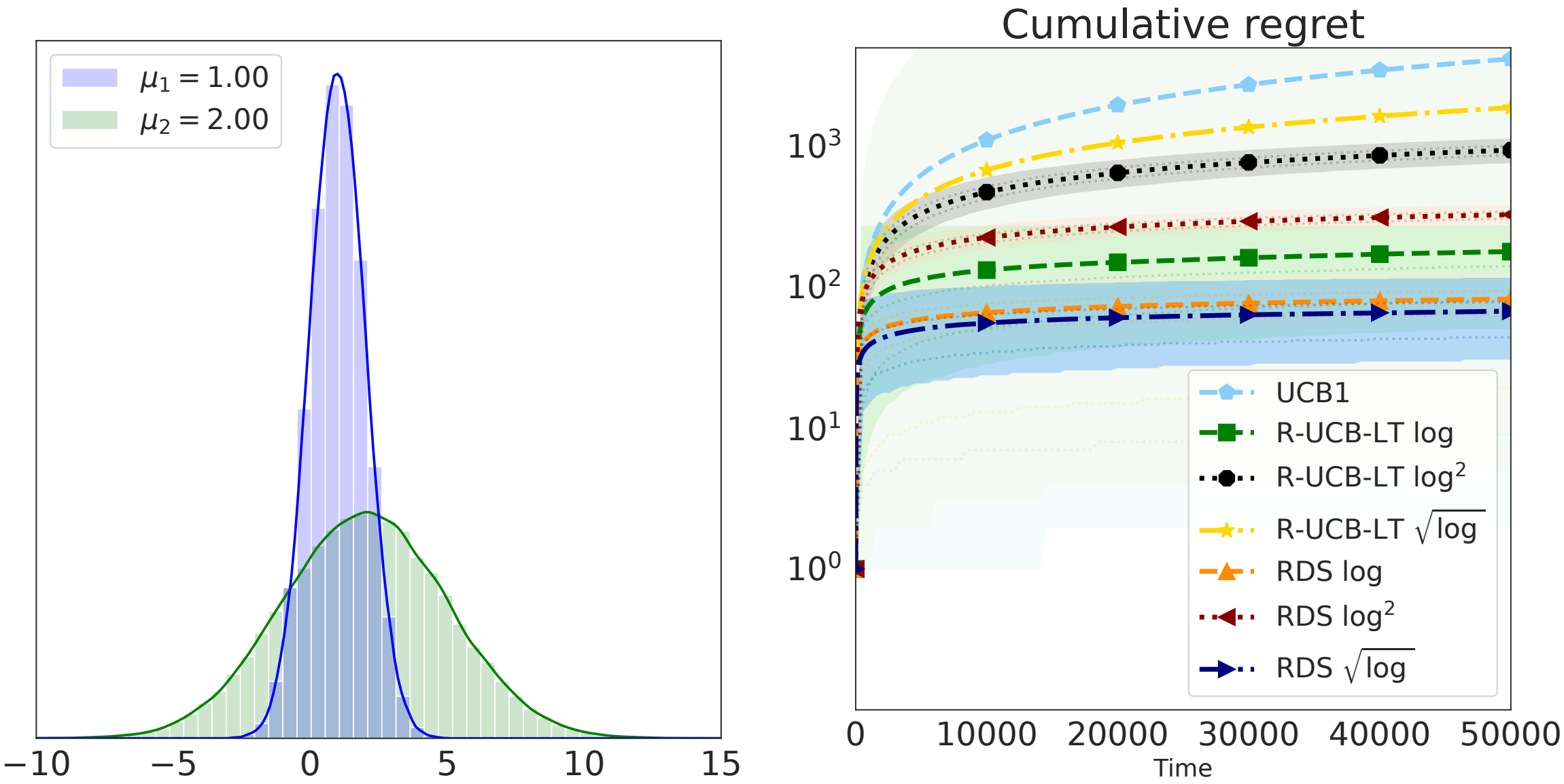

**Figure 6.7** – Left: Gaussian arms $\mathcal{N}(1,1)$ and $\mathcal{N}(2,3)$. Right: Comparison of different instances of leverage sequence $(\rho_n)_{n\in\mathbb{N}}$ for RDS and exploration bonus $f$ for R-UCB-LT (Ashutosh et al., 2021), in logarithmic scale for readability. Thick lines denote mean cumulative regret up to horizon $T = 5 \times 10^4$ over $5000$ independent replicates. Dotted lines denote to respectively 25th and 75th regret percentiles. Shaded areas denote the 5th and 95th percentiles.

## Conclusion

We have introduced a new framework for randomised exploration in stochastic bandits based on resampling of the reward history and a data-dependent bonus, which generalises an optimal Thompson Sampling strategy for bounded distributions to *light tailed* families. We proposed three instances of such Dirichlet Sampling (DS) algorithms, corresponding to different modeling assumptions. In our opinion, these new algorithms are appealing for the practitioner because (i) our theoretical results show strong guarantees under different settings, (ii) DS algorithms are simple to implement despite the technically challenging analysis and achieve strong practical performances, and (iii) they provide alternative robust ways to tackle unbounded distributions in bandit problems. Interesting future directions include extending the DS framework to *heavy tailed* distributions, and tightening the analysis of *boundary crossing probabilities* of Section 6.2 to design sharper bonuses for general families of distributions motivated by real use-cases. Moreover, we believe the duel-based structure associated with the generic regret decomposition of Theorem 6.3 opens up new perspectives to design exploration strategies in bandits. In particular, they allow to analyse policies using the history of two arms in the computation of a single index.

# Chapter 7

# Perspectives on Dirichlet sampling for contextual bandits ()

*Sometimes, you have to roll the hard six.*

— "The Hand of God", *Battlestar Galactica*

The content of this chapter is new and has not been previously published. This work started as a follow-up to the article presented in the previous chapter, in an effort to lift the Dirichlet sampling scheme from multiarmed to contextual bandits. As a testament to the flexibility of the risk-aware framework of Chapter 5, we consider the setting of linear bandits with elicitable risk measures, which extends the classical mean-linear case at minimal cost in terms of formalism. To analyse this new class of algorithms, we leverage an important probabilistic tool, namely *strong approximation of empirical processes*, which has seldom been used in the context of bandits. Beyond the scope of Dirichlet sampling, we hope this work may help popularise these methods for sequential learning.

## Contents

## Outline and contributions

In this chapter, we propose an extension of Dirichlet sampling (DS) to linear bandits. Following the setting of Chapter 5, we consider an agent that sequentially observes a decision set $\mathcal{X}_t \subset \mathbb{R}^d$ at time $t \in \mathbb{N}$, plays an action $X_t \in \mathcal{X}_t$ and receives a random reward $Y_t$, such that a certain elicitable risk measure $\rho_{\mathcal{L}}$ of the reward distribution is parametrised by $\langle \theta^\star, X_t \rangle$, where $\mathcal{L} \colon \mathbb{R} \times \mathbb{R} \to \mathbb{R}$ is a strongly convex mapping. For simplicity, the reader may think of $\mathcal{L}$ as $\mathcal{L} \colon (y, y') \in \mathbb{R} \times \mathbb{R} \mapsto (y - y')^2/2$, in which case $\rho_{\mathcal{L}}$ is simply the expectation risk measure.

## 7.1 Weighted bootstrap regression as a randomised design principle

A key component of many contextual bandit algorithms is to estimate the parameter $\theta^\star$ in an open set $\Theta \subset \mathbb{R}^d$ from a set of $t-1$ pairs of actions and rewards $((X_s, Y_s))_{s=1}^{t-1}$. Drawing inspiration from Dirichlet sampling, we introduce additional pairs $(\widetilde{X}_i^t, \widetilde{Y}_i^t) \in \mathbb{R}^d \times \mathbb{R}$ for $i \in \{1, \ldots, q\}$ and $q \in \mathbb{N}$ which represent (possibly time-dependent) exploration variables, and we define the empirical risk minimisation estimator

$$\widehat{\theta}_t = \underset{\theta \in \Theta}{\operatorname{argmin}} \sum_{s=1}^{t-1} \mathcal{L}(Y_s, \langle \theta, X_s \rangle) + \sum_{i=1}^{q} \mathcal{L}(\widetilde{Y}_i^t, \langle \theta, \widetilde{X}_i^t \rangle)\,. \tag{7.1}$$

For simplicity, we drop the regularisation term and assume all relevant matrices are invertible if need be. We also let $n_t = t + q - 1$ be the total number of variables at time $t$ ($q$ exploration variables in addition to the $t-1$ observations). Letting $\widehat{\mathbb{P}}_t = (\sum_{s=1}^{t-1} \delta_{X_s, Y_s} + \sum_{i=1}^{q} \delta_{(\widetilde{X}_i^t, \widetilde{Y}_i^t)})/n_t$ be the empirical measure on $\mathbb{R}^d \times \mathbb{R}$ induced by these $n_t$ variables, $\widehat{\theta}_t$ is an M-estimator (Huber, 2004; Kosorok, 2008) defined by $\operatorname{argmin}_{\theta \in \Theta} \mathbb{E}_{\widehat{\mathbb{P}}_t}[\mathcal{L}(Y, \langle \theta, X \rangle)]$. The optimism principle (LinUCB) aims to quantify the uncertainty of $\widehat{\mathbb{P}}_t$ using the concentration of measure phenomenon to guide exploration, yielding deterministic algorithms. An alternative approach is to substitute the empirical measure $\widehat{\mathbb{P}}_t$ with a randomised measure $\widetilde{\mathbb{P}}_t$ that aims to approximate the true distribution of the pair $(X, Y)$. This approach was pioneered by the *bootstrap* method (Efron, 1982; Efron and Tibshirani, 1994) and later extended to the *weighted bootstrap* (Rao and Zhao, 1992; Barbe and Bertail, 1995). Given a probability measure $\nu_Z$ on $\mathbb{R}_+$ and a sequence of independent $\nu_Z$-distributed random variables $Z^t = (Z_s^t)_{s=1}^{n_t}$ and $S_t = \sum_{s=1}^{n_t} Z_s^t$, we define the random (with respect to $Z^t$) probability measure

$$\widetilde{\mathbb{P}}_t = \frac{1}{S_t} \left( \sum_{s=1}^{t-1} Z_s^t \delta_{(X_s, Y_s)} + \sum_{i=1}^{q} Z_{t+i-1}^t \delta_{(\widetilde{X}_i^t, \widetilde{Y}_i^t)} \right), \tag{7.2}$$

and the associated M-estimator $\widetilde{\theta}_t \in \operatorname{argmin}_{\theta \in \Theta} \mathbb{E}_{\widetilde{\mathbb{P}}_t}[\mathcal{L}(Y, \langle \theta, X \rangle)]$. The weight measure $\nu_Z$ is typically assumed to be light tailed, either multinomial (bootstrap, corresponding to sampling past observations with replacement) or with unit mean and variance (weighted bootstrap), such as the exponential distribution $\mathcal{E}(1)$. Furthermore, if we define for $\theta \in \Theta$,

$$\begin{aligned} \Psi_\theta \colon \mathcal{X} \times \mathbb{R} &\longrightarrow \mathbb{R}^d & \quad\text{and}\quad \widetilde{F}_t \colon \Theta &\longrightarrow \mathbb{R}^d \\ (X, Y) &\longmapsto \partial \mathcal{L}(Y, \langle \theta, X \rangle) X & \theta &\longmapsto \mathbb{E}_{\widetilde{\mathbb{P}}_t} \left[ \Psi_\theta(X, Y) \right] , \end{aligned} \tag{7.3}$$

and assuming mild regularity assumptions (first order condition), $\widetilde{\theta}_t$ is the solution to the Z-estimation equation (Kosorok, 2008) $\widetilde{F}_t(\widetilde{\theta}_t) = 0$. In particular, under the expectation risk measure, we have the closed-form expression:

$$\begin{aligned} \widetilde{\theta}_t &= \underset{\theta \in \Theta}{\operatorname{argmin}} \sum_{s=1}^{t-1} Z_s^t \left( Y_s - \langle \theta, X_s \rangle \right)^2 + \sum_{i=1}^{q} Z_{t+i-1}^t \left( \widetilde{Y}_i^t - \langle \theta, \widetilde{X}_i^t \rangle \right)^2 \\ &= \left( \sum_{s=1}^{t-1} Z_s^t X_s X_s^\top + \sum_{i=1}^{q} Z_{t+i-1}^t \widetilde{X}_i^t \widetilde{X}_i^{t\top} \right)^{-1} \left( \sum_{s=1}^{t-1} Z_s^t Y_s X_s + \sum_{i=1}^{q} Z_{t+i-1}^t \widetilde{Y}_i^t \widetilde{X}_i^t \right) . \end{aligned} \tag{7.4}$$

For contextual bandits, we propose the linear Dirichlet sampling algorithm (LinDS), an adaptation of the Dirichlet sampling scheme, which is greedy with respect to the weighted bootstrap estimator, without adding an optimistic exploration bonus. Note that we formulate it for a generic convex risk (CR), as it also covers the classical mean-linear case.

**Algorithm 6** LinDS-CR

**Input:** Exploration bonuses $\left( ((\widetilde{X}_i^t, \widetilde{Y}_i^t))_{i=1}^q \right)_{t \in \mathbb{N}}$, projection operator $\Pi$, weight distribution $\nu_Z$.
**Initialisation:** Observe $\mathcal{X}_1$, $t = 1$.
**while** *continue* **do**

- Draw $(Z_s^t)_{s=1}^{t+q-1} \sim \nu_Z$ ; ▷ `Random weights`
- $\widetilde{\theta} \leftarrow \underset{\mathbb{R}^d}{\operatorname{argmin}} \sum_{s=1}^{t-1} Z_s^t \mathcal{L}(Y_s, \langle \theta, X_s \rangle) + \sum_{i=1}^{q} Z_{t+i-1}^t \mathcal{L}(\widetilde{Y}_i^t, \langle \theta, \widetilde{X}_i^t \rangle)$ ; ▷ `Weighted bootstrap`
- $\bar{\theta} \leftarrow \Pi(\widetilde{\theta})$ ; ▷ `Projection`
- $X_t \leftarrow \underset{x \in \mathcal{X}_t}{\operatorname{argmax}} \langle \bar{\theta}, x \rangle$ ; ▷ `Play`
- Observe $Y_t$ and $\mathcal{X}_{t+1}$; ▷ `Observe reward and next action set`
- $t \leftarrow t + 1$.

**Reduction to DS for multiarmed bandits.** To build intuition, we show that the weighted bootstrap strategy with exponential weights of Algorithm 6 actually matches the generic DS index strategy (Definition 6.1) in the case of multiarmed bandits with expectation risk measure.

Let $d = K \in \mathbb{N}$ and $\mathcal{X}_t = (e_k)_{k \in [K]}$ be the canonical basis of $\mathbb{R}^K$ for all $t \in \mathbb{N}$. If at time $t \in \mathbb{N}$ the action $X_t = e_k$ is played for $k \in [K]$, we let $\pi_t = k$ and the matrix $X_t X_t^\top$ has the $k$-th diagonal term equal to 1 and all other entries equal to zero. We choose $q = K$ constant exploration actions $\widetilde{X}_k^t = e_k$ and corresponding exploration rewards $\widetilde{Y}_k^t$ for $k \in [K]$. The weighted bootstrap estimator is

$$\widetilde{\theta}_t = \left( \sum_{s=1}^{t-1} \frac{\mathbb{1}_{\pi_s = k} Z_s^t Y_s}{\sum_{s=1}^{t-1} \mathbb{1}_{\pi_s = k} Z_s^t + Z_{t+k-1}^t} + \frac{Z_{t+k-1}^t \widetilde{Y}_k^t}{\sum_{s=1}^{t-1} \mathbb{1}_{\pi_s = k} Z_s^t + Z_{t+k-1}^t} \right)_{k \in [K]} . \tag{7.5}$$

If the i.i.d. random weights $(Z_s^t)_{s=1}^{t+q-1}$ follow the exponential distribution $\mathcal{E}(1)$ (redrawn at each time $t \in \mathbb{N}$), the $k$-th coordinate is equal in distribution to $\sum_{i=1}^{N_{t-1}^k} W_i Y_i^k + W_{N_{t-1}^k+1} \widetilde{Y}_k^t$, where $N_{t-1}^k = \sum_{s=1}^{t-1} \mathbb{1}_{\pi_s = k}$ and $(W_i^k)_{i=1}^{N_{t-1}^k+1} \sim \mathcal{D}_{N_{t-1}^k+1}$ (Lemma E.1), which is precisely the definition of a DS index for arm $k \in [K]$. In particular, if $\nu_k \in \mathcal{F}_{[\underline{B}_k, \overline{B}_k]}$ (setting (B1), bounded rewards with known upper bound), letting $\widetilde{Y}_k^t = \overline{B}_k$ makes LinDS equivalent to BDS.

Interestingly, this also paves a way for non-contextual multiarmed bandits with elicitable risk measures. Indeed, for a generic strongly convex loss $\mathcal{L}$, we have

$$\widetilde{\theta}_t = \operatorname*{argmin}_{\theta = (\theta_k)_{k \in [K]} \in \Theta} \sum_{k[K]} \sum_{s=1}^{t-1} \mathbb{1}_{\pi_s = k} Z_s^t \mathcal{L}(Y_s, \theta_k) + \sum_{i=1}^{K} \mathbb{1}_{\pi_s = k} Z_{t+i-1}^t \mathcal{L}(\widetilde{Y}_k^t, \theta_k) . \tag{7.6}$$

The right-hand side is the sum of $K$ univariate function in variable $\theta_k$. If the set $\Theta$ has a product structure $\Theta = \prod_{k \in [K]} \Theta_k$, each coordinate of the vector $\widetilde{\theta}_t$ can be estimated individually. By grouping the i.i.d. random variables $(Z_s^t)_{s=1}^{t+K-1}$ into $K$ i.i.d. lists $(Z_i^{k,t})_{i=1}^{N_{t-1}^k+1}$ and denoting by $(Y_i^k)_{i=1}^{N_{t-1}^k}$ the rewards observed for arm $k \in [K]$ at time $t-1$, we obtain the univariate estimators

$$\widetilde{\theta}_{k,t} = \operatorname*{argmin}_{\theta_k \in \Theta_k} \sum_{i=1}^{N_{t-1}^k} Z_i^{k,t} \mathcal{L}(Y_i^k, \theta_k) + Z_{N_{t-1}^k+1}^{k,t} \mathcal{L}(\widetilde{Y}_k^t, \theta_k) . \tag{7.7}$$

**Randomised estimators and empirical processes** Instances of randomised algorithms have been analysed for contextual bandits, starting with linear Thompson sampling Agrawal and Goyal (2013). The key component to the regret analysis is to control the probability of three events, as summarised in Kveton et al. (2020, Theorem 1) : (i) a deterministic estimator $\widehat{\theta}_t$ concentrates around the true parameter $\theta^\star$, (ii) the randomised estimator $\widetilde{\theta}_t$ concentrates around $\widehat{\theta}_t$, and (iii) $\widetilde{\theta}_t$ also *anticoncentrates* away from $\widehat{\theta}_t$. The first condition is easily satisfied if we choose $\widehat{\theta}_t$ to be the empirical risk minimiser (Corollary 5.13), so the technical challenge lies in proving the other two. That last condition is analogous to Assumption 6.4 and is necessary to promote sufficient exploration, without the need for an optimistic exploration bonus.

In stark contrast with $\widehat{\theta}_t$, the weighted bootstrap estimator $\widetilde{\theta}_t$ does not lend itself to the type of martingale analysis (method of mixtures) that we have used throughout this thesis to analyse time-uniform concentration. Indeed, because the random weights are redrawn at each time $t \in \mathbb{N}$, $\widetilde{\theta}_t$ shares little information with $\widetilde{\theta}_{t-1}$ (see Appendix F.1 for a more detailed study). For optimistic algorithms, deriving tight time-uniform confidence sequences was paramount since these were used in the design of the algorithm itself (to tune the exploration bonus); in other words, looser concentration would translate directly to more regret due to overexploration. For randomised strategies, exploration only comes from the variability of $\widetilde{\theta}_t$, and hence, while we still need to analyse the concentration of $\widehat{\theta}_t$, looser bounds may only be detrimental to the theoretical guarantees, *not* the actual empirical performances. Also, using crude union arguments instead of sharp time-uniform sequences typically contributes to an extra $\mathcal{O}(\sqrt{\log T})$ term at horizon $T \in \mathbb{N}$, which would only add a polylogarithmic correction to the minimax optimal regret $\mathcal{O}(\sqrt{T})$. We only mention the dependency in $T$ since the dependency in the dimension $d$ may be suboptimal; for instance, guarantees for linear Thompson sampling (Agrawal and Goyal, 2013; Abeille and Lazaric, 2017) are $\widetilde{\mathcal{O}}(d^{3/2}\sqrt{T})$.

We now sketch an approach to analyse the concentration of the weighted bootstrap estimator $\widetilde{\theta}_t$. Similarly to the analysis of Chapter 5, we start from the Z-estimation equation $\widetilde{F}_t(\widetilde{\theta}_t) = 0$, and try to relate $\widetilde{\theta}_t$ and $\widehat{\theta}_t$. Precisely, we define the following mappings[1]

$$\begin{aligned} \widetilde{H}_t \colon \Theta &\longrightarrow \mathcal{S}_d^+(\mathbb{R}) & \quad\text{and}\quad \bar{\widetilde{H}}_t \colon \Theta \times \Theta &\longrightarrow \mathcal{S}_d^+(\mathbb{R}) \\ \theta &\longmapsto \mathbb{E}_{\widetilde{\mathbb{P}}_t}\left[\partial^2 \mathcal{L}(Y, \langle \theta, X \rangle) X X^\top\right] & (\theta, \theta') &\longmapsto \int_0^1 \widetilde{H}_t(u\theta + (1-u)\theta')du\,, \end{aligned} \tag{7.8}$$

so that $0 = \widetilde{F}_t(\widetilde{\theta}_t) = \widetilde{F}_t(\widehat{\theta}_t) + \bar{\widetilde{H}}_t(\widehat{\theta}_t, \widetilde{\theta}_t)(\widetilde{\theta}_t - \widehat{\theta}_t)$. Assuming $\bar{\widetilde{H}}_t(\widehat{\theta}_t, \widetilde{\theta}_t)$ is invertible, we have

$$\widetilde{\theta}_t - \widehat{\theta}_t = -\bar{\widetilde{H}}_t(\widehat{\theta}_t, \widetilde{\theta}_t)^{-1} \left(\mathbb{E}_{\widetilde{\mathbb{P}}_t}\left[\Psi_{\widehat{\theta}_t}(X, Y)\right]\right)\,. \tag{7.9}$$

Therefore, we would like to control the fluctuations of the expected term on the right-hand side. To this end, we introduce the following stochastic process.

[1] Note that compared to Chapter 5, we define operators $F$ and $H$ as *means* rather than sums, which more naturally fits the description of the weighted bootstrap regression as an expectation under a reweighted measure.

**Definition 7.1** (Weighted bootstrap empirical process). *Assume that for all $\theta \in \Theta$, $\Psi_\theta$ belongs to a given family of mappings $\mathcal{F} \subset \{\psi \colon \mathbb{R}^d \times \mathbb{R} \to \mathbb{R}^d\}$, we define the* ***weighted bootstrap empirical process*** *associated with probability measures $\widehat{\mathbb{P}}_t$ and $\widetilde{\mathbb{P}}_t$ as*

$$\begin{aligned} \widetilde{\alpha}_t \colon \mathcal{F} &\longrightarrow \mathbb{R}^d \\ \psi &\longmapsto \sqrt{n_t}\left(\mathbb{E}_{\widetilde{\mathbb{P}}_t}[\psi(X,Y)] - \mathbb{E}_{\widehat{\mathbb{P}}_t}[\psi(X,Y)]\right). \end{aligned} \tag{7.10}$$

Note that this is a process in the sense that for a fixed $t \in \mathbb{N}$, $(\widetilde{\alpha}_t(\psi))_{\psi \in \mathcal{F}}$ defines a stochastic process indexed by *functions* in $\mathcal{F}$, where the randomness comes from the variables $((X_s, Y_s))_{s=1}^{t-1}$ and the weights $(Z_s^{t})_{s=1}^{n_t}$. Moreover, by definition of the deterministic estimator $\widehat{\theta}_t$, we have $\mathbb{E}_{\widehat{\mathbb{P}}_t}[\Psi_{\widehat{\theta}_t}] = 0$. We then rewrite the Z-estimation equation as

$$\sqrt{n_t}\left(\widetilde{\theta}_t - \widehat{\theta}_t\right) = -\bar{\widetilde{H}}_t(\widehat{\theta}_t, \widetilde{\theta}_t)^{-1}\widetilde{\alpha}_t(\Psi_{\widehat{\theta}_t}). \tag{7.11}$$

Consequently, it is sufficient to study the fluctuations of the weighted bootstrap empirical process. To this end, we briefly introduce the theory of *strong approximation* of empirical processes, which offers powerful tools to generalise the well-known central limit theorem.

## 7.2 Technical detour: strong approximation of empirical processes

Let us first consider a stochastic process $(S_t)_{t \in \mathbb{N}}$ such that $S_t$ is the sum of $t \in \mathbb{N}$ i.i.d. centred random variable with unit variance. It is well-known that $\sqrt{t}S_t$ converges in distribution (i.e. weakly) to a standard Gaussian $\mathcal{N}(0,1)$. The idea behind strong approximation, also known as the strong invariance principle, is to lift this analysis to a uniform (i.e. strong) convergence, either almost surely or in probability. Practically speaking, this allows to replace $(S_t)_{t \in \mathbb{N}}$ with $(B_t)_{t \in \mathbb{N}}$ where $B$ is a Brownian motion (defined on a possibly enriched probability space), up to controlled fluctuations of order almost surely $o(\sqrt{t \log \log t})$ (Strassen, 1964); in addition, this approximation rate is tight (Major, 1979).

More generally, we may consider empirical processes instead of just sums of i.i.d. random variables. We define the classical empirical process $\widehat{\alpha}_t(\psi) = \sqrt{n_t}(\mathbb{E}_{\widehat{\mathbb{P}}_t}[\psi(X,Y)] - \mathbb{E}_{\mathbb{P}}[\psi(X,Y)])$ for $t \in \mathbb{N}$ and $\psi \in \mathcal{F}$, and study its convergence in the strong topology defined by the supremum norm $\|\widehat{\alpha}_t\|_{\mathcal{F}} = \sup_{\psi \in \mathcal{F}} \|\widehat{\alpha}_t(\psi)\|_2$. Obviously, this convergence crucially depends on the structure of the family $\mathcal{F}$ (the smaller the family, the faster the convergence). For $d = 1$ and family $\mathcal{F} = \{\mathbb{1}_{(-\infty,y)},\ y \in \mathbb{R}\}$, $\|\widehat{\alpha}_t\|_{\mathcal{F}}$ corresponds to the Kolmogorov-Smirnov statistic (rescaled by $\sqrt{t}$) between cumulative distribution functions. In this specific case, assuming $\mathbb{P}$ is light tailed, the Komlós-Major-Tusnády theorem (Komlós et al., 1975, 1976) shows that almost surely $\|\widehat{\alpha}_t - \mathbb{G}_t\|_{\mathcal{F}} = \mathcal{O}(t^{-1/2}\log(t))$ (with also a maximal, nonasymptotic inequality in probability), where $(\mathbb{G}_t)_{t\in\mathbb{N}}$ is a sequence of independent $\mathbb{P}$-Brownian bridge indexed by $\mathcal{F}$, i.e. $\mathbb{G}_t$ is the centred Gaussian process with covariance $\mathbb{E}_{\mathbb{P}}[\mathbb{G}_t(\psi)\mathbb{G}_t(\psi')] = \mathbb{E}_{\mathbb{P}}[\psi\psi'] - \mathbb{E}_{\mathbb{P}}[\psi]\mathbb{E}_{\mathbb{P}}[\psi']$, for all $t \in \mathbb{N}$, and $\psi, \psi' \in \mathcal{F}$. This result was later extended by Berthet and Mason (2006) to more general families $\mathcal{F}$ satisfying uniform finite covering number conditions, such a the Vapnik-Chervonenkis condition (VC) and the more general bracketing entropy condition (BR), at the cost of slower approximation rates ; we refer to Albertus (2019) for a recent overview.

In the context of weighted bootstrap however, we are interested in the empirical process $\widetilde{\alpha}_t$ rather than $\widehat{\alpha}_t$. Fortunately, Alvarez-Andrade and Bouzebda (2013) extended the KMT theorem to the strong approximation of $\widetilde{\alpha}_t$ by a sequence of independent $\mathbb{P}$-Brownian bridges, including in the multidimensional case $d \geqslant 1$, with a $\mathcal{O}(t^{-1/(4d-2)}\log(t))$ approximation error. Finally, Albertus and Berthet (2019) provided corresponding results in dimension $d = 1$ for the generalisation of Berthet and Mason, with same approximation rates, which we now detail.

**Assumption 7.2** (Bounded bootstrap weights). *There exists $\underline{\zeta}, \overline{\zeta} \in \mathbb{R}_+^\star$ such that*

$$\mathbb{E}_{Z\sim\nu_Z}[Z] = 1, \quad \mathbb{V}_{Z\sim\nu_Z}[Z] = 1, \quad \textit{and} \quad \operatorname{Supp}(\nu_Z) \subseteq [\underline{\zeta}, \overline{\zeta}]\,. \tag{7.12}$$

*We call the parameter $\kappa_Z = \overline{\zeta}/\underline{\zeta}$ the conditioning of the weight measure $\nu_Z$.*

This assumption is not necessary to establish strong approximation results for the weight bootstrap empirical processes but simplifies otherwise cumbersome measurability arguments (Albertus and Berthet, 2019); we will also resort to it in the regret analysis in the next section. In particular, the boundedness requirement excludes the exponential distribution.

We now state the strong approximation result in dimension $d = 1$ for the (BR) condition, which is the most general.

**Definition 7.3** (Bracketing entropy condition). *Let $\mathcal{Y}$ be a metric space, $\overline{\mathcal{F}} = \{\psi \colon \mathcal{Y} \to \mathbb{R}\}$, $\mathcal{F} \subset \overline{\mathcal{F}}$, $\|\cdot\|$ a norm on $\overline{\mathcal{F}}$ and $\varepsilon > 0$. We let $[\underline{\psi}, \overline{\psi}] = \{\psi \in \overline{\mathcal{F}},\ \underline{\psi} \leqslant \psi \leqslant \overline{\psi}\}$ for $\underline{\psi}, \overline{\psi} \in \overline{\mathcal{F}}$ and define the $\varepsilon$-bracketing number of $\mathcal{F}$ with respect to $\|\cdot\|$ as*

$$N_{[\,]}(\varepsilon, \mathcal{F}, \|\cdot\|) = \inf \left\{ n \in \mathbb{N},\ \exists((\underline{\psi}_i, \overline{\psi}_i))_{i=1}^n,\ \mathcal{F} \subset \bigcup_{i=1}^n [\underline{\psi}_i, \overline{\psi}_i] \text{ and } \forall i \in [n], \|\overline{\psi}_i - \underline{\psi}_i\| \leqslant \varepsilon \right\}, \tag{7.13}$$

*and the $\varepsilon$-bracketing entropy of $\mathcal{F}$ with respect to $\|\cdot\|$ as*

$$H_{[\,]}(\varepsilon, \mathcal{F}, \|\cdot\|) = \log N_{[\,]}(\varepsilon, \mathcal{F}, \|\cdot\|)\,. \tag{7.14}$$

*We say that $\mathcal{F}$ satisfies the* (BR) *condition if there exists $b > 0$ and $r \in (0, 1)$ such that*

$$\forall \varepsilon \in (0,1),\ H_{[\,]}(\varepsilon, \mathcal{F}, \|\cdot\|_{\mathbb{P},2}) \leqslant b^2 \varepsilon^{-2r}\,, \tag{7.15}$$

*where $\|\psi\|_{\mathbb{P},2} = \sqrt{\mathbb{E}_{Y\sim\mathbb{P}}[\psi(Y)^2]}$.*

This condition is specific to the case $d = 1$ since it requires a total order (to define the segment $[\underline{\psi}, \overline{\psi}]$), and extending it to higher dimensional settings is nontrivial and remains, to the best of our knowledge, an open question.

The following lemma, directly borrowed from Vaart and Wellner (2023, Corollary 2.7.2), provides a convenient example of family that satisfies the bracketing entropy condition.

**Lemma 7.4**. *If $\mathcal{Y}$ is a bounded, nonempty interval of $\mathbb{R}$ and $\mathcal{F}$ is a subset of the set of continuously differentiable mappings with Lipschitz derivative, then it satisfies the* (BR) *condition with $r = 1/2$.*

We now state the strong approximation result of Albertus and Berthet (2019, Theorem 2.1) to our setting, which provides the theoretical foundation for the analysis of LinDS.

**Theorem 7.5** (Strong approximation under (BR)). *Let $\alpha > 0$. Under Assumption 7.2, if $\mathcal{F}$ satisfies the (BR) condition with $r \in (0,1)$, there exists $C_\alpha > 0$ and a sequence of independent $\mathbb{P}$-Brownian bridges $(\mathbb{G}_t)_{t\in\mathbb{N}}$ such that for $t$ large enough,*

$$\mathbb{P}\left(\|\widetilde{\alpha}_t - \mathbb{G}_t\|_{\mathcal{F}} \geqslant \frac{C_\alpha}{\log(t)^{\frac{1-r}{2r}}}\right) \leqslant \frac{1}{\log(t)^\alpha}. \tag{7.16}$$

*In particular, $\|\widetilde{\alpha}_t - \mathbb{G}_t\|_{\mathcal{F}} = \mathcal{O}(\log(t)^{-(1-r)/(2r)})$ almost surely when $t \to +\infty$.*

**Remark 7.6** (Strong approximation in statistics). *Arguments based on controlling suprema of empirical processes are ubiquitous in statistics, mainly to establish asymptotic normality of estimators, as many inference problems can be framed as M or Z-estimations, see e.g. Geer (2000); Kosorok (2008); Koltchinskii (2011); Vaart and Wellner (2023). On the topic of strong approximation, we refer to the monograph Csörgo and Révész (2014). However, to the best of our knowledge, these results have been seldom used by the sequential decision-making community. A possible explanation is that optimism has long been the dominating trend in designing algorithms for bandits and model-based reinforcement learning, which requires tight, nonasymptotic concentration, whereas bounds on suprema of empirical processes are typically very asymptotic and little explicit in nature. A noteworthy exception is Waudby-Smith et al. (2021), who recently applied the strong invariance principle of Strassen (1964); Major (1979) to build time-uniform confidence sequences for semiparametric causal inference.*

## 7.3 Regret analysis of LinDS-CR

We now provide a heuristic upper bound on the pseudo regret of Algorithm 6. Assuming that the family $\mathcal{F} = \{\theta \in \Theta,\ \Psi_\theta\}$ satisfies a strong approximation principle (e.g. if $d = 1$, the (BR) condition) we deduce from equation (7.11) that almost surely

$$\widetilde{\theta}_t = \widehat{\theta}_t - \frac{1}{\sqrt{n_t}} \widetilde{\bar{H}}_t(\widehat{\theta}_t, \widetilde{\theta}_t)^{-1} \left(\mathbb{G}_t\left(\Psi_{\widehat{\theta}_t}\right) + v_t\right), \tag{7.17}$$

where $\mathbb{G}_t(\Psi_{\widehat{\theta}_t})$ is a $d$-dimensional Gaussian random variable with covariance matrix $\mathbb{V}_{\mathbb{P}}[\Psi_{\widehat{\theta}_t}(X,Y)]$ and $v_t$ is converging to zero. This serves as an alternative to Corollary 5.13 to control the concentration of the randomised estimator. Note that the strong approximation, as opposed to weak convergence principles such as the central limit theorem, is crucial here for two reasons: (i) it allows to relate *random variables* instead of *distributions*, and (ii) the uniformity over the family $\mathcal{F}$ makes the error term $v_t$ independent of $\widehat{\theta}_t$ and $\widetilde{\theta}_t$. Of note, the (BR) condition is satisfied in our setting provided action sets are convex and the curvature of the loss $\mathcal{L}$ is controlled (Lemma 7.4). We illustrate this in the following (purely algebraic) lemma, where we recall the definition of the global Hessian matrix $V_t = n_t \mathbb{E}_{\widehat{P}_t}[XX^\top]$ (see Chapter 5).

**Lemma 7.7** (Relation between randomised and deterministic Hessian matrices). *Assume Assumptions 5.8 and 7.2, then we have the following almost sure inequalities:*

$$\frac{m}{n_t \kappa_Z} V_t \preccurlyeq \bar{\widetilde{H}}_t(\widehat{\theta}_t, \widetilde{\theta}_t) \preccurlyeq \frac{\kappa_Z M}{n_t} V_t \,. \tag{7.18}$$

*Proof of Lemma 7.7.* First, since $\partial^2 \mathcal{L}$ is nonnegative, we have $m\widetilde{V}_t \preccurlyeq n_t \bar{\widetilde{H}}_t(\widehat{\theta}_t, \widetilde{\theta}_t) \preccurlyeq M\widetilde{V}_t$ with $\widetilde{V}_t = n_t(\sum_{s=1}^{t-1} W_s^t X_s X_s^\top + \sum_{i=1}^{q} W_{t+i-1}^t \widetilde{X}_i \widetilde{X}_i^\top)$ and $W_i^t = Z_i^t / \sum_{j=1}^{n_t} Z_j^t$ for all $t \in \mathbb{N}$ and $i \in \{1, \dots, n_t\}$. Now given the constraint that almost surely $Z_j^t \in [\underline{\zeta}, \overline{\zeta}]$ for all $j \in \{1, \dots, n_t\}$, we see that $W_i^t$ is maximal when $Z_i^t = \overline{\zeta}$ and $Z_j^t = \underline{\zeta}$ for $j \neq i$, and conversely if $W_i^t$ is minimal, i.e.

$$\frac{\underline{\zeta}}{\underline{\zeta} + (n_t - 1)\overline{\zeta}} \leqslant W_i^t \leqslant \frac{\overline{\zeta}}{\overline{\zeta} + (n_t - 1)\underline{\zeta}} \,, \tag{7.19}$$

and thus

$$\frac{n_t}{1 + \kappa_Z(n_t - 1)} V_t \leqslant \widetilde{V}_t \leqslant \frac{n_t}{1 + \frac{n_t - 1}{\kappa_Z}} V_t \,, \tag{7.20}$$

where we recall that $\kappa_Z = \overline{\zeta}/\underline{\zeta}$. To conclude, let $f_\gamma \colon x \in [1, +\infty) \mapsto x/(1 + \gamma(x-1))$ with $\gamma \in \mathbb{R}_+^\star$. This function is differentiable and for $x \geqslant 1$ we have

$$\frac{\mathrm{d}f_\gamma}{\mathrm{d}x}(x) = \frac{1 - \gamma}{(1 + \gamma(x-1))^2} \,. \tag{7.21}$$

Therefore if $\gamma = \kappa_Z > 1$, $f_\gamma$ is nonincreasing and thus $f_\gamma(x) \geqslant \lim_{x \to +\infty} f_\gamma(x) = 1/\gamma = 1/\kappa_Z$, and if $\gamma = 1/\kappa_Z < 1$, $f_\gamma$ is nondecreasing and thus $f_\gamma(x) \leqslant f_\gamma(1) = \lim_{x \to +\infty} f_\gamma(x) = 1/\gamma = \kappa_Z$. We conclude by plugging $x = n_t$ in the above equations. ■

**Regret analysis.** We now sketch a bound on the cumulative pseudo regret for Algorithm 6. Fix $T \in \mathbb{N}$, and for $t \leqslant T$, we let $X_t^\star = \mathrm{argmax}_{x \in \mathcal{X}_t} \langle \theta^\star, x \rangle$ and recall that $X_t = \mathrm{argmax}_{x \in \mathcal{X}_t} \langle \widetilde{\theta}_t, x \rangle$. We first decompose the pseudo regret (as in Chapter 5) to make appear the estimator $\widehat{\theta}_t$ that appears in the definition of the LinDS policy:

$$\mathcal{R}_T = \sum_{t=1}^{T} \langle \theta^\star, X_t^\star - X_t \rangle \leqslant \sum_{t=1}^{T} \left| \left\langle \theta^\star - \widetilde{\theta}_t, X_t^\star \right\rangle \right| + \left| \left\langle \theta^\star - \widetilde{\theta}_t, X_t \right\rangle \right| + \underbrace{\left\langle \widetilde{\theta}_t, X_t^\star - X_t \right\rangle}_{\leqslant 0}$$

$$\leqslant \sum_{t=1}^{T} \left| \left\langle \theta^\star - \widetilde{\theta}_t, X_t^\star \right\rangle \right| + \left| \left\langle \theta^\star - \widetilde{\theta}_t, X_t \right\rangle \right| , \tag{7.22}$$

where the last inequality follows from the definition of $X_t$ as the greedy action with respect to $\widetilde{\theta}_t$. We further split the regret term at round $t$ by introducing the deterministic estimator $\widehat{\theta}_t$:

$$\mathcal{R}_T \leqslant \sum_{t=1}^{T} \left| \left\langle \theta^\star - \widehat{\theta}_t, X_t^\star \right\rangle \right| + \left| \left\langle \theta^\star - \widehat{\theta}_t, X_t \right\rangle \right| + \left| \left\langle \widehat{\theta}_t - \widetilde{\theta}_t, X_t^\star \right\rangle \right| + \left| \left\langle \widehat{\theta}_t - \widetilde{\theta}_t, X_t \right\rangle \right| . \tag{7.23}$$

Note that the first two terms are equivalent to the analysis of LinUCB (Theorem 5.18), relying on the time-uniform concentration of the deterministic estimator and the elliptic potential lemma, contributing to $\widetilde{\mathcal{O}}(\sqrt{T})$ regret (up to polylogarithmic terms in $T$). The remaining two terms are controlled by the strong approximation result of equation (7.17), i.e.

$$\left\langle \widehat{\theta}_t - \widetilde{\theta}_t, X_t \right\rangle = \left\langle \frac{1}{\sqrt{n_t}} \bar{\widetilde{H}}_t(\widehat{\theta}_t, \widetilde{\theta}_t)^{-1} \mathbb{G}_t \left( \Psi_{\widehat{\theta}_t} \right), X_t \right\rangle + \left\langle \frac{1}{\sqrt{n_t}} \bar{\widetilde{H}}_t(\widehat{\theta}_t, \widetilde{\theta}_t)^{-1} v_t, X_t \right\rangle , \tag{7.24}$$

and similarly with $X_t^\star$ instead of $X_t$. We start with the second term. By the law of large numbers, the matrix $\bar{\widetilde{H}}_t(\widehat{\theta}_t, \widetilde{\theta}_t)^{-1}$ converges almost surely, and in particular $\bar{\widetilde{H}}_t(\widehat{\theta}_t, \widetilde{\theta}_t)^{-1} = \mathcal{O}(1)$, and by Assumption 5.17, $X_t$ is bounded, therefore the contribution of that term to the cumulative pseudo regret is $\mathcal{O}(\sum_{t=1}^{T} v_t/\sqrt{t})$. The approximation error $v_t$ scales as $\log(t)^{-(1-r)/(2r)}$, so that this sum scales as $\widetilde{\mathcal{O}}(\sqrt{T})$ and thus does not increase the order of the dominating term in the pseudo regret bound (up to polylogarithmic terms). Under Lemma 7.4, we have $r = 1/2$ and the exact scaling is given by the following technical lemma.

**Lemma 7.8**. *Let $T \geqslant 2$, then $\sum_{t=2}^{T} 1/\sqrt{t \log(t)} = \mathcal{O}(\sqrt{T/\log(T)})$.*

*Proof of Lemma 7.8.* We evaluate the imaginary error function $\mathrm{erfi}\colon z \in \mathbb{R}_+^\star \mapsto 2/\sqrt{\pi} \int_0^z e^{u^2} du$ at $z = \sqrt{\log(T)/2}$, i.e. $\mathrm{erfi}(\sqrt{\log(T)/2}) = 2/\sqrt{\pi} \int_0^{\sqrt{\log(T)/2}} e^{u^2} du$. The change of variable $u = \sqrt{\log(x)/2}$ yields $\mathrm{erfi}(\sqrt{\log(T)/2}) = \int_1^T dx/(2\sqrt{2\pi x \log(x)})$. The imaginary error function admits the asymptotic expansion when $z \to +\infty$ (Abramowitz and Stegun, 1968, Section 7.1) $\mathrm{erfi}(z) = (1/(\sqrt{\pi} z) + \mathcal{O}(1/z^3)) e^{z^2}$, therefore $\int_1^T dx/\sqrt{x \log(x)} = \mathcal{O}(\sqrt{T/\log(T)})$. We conclude by a sum-integral comparison argument. ■

We finally study the contribution of the Gaussian term. With Assumptions 5.8 and 7.2, we observe that $\bar{\bar{H}}_t(\widehat{\theta}_t, \widetilde{\theta}_t)^{-1}\mathbb{G}_t/\sqrt{n_t} \approx \mathcal{N}\left(0, \Theta(V_t^{-1})\right)$ (Lemma 7.7). Conditionally on $X_t$, the covariance matrix of the approximating Gaussian process is $\mathbb{V}_{\mathbb{P}}[\partial\mathcal{L}(Y, \langle\widehat{\theta}_t, X_t\rangle) \mid X_t]X_tX_t^\top$, which is typically controlled by $\sigma^2 X_tX_t^\top$ for some $\sigma \in \mathbb{R}_+^\star$ assuming a sub-Gaussian bound in the spirit of Assumption 5.10. In other words, up to a small approximation error, the randomised estimator $\widetilde{\theta}_t$ is approximately Gaussian centred around the deterministic estimator $\widehat{\theta}_t$, with covariance proportional to the global Hessian. Interestingly, this is exactly the setting of linear Thompson sampling with Gaussian prior. In particular, such a Gaussian prior achieves both the concentration (e.g. Agrawal and Goyal (2013, Lemma 1)) and anticoncentration (Abeille and Lazaric, 2017, Appendix A) properties required in the regret analysis, provided the variance of the prior is high enough. In our case, a natural way to inflate the variance is through the choice of the exploration bonuses $((\widetilde{X}_i^t, \widetilde{Y}_i^t))_{i=1}^q$: for instance, if actions and rewards are bounded by $L > 0$ (Assumption 5.17) and $\overline{B} > 0$ respectively, a sensible heuristic would be to choose constant bonuses $(\widetilde{X}_i, \widetilde{Y}_i)$ in $(\{\pm L\}^d, \overline{B})$ (or use a slowly growing function of $t$ for the exploration reward, in the spirit of RDS, see Chapter 6).

On that account, we propose the following conjecture for the pseudo regret of LinDS.

**Conjecture 7.9** (Pseudo regret upper bound for LinDS-CR). *Let $\delta \in (0,1)$ and $T \in \mathbb{N}$. Under Assumptions 5.8, 5.10, 5.17 and 7.2, if the reward distributions have bounded support in $[\underline{B}, \overline{B}] \subset \mathbb{R}$ and the family $\mathcal{F} = \{(x,y) \in \mathcal{X} \times [\underline{B}, \overline{B}] \mapsto \partial\mathcal{L}(y, \langle\theta, x\rangle)x,\ \theta \in \Theta\}$ satisfies the (BR) condition, then there exists $m \in \mathbb{N}$ and $((\widetilde{X}_i, \widetilde{Y}_i))_{i=1}^q \in (\mathbb{R}^d \times \mathbb{R})^q$ such that Algorithm 6 enjoys similar first order pseudo regret guarantees as linear TS with Gaussian prior, i.e. with probability at least $1-\delta$, its pseudo regret is upper bounded by*

$$\mathcal{R}_T \leqslant \widetilde{\mathcal{O}}\left(d^{\frac{3}{2}}\sqrt{T}\right) \text{ or, if } |\mathcal{X}_t| \leqslant K \text{ for all } t \in \mathbb{N},\ \mathcal{R}_T \leqslant \widetilde{\mathcal{O}}\left(\sqrt{dT\log K}\right), \tag{7.25}$$

*where $\widetilde{\mathcal{O}}$ hides polylogarithmic terms in $T$.*

Of note, the main theoretical roadblock to prove this conjecture is the lack of a multidimensional strong approximation principle for weighted bootstrap under general conditions (we remark that an approximation rate of $\mathcal{O}(\log(t)^{-\beta})$ for some $\beta \in \mathbb{R}_+^\star$ would be sufficient to maintain $\widetilde{\mathcal{O}}(\sqrt{T})$ pseudo regret). In the unstructured $K$-armed bandit case however, we may apply Theorem 7.5 to each of the $K$ univariate estimation equations (7.7) under the (BR) condition, without requiring a multidimensional extension. We leave the technical details, in particular the tuning of the exploration bonus for future works.

Note that the first upper bound adds an extra $\mathcal{O}(\sqrt{d})$ dependency compared to the minimax lower bound of Theorem 1.10, which is a standard caveat of the Thompson sampling analysis.

**Remark 7.10**. *In particular with the quadratic loss, this gives a minimax regret bound for the BDS scheme studied in Chapter 6, which complements the instance-dependent analysis of Theorem 6.15.*

## 7.4 Numerical experiments

We reproduce the three experiments of Section 5.4 in Chapter 5, with the addition of Algorithm 6. For the $K$-armed bandit settings (experiments 1 and 3), we used $q = K$ exploration actions $\widetilde{X}_k = e_k$ (the $k$-th vector of the canonical basis of $\mathbb{R}^K$ for $k \in [K]$, and following the empirical merits of RDS (Chapter 6), we tuned the exploration rewards as $\widetilde{Y}_k^t = \sqrt{\log(1 + N_{t-1}^k)}$ for arm $k$ at time $t \in \mathbb{N}$. For the fully contextual setting (experiment 2), we used a single exploration action $\widetilde{X}_1 = (L)^d$, i.e. the vector where each coordinate is the upper bound on the norm of regular actions (Assumption 5.17), and a corresponding exploration reward $\widetilde{Y}_1^t = \sqrt{\log(1+t)}$. This choice of a single exploration bonus was motivated by the geometry of the decision sets in this setting. Indeed, vectors in $\mathcal{X}_t = \{X_t^1, X_t^2\}$ are drawn from Gaussian distributions with full rank covariance matrices, and thus, almost surely, neither of them are orthogonal with the fixed vector $\widetilde{X}_1$. Hence, that single vector induces exploration in both directions $X_t^1$ and $X_t^2$.

In all three experiments, LinDS-CR exhibited the best performances, with quickly flattening pseudo regret median curves (Figure 7.1). Beyond the strong practical quality of randomised algorithms, as evidenced in Chapter 6, we believe this is also a byproduct of the simplicity of the DS scheme. In LinUCB-CR and its variant, exploration is tuned manually by analysing the concentration of the rewards, and therefore these algorithms pay the price of the possible lack of sharpness in the parameters of that concentration (variance $\sigma$, conditioning $\kappa$, Hessian matrices $V_t$ or $H_t$ — see Appendix D.2 — etc.). Alternatively, they can be made more exploitative, e.g. by reducing the variance parameter $\sigma$ (that typically controls the radius of the confidence ellipsoids), at the cost of invalidating the theoretical guarantees. In stark contrast, the only degrees of freedom in the design of LinDS-CR are the exploration variables, the effect of which decreases naturally over time ($m$ exploration rewards of magnitude $\mathcal{O}(1)$ or $\mathcal{O}(\sqrt{\log t})$ versus $t$ observations). Hence, LinDS-CR is much less prone to overly conservative parametrisations (incidentally, it is also easier to implement as it eschews all matrix computations and inversions). Notwithstanding, we acknowledge that LinDS-CR is solely backed by a heuristic regret bound and that choosing efficient exploration actions, in particular in high dimensions, may be a highly nontrivial problem, that possibly depends on the geometry of the actions sets.

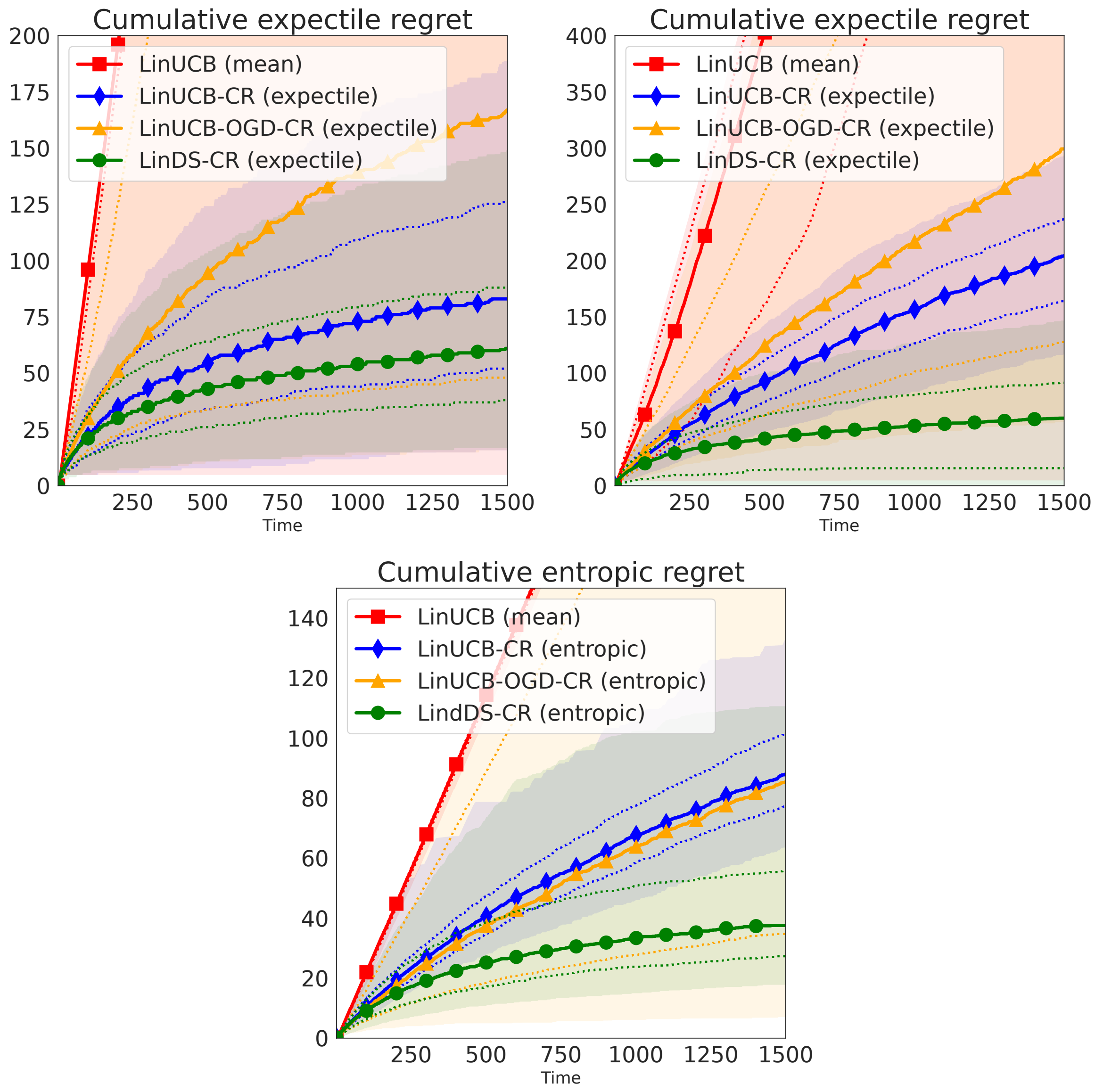

**Figure 7.1 –** Left: two-armed Gaussian expectile bandit. Center: two-armed linear expectile bandit with $\mathbb{R}^3$ contexts and expectile-based asymmetric noises. Right: two-armed Bernoulli entropic risk bandit. Thick lines denote median cumulative pseudo regret over 500 independent replicates. Dotted lines denote the 25 and 75 pseudo regret percentiles. Shaded areas denote the 5 and 95 percentiles.

## Conclusion

We have proposed an extension of Dirichlet sampling to linear bandits (both risk-neutral and risk-aware). While we lack the theoretical tools to formally analyse the resulting LinDS algorithm, it is supported by a probabilistic heuristic, leveraging strong approximation, and exhibits strong empirical performances. We hope this fosters new applications of Dirichlet sampling as a general-purpose design principle for sequential learning (as opposed to a specific scheme for the simple multiarmed bandit problem).

**Take-home message:**

**☞ To solve stochastic bandits, we have access to a range of efficient randomised algorithms, both in theory and practice.**

We have generalised algorithms based on posterior sampling for stochastic bandits to cover existing and novel model specifications, with strong practical performances and in some cases optimal instance-dependent pseudo regret guarantees.

- With no specification other than **light tailed**, we have obtained near-logarithmic regret bounds, supported by robust results on different experimental settings.
- The principle of Dirichlet sampling naturally extends to (risk-aware) linear bandits, with promising empirical results and theoretical insights that call for further research.

We believe that Dirichlet sampling is a strong general-purpose principle to design randomised algorithms for sequential learning. We however acknowledge that deterministic approaches may remain of interest to practitioners concerned with the explainability and reproducibility of the recommended decisions.

# Part V

# Contributions to statistical analysis in bariatric surgery

# Chapter 8

# Development and validation of an interpretable machine learning calculator for predicting five year-weight trajectories after bariatric surgery: a multinational retrospective SOPHIA study ()

*I guess my biggest problem is I've been cursed with the ability to do the math.*
— "Detox", *House, M.D.*

The main part of this chapter (Sections 8.2 to 8.5) was published in *The Lancet Digital Health* in August 2023. Earlier results were communicated at congresses and gatherings on obesity: the 2022 *IFSO-EC Congress on Obesity*, the 2022 *Think-Tank EGID* meeting, the 2022 *SOFFCO-MM*, the 2022 *San Diego Obesity Week*, the 2022 *MSSG AFERO*, the 2023 *IFSO World Congress on Bariatric and Metabolic Surgery* and the *Clubster NSL* webinar on chronic diseases. To broaden its outreach, the publication in the Lancet was followed by a press release in several generalist media outlets in France and beyond (Ireland, Italy); see also https://www.inria.fr/en/surgery-obesity-machine-learning-personalized-medicine for an earlier interview.

Compared to the rest of this thesis, which focused heavily on mathematics and stochastic bandits, we intend this part to be oriented towards a more medical research-inclined readership.

In this regard, we have kept the concise structure of the published article; we have however extended the mathematical details regarding the preprocessing (interpolation of visit dates) and the smoothing of weight trajectories in Appendix G.

We conclude this chapter by mentioning an independent work on the causal analysis of the impact of robotic assistance in bariatric surgery, presented at the 2023 annual meeting of the *American Surgical Association* (ASA) and later published in *Annals of Surgery*, as well as oingoing follow-up efforts to extend the weight trajectory prediction model to other types of interventions (OAGB), other populations (paediatric bariatric surgery) and other predictors (polygenic risk scores).

### Contents

## Outline and contributions

**Background.** Weight loss trajectories after bariatric surgery vary widely between individuals, and predicting weight loss before the operation remains challenging. We aimed to develop a model using machine learning to provide individual preoperative predictions of the five-year weight loss trajectories after surgery.

**Methods.** In this multinational retrospective observational study, we enrolled adult participants (aged 18 years or older) from ten prospective cohorts (including ABOS [NCT01129297], BAREVAL [NCT02310178], the Swedish Obese Subjects study, and a large cohort from the Dutch Obesity Clinic [Nederlandse Obesitas Kliniek]) and two randomised trials (SleevePass [NCT00793143] and SM-BOSS [NCT00356213]) in Europe, the Americas, and Asia, with a 5 year follow-up after Roux-en-Y gastric bypass, sleeve gastrectomy or gastric band. Patients with a previous history of bariatric surgery or large delays between scheduled and actual visits were excluded. The training cohort comprised patients from two centres in France (ABOS and BAREVAL). The primary outcome was BMI at 5 years. A model was developed using LASSO to select variables and CART to build interpretable regression trees. The performances of the model were assessed through the median absolute deviation (MAD) and root mean squared error (RMSE) of BMI.

**Findings.** 10,231 patients were included from 12 centres in 10 countries were included in the analysis, corresponding to 30,602 patient-years. Among participants in all 12 cohorts, 7701 (75.3%) were female, 2530 (24.7%) were male. Among 434 baseline attributes available in the training cohort, seven variables were selected: height, weight, intervention type, age, diabetes status, diabetes duration, and smoking status. At 5 years, the overall mean values (95% CI) of MAD and RMSE of BMI across external testing cohorts were 2.8 (2.6; 3.0) kg/m$^2$ and 4.7 (4.4; 5.0) kg/m$^2$, respectively, while the mean (SD) difference between predicted and observed BMI was -0.3 (4.7) kg/m$^2$. This model is incorporated in an easy to use and interpretable web-based prediction tool to help inform clinical decisions before surgery, which is available at this address: https://bariatric-weight-trajectory-prediction.univ-lille.fr.

## 8.1 Preliminary: a brief introduction to bariatric surgery for non-specialists

**Obesity.** Overweight and obesity, as defined by a body mass index (BMI) equal to respectively 25 and 30 kg/m$^2$ or higher, form a complex, multifactorial, worldwide disease, that has grown to pandemic proportions over the last decades,[1] with a constant increase in prevalence for *all* countries in the past 40 years (Ng et al., 2014). Contrary to long-standing misconceptions, obesity cannot be reduced to a simple behavioural issue, often mocked as a consequence of sloth and gluttony, it is rather the consequence of multiple causal factors (environmental, social, psychological, genetic). In addition to a strained physical condition, it is associated with a vast range of comorbidities (diabetes, hypertension, sleep apnea, fatty liver) and social exclusion.

**Status of bariatric surgery in the world.** Weight loss surgery, also known as *bariatric surgery*, is a set of surgical techniques that aim to

> modify the stomach and intestines to treat obesity and related diseases. The operations may make the stomach smaller and also bypass a portion of the intestine. This results in less food intake and changes how the body absorbs food for energy resulting in decreased hunger and increased fullness. These procedures improve the body's ability to achieve a healthy weight. (American Society for Metabolic and Bariatric Surgery, 2023)

In France, individuals living with severe obesity (class III, i.e. BMI equal to 40 kg/m$^2$ or higher) are eligible for bariatric surgery when first-line therapy proved insufficient or inadequate (Haute Autorité de Santé, 2009). It is also available for patients with BMI 35 kg/m$^2$ or higher associated with at least one comorbidity susceptible to be improved after surgery, for instance type 2 diabetes, high blood pressure or nonalcoholic steatohepatitis (NASH, the most severe form of the fatty liver disease). When surgery is performed specifically to improve such related comorbidities, it is often called *metabolic surgery*. Similar guidelines are applied in most countries following the 1991 National Institutes of Health consensus development conference. Recently, the American Society of Metabolic and Bariatric Surgery (ASMBS) and the International Federation for the Surgery of Obesity and Metabolic Disorders (IFSO) have lowered to respectively 30 and 35 $kg/m^2$ the BMI thresholds for bariatric surgery for patients with and without related comorbidities (Eisenberg et al., 2023); of note, an updated guideline for surgery in France is under development by the *Haute Autorité de Santé*. In France, over 60,000 bariatric interventions are performed each year, which represents a twenty times increase over the past twenty years, (Oberlin and De Peretti, 2018), meaning that 1% of the total adult population is currently living with the consequence of a bariatric procedure. To handle such a large and growing population requiring lifelong care, it is paramount to inform and develop cooperation between actors of the health system, not only specialised bariatric surgery centres but also general hospitals, organised in Territorial Hospital Groups (Gauthiez, 2017), and practitioners, who are often unfamiliar with the specificities of the bariatric postoperative care.

[1] See for instance the editorial of the June 2021 issue of *The Lancet Gastroenterology & Hepatology*, available at https://www.thelancet.com/journals/langas/article/PIIS2468-1253(21)00143-6/fulltext.

**Description of the main interventions.** Among the three most common operations, the least invasive one is the *adjustable gastric banding* (AGB), which consists in placing a ring-shaped silicon device around the upper part of the stomach to reduce its effective volume. Its impact on weight loss and associated diseases has proven limited, and its popularity has been declining in the past fifteen years. The *Roux-en-Y gastric bypass* (RYGB) is a procedure in which a small part of the stomach (the gastric pouch) is directly connected to the small intestine, leaving most of the stomach (the gastric remnant) unused. In addition to restricting the functional volume of the digestive tract, it also alters the way food is processed by the intestine (an organ often called "the second brain"), triggering a hormonal response that inhibits hunger, increases satiety and favours weight loss. It is sometimes associated with long-term complications such as ulcers, rapid gastric emptying leading to sickness ("dumping syndrome"), gallstone formations and nutritional deficiencies. Nowadays, this operation is often performed by laparoscopy ("keyhole surgery") to minimise invasiveness, as opposed to a laparotomic approach ("open surgery"). Finally, the *sleeve gastrectomy* (SG) consists in removing a large portion of the stomach, thus also limiting its functional volume and forcing a faster transition to the small intestine. Technically simpler than the RYGB, it is however non-reversible and has a lower impact on metabolism. It is associated with a higher risk of severe acid reflux. It is typically performed laparoscopically and has become the most common bariatric intervention in the past decade. A schematic representation of all three operations is reported in Figure 8.1.

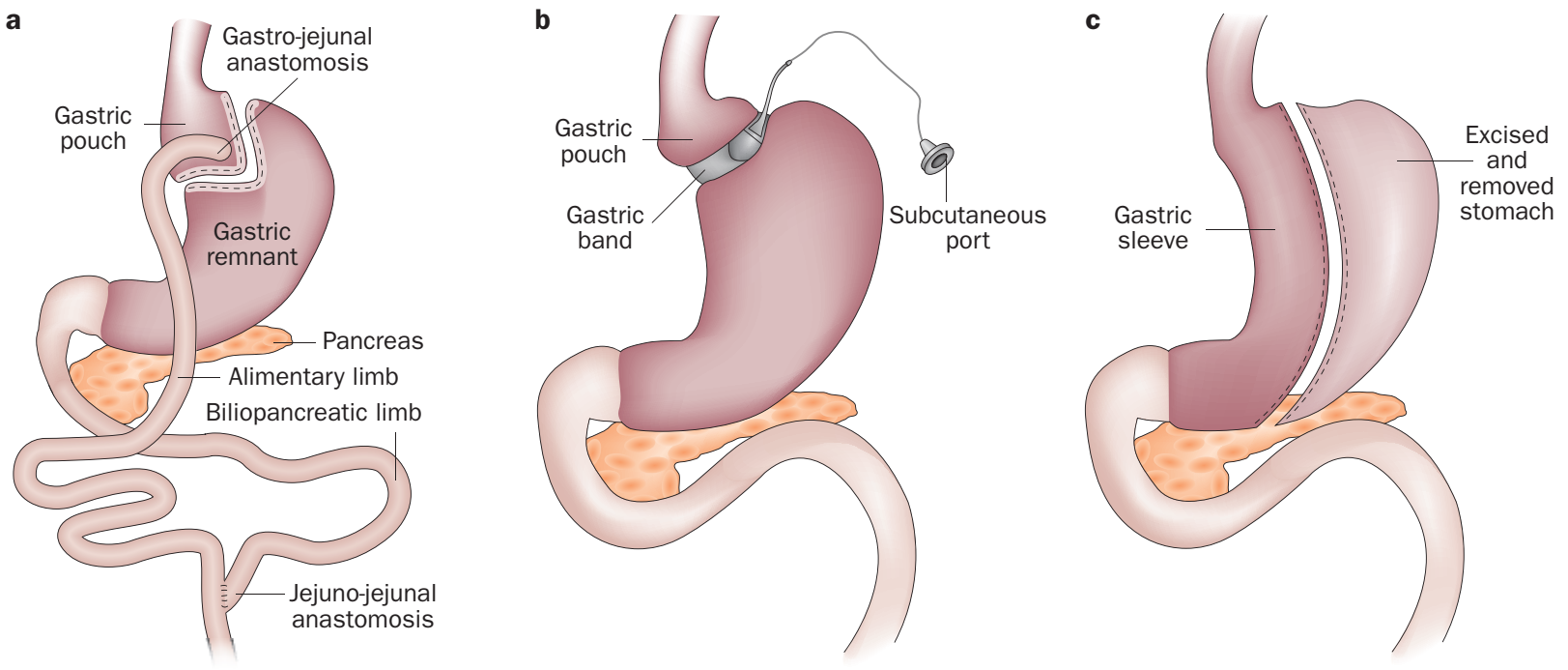

**Figure 8.1** – Schematic representation of the three main bariatric surgery operations: (a) Roux-en-Y gastric bypass (RYGB), (b) adjustable gastric banding (AGB) and (c) sleeve gastrectomy (SG). Credits to Miras and Le Roux (2013).

## 8.2 Introduction

The prevalence of obesity has increased globally over the last two decades (Collaborators, 2017). Obesity is a heterogeneous condition associated with several complications and metabolic manifestations across individuals, ultimately increasing the risk of all-cause mortality (Aune et al., 2016). Bariatric surgery, although not first-line therapy, has emerged as an effective treatment for sustained weight loss (Carlsson et al., 2020), resulting in long-term improvement in obesity-related complications (Colquitt et al., 2014), and prolonged life expectancy (Syn et al., 2021). Despite a comprehensive preoperative assessment of each candidate, it is challenging to forecast the weight loss outcomes following the intervention. Indeed, weight loss is heterogeneous as regards changes over time, differences between procedures, and between individuals (Courcoulas et al., 2018; Peterli et al., 2018).

A reasonable estimation of the expected weight loss trajectory after bariatric surgery would help inform clinical decisions by patients and healthcare providers. Therefore, multiple models have been proposed to predict postoperative weight loss (Karpińska et al., 2021). For such models to be clinically relevant, they need to predict at least five years of weight loss outcome (Colquitt et al., 2014). However, most previous studies were restricted to the early postoperative period (Karpińska et al., 2021) or undermined by the high proportion of patients lost to follow-up (Puzziferri et al., 2014), or both. Some prediction models also incorporated early weight loss (before 6 months), to predict longer outcomes (beyond 2 years, see Tettero et al. (2022); Manning et al. (2015)), and can therefore not be used before the operation.

Previous attempts to predict longer-term weight loss after bariatric surgery have used multivariable regressions (Karpińska et al., 2021). However such methods assume that the relationship between the dependent and the independent variable is linear, which may not always be the case. Additionally, multivariable regressions may lead to difficulties in accounting for interactions (Batterham et al., 2016). Moreover, although coefficients in linear regression and the odds in logistic regression are relatively easy to understand, they are not easy to apply in clinical decision-making. In contrast, machine learning methods have the potential to distinguish subtle, nonlinear patterns in data that are often not accessible using traditional approaches like logistic regression (Finks et al., 2012). Likewise, machine learning (ML) models have outperformed logistic regression in preoperative risk stratification using National Surgical Quality Improvement Program data (Bertsimas et al., 2018). Currently only a few studies have applied artificial intelligence (neural network) to predict early weight loss after bariatric surgery, but none of them were externally validated (Bektaş et al., 2022).

Therefore, the aims of the present study were: first, to use ML to develop a system predicting postoperative weight loss trajectory, utilising information gathered by protocol-driven, comprehensive preoperative assessment and repeated postoperative weight assessments in a large prospective cohort study of patients submitted to bariatric surgery; second, to validate the performance of the proposed model globally, using multiple external prospective cohorts and randomised controlled trials; and third, to incorporate the results into an easy-to-use and interpretable web-based tool providing individual preoperative prediction of postoperative weight loss trajectory.

### Research in context

**Evidence before this study.** Obesity is a heterogeneous condition that increases the risk of all-cause mortality. Bariatric surgery, while not first-line therapy, is an effective treatment for sustained weight loss improving obesity-related complications and life expectancy. However, despite comprehensive preoperative assessment of each candidate, long-term weight loss outcomes are heterogeneous as regards to changes over time, differences between procedures, and between individuals. In January 2021, a review of existing literature was conducted using the following search terms: bariatric surgery, postoperative weight loss, weight loss prediction, and prediction model. We included studies that investigated models of weight loss after Roux-en-Y gastric bypass (RYGB), sleeve gastrectomy (SG), and adjusted gastric banding (AGB) and used a prospective or retrospective design. Most previous studies were restricted to the early postoperative period and/or undermined by the high proportion of patients lost to follow-up. Some prediction models also incorporated early weight loss (before 6 months), to predict longer outcomes (beyond 2 years), and can therefore not be used before the operation. Previous attempts to predict longer-term weight loss after bariatric surgery have used multivariable regressions. However, such methods assume that the relationship between the dependent and the independent variable is linear, which may not always be the case.

**Added value of this study.** In the present study, we developed a machine learning model that provides accurate individual weight trajectories expected during 5 years after bariatric surgery, based on seven simple preoperative variables, including age, weight, height, smoking history, T2D status and duration, and the type of intervention. These variables are readily available in a variety of clinical settings without interpretation and do not require laboratory tests. The model was incorporated into an easy-to-use and interpretable web-based tool. It is the first to provide preoperative predictions of weight trajectories up to 5 years after surgery based on machine learning, simultaneously for three of the most common types of surgery, RYGB, SG, and AGB. The present study also showed the impact of diabetes duration and smoking, which were not previously included in weight loss surgery prediction models.

**Implications of all the available evidence.** This model could help to refine individual weight loss trajectory prediction in routine clinical practice, appearing as an accurate and simple strategy to inform clinical decisions for both healthcare providers and patients prior to surgery, enabling precision medicine and individualised patient management.

## 8.3 Methods

### Study design and participants

In this multinational retrospective observational study, we used data from 10 cohorts of adult patients submitted for the first time to Roux-en-Y gastric bypass (RYGB), sleeve gastrectomy (SG), and adjusted gastric banding (AGB), from eight countries. All patients had up to 5 years of postoperative data available and were aged 18 years or older. We excluded patients with a previous history of bariatric surgery, as the preoperative weights measured before reintervention already accounted for the effect of past interventions, which would have added a bias in the calculated weight loss outcomes. We also excluded patients with large delays between scheduled and actual visits related to postoperative complications. In case of missing follow-up visits, patients were kept in the analysis but censored at the corresponding dates. Patients not expected at a given time (recent interventions) were also censored after the last completed visit (full details are deferred to Appendix G.1).

**Training cohort.** This cohort consisted of patients who were prospectively enrolled at the time of primary bariatric surgery in two longitudinal cohort studies evaluating the long-term outcome of bariatric surgery, ABOS (NCT01129297) in Lille (France), and BAREVAL (NCT02310178) in Montpellier (France), between February 10, 2006 and November 2, 2020, and April 8, 2014 and April 28, 2020, respectively.

**External testing cohorts.** The prediction model was validated, using six cohorts including France (Projet régional de REcherche Clinique en Obésité Sévère [PRECOS], NCT03517072; Lyon, NCT02139813), the Netherlands (Nederlandse Obesitas Kliniek [NOK], see Tettero et al. (2022)), Sweden (the Swedish Obese Subjects [SOS] study, Sjöström et al. (1992); Carlsson et al. (2020)), Italy (NCT01581801 and NCT00888836, see Mingrone et al. (2021)), Singapore (Singapore General Hospital [SGH], see Tan et al. (2021)), Brazil (Center for the treatment of Obesity and Diabetes [COD], Hospital Oswaldo Cruz, Sao Paulo, Brazil, see Cohen et al. (2020, 2022)), and Mexico (Zerrweck et al., 2021).

**Additional testing cohort.** Additional external validation was conducted in participants of two registered and previously published randomised, open-label, multicentre trials which compared patients submitted to RYGB with SG in Finland (SleevePass, NCT00793143, see Salminen et al. (2018, 2022)) and Switzerland (Swiss Multicenter Bypass or Sleeve Study [SM-BOSS], NCT00356213, see Peterli et al. (2018)). The individual-level 5 year data of these two studies has been merged into a single analysis (Wölnerhanssen et al., 2021). Details about study protocols and data collection for these two cohorts are provided in Appendix G.3. Study participants in all cohorts gave written informed consent, and self-reported sex data and were provided with two options (male or female). All centres obtained ethics approval for their respective studies.

## Outcomes

We denote by $W_t$ the weight of a study participant measured at time $t$ after surgery ($t = 0$ represents the preoperative weight, $t > 0$ represents postoperative follow-up visits). The **body mass index** (BMI) was defined at time $t$ from the weight $W_t$ (kg) and the height $h_t$ (m) as

$$\text{BMI}_t = \frac{W_t}{h_t^2}\ \text{kg/m}^2\,. \tag{8.1}$$

The primary study outcome was the prediction of BMI at $t = 5$ years after bariatric surgery (month M60). Secondary outcomes were weight loss at earlier postoperative visits (months M1, M3, M12, M24), expressed as weight (kg) or **percent of total weight loss** (TWL), calculated as:

$$\text{TWL}_t = \frac{W_0 - W_t}{W_0} \times 100\,, \tag{8.2}$$

and **percent of excess weight loss** (EWL), calculated with the formula:

$$\text{EWL}_t = \frac{\text{BMI}_0 - \text{BMI}_t}{\text{BMI}_0 - 25} \times 100\,. \tag{8.3}$$

### Model development

To derive the model, the training cohort was divided into two subsets: a training subset consisting of 80% of randomly selected patients, and an internal testing subset comprising 20% of patients.

We first performed a preprocessing of all patients' features.As the ABOS cohort had a large number of preoperative attributes per patient, we ran a feature selection algorithm on this patient subgroup to extract the most statistically relevant ones concerning outcome prediction using the Least Absolute Shrinkage and Selection Operator (LASSO, see Lim and Hastie (2015)).

To develop the model, we further leveraged a class of ML algorithms called decision trees to learn meaningful subgroups of patients that share statistical similarities in their baseline characteristics, and second, to fit a TWL prediction model for each subgroup. For instance, decision trees can predict weight loss when they are trained on a heterogeneous cohort of different bariatric interventions such as RYGB, SG and AGB, according to the type of intervention as well as using other variables such as the age at intervention, BMI and other clinical features.

To calibrate the decision trees, we used LASSO-extracted features as input for the Classification and Regression Trees (CART) algorithm (Breiman et al., 1984). A workflow diagram of the machine learning process is displayed in Figure G.3. The algorithm was calibrated on the training subset of the training cohort. We further compared the predicted TWL to the observed outcomes of patients in the testing subset of the training cohort (internal validation). Details of the model development are reported in Appendix G.4.

Additionally, we compared that approach with other methods: classical linear models, linear mixed effect model, random forest model, and CART on all variables without LASSO using instead pruning for feature selection (Appendix G.5).

### Model validation in external cohorts

The model was externally validated by comparing the observed ($\mathrm{TWL}_t^i$) and predicted ($\widehat{\mathrm{TWL}}_t^i$) total weight loss at each visit $t$ for each participant $i$ (among $n$) of eight distinct testing cohorts (NOK, SGH, SOS, PRECOS, Roma, Lyon, COD and Mexico). For better readability, weight loss was also computed as BMI (kg/m$^2$) by converting predicted TWL into predicted weights by using the formula:

$$\widehat{W}_t = W_0 \left(1 - \frac{\widehat{\mathrm{TWL}}_t^i}{100}\right) \text{ kg}\,. \tag{8.4}$$

The performances of the prediction model were calculated at each visit date and expressed by the *median absolute deviation* (MAD, defined as the median of $(|\text{TWL}_t^i - \widehat{\text{TWL}_t^i}|)_{i=1}^n$ ) which measures dispersion of predicted TWL around the true values and is robust to outliers.[2] We also calculated the *root mean squared error* (RMSE, defined as $\sqrt{1/n \sum_{i=1}^n (\text{TWL}_t^i - \widehat{\text{TWL}_t^i})^2}$ , which jointly measures the model prediction bias and variance, but is more sensitive to outliers as the square amplifies them. These two metrics were also expressed in terms of normalised ratios as % of observed BMI for each visit. We used Bland-Altman plots with actual versus predicted BMI at each specific time point (M12, M24 and M60) to assess model calibration.

### Model validation in randomised controlled trials

In addition, we used the individual data from the two randomised clinical trials (SleevePass and SM-BOSS) to replicate with our model, the previously reported comparison of RYGB versus SG in terms of weight loss (Wölnerhanssen et al., 2021). The original report combined and analysed weight follow-up data from the two studies for up to 5 years using a linear mixed model. In our study, we replaced the observed individual weight loss values with those predicted at each time point by our ML model and analysed them using the same linear mixed model described in the original report in order to compare the predicted and observed mean (95% CI) difference in weight loss between the two operations.

### Statistical analysis

Patients' characteristics were reported for each cohort as mean (SD) for continuous traits and n (%) for categorical variables. Comparison of median weight loss between two groups were performed using Mann-Whitney U test, and between three or more groups using Kruskal-Wallis one-way ANOVA. MAD and RMSE were displayed as their estimates and 95% confidence intervals (CI), estimated by bootstrap (BCa method, $n = 10{,}000$ replications). Weight loss and BMI median trajectories of participants submitted to each operation in each cohort were illustrated as a function of time using a nonlinear smoothing of the values observed at discrete times (see Appendix G.2), which is not part of the validation and performance assessment. The trajectories of patients are displayed along with prediction intervals of predicted BMI, calculated as [prediction + 25th percentile of error, prediction + 75th percentile of error].

Patient features containing more than 50% of missing values were excluded from the analysis. The remaining missing values were handled in two ways. For the LASSO analysis, the predictive mean matching method was used based on key characteristics at baseline: weight, sex, age, operation type, and presence of type 2 diabetes and its duration (Van Buuren and Groothuis-Oudshoorn, 2011). We imputed $n = 10$ sets of data, and selected variables by pooling the variable selected by LASSO in each dataset. The decision tree algorithm uses surrogate variables for the CART analysis in the case of missing data (Therneau et al., 2015).

[2]Technically, what we use is the *median absolute error* (MAE), whereas MAD is commonly defined as the median of absolute deviations from the median, which is counterpart to standard deviation for absolute errors. We purposefully use this terminology here to avoid confusion with the *mean absolute error*.

The analysis was performed using the R software version 3.6.3, the library rpart for the CART implementation, and the libraries glmnet and glinternet for the LASSO implementation. An online tool was developed based on two components: a front-end graphical user interface coded with JavaScript using the React library, and a back-end prediction and smoothing model coded in Python using the Flask microweb framework. We used the transparent reporting of a multivariable prediction model for individual prognosis or diagnosis (TRIPOD AI) guidelines (Collins et al., 2021) to report the prediction model's development and validation.

For comparison, we also analysed results from previously published postoperative weight loss prediction models (Karpińska et al., 2021) in Appendix G.6.

## 8.4 Results

### Characteristics of training and testing cohorts

The training cohort from France (n=1493), and the testing cohorts from Europe (n=7137), America (n=167), Asia (n=977), and the two randomised controlled trials used for validation (n=457), represented a total of 10,231 patients from 12 centres in 10 countries, corresponding to 30,602 patient-years. The baseline characteristics of participants of each cohort and the proportion of each type of operation performed are summarised in Table 8.1. The overall trajectories of the median (IQR) BMI and TWL observed during the five years after each operation are summarised in Figure 8.2. Individual weights at baseline (SD) ranged from 65 to 295 (25.6) kg and BMI from 26.7 to 94.1 (7.5) kg/m$^2$. Age at intervention ranged from 18 to 74 years. Among participants, 7,701 (75.3%) were female and 2,882 (28.2%) had type 2 diabetes at baseline. RYGB was the most frequent operation (n=6691, 65.4%), followed by SG (n=2872, 28.1%) and AGB (n=668, 6.5%). At 5 years, the median (IQR) TWL was 26.8% (19.8; 34.0), ranging from -13.3% to 62.7%. Overall, the general shapes of weight loss trajectories for each operation were similar among cohorts, with a nadir weight loss reached between 1 and 2 years, followed by limited weight regain afterward: median (IQR) 18.7% (4.1; 33.9) of maximal weight loss across operations. Weight regain was significantly greater after SG compared to RYGB ($p<0.0001$): 21.3% (5.3; 39.1) versus 15.0% (4.6; 27.1). At 5 years, RYGB resulted in significantly higher median (IQR) total weight loss than SG and AGB ($p<0.0001$): 28.2% (21.7; 35.1) versus 23.6% (15.2; 31.6) and 14.9% (7.2; 25.3).

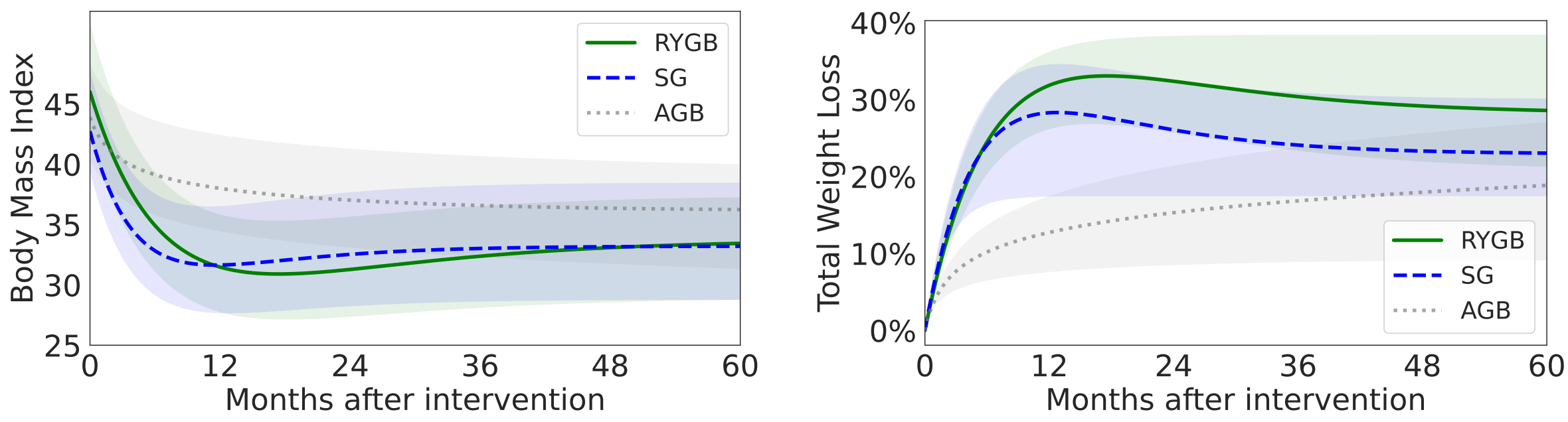

(**a**) Training cohort.

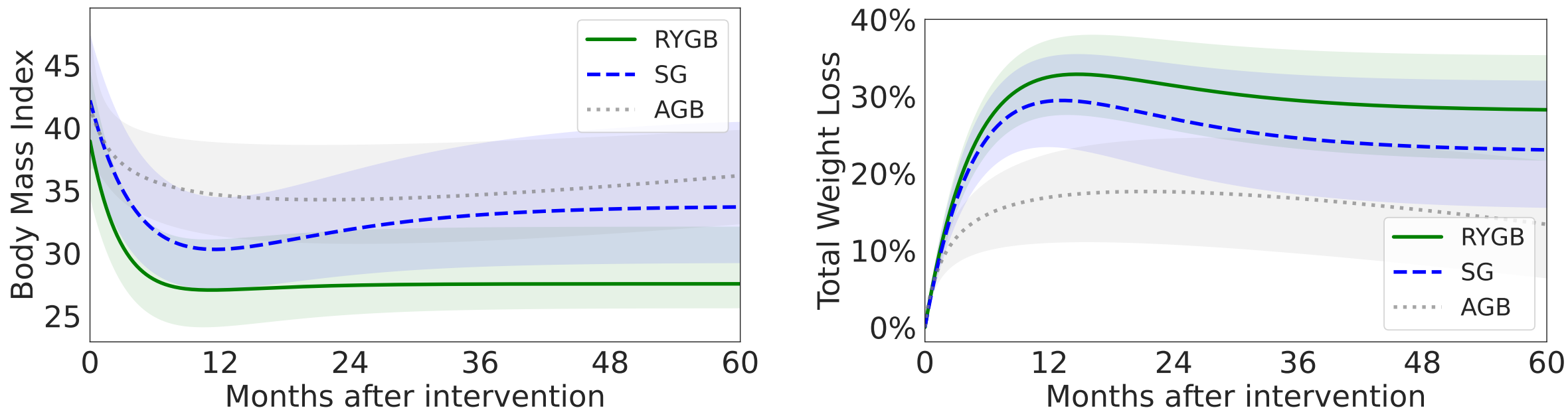

(**b**) Testing cohorts.

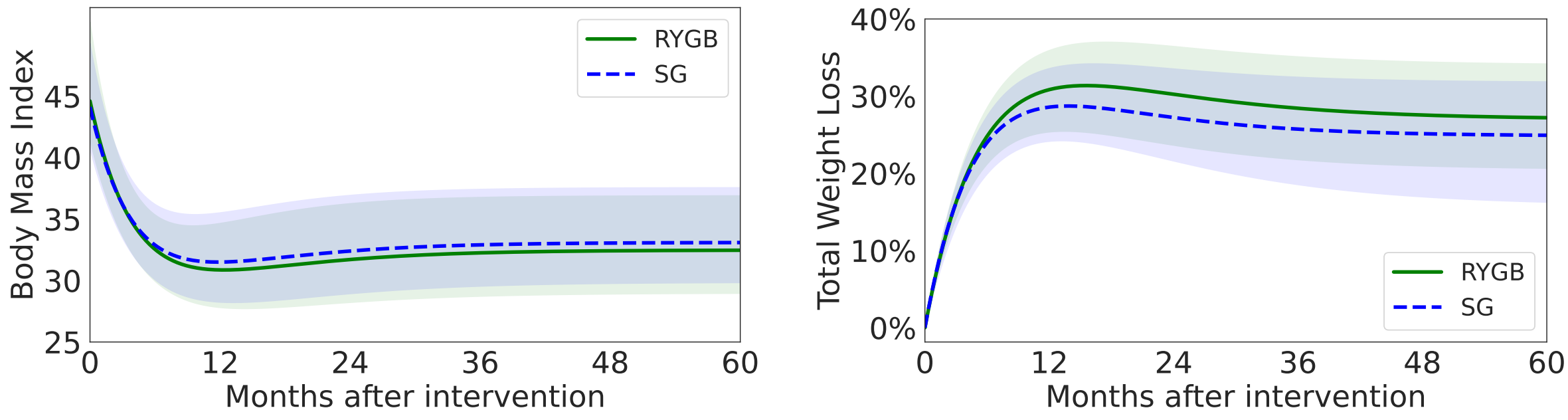

(**c**) Randomised controlled trial cohorts.

**Figure 8.2** – Smoothed observed median BMI (left) and total weight loss trajectories (right) with corresponding IQR for each operation, for the training cohort, testing cohorts, and the randomised controlled trial cohorts.

**Table 8.1** – Participant characteristics at baseline in the derivation and validation cohorts. Data are n (%) or mean (SD) unless stated otherwise. ABOS=Atlas Biologique de l'Obésité Sévère. BAREVAL=Medical Follow-up of Severe or Morbid Obese Patients Undergoing Bariatric Surgery. COD=Center for the treatment of Obesity and Diabetes, Hospital Oswaldo Cruz, Sao Paulo, Brazil. NA=not available. NOK=Nederlandse Obesitas Kliniek. PRECOS=Projet régional de REcherche Clinique en Obésité Sévère. SGH=Singapore General Hospital. SM-BOSS=Swiss Multicenter Bypass or Sleeve Study. SOS=Swedish Obese Subjects.

| | ABOS | BAREVAL | NOK | SGH | SOS | PRECOS | Roma | Lyon | COD | Mexico | SleevePass | SM-BOSS |
|---|---|---|---|---|---|---|---|---|---|---|---|---|
| Cohort size | 1147 | 346 | 5888 | 977 | 642 | 237 | 200 | 170 | 126 | 41 | 240 | 217 |
| Age | 42.1 (11.8) | 41.0 (11.7) | 44.2 (11.2) | 40.7 (10.3) | 47.3 (6.0) | 48.1 (11.6) | 44.1 (9.7) | 43.5 (10.9) | 55.7 (7.4) | 41.0 (7.6) | 48.4 (9.3) | 42.5 (11.2) |
| Sex | | | | | | | | | | | | |
| Female | 848 (73.9%) | 246 (71.1%) | 4683 (79.5%) | 632 (64.7%) | 450 (70.1%) | 178 (75.1%) | 122 (61.0%) | 117 (68.8%) | 66 (52.4%) | 36 (87.8%) | 167 (69.6%) | 156 (71.9%) |
| Male | 299 (26.1%) | 100 (28.9%) | 1205 (20.5%) | 345 (35.3%) | 192 (29.9%) | 59 (24.9%) | 78 (39.0%) | 53 (31.2%) | 60 (47.6%) | 5 (12.2%) | 73 (30.4%) | 61 (28.1%) |
| Type of operation | | | | | | | | | | | | |
| Gastric band | 223 (19.4%) | N/A | 3 (0.1%) | N/A | 376 (58.6%) | 66 (27.8%) | N/A | N/A | N/A | N/A | N/A | N/A |
| Roux-en-Y gastric bypass | 704 (61.4%) | N/A | 4801 (81.5%) | 204 (20.9%) | 266 (41.4%) | 85 (35.9%) | 165 (82.5%) | 69 (40.6%) | 126 (100.0%) | 41 (100.0%) | 119 (49.6%) | 111 (51.2%) |
| Sleeve gastrectomy | 220 (19.2%) | 346 (100.0%) | 1084 (18.4%) | 773 (79.1%) | N/A | 86 (36.3%) | 35 (17.5%) | 101 (59.4%) | N/A | N/A | 121 (50.4%) | 106 (48.8%) |
| BMI at baseline | 47.0 (7.4) | 42.6 (5.9) | 44.2 (5.7) | 42.3 (7.3) | 42.6 (4.8) | 45.2 (7.0) | 45.2 (6.0) | 45.6 (7.6) | 34.3 (2.8) | 41.7 (5.1) | 47.9 (6.9) | 43.9 (5.3) |
| Diabetes status at baseline | | | | | | | | | | | | |
| No type 2 diabetes | 296 (25.7%) | 263 ( 76.0%) | 4528 (76.9%) | 500 (51.2%) | 357 (55.6%) | 172 (72.6%) | 111 (55.5%) | 65 (38.2%) | N/A | 28 (68.3%) | 139 (57.9%) | 163 (75.1%) |
| Pre-type 2 diabetes | 454 (39.7%) | N/A | N/A | 102 (10.4%) | 151 (23.5%) | N/A | 19 (9.5%) | 1 (0.6%) | N/A | N/A | N/A | N/A |
| Type 2 diabetes | 397 (34.7%) | 83 ( 24.0%) | 1360 (23.1%) | 375 (38.4%) | 134 (20.9%) | 65 (27.4%) | 70 (35.0%) | 104 (61.2%) | 126 (100.0%) | 13 (31.7%) | 101 (42.1%) | 54 (24.9%) |
| T2D duration in years at baseline | 21.0 (23.1) | 5.3 (6.4) | N/A | 6.6 (7.7) | 3.0 (5.8) | N/A | 4.1 (4.2) | 1.8 (4.3) | 8.6 (3.0) | N/A | 6.3 (5.6) | 4.3 (6.0) |
| Smoking | 121 (10.5%) | 71 (20.4%) | 1115 (18.9%) | N/A | 166 (25.9%) | N/A | 40 (30.8%) | 12 (7.1%) | 18 (14.3%) | N/A | 52 (21.7%) | N/A |

## Development of the predictive weight loss model

**Feature selection.** Among the 447 attributes available at baseline in ABOS, 62 (14%) were excluded because of missing data, class imbalanced, or free text input. Among the 385 remaining variables, the LASSO algorithm selected seven features that were associated with TWL% at least at one postoperative visit: *preoperative weight*, *height*, *type of intervention*, *age at intervention*, *current smoking history*, *T2D status*, and *diabetes duration*. The hierarchical group-LASSO method selected the same features, as well as two additional interactions: one between *type of intervention* and *T2D*, and one between *type of intervention* and *T2D duration*. Training CART using the 9 variables selected by hierarchical group-LASSO did not modify the decision trees. The final CART algorithm therefore used the seven features selected by LASSO for the TWL regression task.

**Feature stratification.** At all postoperative times, the first and most discriminant branch of decision trees divided the population by the type of intervention, with AGB being separated from RYGB at all times. At one year, the second descendant branch separated patients by age. The following descendant branch distinguished between SG and RYGB, only in older patients, while smoking status distinguished younger patients. At two years, the second descendant branch distinguished SG from RYGB. The following branch separated SG patients by age and RYGB patients according to their diabetes status. At 5 years, AGB and SG were not separated. The second descendant branch was T2D status (RYGB patients) and age (other patients). Overall, RYGB and younger age were consistently associated with greater weight loss. Diabetes status and longer diabetes duration were always associated with less weight loss. Smoking was associated with greater weight loss, but only during the first year. Figure 8.3 illustrates the corresponding trees at the main postoperative dates (M12, M24 and M60).

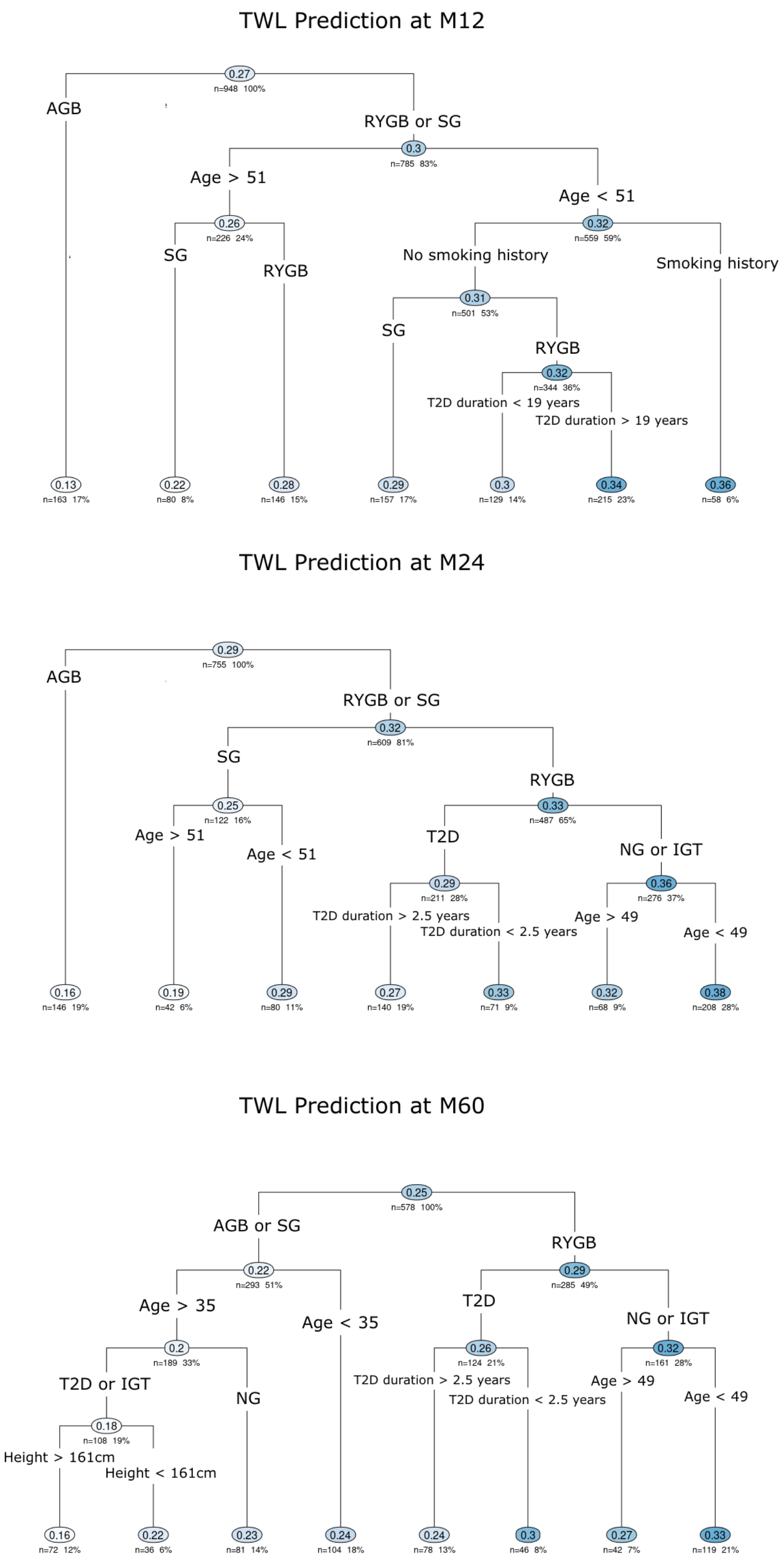

**Figure 8.3 –** Regression trees for the prediction of TWL at month M12, M24 and M60.

**Internal validation.** The model's prediction performances were evaluated at 1, 2 and 5 years in the testing subset of the training cohort (Figure 8.4). At 5 years, the estimates (95% CI) of MAD and RMSE of BMI were 3.1 (2.7 - 3.4) kg/m$^2$ and 4.9 (3.9 - 5.7) kg/m$^2$, respectively, corresponding to normalised estimates (95% CI) of MAD and RMSE of 8.9% (7.8% - 9.7%) and 14.0% (11.2% - 16.3%).

### Prediction model performance in testing cohorts

The performances of the model at 1 year, 2 years, and 5 years in the testing cohorts are detailed in Figure 8.4, as well as in Tables 8.2 and 8.3. At 5 years, the overall mean weighted values of MAD and RMSE of BMI across cohorts were 2.8 kg/m$^2$ and 4.7 kg/m$^2$, respectively, corresponding to 8.8% and 14.7% of BMI (normalised indices). The performances of the model were significantly higher in RYGB (MAD 2.8, RMSE 4.4) than in SG (MAD 3.5, RMSE 5.6) and AGB (MAD 6.1, RMSE 6.7), $p<0.0001$. Overall, the mean (SD) difference between predicted and observed BMI at 5 years was -0.3 (4.3) kg/m$^2$. The model showed satisfying calibration at all dates, as shown in Figure G.4, with no major bias identified.

### Performance of the model in randomised controlled clinical trials

We tested the model's performance to predict the overall outcome of randomised controlled trials comparing weight loss after RYGB and SG during 5 years. The results of our synthetic study based on predicted weights were in overall agreement with those of the published research, concluding to a significantly higher weight loss following RYGB as compared to SG. The mean (95% CI) difference between the two groups was 14.7 (13.7; 15.7) percent of EWL and 6.6 (6.2; 6.9) percent TWL in the synthetic study versus 7.0 (3.5; 10.5) percent of EWL and 3.2 (1.6; 4.7) percent TWL in the combined analysis of the two original studies (Wölnerhanssen et al., 2021).

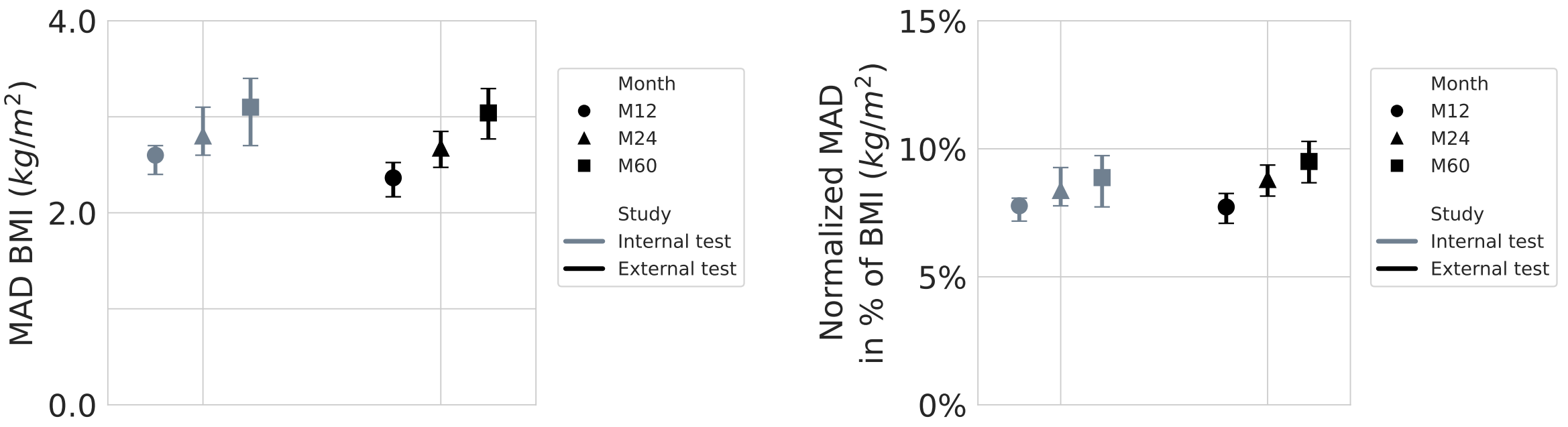

(**a**) By cohort size.

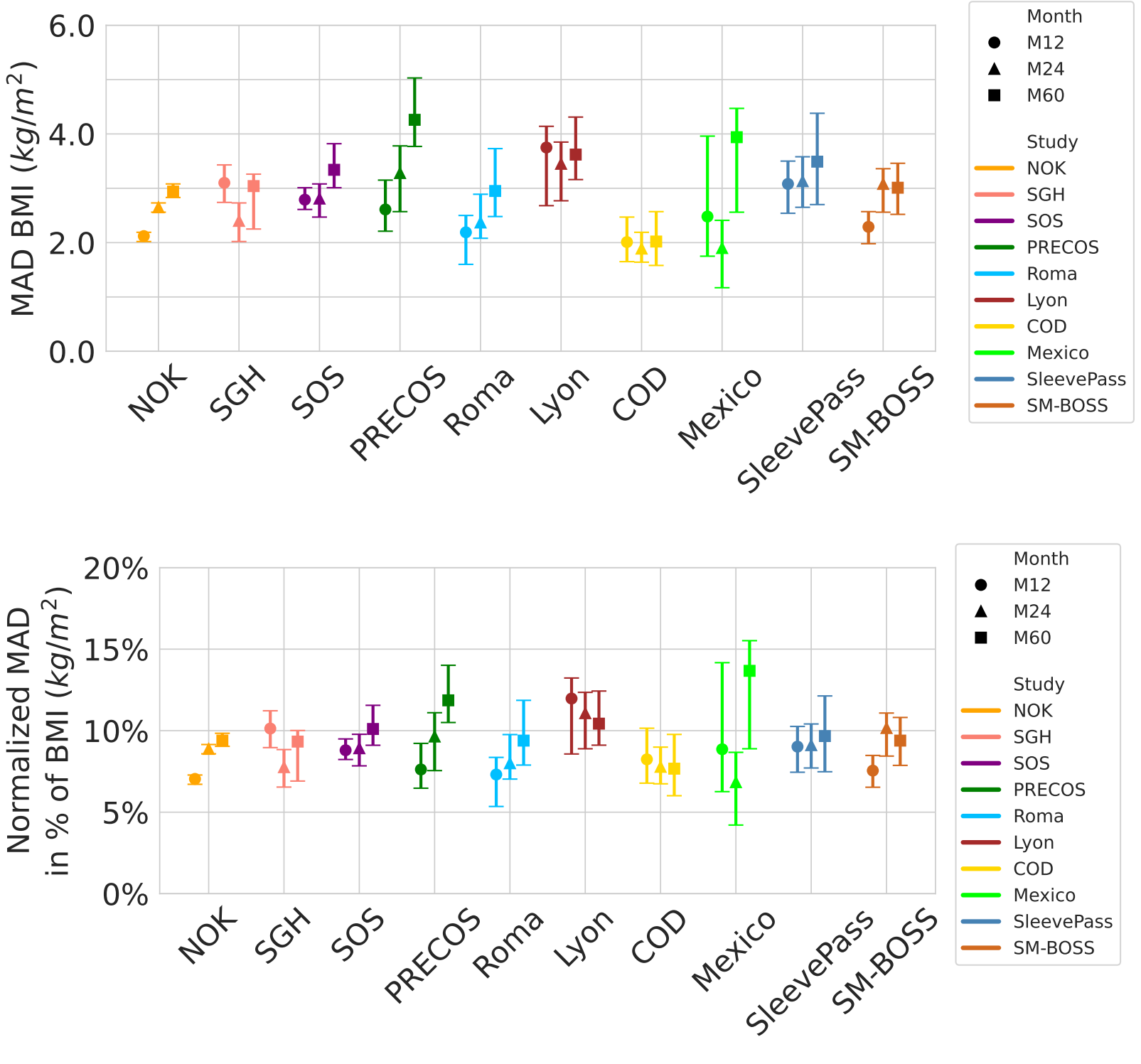

(**b**) Individual validation cohorts.

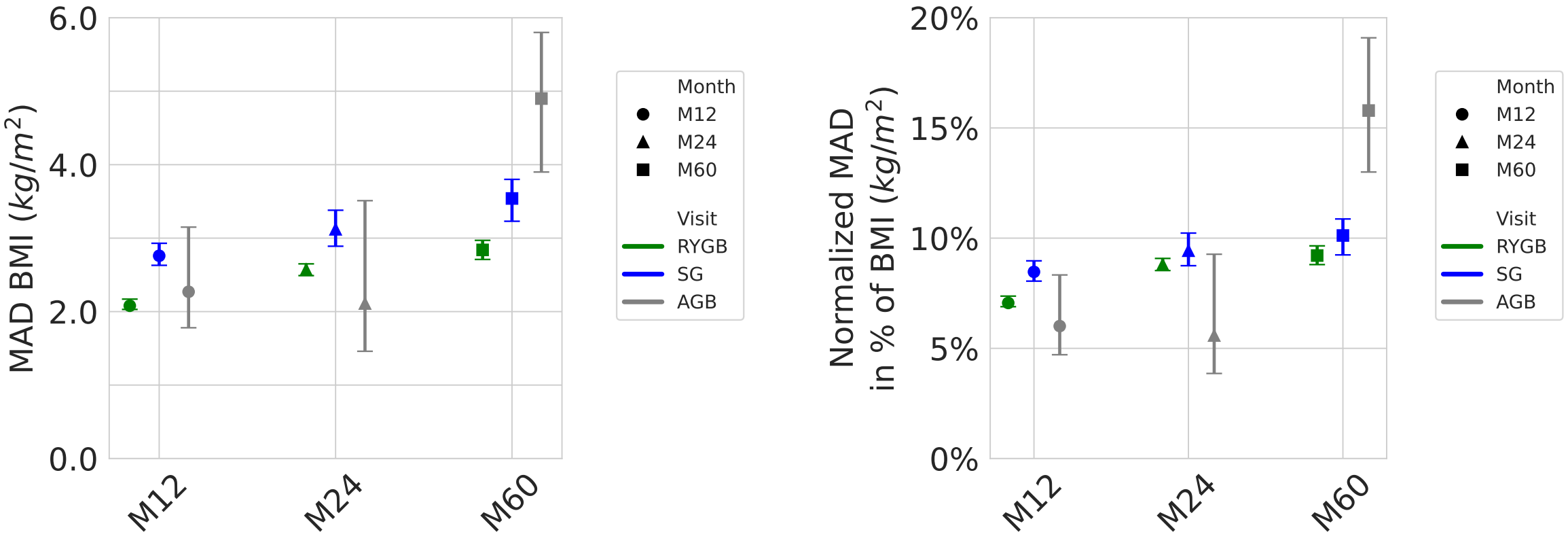

(**c**) By operation.

**Figure 8.4** – MAD (left) and normalised MAD (right) of predicted BMI outcomes, averaged by cohort size (top), individual validation cohorts (centre), and by operation (bottom).

**Table 8.2** – Comparison of predicted outcomes in validation cohorts.

| | BMI difference in kg/m$^2$ (SD) | | | RMSE in kg/m$^2$ (95% CI) | | | Normalized RMSE in % of BMI (95% CI) | | |
|---|---|---|---|---|---|---|---|---|---|
| | M12 | M24 | M60 | M12 | M24 | M60 | M12 | M24 | M60 |
| Testing subset of ABOS+BAREVAL (n=293) | -0.2 (4.1) | -0.5 (4.7) | -1.0 (4.8) | 4.1 (3.8; 4.4) | 4.7 (4.0; 5.4) | 4.9 (3.9; 5.7) | 12.3 (11.4; 13.1) | 14.0 (12.0; 16.1) | 14.0 (11.2; 16.3) |
| External validation | | | | | | | | | |
| NOK (n=5888) | 0.2 (3.3) | -0.1 (4.1) | -0.0 (4.7) | 3.3 (3.2; 3.4) | 4.1 (4.0; 4.2) | 4.7 (4.5; 4.8) | 11.2 (10.8; 11.3) | 13.9 (13.5; 14.2) | 15.0 (14.5; 15.5) |
| SGH (n=977) | -2.5 (3.7) | -0.8 (4.0) | -0.9 (4.0) | 4.4 (4.2; 4.8) | 4.1 (3.7; 4.6) | 4.0 (3.6; 4.6) | 14.5 (13.7; 15.5) | 13.2 (12.1; 14.8) | 12.4 (11.0; 14.2) |
| SOS (n=642) | 1.8 (4.4) | 0.9 (4.9) | -1.1 (4.9) | 4.8 (4.4; 5.1) | 5.0 (4.6; 5.3) | 5.0 (4.7; 5.4) | 15.1 (13.9; 16.1) | 15.9 (14.6; 16.8) | 15.1 (13.9; 16.3) |
| PRECOS (n=237) | -0.7 (4.5) | -0.0 (4.7) | -1.6 (6.0) | 4.5 (4.0; 5.0) | 4.7 (4.3; 5.3) | 6.2 (5.6; 6.8) | 13.2 (11.8; 14.7) | 13.9 (12.5; 15.5) | 17.2 (15.7; 19.0) |
| Roma (n=200) | -0.4 (3.8) | 1.3 (4.0) | 1.1 (4.6) | 3.8 (3.3; 4.4) | 4.2 (3.6; 5.2) | 4.7 (4.3; 5.3) | 12.6 (10.9; 14.6) | 14.1 (12.1; 17.4) | 15.0 (13.6; 16.8) |
| Lyon (n=170) | -1.2 (4.1) | -1.4 (4.8) | -0.5 (5.8) | 4.3 (3.7; 4.9) | 5.0 (4.3; 6.0) | 5.8 (5.1; 6.9) | 13.5 (11.9; 15.8) | 16.0 (13.7; 19.4) | 16.7 (14.8; 19.8) |
| COD (n=126) | -0.4 (3.1) | -1.4 (2.2) | -1.7 (2.2) | 3.1 (2.8; 3.5) | 2.6 (2.3; 2.9) | 2.8 (2.5; 3.1) | 12.7 (11.3; 14.3) | 10.6 (9.6; 11.8) | 10.6 (9.6; 11.7) |
| Mexico (n=41) | -0.9 (4.2) | -1.0 (3.5) | -0.3 (5.5) | 4.3 (3.4; 5.5) | 3.6 (2.7; 4.6) | 5.4 (4.4; 7.1) | 15.4 (12.3; 19.7) | 12.9 (9.9; 16.4) | 18.8 (15.2; 24.6) |
| SleevePass (n=240) | -1.0 (4.4) | -0.2 (4.7) | 0.2 (5.1) | 4.5 (4.1; 4.9) | 4.7 (4.3; 5.1) | 5.1 (4.7; 5.7) | 13.2 (12.1; 14.5) | 13.6 (12.4; 15.0) | 14.2 (12.9; 15.8) |
| SM-BOSS (n=217) | -1.1 (3.6) | 0.1 (4.4) | 0.2 (5.0) | 3.7 (3.4; 4.2) | 4.4 (4.0 ; 4.9) | 4.9 (4.5; 5.7) | 12.4 (11.2; 13.8) | 14.4 (13.1; 16.0) | 15.5 (14.0; 17.7) |
| Average (weighted by cohort sizes) | -0.1 (3.5) | -0.1 (4.2) | -0.3 (4.7) | 3.7 (3.5; 3.9) | 4.2 (4.0; 4.5) | 4.7 (4.4; 5.0) | 12.0 (11.4; 12.6) | 14.0 (13.2; 14.8) | 14.7 (13.8; 15.7) |

**Table 8.3** – Comparison of predicted outcomes by operation in validation cohorts.

| | BMI difference in kg/m$^2$ (SD) | | | RMSE in kg/m$^2$ (95% CI) | | | Normalized RMSE in % of BMI (95% CI) | | |
|---|---|---|---|---|---|---|---|---|---|
| | M12 | M24 | M60 | M12 | M24 | M60 | M12 | M24 | M60 |
| RYGB | -0.0 (3.2) | -0.4 (3.9) | -0.3 (4.5) | 3.2 (3.2; 3.3) | 3.9 (3.9; 4.0) | 4.5 (4.3; 4.6) | 11.0 (10.8; 11.2) | 13.5 (13.2; 13.8) | 14.6 (14.1; 15.0) |
| SG | -0.4 (4.3) | 1.0 (4.8) | 0.9 (5.6) | 4.3 (4.2; 4.5) | 4.9 (4.7; 5.2) | 5.7 (5.4; 6.0) | 13.2 (12.7; 13.8) | 14.9 (14.2; 15.6) | 16.2 (15.3; 17.2) |
| AGB | 1.7 (3.9) | 0.7 (4.1) | -2.8 (4.3) | 4.7 (4.2; 5.4) | 4.7 (4.1; 5.3) | 6.0 (5.4; 6.7) | 13.6 (11.9; 15.3) | 13.6 (12.0; 15.3) | 16.6 (14.9; 18.4) |

### Comparison with alternative models

The performances of our decision tree model were compared in training and testing cohorts with benchmark models based on simple regression, linear mixed effect, random forest, and CART with pruning. The decision tree approach outperformed simple regression at M12, M24 and M60 in both internal and external test data. The linear mixed model using the seven selected variables, and modelling time as restricted cubic splines, with random intercept and slope, did not outperform the decision tree model on both internal and external test data. Random forest resulted in only marginally lower MAD estimates (confidence intervals largely overlapped). CART with pruning selected only six variables: weight, height, age, T2D status and duration, and the type of intervention, as compared to the seven variables with LASSO+CART, which additionally selected smoking history. Triaging the variables with LASSO in the first place resulted in better performances at M12, the only time point where smoking history appears in the decision trees. Full results are reported in Figure G.5.

### Comparison with models previously published in the literature

We identified 12 models that have been previously proposed to predict weight loss following one or more of the three interventions analysed in the present study, during 1 to 5 years (Baltasar et al., 2011; Wise et al., 2016; Goulart et al., 2016; Seyssel et al., 2018; Janik et al., 2019; Velázquez-Fernández et al., 2019; Cottam et al., 2018). The accuracy of weight loss predicted by these models, estimated by RMSE and normalised RMSE in the appropriate subset of patients from the ABOS cohort, ranged from 3.8 to 6.1 kg/m$^2$ and from 11.7 to 17.4 percent of BMI at one year (6 models), compared to averages of 3.7 kg/m$^2$ and 12.0 percent of BMI for our current model. At 2 years, existing models ranged from 4.9 to 7.0 kg/m$^2$ and 15.5 to 20.2 percent of BMI (5 models), compared to 4.2 kg/m$^2$ and 14.0 percent of BMI for our current model. Finally, RMSE and normalised RMSE evaluated at 5 years among existing models were 5.4 kg/m$^2$ and 15.8 percent of BMI (only 1 model), compared to 4.7 kg/m$^2$ and 14.7 percent of BMI for our current model. Moreover, using the Diebold-Mariano test (Harvey et al., 1997) to compare forecasts from two competing prediction models, we established that the increased accuracy of our model was significant compared to at least half the benchmark models. The individual results of these models are detailed in Tables G.1 and G.2.

### Online weight loss trajectory calculator

The machine learning model developed in the present study was then integrated into a software[3] allowing to display the 5-year weight trajectory that can be expected for a given patient prior to the intervention according to the seven key baseline characteristics included in the model. The graphical output of the model was presented and discussed among investigators and patient representatives. The resulting user-friendly calculator displays the predicted weight trajectory at any given time, alongside prediction intervals corresponding to IQR of prediction errors. By default, individual predicted trajectories are expressed in kg. According to the user's choice, results can also be displayed in kg/m$^2$, % of TWL, or % of EWL. To improve readability, predicted trajectories over time for each of these metrics are displayed as smooth curves. Figure 8.5 displays two illustrative examples of BMI and TWL trajectories predicted for individual patients from the current version of the calculator available online (https://bariatric-weight-trajectory-prediction.univ-lille.fr, see Figure 8.6).

[3]This software is owned by Inria, CHU Lille and Université de Lille, and is protected by the digital deposit IDDN.FR.001.370002.000.S.P.2022.000.31230, Agence pour la Protection des Programmes (APP).

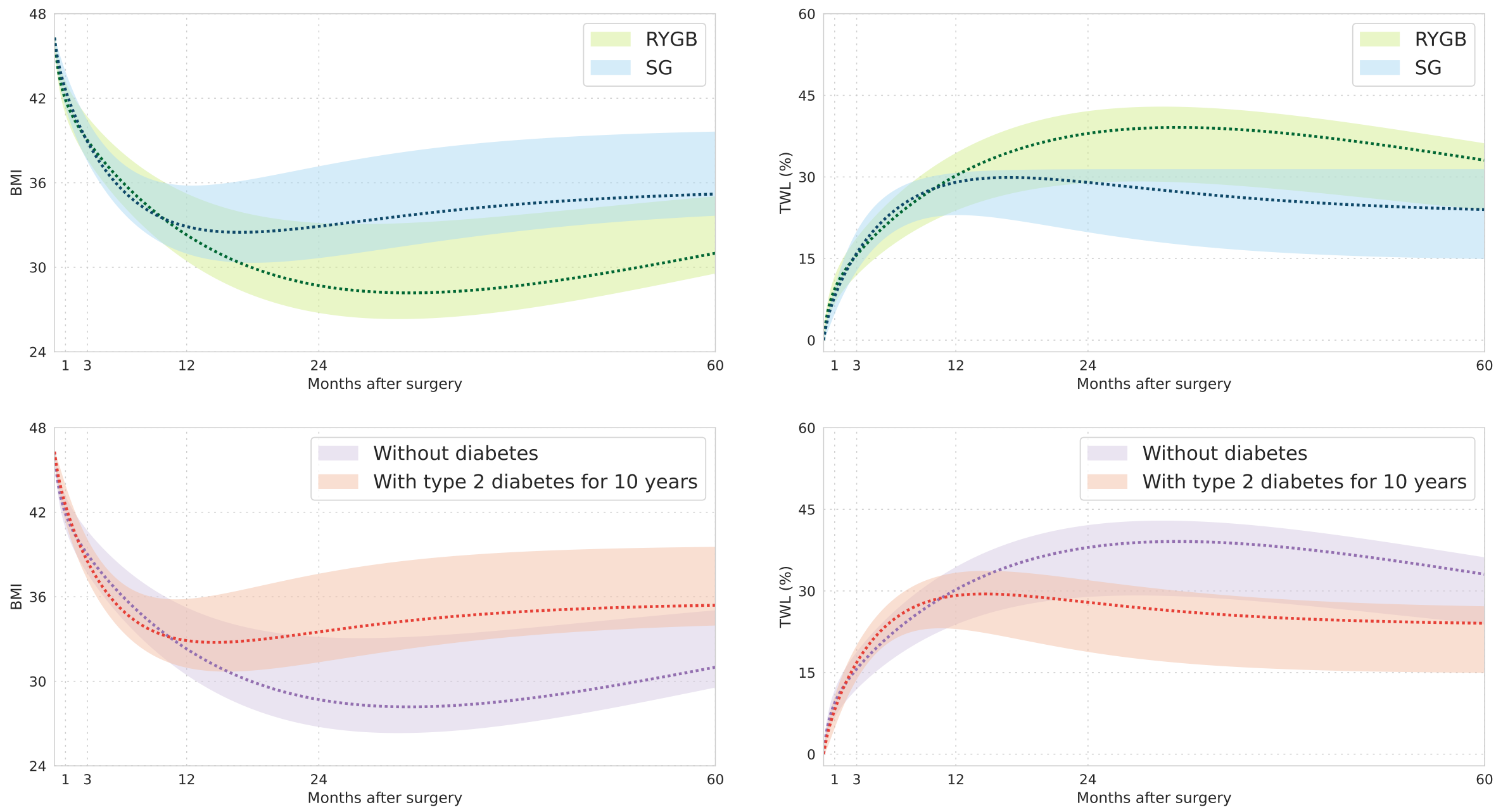

**Figure 8.5 –** Predicted trajectory and IQR of BMI (left) and total weight loss (right) for a 30-year-old patient with a weight of 150 kg, a height of 1·80 m, who was a non-smoker, without diabetes, undergoing Roux-en-Y gastric bypass and sleeve gastrectomy (top); and for a 30 years old patient with a weight of 150 kg, height of 1·80 m, who was a non-smoker undergoing Roux-en-Y gastric bypass without diabetes and with type 2 diabetes with 10 years duration (bottom).

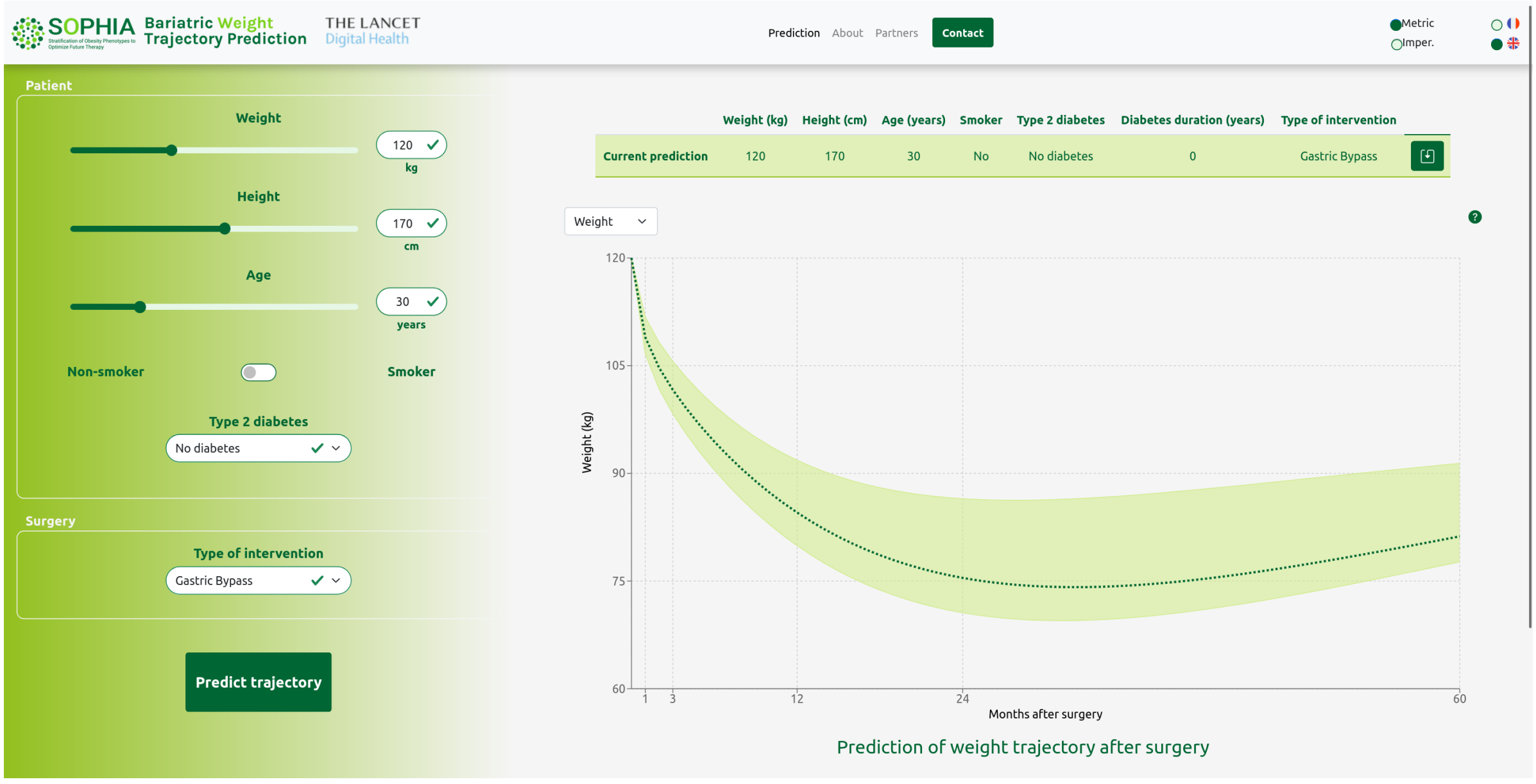

**Figure 8.6** – Bariatric Weight Trajectory Prediction website.

## 8.5 Discussion

We developed a machine learning model that provides accurate individual weight trajectories expected during 5 years after bariatric surgery, based on seven simple preoperative variables, namely age, weight, height, smoking history, T2D status and duration, and the type of intervention. These variables are readily available in a variety of clinical settings without interpretation and do not require laboratory tests. The model was validated globally, in 8 cohorts and two randomised controlled trials, in Europe, the Americas, and Asia, and incorporated in an easy-to-use and interpretable web-based tool providing individual preoperative prediction of postoperative weight loss trajectory.

This accessible and interpretable model is the first to provide preoperative predictions of weight trajectories up to 5 years after surgery, simultaneously for three of the most common types of surgery, i.e. RYGB, SG and AGB. Our results highlighted the association of the type of operation and diabetes status with weight trajectories. The present study also showed the impact of diabetes duration and smoking, which were not previously included in weight loss surgery prediction models.

As expected, the type of surgery was the first node to appear in the decision tree. Interestingly, SG and RYGB were not distinguishable at one year but separated soon after. This is consistent with recent literature (Peterli et al., 2012). While some early reports showed similar weight loss after RYGB and SG at one (Peterli et al., 2012) and two years (Fischer et al., 2012), a meta-analysis of randomised clinical controlled trials (Wölnerhanssen et al., 2021) as well as a large matched controlled cohort study have shown the superiority of RYGB as compared with SG at 5 years. Additionally, a synthetic emulation of the combined analysis of the two randomised controlled trials (Wölnerhanssen et al., 2021) based on our model, resulted in a similar conclusion to the original report, albeit resulting in a larger difference between the two operations. This illustrates that differences in weight loss outcomes may be larger in non-randomised settings where patient and care provider preferences drive procedure selection (Arterburn et al., 2021).

Several studies have already suggested that weight loss is lower in patients with T2D, particularly in those with uncontrolled diabetes (Parri et al., 2015; Shantavasinkul et al., 2016; Diedisheim et al., 2021). In the Longitudinal Assessment of Bariatric Surgery (LABS) study, the presence of T2D at baseline was associated with reduced weight loss after surgery (Courcoulas et al., 2015). In the present study we found that TWL also lowers when duration of diabetes increases. Diabetes duration is a predictor of disease severity, and a proxy of declined $\beta$-cell function, which was also associated with lower weight loss one year post bariatric surgery (Borges-Canha et al., 2021).

Most of the studies that looked at preoperative smoking patterns did not demonstrate any association between smoking and postoperative weight loss (Mohan et al., 2021). A minority of studies reported small differences in postoperative weight loss between smokers and nonsmokers (Mohan et al., 2021); these differences became non-significant with longer follow-up, which is consistent with our model, which splits the population based on history of smoking only during the first year after surgery. Of note, our prediction algorithm did not identify sex as a significant predictor of postoperative weight loss, in line with several previous reports (Parri et al., 2015; Mousapour et al., 2021; Tankel et al., 2020).

In contrast to previous studies, relying on handpicked features and linear models, the present study leverages a machine learning approach. LASSO is an alternative to multivariate regression that enforces sparsity in the covariates used for prediction. Notably, all variables selected by LASSO in the present study were clinical traits, as opposed to the many continuous biological variables available in ABOS. The CART algorithm learns a tree stratification of covariates that identifies relationships beyond the scope of traditional analysis techniques (Barnholtz-Sloan et al., 2011). Tree-based models are well-suited to capture nonlinear effects of mixed nominal, ordinal, and continuous attributes and have been recently shown to outperform deep learning on tabular data (Grinsztajn et al., 2022). In addition, this learned stratification of patient attributes allows for a clear interpretation of the predicted outcomes. CART also provides critical variable thresholds and their directional influences on the outcomes. Unlike multivariate linear regression approaches, CART learns nonparametric models that do not require a strong specification of the mapping between covariates and outcomes. These strengths are especially valuable in addressing the data heterogeneity commonly associated with clinical datasets. Descending in the decision tree, it is possible to draw the involvement of different features, the combination of which allows to define with higher accuracy the clinico-biological characteristics of an individual with higher or lower weight loss. Another remarkable feature is that by exploring the tree from root to leaves, CART divided the population based on diabetes status or diabetes duration, only for patients submitted to RYGB.

Our study has several limitations. First, the machine learning algorithm selected only seven simple clinical features. It is possible that the prediction can be improved with more performant classes of algorithms, such as random forest or deep learning (Ge and Wong, 2008). However, such methods would require more training data and provide a less interpretable model, which we consider as a decisive criterion for implementation in patients' care settings. Indeed, more complex models are not straightforwardly interpretable by humans, undermining trust (Stiglic et al., 2020). Concerns about such black box algorithms are increasingly highlighted as one of the primary barriers to the adoption of ML in the healthcare context (Petch et al., 2022). Second, the extensive clinical and biological data set used for model development only included a limited number of socioeconomic, ethnic, behavioural, and nutritional aspects, which may influence postoperative weight loss trajectories. Likewise, we did not evaluate the added value of genetic analyses (de Toro-Martin et al., 2018; Antoine et al., 2022) or of new disease stratification (Raverdy et al., 2022). Third, our analysis was limited to the three most performed operations worldwide. AGB has been less frequently performed in recent years, and several new and rapidly increasing operations, such as one anastomosis gastric bypass or endoscopic sleeve gastrectomy, were not included because of a scarcity of a large number of patients with 5 years of follow-up data. In addition, we did not include reoperations in our model. Fourth, except for the SGH, COD, and Mexico cohorts, the majority of individuals enrolled in our study were White, especially those submitted to AGB. Therefore, our results should be further replicated in non-White populations. Finally, our study was focused primarily on weight loss and not obesity complications, like type 2 diabetes, hypertension, or non-alcoholic fatty liver disease (Bauvin, 2020). We also did not analyse the risks associated with surgery (Thereaux et al., 2019), but we appreciate that these are also essential to inform clinical decision.

In summary, we have developed and validated an easy to use and interpretable model that provides individual predictions of weight loss trajectory after bariatric surgery. We have shown its generalisability and transportability across multiple cohorts in Europe, America, and Asia, as well as its performance in intervention clinical trials.

Individual weight loss trajectory prediction appears to be an accurate and simple strategy to inform clinical decisions for both healthcare providers and patients prior to surgery. Our model can also be used postoperatively to identify patients whose actual weight loss trajectories differ from their predicted trajectory, thus allowing the timely implementation of appropriate clinical interventions.

## 8.6 Extensions and related works

We conclude by summarising several follow-up works around the use of machine learning in bariatric surgery and various extensions of our weight loss prediction model. Except the first one, these studies have not yet been published and are awaiting further data collection and analyses. Accordingly, we only provide a high-level description and share early results.

**Causal analysis of the impact of robotic assistance in bariatric surgery.** While the benefits of robotic assistance were established at the beginning of surgical training, there is limited data on the robot's influence on experienced bariatric laparoscopic surgeons. We conducted a retrospective study using the BRO clinical database (2008-2022), collecting data from patients operated on in expert centres. We compared the rate of serious postoperative complications (defined as a Clavien score equal to 3 or higher) in patients undergoing metabolic bariatric surgery with and without robotic assistance. From a set of 14 explanatory variables, we constructed a *directed acyclic graph* (DAG) from expert knowledge to identify an adjustment set used in a multivariate linear regression to account for causality, as well as a propensity score matching to calculate the *average treatment effect* (ATE) of robotic assistance. We further validated the conclusions of these standard methods with recent techniques for causal inference, namely *targeted maximum likelihood* (TMLE, Schuler and Rose (2017)) and *double debiased machine learning* with boosted trees learners (DML, Chernozhukov et al. (2018)), that are able to accurately estimate both outcomes and confounding mechanisms simultaneously.

The study included 35,043 patients (24,428 SG, 10,452 RYGB and 163 single anastomosis duodenal-ileal bypass with sleeve gastrectomy [SADI-S]), with 938 operated on with robotic assistance (801 SG, 134 RYGB, and 3 SADI-S), among 142 centres in France. Overall, we found no benefit of robotic assistance regarding the risk of complications (ATE $= -0.05$, $p = 0.794$), with no difference among derivation surgeries (RYGB and SADI-S, $p = 0.322$) but a negative trend in the SG group towards a higher risk of complications ($p = 0.060$). Length of hospital stay was significantly decreased in the robot group ($3.7 \pm 11.1$ vs $4.0 \pm 9.0$ days, $p < 0.001$). Further details may be found in the published version of this study (Caiazzo et al., 2023).

**Extension to OAGB.** In an effort to support the ever-growing prevalence of severe obesity, other types of bariatric interventions have been developed to complement the existing range of surgical options. An example of such recent advances is the laparoscopic *one-anastomosis gastric bypass* (OAGB), a derivation surgery similar to RYGB. We are currently gathering a cohort of patients to (i) investigate the accuracy of the RYGB weight predictor for this operation, and (ii) retrain the tree-based model described above to provide OAGB-specific predictions.

**Extension to teenagers and young adults.** While the ABOS cohort comprised patients aged 18 or older, young adults (25 or younger) represented only 10% of the study population, which raises the question of the applicability of the prediction model for paediatric bariatric surgery. Weight loss surgical interventions have been performed on teenagers in France since the late 2000s in expert centres and under strict conditions for preparation and follow-up, to address the issue of increased prevalence of severe obesity at a young age. We have enrolled 329 patients from 5 centres in France and we are looking to expand this dataset by collaborating with other centres across Europe. This initial cohort was presented at the 32nd Annual Congress of the *European Childhood Obesity Group* (Thivel et al., 2023). Here again, we expect to retrain the regression model to provide predictions tailored to the case of paediatric surgery.

**Added value of genetic analysis.** Finally, we mention a natural extension of our model to incorporate genetics to further personalise and refine weight loss predictions. The elementary block of genetic information is the *single nucleotide polymorphism* (SNP), representing a genomic variant at a single base position in the DNA, which can be acquired e.g. by genotyping from a blood sample. A rigorous way to identify significant correlations between SNPs and phenotype (i.e. the observable traits, in our case weight loss) is to perform a *genome-wide association study* (GWAS). Such studies are however difficult to implement as they require large samples of individuals to support the high-dimensional SNP statistics and prevent the discovery of spurious correlations. An alternative to GWAS consists in calculating a *polygenic risk score* (PRS) for a specific outcome, typically the average of a subset of SNPs weighted by the odds ratio or regression weights of a logistic or linear regression on said SNPs. For instance, de Toro-Martin et al. (2018) considered 186 SNPs previously associated with BMI and predicted the long-term success or failure of bariatric surgery, as defined by a target of excess weight loss (EWL), for an intervention called *biliopancreatic diversion with duodenal switch* in a population of n=865 patients. Denoting by $\mathrm{SNP}_i \in \mathbb{R}^{186}$ the vector of SNPs and by $Y_i$ the Bernoulli random variable encoding the success of surgery for patient $i \in \{1, \dots, n\}$, they fitted a logistic model

$$\mathbb{E}\left[Y_i \mid \mathrm{SNP}_i\right] = \sigma\left(\langle \beta, \mathrm{SNP}_i \rangle\right) , \tag{8.5}$$

where $\beta \in \mathbb{R}^{186}$ and $\sigma \colon z \mapsto 1/(1+\exp(-z))$ is the sigmoid function. The corresponding PRS for patient $i$ was then derived by the formula

$$\mathrm{PRS}_i = \sum_{j=1}^{186} \mathrm{OR}_j \mathrm{SNP}_{i,j}, \quad \text{where} \quad \mathrm{OR_j} = e^{\beta_j} . \tag{8.6}$$

As part of an ongoing work, we are investigating the added value of this PRS in the ABOS cohort for predicting weight loss (which is a continuous outcome, rather than a binary one as in the logistic model). Early results suggest a small improvement at M60 for SG and AGB, effectively reducing prediction bias, but no significant effect for RYGB and earlier postoperative dates. Of note, we arrived at a similar conclusion by retraining the PRS in the ABOS cohort on continuous weight loss prediction directly. Interestingly, RYGB is expected to be less sensitive to genetic information as it radically alters the metabolism (more so than competing interventions); similarly, initial weight loss (up to 2 years after surgery) is conditioned by the intervention and the phenotype of the patient (such as the presence of type 2 diabetes), while weight regain (between 2 and 5 years) appears to be affected by the environment and genetic variations.

**Take-home message:**

## ☞ We can forecast weight loss trajectories after bariatric surgery from a small number of simple clinical features using a white box model.

We have implemented a simple, highly interpretable machine learning pipeline to predict postoperative weights in a heterogeneous population and validated it internationally. This tree-based approach offers an alternative to the widespread linear models ubiquitous in the medical literature on obesity. In particular, our model was able to discriminate patients based on a small number of attributes, including:

- **intervention type**: gastric bypass and sleeve gastrectomy induce differentiated long term weight trajectories, with higher weight loss potential for the former;
- **age**: surgery is more efficient on younger patients;
- **type 2 diabetes**: hinders weight loss, more so for severe (longer duration) diabetes.

We foresee two potential uses for this tool:

- At the **preoperative** stage: to help surgery candidates visualise and anticipate their future weight loss;
- At the **postoperative** stage: to help medical staff identify complications (significant deviations from weight trajectories predicted before surgery).

The latter may pave a way towards building a data-driven recommender system to improve postoperative care.

# Chapter 9

# Conclusion and perspectives

*Anicetus villam statione circumdat refractaque ianua obvios servorum abripit, donec ad fores cubiculi veniret; cui pauci adstabant, ceteris terrore inrumpentium exterritis. cubiculo modicum lumen inerat et ancillarum una, magis ac magis anxia Agrippina, quod nemo a filio ac ne Agermus quidem: aliam fore laetae rei faciem; nunc solitudinem ac repentinos strepitus et extremi mali indicia. abeunte dehinc ancilla, "tu quoque me deseris?"*

— Tacitus, *The Annals*

In this concluding chapter, we address potential future research directions for both the mathematical foundations and the medical applications of sequential learning.

**Short-term.** We would like to further develop the second order sub-Gaussian concentration studied in Chapter 4 and the Dirichlet sampling scheme for linear bandits introduced in Chapter 7. For the former, we believe a better understanding of this new model specification (in particular, the role of the parameter $\rho$) would help strengthen intuition regarding which probability distribution fits within this new scope and facilitate its adoption as a practical modelling alternative. For the latter, we plan to liaise with experts on strong approximation to lift the theoretical roadblocks towards a formal regret analysis. Both are intended to be submitted for publication in the near future.

On the medical front, we have hinted at several direct extensions of the weight trajectory prediction model in Chapter 8, to include more operations (OAGB), more patients (paediatrics) and more predictor variables (genetics). These are currently ongoing works and are expected for publication in the coming year.

**Medium-term.** We intend this thesis to fit within a recent trend in machine learning and statistics that calls for more transparent and efficient empirical studies. We hope to draw attention to the importance of safe, anytime-valid statistics, heralded by the recent advances in $E$-processes. We also advocate for nonparametric methods, including Dirichlet sampling strategies but also other subsampling and resampling schemes, permutation tests, etc., as a powerful way to mitigate model misspecification. We hope the contributions we made in this thesis foster more research in these directions and raise awareness among practitioners.

Furthermore, we emphasise that the weight prediction model presented in Chapter 8 is only the first step in the collaboration between Inria Scool and Lille University Hospital. Going further, we hope to make it a proof of concept to develop prediction models for other outcomes, starting with blood glucose for patients living with type 2 diabetes, and progressively building towards long-term postoperative complications. In addition to the problem of predicting, we also intend to address the *learning* of follow-up policies, i.e. recommending visits and postoperative pathways to reduce complications and optimise costs for both the medical staff and the patients. During the course of this thesis, Inria Scool and Lille University Hospital have obtained a grant from the French national research agency (ANR) to fund future work on these aspects under the *BIP-UP* project (https://anr.fr/Project-ANR-22-CE23-0031).

**Long-term.** Last, a major practical limitation of the prediction model we developed for bariatric surgery is that it only operates at the preoperative stage. Ideally, we would like to be able to add new information after surgery, for instance weights measured during follow-up visits, to refine the prediction at later visits. Beyond the development and validation of such a new model, this raises the question of the integration of this tool within a digital health platform. Indeed, healthcare policymakers in France recently promoted the use of shared medical records (the nationwide *Dossier Médical Partagé*, as well as regional initiatives such as the *Prédice* programme in the Hauts-de-France), which we believe could constitute a promising venue for the dissemination of our tool; for instance, weights measured at any visit to the general practitioner or the hospital could feed automatically into a postoperative model and update future predictions on the fly. We anticipate that this would require a significant, long-term effort of coordination and communication across actors in academic research, health administration, medical professionals, IT service providers, data regulation and politics.

# Appendix A

# Background material on statistical sequential decision-making models

## Contents

## A.1 Technical results on stochastic bandits

**Comments on the definition of stochastic bandits.** Alternatively to the reward process introduced in Definition 1.2, we could have defined a reward process for all arms $\widetilde{\mathbb{Y}} = (\widetilde{Y}_t^k)_{k\in[K],t\in\mathbb{N}} \sim \boldsymbol{\nu}^{\otimes\mathbb{N}}$ and considered instead the reward process $\widetilde{\mathbb{Y}}^\pi = (\widetilde{Y}_t^{\pi_t})_{t\in\mathbb{N}}$. Importantly, the process $\widetilde{\mathbb{Y}}$ is *not* adapted to the natural bandit filtration since for $t \in \mathbb{N}$ and $k \in \mathbb{K}\backslash\{\pi_t\}$, $\widetilde{Y}_t^k$ is not $\mathcal{G}_t^\pi$-measurable. In other words, a bandit model represents a partial information setting: the random variable $\pi_t$ is the arm pulled by the agent at time $t \in \mathbb{N}$, and only the corresponding reward random variable is observed, not the rewards corresponding to the *arms not taken*. Both formulations are actually equivalent. Indeed, let $(\tau_n^{k,\pi})_{k\in[K],n\in\mathbb{N}} = (\inf\{t \in \mathbb{N},\ N_t^{k,\pi} = n\})_{k\in[K],n\in\mathbb{N}}$ be the sequence of times when each arm is pulled according to policy $\pi$. Then for $k \in [K]$, Doob's optional skipping theorem (Kallenberg, 1997; Chow and Teicher, 2003) applied to the i.i.d. sequence $(\widetilde{Y}_t^k)_{t\in\mathbb{N}}$ and increasing sequence $(\tau_n^{k,\pi})_{n\in\mathbb{N}}$ shows that $(\widetilde{Y}^k_{\tau_i^{k,\pi}})_{i=1}^n \overset{\mathcal{D}}{=} (\widetilde{Y}_i^k)_{i=1}^n$, i.e. they are equally distributed as $\nu_k^{\otimes n}$ for all $n \in \mathbb{N}$, and thus the processes $\mathbb{Y}^\pi$ and $\widetilde{\mathbb{Y}}^\pi$ are equivalent. Both the global ($\mathbb{Y}$, running time $t$) and local ($\widetilde{\mathbb{Y}}$, counting pulls $N_t^k = n$) points of view may be convenient depending on the situation. For clarity, we use the following notations throughout this thesis, where $k \in [K]$ represents an arm.

- **Global**: at time $t \in \mathbb{N}$,
  - reward history: $\mathbb{Y}_t^k = (\widetilde{Y}_{\tau_i^k}^k)_{i=1}^{N_t^k} \overset{\mathcal{D}}{=} (Y_i^k)_{i=1}^{N_t^k} \sim \nu_k^{\otimes N_t^k}$;
  - empirical measure $\widehat{\nu}_t^k = \frac{1}{N_t^k} \sum_{i=1}^{N_t^k} \delta_{\widetilde{Y}_{\tau_i^k}^k} \overset{\mathcal{D}}{=} \frac{1}{N_t^k} \sum_{i=1}^{N_t^k} \delta_{Y_i^k}$;
  - empirical mean $\widehat{\mu}_t^k = \frac{1}{N_t^k} \sum_{i=1}^{N_t^k} \widetilde{Y}_{\tau_i^k}^k \overset{\mathcal{D}}{=} \frac{1}{N_t^k} \sum_{i=1}^{N_t^k} Y_i^k$.
- **Local**: at sample size $n \in \mathbb{N}$,
  - reward history: $\mathbb{Y}_{(n)}^k = (\widetilde{Y}_{\tau_i^k}^k)_{i=1}^{n} \overset{\mathcal{D}}{=} (Y_i^k)_{i=1}^{n} \sim \nu_k^{\otimes n}$;
  - empirical measure $\widehat{\nu}_{(n)}^k = \frac{1}{n} \sum_{i=1}^{n} \delta_{\widetilde{Y}_{\tau_i^k}^k} \overset{\mathcal{D}}{=} \frac{1}{n} \sum_{i=1}^{n} \delta_{Y_i^k}$;
  - empirical mean $\widehat{\mu}_{(n)}^k = \frac{1}{n} \sum_{i=1}^{n} \widetilde{Y}_{\tau_i^k}^k \overset{\mathcal{D}}{=} \frac{1}{n} \sum_{i=1}^{n} Y_i^k$.

Going further, to avoid cluttering, we only use the second notation $\mathbb{Y}$, thus implicitly referring to the skipped process $(\widetilde{Y}_{\tau_n^k}^k)_{n\in\mathbb{N}}$ when adopting the global point of view. Moreover, for $T \in \mathbb{N}$, we use the notations $\mathbb{Y}_T^\pi = (Y_t^{\pi_t})_{t=1}^T$ and $\boldsymbol{\nu}_\pi^{\otimes T} \in \mathcal{M}_1^+(\mathbb{R}^T)$ to denote the distribution of this sequence. Note that this is *not* a product measure (the sequence $\mathbb{Y}_T^\pi$ is not independent) even though $\boldsymbol{\nu}$ is, since the policy $\pi$ introduces a coupling between the arms. In simpler words, if for instance $\pi$ is designed to select arms that generated high rewards in previous rounds, future rewards are partially *caused* by past rewards — if arm $k$ is pulled at time $t$, it is *because* arm $k$ performed well at times $s < t$.

**Equivalence between high probability and expected pseudo regret bounds.**

**Lemma A.1**. *Let $T \in \mathbb{N}$. Assume that $\mathbb{E}_{\pi,\boldsymbol{\nu}_\pi^{\otimes T}}[\mathcal{R}_T^{\pi,\rho,\boldsymbol{\nu}}] \leqslant f(T) \in \mathbb{R}_+^\star$. Then for any $\delta \in (0,1)$, we have*

$$\mathbb{P}_{\pi,\boldsymbol{\nu}_\pi^{\otimes T}}\left(\mathcal{R}_T^{\pi,\rho,\boldsymbol{\nu}} \leqslant \frac{f(T)}{\delta}\right) \geqslant 1-\delta\,. \tag{A.1}$$

*Conversely, if for all $\delta \in (0,1)$, $\mathbb{P}_{\pi,\boldsymbol{\nu}_\pi^{\otimes T}}\left(\mathcal{R}_T^{\pi,\rho,\boldsymbol{\nu}} \leqslant g(\delta, T)\right) \geqslant 1-\delta$, the following bounds hold.*

*(i) If there exists $\bar{\Delta} \in \mathbb{R}_+^\star$ such that $\max_{1\leqslant t\leqslant T}\left(\max_{k\in[K]} \rho_t(\nu_t^k) - \min_{k\in[K]} \rho_t(\nu_t^k)\right) \leqslant \bar{\Delta}$, then*

$$\mathbb{E}_{\pi,\boldsymbol{\nu}_\pi^{\otimes T}}[\mathcal{R}_T^{\pi,\rho,\boldsymbol{\nu}}] \leqslant g\left(\frac{1}{T}, T\right) + \bar{\Delta}\,. \tag{A.2}$$

*(ii) If there exists $g_T^{-1}\colon u \in \mathbb{R}_+^\star \to \mathbb{R}_+^\star$ such that $g(g_T^{-1}(u), T) \leqslant u$ for all $u \in \mathbb{R}_+^\star$, then*

$$\mathbb{E}_{\pi,\boldsymbol{\nu}_\pi^{\otimes T}}[\mathcal{R}_T^{\pi,\rho,\boldsymbol{\nu}}] \leqslant \int_0^{+\infty} g_T^{-1}(u)du\,. \tag{A.3}$$

*(iii) If there exists $g(T) \in \mathbb{R}_+^\star$ such that $g(\delta, T) = \sqrt{2\log(1/\delta)}g(T)$ for all $\delta \in (0,1)$, then*

$$\mathbb{E}_{\pi,\boldsymbol{\nu}_\pi^{\otimes T}}[\mathcal{R}_T^{\pi,\rho,\boldsymbol{\nu}}] \leqslant \sqrt{\frac{\pi}{2}}g(T)\,. \tag{A.4}$$

*Proof of Lemma A.1.* First, note that $\mathcal{R}_T^{\pi,\rho,\boldsymbol{\nu}}$ is a nonnegative random variable. The first result follows immediately from Markov's inequality, i.e. for any $u \in \mathbb{R}_+^\star$, we have

$$\mathbb{P}_{\pi,\boldsymbol{\nu}_\pi^{\otimes T}}\left(\mathcal{R}_T^{\pi,\rho,\boldsymbol{\nu}} \geqslant u\right) \leqslant \frac{\mathbb{E}_{\pi,\boldsymbol{\nu}_\pi^{\otimes T}}[\mathcal{R}_T^{\pi,\rho,\boldsymbol{\nu}}]}{u} \leqslant \frac{f(T)}{u}\,, \tag{A.5}$$

and we can set $u = f(T)/\delta$ for any $\delta \in (0,1)$. Conversely, for (i), we have

$$\begin{aligned} \mathcal{R}_T^{\pi,\rho,\boldsymbol{\nu}} &= \mathcal{R}_T^{\pi,\rho,\boldsymbol{\nu}} \mathbb{1}_{\mathcal{R}_T^{\pi,\rho,\boldsymbol{\nu}} \leqslant g(\delta,T)} + \mathcal{R}_T^{\pi,\rho,\boldsymbol{\nu}} \mathbb{1}_{\mathcal{R}_T^{\pi,\rho,\boldsymbol{\nu}} > g(\delta,T)} \\ &\leqslant g(\delta) + \mathcal{R}_T^{\pi,\rho,\boldsymbol{\nu}} \mathbb{1}_{\mathcal{R}_T^{\pi,\rho,\boldsymbol{\nu}} > g(\delta,T)} \\ &\leqslant g(\delta) + \bar{\Delta} T \mathbb{1}_{\mathcal{R}_T^{\pi,\rho,\boldsymbol{\nu}} > g(\delta,T)}\,. \end{aligned} \tag{A.6}$$

Taking the expectation on both sides and setting $\delta = 1/T$ yields the result. For (ii), we use the classical identity $\mathbb{E}[Y] = \int_0^{+\infty} \mathbb{P}(Y \geqslant u)du$ for a nonnegative random variable $Y$ to obtain

$$\mathbb{E}_{\pi,\boldsymbol{\nu}_\pi^{\otimes T}}[\mathcal{R}_T^{\pi,\rho,\boldsymbol{\nu}}] = \int_0^{+\infty} \mathbb{P}_{\pi,\boldsymbol{\nu}_\pi^{\otimes T}}\left(\mathcal{R}_T^{\pi,\rho,\boldsymbol{\nu}} \geqslant u\right) du \leqslant \int_0^{+\infty} \mathbb{P}_{\pi,\boldsymbol{\nu}_\pi^{\otimes T}}\left(\mathcal{R}_T^{\pi,\rho,\boldsymbol{\nu}} \geqslant g(g_T^{-1}(u), T)\right) du$$

$$\leqslant \int_0^{+\infty} g_T^{-1}(u)du\,. \tag{A.7}$$

Finally for (iii), we use the above inequality with $g_T^{-1}\colon u \in \mathbb{R}_+^\star \mapsto e^{-\frac{u^2}{2g(T)}}$ such that

$$\mathbb{E}_{\pi,\boldsymbol{\nu}_\pi^{\otimes T}}[\mathcal{R}_T^{\pi,\rho,\boldsymbol{\nu}}] \leqslant \int_0^{+\infty} e^{-\frac{u^2}{2g(T)}} du = \sqrt{\frac{\pi}{2}}g(T)\,, \tag{A.8}$$

where we used the expression of the c.d.f. of the Gaussian distribution $\mathcal{N}(0, g(T))$. ■

The $\mathcal{O}(\sqrt{\log 1/\delta})$ dependency of the high probability regret bound is standard and is a typical byproduct of concentration inequalities at level $\delta$ (see Section 1.4).

**Pseudo regret decomposition.**

*Proof of Proposition 1.5.* The first formula follows from a simple inversion of sums:

$$\mathcal{R}_T^{\pi,\rho,\boldsymbol{\nu}} = \sum_{t=1}^T \rho^\star(\boldsymbol{\nu}) - \rho(\nu_{\pi_t}) = \sum_{t=1}^T \sum_{k\in[K]} (\rho^\star(\boldsymbol{\nu}) - \rho(\nu_k))\, \mathbb{1}_{\pi_t=k} = \sum_{k\in[K]} \Delta_k^{\rho,\boldsymbol{\nu}} N_T^k\,, \tag{A.9}$$

and noting that $\Delta_{k^\star}^{\rho,\boldsymbol{\nu}} = 0$ by definition of the optimal arm $k^\star$. The second one is immediate using the linearity of expectation and the fact that $\Delta_k^{\rho,\boldsymbol{\nu}}$ is a constant for each arm $k \in [K]$. ■

**Instance-dependent lower bound for multiarmed bandits. (💡).**

*Proof of Theorem 1.6.* Let $\boldsymbol{\nu}' \in \mathcal{F}$ be another measure such that $\boldsymbol{\nu}$ is absolutely continuous with respect to $\boldsymbol{\nu}'$, and let $\mathbb{Y}_T^{\pi,\boldsymbol{\nu}} = (Y_t^{\pi_t,\boldsymbol{\nu}})_{t=1}^T$ and $\mathbb{Y}_T^{\pi,\boldsymbol{\nu}'} = (Y_t^{\pi_t,\boldsymbol{\nu}'})_{t=1}^T$ be sequences of rewards drawn from $T$ rounds of the bandit model with measure $\boldsymbol{\nu}$ and $\boldsymbol{\nu}'$ respectively. We assume that $\pi$ is a deterministic policy (the general case follows from simple conditional arguments on the external randomisation $(\mathcal{G}_t^\perp)_{t\in\mathbb{N}}$). The fundamental change of measure inequality of Maillard (2019b, Lemma 3.3) shows that

$$\mathbb{E}_{\boldsymbol{\nu}_\pi^{\otimes T}}\left[\log \frac{d\boldsymbol{\nu}_\pi^{\otimes T}}{d\boldsymbol{\nu}_\pi'^{\otimes T}}(\mathbb{Y}_T^{\pi,\boldsymbol{\nu},\pi})\right] = \mathrm{KL}(\boldsymbol{\nu}_\pi^{\otimes T} \,\|\, \boldsymbol{\nu}_\pi'^{\otimes T}) \geqslant \sup_{g\colon \mathbb{R}^T\to[0,1]} \mathrm{kl}\left(\mathbb{E}_{\mathbb{Y}'\sim\boldsymbol{\nu}_\pi'^{\otimes T}}[g(\mathbb{Y}')], \mathbb{E}_{\mathbb{Y}\sim\boldsymbol{\nu}_\pi^{\otimes T}}[g(\mathbb{Y})]\right)\,, \tag{A.10}$$

where we recall the definition of the Bernoulli Kullback-Leibler divergence:

$$\begin{aligned} \mathrm{kl}\colon [0,1]\times[0,1] &\longrightarrow \mathbb{R} \\ (p,q) &\longmapsto p\log\frac{p}{q} + (1-p)\log\frac{1-p}{1-q}\,. \end{aligned} \tag{A.11}$$

Moreover, the following likelihood factorisation result holds:

$$
\begin{aligned}
\mathbb{E}_{\boldsymbol{\nu}_\pi^{\otimes T}}\left[\log \frac{d\boldsymbol{\nu}_\pi^{\otimes T}}{d\boldsymbol{\nu}_\pi'^{\otimes T}}\left(\mathbb{Y}_T^{\pi,\boldsymbol{\nu},\pi}\right)\right] &= \mathbb{E}_{\boldsymbol{\nu}_\pi^{\otimes T}}\left[\sum_{t=1}^T \log \frac{d\nu_{\pi_t}}{d\nu'_{\pi_t}}(Y_t^{\pi_t,\boldsymbol{\nu}})\right] && \text{(independent arms)} \\
&= \mathbb{E}_{\boldsymbol{\nu}_\pi^{\otimes T}}\left[\sum_{k\in[K]}\sum_{t=1}^T \log \frac{d\nu_k}{d\nu'_k}(Y_t^{k,\boldsymbol{\nu}})\mathbb{1}_{\pi_t=k}\right] \\
&= \sum_{k\in[K]}\sum_{t=1}^T \mathbb{E}_{\boldsymbol{\nu}_\pi^{\otimes T}}\left[\mathbb{E}_{\boldsymbol{\nu}_\pi^{\otimes T}}\left[\log \frac{d\nu_k}{d\nu'_k}(Y_t^{k,\boldsymbol{\nu}})\mathbb{1}_{\pi_t=k} \,\middle|\, \pi_t\right]\right] \\
&= \sum_{k\in[K]}\sum_{t=1}^T \mathbb{E}_{\boldsymbol{\nu}_\pi^{\otimes T}}\left[\mathbb{1}_{\pi_t=k}\mathbb{E}_{\boldsymbol{\nu}_\pi^{\otimes T}}\left[\log \frac{d\nu_k}{d\nu'_k}(Y_t^{k,\boldsymbol{\nu}})\right]\right] \\
&= \sum_{k\in[K]}\sum_{t=1}^T \mathbb{E}_{\boldsymbol{\nu}_\pi^{\otimes T}}\left[\mathbb{1}_{\pi_t=k}\mathrm{KL}(\nu_k \parallel \nu'_k)\right] \\
&= \sum_{k\in[K]} \mathbb{E}_{\boldsymbol{\nu}_\pi^{\otimes T}}\left[N_T^{k,\pi}\right]\mathrm{KL}(\nu_k \parallel \nu'_k)\,. && \text{(A.12)}
\end{aligned}
$$

We now fix a suboptimal arm $k \in [K] \setminus \{k^\star\}$. Combining the two results above and singling out the number of pulls to arm $k$, which yields:

$$
\mathbb{E}_{\boldsymbol{\nu}_\pi^{\otimes T}}\left[N_T^{k,\pi}\right] \geqslant \sup_{g\colon \mathbb{R}^T\to[0,1]} \frac{\mathrm{kl}\left(\mathbb{E}_{\mathbb{Y}'\sim\boldsymbol{\nu}_\pi'^{\otimes T}}[g(\mathbb{Y}')], \mathbb{E}_{\mathbb{Y}\sim\boldsymbol{\nu}_\pi^{\otimes T}}[g(\mathbb{Y})]\right) - \sum_{j\in[K]\setminus\{k\}}\mathbb{E}_{\boldsymbol{\nu}_\pi^{\otimes T}}\left[N_T^{j,\pi}\right]\mathrm{KL}(\nu_j \parallel \nu'_j)}{\mathrm{KL}(\nu_k \parallel \nu'_k)}\,. \tag{A.13}
$$

Let $\varepsilon \in [0,1)$ and define the event $\Omega_\varepsilon = \{N_T^{k,\pi} > T^{1-\varepsilon}\}$ and $g = \mathbb{1}_{\Omega_\varepsilon}$. By Markov's inequality, we have

$$
\mathbb{E}_{\mathbb{Y}\sim\boldsymbol{\nu}_\pi^{\otimes T}}[g(\mathbb{Y})] = \mathbb{P}_{\boldsymbol{\nu}_\pi^{\otimes T}}\left(N_T^{k,\pi} > T^{1-\varepsilon}\right) \leqslant T^{-(1-\varepsilon)}\mathbb{E}_{\boldsymbol{\nu}_\pi^{\otimes T}}\left[N_T^{k,\pi}\right] = o(1)\,, \tag{A.14}
$$

since $\pi$ is strongly consistent over $\mathcal{F}$. Now, because $\rho$ is confusable over $\mathcal{F}$, $\boldsymbol{\nu}'$ can be chosen so that $\nu'_j = \nu_j$ for $j \in [K] \setminus \{k\}$ and $\rho(\nu'_k) > \rho^\star(\boldsymbol{\nu})$. Using again Markov's inequality and the trivial identity $T = N_T^{k,\pi} + \sum_{j\in[K]\setminus\{k\}} N_T^{j,\pi}$, we have

$$
\begin{aligned}
\log\left(1 - \mathbb{P}_{\boldsymbol{\nu}_\pi'^{\otimes T}}(\Omega_\varepsilon)\right) &= \log \mathbb{P}_{\boldsymbol{\nu}_\pi'^{\otimes T}}\left(\sum_{j\in[K]\setminus\{k\}} N_T^{j,\pi} \geqslant T - T^{1-\varepsilon}\right) \\
&\leqslant -\log\left(T - T^{1-\varepsilon}\right) + \log\left(\sum_{j\in[K]\setminus\{k\}} \mathbb{E}_{\boldsymbol{\nu}_\pi'^{\otimes T}}\left[N_T^{j,\pi}\right]\right) \\
&= -\log T + o(\log T)\,, && \text{(A.15)}
\end{aligned}
$$

since the uniform efficiency also applies to $\boldsymbol{\nu}' \in \mathcal{F}$. Therefore, the kl terms simplifies asymptotically when $T \to +\infty$:

$$\begin{aligned}
&\mathrm{kl}\left(\mathbb{E}_{\mathbb{Y}' \sim \boldsymbol{\nu}_\pi'^{\otimes T}}[g(\mathbb{Y}')], \mathbb{E}_{\mathbb{Y} \sim \boldsymbol{\nu}_\pi^{\otimes T}}[g(\mathbb{Y})]\right) \\
&\quad = \mathbb{P}_{\boldsymbol{\nu}_\pi^{\otimes T}}(\Omega_\varepsilon) \log \frac{\mathbb{P}_{\boldsymbol{\nu}_\pi^{\otimes T}}(\Omega_\varepsilon)}{\mathbb{P}_{\boldsymbol{\nu}_\pi'^{\otimes T}}(\Omega_\varepsilon)} + \left(1 - \mathbb{P}_{\boldsymbol{\nu}_\pi^{\otimes T}}(\Omega_\varepsilon)\right) \log \left(\frac{1 - \mathbb{P}_{\boldsymbol{\nu}_\pi^{\otimes T}}(\Omega_\varepsilon)}{1 - \mathbb{P}_{\boldsymbol{\nu}_\pi'^{\otimes T}}(\Omega_\varepsilon)}\right) \\
&= \log T + o(\log T) \ . \qquad \text{(A.16)}
\end{aligned}$$

Finally, since $\nu_j = \nu'_j$ for $j \in [K] \setminus \{k\}$, the corresponding KL terms vanish, which gives

$$\liminf_{T \to +\infty} \frac{\mathbb{E}_{\boldsymbol{\nu}_\pi^{\otimes T}}\left[N_T^{k,\pi}\right]}{\log T} \geqslant \frac{1}{\mathrm{KL}(\nu_k \parallel \nu'_k)} \, . \qquad \text{(A.17)}$$

This inequality holds for any measure $\nu'_k$ that satisfies the confusability property of $\rho$, therefore we can take the sup of such measures, which gives the lower bound on the expected number of pulls. The lower bound on the expected pseudo regret follows immediately from Proposition 1.5.

■

**Minimax lower bound for contextual bandits. (💡).**

*Proof of Theorem 1.10.* We follow the steps of Lattimore and Szepesvári (2020, Theorems 24.1 and 24.2), but extend the proofs beyond Gaussian bandit models. In particular, we construct an explicit instance of linear bandit parametrised by some $\theta \in \mathbb{R}^d$ that suffers $\Omega(d\sqrt{T})$ expected pseudo regret. Let $\Delta \in \mathbb{R}_+^\star$ and fix $\theta \in \{\pm\Delta\}^d$. For $t \in \mathbb{N}$, under the contextual bandit model $\nu^\theta$, by definition of the optimal action, we recall that $X_t^\star = \operatorname{argmax}_{x \in \mathcal{X}} \langle \theta, x \rangle$. For a fixed $i \in \{1, \dots, d\}$, we consider $\bar{\theta}^i \in \{\pm\Delta\}^d$ built from $\theta$ by flipping its $i$-th coordinate, i.e. $\theta_j = \bar{\theta}^i_j$ for $j \neq i$ and $\bar{\theta}^i_i = -\theta_i$. Note that this defines an equivalence relation on $\{\pm\Delta\}^d$ where $\theta \sim \theta'$ if and only if $\theta' = \bar{\theta}^i$, which induces a partition $\{D_n^i\}_{n=1}^{2^{d-1}}$ of $\{\pm\Delta\}^d$ where $D_n^i = \{\theta_n, \bar{\theta}_n^i\}$ for all $n \in \{1, \dots, 2^{d-1}\}$.

**Case (i):** $\mathcal{X} = \mathbb{B}^d_{\|\cdot\|_\infty}(0,1)$. We have $X_t^\star = (X_{t,i}^\star)_{i=1}^d$ and $X_{t,i}^\star = \operatorname{sign}(\theta_i)$ for all $i \in \{1, \dots, d\}$, and thus the pseudo regret takes the form

$$\begin{aligned}
\mathbb{E}_{\mathbb{X},\theta}[\mathcal{R}_T^{\mathbb{X},\rho,\nu_\theta}] = \mathbb{E}_{\mathbb{X},\theta}\left[\sum_{t=1}^T \rho(\nu_\theta(X_t^\star)) - \rho(\nu_\theta(X_t))\right] &= \mathbb{E}_{\mathbb{X},\theta}\left[\sum_{t=1}^T \langle \theta, X_t^\star - X_t \rangle\right] \\
&= \mathbb{E}_{\mathbb{X},\theta}\left[\sum_{t=1}^T \sum_{i=1}^d (\operatorname{sign}(\theta_i) - X_{t,i})\theta_i\right] \, . \qquad \text{(A.18)}
\end{aligned}$$

Therefore, regret is accumulated along the coordinate $i$ whenever $X_{t,i}$ is not aligned with $\theta_i$. Moreover, each summand can be lower bounded as $(\text{sign}(\theta_i) - X_{t,i})\theta_i \geqslant \Delta \mathbb{1}_{\text{sign}(\theta_i)\neq \text{sign}(X_{t,i})}$. Hence, it is natural to consider the event

$$\mathcal{E}_{\theta,i} = \left( \sum_{t=1}^{T} \mathbb{1}_{\text{sign}(\theta_i)\neq \text{sign}(X_{t,i})} \geqslant \frac{T}{2} \right), \tag{A.19}$$

and similarly for $\mathcal{E}_{\bar{\theta}^i,i}$. Simple algebra coupled with the definition of the total variation distance and Pinsker's inequality (recalled in Section 1.1) yields the following inequalities:

$$\begin{aligned} \mathbb{P}_{\mathbb{X},\bar{\theta}^i}(\mathcal{E}_{\theta,i}) - \mathbb{P}_{\mathbb{X},\theta}(\mathcal{E}_{\theta,i}) &\leqslant \delta_{\text{TV}}(\mathbb{P}_{\mathbb{X},\theta}, \mathbb{P}_{\mathbb{X},\bar{\theta}^i}) \\ &\leqslant \sqrt{\frac{1}{2}\text{KL}(\mathbb{P}_{\mathbb{X},\theta} \parallel \mathbb{P}_{\mathbb{X},\bar{\theta}^i})} \\ &= \mathbb{P}_{\mathbb{X},\bar{\theta}^i}(\mathcal{E}_{\theta,i}) + \mathbb{P}_{\mathbb{X},\bar{\theta}^i}(\bar{\mathcal{E}}_{\theta,i}) - 1 + \sqrt{\frac{1}{2}\text{KL}(\mathbb{P}_{\mathbb{X},\theta} \parallel \mathbb{P}_{\mathbb{X},\bar{\theta}^i})}, \end{aligned} \tag{A.20}$$

where $\bar{\mathcal{E}}_{\theta,i}$ is the complement event of $\mathcal{E}_{\theta,i}$. Since $\theta_i$ and $\bar{\theta}^i_i$ are of opposite signs by construction, we have $\bar{\mathcal{E}}_{\theta,i} = \mathcal{E}_{\bar{\theta}^i,i}$. Therefore, we deduce that

$$\begin{aligned} \mathbb{P}_{\mathbb{X},\theta}(\mathcal{E}_{\theta,i}) + \mathbb{P}_{\mathbb{X},\bar{\theta}^i}(\mathcal{E}_{\bar{\theta}^i,i}) &\geqslant 1 - \sqrt{\frac{1}{2}\text{KL}(\mathbb{P}_{\mathbb{X},\theta} \parallel \mathbb{P}_{\mathbb{X},\bar{\theta}^i})} \\ &= 1 - \sqrt{\frac{1}{2}\sum_{t=1}^{T} \mathbb{E}_{\mathbb{X},\theta}\left[\log \frac{\mathrm{d}\nu_\theta(X_t)}{\mathrm{d}\nu_{\bar{\theta}^i}(X_t)}(Y_t)\right]} \\ &= 1 - \sqrt{\frac{1}{2}\sum_{t=1}^{T} \mathbb{E}_{\mathbb{X},\theta}\left[\mathbb{E}_{\mathbb{X},\theta}\left[\log \frac{\mathrm{d}\nu_\theta(X_t)}{\mathrm{d}\nu_{\bar{\theta}^i}(X_t)}(Y_t) \,\middle|\, X_t\right]\right]} \\ &\geqslant 1 - \sqrt{\frac{1}{2}\sum_{t=1}^{T} \mathbb{E}_{\mathbb{X},\theta}\left[g\left(\frac{1}{\sigma}\langle \theta - \bar{\theta}^i, X_t\rangle\right)\right]} \\ &= 1 - \sqrt{\frac{1}{2}\sum_{t=1}^{T} \mathbb{E}_{\mathbb{X},\theta}\left[g\left(\frac{2\Delta X_{t,i}}{\sigma}\right)\right]} \\ &= \Omega\left(\frac{T\Delta^2}{\sigma^2}\right). \end{aligned} \tag{A.21}$$

which holds if $\Delta = \mathcal{O}(\sigma T^{-1/2})$. Note that this inequality involves two parameters $\theta$ and $\bar{\theta}^i$ simultaneously. To disentangle this, we sum across all possible $2^d$ parameters in $\{\pm\Delta\}^d$ by pairing each $\theta$ with its corresponding $\bar{\theta}^i$. Formally, using the partition $\{D_n\}_{n=1}^{2^{d-1}}$, we have

$$\frac{1}{2^d}\sum_{i=1}^{d}\sum_{\theta' \in \{\pm\Delta\}^d} \mathbb{P}_{\mathbb{X},\theta}(\mathcal{E}_{\theta,i}) = \frac{1}{2^d}\sum_{i=1}^{d}\sum_{n=1}^{2^{d-1}} \mathbb{P}_{\mathbb{X},\theta_n}(\mathcal{E}_{\theta_n,i}) + \mathbb{P}_{\mathbb{X},\bar{\theta}^i_n}(\mathcal{E}_{\bar{\theta}^i_n,i}) \geqslant \frac{1}{2^d}\sum_{i=1}^{d} 2^{d-1}\Omega(T\Delta^2)$$

$$= \Omega\left(\frac{dT\Delta^2}{\sigma^2}\right) . \tag{A.22}$$

In others words, the average across all $\theta$ of $\sum_{i=1}^d \mathbb{P}_{\mathbb{X},\theta}(\mathcal{E}_{\theta,i})$ is lower bounded by $\Omega(dT\Delta^2/\sigma^2)$ and therefore at least one $\theta$ must satisfy $\sum_{i=1}^d \mathbb{P}_{\mathbb{X},\theta}(\mathcal{E}_{\theta,i}) \geqslant \Omega(dT\Delta^2/\sigma^2)$. For this particular instance of $\theta$, this corresponding bandit model suffers an expected pseudo regret of

$$\mathbb{E}_{\mathbb{X},\theta}[\mathcal{R}_T^{\mathbb{X},\rho,\nu_\theta}]d \geqslant \Delta \sum_{i=1}^d \mathbb{E}_{\mathbb{X},\theta}\left[\sum_{t=1}^T \mathbb{1}_{\mathrm{sign}(\theta_i)\neq\mathrm{sign}(X_{t,i})}\right] \geqslant \frac{T\Delta}{2}\sum_{i=1}^d \mathbb{P}_{\mathbb{X},\theta}\left(\mathcal{E}_{\theta,i}\right)$$

(Markov's inequality)

$$\geqslant \Omega\left(\frac{dT^2\Delta^3}{\sigma^2}\right) . \tag{A.23}$$

We conclude by tuning the hyperparameter $\Delta \in \mathbb{R}_+^\star$, with the constraint that $\Delta = \mathcal{O}(\sigma T^{-1/2})$. In particular, $\Delta = \sigma T^{-1/2}$ provides the $\Omega(\sigma d\sqrt{T})$ lower bound.

**Case (ii):** $\mathcal{X} = \mathbb{B}^d_{\|\cdot\|_2}(0,1)$**.** It follows from the properties of the Euclidean scalar product that $X_t^\star = \theta^\star/\|\theta^\star\|$. The specific form of $\theta$ also ensures that $\|\theta\|_2 = \Delta\sqrt{d}$. We deduce that

$$\begin{aligned}\mathbb{E}_{\mathbb{X},\theta}[\mathcal{R}_T^{\mathbb{X},\rho,\nu_\theta}] = \mathbb{E}_{\mathbb{X},\theta}\left[\sum_{t=1}^T \langle\theta, X_t^\star - X_t\rangle\right] &= \mathbb{E}_{\mathbb{X},\theta}\left[\sum_{t=1}^T \|\theta\|_2 - \sum_{i=1}^d \theta_i X_{t,i}\right] \\ &= \Delta\mathbb{E}_{\mathbb{X},\theta}\left[\sum_{t=1}^T\sum_{i=1}^d \frac{1}{\sqrt{d}} - \mathrm{sign}(\theta_i)X_{t,i}\right] \\ &\geqslant \frac{\Delta\sqrt{d}}{2}\mathbb{E}_{\mathbb{X},\theta}\left[\sum_{t=1}^T\sum_{i=1}^d\left(\frac{1}{\sqrt{d}} - \mathrm{sign}(\theta_i)X_{t,i}\right)^2\right] ,\end{aligned} \tag{A.24}$$

where the last line follows from simple algebra using the fact that $\|X_t\|_2^2 \leqslant 1$ for all $t \in \mathbb{N}$. Since the right-hand side is a sum of $T$ nonnegative terms, it is lower bounded by the sum of the first $\tau$ terms for some stopping $\tau$ such that $\tau \leqslant T$ almost surely, i.e.

$$\mathbb{E}_{\mathbb{X},\theta}[\mathcal{R}_T^{\mathbb{X},\rho,\nu^\theta}] \geqslant \frac{\Delta\sqrt{d}}{2}\sum_{i=1}^d \mathbb{E}_{\mathbb{X},\theta}\left[\sum_{t=1}^\tau\left(\frac{1}{\sqrt{d}} - \mathrm{sign}(\theta_i)X_{t,i}\right)^2\right] . \tag{A.25}$$

In particular, we may define the stopping time $\tau_i = T \wedge \inf\{t \in \mathbb{N},\ \sum_{s=1}^t X_{s,i}^2 \geqslant T/d\}$ and let $U_i^\pm = \sum_{t=1}^{\tau_i}(1/\sqrt{d} - X_{t,i})^2$. The term $U_i^+$ can be upper bounded using the inequality

$(a-b)^2 \leqslant 2(a^2+b^2)$ for all $a, b \in \mathbb{R}$:

$$0 \leqslant U_i^+ \leqslant 2\sum_{t=1}^{\tau_i} \frac{1}{d} + X_{t,i}^2 \leqslant \frac{2\tau_i}{d} + 2\sum_{t=1}^{\tau_i} X_{t,i}^2 \leqslant \frac{2T}{d} + 2\left( \underbrace{\sum_{t=1}^{\tau_i - 1} X_{t,i}^2}_{\leqslant T/d} + \underbrace{X_{\tau_i,i}^2}_{\leqslant |X_{\tau_i}|_2^2 \leqslant 1} \right) \leqslant \frac{4T}{d} + 2\,. \tag{A.26}$$

Hence, by the properties of the total variation distance and Pinsker's inequality, we have that

$$|\mathbb{E}_{\mathbb{X},\theta}[U_i^+] - \mathbb{E}_{\mathbb{X},\bar{\theta}^i}[U_i^+]| \leqslant (4T/d+2)\sqrt{\frac{1}{2}\mathrm{KL}(\mathbb{P}_{\mathbb{X},\theta} \parallel \mathbb{P}_{\mathbb{X},\bar{\theta}^i})}\,. \tag{A.27}$$

We now use the generic divergence decomposition theorem with stopping times exposed in Lattimore and Szepesvári (2020, Exercises 15.7 and 15.8) to obtain the following bounds:

$$\begin{aligned}
\mathrm{KL}(\mathbb{P}_{\mathbb{X},\theta} \parallel \mathbb{P}_{\mathbb{X},\bar{\theta}^i}) \leqslant \mathbb{E}_{\mathbb{X},\theta}\left[\sum_{t=1}^{\tau_i} \log \frac{\mathrm{d}\nu_\theta(X_t)}{\mathrm{d}\nu_{\bar{\theta}^i}(X_t}(Y_t)\right] &= \mathbb{E}_{\mathbb{X},\theta}\left[\sum_{t=1}^{\tau_i} g\left(\frac{2\Delta X_{t,i}}{\sigma}\right)\right] \\
&= \mathcal{O}\left(\frac{\Delta^2}{\sigma^2}\mathbb{E}_{\mathbb{X},\theta}\left[\sum_{t=1}^{\tau_i} X_{t,i}^2\right]\right)\,,
\end{aligned} \tag{A.28}$$

and thus, by construction of the stopping time $\tau_i$,

$$\mathrm{KL}(\mathbb{P}_{\mathbb{X},\theta} \parallel \mathbb{P}_{\mathbb{X},\bar{\theta}^i}) \leqslant \mathcal{O}\left(\frac{T\Delta^2}{\sigma^2 d}\right)\,. \tag{A.29}$$

Putting these together, we obtain the following inequality:

$$\mathbb{E}_{\mathbb{X},\theta}[U_i^+] \geqslant \mathbb{E}_{\mathbb{X},\bar{\theta}^i}[U_i^+] - \Omega\left(\frac{\Delta}{\sigma}\left(\frac{T}{d}\right)^{\frac{3}{2}}\right)\,. \tag{A.30}$$

Adding $\mathbb{E}_{\mathbb{X},\bar{\theta}^i}[U_i^-]$ on both sides yields

$$\begin{aligned}
\mathbb{E}_{\mathbb{X},\theta}[U_i^+] + \mathbb{E}_{\mathbb{X},\bar{\theta}^i}[U_i^-] &\geqslant \mathbb{E}_{\mathbb{X},\bar{\theta}^i}[U_i^+ + U_i^-] - \Omega\left(\frac{\Delta}{\sigma}\left(\frac{T}{d}\right)^{\frac{3}{2}}\right) \\
&= 2\mathbb{E}_{\mathbb{X},\bar{\theta}^i}\left[\frac{\tau_i}{d} + \sum_{t=1}^{\tau_i} X_{t,i}^2\right] - \Omega\left(\frac{\Delta}{\sigma}\left(\frac{T}{d}\right)^{\frac{3}{2}}\right) \\
&\geqslant \frac{2T}{d} - \Omega\left(\frac{\Delta}{\sigma}\left(\frac{T}{d}\right)^{\frac{3}{2}}\right)\,.
\end{aligned} \tag{A.31}$$

Finally, we use the same partitioning argument as in case (i) to obtain the following bound on the expected pseudo regret averaged over all possible $\theta \in \{\pm\Delta\}^d$:

$$\begin{aligned}
\frac{1}{2^d} \sum_{\theta \in \{\pm\Delta\}^d} \mathbb{E}_{\mathbb{X},\theta}[\mathcal{R}_T^{\mathbb{X},\rho,\nu_\theta}] &\geqslant \frac{\Delta\sqrt{d}}{2^{d+1}} \sum_{i=1}^{d} \sum_{\theta \in \{\pm\Delta\}^d} \mathbb{E}_{\mathbb{X},\theta}[U_i^{\mathrm{sign}(\theta_i)}] \\
&= \frac{\Delta\sqrt{d}}{2^{d+1}} \sum_{i=1}^{d} \sum_{n=1}^{2^{d-1}} \mathbb{E}_{\mathbb{X},\theta_n}[U_i^{\mathrm{sign}(\theta_{n,i})}] + \mathbb{E}_{\mathbb{X},\bar{\theta}_n^i}[U_i^{\mathrm{sign}(\bar{\theta}_{n,i}^i)}] \\
&\geqslant \frac{\Delta\sqrt{d}}{4}\left(2T - \Omega\left(\frac{\Delta T^{\frac{3}{2}}}{\sigma\sqrt{d}}\right)\right)
\end{aligned} \tag{A.32}$$

Setting $\Delta = \sigma\sqrt{d/T}$ turns the right-hand side into $\Omega(\sigma d\sqrt{T})$ for $T$ large enough. Finally, since the average over $\{\pm\Delta\}^d$ of the expected pseudo regret is lower bounded by this quantity, at least one $\theta \in \{\pm\Delta\}^d$ must satisfy this lower bound, which concludes the proof. ■

## A.2 Technical results on statistical models.

**Control of the variance for sub-Gaussian distributions.**

*Proof of Lemma 1.12.* Let $\mu = \mathbb{E}_{Y\sim\nu}[Y]$ and $\gamma\colon \lambda \in \mathbb{R} \mapsto \mathbb{E}_{Y\sim\nu}[\exp(\lambda(Y-\mu))] - \exp(\lambda^2 R^2/2)$. The sub-Gaussian condition implies that $\gamma(\lambda) \leqslant 0$ for all $\lambda \in \mathbb{R}$, and a second order expansion around $\lambda = 0$ shows that $\gamma(\lambda) = \lambda^2/2(\mathbb{E}_{Y\sim\nu}[(Y-\mu)^2] - R^2) + o(\lambda^2)$. Therefore, the dominating term must be nonpositive, which is precisely the statement of the lemma. ■

**Obstruction to logarithmic pseudo regret.**

*Proof of Lemma 1.13.* Note that bounded distributions are sub-Gaussian, sub-Gaussian distributions are light tailed, and light tailed distributions have finite moments of any order, therefore it is sufficient to show the result for the smallest of these families, i.e. $\mathcal{F} = \cup_{\underline{B},\overline{B}\in\mathbb{R}} \mathcal{F}_{[\underline{B},\overline{B}]}$.

Let $\nu \in \mathcal{F}$. $\mu = \mathbb{E}_{Y\sim\nu}[Y]$ and $\mu^\star > \mu$. We construct an explicit sequence of distributions $(\nu'_n)_{n\in\mathbb{N}} \in \mathcal{F}^{\mathbb{N}}$ such that (i) $\mathbb{E}_{Y\sim\nu'_n} > \mu^\star$ and (ii) $\mathrm{KL}(\nu \,\|\, \nu'_n) \to 0$. Indeed, for $n \in \mathbb{N}^\star$, we define

$$\Delta = \mu^\star - \mu \quad \text{and} \quad \nu'_n = (1 - \frac{1}{n})\nu + \frac{1}{n}\delta_{\mu+2n\Delta} \,. \tag{A.33}$$

Assuming $n$ is large enough, $\mu + 2n\Delta$ lies outside the upper bounded set $\mathrm{Supp}\,\nu$, in which case the density of $\nu$ with respect to $\nu'_n$ becomes $1/(1-1/n)\mathbb{1}_{\mathrm{Supp}\,\nu}$, and thus $\mathrm{KL}(\nu \,\|\, \nu'_n) = \log\frac{1}{1-1/n}$. Moreover, the expectation of $\nu'_n$ is $\mu + 2\Delta > \mu^\star$, and therefore $0 \leqslant \mathcal{K}_{\inf}(\nu;\mu^\star) \leqslant \log\frac{1}{1-1/n}$. Since this upper bound holds for any $n \in \mathbb{N}^\star$, we deduce that $\mathcal{K}_{\inf}(\nu;\mu^\star) = 0$. ■

## A.3 Technical results in concentration of measures

### Fixed sample concentration and the Cramér-Chernoff method

*Proof of Proposition 1.15.* We propose an alternative to the classical proof of this result, based on a supermartingale construction, in order to highlight the profound link between martingales and concentration which we will constantly use in the sequel. Indeed, $M_s^\lambda = \exp(\lambda S_s - s\psi(\lambda))$ for $s \in \{1, \dots, t\}$ and $\lambda \in \mathcal{I}$ defines a $(\mathcal{G}_s)_{s=1}^t$-adapted, integrable process which satisfies

$$\begin{aligned}\mathbb{E}\left[e^{\lambda S_{s+1}-(s+1)\psi(\lambda)} \,\middle|\, \mathcal{G}_s\right] &= \mathbb{E}\left[\underbrace{M_s^\lambda}_{\mathcal{G}_s\text{-measurable}} \exp\left(\lambda(S_{s+1}-S_s)-\psi(\lambda)\right) \,\middle|\, \mathcal{G}_s\right] \\ &= M_s^\lambda \mathbb{E}\left[\exp\left(\lambda(S_{s+1}-S_s)-\psi_R(\lambda)\right) \mid \mathcal{G}_s\right] \\ &\leqslant M_s^\lambda\,, \end{aligned} \tag{A.34}$$

therefore $(M_s^\lambda)_{s=1}^t$ defines a nonnegative supermartingale. Now, by Markov's inequality we have $\mathbb{P}(M_t^\lambda \geqslant 1/\delta) \leqslant \mathbb{E}[M_t^\lambda]\delta \leqslant \delta$. Rewriting this in terms of $S_t$ yields the inequality

$$\forall \lambda \in \mathcal{I}_+,\ \mathbb{P}\left(S_t \geqslant t\frac{\psi(\lambda) + \frac{1}{t}\log\frac{1}{\delta}}{\lambda}\right) \leqslant \delta\,. \tag{A.35}$$

Optimising in $\lambda \in \mathcal{I}_+$ the right-hand side in the probability makes appear the Fenchel-Legendre transform $\psi^{\star,+}$. Indeed, following Pinelis (2013, Proposition 1.5), its generalised inverse satisfies the relation $(\psi^{\star,+})^{-1}(v) = \inf_{\lambda\in\mathcal{I}_+}(\psi(\lambda)+v)/\lambda$ for any $v \in \mathbb{R}_+$ (note that $\psi^{\star,+}$ is nondecreasing on $\mathbb{R}_+^\star$ and continuous by convexity, therefore its generalised inverse exists). Hence, we have

$$\mathbb{P}\left(S_t \geqslant t(\psi_R^{\star,+})^{-1}\left(\frac{1}{t}\log\frac{1}{\delta}\right)\right) \leqslant \delta\,. \tag{A.36}$$

The reverse bound follows from a symmetric argument using $\psi^{\star,-}$ instead. ∎

*Proof of Corollary 1.16.* We apply Proposition 1.15 with $S_t = \sum_{s=1}^t Y_s - \mu$, $\mathcal{I} = \mathbb{R}$ and $\psi\colon \lambda \in \mathbb{R} \mapsto R^2\lambda^2/2$. A straightforward calculation shows that $\psi_R^{\star,+}\colon \mathbb{R}_+ \mapsto u^2/(2R^2)$ (attained at $\lambda = u/R^2$ in the definition of the Fenchel-Legendre transform), which yields the expected result. In particular, isolating $\mu$ shows that $\mu \leqslant \widehat{\mu}_t - R\sqrt{2/t\log(1/\delta)}$ holds with probability at most $\delta$. A similar calculation for $\psi^{\star,-}$ gives $\mu \geqslant \widehat{\mu}_t + R\sqrt{2/t\log(1/\delta)}$ with probability at most $\delta$. We conclude by invoking a union bound on both events. ∎

### Time-uniform concentration and the method of mixtures

**Method of mixtures.**

*Proof of Theorem 1.20.* First, $(M_t)_{t\geqslant t_0}$ is a nonnegative supermartingale, hence the random variable $M_\infty = \lim_{t\to+\infty} M_t$ is almost surely well-defined (Doob's supermartingale convergence theorem, Williams (1991, Chapter 11)), and thus $M_\tau$ is almost surely well-defined. By Fatou's lemma, we have that

$$M_\tau = \mathbb{E}[\liminf_{t\to+\infty} M_{t\wedge\tau}] \leqslant \liminf_{t\to+\infty} \mathbb{E}[M_{t\wedge\tau}] \leqslant \mathbb{E}[M_{t_0}] \leqslant 1\,. \tag{A.37}$$

Then, Markov's inequality provides the following bound:

$$\mathbb{P}\left(M_\tau \geqslant \frac{1}{\delta}\right) \leqslant \mathbb{E}[M_\tau]\delta \leqslant \delta\,. \tag{A.38}$$

The order-preserving property of the generalised inverse implies that

$$\mathbb{P}\left(S_\tau \geqslant F_\tau^{-1}\left(\frac{1}{\delta}\right)\right) \leqslant \mathbb{P}\left(M_\tau \geqslant \frac{1}{\delta}\right) \leqslant \delta\,. \tag{A.39}$$

In particular for the $(\mathcal{G}_t)_{t\geqslant t_0}$-stopping time $\tau = \inf\{t \geqslant t_0,\ S_t \geqslant F_t^{-1}(1/\delta)\}$, we have

$$\mathbb{P}(\tau < \infty) = \mathbb{P}\left(\exists t \geqslant t_0,\ S_t \geqslant F_t^{-1}\left(\frac{1}{\delta}\right)\right) = \mathbb{P}\left(S_\tau \geqslant F_\tau^{-1}\left(\frac{1}{\delta}\right)\right) \leqslant \delta\,. \tag{A.40}$$

We obtain the reverse bound in the case where the $(F_t)_{t\geqslant t_0}$ are nonincreasing. ∎

*Proof of Corollary 1.21.* We consider the same nonnegative supermartingale construction as in the proof of Proposition 1.15, i.e. if $\log \mathbb{E}_{Y\sim\nu}[\exp(\lambda(Y-\mu))] \leqslant \psi(\lambda)$ for $\lambda = \mathcal{I} \subseteq \mathbb{R}$, we define for $t \in \mathbb{N}$, $S_t = \sum_{s=1}^t Y_s - \mu$ and $M_t^\lambda = \exp(\lambda S_t - t\psi(\lambda))$. However, contrary to Proposition 1.15, we cannot optimise $\lambda$ without breaking the supermartingale property. Indeed, for a given $t \in \mathbb{N}$, the optimal $\lambda = \lambda_t^\star$ satisfies the $(\psi^{\star,+})^{-1}(\log(1/\delta)/t) = (\psi(\lambda_t^\star) + \log(1/\delta)/t)/\lambda_t^\star$ and therefore depends on the time $t$ (the process $(M_t^\lambda)_{t\in\mathbb{N}}$ is only a supermartingale for a *fixed* $\lambda \in \mathbb{R}$).

Instead, we consider a *mixture* supermartingale by randomising over possible values of $\lambda$. Formally, let $(\mathcal{G}_t)_{t\in\mathbb{N}}$ the natural filtration of the process $(S_t)_{t\in\mathbb{N}}$, $\Lambda$ a random variable independent on $(\mathcal{G}_t)_{t\in\mathbb{N}}$ and consider the $\sigma$-algebra $\mathcal{G}_\infty = \sigma(\bigcup_{t\in\mathbb{N}} \mathcal{G}_t)$. We define the *mixture* process by $M_t = \mathbb{E}[M_t^\Lambda \mid \mathcal{G}_\infty]$, which preserves the nonnegative supermartingale property since

$$\begin{aligned}
\mathbb{E}\left[M_{t+1} \mid \mathcal{G}_t\right] = \mathbb{E}\left[\mathbb{E}\left[M_{t+1}^\Lambda \,\middle|\, \mathcal{G}_\infty\right] \,\middle|\, \mathcal{G}_t\right] &= \mathbb{E}\left[M_{t+1}^\Lambda \,\middle|\, \mathcal{G}_t\right] && \text{(tower property)}\\
&\leqslant \mathbb{E}\left[M_t^\Lambda \,\middle|\, \mathcal{G}_t\right] && \text{(supermartingale)}\\
&= \mathbb{E}\left[\mathbb{E}\left[M_t^\Lambda \,\middle|\, \mathcal{G}_\infty\right] \,\middle|\, \mathcal{G}_t\right] && \text{(tower property)}\\
&= M_t\,. && \text{(A.41)}
\end{aligned}$$

Moreover, $\mathbb{E}[M_0] = \mathbb{E}[\mathbb{E}[M_0^\Lambda \mid \Lambda]] \leqslant \mathbb{E}[1] = 1$, therefore we may apply Theorem 1.20.

It remains to compute an explicit expression for $M_t$. If we denote by $\nu_\Lambda$ the measure of the mixing random variable $\Lambda$, then $M_t = \int_{\mathcal{I}} M_t^\lambda d\nu_\Lambda(\lambda)$. In the sub-Gaussian case with $R=1$, we choose $\Lambda \sim \mathcal{N}(0, 1/\alpha)$ and thus $M_t = F_t(S_t)$ with

$$\begin{aligned} F_t \colon \mathbb{R} &\longrightarrow \mathbb{R}_+ \\ y &\longmapsto \sqrt{\frac{\alpha}{2\pi}} \int_{-\infty}^{\infty} e^{\lambda y - \frac{t\lambda^2}{2} - \frac{\alpha\lambda^2}{2}} d\lambda \,. \end{aligned} \tag{A.42}$$

To compute this integral, we resort to the standard "completion of the square" method, abundantly used in the calculation of conjugate prior in Bayesian statistics. Indeed, let $\bar\sigma^2 = 1/(t+\alpha)$, then for $\lambda \in \mathbb{R}$ and $y \in \mathbb{R}$, we have

$$\frac{t+\alpha}{2}\lambda^2 - \lambda y = \frac{1}{2\bar\sigma^2}\left(\lambda - \bar\sigma^2 y\right)^2 - \frac{\bar\sigma^2 y^2}{2}\,, \tag{A.43}$$

and consequently, by identifying the density of $\mathcal{N}(\bar\sigma^2 y, \bar\sigma^2)$ we obtain

$$F_t(y) = \sqrt{\alpha}\bar\sigma e^{\frac{\bar\sigma^2 y^2}{2}} \underbrace{\frac{1}{\sqrt{2\pi}\bar\sigma}\int_{-\infty}^{\infty} e^{-\frac{(\lambda - \bar\sigma^2 y)}{2\bar\sigma^2}} d\lambda}_{=1} = \frac{e^{\frac{y^2}{t+\alpha}}}{\sqrt{1+\frac{t}{\alpha}}}\,. \tag{A.44}$$

We obtain the bound for $R=1$ by inverting $F_t(y) = 1/\delta$ for $\delta \in (0,1)$ and $y \in \mathbb{R}_+$. i.e. for any random time $\tau$ in $\mathbb{N}$:

$$\mathbb{P}\left(S_\tau \geqslant \sqrt{2(\tau+\alpha)\log\left(\frac{1}{\delta}\sqrt{1+\frac{\tau}{\alpha}}\right)}\right)\,. \tag{A.45}$$

For an arbitrary $R \in \mathbb{R}_+^\star$, it is straightforward to show that $(Y_t/R)_{t\in\mathbb{N}}$ is an i.i.d. sequence of 1-sub-Gaussian random variables with mean $\mu/R$, and therefore the general result can be deduced by scaling. Finally, we obtain the reverse bound by considering $y \in \mathbb{R}_-$ instead, and the expression of the confidence sequence results from a simple union bound. ■

**Tuning of the mixing parameter.**

*Proof of Lemma 1.23.* We find the optimal mixing parameter $\alpha^\star$ by differentiating the mapping $\alpha \in \mathbb{R}_+^\star \mapsto (t_0+\alpha)\log(\sqrt{1+t_0/\alpha}/\delta)$, which gives $\log(\sqrt{1+t_0/\alpha^\star}/\delta) = t_0/(2\alpha^\star)$. After simple algebra, we obtain $we^w = -\delta^2/e$ with $w = 1 + t_0/\alpha^\star$. Since $-\delta^2/e \in (-1/e, 0)$, the unique solution in $w \in \mathbb{R}$ to this equation is given by the Lambert function $W_{-1}$. ■

We also report in Figure A.1 a numerical experiment to illustrate the effect of the mixing parameter $\alpha$ on the width of the corresponding sub-Gaussian confidence sequence.

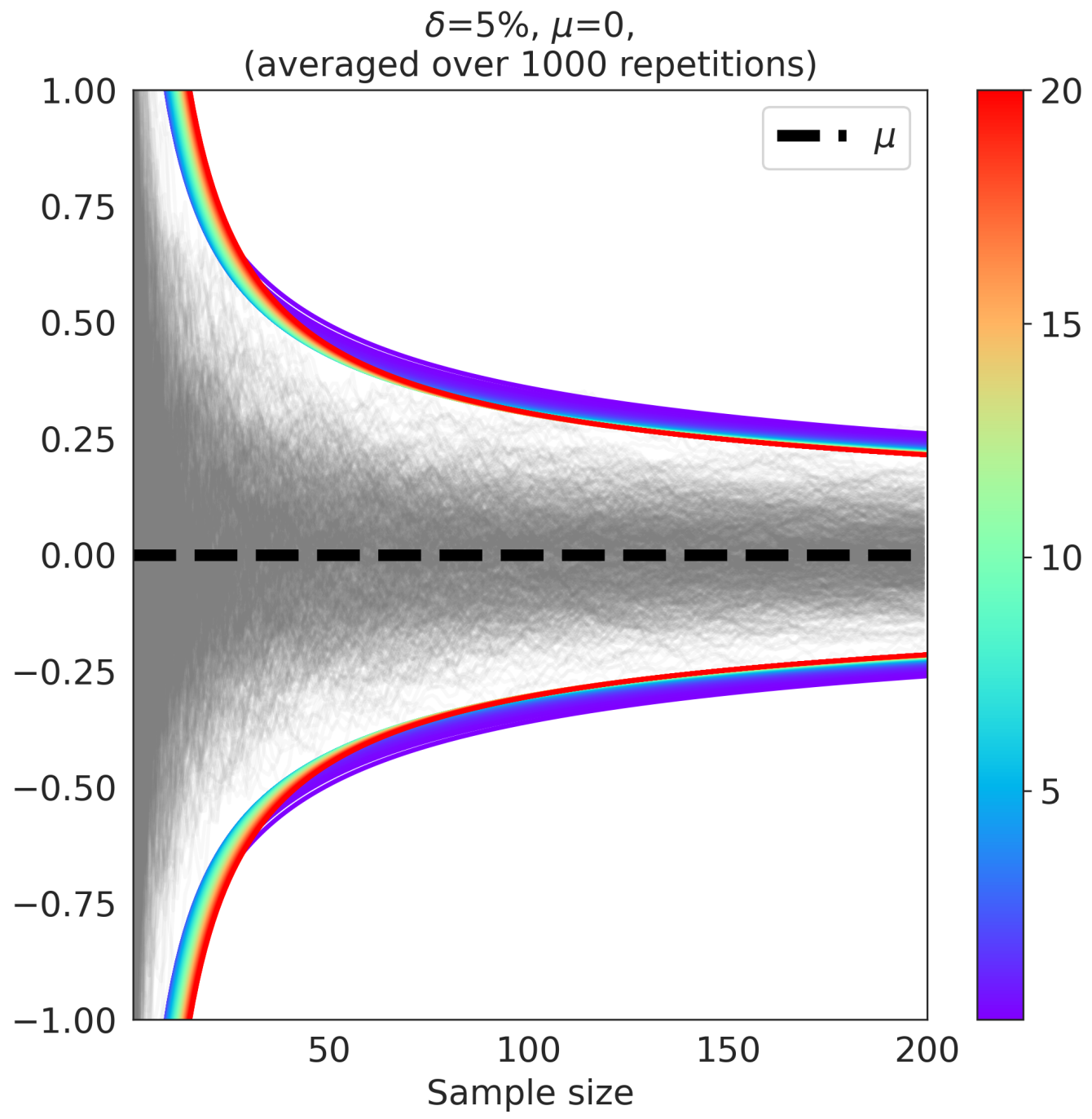

**Figure A.1 –** Comparison of the time-uniform sub-Gaussian confidence envelopes for varying mixing parameter $\alpha \in [0.1, 10]$, as a function of the sample size $t$, over 1000 independent replicates. Grey lines are trajectories of empirical means $\widehat{\mu}_t$, drawn for $\mathcal{N}(\mu, 1)$ with $\mu = 0$. Thick black dashed line: $\mu$.

**Extension to multivariate distributions.**

*Proof of Proposition 1.26.* As in the one-dimensional method of mixtures of Corollary 1.21, we first leverage the sub-Gaussian property to construct a nonnegative supermartingale. Indeed, for $\lambda \in \mathbb{R}^d$ and $t \in \mathbb{N}$, we let $M_t^\lambda = \exp(\lambda^\top S_t - R^2/2\|\lambda\|^2_{V_t^0})$, where we let $V_t^0 = \sum_{s=1}^{t-1} X_s X_s^\top$. This defines a $(\mathcal{G}_t)_{t\in\mathbb{N}}$-adapted and integrable process and

$$
\begin{aligned}
\mathbb{E}\left[M_{t+1}^\lambda \,\middle|\, \mathcal{G}_t\right] &= \mathbb{E}\left[\exp\left(\lambda^\top S_{t+1} - \frac{R^2}{2}\|\lambda\|^2_{V_{t+1}^0}\right) \,\middle|\, \mathcal{G}_t\right] \\
&= \mathbb{E}\left[\exp\left(\lambda^\top S_t - \frac{R^2}{2}\lambda^\top V_t^0 \lambda + \eta_t \lambda^\top X_t - \frac{R^2}{2}\lambda^\top \left(X_t X_t^\top\right)\lambda\right) \,\middle|\, \mathcal{G}_t\right] \\
&= \mathbb{E}\left[\underbrace{M_t^\lambda}_{\mathcal{G}_t\text{-adapted}} \underbrace{\exp\left(\eta_t \lambda^\top X_t - \frac{R^2}{2}\left(\lambda^\top X_t\right)^2\right)}_{\leqslant 1} \,\middle|\, \mathcal{G}_t\right] \\
&\qquad\qquad (R\text{-sub-Gaussian control with parameter } \lambda^\top X_t \in \mathbb{R}) \\
&\leqslant M_t^\lambda \,. 
\end{aligned} \tag{A.46}
$$

Hence $(M_t^\lambda)_{t\in\mathbb{N}}$ is a nonnegative supermartingale. Let $\Lambda$ a $\mathbb{R}^d$-valued random variable independent of the rest and $M_t = \mathbb{E}[M_t^\Lambda \mid \mathcal{G}_\infty]$ with $\mathcal{G}_\infty = \sigma(\bigcup_{t\in\mathbb{N}} \mathcal{G}_t)$. As in the proof of Corollary 1.21, $(M_t)_{t\in\mathbb{N}}$ remains a $(\mathcal{G}_t)_{t\in\mathbb{N}}$-nonnegative supermartingale. If $R = 1$, we choose for mixing measure $\Lambda \sim \mathcal{N}(0, 1/\alpha)$ and thus $M_t = F_t(S_t)$ with

$$\begin{aligned} F_t \colon \mathbb{R}^d &\longrightarrow \mathbb{R}_+ \\ y &\longmapsto \left(\frac{\alpha}{2\pi}\right)^{\frac{d}{2}} \int_{\mathbb{R}^d} e^{\lambda^\top y - \frac{1}{2}(\lambda^\top (V_t^0 + \alpha I_d)\lambda)} d\lambda \,. \end{aligned} \tag{A.47}$$

By completing the square in the exponential and letting $\bar{\lambda} = (V_t^\alpha)^{-1} y$, we obtain the following expression for $y \in \mathbb{R}^d$:

$$\begin{aligned} F_t(y) &= \left(\frac{\alpha}{2\pi}\right)^{\frac{d}{2}} \exp\left(\frac{1}{2}\bar{\lambda}^\top V_t^\alpha \bar{\lambda}\right) \int_{\mathbb{R}^d} e^{-\frac{1}{2}(\lambda - \bar{\lambda})^\top V_t^\alpha (\lambda - \bar{\lambda})} d\lambda \quad \text{(Multivariate Gaussian integral)} \\ &= \sqrt{\frac{\alpha^d}{\det V_t^\alpha}} \exp\left(\frac{1}{2}\bar{\lambda}^\top V_t^\alpha \bar{\lambda}\right) . \end{aligned} \tag{A.48}$$

Note that $V_t^\alpha \succcurlyeq \alpha I_d$ and is therefore positive definite (since $\alpha > 0$), and in particular invertible. Consequently, this expression further simplifies as:

$$F_t(y) = \sqrt{\frac{\alpha^d}{\det V_t^\alpha}} \exp\left(\frac{1}{2}\|y\|^2_{(V_t^\alpha)^{-1}}\right) . \tag{A.49}$$

From there, we follow the framework of Theorem 1.20 to obtain the concentration bound for any $\mathbb{N}$-valued $(\mathcal{G}_t)_{t\in\mathbb{N}}$-stopping time. In particular, the choice of $\tau = \inf\{t \in \mathbb{N},\ \|S_t\|^2_{(V_t^\alpha)^{-1}} \geqslant (2\log(1/\delta) + \log(\det V_t^\alpha / \alpha^d))\}$ provides the time-uniform result. Finally, we recover the general case $R \in \mathbb{R}_+^\star$ by considering $(\eta_t/R)_{t\in\mathbb{R}}$ instead. ■

**Extension to continuous time.** As a testament to the flexibility and the generality of the method of mixtures, we briefly extend it to continuous time stochastic processes. We refer to Revuz and Yor (2013) for background material on stochastic differential equations.

**Proposition A.2** (Time-uniform concentration for SDE). *Let two mappings $b\colon \mathbb{R}_+ \times \mathbb{R} \to \mathbb{R}$ and $\sigma\colon \mathbb{R}_+ \times \mathbb{R} \to \mathbb{R}_+^\star$ such that the stochastic differential equation (SDE) represented by $dS_t = b(t,S_t)dt + \sigma(t,S_t)dW_t$ admits a unique strong solution $(S_t)_{t\in\mathbb{R}_+}$ with initial condition $S_0$ (possibly random) with respect to the (augmented) natural filtration $(\mathcal{G}_t)_{t\in\mathbb{R}_+}$ of the Brownian motion $(W_t)_{t\in\mathbb{R}_+}$ (e.g. $(t,y) \mapsto b(t,y)$ and $(t,y) \mapsto \sigma(t,y)$ are Lipschitz continuous in $y$ uniformly in $t$). Assume that almost surely, (i) for all $t \in \mathbb{R}_+$, the mapping $\Psi\colon (t,y) \in \mathbb{R}_+ \times \mathbb{R} \mapsto \int_{S_0}^{y} dy'/\sigma(t,y')$ is well-defined, continuously differentiable in $t$ and twice continuously differentiable in $y$, and (ii) the SDE $d\bar{S}_t = (b(t,S_t)/\sigma(t,S_t) - 1/2\partial_y\sigma(t,S_t) - \int_{S_0}^{S_t} \partial_t\sigma(t,S_y)/\sigma^2(t,y)dy)dt$ admits a unique strong solution with initial condition $\bar{S}_0 = 0$. Then, for a given $\alpha \in \mathbb{R}_+^\star$,*

(i) *the process $(M_t^\lambda)_{t\in\mathbb{R}}$ defined by $M_t = \exp(\lambda(\Psi(t,S_t) - \bar{S}_t) - \lambda^2 t/2)$ for $\lambda \in \mathbb{R}$ and $t \in \mathbb{R}_+$ is a nonnegative martingale;*

(ii) *for $\delta \in (0,1)$, we have the following inequality*

$$\mathbb{P}\left(\forall t \in \mathbb{R}_+,\ \Psi(t,S_t) - \bar{S}_t \leqslant \sqrt{2(t+\alpha)\log\left(\frac{1}{\delta}\sqrt{1+\frac{t}{\alpha}}\right)}\right) \geqslant 1-\delta\,. \tag{A.50}$$

*In particular, if $b = b_{\theta^\star}$ and $\sigma = \sigma_{\theta^\star}$ for $\theta^\star \in \Theta \subseteq \mathbb{R}$, we denote by $\Psi_\theta$, $(\bar{S}_t^\theta)_{t\in\mathbb{R}_+}$ and $\mathbb{P}_\theta$ the corresponding mapping, process and measure associated with $\theta \in \Theta$ and define for $t \in \mathbb{R}_+$ the set*

$$\widehat{\Theta}_{t,\alpha}^\delta = \left\{\theta \in \Theta,\ \left|\Psi_\theta(t,S_t) - \bar{S}_t^\theta\right| \leqslant \sqrt{2(t+\alpha)\log\left(\frac{2}{\delta}\sqrt{1+\frac{t}{\alpha}}\right)}\right\}\,. \tag{A.51}$$

*Then $(\widehat{\Theta}_{t,\alpha}^\delta)_{t\in\mathbb{R}_+}$ is a* ***time-uniform confidence process*** *at level $\delta$ for $\theta^\star$, in the sense that*

$$\mathbb{P}_{\theta^\star}\left(\forall t \in \mathbb{R}_+,\ \theta^\star \in \widehat{\Theta}_{t,\alpha}^\delta\right) \geqslant 1-\delta\,, \tag{A.52}$$

*Proof of Proposition A.2.* Standard applications of Itô's formula (thanks to the regularity assumptions on $\Psi$) shows that the process defined for $t \in \mathbb{R}_+$ by $\widetilde{S}_t = \Psi(t,S_t)$ (known as the Lamperti transform of $S_t$) satisfies the equality $\widetilde{S}_t = S_0 + \bar{S}_t + W_t$ for all $t \in \mathbb{R}_+$ almost surely. Furthermore, the Doleans-Dade formula shows that $M_t^\lambda = \exp(\lambda W_t - \lambda^2 t/2)$ defines a nonnegative martingale for any $\lambda \in \mathbb{R}$. The rest follows from the method of mixtures with mixing measure $\mathcal{N}(0,1/\alpha)$ as per Theorem 1.20. ∎

**Corollary A.3** (Time-uniform confidence processes for classical SDE). *Fix $\alpha \in \mathbb{R}_+^\star$ and $t \in \mathbb{R}_+^\star$, and let $y_0 \in \mathbb{R}$, $\sigma \in \mathbb{R}_+^\star$, $\mu \in \mathbb{R}$ and $\kappa \in \mathbb{R}_+^\star$.*

- ***Brownian motion.*** *Let $dS_t = \mu dt + \sigma dW_t$, $S_0 = y_0$ and define*

$$\widehat{\mu}_t = \frac{S_t - y_0}{t} \quad \text{and} \quad \widehat{\Theta}_{t,\alpha}^\delta = \left[\widehat{\mu}_t \pm \sigma\sqrt{\frac{2}{t}\left(1+\frac{\alpha}{t}\right)\log\left(\frac{2}{\delta}\sqrt{1+\frac{t}{\alpha}}\right)}\right]. \tag{A.53}$$

- ***Geometric Brownian motion.*** *Let $dS_t = \mu S_t dt + \sigma S_t dW_t$, $S_0 = y_0$ and define*

$$\widehat{\mu}_t^\sigma = \frac{1}{t}\log\frac{S_t}{y_0} + \frac{\sigma^2}{2} \quad \text{and} \quad \widehat{\Theta}_{t,\alpha}^\delta = \left[\widehat{\mu}_t^\sigma \pm \sigma\sqrt{\frac{2}{t}\left(1+\frac{\alpha}{t}\right)\log\left(\frac{2}{\delta}\sqrt{1+\frac{t}{\alpha}}\right)}\right]. \tag{A.54}$$

- ***Ornstein-Uhlenbeck.*** *Let $dS_t = \kappa(\mu - S_t)dt + \sigma dW_t$, $S_0 = y_0$ and define*

$$\widehat{\mu}_t^\kappa = \frac{S_t - y_0 e^{-\kappa t}}{1 - e^{-\kappa t}} \quad \text{and}$$
$$\widehat{\Theta}_{t,\alpha}^\delta = \left[\widehat{\mu}_t^\kappa \pm \frac{\sigma}{1-e^{-\kappa t}}\sqrt{2\alpha\left(\frac{1}{2\kappa\alpha} + \left(1 - \frac{1}{2\kappa\alpha}\right)e^{-2\kappa t}\right)\log\left(\frac{2}{\delta}\sqrt{1+\frac{1}{2\kappa\alpha}(e^{2\kappa t}-1)}\right)}\right]. \tag{A.55}$$

*In all cases, $(\widehat{\Theta}_{t,\alpha}^\delta)_{t\in\mathbb{R}_+^\star}$ is a time-uniform confidence process for $\mu$ at level $\delta$.*

*Proof of Corollary A.3.* The case of Brownian motion follows directly from the Gaussian method of mixture, noting that $M_t^\lambda = \exp(\lambda(S_t - \mu t) - \lambda^2\sigma^2/2)$ defines a nonnegative martingale for all $\lambda \in \mathbb{R}$ and $t \in \mathbb{R}_+$ (i.e. exactly as in the discrete time Gaussian case).

For the geometric Brownian motion, the Lamperti transformation takes the (stationary) form $\Psi\colon y \in \mathbb{R}_+^\star \mapsto \int_{y_0}^y dy'/(\sigma y') = \log(y)/\sigma$, the results follows directly from Proposition A.2.

For the Ornstein-Uhlenbeck process, we simplify the problem by considering the transformation $\widetilde{S}_t = e^{\kappa t}(S_t - \mu)/\sigma$, which satisfies the SDE $d\widetilde{S}_t = e^{\kappa t}dW_t$ with $\widetilde{S}_0 = (y_0 - \mu)/\sigma$. Integrating this SDE yields the equation $\widetilde{S}_t = \widetilde{S}_0 + Z_t$ with $Z_t = \int_0^t e^{\kappa s}dW_s$. By the properties of Itô integration with a deterministic integrand, $(Z_t)_{t\in\mathbb{R}_+}$ is a martingale with Gaussian marginals with variance $\sigma_{\kappa,t}^2 = \int_0^t e^{2\kappa s}ds = (e^{2\kappa t} - 1)/(2\kappa)$ (Itô isometry). Therefore, $M_t^\lambda = \exp(\lambda Z_t - \lambda^2\sigma_{\kappa,t}^2/2)$ defines a nonnegative martingale. We conclude by invoking the Gaussian method of mixtures with mixing measure $\mathcal{N}(0, 1/\alpha)$ and rearranging the terms. ∎

**Remark A.4** (The time-uniform Ornstein-Uhlenbeck confidence process does not shrink). *Note that when $t \to +\infty$, in the Ornstein-Uhlenbeck case, $|\widehat{\Theta}^{\delta}_{t,\alpha}| \sim \sigma\sqrt{t}$. In particular, the interval $\widehat{\Theta}^{\delta}_{t,\alpha}$ does not shrink around the point estimate $\widehat{\mu}^{\kappa}_{t}$. This is actually quite intuitive: solving the SDE yields the representation $S_t = \mu + e^{-\kappa t}(y_0 - \mu) + \sigma/\sqrt{2\kappa} W_{1-e^{-2\kappa t}}$, and therefore asymptotically $S_t$ behaves like a Gaussian random variable $\mathcal{N}(\mu, \sigma^2/(2\kappa))$. Consequently, $S_t$ will cross any constant barrier in $t$ with probability 1 if $t$ is large enough, which makes it impossible to have a time-uniform constant (let alone shrinking) confidence process. The other two processes, which are nonstationary, exhibit the shrinking rate $\mathcal{O}(\sqrt{\log(t)/t})$, which is standard for the method of mixtures. Note that the averaged integrated Ornstein-Uhlenbeck process $\int_0^t S_s ds/t$ would lead to a similar concentration.*

**Optimal time-uniform concentration rate.**

*Proof of Proposition 1.28.* First, note that it enough to prove the $\limsup$ bound with $R = 1$, the other cases being deduced easily by symmetry by considering $S_t/R$ or $-S_t$. Again, we make use of the nonnegative supermartingale defined for $t \in \mathcal{T}$ and $\lambda \in \mathbb{R}$ by $M_t^{\lambda} = \exp(\lambda S_t - t\lambda^2/2)$. It follows from Corollary 1.21 or Corollary A.3 that for any $\delta \in (0,1)$,

$$\mathbb{P}\left(\sup_{s\in[0,t]\cap\mathcal{T}} M_s^{\lambda} \geqslant \frac{1}{\delta}\right) \leqslant \delta \tag{A.56}$$

We now construct three sequences $(t_n)_{n\in\mathbb{N}} \in (cT)^{\mathbb{N}}$, $(\lambda_n)_{n\in\mathbb{N}} \in (\mathbb{R}_+^{\star})^{\mathbb{N}}$ and $(\delta_n)_{n\in\mathbb{N}} \in (0,1)^{\mathbb{N}}$ to help us prove the asymptotic upper bound. First, we set choose a geometric time grid, i.e. $t_n = (1+\eta)^n$ for some $\eta \in \mathbb{R}_+^{\star}$ and $n \in \mathbb{N}$. We want the sequence of probabilities defined by the above equation for $t_n$ to be summable, i.e. $\sum_{n\geqslant 1}\delta_n < \infty$; a natural choice is thus $\delta_n = (\log t_n)^{-(1+\varepsilon)} = \mathcal{O}(n^{-(1+\varepsilon)})$ for some $\varepsilon \in \mathbb{R}_+^{\star}$. We now choose $\lambda_n$ to make appear the $\log\log$ term of the upper bound, i.e. $\lambda_n = (1+\varepsilon)\sqrt{2\log\log(t_n)/t_n}$. By extracting $S_s$ from the expression of $M_s^{\lambda_n}$ for $s \in [0, t_n] \cap \mathcal{T}$, we obtain the following bound:

$$\mathbb{P}\left(\sup_{s\in[0,t_n]\cap\mathcal{T}} S_s - \frac{\lambda_n s}{2} \geqslant \sqrt{\frac{t_n \log\log t_n}{2}}\right) = \mathcal{O}\left(\frac{1}{n^{1+\varepsilon}}\right). \tag{A.57}$$

Consequently, the Borel-Cantelli lemma shows that almost surely a finite number of these occur, i.e. there exists a random variable $N$ in $\mathcal{T}$ such that $\sup_{s\in[0,t_n]\cap\mathcal{T}} S_s - \frac{\lambda_n s}{2} \geqslant \sqrt{\frac{t_n \log\log t_n}{2}}$ for all $n \geqslant N$. For $t \in [t_n, t_{n+1}] \cap \mathcal{T}$, we therefore have the following control:

$$S_t \leqslant \sqrt{\frac{t_n \log\log t_n}{2}} + \frac{\lambda_n t_{n+1}}{2} = (1 + (1+\eta)(1+\varepsilon))\sqrt{\frac{t_n \log\log t_n}{2}}$$

$$\leqslant (1+(1+\eta)(1+\varepsilon))\sqrt{\frac{t\log\log t}{2}}\,, \tag{A.58}$$

which holds for any $\varepsilon, \eta \in \mathbb{R}_+^\star$. Hence, by letting $n \to +\infty$, we have that almost surely:

$$\limsup_{t\to+\infty} \frac{S_t}{\sqrt{2t\log\log t}} \leqslant 1\,. \tag{A.59}$$

■

### Catching the optimal rate with mixtures

*Proof of Proposition 1.29.* Let $(u_t)_{t\in\mathbb{N}}$ be an arbitrary (for now) nondecreasing sequence. The strategy to control $\mathbb{P}\left(S_\tau \geqslant u_\tau\right)$ is to split the event $\{S_\tau \geqslant u_\tau\}$ over the geometric grid $(t_n)_{n\in\mathbb{N}}$ and apply the supermartingale transform $F^{\alpha_n}$. Compared to the standard method of mixtures, this allows to use a split-dependent hyperparameter $\alpha_n$ rather than a constant $\alpha$. We have:

$$\begin{aligned}
\mathbb{P}\left(S_\tau \geqslant u_\tau\right) &\leqslant \sum_{n=1}^{+\infty} \mathbb{P}\left(\exists t \in [t_{n-1}, t_n),\ S_t \geqslant u_t\right) \\
&= \sum_{n=1}^{+\infty} \mathbb{P}\left(\exists t \in [t_{n-1}, t_n),\ M_t^{\alpha_{n-1}} \geqslant F_t^{\alpha_{n-1}}(u_t)\right) \quad (x \mapsto F_t^{\alpha_{n-1}}(x) \text{ is nondecreasing}) \\
&\leqslant \sum_{n=1}^{+\infty} \mathbb{P}\left(\exists t \in [t_{n-1}, t_n),\ M_t^{\alpha_{n-1}} \geqslant F_{t_n}^{\alpha_{n-1}}(u_{t_{n-1}})\right) \\
&\qquad\qquad (t \mapsto F_t^{\alpha_{n-1}} \text{ is nonincreasing and } u_t \text{ is nondecreasing}) \\
&\leqslant \sum_{n=1}^{+\infty} \frac{\mathbb{E}\left[M_{t_{n-1}}^{\alpha_{n-1}}\right]}{F_{t_n}^{\alpha_{n-1}}(u_{t_{n-1}})} \qquad (\text{Theorem 1.20}) \\
&\leqslant \sum_{n=1}^{+\infty} \frac{1}{F_{t_n}^{\alpha_{n-1}}(u_{t_{n-1}})}\,. \qquad (M^{\alpha_{n-1}} \text{ is a supermartingale})
\end{aligned} \tag{A.60}$$

It remains to choose the sequence $(u_t)_{t\in\mathbb{N}}$ such that the series above converges to $\delta$. A natural choice is $F_{t_n}^{\alpha_{n-1}}(u_{t_{n-1}}) = \frac{1}{\delta}\zeta(\eta)n^\eta$, which leads to

$$u_{t_{n-1}} = \left(F_{t_n}^{\alpha_{n-1}}\right)^{-1}\left(\frac{\zeta(\eta)n^\eta}{\delta}\right) \quad \text{and} \quad u_t = u_{t_{n-1}} \text{ for } t \in [t_{n-1}, t_n)\,, \tag{A.61}$$

where $\left(F_{t_n}^{\alpha_{n-1}}\right)^{-1}$ is the inverse of $\left(F_{t_n}^{\alpha_{n-1}}\right)^{-1}$ if it is strictly increasing, and the appropriate pseudo-inverse (which is well-defined since it is a nondecreasing function by assumption) otherwise. Finally, note that $t \in [t_{n-1}, t_n)$ is equivalent to $n = \lceil \log t / \log \Delta \rceil = n_t$. We conclude by invoking the supermartingale construction of Theorem 1.20. ■

*Proof of Corollary 1.30.* In the $R$-sub-Gaussian case with $(S_t)_{t\in\mathbb{N}} = (\sum_{s=1}^t Y_s)_{t\in\mathbb{N}}$, inverting the supermartingale mapping of Proposition 1.29 yields the expression (similar to Corollary 1.21) $(F_t^\alpha)^{-1}\colon y \in \mathbb{R}_+^\star \mapsto (R\sqrt{2(t+\alpha)\log(y\sqrt{1+t/\alpha})}$), and thus

$$u_t = R\sqrt{2(t_{n_t} + \alpha_{n_{t_n}-1})\log\left(\frac{\zeta(\eta)n_t^\eta}{\delta}\sqrt{1+\frac{t_{n_t}}{\alpha_{n_{t_n}-1}}}\right)}\,. \tag{A.62}$$

Optimising $\alpha_{n-1}$ to tighten the bound at $t = t_n$ yields $\alpha_{n-1} = \gamma_\delta t_n = \operatorname{argmin}_\alpha (F_{t_n}^\alpha)^{-1}(\zeta(\eta)n^\eta/\delta)$ with $\gamma_\delta = -1/(1+W_{-1}(-\delta^2/e))$. By plugging this in the above bound, we obtain the expression

$$u_t = R\sqrt{2(1+\gamma_\delta)t_{n_t}\log\left(\frac{\zeta(\eta)n_t^\eta}{\delta}\sqrt{1+\frac{1}{\gamma_\delta}}\right)}\,. \tag{A.63}$$

In particular, making $\alpha_{n_{t_n}-1}$ proportional to $t_{n_t}$ cancels out the dependency on $t$ in the term $t_{n_t}/\alpha_{n_{t_n}-1} = 1/\gamma_\delta$, and thus $u_t = \mathcal{O}(\sqrt{t\log(n_t/\delta)}) = \mathcal{O}(\sqrt{t\log(\log(t)/\delta)})$ for $t \to +\infty$. We conclude by considering $(\widehat{\mu}_t)_{t\in\mathbb{N}} = (S_t/t)_{t\in\mathbb{N}}$ and the bounds $(\pm u_t/t)_{t\in\mathbb{N}}$. ■

## Generic UCB algorithm for multiarmed bandits

### Generic pseudo regret bound.

*Proof of Theorem 1.33.* We write $(\widehat{\Theta}_{k,n}^\delta)_{n\in\mathbb{N}} = ([\mathrm{LCB}_{k,n}^\delta, \mathrm{UCB}_{k,n}^\delta])_{n\in\mathbb{N}}$ and define the following good events for $k \in [K]$ and $k^\star = \operatorname{argmax}_{k\in[K]} \rho_k$:

$$\mathcal{G}_k^1 = \left\{\rho^\star < \min_{1\leqslant t\leqslant T} \mathrm{UCB}_{k^\star,N_t^{k^\star}}^\delta\right\}\,, \quad \mathcal{G}_k^2 = \left\{\rho^\star > \mathrm{UCB}_{k,\tau_k^\delta}^\delta\right\}\,, \quad \text{and} \quad \mathcal{G}_k = \mathcal{G}_k^1 \bigcap \mathcal{G}_k^2\,. \tag{A.64}$$

On the event $\mathcal{G}_k$, assume that $N_T^k > \tau_k^\delta$, then there exists $t \leqslant T$ such that $N_{t-1}^k = \tau_k^\delta$ and $\pi_t = k$ (there exists a time at which arm $k$ has been played $\tau_k^\delta$ time and the policy recommended to play arm $k$ one more time). In that case, we have

$$\mathrm{UCB}_{k,N_{t-1}^k}^\delta = \mathrm{UCB}_{k,\tau_k^\delta}^\delta < \rho^\star < \mathrm{UCB}_{k^\star,N_{t-1}^{k^\star}}^\delta\,. \tag{A.65}$$

Therefore $\mathrm{UCB}_{k,N_{t-1}^K}^\delta < \mathrm{UCB}_{k^\star,N_{t-1}^{k^\star}}^\delta$, and thus arm $k$ cannot be played at time $t$, which contradicts $\pi_t = k$. Consequently, we have $N_T^k \leqslant \tau_k^\delta$ on the event $\mathcal{G}_k$.

We now control the probability of the complement event $\bar{\mathcal{G}_k} = \bar{\mathcal{G}_k^1} \cup \bar{\mathcal{G}_k^2}$. By definition of the confidence sequence for arm $k^\star$, we have

$$\begin{aligned}\mathbb{P}\left(\bar{\mathcal{G}_k^1}\right) = \mathbb{P}\left(\exists t \in \{1,\dots,T\},\ \rho^\star \geqslant \mathrm{UCB}_{k^\star,N_t^{k^\star}}^\delta\right) &\leqslant \mathbb{P}\left(\exists t \in \mathbb{N},\ \rho^\star \notin \widehat{\Theta}_{k^\star,N_t^{k^\star}}^\delta\right)\\ &\leqslant \mathbb{P}\left(\exists n \in \mathbb{N},\ \rho^\star \notin \widehat{\Theta}_{k^\star,n}^\delta\right)\end{aligned}$$

$$\leqslant \delta\,. \tag{A.66}$$

For the second complement event, we note that $\rho^\star = \rho_k + \Delta_k$ and

$$\mathbb{P}\left(\bar{\mathcal{G}_k^2}\right) = \mathbb{P}\left(\rho_k \leqslant \mathrm{UCB}^\delta_{k,\tau_k^\delta} - \Delta_k\right) = \mathbb{P}\left(\rho_k \leqslant \mathrm{LCB}^\delta_{k,\tau_k^\delta} + \underbrace{\mathrm{UCB}^\delta_{k,\tau_k^\delta} - \mathrm{LCB}^\delta_{k,\tau_k^\delta} - \Delta_k}_{<0 \text{ by definition of } \tau_k^\delta}\right)$$
$$\leqslant \mathbb{P}\left(\rho_k \notin \widehat{\Theta}^\delta_{k,\tau_k^\delta}\right)$$
$$\leqslant \delta\,. \tag{A.67}$$

Therefore, using a simple union bound yields $\mathbb{P}(\bar{\mathcal{G}_k}) \leqslant 2\delta$. The bound in probability follows from another union argument over all arms $k \in [K]$. Moreover, using the trivial inequality $N_T^k \leqslant T$, we obtain

$$\mathbb{E}_{\pi,\nu_\pi^{\otimes T}}\left[N_T^k\right] = \mathbb{E}_{\pi,\nu_\pi^{\otimes T}}\left[N_T^k \mathbb{1}_{\mathcal{G}_k}\right] + \mathbb{E}_{\pi,\nu_\pi^{\otimes T}}\left[N_T^k \mathbb{1}\bar{\mathcal{G}_k}\right] \leqslant \mathbb{E}_{\pi,\nu_\pi^{\otimes T}}\left[\tau_k^\delta\right] + T\mathbb{E}_{\pi,\nu_\pi^{\otimes T}}\left[\mathbb{1}\bar{\mathcal{G}_k}\right]$$
$$\leqslant \mathbb{E}_{\pi,\nu_\pi^{\otimes T}}\left[\tau_k^\delta\right] + 2\delta T\,. \tag{A.68}$$

■

**Sub-Gaussian UCB.**

*Proof of Corollary 1.34.* For $\alpha \in \mathbb{R}_+^\star$, the width of the $R$-sub-Gaussian confidence sequence at sample size $n \in \mathbb{N}$ is $2R\sqrt{2/n(1+\alpha/n)\log(2\sqrt{1+n/\alpha}/\delta)}$, which is asymptotically equivalent to $D(n) = 2R\sqrt{1/n\log(4n/(\alpha\delta^2))}$ for $n \to +\infty$. For $x \in \mathbb{R}_+$, we solve the equation $D(x) = \Delta_k$, which after simple algebra yields, with $\gamma = \Delta_k^2/(4R^2)$,

$$-\gamma x e^{-\gamma x} = -\frac{\gamma\alpha\delta^2}{4}\,. \tag{A.69}$$

For $\delta = 1/T$ and $T$ large enough, the right-hand side is in $(-1/e, 0)$, and therefore the solution to this equation is given by the first negative branch of the Lambert $W$ function:

$$x = -\frac{1}{\gamma}W_{-1}\left(-\frac{\gamma\alpha}{4T^2}\right)\,. \tag{A.70}$$

Substituting $\gamma$ and using the asymptotic $W_{-1}(z) = \log(-z) - \log(-\log(-z))$ for $z \to 0^-$ yields the result. ■

**Iterated logarithm UCB.**

*Proof of Corollary 1.35.* We use the same technique as in the proof of Corollary 1.34 and seek to find the solution in $x \in \mathbb{R}_+$ to the equation $C\sqrt{\log(\log(x)/\delta)/x} = \Delta_k$. For $\delta = 1/T$, we obtain

$$\log(x)e^{-\gamma x} = \frac{1}{T}\,, \tag{A.71}$$

where $\gamma = \Delta_k^2/C^2$. Let $\omega\colon \mathbb{R}_+^\star \to \mathbb{R}_+^\star$ defined implicitly by $\log \omega(z) e^{-\gamma\omega(z)} = 1/T$. This does not correspond to a known special function, so we have to do an ad hoc asymptotic analysis (details are essentially the same as the proof of the asymptotic expansion of the Lambert $W$ function). To this end, we write for $z \in \mathbb{R}_+^\star$,

$$\begin{aligned}\omega(z) = \frac{1}{\gamma}\log\left(\frac{\log\omega(z)}{z}\right) &= \frac{1}{\gamma}\log\left(\frac{\log\left(\frac{1}{\gamma}\log\left(\frac{\log\omega(z)}{z}\right)\right)}{z}\right)\\ &= \frac{1}{\gamma}\log\frac{1}{z} + \frac{1}{\gamma}\log\log\left(\frac{1}{\gamma}\log\left(\frac{\log\omega(z)}{z}\right)\right)\,. \end{aligned} \tag{A.72}$$

When $z \to 0^+$, we obtain $\omega(z) = 1/\gamma\log(1/z) + 1/\gamma\log\log(1/\gamma\log(1/z)) + o(\log\log\log(1/z))$ (after carefully bounding the remainder). We conclude by substituting $\gamma$. ∎

**Obstruction to $\mathcal{O}(\sqrt{\log(n)/n})$ concentration rate.**

*Proof of Corollary 1.36.* Assume the opposite, i.e. almost surely there exists $C \in \mathbb{R}_+^\star$ such that for all $n_0 \in \mathbb{N}$, there exists $n \geqslant n_0$ satisfying the inequality $|\Theta_n^\delta| \leqslant Cn^{-\alpha}\log(n/\delta)^\beta$. The sequence $(\widehat{\Theta}_n^\delta)_{n\in\mathbb{N}}$ can be made nonincreasing by taking the running intersections, and so this inequality implies that $|\Theta_n^\delta| \leqslant Cn^{-\alpha}\log(n/\delta)^\beta$ for all $n \in \mathbb{N}$ large enough. Now, consider a two-armed bandit model $(\nu_1 \otimes \nu_2, \mathcal{F} \otimes \mathcal{F})$ such that $\Delta = \rho(\nu_1) - \rho(\nu_2) > 0$. Then the policy implemented by Algorithm 1 with $\mathrm{UCB}_{k,n}^\delta = \max\widehat{\Theta}_n^\delta(\mathbb{Y}_{(n)}^k)$ and $\mathbb{Y}_{(n)}^k = (Y_i^k)_{i=1}^n$ an i.i.d. sample drawn from $\nu_k$, for $k \in \{1,2\}$ and $n \in \mathbb{N}$, suffers a pseudo regret bounded by $\mathcal{O}(\log T)$. Indeed, solving for $x \in \mathbb{R}_+^\star$ the equation $Cx^{-\alpha}\log(x/\delta)^\beta = \Delta$ yields $xe^{-\gamma x^{\alpha/\beta}} = \delta$ with $\gamma = (\Delta/C)^{1/\beta}$. Letting $\omega(z)$ be the solution to $\omega(z)e^{-\omega(z)^{\alpha/\beta}} = z \in \mathbb{R}_+^\star$ yields

$$\omega(z) = \left(\frac{1}{\gamma}\log\left(\frac{\omega(z)}{z}\right)\right)^{\frac{\beta}{\alpha}} = \left(\frac{1}{\gamma}\log\frac{1}{z} + \frac{\beta}{\alpha\gamma}\log\left(\frac{1}{\gamma}\log\left(\frac{\omega(z)}{z}\right)\right)\right)^{\frac{\beta}{\alpha}}\,, \tag{A.73}$$

which, similarly the the proofs above, yields $\omega(z) = (1/\gamma\log(1/z))^{\beta/\alpha} + o((\log(1/z))^{\beta/\alpha})$. Setting $\delta = 1/T$ and applying the analysis of Theorem 1.33 shows that the number of pulls to arm 2 is asymptotically upper bounded by $\mathcal{O}(\log(T)^{\beta/\alpha}) = \mathcal{O}(\log T)$. This is incompatible with the lower bound of Theorem 1.6 since $\mathcal{K}_{\inf}^{\mathcal{F},\rho}(\nu_2;\rho_1) = 0$. ∎

**Heavy tailed UCB.**

*Proof of Corollary 1.37.* Again, we seek to find $x \in \mathbb{R}_+^\star$ such that $CM^{1/(1+\varepsilon)}(1/x\log(x/\delta))^{\varepsilon/(1+\varepsilon)} = \Delta_k$, i.e. $-\gamma x e^{-\gamma x} = -\gamma\delta$ with $\gamma = M^{-1/\varepsilon}(\Delta_k/C)^{1+1/\varepsilon}$. We conclude by invoking the asymptotic expansion of $W_{-1}$ just as in the proof of Corollary 1.34. ∎

**Heavy tailed UCB corrupted by Nature.**

*Proof of Corollary 1.38.* Let $n_k^{\delta,\delta'} = \lceil \log(4/(\delta\delta'))/\eta \rceil$ and define the good event

$$\mathcal{H}_k = \left\{ |\Theta^\delta_{k,n_k^{\delta,\delta'}}| \leqslant C\kappa^{\frac{1}{1+\varepsilon}}\eta^{\frac{\varepsilon}{1+\varepsilon}} \right\} \subseteq \bigcap_{n \geqslant n_k^{\delta,\delta'}} \left\{ |\Theta^\delta_{k,n}| \leqslant C\kappa^{\frac{1}{1+\varepsilon}}\eta^{\frac{\varepsilon}{1+\varepsilon}} \right\} \subseteq \bigcap_{n \geqslant n_k^{\delta,\delta'}} \left\{ |\Theta^\delta_{k,n}| < \Delta_k \right\}, \tag{A.74}$$

where the first inclusion holds since $(\widehat{\Theta}^\delta_{k,n})_{n\in\mathbb{N}}$ is nonincreasing for the set inclusion.

Using the notations of the proof of Theorem 1.33, we obtain

$$\begin{aligned}
\mathbb{E}_{\pi,\nu_\pi^{\eta\otimes T}}\left[N_T^k\right] &= \mathbb{E}_{\pi,\nu_\pi^{\eta\otimes T}}\left[N_T^k \mathbb{1}_{\mathcal{G}_k}\right] + \mathbb{E}_{\pi,\nu_\pi^{\eta\otimes T}}\left[N_T^k \mathbb{1}_{\bar{\mathcal{G}}_k}\right] \\
&\leqslant \mathbb{E}_{\pi,\nu_\pi^{\eta\otimes T}}\left[\tau_k^\delta\right] + \mathbb{E}_{\pi,\nu_\pi^{\eta\otimes T}}\left[N_T^k \mathbb{1}_{\bar{\mathcal{G}}_k}\mathbb{1}_{\mathcal{H}_k}\right] + \mathbb{E}_{\pi,\nu_\pi^{\eta\otimes T}}\left[N_T^k \mathbb{1}_{\bar{\mathcal{G}}_k}\mathbb{1}_{\bar{\mathcal{H}}_k}\right] \\
&= \mathbb{E}_{\pi,\nu_\pi^{\eta\otimes T}}\left[\tau_k^\delta \mathbb{1}_{\mathcal{H}_k}\right] + \mathbb{E}_{\pi,\nu_\pi^{\eta\otimes T}}\left[\tau_k^\delta \mathbb{1}_{\bar{\mathcal{H}}_k}\right] + \mathbb{E}_{\pi,\nu_\pi^{\eta\otimes T}}\left[N_T^k \mathbb{1}_{\bar{\mathcal{G}}_k}\mathbb{1}_{\mathcal{H}_k}\right] + \mathbb{E}_{\pi,\nu_\pi^{\eta\otimes T}}\left[N_T^k \mathbb{1}_{\bar{\mathcal{G}}_k}\mathbb{1}_{\bar{\mathcal{H}}_k}\right].
\end{aligned} \tag{A.75}$$

On $\mathcal{H}_k$, we have $|\Theta^\delta_{k,n_k^{\delta,\delta'}}| < \Delta_k$ and therefore $\tau_k^\delta \leqslant n_k^{\delta,\delta'}$. Moreover, still on $\mathcal{H}_k$, we have the upper bound $\mathbb{P}_{\pi,\nu_\pi^{\eta\otimes T}}(\bar{\mathcal{G}}_k \cup \mathcal{H}_k) \leqslant 2\delta$ (the proof reads identically to that of Theorem 1.33). On the complement event $\bar{\mathcal{H}}_k$, which holds with probability at most $\delta'$, we use the trivial upper bounds $\tau_k^\delta \leqslant T$ and $N_T^k \leqslant T$. Combining these arguments, we obtain

$$\mathbb{E}_{\pi,\nu_\pi^{\eta\otimes T}}\left[N_T^k\right] \leqslant n_k^{\delta,\delta'} + (3\delta+\delta')T \leqslant \frac{1}{\eta}\log\left(\frac{4}{\delta\delta'}\right) + 1 + (3\delta+\delta')T\,. \tag{A.76}$$

We conclude by setting $\delta = \delta' = 1/T$. ∎

## A.4 Some benchmarks for concentration

**Hedged capital.** We reproduce here the construction of Waudby-Smith and Ramdas (2023), which eschews the need for a suitable mixing measure while still preserving the supermartingale property by using *predictable weights* and provide state-of-the art concentration for bounded distributions.

**Proposition A.5** (Hedged capital concentration). *Let $\underline{B}, \overline{B} \in \mathbb{R}$, and $(Y_t)_{t\in\mathbb{N}}$ an i.i.d. sequence drawn from $\nu \in \mathcal{F}_{[\underline{B},\overline{B}]}$. We define for $m \in [\underline{B}, \overline{B}]$, $t \in \mathbb{N}$ and $\delta \in (0,1)$ the following quantities:*

$$
\begin{aligned}
&\mathcal{K}_t(m) = \mathcal{K}_t^+(m) \vee \mathcal{K}_t^-(m)\,, \\
&\mathcal{K}_t^+(m) = \prod_{s=1}^{t}\left(1 + \lambda_s^+(m)\frac{Y_s - m}{\overline{B} - \underline{B}}\right) \quad \textit{and} \quad \mathcal{K}_t^-(m) = \prod_{s=1}^{t}\left(1 - \lambda_s^-(m)\frac{Y_s - m}{\overline{B} - \underline{B}}\right)\,, \\
&\lambda_t^+(m) = |\lambda_t| \wedge \frac{\overline{B} - \underline{B}}{2(m - \underline{B})} \quad \textit{and} \quad \lambda_t^-(m) = |\lambda_t| \wedge \frac{\overline{B} - \underline{B}}{2(\overline{B} - m)}\,, \\
&\lambda_t = \sqrt{\frac{2\log\frac{2}{\delta}}{\widehat{\sigma}_{t-1}^2\, t\log(t+1)}}\,, \quad \widehat{\sigma}_t^2 = \frac{\frac{(\overline{B}-\underline{B})^2}{4} + \sum_{s=1}^{t}(Y_s - \widehat{\mu}_s)^2}{t+1} \quad \textit{and} \quad \widehat{\mu}_t = \frac{\frac{\underline{B}+\overline{B}}{2} + \sum_{s=1}^{t} Y_s}{t+1}\,, \\
&\widehat{\Theta}_t^\delta = \left\{ m \in \left[\underline{B}, \overline{B}\right],\ \mathcal{K}_t(m) < \frac{1}{\delta} \right\}. \qquad \text{(A.77)}
\end{aligned}
$$

*Then $(\widehat{\Theta}_t^\delta)_{t\in\mathbb{N}}$ is a **time-uniform confidence sequence** at level $\delta$ for $\mu$.*

We refer to Waudby-Smith and Ramdas (2023, Theorem 3) for more details and the proof of this result. Note that similarly to many of the confidence sequences introduced in this thesis (Chapters 3 and 4), the sets $\widehat{\Theta}_t^\delta$ are computed numerically rather than using an explicit formula. In the hedged capital case, these sets are provably convex, so it is sufficient to apply a numerical root search to find the boundary points, determined by the equation $\mathcal{K}_t(m) = 1/\delta$.

To better understand their construction, we also adapt it to the sub-Gaussian case, and compare it with the sub-Gaussian method of mixtures (Corollary 1.21).

**Proposition A.6** (Sub-Gaussian hedged capital concentration). *Let $R \in \mathbb{R}_+^\star$, $(Y_t)_{t\in\mathbb{N}}$ an i.i.d. sequence drawn from $\nu \in \mathcal{F}_{\mathcal{G},R}$ and $(\mathcal{G}_t)_{t\in\mathbb{N}}$ the corresponding natural filtration. For any $(\mathcal{G}_t)_{t\in\mathbb{N}}$-predictable process $(\lambda_t)_{t\in\mathbb{N}}$, we define*

$$\begin{aligned} \mathcal{K}_t^\lambda \colon \mathbb{R} &\longrightarrow \mathbb{R}_+ \\ m &\longmapsto \prod_{s=1}^{t} \exp\left(\lambda_s(Y_s - m) - \frac{R^2\lambda_s^2}{2}\right) . \end{aligned} \tag{A.78}$$

*Then $(\mathcal{K}_t^\lambda(\mu))_{t\in\mathbb{N}}$ defines a nonnegative supermartingale. Moreover, we consider*

$$\widehat{\mu}_t^\lambda = \frac{\sum_{s=1}^t \lambda_s Y_s}{\sum_{s=1}^t \lambda_s} \quad \text{and} \quad \widehat{\Theta}_t^\delta = \left[\widehat{\mu}_t^\lambda \pm \frac{R^2\sum_{s=1}^t \lambda_s^2 + 2\log\frac{2}{\delta}}{\sum_{s=1}^t \lambda_s}\right] . \tag{A.79}$$

*Then $(\widehat{\Theta}_t^\delta)_{t\in\mathbb{N}}$ is a* ***time-uniform confidence sequence*** *at level $\delta$ for $\mu$. In particular, we recommend*

$$\lambda_t = \sqrt{\frac{2\log\frac{2}{\delta}}{\widehat{\sigma}_{t-1}^2 t\log(t+1)}}, \quad \widehat{\sigma}_t^2 = \frac{R + \sum_{s=1}^t (Y_s - \widehat{\mu}_s)^2}{t+1} \quad \text{and} \quad \widehat{\mu}_t = \frac{1}{t}\sum_{s=1}^t Y_s \tag{A.80}$$

*Proof of Proposition A.6.* We first prove that $(\mathcal{K}_t^\lambda(\mu))_{t\in\mathbb{N}}$ is a nonnegative supermartingale. Indeed, for $t \in \mathbb{N}$, we have

$$\begin{aligned} \mathbb{E}\left[\mathcal{K}_{t+1}^\lambda(\mu) \,\middle|\, \mathcal{G}_t\right] &= \mathbb{E}\left[\underbrace{\mathcal{K}_t^\lambda(\mu)}_{\mathcal{G}_t\text{-measurable}} \exp\left(\lambda_{t+1}(Y_{t+1} - \mu) - \frac{R^2\lambda_{t+1}^2}{2}\right) \,\middle|\, \mathcal{G}_t\right] \\ &= \mathcal{K}_t^\lambda(\mu)\underbrace{\mathbb{E}\left[\exp\left(\lambda_{t+1}(Y_{t+1} - \mu) - \frac{R^2\lambda_{t+1}^2}{2}\right) \,\middle|\, \mathcal{G}_t, \lambda_{t+1}\right]}_{\leqslant 1} \\ &\qquad\qquad (\lambda_{t+1} \text{ is } \mathcal{G}_t\text{-measurable}) \\ &\leqslant \mathcal{K}_t^\lambda(\mu) . \end{aligned} \tag{A.81}$$

We then notice that $\widehat{\Theta}_t^\delta = \{m \in \mathbb{R},\ \mathcal{K}_t^\lambda(m) < \frac{2}{\delta}\} \cap \{m \in \mathbb{R},\ \mathcal{K}_t^{-\lambda}(m) < \frac{2}{\delta}\}$, and we conclude by invoking Theorem 1.20. Finally, the recommended tuning of the predictable sequence $(\lambda_t)_{t\in\mathbb{N}}$ follows the a simple heuristic. Let $t \in \mathbb{N}$ and assume $\lambda_1 = \cdots = \lambda_t = \lambda^\star$. Optimising the width of the confidence set at time $t$, namely $|\widehat{\Theta}_t^\delta| = \frac{R^2(\lambda^\star)^2 + 2/t\log(2/\delta)}{\lambda^\star}$ yields $\lambda^\star = \sqrt{2/(R^2 t)\log\frac{2}{\delta}}$. We then simply substitute $R$ with a standard predictable estimator (biased towards $R$) and replace $t$ with $t\log(t)$ as suggested in Waudby-Smith and Ramdas (2023) in order to mimick the law of iterated logarithm. ■

Note that both methods produce a *weighted* mean estimator $\widehat{\mu}_t^\lambda$ rather than the classical empirical mean $\widehat{\mu}_t$. In particular, $\widehat{\mu}_t^\lambda$ is dependent of the order of the observations $(Y_s)_{s=1}^t$. It is possible to remove this exogenous dependency by averaging over shuffled sequences $(Y_{\sigma(s)})_{s=1}^t$ for $\sigma \in \mathfrak{S}_t$, at the cost of making the confidence sequences even less explicit.

In Figure A.2, we empirically compare the sub-Gaussian hedged capital approach of Proposition A.6 with the method of mixtures of Corollary 1.21 for various mixing parameter $\alpha$ in four different sub-Gaussian distributions: $\mathcal{N}(0,1)$ $(R=1)$, $\mathrm{Beta}(0.5,1)$ $(R=1/2)$, the triangular distribution over $[-1,1]$ peaked at $0$ $(R=1)$, and finally an example of simulated real-world distribution representing maize grain yield, which we previously used in Chapter 5, Figure 4.6b (see also Section 6.4 in Chapter 6 for more details on the DSSAT simulator). In all cases, the method of mixtures performed better, resulting in tighter confidence sequences. This illustrates that the hedged capital approach shines in nonparametric settings where, contrary to the sub-Gaussian case, no natural mixing measure are available, such as the bounded support case.

**Discrete mixtures with inverted stitching.** Another construction suggests a compromise between time peeling and the method of mixtures with Gaussian priors by considering *discrete* mixing measure. The proposition below is a special instance of Howard et al. (2021, Theorem 3).

**Proposition A.7** (Sub-Gaussian discrete mixture with inverted stitching concentration). *Let $R \in \mathbb{R}_+^\star$, $(Y_t)_{t\in\mathbb{N}}$ an i.i.d. sequence drawn from $\nu \in \mathcal{F}_{\mathcal{G},R}$. For $\delta \in (0,1)$, we consider $(\widehat{\mu}_t)_{t\in\mathbb{N}} = (1/t\sum_{s=1}^t Y_s)$ and*

$$\left(\widehat{\Theta}_t^\delta\right)_{t\in\mathbb{N}} = \left(\left[\widehat{\mu}_t \pm 1.7R\sqrt{\frac{\log\log(2t) + 0.72\log\left(\frac{5.2}{\alpha}\right)}{t}}\right]\right)_{t\in\mathbb{N}} \tag{A.82}$$

*Then $(\widehat{\Theta}_t^\delta)_{t\in\mathbb{N}}$ is a* ***time-uniform confidence sequence*** *at level $\delta$ for $\mu$.*

Interestingly, this time-uniform confidence bound matches the optimal concentration rate of the law of iterated logarithms (Proposition 1.28). However, as argued in Section 1.4, this often comes at the cost of higher leading term, illustrated here by the seemingly ad hoc constants $1.7$, $0.72$ and $5.2$, making the improvement over the standard method of mixtures only noticeable for extremely large sample sizes. We confirm this remark in Figure A.3 where we compare this discrete mixture bound to the classical Gaussian method of mixture (Corollary 1.21). Indeed, the tighter rate $\mathcal{O}(\sqrt{\log\log(t)/t})$ rate only shows for large $t$ (note that if we know in advance the approximate final sample size $T$, setting $\alpha = 0.12T$ makes the sub-Gaussian mixture bound sharper around $t = T$, thus further postponing the time where the discrete mixture shines).

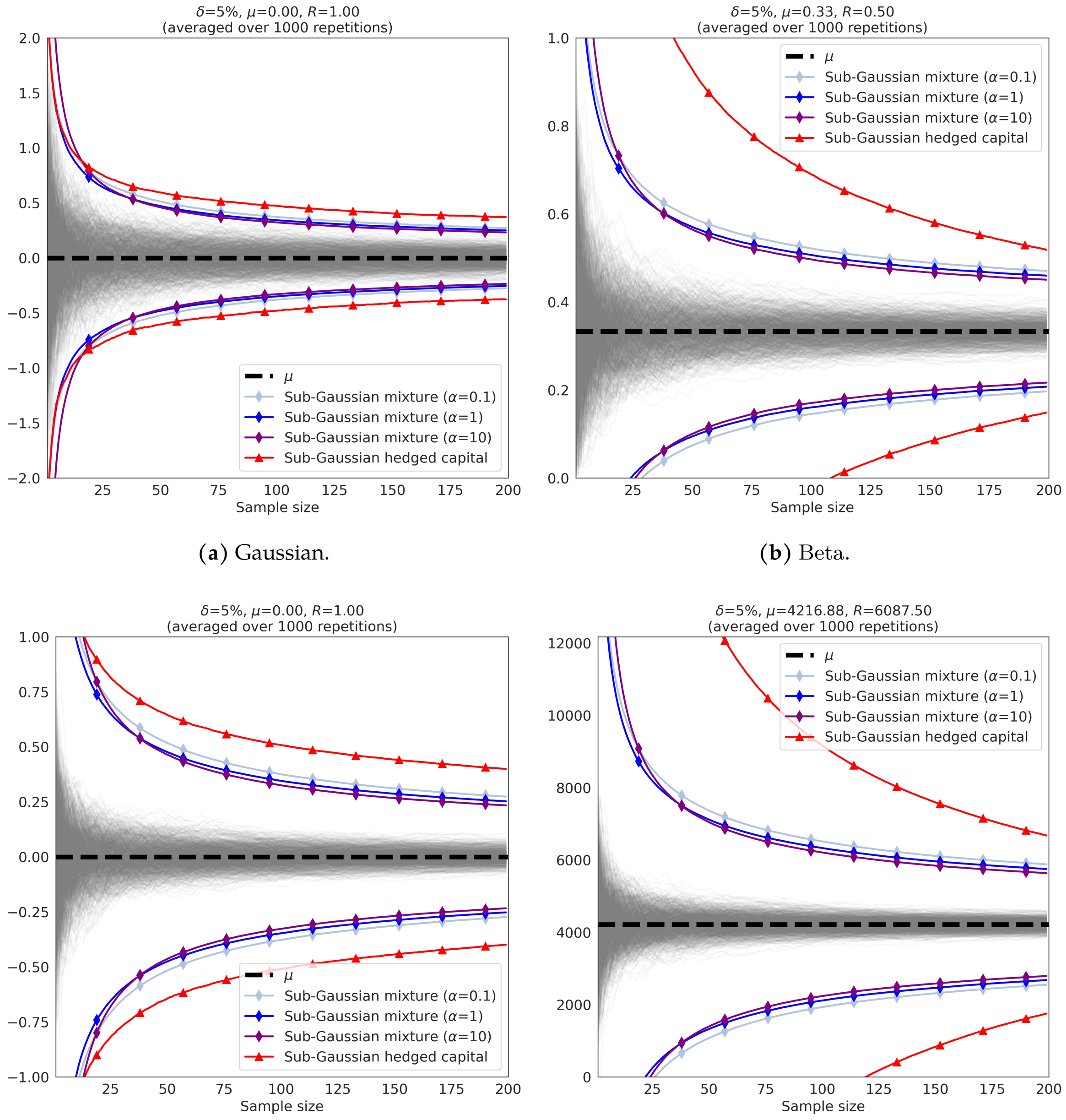

(**a**) Gaussian. (**b**) Beta. (**c**) Triangular. (**d**) DSSAT.

**Figure A.2** – Comparison of median confidence envelopes around the mean for distributions $\mathcal{N}(0, 1)$, $\mathrm{Beta}(0.5, 1)$, triangular on $(-1, 0, 1)$ and a simulated DSSAT yield distribution, as a function of the sample size $t$, over 1000 independent replicates. Grey lines are trajectories of empirical means $\widehat{\mu}_t$.

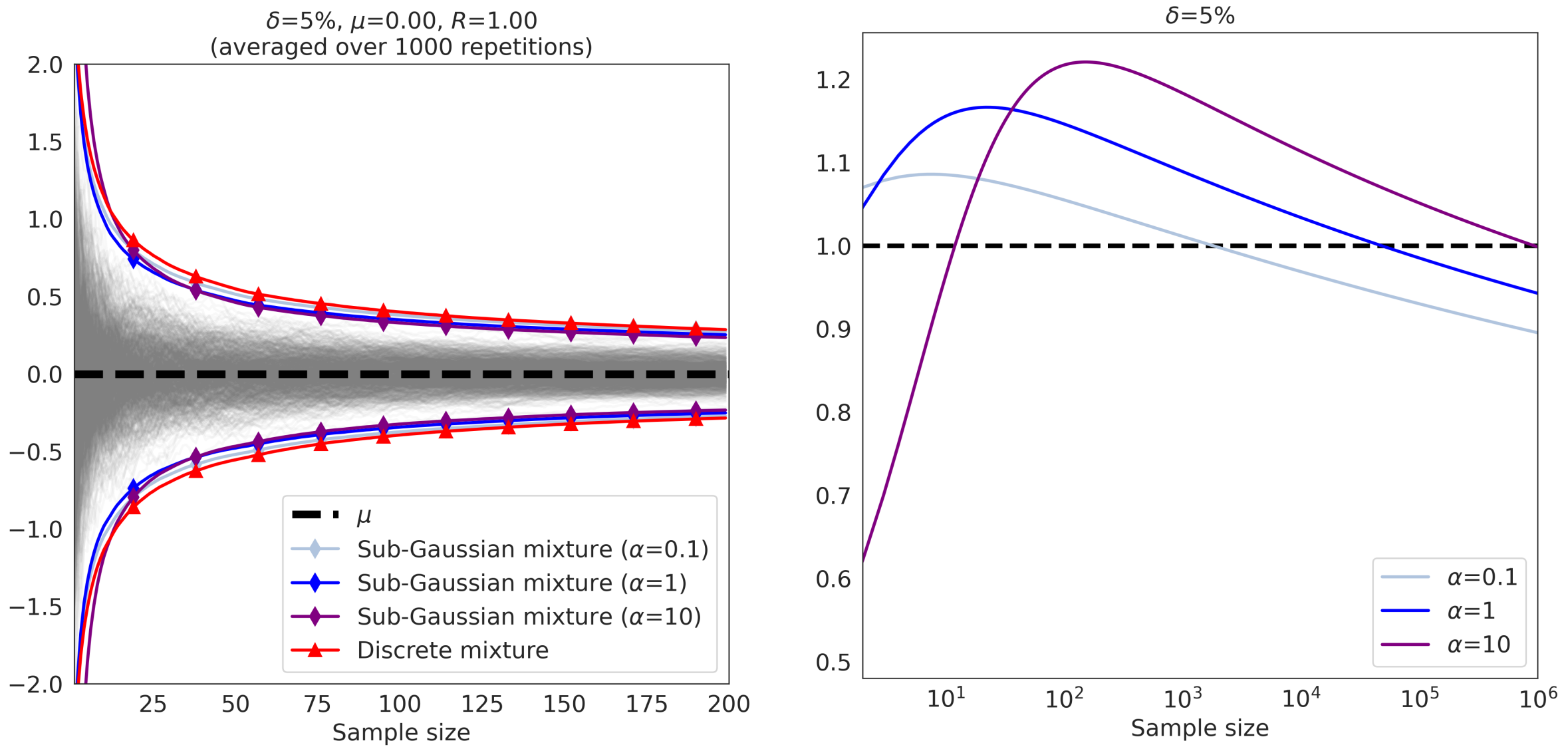

**Figure A.3** – Left: Comparison of confidence envelopes around the mean for $\mathcal{N}(0, 1)$, as a function of the sample size $t$, over $1000$ independent replicates. Grey lines are trajectories of empirical means $\widehat{\mu}_t$. Right: ratio of the upper discrete mixture bound to the various sub-Gaussian mixture bounds (less than $1$ means the discrete mixture is tighter).

# Appendix B

# Bregman deviations of generic exponential families

This chapter presents supplementary material for Chapter 3.

**Contents**

## B.1 Bregman concentration with global Legendre regularisation ()

In this section, we state an alternative result to Theorem 3.3 that extends to the case of sequence $(Y_t)_{t\in\mathbb{N}}$ of random variables that are not independent, each having possibly different distribution from others.

**Theorem B.1** (Method of mixtures with global Legendre regularisation, general case). *Let $(\mathcal{G}_t)_{t\in\mathbb{N}}$ be a filtration, $(F_t)_{t\in\mathbb{N}}$ and $(h_t)_{t\in\mathbb{N}}$ two sequences of functions such that for each $t \in \mathbb{N}$,*

*(i) $Y_t$ is $\mathcal{G}_t$-measurable,*

*(ii) $F_t$ and $h_t$ are $\mathcal{G}_{t-1}$-measurable,*

*(iii) given $\mathcal{G}_{t-1}$, $Y_t \sim p_{\theta^\star,t}$ where $p_{\theta^\star,t} \in \mathcal{G}$ belongs to an exponential family with parameter $\theta^\star$, feature function $F_t$ and base function $h_t$.*

*Let $\mathcal{L}_t$ be the log-partition function corresponding to $p_{\theta^\star,t}$. For any Legendre (i.e. strictly convex and continuously differentiable) function $\mathcal{L}_0 : \Theta \to \mathbb{R}$ such that $\exp(-\mathcal{L}_0)$ is integrable, we introduce the parameter estimate and Bregman information gain*

$$\widehat{\theta}_{t,\mathcal{L}_0} = \left(\sum_{s=0}^{t} \nabla \mathcal{L}_t\right)^{-1} \left(\sum_{s=1}^{t} F_s(Y_s)\right), \tag{B.1}$$

$$\gamma_{t,\mathcal{L}_0} = \log\left(\frac{\int_\Theta \exp\left(-\mathcal{L}_0(\theta')\right) d\theta'}{\int_\Theta \exp\left(-\sum_{s=0}^{t} \mathcal{B}_{\mathcal{L}_s}(\theta', \widehat{\theta}_{t,\mathcal{L}_0})\right) d\theta'}\right), \tag{B.2}$$

*and then, for any $\delta \in (0,1]$, consider the set*

$$\widehat{\Theta}_{t,\mathcal{L}_0}(\delta) = \left\{\theta_0 : \sum_{s=0}^{t} \mathcal{B}_{\mathcal{L}_s}\left(\theta_0, \widehat{\theta}_{t,\mathcal{L}_0}\right) - \mathcal{L}_0(\theta_0) \leqslant \log\frac{1}{\delta} + \gamma_{t,\mathcal{L}_0}\right\}. \tag{B.3}$$

*Then $\left(\widehat{\Theta}^\delta_{t,\mathcal{L}_0}\right)_{t\in\mathbb{N}}$ is a **time-uniform confidence sequence** at level $\delta$ for $\theta^\star$, i.e.*

$$\mathbb{P}_{\theta^\star}\left(\forall t \in \mathbb{N},\ \theta^\star \in \widehat{\Theta}^\delta_{t,\mathcal{L}_0}\right) \geqslant 1 - \delta\,, \tag{B.4}$$

*or equivalently, for any random time $\tau$ in $\mathbb{N}$,*

$$\mathbb{P}_{\theta^\star}\left(\theta^\star \in \widehat{\Theta}^\delta_{\tau,\mathcal{L}_0}\right) \geqslant 1 - \delta\,. \tag{B.5}$$

In contrast to Theorem 3.3 that involves a local regularisation using the true parameter $\theta^\star \in \Theta$ of the family and a constant $c > 0$, here we make use of a global regularisation, in the form of the Legendre function $\mathcal{L}_0$. Concretely, the regularised parameter estimate $\widehat{\theta}_{t,\mathcal{L}_0}$ and the Bregman information gain $\gamma_{t,\mathcal{L}_0}$ do not depend on the true parameter $\theta^\star$, thus making for a more explicit confidence set $\widehat{\Theta}_{t,\mathcal{L}_0}$. The tradeoff here is that the choice of regulariser is limited by the integrability assumption on $\exp(-\mathcal{L}_0)$, which is critical to build an appropriate prior for the method of mixtures.

*Proof of Theorem B.1.* The proof of this result follows a similar line of proof as we used for Theorem 3.3. However the use of $\mathcal{L}_0$ instead of $c$ induces a few changes that we detail below. In particular, we use a different prior to build the mixture of martingales.

**Martingale and mixture martingale construction.** For any $\lambda \in \mathbb{R}^d$ and $t \in \mathbb{N}$, we define

$$M_t^\lambda = \exp\left(\sum_{s=1}^{t}\left(\lambda^\top\left(F_s(Y_s) - \mathbb{E}_{\theta^\star}\left[F_s(Y_s)\right]\right) - B_{\mathcal{L}_s,\theta^\star}(\lambda)\right)\right), \tag{B.6}$$

where we introduced $B_{\mathcal{L}_s,\theta^\star}(\lambda) = \mathcal{L}_s(\theta^\star + \lambda) - \mathcal{L}_s(\theta^\star) - \langle \lambda, \nabla\mathcal{L}_s(\theta^\star)\rangle$ for convenience.

Note that $M_t^\lambda$ is nonnegative and

$$\mathbb{E}_{\theta^\star}\left[\exp\left(\lambda^\top F_s(Y_s)\right)\right] = \exp\left(\mathcal{L}_s(\theta^\star + \lambda) - \mathcal{L}_s(\theta^\star)\right), \tag{B.7}$$

$$\mathbb{E}_{\theta^\star}\left[\exp\left(\lambda^\top \mathbb{E}_{\theta^\star}\left[F_s(Y_s)\right]\right)\right] = \exp\left(\lambda^\top \mathcal{L}_s(\theta^\star)\right). \tag{B.8}$$

Note that $M_t^\lambda$ is $\mathcal{G}_t$-measurable and in fact $\mathbb{E}_{\theta^\star}[M_t^\lambda \mid \mathcal{G}_{t-1}] = M_{t-1}^\lambda$. Therefore $(M_t^\lambda)_{t\in\mathbb{N}}$ is a nonnegative martingale adapted to the filtration $(\mathcal{G}_t)_{t\in\mathbb{N}}$ and actually satisfies the equality $\mathbb{E}_{\theta^\star}\left[M_t^\lambda\right] = 1$. For any prior density $q(\theta)$ over $\theta \in \Theta$, we now define a mixture of martingales

$$M_t = \int_{\Lambda_{\theta^\star}} M_t^\lambda q(\theta + \lambda) d\lambda\,. \tag{B.9}$$

where $\Lambda_{\theta^\star} = \{\lambda,\ \theta^\star + \lambda \in \Theta\}$. Then $(M_t)_{t\in\mathbb{N}}$ is also a martingale and also satisfies the equality $\mathbb{E}_{\theta^\star}\left[M_t\right] = 1$. Now consider the prior density

$$q(\theta^\star + \lambda) = c_0 \exp\left(-\mathcal{L}_0(\theta^\star + \lambda)\right), \tag{B.10}$$

where $c_0 = \frac{1}{\int_\Theta \exp(-\mathcal{L}_0(\theta'))d\theta'}$ (which is well-defined since $\exp(-\mathcal{L}_0)$ is integrable over $\Theta$). We then have

$$M_t = c_0 \int_{\Lambda_{\theta^\star}} \exp\left(\lambda^\top S_t - \sum_{s=1}^{t} \mathcal{B}_{\mathcal{L}_s,\theta^\star}(\lambda) - \mathcal{L}_0(\theta^\star + \lambda)\right) d\lambda\,, \tag{B.11}$$

where we denote $S_t = \sum_{s=1}^{t}\left(F_s(Y_s) - \mathbb{E}_{\theta^\star}\left[F_s(Y_s)\right]\right)$. Now, from the formula of parameter estimate, we have

$$\sum_{s=1}^{t} \nabla\mathcal{L}_s(\widehat{\theta}_{t,\mathcal{L}_0}) + \nabla\mathcal{L}_0(\widehat{\theta}_{t,\mathcal{L}_0}) = \sum_{s=1}^{t} F_s(Y_s)\,. \tag{B.12}$$

This yields

$$S_t = \sum_{s=1}^{t} \Big(\nabla \mathcal{L}_s(\widehat{\theta}_{t,\mathcal{L}_0}) - \nabla \mathcal{L}_s(\theta)\Big) + \nabla \mathcal{L}_0(\widehat{\theta}_{t,\mathcal{L}_0})\,. \tag{B.13}$$

**Legendre function and Bregman properties.** We now introduce the regularisation function $\mathcal{L}_{1:t}(\theta^\star) = \sum_{s=1}^t \mathcal{L}_s(\theta^\star)$. Note that $\mathcal{L}_{1:t}$ is also a Legendre function and its associated Bregman divergence satisfies, for any $\theta, \theta' \in \Theta$,

$$\mathcal{B}_{\mathcal{L}_{1:t}}(\theta', \theta) = \sum_{s=1}^{t} \Big(\mathcal{L}_s(\theta') - \mathcal{L}_s(\theta) - (\theta' - \theta)^\top \nabla \mathcal{L}_s(\theta)\Big) = \sum_{s=1}^{t} \mathcal{B}_{\mathcal{L}_s}(\theta', \theta). \tag{B.14}$$

In this notation, we can rewrite $\sum_{s=1}^t \mathcal{B}_{\mathcal{L}_s,\theta^\star}(\lambda) = \mathcal{B}_{\mathcal{L}_{1:t},\theta^\star}(\lambda)$ and $S_t = \nabla \mathcal{L}_{1:t}(\widehat{\theta}_{t,\mathcal{L}_0}) - \nabla \mathcal{L}_t(\theta^\star) + \nabla \mathcal{L}_0(\widehat{\theta}_{t,\mathcal{L}_0})$. We then obtain

$$M_t = c_0 \exp\left(-\mathcal{L}_0(\theta^\star)\right) \int_{\Lambda_{\theta^\star}} \exp\Big(\lambda^\top x_t - \mathcal{B}_{\mathcal{L}_{1:t},\theta^\star}(\lambda) + \lambda^\top x_0 - \mathcal{B}_{\mathcal{L}_0,\theta^\star}(\lambda)\Big)\, d\lambda\,, \tag{B.15}$$

where we have introduced $x_0 = \nabla \mathcal{L}_0(\widehat{\theta}_{t,\mathcal{L}_0}) - \nabla \mathcal{L}_0(\theta^\star)$ and $x_t = \nabla \mathcal{L}_{1:n}(\widehat{\theta}_{t,\mathcal{L}_0}) - \nabla \mathcal{L}_{1:t}(\theta^\star)$.

We now have from the Bregman duality property recalled in Lemma 3.1 that

$$\sup_{\lambda \in \Lambda_{\theta^\star}} \Big(\lambda^\top x_t - \mathcal{B}_{\mathcal{L}_{1:t},\theta^\star}(\lambda)\Big) = \mathcal{B}^\star_{\mathcal{L}_{1:t},\theta^\star}(x_t) \tag{B.16}$$

$$= \mathcal{B}^\star_{\mathcal{L}_{1:t},\theta^\star}(\nabla \mathcal{L}_{1:t}(\widehat{\theta}_{t,\mathcal{L}_0}) - \nabla \mathcal{L}_{1:t}(\theta^\star)) = \mathcal{B}_{\mathcal{L}_{1:t}}(\theta^\star, \widehat{\theta}_{t,\mathcal{L}_0})\,. \tag{B.17}$$

Furthermore, any optimal $\lambda$ must satisfy

$$\nabla \mathcal{L}_{1:t}(\theta^\star + \lambda) - \nabla \mathcal{L}_{1:t}(\theta^\star) = x_t \implies \nabla \mathcal{L}_{1:t}(\theta^\star + \lambda) = \nabla \mathcal{L}_{1:t}(\widehat{\theta}_{t,\mathcal{L}_0})\,. \tag{B.18}$$

One possible solution is $\lambda^\star = \widehat{\theta}_{t,\mathcal{L}_0} - \theta^\star$, in which case we deduce that

$$\begin{aligned}
&\lambda^\top x_t - \mathcal{B}_{\mathcal{L}_{1:t},\theta^\star}(\lambda)\\
&= \lambda^\top x_t - \mathcal{B}_{\mathcal{L}_{1:t},\theta^\star}(\lambda) + \mathcal{B}_{\mathcal{L}_{1:t}}(\theta^\star, \widehat{\theta}_{t,\mathcal{L}_0}) - \Big(\lambda^{\star\top} x_t - \mathcal{B}_{\mathcal{L}_{1:t},\theta^\star}(\lambda^\star)\Big)\\
&= \mathcal{B}_{\mathcal{L}_{1:t}}(\theta^\star, \widehat{\theta}_{t,\mathcal{L}_0}) + (\lambda - \lambda^\star)^\top \nabla \mathcal{L}_t(\theta^\star + \lambda^\star) + \mathcal{B}_{\mathcal{L}_{1:t},\theta^\star}(\lambda^\star) - \mathcal{B}_{\mathcal{L}_{1:t},\theta^\star}(\lambda) - (\lambda - \lambda^\star)^\top \nabla \mathcal{L}_{1:t}(\theta^\star)\\
&= \mathcal{B}_{\mathcal{L}_{1:t}}(\theta^\star, \widehat{\theta}_{t,\mathcal{L}_0}) + (\lambda - \lambda^\star)^\top \nabla \mathcal{L}_{1:t}(\theta^\star + \lambda^\star) + \mathcal{L}_{1:t}(\theta^\star + \lambda^\star) - \mathcal{L}_{1:t}(\theta^\star + \lambda)\,.
\end{aligned} \tag{B.19}$$

Similarly, we have

$$\lambda^\top x_0 - \mathcal{B}_{\mathcal{L}_0,\theta^\star}(\lambda) = \mathcal{B}_{\mathcal{L}_0}(\theta^\star, \widehat{\theta}_{t,\mathcal{L}_0}) + (\lambda - \lambda^\star)^\top \nabla \mathcal{L}_0(\theta^\star + \lambda^\star) + \mathcal{L}_0(\theta^\star + \lambda^\star) - \mathcal{L}_0(\theta^\star + \lambda)\,. \tag{B.20}$$

**Martingale rewriting and conclusion.** Plugging in (B.20) and (B.19) in (B.15), we now obtain

$$\begin{aligned}
M_t &= c_0 \exp\left(\sum_{s\in\{0,1:t\}} \mathcal{B}_{\mathcal{L}_s}(\theta^\star, \widehat{\theta}_{t,\mathcal{L}_0}) - \mathcal{L}_0(\theta^\star)\right) \\
&\quad \times \int_{\Lambda_{\theta^\star}} \exp\left(\sum_{s\in\{0,1:t\}} \Big((\lambda-\lambda^\star)^\top \nabla\mathcal{L}_s(\theta^\star+\lambda^\star) + \mathcal{L}_s(\theta^\star+\lambda^\star) - \mathcal{L}_s(\theta^\star+\lambda)\Big)\right) d\lambda \\
&= c_0 \exp\left(\sum_{s\in\{0,1:t\}} \mathcal{B}_{\mathcal{L}_s}(\theta^\star, \widehat{\theta}_{t,\mathcal{L}_0}) - \mathcal{L}_0(\theta^\star)\right) \\
&\quad \times \exp\left(-\sum_{s\in\{0,1:t\}} \Big((\theta^\star+\lambda^\star)^\top \nabla\mathcal{L}_s(\theta^\star+\lambda^\star) - \mathcal{L}_s(\theta^\star+\lambda^\star)\Big)\right) \\
&\quad \times \int_{\Lambda_{\theta^\star}} \exp\left(\sum_{s\in\{0,1:t\}} \Big((\theta^\star+\lambda)^\top \nabla\mathcal{L}_s(\theta^\star+\lambda^\star) - \mathcal{L}_s(\theta^\star+\lambda)\Big)\right) d\lambda \\
&= \frac{c_0}{c_t} \exp\left(\sum_{s\in\{0,1:t\}} \mathcal{B}_{\mathcal{L}_s}(\theta^\star, \widehat{\theta}_{t,\mathcal{L}_0}) - \mathcal{L}_0(\theta^\star)\right) \\
&\quad \times \frac{\int_{\Lambda_{\theta^\star}} \exp\left(\sum_{s\in\{0,1:t\}} \Big((\theta^\star+\lambda)^\top \nabla\mathcal{L}_s(\theta^\star+\lambda^\star) - \mathcal{L}_s(\theta^\star+\lambda)\Big)\right) d\lambda}{\int_\Theta \exp\left(\sum_{s\in\{0,1:t\}} \big((\theta')^\top \nabla\mathcal{L}_s(\theta^\star+\lambda^\star) - \mathcal{L}_s(\theta')\big)\right) d\theta'} \\
&= \frac{c_0}{c_t} \exp\left(\sum_{s=1}^t \mathcal{B}_{\mathcal{L}_s}(\theta^\star, \widehat{\theta}_{t,\mathcal{L}_0}) + \mathcal{B}_{\mathcal{L}_0}(\theta^\star, \widehat{\theta}_{t,\mathcal{L}_0}) - \mathcal{L}_0(\theta^\star)\right), \qquad \text{(B.21)}
\end{aligned}$$

where we have introduced

$$c_t = \frac{\exp\left(\sum_{s\in\{0,1:t\}} \Big((\theta^\star+\lambda^\star)^\top \nabla\mathcal{L}_s(\theta^\star+\lambda^\star) - \mathcal{L}_s(\theta^\star+\lambda^\star)\Big)\right)}{\int_\Theta \exp\left(\sum_{s\in\{0,1:t\}} \big((\theta')^\top \nabla\mathcal{L}_s(\theta^\star+\lambda^\star) - \mathcal{L}_s(\theta')\big)\right) d\theta'}. \qquad \text{(B.22)}$$

A simple application of Theorem 1.20 yields, for any random time $\tau$ in $\mathbb{N}$,

$$\begin{aligned}
\mathbb{P}_{\theta^\star}\left(\sum_{s=1}^\tau \mathcal{B}_{\mathcal{L}_s}(\theta^\star, \widehat{\theta}_{\tau,\mathcal{L}_0}) + \mathcal{B}_{\mathcal{L}_0}(\theta^\star, \widehat{\theta}_{\tau,\mathcal{L}_0}) - \mathcal{L}_0(\theta^\star) \geqslant \log\left(\frac{c_\tau}{c_0\delta}\right)\right) &= \mathbb{P}_{\theta^\star}\left(M_\tau \geqslant \frac{1}{\delta}\right) \\
&\leqslant \delta \mathbb{E}_{\theta^\star}[M_\tau] \\
&= \delta\,. \qquad \text{(B.23)}
\end{aligned}$$

Finally, since $\lambda^\star = \widehat{\theta}_{\tau,\mathcal{L}_0} - \theta^\star$, we have the more explicit form

$$\frac{c_\tau}{c_0} = \frac{\int_\Theta \exp\left(-\mathcal{L}_0(\theta')\right) d\theta'}{\int_\Theta \exp\left(-\sum_{s\in\{0,1:\tau\}} \mathcal{B}_{\mathcal{L}_s}(\theta', \theta^\star+\lambda^\star)\right) d\theta'} \qquad \text{(B.24)}$$

$$= \frac{\int_\Theta \exp\left(-\mathcal{L}_0(\theta')\right) d\theta'}{\int_\Theta \exp\left(-\sum_{s=1}^{\tau} \mathcal{B}_{\mathcal{L}_s}(\theta', \widehat{\theta}_{\tau,\mathcal{L}_0}) - \mathcal{B}_{\mathcal{L}_0}(\theta', \widehat{\theta}_{\tau,\mathcal{L}_0})\right) d\theta'} \,. \tag{B.25}$$

We conclude on the time-uniform confidence sequence as in Theorem 3.3. ■

The general result of Theorem B.1 specifies straightforwardly to the case when all observations have same law, yielding the following corollary that we state now for completeness.

**Corollary B.2** (Method of mixtures with global Legendre regularisation, i.i.d. case). *Let $t \in \mathbb{N}$ and $\widehat{\mu}_t = 1/t \sum_{s=1}^{t} F(Y_s)$. For all Legendre function $\mathcal{L}_0$ such that $\exp(-\mathcal{L}_0)$ is integrable over $\Theta$, let*

$$\widehat{\theta}_{t,\mathcal{L}_0} = \left(\nabla\mathcal{L} + \frac{1}{t}\nabla\mathcal{L}_0\right)^{-1} (\widehat{\mu}_t)\,, \tag{B.26}$$

$$\gamma_{t,\mathcal{L}_0} = \log\left(\frac{\int_\Theta \exp\left(-\mathcal{L}_0(\theta')\right) d\theta'}{\int_\Theta \exp\left(-t\mathcal{B}_{\mathcal{L}}\left(\theta', \widehat{\theta}_{t,\mathcal{L}_0}\right) - \mathcal{B}_{\mathcal{L}_0}\left(\theta', \widehat{\theta}_{t,\mathcal{L}_0}\right)\right) d\theta'}\right)\,, \tag{B.27}$$

*and then, for any $\delta \in (0,1]$, consider the set*

$$\widehat{\Theta}^{\delta}_{t,\mathcal{L}_0} = \left\{\theta_0 : t\mathcal{B}_{\mathcal{L}}\left(\theta_0, \widehat{\theta}_{t,\mathcal{L}_0}\right) + \mathcal{B}_{\mathcal{L}_0}\left(\theta_0, \widehat{\theta}_{t,\mathcal{L}_0}\right) \leqslant \log\frac{1}{\delta} + \mathcal{L}_0(\theta_0) + \gamma_{t,\mathcal{L}_0}\right\}\,. \tag{B.28}$$

*Then $\left(\widehat{\Theta}^{\delta}_{t,\mathcal{L}_0}\right)_{t\in\mathbb{N}}$ is a* ***time-uniform confidence sequence*** *at level $\delta$ for $\theta^\star$.*

**Choice of Legendre function $\mathcal{L}_0$.** A natural choice for the Legendre regulariser $\mathcal{L}_0$ is to use the log-partition function of the exponential family at hand. If $\mathcal{L}_0 = c\mathcal{L}$, where $c > 0$ is a scaling coefficient, the formula above simplify to

$$\widehat{\theta}_{t,\mathcal{L}_0} = \frac{\widehat{\theta}_t}{1 + \frac{c}{t}} \text{ and } \gamma_{t,\mathcal{L}_0} = \log\left(\frac{\int_\Theta \exp\left(-c\mathcal{L}(\theta')\right) d\theta'}{\int_\Theta \exp\left(-(t+c)\mathcal{B}_{\mathcal{L}}\left(\theta', \widehat{\theta}_{t,\mathcal{L}_0}\right)\right) d\theta'}\right)\,, \tag{B.29}$$

where $\widehat{\theta}_t = (\nabla\mathcal{L})^{-1}(\widehat{\mu}_t)$ is the standard MLE of the parameter $\theta_0$.

In the special case of the one-dimensional Gaussian family $\mathcal{G} = \{\mathcal{N}(\theta, \sigma^2), \theta \in \mathbb{R}\}$ where the variance $\sigma^2$ is known, the log-partition function regulariser defined by $\mathcal{L}_0(\theta) = \frac{\theta^2}{2\sigma^2}$ satisfies the integrability assumption $\int_{\mathbb{R}} e^{-c\mathcal{L}_0(\theta)} d\theta < \infty$ with $c > 0$. Straightforward calculations show that the resulting confidence set is the same as the one derived from Theorem 3.3 with local regularisation $c$ (see Appendix B.2). For many other standard families however, this integrabil-ity assumption may fail, as in the case of the exponential distributions $\mathcal{G} = \{\mathcal{E}(-\theta), \theta \in \mathbb{R}^*_+\}$, for which $\mathcal{L}(\theta) = -\log(-\theta)$ (see Appendix B.2).

For other choices of Legendre function, computing the regularised parameter estimate $\widehat{\theta}_{t,\mathcal{L}_0}$ and the information gain $\gamma_{t,\mathcal{L}_0}$ requires inverting the function $\nabla\mathcal{L} + \frac{1}{t}\nabla\mathcal{L}_0$ and computing a tedious integral, both of which seldom result in closed-form expressions. For these reasons, we recommend as a first step the use of local regularisation (Theorem 3.3) over Legendre regularisers.

**Application to linear contextual bandits.** We now detail an application of Theorem B.1 to build vector confidence sets for linear bandit using a quadratic Legendre regulariser. We consider the setting of linear bandits (see Chapter 1), but with possibly action-dependent variance (heteroscedastic noise). We recall that an algorithm for this problem chooses, at each round $t \in \mathbb{N}$, an action $X_t \in \mathcal{X}_t$, and subsequently observes a random reward $Y_t = \langle \theta^\star, \phi(X_t) \rangle + \varepsilon_t$, where $\phi \colon \mathcal{X} \to \mathbb{R}^d$ is a (fixed) feature map, and $\theta^\star \in \mathbb{R}^d$ is a vector of weights unknown to the algorithm. The noise $\varepsilon_t \sim \mathcal{N}(0, \sigma^2_{X_t} I)$ is assumed to be Gaussian with known, action-dependent variance $\sigma^2_{X_t}$. The goal is to select such actions to maximise the expected cumulative reward $\mathbb{E}[\sum_{t=1}^T \langle \theta^\star, \phi(X_t)\rangle]$ up to a (possibly known beforehand) horizon $T$. An optimistic policy $\mathbb{X} = (X_t)_{t\in\mathbb{N}}$ for this problem selects actions sequentially depending on the available history of past actions and rewards. Naturally, for each $t \in \mathbb{N}$, this amounts to controlling the deviation between the unknown parameter $\theta^\star$ and a suitable estimate $\widehat{\theta}_t$ built from $t$ observations $(X_s, Y_s)_{s \leqslant t}$.

Under this statistical model, the feature and log-partition functions are given respectively by $F_t(x) = x\phi(X_t)/\sigma^2_{X_t}$ and $\mathcal{L}_t(\theta) = 1/(2\sigma^2_{X_t}) \left\|\theta\right\|^2_{\phi(X_t)\phi(X_t)^\top}$. Let us define the (Hessian) matrix

$$V_t = \sum_{s=1}^{t-1} \frac{1}{\sigma^2_{X_t}} \phi(X_t)\phi(X_t)^\top \,, \tag{B.30}$$

and introduce the Legendre function $\mathcal{L}_0(\theta) = 1/2 \left\|\theta\right\|^2_{V_0}$, where $V_0$ is some fixed positive definite matrix. Then, the parameter estimate and the Bregman information gain take the form

$$\widehat{\theta}_{t,\mathcal{L}_0} = (V_t + V_0)^{-1} \sum_{s=1}^{t-1} \frac{Y_s}{\sigma^2_{X_s}} \phi(X_s) \,, \quad \gamma_{t,\mathcal{L}_0} = \log \frac{\det(V_0 + V_t)^{1/2}}{\det(V_0)^{1/2}} \,. \tag{B.31}$$

Note that, the notion of Bregman information gain coincides with the classical information gain for (sub-Gaussian) linear bandits (Srinivas et al., 2010; Russo and Van Roy, 2016). Then the confidence sequence $(\widehat{\Theta}^\delta_{t,\mathcal{L}_0})_{t\in\mathbb{N}}$ of Theorem B.1 associated with the adapted bandit filtration is

$$\left(\widehat{\Theta}^\delta_{t,\mathcal{L}_0}\right)_{t\in\mathbb{N}} = \left(\left\{\theta_0 \in \Theta,\ \left\|\theta - \theta_{t,\mathcal{L}_0}\right\|^2_{V_t+V_0} \leqslant \left\|\theta\right\|^2_{V_0} + 2\log \frac{\det(V_0 + V_t)^{1/2}}{\delta \det(V_0)^{1/2}} \right\}\right)_{t\in\mathbb{N}} \,. \tag{B.32}$$

Under the assumption that the noise process $(\varepsilon_t)_{t\in\mathbb{N}}$ is conditionally $\sigma_{X_t}$-sub-Gaussian, Abbasi-Yadkori et al. (2011) obtained a high probability confidence sequence defined as

$$\left(\widetilde{\Theta}^{\delta}_{t,\mathcal{L}_0}\right)_{t\in\mathbb{N}} = \left(\left\{\theta_0 \in \Theta,\ \|\theta - \theta_{t,\mathcal{L}_0}\|_{V_t+V_0} \leqslant \|\theta\|_{V_0} + \sqrt{2\log\frac{\det(V_0+V_t)^{1/2}}{\delta\det(V_0)^{1/2}}}\right\}\right)_{t\in\mathbb{N}} . \tag{B.33}$$

Interestingly, since $\sqrt{a+b} \leqslant \sqrt{a} + \sqrt{b}$ for all $a, b \in \mathbb{R}^{\star}_{+}$, it is clear that $\Theta_{t,\mathcal{L}_0}(\delta) \subset \widetilde{\Theta}_{t,\mathcal{L}_0}(\delta)$ for all $t \in \mathbb{N}$, yielding a slight improvement over the standard multivariate Gaussian mixture bounds.

## B.2 Specification to illustrative exponential families

In this section, we provide the technical derivations to specify our generic concentration result to a few classical distributions. The results are summarised in Table 3.1. In what follows, $(Y_t)_{t\in\mathbb{N}}$ denotes an i.i.d. sequence drawn from the exponential family at hand with a fixed parameter $\theta^\star \in \Theta$, and $t \in \mathbb{N}$ a fixed integer.

### Gaussian with unknown mean, known variance

Consider $Y \sim \mathcal{N}(\mu, \sigma^2)$, where $\mu$ is unknown and $\sigma$ is known. This corresponds to an exponential family of distributions, with parameter $\theta^\star = \mu$, feature function $F(y) = \frac{y}{\sigma^2}$ and log-partition function $\mathcal{L}(\theta) = \frac{\theta^2}{2\sigma^2}$. The Bregman divergence between two parameters $\theta'$ and $\theta$ associated with $\mathcal{L}$ is given by $\mathcal{B}_{\mathcal{L}}(\theta', \theta) = \mathrm{KL}(p_\mu \parallel p_{\mu'}) = \frac{(\mu-\mu')^2}{2\sigma^2}$. We further have $\mathcal{L}'(\theta) = \frac{\theta}{\sigma^2} = \frac{\mu}{\sigma^2}$ and $\mathcal{L}''(\theta) = \frac{1}{\sigma^2}$, implying that $\mathcal{L}'$ is invertible. Denoting $S_t = \sum_{s=1}^{t} Y_s$, we obtain that

$$\widehat{\mu}_{t,c}(\mu) = \widehat{\theta}_{t,c}(\theta) = \frac{S_t + c\mu}{t+c} . \tag{B.34}$$

The Bregman deviations simplify as follows

$$(t+c)\mathcal{B}_{\mathcal{L}}(\theta, \widehat{\theta}_{t,c}(\theta)) = (t+c)\frac{(\widehat{\theta}_{t,c}(\theta) - \theta)^2}{2\sigma^2} = \frac{1}{t+c}\frac{(S_t - t\mu)^2}{2\sigma^2} . \tag{B.35}$$

Now, on the other hand, let us see that the information gain is explicitly given by

$$\gamma_{t,c}(\mu) = \frac{1}{2}\log\frac{t+c}{c} . \tag{B.36}$$

We obtain from Theorem 3.3 with probability at least $1-\delta$, for all $t \in \mathbb{N}$,

$$\frac{1}{t+c}\frac{(S_t - t\mu)^2}{2\sigma^2} \leqslant \log\frac{1}{\delta} + \frac{1}{2}\log\frac{t+c}{c} . \tag{B.37}$$

### Gaussian with known mean, unknown variance

We consider $Y \sim \mathcal{N}(\mu, \sigma^2)$, where $\mu$ is known and $\sigma$ is unknown. This corresponds to a one-dimensional exponential family model with parameter $\theta^\star = -\frac{1}{2\sigma^2} \in \mathbb{R}_-^*$, feature function $F(y) = (y-\mu)^2$ and log-partition function $\mathcal{L}(\theta) = -\frac{1}{2}\log(-2\theta)$. The first two derivatives of $\mathcal{L}$ are given by $\mathcal{L}'(\theta) = -\frac{1}{2\theta}$ and $\mathcal{L}''(\theta) = \frac{1}{2\theta^2}$, which shows that $\mathcal{L}'$ is invertible on the domain $\Theta = \mathbb{R}_-^*$. The Bregman divergence between two parameters $\theta'$ and $\theta$ is therefore $\mathcal{B}_{\mathcal{L}}(\theta', \theta) = \frac{1}{2}\log(\theta/\theta') + \frac{1}{2}(\theta'/\theta - 1)$.

Let $Q_t = \sum_{s=1}^t (Y_s - \mu)^2$. A short calculation shows that the following expression holds:

$$\widehat{\theta}_{t,c}(\theta) = -\frac{t+c}{c/\theta - 2Q_t}\,. \tag{B.38}$$

To compute the Bregman information gain, note that the above expression and a change of variable implies

$$\int_{-\infty}^{0} \exp\left(-c\mathcal{B}_{\mathcal{L}}\left(\theta', \theta\right)\right) d\theta' = -\theta\,(2/c)^{c/2+1}\, e^{c/2}\Gamma\left(c/2+2\right)\,. \tag{B.39}$$

Combining the last two lines yields

$$\begin{aligned}
\gamma_{t,c}(\theta) &= \log\frac{\theta}{\widehat{\theta}_{t,c}(\theta)} - (1+\log 2)\frac{t}{2} - (\frac{c}{2}+1)\log c + (\frac{t+c}{2}+1)\log(t+c) - \frac{\log\Gamma\left(\frac{t+c}{2}+2\right)}{\log\Gamma\left(\frac{c}{2}+2\right)} && \text{(B.40)}\\
&= \log\frac{c-2\theta Q_t}{t+c} - (1+\log 2)\frac{t}{2} - (\frac{c}{2}+1)\log c + (\frac{t+c}{2}+1)\log(t+c) - \frac{\log\Gamma\left(\frac{t+c}{2}+2\right)}{\log\Gamma\left(\frac{c}{2}+2\right)} && \text{(B.41)}\\
&= \log\frac{Q_t/\sigma^2+c}{t+c} - (1+\log 2)\frac{t}{2} - (\frac{c}{2}+1)\log c + (\frac{t+c}{2}+1)\log(t+c) - \frac{\log\Gamma\left(\frac{t+c}{2}+2\right)}{\log\Gamma\left(\frac{c}{2}+2\right)}\,. && \text{(B.42)}
\end{aligned}$$

Moreover, we deduce from the expression of $\widehat{\theta}_{t,c}(\theta)$ and the Bregman divergence that:

$$\mathcal{B}_{\mathcal{L}}(\theta', \widehat{\theta}_{t,c}(\theta)) = -\frac{1}{2}\log\left(\frac{c\theta'/\theta - 2\theta' Q_t}{t+c}\right) + \frac{1}{2}\left(\frac{c\theta'/\theta - 2\theta' Q_t}{t+c} - 1\right)\,. \tag{B.43}$$

So, we have

$$(t+c)\mathcal{B}_{\mathcal{L}}(\theta, \widehat{\theta}_{t,c}(\theta)) = -\frac{t+c}{2}\log\left(\frac{Q_t/\sigma^2+c}{t+c}\right) + \frac{Q_t/\sigma^2+c}{2} - \frac{t}{2}\,. \tag{B.44}$$

After some simple algebra and using the natural parametrisation $\theta = -\frac{1}{2\sigma^2}$, we derive from Theorem 3.3 that with probability at least $\geqslant 1-\delta$, for all $t \in \mathbb{N}$,

$$
\begin{aligned}
&-(\frac{t+c}{2}+1)\log\frac{Q_t/\sigma^2+c}{t+c}+\frac{Q_t}{2\sigma^2}\\
&\leqslant \log\frac{1}{\delta}-\frac{t}{2}\log 2-(\frac{c}{2}+1)\log c+(\frac{t+c}{2}+1)\log(t+c)+\log\Gamma\left(\frac{c}{2}+2\right)-\log\Gamma\left(\frac{t+c}{2}+2\right).
\end{aligned}
\tag{B.45}
$$

### Gaussian with unknown mean and variance

We consider $Y \sim \mathcal{N}(\mu, \sigma^2)$, where both $\mu \in \mathbb{R}$ and $\sigma > 0$ are unknown. These distributions form a two-dimensional exponential family with parameter $\theta^\star = \begin{pmatrix}\theta_1^\star \\ \theta_2^\star\end{pmatrix} = \begin{pmatrix}\frac{\mu}{\sigma^2} \\ -\frac{1}{2\sigma^2}\end{pmatrix}$ belonging to the domain $\Theta = \mathbb{R} \times \mathbb{R}_-^*$. The corresponding feature and log-partition functions writes $F(y) = \begin{pmatrix} y \\ y^2\end{pmatrix}$ and $\mathcal{L}(\theta) = -\frac{1}{2}\log(-\theta_2) - \frac{\theta_1^2}{4\theta_2}$ respectively. The gradient and Hessian of $\mathcal{L}$ follows from straightforward calculations and write:

$$
\nabla\mathcal{L}(\theta) = \begin{pmatrix} -\frac{\theta_1}{2\theta_2} \\ -\frac{1}{2\theta_2}+\frac{\theta_1^2}{4\theta_2^2}\end{pmatrix}, \nabla^2\mathcal{L}(\theta) = \begin{pmatrix} -\frac{1}{2\theta_2} & \frac{\theta_1}{2\theta_2^2} \\ \frac{\theta_1}{2\theta_2^2} & \frac{1}{2\theta_2^2}-\frac{\theta_1^2}{2\theta_2^3}\end{pmatrix}. \tag{B.46}
$$

In particular, $\mathrm{Tr}\ \nabla^2\mathcal{L}(\theta) = -\frac{1}{2\theta_2}+\frac{1}{2\theta_2^2}-\frac{\theta_1^2}{2\theta_2^3} > 0$ and $\det \nabla^2\mathcal{L}(\theta) = -\frac{1}{4\theta_2^3} > 0$, therefore the symmetric matrix $\nabla^2\mathcal{L}(\theta)$ is positive definite (both its eigenvalues are positive). The inverse of the gradient mapping $\nabla\mathcal{L}\colon \Theta \to \mathcal{P}$, where the image set is $\mathcal{P} = \{(x,y) \in \mathbb{R}^2, y > x^2\}$, can be explicitly computed as:

$$
(\nabla)^{-1}\mathcal{L}\begin{pmatrix} x \\ y\end{pmatrix} = \frac{1}{y-x^2}\begin{pmatrix} x \\ -\frac{1}{2}\end{pmatrix}. \tag{B.47}
$$

The expression of the Bregman divergence between two parameters $\theta$ and $\theta'$ can be calculated from the above results and reads:

$$
\mathcal{B}_{\mathcal{L}}(\theta',\theta) = \frac{1}{2}\log\frac{\theta_2}{\theta_2'}+\frac{\theta_2'}{2\theta_2}-\theta_2'\left(\frac{\theta_1'}{2\theta_2'}-\frac{\theta_1}{2\theta_2}\right)^2-\frac{1}{2}. \tag{B.48}
$$

Let $S_t = \sum_{s=1}^t Y_s$ and $Q_t = \sum_{s=1}^t Y_s^2$ and $V_{t,c} = \frac{1}{t+c}Q_t - \left(\frac{1}{t+c}S_t\right)^2$. We have that

$$
\widehat{\theta}_{t,c}(\theta) = \frac{1}{V_{t,c}-\frac{c}{t+c}\frac{1}{2\theta_2}+\frac{tc}{(t+c)^2}\frac{\theta_1^2}{4\theta_2^2}+\frac{c}{(t+c)^2}\frac{\theta_1}{\theta_2}S_t}\begin{pmatrix}\frac{1}{t+c}S_t-\frac{c}{2(t+c)}\frac{\theta_1}{\theta_2} \\ -\frac{1}{2}\end{pmatrix}. \tag{B.49}
$$

To compute the Bregman information gain, we need to evaluate the integral $\int_\Theta \exp\left(-c\mathcal{B}_\mathcal{L}(\theta',\theta)\right) d\theta'$. As the integrand is nonnegative, we can integrate first along $\theta'_1$ (Fubini's theorem), which writes:

$$\begin{aligned}\int_\Theta \exp\left(-c\mathcal{B}_\mathcal{L}(\theta',\theta)\right) d\theta' &= \int_{-\infty}^{0}\int_{-\infty}^{\infty} \exp\left(-\frac{c}{2}\log\left(\frac{\theta_2}{\theta'_2}\right) - \frac{c}{2}\frac{\theta'_2}{\theta_2} + c\theta'_2\left(\frac{\theta'_1}{2\theta'_2} - \frac{\theta_1}{2\theta_2}\right)^2 + \frac{c}{2}\right) d\theta'_1 d\theta'_2 \\ &= e^{\frac{c}{2}}\int_{-\infty}^{0}\left(\frac{\theta'_2}{\theta_2}\right)^{\frac{c}{2}} e^{-\frac{c}{2}\frac{\theta'_2}{\theta_2}}\int_{-\infty}^{\infty}\exp\left(c\theta'_2\left(\frac{\theta'_1}{2\theta'_2} - \frac{\theta_1}{2\theta_2}\right)^2\right) d\theta'_1 d\theta'_2\,.\end{aligned} \tag{B.50}$$

Let $m = \frac{\theta_1}{2\theta_2}$ and $s = \frac{1}{\sqrt{-2c\theta'_2}}$. The integral with respect to $\theta'_1$ can be rewritten using the change of variable $y = \frac{\theta'_1}{2\theta'_2}$ as:

$$\int_{-\infty}^{\infty}\exp\left(c\theta'_2\left(\frac{\theta'_1}{2\theta'_2} - \frac{\theta_1}{2\theta_2}\right)^2\right) d\theta'_1 = -2\theta'_2\sqrt{2\pi}s\underbrace{\int_{-\infty}^{\infty}\frac{1}{\sqrt{2\pi}s}e^{-\frac{(y-m)^2}{2s^2}}dy}_{=1}\,. \tag{B.51}$$

Therefore, the above calculation simplifies to:

$$\int_\Theta \exp\left(-c\mathcal{B}_\mathcal{L}(\theta',\theta)\right) d\theta' = 2\sqrt{\frac{\pi}{c}}e^{\frac{c}{2}}\int_{-\infty}^{0}\left(\frac{\theta'_2}{\theta_2}\right)^{\frac{c}{2}} e^{-\frac{c}{2}\frac{\theta'_2}{\theta_2}}\sqrt{-\theta'_2}d\theta'_2\,, \tag{B.52}$$

which after a linear change of variable on $\theta'_2$ can be related to the Gamma function as follows:

$$\int_\Theta \exp\left(-c\mathcal{B}_\mathcal{L}(\theta',\theta)\right) d\theta' = 2\sqrt{\frac{\pi}{c}}\left(\frac{2}{c}\right)^{\frac{c+1}{2}+1}\Gamma\left(\frac{c+1}{2}+1\right)(-\theta_2)^{\frac{3}{2}}\,. \tag{B.53}$$

The Bregman information gain is thus:

$$\begin{aligned}\gamma_{t,c}(\theta) = &-\left(\frac{1+\log 2}{2}\right)t - \left(\frac{c}{2}+2\right)\log c + \left(\frac{t+c}{2}+2\right)\log(t+c) \\ &+ \log\Gamma\left(\frac{c+3}{2}\right) - \log\Gamma\left(\frac{t+c+3}{2}\right) + \frac{3}{2}\log\frac{\theta_2}{\widehat{\theta}_{t,c}(\theta)_2}\,,\end{aligned} \tag{B.54}$$

with $\frac{\theta_2}{\widehat{\theta}_{t,c}(\theta)_2} = -2\theta_2 V_{t,c} + \frac{c}{t+c} - \frac{tc}{(t+c)^2}\frac{\theta_1^2}{2\theta_2} - \frac{2c\theta_1}{(t+c)^2}S_t t$. Now, applying the result of Theorem 3.3 shows that with probability at least $1-\delta$, for all $t \in \mathbb{N}$,

$$\begin{aligned}&-\frac{t+c+3}{2}\log\left(\frac{\theta_2}{\widehat{\theta}_{t,c}(\theta)_2}\right) - \theta_2(t+c)V_{t,c} - \frac{t\theta_1^2}{4\theta_2} - \frac{\theta_2}{t+c}S_t^2 - \theta_1 S_t \\ &\leqslant \log\frac{1}{\delta} - \frac{t}{2}\log 2 - \left(\frac{c}{2}+2\right)\log c + \left(\frac{t+c}{2}+2\right)\log(t+c) + \log\Gamma\left(\frac{c+3}{2}\right) - \log\Gamma\left(\frac{t+c+3}{2}\right)\,.\end{aligned} \tag{B.55}$$

After expanding the ratio $\frac{\theta_2}{\widehat{\theta}_{t,c}(\theta)_2}$ and substituting the natural parametrisation in terms of $\mu$ and $\sigma$, we finally obtain that with probability at least $1-\delta$, for all $t \in \mathbb{N}$,

$$
\begin{aligned}
&-\frac{t+c+3}{2}\log\left(\frac{1}{\sigma^2}V_{t,c}+\frac{c}{t+c}+\frac{tc}{(t+c)^2}\frac{\mu^2}{\sigma^2}-\frac{2c}{(t+c)^2}\frac{\mu}{\sigma^2}S_t\right)\\
&+\frac{t+c}{2\sigma^2}V_{t,c}+\frac{t\mu^2}{2\sigma^2}+\frac{1}{2(t+c)\sigma^2}S_t^2-\frac{\mu}{\sigma^2}S_t\\
&\leqslant \log\frac{1}{\delta}-\frac{t}{2}\log 2-\left(\frac{c}{2}+2\right)\log c+\left(\frac{t+c}{2}+2\right)\log(t+c)+\log\Gamma\left(\frac{c+3}{2}\right)-\log\Gamma\left(\frac{t+c+3}{2}\right). \qquad \text{(B.56)}
\end{aligned}
$$

To simplify this formula, we introduce the standardised sum of squares $Z_t(\mu',\sigma')=\sum_{s=1}^t\left(\frac{Y_s-\mu'}{\sigma'}\right)^2$ for $(\mu',\sigma')\in\mathbb{R}\times\mathbb{R}_+^*$. After rearranging terms and denoting $\widehat{\mu}_t = S_t/t$, the above formula reads:

$$
\begin{aligned}
&\frac{1}{2}Z_t(\mu,\sigma)-\frac{t+c+3}{2}\log\left(\frac{t}{t+c}Z_t(\widehat{\mu}_t,\sigma)+\frac{c}{t+c}Z_t(\mu,\sigma)+c\right)\\
&\leqslant \log\frac{1}{\delta}-\frac{t}{2}\log 2-\left(\frac{c}{2}+2\right)\log c+\frac{1}{2}\log(t+c)+\log\Gamma\left(\frac{c+3}{2}\right)-\log\Gamma\left(\frac{t+c+3}{2}\right). \qquad \text{(B.57)}
\end{aligned}
$$

### Bernoulli

We consider $Y\sim\mathcal{B}(\mu)$. This corresponds to an exponential family model, with parameter $\theta^\star=\log\frac{\mu}{1-\mu}$, feature function $F(y)=y$ and log-partition function $\mathcal{L}(\theta)=\log(1+\exp(\theta))$. The Bregman divergence between two parameters $\theta'$ and $\theta$ associated with $\mathcal{L}$ is given by $\mathcal{B}_{\mathcal{L}}(\theta',\theta)=\mathrm{kl}(\mu,\mu')=\mu\log\frac{\mu}{\mu'}+(1-\mu)\log\frac{1-\mu}{1-\mu'}$. We further have $\mathcal{L}'(\theta)=\frac{\exp(\theta)}{1+\exp(\theta)}=\mu$ and $\mathcal{L}''(\theta)=\frac{\exp(\theta)}{(1+\exp(\theta))^2}=\mu(1-\mu)$. Therefore $\mathcal{L}'$ is invertible and we have the expression $\widehat{\theta}_{t,c}(\theta)=(\mathcal{L}')^{-1}\left(\frac{1}{t+c}\sum_{s=1}^t Y_s+\frac{c}{t+c}\frac{\exp(\theta)}{1+\exp(\theta)}\right)$. Then, denoting $S_t=\sum_{s=1}^t Y_t$, we get

$$
\widehat{\mu}_{t,c}(\mu)=\mathcal{L}'(\widehat{\theta}_{t,c}(\theta))=\frac{\sum_{s=1}^t Y_s+c\mu}{t+c}=\frac{S_t+c\mu}{t+c}\,. \qquad \text{(B.58)}
$$

Therefore, the Bregman deviation specifies to the following closed-form formula

$$
\begin{aligned}
(t+c)\mathcal{B}_{\mathcal{L}}(\theta,\widehat{\theta}_{t,c}(\theta)) &= (t+c)\mathtt{kl}(\widehat{\mu}_{t,c}(\mu),\mu) \qquad \text{(B.59)}\\
&= (S_t+c\mu)\log\frac{\widehat{\mu}_{t,c}(\mu)}{\mu}+(t-S_t+c(1-\mu))\log\frac{1-\widehat{\mu}_{t,c}(\mu)}{1-\mu}\,. \qquad \text{(B.60)}
\end{aligned}
$$

Now, we observe that

$$
\int_{\mathbb{R}^d}\exp\left(-c\mathcal{B}_{\mathcal{L}}(\theta',\theta)\right)d\theta'=\frac{\mathrm{B}(c\mu,c(1-\mu))}{\mu^{c\mu}(1-\mu)^{c(1-\mu)}}\,, \qquad \text{(B.61)}
$$

$$
\int_{\mathbb{R}^d}\exp\left(-(t+c)\mathcal{B}_{\mathcal{L}}(\theta',\widehat{\theta}_{t,c}(\theta))\right)d\theta'=\frac{\mathrm{B}((t+c)\widehat{\mu}_{t,c}(\mu),(t+c)(1-\widehat{\mu}_{t,c}(\mu)))}{\widehat{\mu}_{t,c}(\mu)^{(t+c)\widehat{\mu}_{t,c}(\mu)}(1-\widehat{\mu}_{t,c}(\mu))^{(t+c)(1-\widehat{\mu}_{t,c}(\mu))}}\,. \qquad \text{(B.62)}
$$

Therefore, we deduce that the Bregman information gain rewrites

$$\gamma_{t,c}(\mu) = (S_t + c\mu)\log\widehat{\mu}_{t,c}(\mu) + (t - S_t + c(1-\mu))\log(1-\widehat{\mu}_{t,c}(\mu)) - c\mu\log\mu - c(1-\mu)\log(1-\mu) \tag{B.63}$$

$$+ \log\frac{\Gamma(c\mu)\Gamma(c(1-\mu))}{\Gamma(S_t + c\mu)\Gamma(t - S_t + c(1-\mu))} + \log\frac{\Gamma(t+c)}{\Gamma(c)}\,. \tag{B.64}$$

Combining the above and using Theorem 3.3, we obtain that with probability at least $1-\delta$, for all $t \in \mathbb{N}$,

$$S_t\log\frac{1}{\mu} + (t - S_t)\log\frac{1}{1-\mu} + \log\frac{\Gamma(S_t + c\mu)\Gamma(t - S_t + c(1-\mu))}{\Gamma(c\mu)\Gamma(c(1-\mu))} \leqslant \log\frac{1}{\delta} + \log\frac{\Gamma(t+c)}{\Gamma(c)}\,. \tag{B.65}$$

**Interpretation.** At this point, we can get more intuition about these terms using Stirling's approximation

$$\Gamma(z) = \sqrt{\frac{2\pi}{z}}\left(\frac{z}{e}\right)^z\left(1 + O\left(\frac{1}{z}\right)\right)\,. \tag{B.66}$$

Indeed, we derive the following asymptotic approximation

$$\gamma_{t,c}(\theta) \approx \frac{1}{2}\log\frac{t+c}{c} + \frac{1}{2}\log\frac{\widehat{\mu}_{t,c}(\mu)(1-\widehat{\mu}_{t,c}(\mu))}{\mu(1-\mu)}\,. \tag{B.67}$$

Hence, we obtain with probability at least $1-\delta$, that for all $t \in \mathbb{N}$,

$$\mathrm{kl}\left(\widehat{\mu}_{t,c}(\mu), \mu\right) \lessapprox \frac{1}{t+c}\left(\log\frac{1}{\delta} + \frac{1}{2}\log\frac{t+c}{c} + \frac{1}{2}\log\frac{\widehat{\mu}_{t,c}(\mu)(1-\widehat{\mu}_{t,c}(\mu))}{\mu(1-\mu)}\right)\,. \tag{B.68}$$

Note that the last term in the right-hand side of the above equation is given by

$$\frac{1}{2}\log\frac{\mathbb{V}_{\widehat{\mu}_{t,c}(\mu)}\left[Y\right]}{\mathbb{V}_{\mu}\left[Y\right]}\,, \tag{B.69}$$

which shows that the Bregman information gain induces a variance-dependent bound in a similar spirit as Bernstein's inequality.

### Exponential

We consider $Y \sim \mathcal{E}(1/\mu)$ with unknown mean $\mu$. The distribution of $Y$ is supported on $[0, +\infty)$ with density $p_\mu(y) = \frac{1}{\mu}e^{-\frac{y}{\mu}}$. This corresponds to an exponential family model with parameter $\theta^\star = -\frac{1}{\mu}$, feature function $F(y) = y$ and log-partition function $\mathcal{L}(\theta) = \log(-\frac{1}{\theta})$.

The Bregman divergence between two parameters $\theta'$ and $\theta$ associated with $\mathcal{L}$ is given by $\mathcal{B}_{\mathcal{L}}(\theta', \theta) = \mathrm{KL}(p_\mu \parallel p_{\mu'}) = \frac{\mu}{\mu'} - 1 - \log \frac{\mu}{\mu'}$.

We have $\mathcal{L}'(\theta) = -\frac{1}{\theta} = \mu$ and $\mathcal{L}''(\theta) = \frac{1}{\theta^2} = \mu^2$. Therefore $\mathcal{L}'$ is invertible and we have

$$\widehat{\mu}_{t,c}(\mu) := \mathcal{L}'(\widehat{\theta}_{t,c}(\theta)) = \frac{S_t + c\mu}{t+c}\,. \tag{B.70}$$

Therefore, we deduce that the Bregman divergence takes the following form

$$(t+c)\mathcal{B}_{\mathcal{L}}(\theta, \widehat{\theta}_{t,c}(\theta)) = (t+c)\mathrm{KL}(p_{\widehat{\mu}_{t,c}(\mu)} \parallel p_\mu) = \frac{S_t}{\mu} - (t+c)\log\frac{\widehat{\mu}_{t,c}(\mu)}{\mu} - t\,. \tag{B.71}$$

Now, we observe that

$$\int_{\mathbb{R}^d} \exp\left(-c\mathcal{B}_{\mathcal{L}}(\theta', \theta)\right) d\theta' = \frac{\Gamma(c+1)}{c\mu}\left(\frac{e}{c}\right)^c = \frac{\Gamma(c)}{\mu}\left(\frac{e}{c}\right)^c, \tag{B.72}$$

$$\int_{\mathbb{R}^d} \exp\left(-(t+c)\mathcal{B}_{\mathcal{L}}(\theta', \widehat{\theta}_{t,c}(\theta))\right) d\theta' = \frac{\Gamma(t+c+1)}{(t+c)\widehat{\mu}_{t,c}(\mu)}\left(\frac{e}{t+c}\right)^{t+c} = \frac{\Gamma(t+c)}{\widehat{\mu}_{t,c}(\mu)}\left(\frac{e}{t+c}\right)^{t+c}. \tag{B.73}$$

Therefore, the Bregman information gain writes explicitly as follows

$$\gamma_{t,c}(\theta) = \log\frac{\widehat{\mu}_{t,c}(\mu)}{\mu} + \log\left(\Gamma(c)\left(\frac{e}{c}\right)^c\right) - \log\left(\Gamma(t+c)\left(\frac{e}{t+c}\right)^{t+c}\right). \tag{B.74}$$

We can now specify the inequality $(t+c)\mathcal{B}_{\mathcal{L}}(\theta, \widehat{\theta}_{t,c}(\theta)) \leqslant \log(1/\delta) + \gamma_{t,c}(\theta)$. Combining the above, we obtain with probability at least $1-\delta$, for all $t \in \mathbb{N}$,

$$\frac{S_t}{\mu} - (t+c+1)\log\left(\frac{S_t + c\mu}{(t+c)\mu}\right) \leqslant (t+c)\log(t+c) + \log\frac{\Gamma(c)}{\Gamma(t+c)} + \log\frac{1}{\delta} - c\log c\,. \tag{B.75}$$

**Interpretation.** As in the Bernoulli case, Stirling's approximation provides further intuition. First, we have the following asymptotic approximation

$$\gamma_{t,c}(\theta) \approx \frac{1}{2}\log\frac{t+c}{c} + \log\frac{\widehat{\mu}_{t,c}(\mu)}{\mu}\,, \tag{B.76}$$

and hence, with probability at least $1-\delta$, for all $t \in \mathbb{N}$, we have

$$\mathrm{KL}(p_{\widehat{\mu}_{t,c}(\mu)} \parallel p_\mu) \lessapprox \frac{1}{t+c}\left(\log\frac{1}{\delta} + \frac{1}{2}\log\frac{t+c}{c} + \log\frac{\widehat{\mu}_{t,c}(\mu)}{\mu}\right). \tag{B.77}$$

Note that, as for Bernoulli distributions, the last term in the right-hand side of the above equation is given by the Bernstein-like term

$$\frac{1}{2}\log\frac{\mathbb{V}_{\widehat{\mu}_{t,c}(\mu)}[Y]}{\mathbb{V}_{\mu}[Y]}\,. \tag{B.78}$$

### Gamma with fixed shape

We consider $Y \sim \mathrm{Gamma}(k,\lambda)$ with fixed shape $k>0$ and unknown scale $\lambda>0$.[1] The distribution of $Y$ is supported on $[0,\infty)$ with density $p_\lambda(y)=\frac{1}{\Gamma(k)\lambda^k}y^{k-1}e^{-\frac{y}{\lambda}}$. This corresponds to an exponential family model with parameter $\theta^\star=-\frac{1}{\lambda}$, feature function $F(y)=y$ and log-partition function $\mathcal{L}(\theta)=\log(\lambda^k)=k\log(-\frac{1}{\theta})$. The Bregman divergence between two parameters $\theta'$ and $\theta$ associated with $\mathcal{L}$ is given by $\mathcal{B}_{\mathcal{L}}(\theta',\theta)=\mathrm{KL}(p_\lambda \parallel p_{\lambda'})=k\left(\frac{\lambda}{\lambda'}-1-\log\left(\frac{\lambda}{\lambda'}\right)\right)$.

Note that $\mathcal{L}'(\theta)=-\frac{k}{\theta}=k\lambda$ and $\mathcal{L}''(\theta)=\frac{k}{\theta^2}=k\lambda$. Therefore $\mathcal{L}'$ is invertible, and we get

$$k\widehat{\lambda}_{t,c}(\lambda)=\mathcal{L}'(\widehat{\theta}_{t,c}(\theta))=\frac{\sum_{s=1}^t Y_s+ck\lambda}{t+c} \tag{B.79}$$

Therefore, the Bregman divergence takes the following form

$$\begin{aligned}(t+c)\mathcal{B}_{\mathcal{L}}(\theta,\widehat{\theta}_{t,c}(\theta)) &= (t+c)\mathrm{KL}(p_{\widehat{\lambda}_{t,c}(\lambda)}\parallel p_\lambda)\\ &= k(t+c)\left(\frac{\sum_{s=1}^t Y_s+ck\lambda}{(t+c)k\lambda}\right)-k(t+c)\log\left(\frac{\sum_{s=1}^t Y_s+ck\lambda}{(t+c)k\lambda}\right)-k(t+c)\,.\end{aligned} \tag{B.80}$$

Now, we observe that

$$\int_{\mathbb{R}^d}\exp\left(-c\mathcal{B}_{\mathcal{L}}(\theta',\theta)\right)d\theta'=\frac{\Gamma(ck)}{\lambda}\left(\frac{e}{ck}\right)^{ck}, \tag{B.81}$$

$$\int_{\mathbb{R}^d}\exp\left(-(t+c)\mathcal{B}_{\mathcal{L}}(\theta',\widehat{\theta}_{t,c}(\theta))\right)d\theta'=\frac{\Gamma((t+c)k)}{\widehat{\lambda}_{t,c}(\lambda)}\left(\frac{e}{(t+c)k}\right)^{(t+c)k}. \tag{B.82}$$

Therefore, the Bregman information gain writes explicitly as follows

$$\gamma_{t,c}(\theta)=\log\left(\frac{\sum_{s=1}^t Y_s+ck\lambda}{(t+c)k\lambda}\right)+\log\left(\Gamma(ck)\left(\frac{e}{ck}\right)^{ck}\right)-\log\left(\Gamma((t+c)k)\left(\frac{e}{(t+c)k}\right)^{(t+c)k}\right). \tag{B.83}$$

[1]Note that $\mathrm{Gamma}(\lambda,k)$ with $k=1$ is $\mathcal{E}(1/\lambda)$.

Using the above with Theorem 3.3, we obtain with probability at least $1 - \delta$, for all $t \in \mathbb{N}$,

$$k(t+c)\left(\frac{\sum_{s=1}^t Y_s + ck\lambda}{(t+c)k\lambda}\right) - (k(t+c)+1)\log\left(\frac{\sum_{s=1}^t Y_s + ck\lambda}{(t+c)k\lambda}\right)$$
$$\leqslant \log\frac{1}{\delta} + \log\frac{\Gamma(ck)}{\Gamma((t+c)k)} + (t+c)k\log((t+c)k) + ck - ck\log ck\,. \tag{B.84}$$

### Weibull with fixed shape

We consider $Y \sim \text{Weibull}(\lambda, k)$ with fixed shape $k > 0$ and unknown scale $\lambda > 0$.[2] The distribution of $Y$ is supported on $[0,\infty)$ with density $p_\lambda(y) = \frac{k}{\lambda}\left(\frac{y}{\lambda}\right)^{k-1} e^{-\left(\frac{y}{\lambda}\right)^k}$. This corresponds to an exponential family model with parameter $\theta^\star = -\frac{1}{\lambda^k}$, feature function $F(y) = y^k$ and log-partition function $\mathcal{L}(\theta) = \log(\lambda^k) = \log(-\frac{1}{\theta})$. The Bregman divergence between two parameters $\theta'$ and $\theta$ associated with $\mathcal{L}$ is given by $\mathcal{B}_\mathcal{L}(\theta',\theta) = \text{KL}(p_\lambda \,\|\, p_{\lambda'}) = \left(\frac{\lambda}{\lambda'}\right)^k - 1 - \log\left(\frac{\lambda}{\lambda'}\right)^k$.

Note that $\mathcal{L}'(\theta) = -\frac{1}{\theta} = \lambda^k$ and $\mathcal{L}''(\theta) = \frac{1}{\theta^2} = \lambda^{2k}$. Therefore $\mathcal{L}'$ is invertible, and we get

$$(\widehat{\lambda}_{t,c}(\lambda))^k = \mathcal{L}'(\widehat{\theta}_{t,c}(\theta)) = \frac{\sum_{s=1}^t Y_s^k + c\lambda^k}{t+c}\,. \tag{B.85}$$

Therefore, the Bregman divergence takes the following form

$$\begin{aligned}(t+c)\mathcal{B}_\mathcal{L}(\theta, \widehat{\theta}_{t,c}(\theta)) &= (t+c)\text{KL}(p_{\widehat{\lambda}_{t,c}(\lambda)} \,\|\, p_\lambda)\\ &= (t+c)\left(\frac{\sum_{s=1}^t Y_s^k + c\lambda^k}{(t+c)\lambda^k}\right) - (t+c)\log\left(\frac{\sum_{s=1}^t Y_s^k + c\lambda^k}{(t+c)\lambda^k}\right) - (t+c)\,.\end{aligned} \tag{B.86}$$

Now, we observe that

$$\int_{\mathbb{R}^d} \exp\left(-c\mathcal{B}_\mathcal{L}(\theta',\theta)\right) d\theta' = \frac{\Gamma(c)}{\lambda^k}\left(\frac{e}{c}\right)^c, \tag{B.87}$$
$$\int_{\mathbb{R}^d} \exp\left(-(t+c)\mathcal{B}_\mathcal{L}(\theta', \widehat{\theta}_{t,c}(\theta))\right) d\theta' = \frac{\Gamma(t+c)}{(\widehat{\lambda}_{t,c}(\lambda))^k}\left(\frac{e}{t+c}\right)^{t+c}. \tag{B.88}$$

Therefore, the Bregman information gain writes explicitly as follows

$$\gamma_{t,c}(\theta) = \log\left(\frac{\sum_{s=1}^t Y_s^k + c\lambda^k}{(t+c)\lambda^k}\right) + \log\left(\Gamma(c)\left(\frac{e}{c}\right)^c\right) - \log\left(\Gamma(t+c)\left(\frac{e}{t+c}\right)^{t+c}\right). \tag{B.89}$$

[2] Note that $\text{Weibull}(\lambda, k)$ with $k = 1$ is $\mathcal{E}(1/\lambda)$.

Using the above with Theorem 3.3, we obtain with probability at least $1-\delta$, for all $t \in \mathbb{N}$,

$$
\begin{aligned}
&(t+c)\left(\frac{\sum_{s=1}^t Y_s^k + c\lambda^k}{(t+c)\lambda^k}\right) - (t+c+1)\log\left(\frac{\sum_{s=1}^t Y_s^k + c\lambda^k}{(t+c)\lambda^k}\right) \\
&\leqslant \log\frac{1}{\delta} + \log\frac{\Gamma(c)}{\Gamma(t+c)} + (t+c)\log(t+c) + c - c\log c\,. \qquad \text{(B.90)}
\end{aligned}
$$

### Pareto with fixed scale

We consider $Y \sim \text{Pareto}\,(\alpha)$, where $\alpha > 0$ is unknown.[3] The distribution of $Y$ is supported in $[1, +\infty)$, with density $p_\alpha(y) = \frac{\alpha}{y^{\alpha+1}}$, which corresponds to a one-dimensional exponential family model with parameter $\theta^\star = -\alpha - 1 \in (-\infty, -1)$, feature function $F(y) = \log y$ and log-partition function $\mathcal{L}(\theta) = -\log(-1-\theta)$. The first two derivatives of $\mathcal{L}$ are given by $\mathcal{L}'(\theta) = -\frac{1}{1+\theta}$ and $\mathcal{L}''(\theta) = \frac{1}{(1+\theta)^2}$, therefore $\mathcal{L}'$ is invertible on the domain $(-\infty, -1)$. Using these expressions, the Bregman divergence between two parameters $\theta'$ and $\theta$ writes $\mathcal{B}_\mathcal{L}(\theta', \theta) = -\log(-1-\theta') + \log(-1-\theta) + \frac{\theta'-\theta}{1+\theta}$. Using the shorthand $L_t = \sum_{s=1}^t \log Y_s$, it follows from the definition that:

$$
\widehat{\theta}_{t,c}(\theta) = -1 + \frac{t+c}{\frac{c}{1+\theta} - L_t}\,. \qquad \text{(B.91)}
$$

To compute the Bregman information gain, we rewrite the following integral thanks to an affine change of variable in order to relate it to the Gamma function:

$$
\begin{aligned}
\int_{-\infty}^{-1} \exp\left(-c\mathcal{B}_\mathcal{L}(\theta', \theta)\right) d\theta' &= (-1-\theta)^{-c} e^{\frac{c\theta}{1+\theta}} \int_{-\infty}^{-1} (-1-\theta')^c e^{-\frac{c\theta'}{1+\theta}} d\theta' \\
&= (-1-\theta)\,\Gamma(c)\left(\frac{e}{c}\right)^c\,. \qquad \text{(B.92)}
\end{aligned}
$$

The expression of the Bregman information then follows immediately:

$$
\gamma_{t,c}(\theta) = -\log\left(\frac{t+c}{(-1-\theta)L_t + c}\right) - t - c\log c + (t+c)\log(t+c) + \log\Gamma(c) - \log\Gamma(t+c)\,. \qquad \text{(B.93)}
$$

Moreover, we deduce from the expression of $\widehat{\theta}_{t,c}(\theta)$ and the Bregman divergence that:

$$
\mathcal{B}_\mathcal{L}(\theta', \widehat{\theta}_{t,c}(\theta)) = -\log(-1-\theta') + \log\left(\frac{t+c}{L_t - \frac{c}{1+\theta}}\right) - \frac{\left(L_t - \frac{c}{1+\theta}\right)(\theta'+1)}{t+c} - 1\,. \qquad \text{(B.94)}
$$

[3]We assume scale is fixed to the value 1.

Therefore, Theorem 3.3 combined with the natural parameter $\theta = -1 - \alpha$ yields that with probability at least $1 - \delta$, for all $t \in \mathbb{N}$,

$$\alpha L_t - (t{+}c{+}1) \log\left(\alpha L_t + c\right) \leqslant \log \frac{1}{\delta} - c \log c - \log(t{+}c) + \log \Gamma(c) - \log \Gamma(t{+}c)\,. \tag{B.95}$$

### Chi-square

We consider $Y \sim \chi^2(k)$ or, equivalently, $Y \sim \mathrm{Gamma}\left(\frac{k}{2}, \frac{1}{2}\right)$, i.e. $p_k(y) = \frac{(\frac{1}{2})^{\frac{k}{2}}}{\Gamma(\frac{k}{2})} y^{\frac{k}{2}-1} e^{-\frac{y}{2}}$, $x \geqslant 0$, $k \in \mathbb{N}$ (or $k \in \mathbb{R}_+^*$ if one considers `Gamma` distributions). This corresponds to an exponential family model with parameter $\theta^\star = \frac{k}{2} - 1$, feature function $F(y) = \log y$ and log-partition function $\mathcal{L}(\theta) = (\theta + 1)\log 2 + \log \Gamma(\theta + 1)$. The Bregman divergence between two parameters $\theta'$ and $\theta$ associated with $\mathcal{L}$ is given by

$$\mathcal{B}_{\mathcal{L}}(\theta', \theta) = \mathrm{KL}(p_k \parallel p_{k'}) = \frac{1}{2}(k - k')\psi_0(k/2) - \log \frac{\Gamma(k/2)}{\Gamma(k'/2)}\,, \tag{B.96}$$

where $\psi_0(z) = \frac{d}{dz} \log \Gamma(z)$ denotes the digamma function.

We further have $\mathcal{L}'(\theta) = \log 2 + \psi_0(\theta + 1)$ and $\mathcal{L}''(\theta) = \psi_1(\theta + 1)$, where $\psi_1(z) = \frac{d^2}{dz^2} \log \Gamma(z)$ denotes the trigamma function. Therefore $\mathcal{L}'$ is invertible, and the parameter estimate is given by

$$\log 2 + \psi_0(\widehat{\theta}_{t,c}(\theta) + 1) = \frac{1}{t+c} \sum_{s=1}^{t} \log Y_s + \frac{c}{t+c}(\log 2 + \psi_0(\theta + 1))\,, \tag{B.97}$$

yielding

$$\begin{aligned}\widehat{k}_{t,c}(k) := 2(1 + \widehat{\theta}_{t,c}(\theta)) &= 2\psi_0^{-1}\left(\frac{1}{t+c}\sum_{s=1}^{t} \log Y_s + \frac{c}{t+c}\psi_0(k/2) - \frac{t}{t+c}\log 2\right) \\ &= 2\psi_0^{-1}\left(\frac{1}{t+c}K_t + \frac{c}{t+c}\psi_0(k/2)\right)\,,\end{aligned} \tag{B.98}$$

where $K_t = \sum_{s=1}^{t} \log \frac{Y_s}{2}$. Therefore, the Bregman divergence rewrites as follows

$$(t + c)\mathcal{B}_{\mathcal{L}}(\theta, \widehat{\theta}_{t,c}(\theta)) = (t + c)\mathrm{KL}(p_{\widehat{k}_{t,c}(k)} \parallel P_k) \tag{B.99}$$

$$= \frac{1}{2}(\widehat{k}_{t,c}(k) - k)\left(K_t + c\,\psi_0\left(\frac{k}{2}\right)\right) - (t + c)\log \frac{\Gamma\left(\frac{\widehat{k}_{t,c}(k)}{2}\right)}{\Gamma\left(\frac{k}{2}\right)}\,. \tag{B.100}$$

Now, we see that

$$\int_{\mathbb{R}^d} \exp\left(-c\mathcal{B}_{\mathcal{L}}(\theta', \theta)\right) d\theta' = \int \frac{1}{2} \exp\left(-\frac{c}{2}(k-k')\psi_0(k/2)\right)\left(\frac{\Gamma(k/2)}{\Gamma(k'/2)}\right)^c dk' \,, \tag{B.101}$$

$$\int_{\mathbb{R}^d} \exp\left(-(n+c)\mathcal{B}_{\mathcal{L}}(\theta', \widehat{\theta}_{t,c}(\theta))\right) d\theta' = \int \frac{1}{2} \exp\left(-\frac{t+c}{2}(\widehat{k}_{t,c}(k)-k')\psi_0(\widehat{k}_{n,c}(k)/2)\right)\left(\frac{\Gamma(\widehat{k}_{t,c}(k)/2)}{\Gamma(k'/2)}\right)^{t+c} dk' \,. \tag{B.102}$$

Therefore, the Bregman information gain writes

$$\begin{aligned}
\gamma_{t,c}(\theta) &= c\log\Gamma\left(\frac{k}{2}\right) - (t+c)\log\Gamma\left(\frac{\widehat{k}_{t,c}(k)}{2}\right) - c\,\frac{k}{2}\psi_0\left(\frac{k}{2}\right) + (t+c)\frac{\widehat{k}_{t,c}(k)}{2}\psi_0\left(\frac{\widehat{k}_{t,c}(k)}{2}\right) \\
&\quad + \log\frac{\int\left(\Gamma\left(k'/2\right)\right)^{-c}\exp\left(c\,\frac{k'}{2}\psi_0\left(\frac{k}{2}\right)\right)dk'}{\int\left(\Gamma\left(k'/2\right)\right)^{-(t+c)}\exp\left((t+c)\frac{k'}{2}\psi_0\left(\frac{\widehat{k}_{t,c}(k)}{2}\right)\right)dk'} \\
&= c\log\Gamma\left(\frac{k}{2}\right) - (t+c)\log\Gamma\left(\frac{\widehat{k}_{t,c}(k)}{2}\right) - c\,\frac{k}{2}\psi_0\left(\frac{k}{2}\right) + \frac{\widehat{k}_{t,c}(k)}{2}\left(K_t + c\psi_0\left(\frac{k}{2}\right)\right) \\
&\quad + \log\frac{\int\left(\Gamma\left(k'/2\right)\right)^{-c}\exp\left(c\,\frac{k'}{2}\psi_0\left(\frac{k}{2}\right)\right)dk'}{\int\left(\Gamma\left(k'/2\right)\right)^{-(t+c)}\exp\left(\frac{k'}{2}\left(K_t + c\psi_0\left(\frac{k}{2}\right)\right)\right)dk'} \\
&= c\log\Gamma\left(\frac{k}{2}\right) - (t+c)\log\Gamma\left(\frac{\widehat{k}_{t,c}(k)}{2}\right) - c\,\frac{k}{2}\psi_0\left(\frac{k}{2}\right) + \frac{\widehat{k}_{t,c}(k)}{2}\left(K_t + c\psi_0\left(\frac{k}{2}\right)\right) \\
&\quad + \log J\left(c, c\psi_0\left(\frac{k}{2}\right)\right) - \log J\left(t{+}c, K_t{+}c\psi_0\left(\frac{k}{2}\right)\right) \,,
\end{aligned} \tag{B.103}$$

where we define the auxiliary function $J(a,b) = \int \exp\left(-a\log\Gamma\left(\frac{k'}{2}\right) + b\frac{k'}{2}\right)dk'$. Combining the above with Theorem 3.3, and after some simple algebra, we obtain that with probability at least $1-\delta$, for all $t \in \mathbb{N}$,

$$n\log\Gamma\left(\frac{k}{2}\right) - \frac{k}{2}K_t - \log J\left(c, c\psi_0\left(\frac{k}{2}\right)\right) + \log J\left(t{+}c, K_t{+}c\psi_0\left(\frac{k}{2}\right)\right) \leqslant \log\frac{1}{\delta} \,. \tag{B.104}$$

**Remark B.3** (Discrete or continuous parameters). *The integral terms* $\int dk'$ *in the above derive from the martingale construction in the proof of Theorem 3.3 and the mixture distribution* $q(\theta|\alpha, \beta)$ *over the parameter* $\theta \in \Theta$ *of the exponential family. In the case of* $\mathrm{Gamma}\left(\frac{k}{2}, \frac{1}{2}\right)$ *with unknown shape* $\frac{k}{2} > 0$, *we have* $\theta = \frac{k}{2}$ *and* $\Theta = (0, +\infty)$, *therefore* $dk'$ *corresponds to the Lebesgue measure over* $(0, +\infty)$. *When restricted to the Chi-square family,* $\Theta = \mathbb{N}$, *and* $dk'$ *is instead the counting measure, effectively turning integrals into discrete sums. In Appendix B.3, we report figures using both versions, see Figure B.4 and Figure B.9.*

**Remark B.4** (Computing the integral $J$). *The ratio of integrals (or infinite sums) in (B.104) can be efficiently implemented using a simple integration scheme (or by truncation). Indeed, for a given* $k_{\max} \in \mathbb{N}$, $b \ll 1$ *and* $B \gg 1$, *we define* $x_{k'} = b + \frac{k'}{k_{\max}}(B-b)$ *for* $k' = 0, \dots, k_{\max}$, *so that*

$$\log \frac{\int_0^{+\infty} \left(\Gamma\left(\frac{k'}{2}\right)\right)^{-c} \exp\left(\frac{k'}{2} c\psi_0\left(\frac{k}{2}\right)\right) dk'}{\int_0^{+\infty} \left(\Gamma\left(\frac{k'}{2}\right)\right)^{-(n+c)} \exp\left(\frac{k'}{2}\left(c\psi_0\left(\frac{k}{2}\right) + \sum_{t=1}^n \log \frac{Y_t}{2}\right)\right) dk'}$$
$$\approx \operatorname*{logsumexp}_{k'=1,\dots,k_{\max}} \left(-c \log \Gamma\left(\frac{x_{k'}}{2}\right) + \frac{c x_{k'}}{2}\psi_0\left(\frac{k}{2}\right) + \log\left(x_{k'} - x_{k'-1}\right)\right)$$
$$- \operatorname*{logsumexp}_{k'=1,\dots,k_{\max}} \left(-(n+c)\Gamma\left(\frac{x_{k'}}{2}\right) + \frac{x_{k'}}{2}\left(c\psi_0\left(\frac{k}{2} + \sum_{t=1}^n \log \frac{Y_t}{2}\right)\right) + \log\left(x_{k'} - x_{k'-1}\right)\right). \tag{B.105}$$

*Similarly, we have*

$$\log \frac{\sum_{k'=1}^{+\infty} \left(\Gamma\left(\frac{k'}{2}\right)\right)^{-c} \exp\left(\frac{k'}{2} c\psi_0\left(\frac{k}{2}\right)\right)}{\sum_{k'=1}^{+\infty} \left(\Gamma\left(\frac{k'}{2}\right)\right)^{-(t+c)} \exp\left(\frac{k'}{2}\left(c\psi_0\left(\frac{k}{2}\right) + \sum_{s=1}^t \log \frac{Y_s}{2}\right)\right)}$$
$$\approx \operatorname*{logsumexp}_{k'=1,\dots,k_{\max}} \left(-c \log \Gamma\left(\frac{k'}{2}\right) + \frac{k'}{2} c\psi_0\left(\frac{k}{2}\right)\right)$$
$$- \operatorname*{logsumexp}_{k'=1,\dots,k_{\max}} \left(-(t+c)\Gamma\left(\frac{k'}{2}\right) + \frac{k'}{2}\left(c\psi_0\left(\frac{k}{2} + \sum_{s=1}^t \log \frac{Y_s}{2}\right)\right)\right), \tag{B.106}$$

*The final steps correspond to the right-rectangular scheme over* $(b, B)$ *with* $k_{\max}$ *steps and the truncation to the first* $k_{\max}$ *terms respectively. Note the use of* $\operatorname*{logsumexp}_{k'=1,\dots,k_{\max}}(z) = \log \sum_{k'=1}^{k_{\max}} \exp(z_{k'})$ *for* $z \in \mathbb{R}^{k_{\max}}$, *which is efficiently implemented in many libraries for scientific computing and better handles summation of large numbers. Empirically, we found that* $\log b = -10$, $\log B = 10$ *and* $k_{\max} = 2000$ *provided sufficient accuracy and that using finer approximation schemes did not significantly impact the numerical results.*

### Poisson

We finally consider $Y \sim \mathcal{P}(\lambda)$, where $\lambda > 0$ is unknown. We recall that the distribution of $Y$ is supported on $\mathbb{N}$ with probability mass function $\mathbb{P}(Y = k) = \frac{\lambda^k e^{-\lambda}}{k!}$. This corresponds to a one-dimensional exponential family model with parameter $\theta^\star = \log \lambda \in \mathbb{R}$, feature function $F(y) = y$ and log-partition function $\mathcal{L}(\theta) = e^\theta$ (which is invertible on $\mathbb{R}$). The Bregman divergence between two parameters $\theta'$ and $\theta$ is therefore $\mathcal{B}_{\mathcal{L}}(\theta', \theta) = e^{\theta'} - e^{\theta} - (\theta' - \theta)e^{\theta}$. Using the shorthand $S_t = \sum_{s=1}^{t} Y_s$, it follows from the definition that:

$$\widehat{\theta}_{t,c}(\theta) = \log\left(\frac{S_t + ce^{\theta}}{t+c}\right) . \tag{B.107}$$

The Bregman information gain is expressed using the auxiliary function $I(a,b) = \int_{-\infty}^{+\infty} e^{-ae^{\theta}+b\theta} d\theta$ as:

$$\begin{aligned}\gamma_{t,c}(\theta) &= c(1-\theta)e^{\theta} - (t+c)\left(1 - \widehat{\theta}_{t,c}(\theta)\right)e^{\widehat{\theta}_{t,c}(\theta)} + \log I\left(c, ce^{\theta}\right) - \log I\left(t+c, (t+c)e^{\widehat{\theta}_{t,c}(\theta)}\right) \\ &= c(1-\theta)e^{\theta} - \left(1 - \log\left(\frac{S_t+ce^{\theta}}{t+c}\right)\right)\left(S_t+ce^{\theta}\right) + \log I\left(c, ce^{\theta}\right) - \log I\left(t+c, S_t+ce^{\theta}\right) .\end{aligned} \tag{B.108}$$

Moreover, we deduce from the expression of $\widehat{\theta}_{t,c}(\theta)$ and the Bregman divergence that:

$$\mathcal{B}_{\mathcal{L}}(\theta', \widehat{\theta}_{t,c}(\theta)) = e^{\theta'} - \left(\frac{S_t+ce^{\theta}}{t+c}\right) - \left(\theta' - \log\left(\frac{S_t+ce^{\theta}}{t+c}\right)\right)\left(\frac{S_t+ce^{\theta}}{t+c}\right) . \tag{B.109}$$

Therefore, Theorem 3.3 combined with the natural parametrisation $\theta = \log\lambda$ yields that with probability at least $1-\delta$, for all $t \in \mathbb{N}$,

$$t\lambda - S_t \log\lambda \leqslant \log\frac{1}{\delta} + \log I\left(c, c\lambda\right) - \log I\left(t+c, S_t+c\lambda\right) . \tag{B.110}$$

**Remark B.5** (Computing the integral $I$). *Although, to the best of our knowledge, the integral $I(a,b)$ does not have a closed-form expression, it can be numerically estimated up to arbitrary precision. We recommend the same implementation as discussed in Remark B.4, using the* logsumexp *operator for stability, and refer to the code for further details.*

## B.3 Empirical comparison with existing time-uniform confidence sequences

We illustrate the time-uniform confidence sequences derived from Bregman concentration on several instances of classical exponential families detailed in 3.3. In each setting, when available, we also report confidence sequences based on existing methods in the literature.

In what follows, we fix $\delta \in (0,1)$ the uniform confidence level. For each confidence sequence $\left(\widehat{\Theta}_t^\delta\right)_{t\in\mathbb{N}}(\delta)$, we report in the figures the intersection sequence $\left(\cap_{s\leqslant t}\widehat{\Theta}_s^\delta\right)_{t\in\mathbb{N}}$, which also holds with confidence $1-\delta$, for $t$ up to $200$. Typical realisations of Bregman confidence sequences are reported in Figure B.1 (Gaussian), Figure B.2 (Bernoulli, Poisson), Figure B.3 (Exponential, Gamma, Weibull, Pareto), and Figure B.4 (Chi-square).

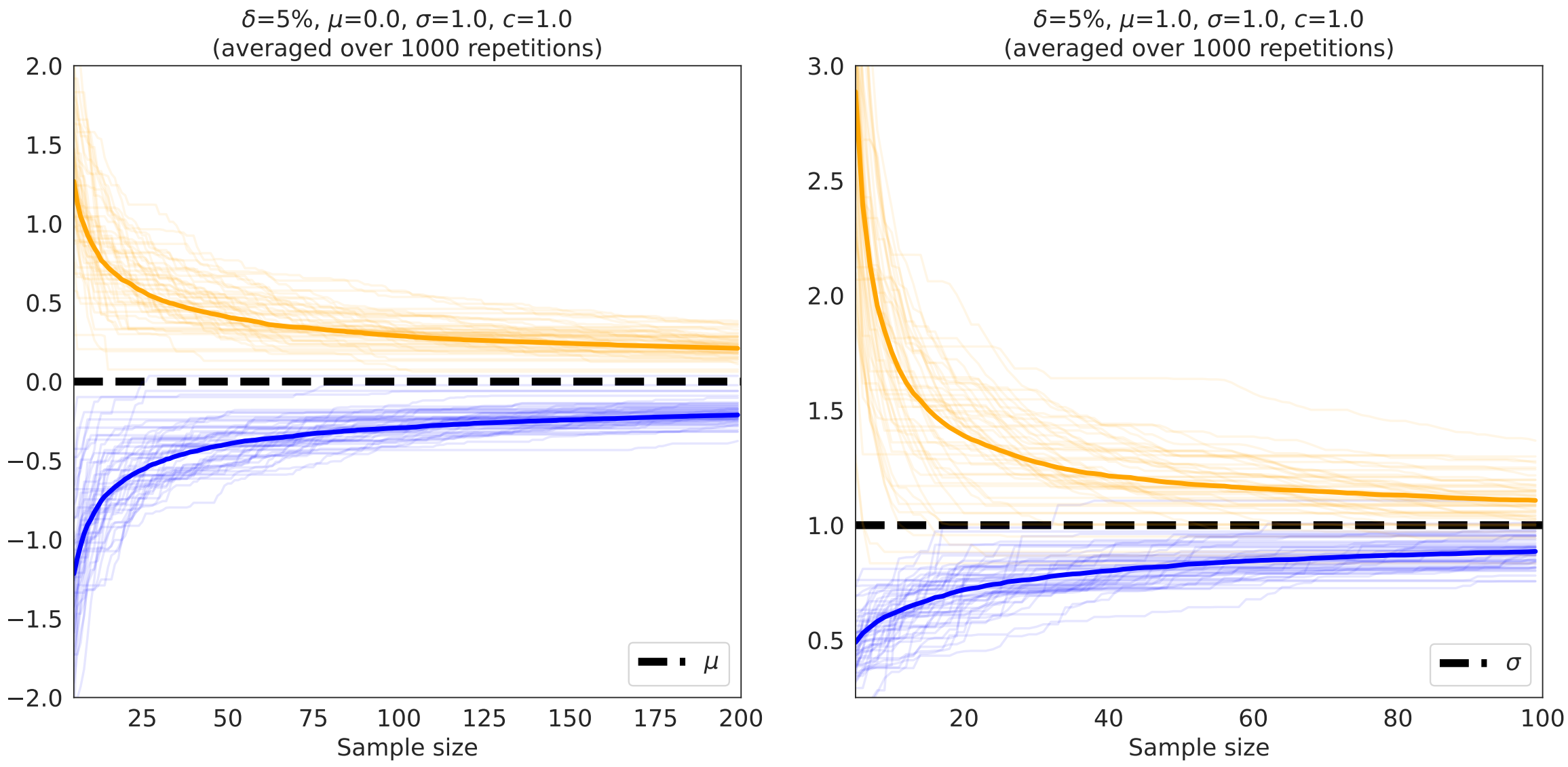

**Figure B.1 –** Examples of confidence upper and lower envelopes around unknown expectation $\mu = 0$ for $\mathcal{N}(\mu, 1)$ (left, cf. Table 3.1) and unknown standard deviation $\sigma = 1$ for $\mathcal{N}(1, \sigma)$ (right, cf. Table 3.1), as a function of the number of observations $t$. The thick lines indicate the median curve over 1000 replicates.

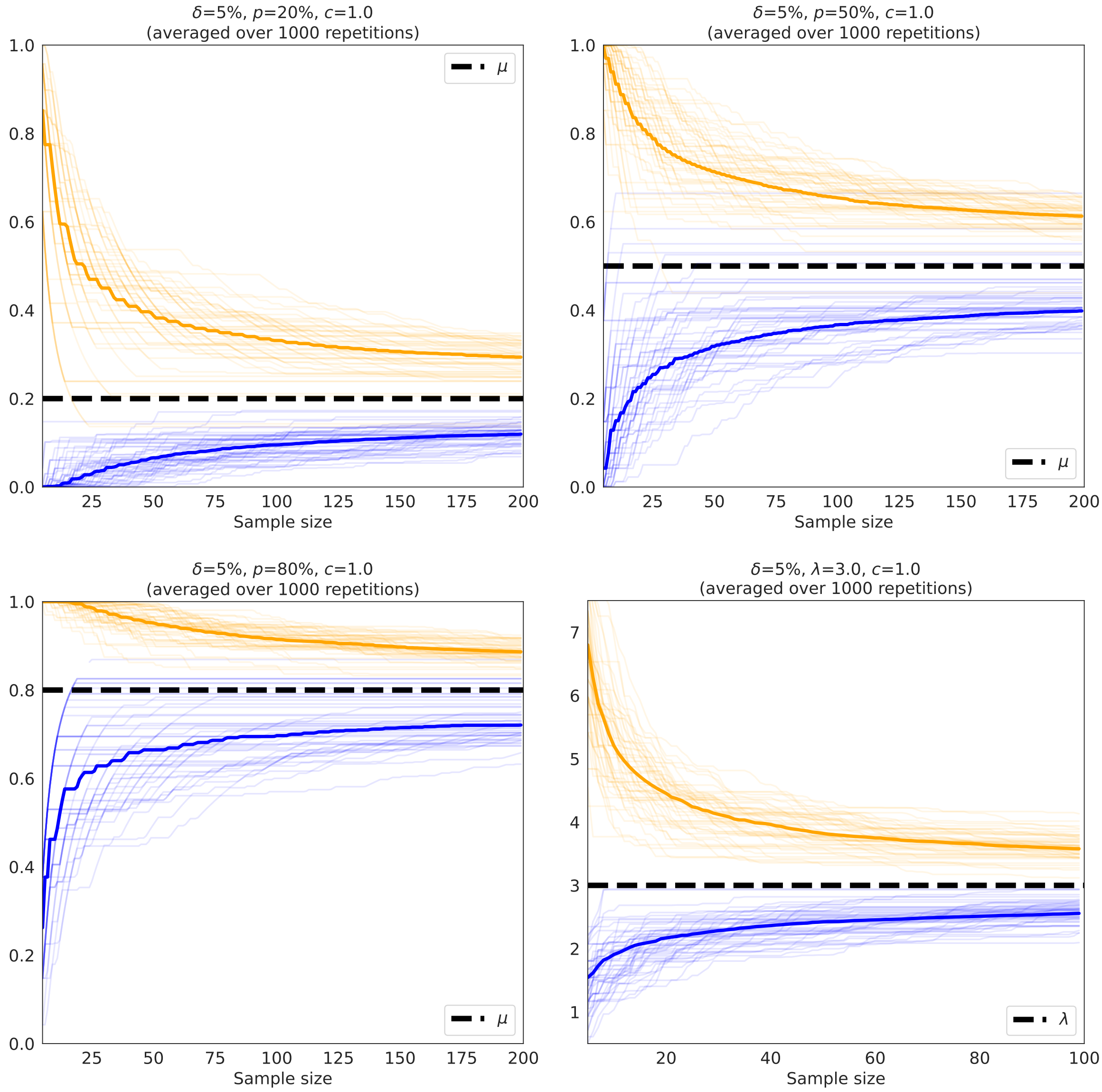

**Figure B.2** – Examples of confidence upper and lower envelopes around expectation $\mu$ for discrete distributions. From top to bottom: Bernoulli$(p)$, $p \in \{0.2, 0.5, 0.8\}$ (cf. Table (3.1)) and Poisson$(3)$ (cf. Table 3.1) on several realisations (each dashed lines) as a function of the number of observations $t$. The thick lines indicate the median curve over 1000 replicates.

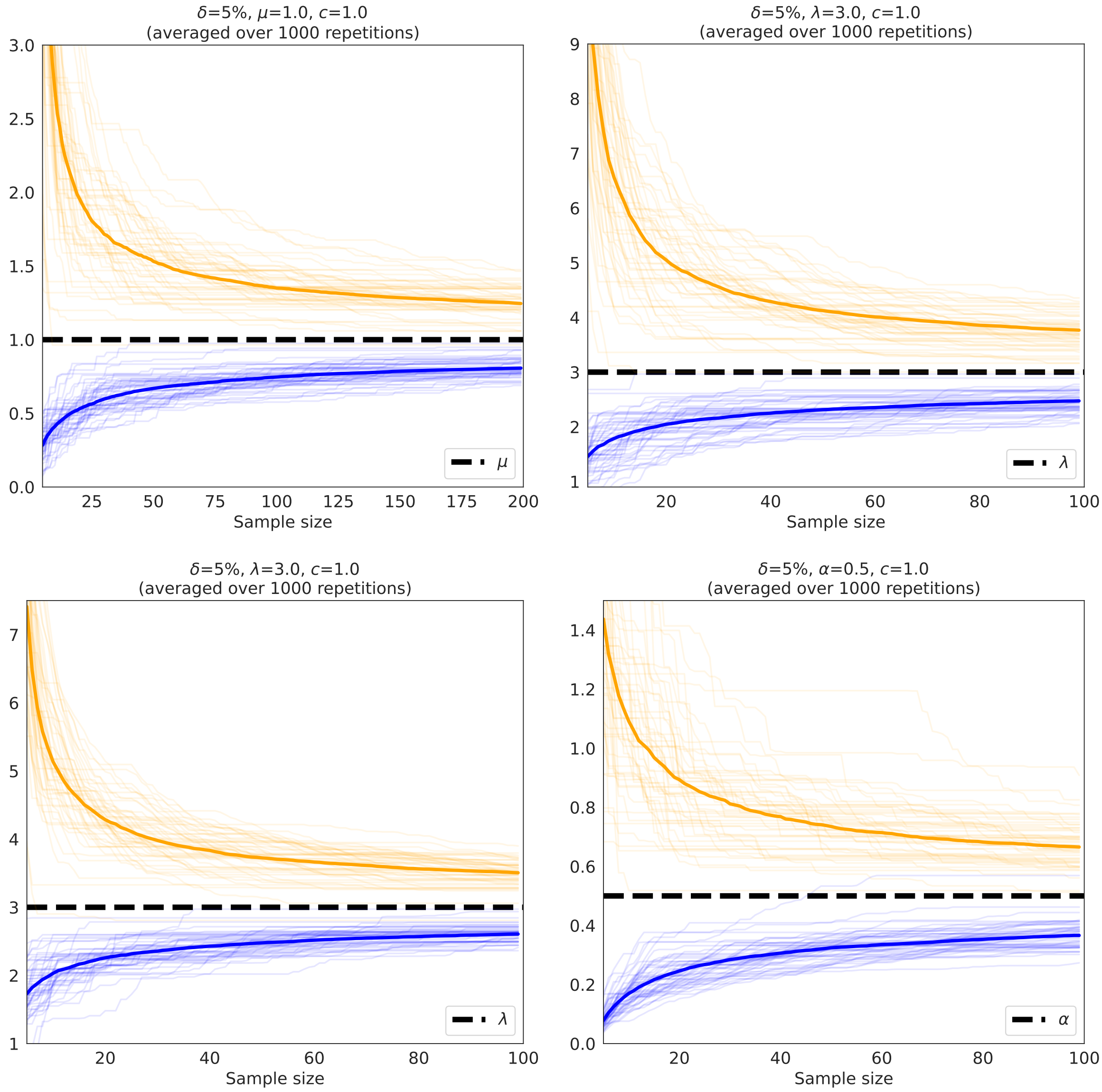

**Figure B.3 –** Examples of confidence upper and lower envelopes (from top to bottom) around scale parameter $\lambda$ for $\mathrm{Exp}(1)$ (cf. Table (3.1)), fixed shape $\mathrm{Gamma}(3, 2)$ (cf. Table 3.1), fixed shape $\mathrm{Weibull}(3, 2)$ (cf. Table 3.1), and exponent $\alpha$ for $\mathrm{Pareto}(\frac{1}{2})$ (cf. Table 3.1), on several realisations (each dashed lines) as a function of the number of observations $t$. The thick lines indicate the median curve over 1000 replicates.

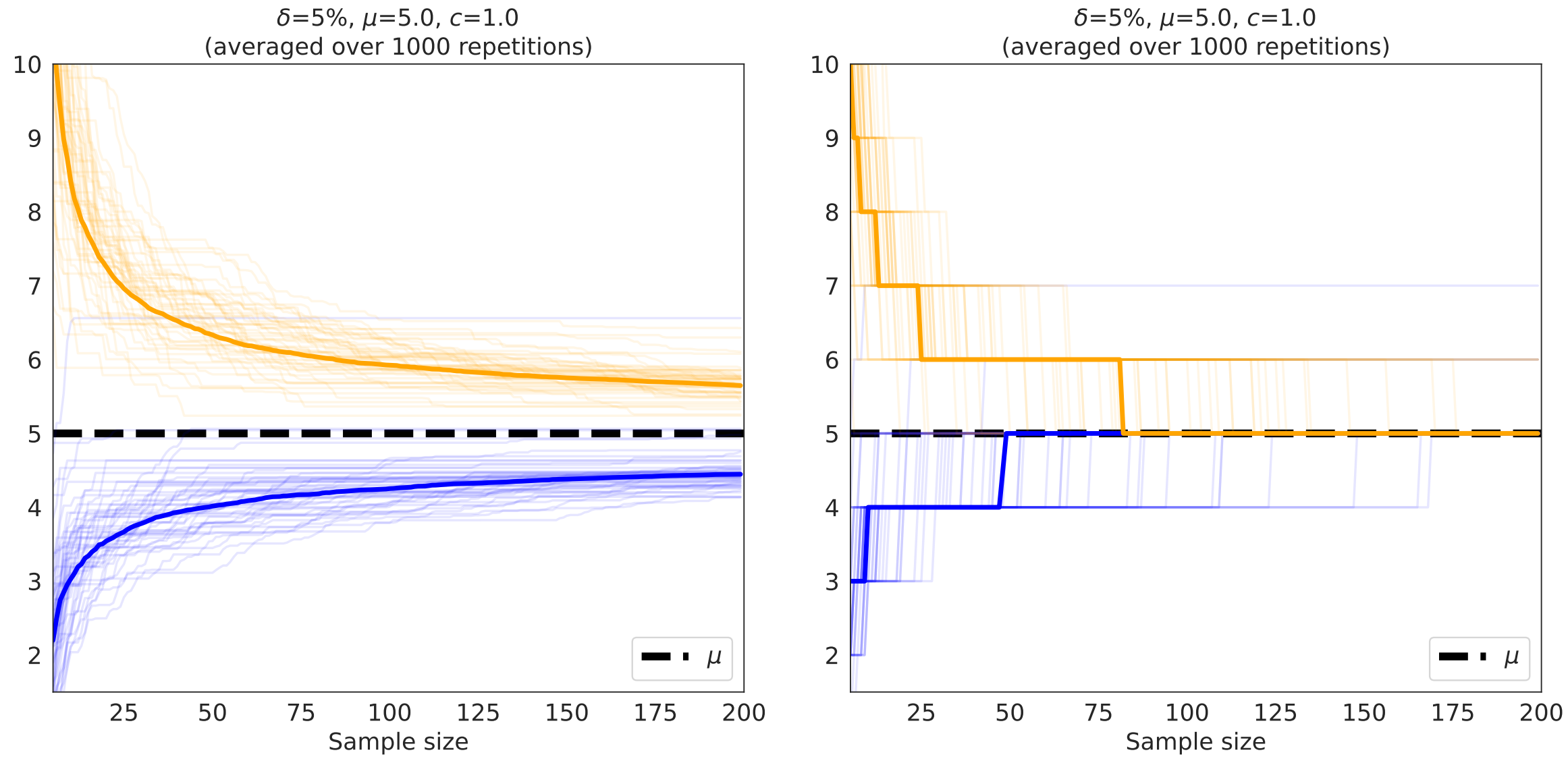

**Figure B.4** – Examples of confidence upper and lower envelopes around expectation $\mu$ for (continuous prior) Gamma $\left(\frac{k}{2}\right)$ with unknown shape $\frac{k}{2} > 0$ and (discrete prior) Chi-square $\chi^2(k)$, $k \in \mathbb{N}^*$ (cf. Table (3.1) and Remark B.3) on several realisations (each dashed lines), as a function of the number of observations $t$. In particular for the Chi-square, confidence lower and upper bounds are ceiled and floored to integers. The thick lines indicate the median curve over 1000 replicates.

## Gaussian with unknown mean and variance

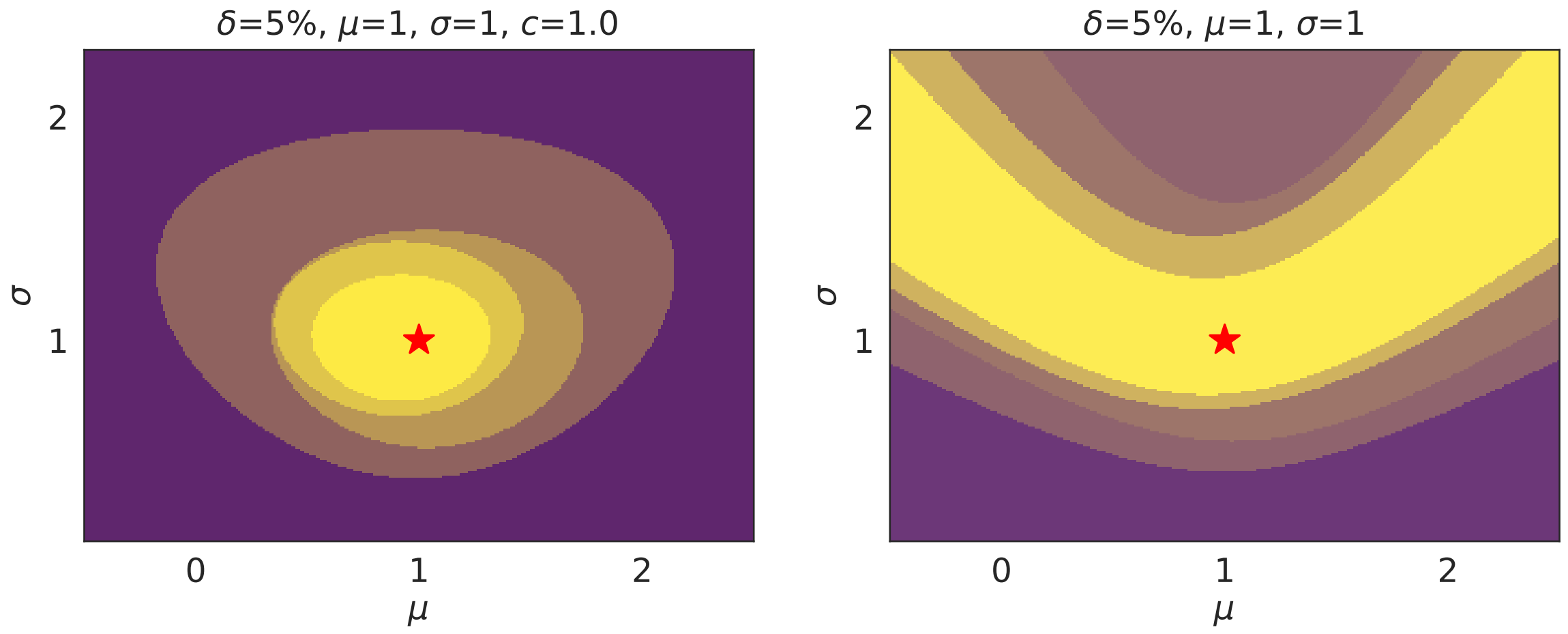

**Figure B.5** – Example of time-uniform, joint confidence sets around $(\mu, \sigma)$ for $\mathcal{N}(1,1)$ (cf. Table 3.1 for $t \in \{10, 25, 50, 100\}$ observations (smaller confidence sets correspond to larger sample sizes). The red star indicates the true parameters $(\mu, \sigma) = (1, 1)$.

We consider the two-dimensional family $\{\mathcal{N}(\mu, \sigma^2), (\mu, \sigma) \in \mathbb{R} \times \mathbb{R}_+^*\}$. To the best of our knowledge, there does not exist time-uniform joint confidence sets for $(\mu, \sigma)$ prior to this work. In order to compare ourselves against some baseline, we derive a crude one based on Chi-square quantiles and a union bound.

Let $t \in \mathbb{N}, \delta \in (0,1)$ and $(Y_s)_{s=1}^t$ i.i.d. samples drawn from $\mathcal{N}(\mu_0, \sigma_0^2)$. The standardised sum of squares

$$Z_t(\mu_0, \sigma_0) = \sum_{s=1}^{t} \left( \frac{Y_s - \mu_0}{\sigma_0} \right)^2$$

follows a $\chi^2(t)$ distribution, therefore, denoting by $q_{\chi^2(t)}$ the corresponding quantile function, we have:

$$\mathbb{P}\left( q_{\chi^2(t)}\left(\frac{\delta}{2}\right) \leqslant Z_t(\mu_0, \sigma_0) \leqslant q_{\chi^2(t)}\left(1 - \frac{\delta}{2}\right) \right) \geqslant 1 - \delta\,. \tag{B.111}$$

By a union bound argument, the intersection of $n$ such events with confidence $\frac{\delta}{t}$, i.e.

$$\widehat{\Theta}_{s,t}^Z(\delta) = \bigcap_{r=1}^{s} \left\{ (\mu, \sigma) \in \mathbb{R} \times \mathbb{R}_+^* :\ q_{\chi^2(r)}\left(\frac{\delta}{2t}\right) \leqslant Z_r(\mu, \sigma) \leqslant q_{\chi^2(r)}\left(1 - \frac{\delta}{2t}\right) \right\}, \tag{B.112}$$

describes a confidence sequence at level $\delta$ that hold uniformly over $r \in \{1, \ldots, t\}$.

We report our confidence sets (cf. equation 3.1) and the above on Figure B.5. The most striking drawback of $\widehat{\Theta}_{r,t}^Z(\delta)$ is that it is not convex nor even bounded; in particular, projecting onto the axes of $\mu$ and $\sigma$ only provides trivial confidence sets, $\mathbb{R}$ and $\mathbb{R}_+^*$ respectively, rendering this result vacuous. To better grasp this phenomenon, let us informally consider $\mu = \alpha + \mu_0$ and $\sigma = \alpha$ for some $\alpha > 0$. We have:

$$Z_t(\mu, \sigma) = \sum_{r=1}^{t} \left( \frac{Y_r - \mu_0}{\alpha} - 1 \right)^2 \tag{B.113}$$

$$= \underbrace{\frac{1}{\alpha^2} \sum_{r=1}^{t} (Y_r - \mu_0)^2}_{\approx \frac{\sigma_0^2}{\alpha^2} t} - \underbrace{\frac{2}{\alpha} \sum_{r=1}^{t} (Y_r - \mu_0)}_{\approx \frac{2}{\alpha}\sqrt{t}} + t\,. \tag{B.114}$$

Hence, for $\alpha \to +\infty$, $Z_t(\mu_0 + \alpha, \alpha) \approx t$. Moreover, Lemma 1 in Laurent and Massart (2000) shows that[4] $q_{\chi^2(r)}(1 - \delta/(2t)) \approx t + 2\sqrt{t \log 2t/\delta} + 2\log 2t\delta$ and $q_{\chi^2(r)}(\delta/(2t)) \approx t - 2\sqrt{t \log 2t/\delta}$ for $r \in \{1, \ldots, s\}$. Therefore, even with increasing the sample size $t$, there exists arbitrary large $\alpha$ such that $(\mu_0 + \alpha, \alpha)$ may belong to $\widehat{\Theta}_{s,t}^Z(\delta)$.

[4] Actually, they only show upper and lower bounds respectively using the Chernoff method, which is typically tight up to logarithmic factor here, so it does not change our conclusion regarding the unboundness of $\widehat{\Theta}_{s,t}^Z(\delta)$.

By contrast, our Bregman confidence sets are convex and bounded, which we interpret as the result of exploiting the true geometry of the two-dimensional Gaussian family. In addition to the unboundedness, $\widehat{\Theta}^{Z}_{s,t}(\delta)$ is built using a crude union bound, which is not anytime (depends on the terminal time $t$) and rather loose.

### Bernoulli

We consider the Bernoulli distribution $\mathcal{B}(\mu)$ for some unknown $\mu \in [0,1]$. Confidence bounds are displayed in Figure B.6 for $\mu \in \{0.2, 0.5, 0.8\}$. The sub-Gaussian mixture time-uniform bound is extracted from Corollary 1.21 with mixing parameter equal to the local regularisation parameter $c$ and sub-Gaussian parameter $R = 1/2$ (Hoeffding's lemma), i.e.

$$\widehat{\Theta}^{\delta}_{t,c} = \left[\widehat{\mu}_{t,c} \pm R\sqrt{\frac{2}{t}\left(1+\frac{c}{t}\right)\log\frac{2\sqrt{1+\frac{t}{c}}}{\delta}}\right] . \tag{B.115}$$

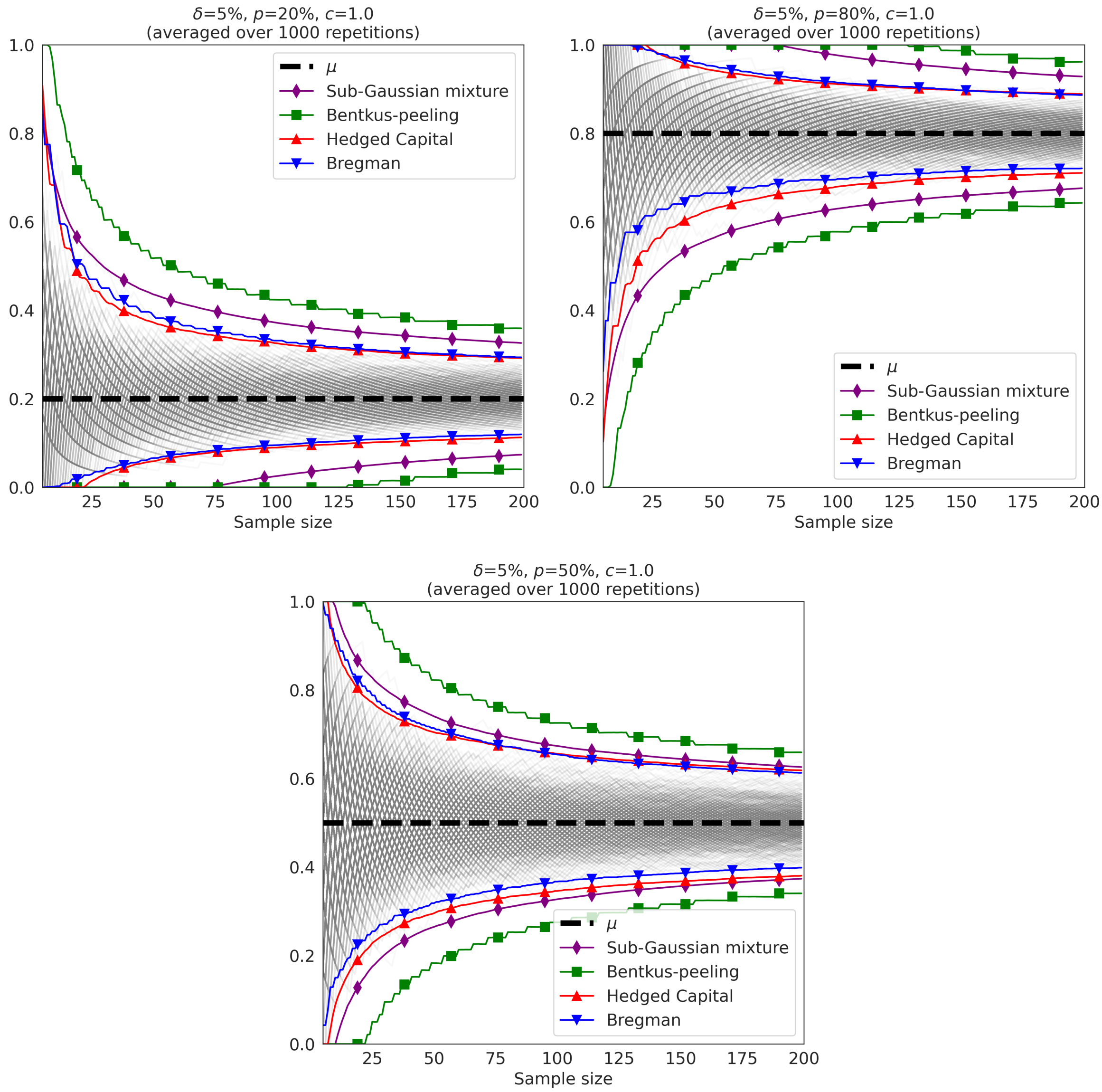

**Figure B.6 –** Confidence upper and lower envelopes built for $\mathcal{B}(\mu)$, $\mu \in \{0.2, 0.5, 0.8\}$, as a function of the number of observations $t$, averaged over $1000$ independent simulations. Grey lines are trajectories of empirical means $\widehat{\mu}_t$. Confidence bounds are clipped between $0$ and $1$.

## Gaussian

We consider Gaussian distributions $\mathcal{N}(\mu, \sigma^2)$ for some unknown $\mu \in \mathbb{R}$ and known variance $\sigma^2$. The confidence bounds are displayed in Figure B.7 for $\mu = 0$.

**Kaufmann-Koolen.** Kaufmann and Koolen (2021) introduces a martingale construction for exponential families to derive time-uniform deviation inequalities under bandit sampling.

However, application of their result is limited to Gaussian distribution with known variance and Gamma distribution with known shape, which is just a scaled version of exponential distribution. Restricting Corollary 10 of Kaufmann and Koolen (2021) to the case of a single arm yields $\mathbb{P}\left(\forall t \in \mathbb{N}, \mu \in \widehat{\Theta}_t^{\delta,\mathrm{KK}}\right) \geqslant 1-\delta$, with

$$\widehat{\Theta}_t^{\delta,\mathrm{KK}} = \left\{\mathcal{B}_{\mathcal{L}}\left(\widehat{\mu}_n, \mu\right) \leqslant \frac{2}{n}\log\left(4+\log n\right) + \frac{1}{n}C^g\left(\log 1/\delta\right)\right\}, \tag{B.116}$$

$$g\colon \lambda \in (1/2, 1] \mapsto 2\lambda\left(1-\log(4\lambda)\right) + \log\zeta(2\lambda) - \frac{1}{2}\log(1-\lambda), \tag{B.117}$$

$$C^g\colon x \in (0, +\infty) \mapsto \min_{\lambda\in(1/2,1]} \frac{g(\lambda)+x}{\lambda}, \tag{B.118}$$

$$\mathcal{L}(\theta) = \frac{\theta^2}{2\sigma^2} \quad \text{(log-partition function of } \mathcal{N}(\cdot, \sigma^2)\text{)}\,. \tag{B.119}$$

The sub-Gaussian mixture is again extracted from Corollary 1.21 with sub-Gaussian parameter equal to the (known) variance of the Gaussian distribution, i.e. $R = \sigma$.

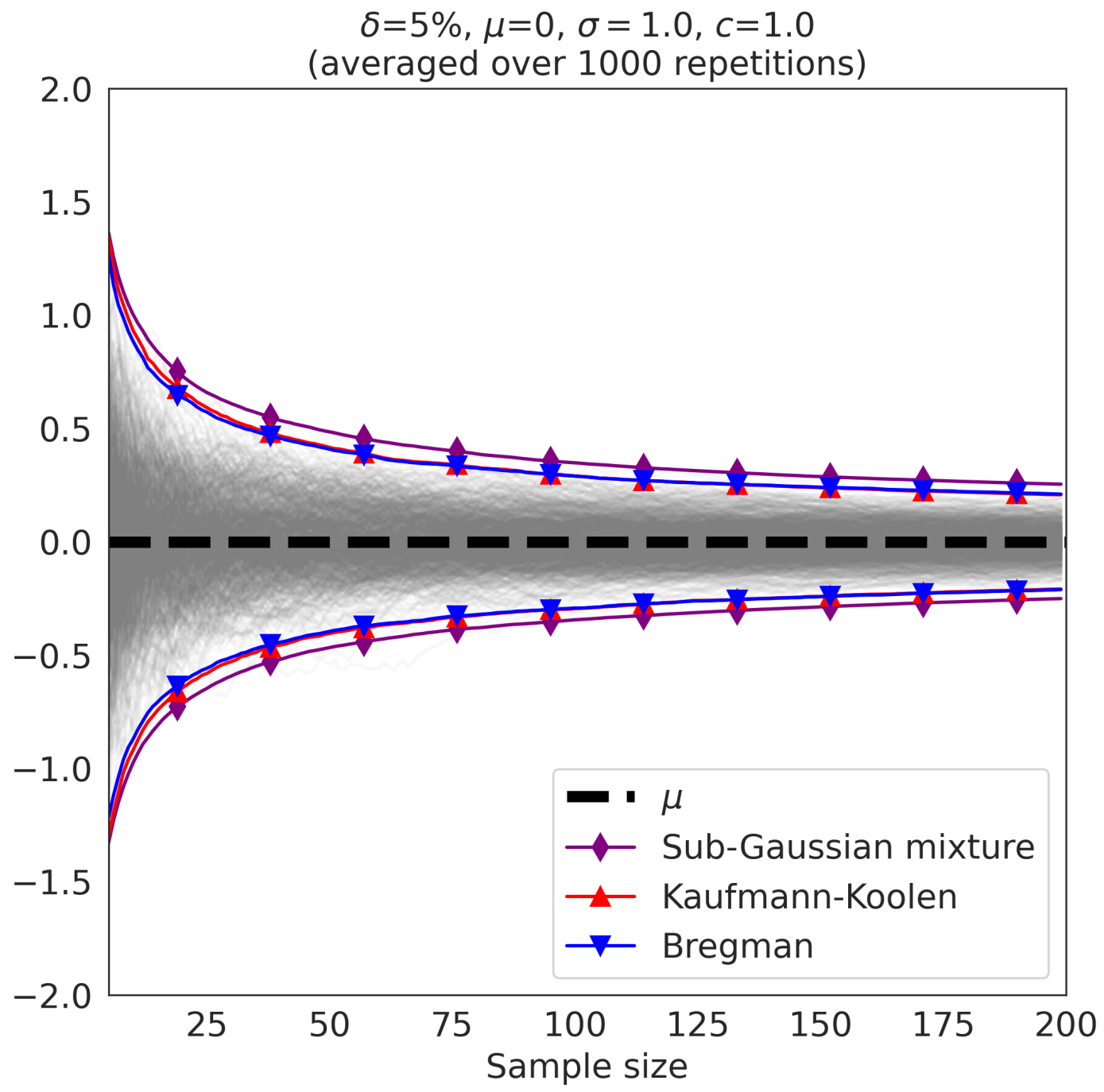

**Figure B.7** – Confidence upper and lower envelopes built for Gaussian distributions $\mathcal{N}(0, 1)$ as a function of the number of observations $t$, averaged over 1000 independent simulations. Grey lines are trajectories of empirical means $\widehat{\mu}_t$.

## Exponential

We now consider exponential distributions $\mathcal{E}(1/\mu)$ for some unknown mean $\mu \in \mathbb{R}$. We report in Figure B.8 the confidence bounds for the case when $\mu = 1$.

**Kaufmann-Koolen.** Again, Kaufmann and Koolen (2021, Corollary 12) show the similar time-uniform bound $\mathbb{P}\left(\forall t \in \mathbb{N}, \mu \in \widehat{\Theta}_t^{\delta,\mathrm{KK}}\right) \geqslant 1 - \delta$, with the same definition as for the Gaussian case except

$$g\colon \lambda \in (1/2, 1] \mapsto 2\lambda\,(1 - \log(4\lambda)) + \log\zeta(2\lambda) - \log(1-\lambda), \tag{B.120}$$

$$\mathcal{L}(\theta) = \log(-1/\mu) \quad (\text{log-partition function of } \mathcal{E}(1/\mu)). \tag{B.121}$$

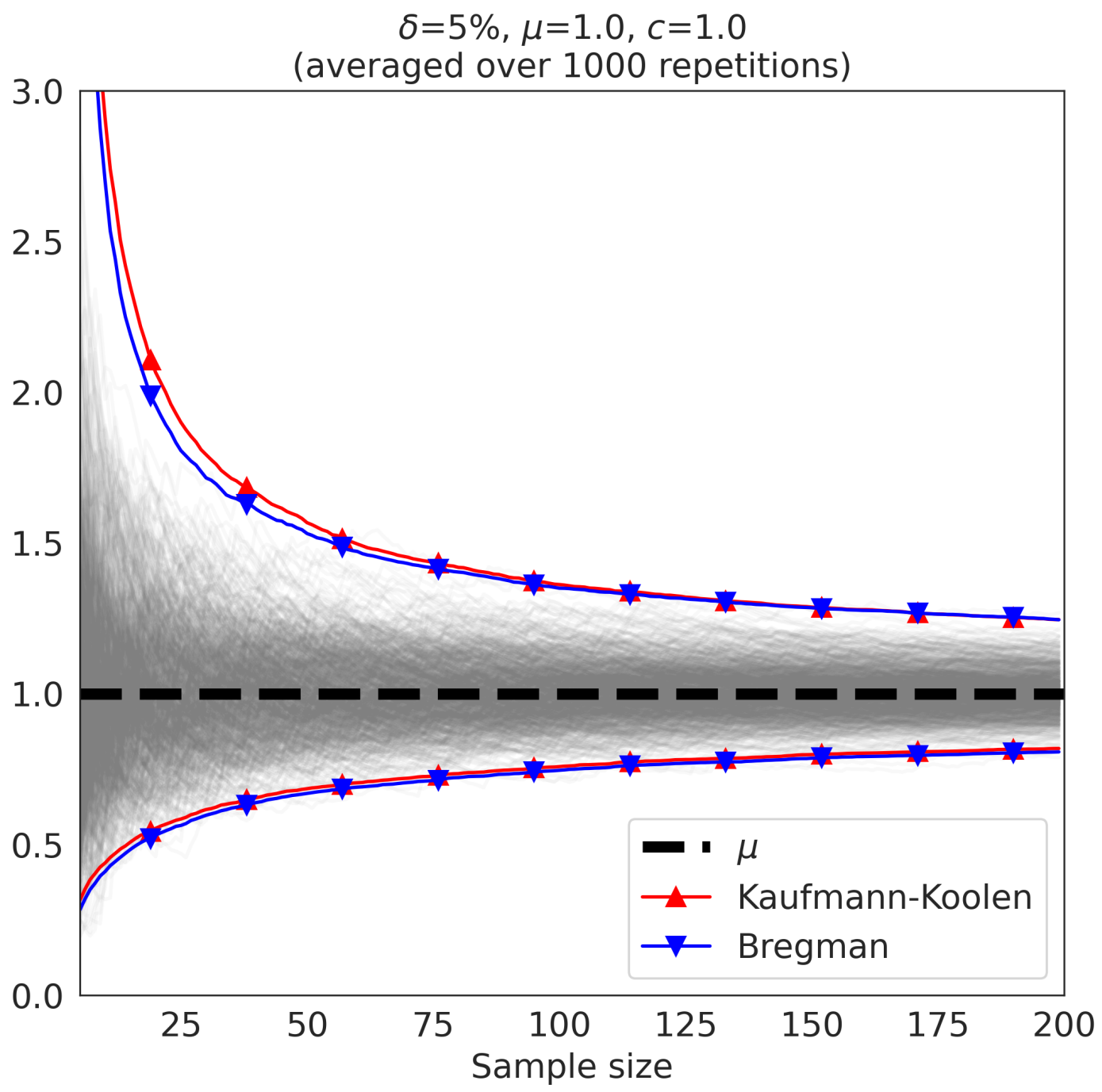

**Figure B.8** – Confidence upper and lower envelopes built for Exponential distributions $\mathcal{E}(1)$ as a function of the number of observations $t$, averaged over 1000 independent simulations. Grey lines are trajectories of empirical means $\widehat{\mu}_t$.

## Chi-square

We consider Gamma $\left(\frac{k}{2}, \frac{1}{2}\right)$ and $\chi^2(k)$ distributions for some unknown $k > 0$ and $k \in \mathbb{N}^*$ respectively. They are in fact the same distributions, however we distinguish both as the restriction on the domain for $k$ (real or integer) bears two consequences:

- the mixture distribution in the martingale construction is either continuous or discrete, resulting in integrals or sums in the expression of the confidence sequence (equation 3.1);
- confidence bounds are constrained to be integers in the discrete case, thus ceiling and flooring the lower and upper bounds respectively.

We find that the former is negligible as both the continuous and discrete mixtures yield the same bound within numerical precision. The latter however allows to drastically shrink the size of the confidence sequences, resulting in perfect identification the mean within $95\%$ confidence with less than $100$ observations in half the simulations (see Figure B.9).

### Key observations

On the studied examples, confidence intervals based on time-uniform Bregman concentration are either comparable with state-of-the art methods for the corresponding setting or result in sharper bounds, especially for small sample sizes (lower bounds for Bernoulli compared to the hedged capital process, upper bound for Exponential compared to Kaufmann-Koolen). In particular, due to the formulation in terms of Bregman divergence, our intervals are naturally asymmetric when the underlying distribution is, and respect the support constraints (for instance Bernoulli Bregman bounds are in $[0, 1]$ without the need for clipping). Moreover, we also provide bounds in novel settings for which, to the best of our knowledge, time-uniform confidence sets are lacking (Chi-square, Poisson, Weibull, Pareto, mean-variance for Gaussian).

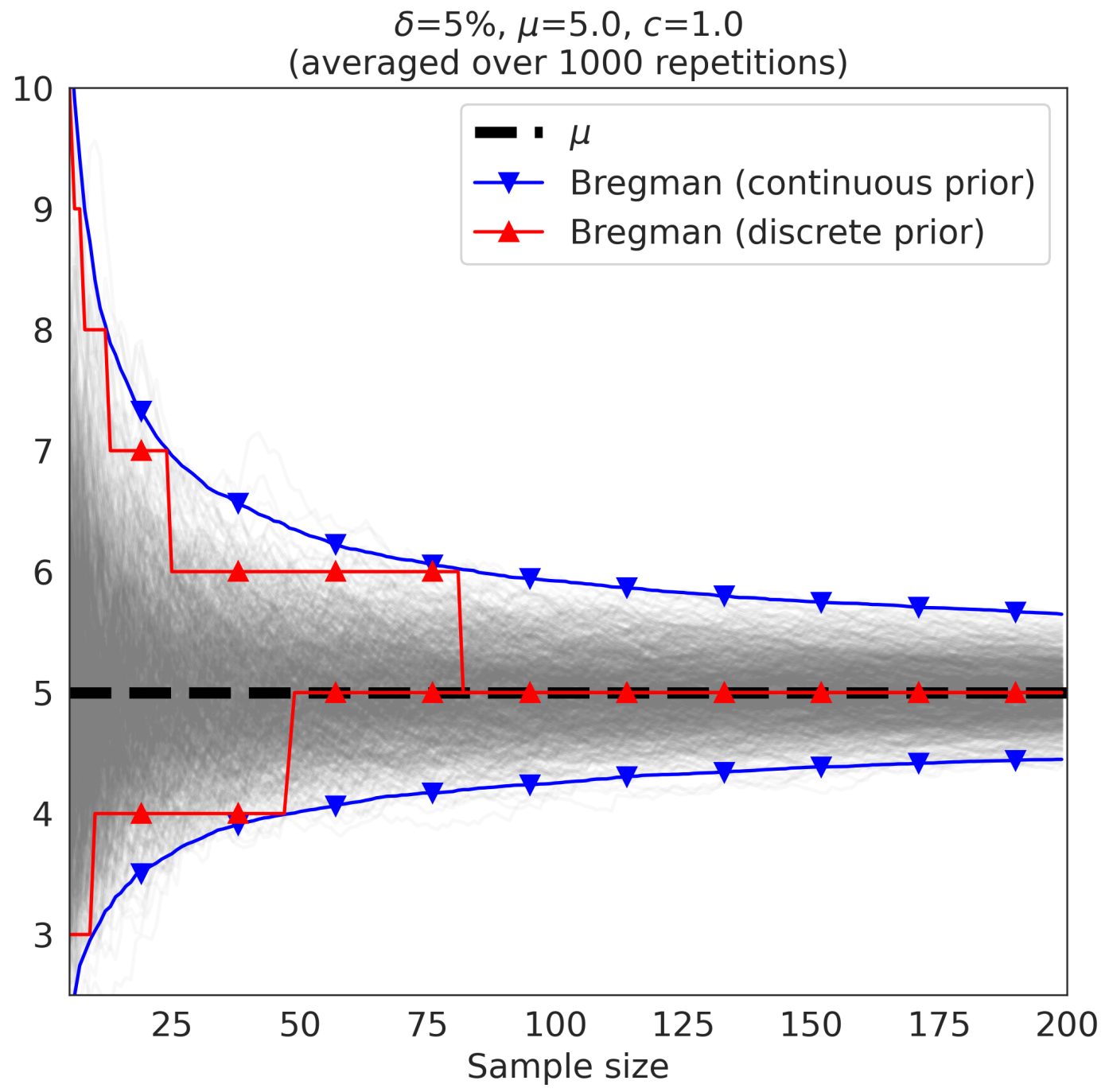

**Figure B.9 –** Confidence upper and lower envelopes built for Gamma $\left(\frac{5}{2}, \frac{1}{2}\right)$ (blue) and $\chi^2(5)$ (red) distributions as a function of the number of observations $t$, over 1000 independent simulations (colored lines are median of simulated bounds). Grey lines are trajectories of empirical means $\widehat{\mu}_t$.

## B.4 Technical details

### Proof of the pseudo regret upper bound for generic exponential families

*Proof of Proposition 3.5.* We first consider a time-uniform confidence sequence at level $\delta$ for $\eta_k = \mathbb{E}_{\theta_k}[F(Y)] = \nabla\mathcal{L}(\theta_k)$ defined as $\widehat{\mathcal{E}}^{\delta}_{k,n,c} = \nabla\mathcal{L}(\widehat{\Theta}^{\delta}_{k,n,c})$, i.e.

$$\widehat{\mathcal{E}}^{\delta}_{k,n,c} = \left\{\eta_0 \in \nabla\mathcal{L}(\Theta),\ (n+c)\mathcal{B}_{\mathcal{L}}\left(\nabla\mathcal{L}^{-1}(\eta_0), \nabla\mathcal{L}^{-1}(\widehat{\eta}_{k,n,c}(\eta_0))\right) \leqslant \log\frac{1}{\delta} + \gamma_{k,n,c}(\nabla\mathcal{L}^{-1}(\eta_0))\right\}, \tag{B.122}$$

where $\widehat{\eta}_{k,n,c}(\eta_0) = \nabla\mathcal{L}(\widehat{\theta}_{k,n,c}(\nabla\mathcal{L}^{-1}(\eta_0)))$ for $n \in \mathbb{N}$ and $\eta_0 \in \nabla\mathcal{L}(\Theta)$. When $n \to +\infty$, the asymptotic behaviour of $\widehat{\Theta}^{\delta}_{k,n,c}$ shows that with probability at least $1-\delta$, $\theta_k \in \bar{\Theta}^{\delta}_{k,n,c}$, where

$$\bar{\Theta}^{\delta}_{k,n,c} = \left\{\theta_0 \in \Theta,\ \|\theta_0 - \widehat{\theta}_{k,n,c}(\theta_0)\|^2_{\nabla^2\mathcal{L}(\widehat{\theta}_{n,c}(\theta_0))} \leqslant \frac{2}{n+c}\log\frac{(1+\frac{n}{c})^{\frac{d}{2}}}{\delta} + \mathcal{O}\left(\frac{1}{n}\right)\right\}. \tag{B.123}$$

For $\theta_0, \theta \in \Theta$, we let $\eta_0 = \nabla\mathcal{L}(\theta_0)$ and $\eta = \nabla\mathcal{L}(\theta)$, and we note that $\eta_0 - \eta = \overline{\nabla^2\mathcal{L}}(\theta_0, \theta)(\theta_0 - \theta)$, where $\overline{\nabla^2\mathcal{L}}(\theta_0, \theta) = \int_0^1 \nabla^2\mathcal{L}(u\theta_0 + (1-u)\theta)du$ is the average of the Hessian matrices of $\mathcal{L}$ along the segment $[\theta, \theta_0]$. Now, we define the following two sets:

$$
\begin{aligned}
\bar{\mathcal{E}}^{\delta}_{k,n,c} &= \nabla \mathcal{L}\left(\bar{\Theta}^{\delta}_{k,n,c}\right) \\
&= \Bigg\{ \eta_0 \in \nabla \mathcal{L}\left(\Theta\right), \ \|\eta_0 - \widehat{\eta}_{k,n,c}(\eta_0)\|^2_{\overline{\nabla^2 \mathcal{L}}(\theta_0, \widehat{\theta}_{k,n,c}(\theta_0))^{-1} \nabla^2 \mathcal{L}(\widehat{\theta}_{k,n,c}(\theta_0)) \overline{\nabla^2 \mathcal{L}}(\theta_0, \widehat{\theta}_{k,n,c}(\theta_0))^{-1}} \\
&\qquad\qquad \leqslant \frac{2}{n+c} \log \frac{(1+\frac{n}{c})^{\frac{d}{2}}}{\delta} + \mathcal{O}\left(\frac{1}{n}\right) \Bigg\},
\end{aligned}
\tag{B.124}
$$

and

$$
\widetilde{\mathcal{E}}^{\delta}_{k,n,c} = \left\{ \eta_0 \in \nabla \mathcal{L}\left(\Theta\right), \|\eta_0 - \widehat{\eta}_{k,n,c}(\eta_0)\|_2 \leqslant \frac{M}{\sqrt{m}} \sqrt{\frac{2}{n+c} \log \frac{(1+\frac{n}{c})^{\frac{d}{2}}}{\delta}} + \mathcal{O}\left(\frac{M}{\sqrt{mn}}\right) \right\}.
\tag{B.125}
$$

For all $n$ large enough, by the curvature assumption on $\mathcal{L}$, we have that $\widehat{\mathcal{E}}^{\delta}_{k,n,c} \subseteq \widetilde{\mathcal{E}}^{\delta}_{k,n,c}$. We now define confidence sequences for $\rho(\theta_k) = \varphi(\eta_k)$ as $\widehat{\mathcal{C}}^{\delta}_{k,n,c} = \varphi(\widehat{\mathcal{E}}^{\delta}_{k,n,c})$ and $\widetilde{\mathcal{C}}^{\delta}_{k,n,c} = \varphi(\widetilde{\mathcal{E}}^{\delta}_{k,n,c})$, which also satisfy the inclusion $\widehat{\mathcal{C}}^{\delta}_{k,n,c} \subseteq \widetilde{\mathcal{C}}^{\delta}_{k,n,c}$. In order to use the generic UCB analysis of Theorem 1.33, we note that the sample size $n$ required to observe $|\widehat{\mathcal{C}}^{\delta}_{k,n,c}| < \Delta_k$ is at most the sample size required for $|\widetilde{\mathcal{C}}^{\delta}_{k,n,c}| < \Delta_k$. Note that the Lipschitz assumption on $\varphi$ shows that $|\widetilde{\mathcal{C}}^{\delta}_{k,n,c}| \leqslant L|\widetilde{\mathcal{E}}^{\delta}_{k,n,c}|$. Furthermore, the UCB induced by $(\widetilde{\mathcal{C}}^{\delta}_{k,n,c})_{n\in\mathbb{N}}$ is equivalent to the $R$-sub-Gaussian setting of Corollary 1.21 with $R = LM/\sqrt{m}$, up to $\mathcal{O}(M/\sqrt{mn})$, which only affects the higher order terms of the regret bound (using a similar reasoning on the Lambert $W$ function as in the proof of this corollary). ■

### Computing the inflating factor $g(t)$ in Theorem 3.8

The right-hand side of the doubly time-uniform confidence set of Theorem 3.8 involves a $\log g(t)/\delta$ term instead of $\log 1/\delta$ as in Theorem 3.3, which is a byproduct of the peeling-like argument used in the proof. Ideally, $g(t)$ should grow as slow as possible with $t$ to avoid unnecessary looseness in the confidence bound. That is however limited by the constraint $\sum_{t=1}^{\infty} 1/g(t) \leqslant 1$, which prohibits the use of a linearly growing $g(t)$ (since the harmonic series $\sum_{t=1}^{\infty} 1/t$ diverges). The choice $g(t) = \kappa(1+t)\log^{1+\eta}(1+t)$ for some $\eta > 0$ and $\kappa > 0$ guarantees that the series of inverses converges, although its limit is not available in closed-form. However, it is sufficient to set $\kappa$ to an upper bound on $\sum_{t=1}^{\infty} \frac{1}{(1+t)\log^{1+\eta}(1+t)}$. The following elementary lemma explains how to compute a tight value for $\kappa$.

**Lemma B.6.** *Let* $f : t \in \mathbb{N} \mapsto (1+t)\log^{1+\eta}(1+t)$ *and define* $S_p = \sum_{t=1}^{p} 1/f(t)$ *for* $p \in \mathbb{N} \cup \{+\infty\}$. *Then* $S_\infty \leqslant S_p + \frac{1}{\eta \log(1+p)^\eta}$.

*Proof of Lemma B.6.* The function $f$ is nondecreasing and positive, therefore for $t \in \mathbb{N}$ and $x \in [t, t+1)$, it holds that $1/f(t+1) \leqslant 1/f(x)$. Integrating both terms as functions of $x$ over the interval $[t, t+1)$ yields $1/f(t+1) \leqslant \int_t^{t+1} 1/f(x)dx$. Summing starting at $t = p$, we further get the following sum-integral comparison:

$$S_\infty - S_p \leqslant \int_p^\infty \frac{dx}{f(x)} .$$

Finally, note that $\frac{d}{dx}\frac{1}{\eta \log^\eta(1+x)} = -\frac{1}{f(x)}$, and thus $\int_p^\infty \frac{dx}{f(x)} = \frac{1}{\eta \log(1+p)^\eta}$. ∎

This lemma shows that $\kappa = S_p + 1/\log(1+p)$ for some $p \geqslant 1$ ($\eta = 1$) is a valid choice for the definition of the inflating factor $g(t)$, and is straightforward to compute numerically. For $p = 100$, we obtain $\kappa \lesssim 2.10974$ and increasing $p$ only changes further digits.

# Appendix C

# Empirical Chernoff concentration: beyond bounded distributions

## Contents

## C.1 Examples of second order sub-Gaussian distributions

In this appendix, we show that several classical distributions are second order sub-Gaussian, with explicit ratio $\rho$.

*Proof of Proposition 4.5.* As a general remark, we first note that the family of second order sub-Gaussian distributions is stable under translation and scaling, i.e. if $\nu \in \mathcal{F}^2_{\mathcal{G},\rho,R}$, then for $m \in \mathbb{R}$ and $c \in \mathbb{R}^\star_+$, $c(\nu + m) \in \mathcal{F}^2_{\mathcal{G},\rho,cR}$.

**Gaussian.** This case corresponds to Lemma 4.2.

**Bernoulli.** Let $\nu = p\delta_q + q\delta_{-p}$ and assume $p < q = 1 - p$. By translation and scaling, if this distribution is second order sub-Gaussian, then so is any distribution supported on two points. $\nu$ is known to be $R$-sub-Gaussian with optimal parameter $R_p = \sqrt{\frac{1/2-p}{\log(q/p)}}$, which we extend by continuity as $R_{1/2} = 1/4$ (see Kearns and Saul (1998); Berend and Kontorovich (2013); Raginsky et al. (2013)). It is also straightforward to note that $R$ is equal to the standard deviation $\sqrt{pq}$ if and only if $\nu$ is symmetric, i.e. $p = q = 1/2$.

For $\lambda \in \mathbb{R}_-^\star$, using the fact that $q^2 - p^2 = 1 - 2p + p^2 - p^2 = q - p$, we obtain

$$\mathbb{E}_{Y\sim\nu}[e^{\lambda Y^2}] = pe^{\lambda q^2} + qe^{\lambda p^2} = e^{\lambda p^2}\left(1 - p(1 - e^{\lambda(q-p)})\right) . \tag{C.1}$$

Let $\rho \in (0, 1]$. The left tail CGF control of Definition 4.1 is equivalent to $\forall \lambda \in \mathbb{R}_-^\star,\ f(\lambda) \leqslant 0$, where $f$ is defined by taking the logarithm in the above expression:

$$f(\lambda) = \lambda p^2 + \log\left(1 - p(1 - e^{\lambda(q-p)})\right) + \frac{1}{2}\log\left(1 - 2\lambda(\rho R_p)^2\right) . \tag{C.2}$$

Using the inequality $\forall x \in (-1, +\infty),\ \log(1 + x) \leqslant x$, we further get:

$$f(\lambda) \leqslant g(\lambda) \quad \text{and} \quad g(\lambda) = -\lambda((\rho R_p)^2 - p^2) - p\left(1 - e^{\lambda(q-p)}\right) , \tag{C.3}$$

which defines a differentiable function with $\mathrm{d}g/\mathrm{d}\lambda(\lambda) = p^2 - (\rho R_p)^2 + p(q - p)e^{\lambda(q-p)}$. The function $\mathrm{d}g/\mathrm{d}\lambda$ is nondecreasing (since $q - p > 0$) and $\lim_{\lambda\to-\infty} \mathrm{d}g/\mathrm{d}\lambda(\lambda) = p^2 - (\rho R_p)^2$. Therefore, $g$ is nondecreasing if $\rho R_p \leqslant p$, in which case we have $\forall \lambda \in \mathbb{R}_-^\star,\ g(\lambda) \leqslant g(0) = 0$. Consequently, $\nu$ is $(\rho_p, R_p)$-second order sub-Gaussian with $\rho_p = p/R_p$.

By symmetry, the case $p > q$ leads to $(\rho R_p) \leqslant q$ and thus $\nu$ is $(\rho_p, R_p)$-second order sub-Gaussian with $\rho_p = q/R_p$.

For the critical case $p = q = \frac{1}{2}$ it is enough to notice that $g(\lambda) = \lambda((\rho R_{1/2})^2 - \frac{1}{4}) = \lambda/4(\rho^2 - 1)$, and hence $g(0) = 0$ if $\rho = 1$. In this situation, $\nu$ is $(1, 1/2)$-second order sub-Gaussian.

**Uniform.** Let $\nu(dy) = 1/2\mathbb{1}_{y\in[-1,1]}dy$ denote the uniform distribution over $[-1, 1]$. It is centred $(bE_{Y\sim\nu}[Y] = 0)$ and by scaling and translation, it is sufficient to prove that it is second order sub-Gaussian to deduce the result for all uniform distributions.

It it proved in Arbel et al. (2020) that $\nu$ is *strictly* sub-Gaussian, i.e. $\nu$ is $R$-sub-Gaussian with $R = \frac{1}{\sqrt{3}} = \sqrt{\mathbb{V}_{Y\sim\nu}[Y]}$. For the second order control, let $\lambda \in \mathbb{R}_-^\star$. We have

$$\begin{aligned}
\mathbb{E}_{Y\sim\nu}\left[e^{\lambda Y^2}\right] = \frac{1}{2}\int_{-1}^{1} e^{\lambda y^2} dy = \int_0^1 e^{\lambda y^2} dy \qquad & \text{(symmetry)} \\
= \int_0^{\sqrt{-2\lambda}} e^{-z^2/2} \frac{dz}{\sqrt{-2\lambda}} \qquad & (z = \sqrt{-2\lambda}y) \\
= \sqrt{\frac{\pi}{-\lambda}} P(\sqrt{-2\lambda}) , \qquad & \text{(C.4)}
\end{aligned}$$

where $P\colon z \in \mathbb{R}_+ \mapsto \Phi(z) - 1/2$ and $\Phi$ is the standard Gaussian c.d.f.. We use the sufficient condition (iii) of Proposition 4.4. Taking the logarithm of the above expression, we obtain:

$$\mathcal{K}_2(\lambda) = \frac{1}{2}\log \pi - \frac{1}{2}\log(-\lambda) + \log P(\sqrt{-2\lambda}). \tag{C.5}$$

Furthermore, the derivative of $\mathcal{K}_2$ is given by:

$$\frac{\mathrm{d}\mathcal{K}_2}{\mathrm{d}\lambda}(\lambda) = -\frac{1}{2\lambda} - \frac{\phi(\sqrt{-2\lambda})}{\sqrt{-2\lambda}P(\sqrt{-2\lambda})}\,, \tag{C.6}$$

where $\phi\colon z \in \mathbb{R} \mapsto \frac{1}{\sqrt{2\pi}}e^{-z^2/2}$ is the standard Gaussian p.d.f.. Therefore, letting $z = \sqrt{-2\lambda}$ for simplicity, the sufficient condition (iii) of Proposition 4.4 can be written as $f(z) \leqslant 1$, with:

$$f(z) = \frac{z^3\phi(z)}{2\pi P(z)^3}\,. \tag{C.7}$$

We note that $\mathrm{d}P/\mathrm{d}z(z) = \phi(z)$ and $\mathrm{d}\phi/\mathrm{d}z(z) = -z\phi(z)$ for $z \in \mathbb{R}_+$, and that $P(0) = 0$, and hence the Taylor expansion of $P$ around $z = 0$ is $P(z) = \frac{z}{\sqrt{2\pi}}\left(1 + \mathcal{O}\left(z^2\right)\right)$. Similarly, $\phi(z) = \frac{1}{\sqrt{2\pi}} + \mathcal{O}(z)$. Therefore, $f$ can be extended by continuity at zero with $f(0) = 1$. It is also differentiable on $(0, +\infty)$ and

$$\frac{\mathrm{d}f}{\mathrm{d}z}(z) = \frac{z^2\phi(z)}{2\pi P(z)^4}\left(\underbrace{3P(z) - z^2P(z) - 3z\phi(z)}_{=:g(x)}\right)\,, \tag{C.8}$$

$$\frac{\mathrm{d}g}{\mathrm{d}z}(z) = 2z\left(\underbrace{z\phi(z) - P(z)}_{=:h(z)}\right)\,, \tag{C.9}$$

$$\frac{\mathrm{d}h}{\mathrm{d}z}(z) = -z^2\phi(z) \leqslant 0\,. \tag{C.10}$$

Consequently, $h$ is nonincreasing with $h(z) \leqslant h(0) = 0$, $g$ is nonincreasing with $g(z) \leqslant g(0) = 0$, and finally $f$ is nonincreasing with $f(z) \leqslant f(0) = 1$, which corresponds to the sufficient condition to prove.

**Symmetric triangular.** Let $\nu(dy) = (1+y)\mathbb{1}_{y\in[-1,0]}dy + (1-y)\mathbb{1}_{y\in(0,1]}dy$ denote the symmetric triangular distribution over $[-1, 1]$. It is centred ($bE_{Y\sim\nu}[Y] = 0$) and by scaling and translation, it is sufficient to prove that it is second order sub-Gaussian to deduce the result for all symmetric triangular distributions. It it also proved in Arbel et al. (2020) that $\nu$ is strictly sub-Gaussian with parameter $R = \sqrt{\mathbb{V}_{Y\sim\nu}[Y]} = 1/\sqrt{6}$. For $\lambda \in \mathbb{R}_-^\star$, we have

$$\begin{aligned}\mathbb{E}_{Y\sim\nu}\left[e^{\lambda Y^2}\right] &= \int_{-1}^{0}(1+y)e^{\lambda y^2}dy + \int_0^1(1-y)e^{\lambda y^2}dy = 2\int_0^1 e^{\lambda y^2}dy - 2\int_0^1 ye^{\lambda y^2}dy \\ &\qquad\qquad\qquad\qquad\qquad\qquad\qquad\qquad\qquad\qquad\qquad\qquad\text{(symmetry)} \\ &= 2\sqrt{\frac{\pi}{-\lambda}}P(\sqrt{-2\lambda}) - \frac{1}{\lambda}e^{\lambda} + \frac{1}{\lambda}\,.\end{aligned} \tag{C.11}$$

This time however, the mapping $\mathcal{H}_2$ is not monotonic on $\mathbb{R}_-$ but rather increasing for large negative values of $\lambda$ and decreasing when $\lambda$ approaches zero (see Figure C.1, we omit the rather tedious calculations here). Therefore its minimum is attained on the boundary of $(-\infty, 0]$, i.e. $\inf_{\lambda\in\mathbb{R}_-} \mathcal{H}_2(\lambda) = \mathcal{H}_2(0) \wedge \lim_{\lambda\to-\infty} \mathcal{H}_2(\lambda)$. By definition, $\mathcal{H}_0(\lambda) = \sigma^2 = 1/6$, and for $\lambda \in \mathbb{R}_-^\star$ we have

$$\begin{aligned}\mathcal{H}_2(\lambda) &= \frac{1}{2\lambda}\left(1 - \left(2\sqrt{\frac{\pi}{-\lambda}}P(\sqrt{-2\lambda} - \frac{1}{\lambda}e^\lambda + \frac{1}{\lambda})\right)^{-2}\right) \\ &= \frac{1}{2\lambda} - \frac{1}{2}\left(2\sqrt{\pi}P(\sqrt{-2\lambda}) + \frac{1}{\sqrt{-\lambda}}e^\lambda - \frac{1}{\sqrt{-\lambda}}\right)^{-2} . \end{aligned} \tag{C.12}$$

Note that $\lim_{z\to+\infty} P(z) = 1/2$ and hence $\lim_{\lambda\to-\infty} \mathcal{H}_2(\lambda) = 1/(2\pi) < 1/6$. Therefore $\nu$ is $(\rho, R)$-sub-Gaussian with $R = 1/\sqrt{6}$ and $\rho = \sqrt{3/\pi} \approx 0.977$.

**Gaussian mixture.** We follow the proof technique of Chafai and Malrieu (2010). In general, for a compact subset $\mathbb{J} \subset \mathbb{R}$, we consider $(\mu_k)_{k\in\mathbb{J}} \in \mathbb{R}^{\mathbb{J}}$, $(\sigma_j)_{j\in\mathbb{J}} \in (\mathbb{R}_+^\star)^{\mathbb{J}}$ and a probability measure $\pi$ on $\mathbb{J}$. For $j \in \mathbb{J}$, we let $\nu_j = \mathcal{N}(\mu_j, \sigma_j^2)$ and denote its density (with respect to the Lebesgue measure) by $p_j$. We define the mixture distribution $\nu$ by its density $p\colon y \in \mathbb{R} \mapsto \mathbb{E}_{J\sim\pi}[p_J(y)]$, where $J$ denotes a random variable distributed as $\pi$. Assume that $\nu$ is centred, i.e. $\mathbb{E}_{J\sim\pi}[\mu_J] = 0$ (again, the general case may be recovered by translation), and that $\max_{j\in\mathbb{J}} \sigma_j < \infty$. The moment generating function of $\nu$ at $\lambda \in \mathbb{R}$ is given by

$$\mathbb{E}_{Y\sim\nu}\left[e^{\lambda Y}\right] = \mathbb{E}_{J\sim\pi}\left[\mathbb{E}_{J\sim\nu_J}\left[e^{\lambda Y} \mid J\right]\right] = \mathbb{E}_{J\sim\pi}\left[e^{\frac{\lambda^2}{2}\sigma_J^2+\lambda\mu_J}\right] \leqslant e^{\frac{\lambda^2}{2}\max_{j\in J}\sigma_j^2}\mathbb{E}_{J\sim\pi}\left[e^{\lambda\mu_j}\right] . \tag{C.13}$$

We now consider the specific case of a discrete mixture with two components, i.e. $\mathbb{J} = \{0, 1\}$ and $\pi$ is the Bernoulli distribution $\mathcal{B}(p)$ for some $p \in (0, 1)$ and $q = 1 - p$. We let $\Delta = \mu_1 - \mu_0$. The above expectation now may be rewritten as

$$\mathbb{E}_{J\sim\pi}\left[e^{\lambda\mu_J}\right] = pe^{\lambda\mu_1} + qe^{\lambda\mu_0} = pe^{q\lambda\Delta} + qe^{-p\lambda\Delta} , \tag{C.14}$$

where the last line comes from the condition $\mathbb{E}_{J\sim\pi}[\mu_J] = 0$, which translates to $\mu_0 = -p\Delta$ and $\mu_1 = q\Delta$. The two points lemma (Chafai and Malrieu, 2010, Lemma 3.3)) states that

$$\mathbb{E}_{J\sim\pi}\left[e^{\lambda\mu_J}\right] \leqslant e^{c_p\lambda^2\Delta^2} \quad \text{and} \quad c_p = \frac{q-p}{4\left(\log q - \log p\right)} , \tag{C.15}$$

which concludes the proof of the first order sub-Gaussian part.

For the second order condition, we have, for $\lambda \in \mathbb{R}_-^\star$,

$$\mathbb{E}_{Y\sim\nu_j}\left[e^{\lambda Y^2}\right] = \frac{1}{\sqrt{2\pi}\sigma_j}\int_{-\infty}^{\infty} e^{\lambda y^2 - \frac{(y-\mu_j)^2}{2\sigma_j^2}}\,dy , \tag{C.16}$$

and letting $\bar{\sigma}_j^2 = \sigma_j^2/(1-2\lambda\sigma_j^2)$, we obtain, by completing the square,

$$\mathbb{E}_{Y\sim\nu_j}\left[e^{\lambda Y^2}\right] = \frac{\bar{\sigma}_j}{\sigma_j} e^{\frac{\mu_j^2}{2\sigma_j^2}\left(\frac{\bar{\sigma}_j^2}{\sigma_j^2}-1\right)} \underbrace{\frac{1}{\sqrt{2\pi}\bar{\sigma}_j}\int_{-\infty}^{\infty} e^{-\frac{1}{2\bar{\sigma}_j^2}\left(y-\frac{\bar{\sigma}_j^2}{\sigma_j^2}\mu_j\right)^2} dy}_{=1} = \frac{e^{\frac{\lambda\mu_j^2}{1-2\lambda\sigma_j^2}}}{\sqrt{1-2\lambda\sigma_j^2}}. \tag{C.17}$$

We conclude by noting that since $\lambda < 0$, this implies $\mathbb{E}_{Y\sim\nu_j}[\exp(\lambda Y^2)] \leqslant 1/\sqrt{1+2\lambda\min_{j\in\mathbb{J}}\sigma_j^2}$. ■

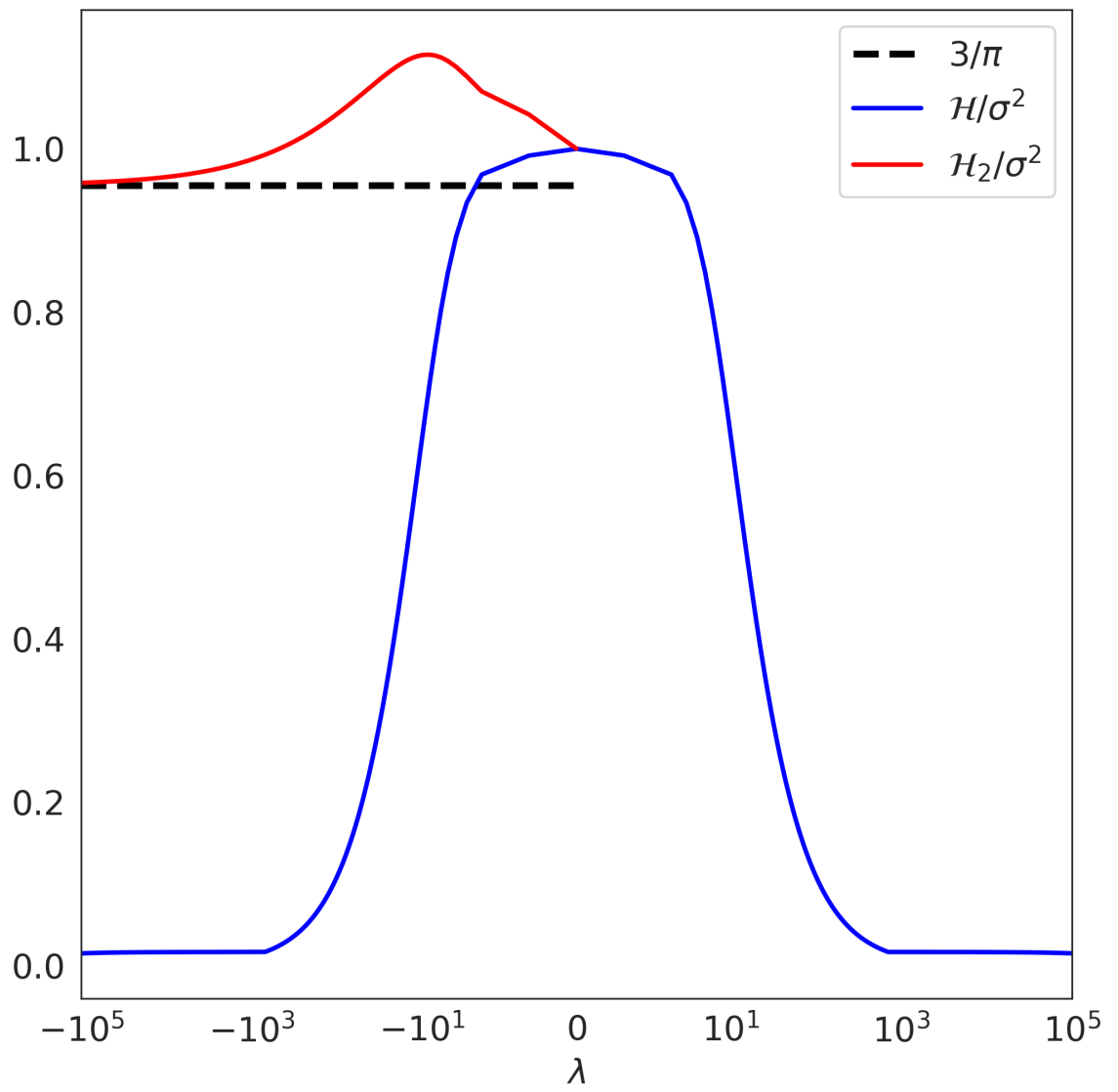

**Figure C.1** – First and second order mappings $\mathcal{H}$ and $\mathcal{H}_2$ for the symmetric triangular distribution over $[-1,1]$. The optimal first order sub-Gaussian parameter is given by $\sigma^2 = \sup_{\mathbb{R}}\mathcal{H} = 1/6$ and the optimal second order parameter is given by $(\rho\sigma)^2 = \inf_{\mathbb{R}_-}\mathcal{H}_2$ with $\rho = \sqrt{3/\pi}$.

**Conjecture on the symmetric** Beta **distributions.** In addition to the examples of Proposition 4.5, we believe that Beta distributions are also second order sub-Gaussian with $\rho = 1$ (see Conjecture 4.6). We now detail calculations and empirical evidence to support this claim, which serve as an illustration of how to check whether distributions are second order sub-Gaussian.

First, it was proved in Marchal and Arbel (2017) that symmetric Beta distributions, i.e. Beta $(\alpha,\alpha)$ for some $\alpha\in\mathbb{R}_+^\star$ are strictly sub-Gaussian, i.e. Beta $(\alpha,\alpha)\in\mathcal{F}_{\mathcal{G},\sigma}$ with parameter $\sigma^2 = \mathbb{V}_{Y\sim\text{Beta}(\alpha,\alpha)}[Y] = 1/(4(2\alpha+1))$. We now turn to the computation of the second order CGF. First, note that the expectation of Beta $(\alpha,\alpha)$ is $1/2$ and the even central moments of are given, for $n\in\mathbb{N}$, by (Arbel et al., 2020):

$$\mathbb{E}_{Y\sim\text{Beta}(\alpha,\alpha)}\left[(Y-\frac{1}{2})^{2n}\right] = \frac{\Gamma(2n+1)}{2^{2n}\Gamma(n+1)}\frac{\Gamma(2\alpha)\Gamma(\alpha+n)}{\Gamma(\alpha)\Gamma(2(\alpha+n))}, \tag{C.18}$$

and hence the power series expansion of the second order MGF is given, for $\lambda \in \mathbb{R}_-$ by

$$\begin{aligned}
\mathbb{E}_{Y\sim \mathrm{Beta}(\alpha,\alpha)}\left[e^{\lambda(Y-\frac{1}{2})^2}\right] &= \sum_{n=0}^{+\infty} \frac{\Gamma(2n+1)}{2^{2n}\Gamma(n+1)} \frac{\Gamma(2\alpha)\Gamma(\alpha+n)}{\Gamma(\alpha)\Gamma(2(\alpha+n))} \frac{\lambda^n}{n!} \\
&= \frac{\Gamma(2\alpha)}{\Gamma(\alpha)} \sum_{n=0}^{+\infty} \frac{\Gamma(n+\frac{1}{2})}{\Gamma(\alpha+n+\frac{1}{2})} 2^{1-2\alpha-2n} \frac{\lambda^n}{n!} \quad \text{(duplication formula for } \Gamma) \\
&= \underbrace{\frac{\Gamma(2\alpha)\sqrt{\pi}}{\Gamma(\alpha)\Gamma(\alpha+\frac{1}{2})} 2^{1-\alpha}}_{=1} \sum_{n=0}^{+\infty} \frac{(\frac{1}{2})_n}{(\alpha+\frac{1}{2})_n} \frac{1}{n!} \left(\frac{\lambda}{4}\right)^n , \qquad \text{(C.19)}
\end{aligned}$$

where we used the Pochhammer symbol $(z)_n = \Gamma(z+n)/\Gamma(z)$ for $z \in \mathbb{R}_+^\star$ and $n \in \mathbb{N}$ and Legendre's duplication formula $\Gamma(z)\Gamma(z+1/2) = 2^{1-2z}\sqrt{\pi}\Gamma(2z)$. Interestingly, the series of the right-hand side defines the confluent hypergeometric function of the first kind, known as Kummer's function (see Definition 4.9). Therefore, for all $\lambda \in \mathbb{R}_-$, we have

$$\mathcal{K}_2(\lambda) = \log M\left(\frac{1}{2}, \frac{1}{2}+\alpha; \frac{\lambda}{4}\right) , \qquad \text{(C.20)}$$

from which we deduce the form of the $\mathcal{H}_2$ mapping of Proposition 4.4:

$$\mathcal{H}_2(\lambda) = \frac{1 - M\left(\frac{1}{2}, \frac{1}{2}+\alpha; \frac{\lambda}{4}\right)^{-2}}{2\lambda} . \qquad \text{(C.21)}$$

Similarly, Kummer's function is related to the CGF of $\mathrm{Beta}\,(\alpha,\alpha)$, in the sense that for all $\lambda \in \mathbb{R}$ we have

$$\mathcal{K}(\lambda) = -\frac{\lambda}{2} + \log M\left(\alpha, 2\alpha; \lambda\right) \quad \text{and} \quad \mathcal{H}(\lambda) = -\frac{1}{\lambda} + 2\frac{\log M\left(\alpha, 2\alpha; \lambda\right)}{\lambda^2} . \qquad \text{(C.22)}$$

Although seemingly complicated, these functions is easily implementable (the function $M$ is available e.g. in Python as *scipy.special.hyp1f1*). We report an example of $\mathcal{H}$ and $\mathcal{H}_2$ in Figure C.2. Of note, the mapping $\mathcal{H}_2$ seems nonincreasing, which according to Proposition 4.4 is sufficient to ensure that $\mathrm{Beta}\,(\alpha,\alpha)$ is second order sub-Gaussian with $\rho = 1$.

To further illustrate this, we note that $\mathcal{K}_2$ is differentiable and satisfies (using the recurrence formulas of Abramowitz and Stegun (1968, 13.4)), the expression

$$\frac{\mathrm{d}\mathcal{K}_2}{\mathrm{d}\lambda}(\lambda) = \sigma^2 \frac{M\left(\frac{3}{2}, \frac{3}{2}+\alpha; \frac{\lambda}{4}\right)}{M\left(\frac{1}{2}, \frac{1}{2}+\alpha; \frac{\lambda}{4}\right)} . \qquad \text{(C.23)}$$

We now define the mapping $f$ as

$$f(\lambda) = \left(1 + 2\lambda \frac{\mathrm{d}\mathcal{K}_2}{\mathrm{d}\lambda}(\lambda)\right) e^{-2\mathcal{K}_2(\lambda)}$$

$$= \left(1 + 2\lambda\sigma^2 \frac{M\left(\frac{3}{2}, \frac{3}{2}+\alpha; \frac{\lambda}{4}\right)}{M\left(\frac{1}{2}, \frac{1}{2}+\alpha; \frac{\lambda}{4}\right)}\right) \frac{1}{M\left(\frac{1}{2}, \frac{1}{2}+\alpha; \frac{\lambda}{4}\right)^2}. \tag{C.24}$$

The sufficient condition (iii) in Proposition 4.4 states that if $f(\lambda) \leqslant 1$ for all $\lambda \in \mathbb{R}_-^\star$ then $\mathcal{H}_2$ is nonincreasing, implying that $\mathrm{Beta}\,(\alpha, \alpha)$ is second order sub-Gaussian with $\rho = 1$. We also report in Figure C.2 this mapping $f$ for varying values of $\alpha$ and observe that it is indeed lower than $1$ across the tested range of $\lambda$. We were however unable to formally prove this statement.

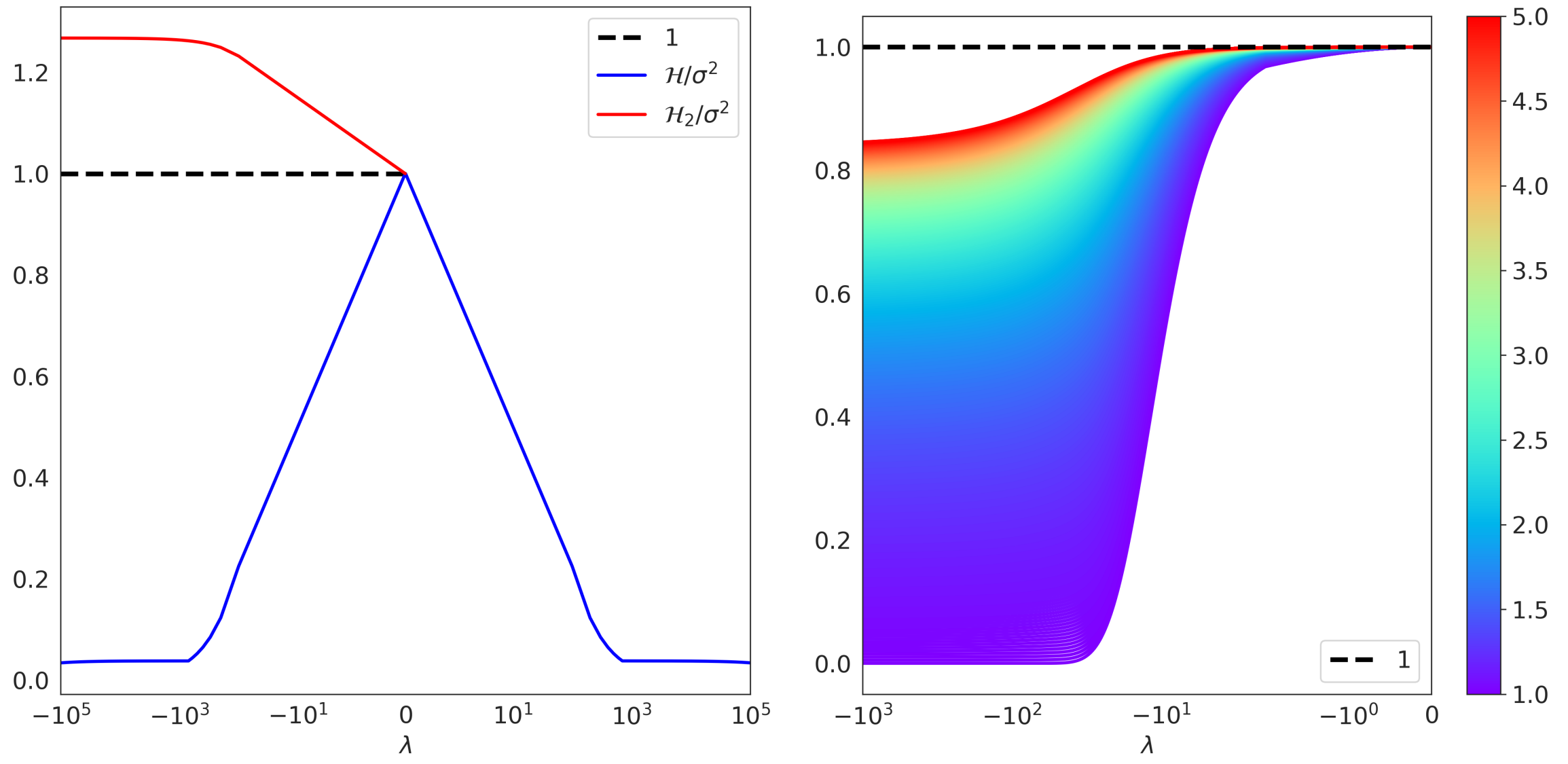

**Figure C.2** – Left: first and second order mappings $\mathcal{H}$ and $\mathcal{H}_2$ for the symmetric $\mathrm{Beta}\,(\alpha, \alpha)$ distribution with $\alpha = 3$. The optimal first order sub-Gaussian parameter is given by $\sigma^2 = \sup_{\mathbb{R}} \mathcal{H} = 1/(4(2\alpha+1))$ and the optimal second order parameter is given by $(\rho\sigma)^2 = \inf_{\mathbb{R}_-} \mathcal{H}_2$. We conjecture that $\rho = 1$. Right: mapping $f$ of equation (C.24) for $\alpha \in [1, 5]$. $f(\lambda) \leqslant 1$ for all $\lambda \in \mathbb{R}_-^\star$ indicates that $\mathcal{H}_2$ is nonincreasing, implying that $\mathrm{Beta}\,(\alpha, \alpha) \in \mathcal{F}^2_{\mathcal{G},1}$.

## Proof of asymptotic expansion of the inverse Tricomi function

*Proof of Lemma 4.12.* The strategy here is to leverage the uniform asymptotic expansion of the Whittaker function $z \in \mathbb{R}_+^\star \mapsto W_{\kappa,\mu}(z)$, where $\kappa, \mu \in \mathbb{R}_+^\star$, which was derived in Olver (1980) (by uniform, we mean that the expansion is valid when both parameters $\mu$ and $\kappa$ and the argument $z$ go to $+\infty$ simultaneously). Indeed, this special function is related to Tricomi's function by the relation $W_{\kappa,\mu}(x) = e^{-x/2}x^{1/2+\mu}U(1/2+\mu-\kappa, 1+2\mu; x)$ for all $x \in \mathbb{R}_+^\star$, from which we may recover the definition of $G^{\mathrm{T}}_{\beta,\gamma,\zeta,t}$. For simplicity, we assume $\beta = 1/2$: not only is it the tuning we recommended in Section 4.3, it also allows to consider $\mu = \kappa$, which greatly simplify downstream calculations (we do not detail the general case, which just amounts to more intricate expressions, with the same conclusion for Tricomi's function). For $x > 2\mu$, we have the following asymptotic expansion for the Whittaker function:

$$W_{\mu,\mu}(x) = \left(\frac{\mu}{2}\right)^{-\frac{1}{4}} \left(\frac{2\mu}{e}\right)^{\mu} \Phi(\mu,\mu,x)\mathcal{U}\left(0, \xi\sqrt{2\mu}\right)\left(1 + \mathcal{O}\left(\frac{\log\mu}{\mu}\right)\right), \tag{C.25}$$

where $\xi = \sqrt{x/\mu - 2 - 2\log(x/(2\mu))}$, $\mathcal{U}$ is the parabolic cylinder function satisfying the following asymptotic expansion when $\mu, x \to +\infty$ (Abramowitz and Stegun, 1968, 19.8.1):

$$\mathcal{U}(0,x) \sim x^{-\frac{1}{2}}e^{-\frac{x^2}{4}}, \tag{C.26}$$

and $\Phi$ is defined as

$$\Phi(\mu,\mu,x) = \sqrt{\frac{\mu\xi x}{x - 2\mu}}. \tag{C.27}$$

Combining all these equations together, we obtain

$$\begin{aligned} U\left(\frac{1}{2}, 1+2\mu; x\right) &\sim e^{\frac{x}{2}}x^{-\frac{1}{2}-\mu}\left(\frac{\mu}{2}\right)^{-\frac{1}{4}}\left(\frac{2\mu}{e}\right)^{\mu}\sqrt{\frac{\mu\xi x}{x-2\mu}}\xi^{-\frac{1}{2}}(2\mu)^{-\frac{1}{4}}e^{-\frac{\xi^2\mu}{2}} \\ &\sim e^{\mu+\mu\log\frac{x}{2\mu}-\mu\log x+\mu\log\frac{2\mu}{e}}(x-2\mu)^{-\frac{1}{2}} \\ &\sim (x-2\mu)^{-\frac{1}{2}}. \end{aligned} \tag{C.28}$$

Now, let $\mu = (\gamma-1)/2 + t/4$, $x = \zeta + z/2$ and $\delta \in (0,1)$. For $t \to +\infty$, $x \to +\infty$, the equation $G^{\mathrm{T}}_{\beta,\gamma,\zeta,t}(z) = 1/\delta$ has the asymptotically equivalent expression:

$$\left(\zeta - \gamma + 1 + \frac{z-t}{2}\right)^{-\frac{1}{2}} \sim \frac{U(\beta,\gamma;\zeta)}{\delta}, \tag{C.29}$$

and after simple algebra we have $z \sim z(t,\delta)$ with

$$z(t,\delta) = t\left(1 + \frac{2\delta^2}{U(\beta,\gamma;\zeta)^2 t} + \frac{2(\gamma-1+\zeta)}{t}\right) \sim t + \frac{2\delta^2}{U(\beta,\gamma;\zeta)^2}. \tag{C.30}$$

Note that the expression of $z(t,\delta)$ ensures that $x > 2\mu$ in the analysis above. ■

### Proof of the pseudo regret upper bound for SOSG distributions

*Proof of Corollary 4.19.* We consider the symmetrised family $\mathring{\mathcal{F}}^2_{\mathcal{G},\rho_k}$ (the other case is proven similarly by observing that the widths of the resulting confidence sequences are essentially the same). Let $k \in [K] \setminus \{k^\star\}$ and denote by $(Y^k_n)_{n\in\mathbb{N}}$ a sequence of i.i.d. random variables drawn from $\nu_k$, and by $(V^I_{k,n})_{n\in\mathbb{N}}$ the associated sequence of U-statistics defined by $V^I_{k,n} = 1/\lfloor n/2 \rfloor \sum_{i=1}^{\lfloor n/2 \rfloor}(Y^k_{2i} - Y^k_{2i-1})^2/2$. Second order sub-Gaussian distributions have fi-nite variance, therefore $V^I_{k,n} = \sigma_k^2 + o(1)$ (law of large numbers). By Lemma 4.12, we have $(G^{\mathrm{T}}_{\beta,\gamma,\zeta,\lfloor n/2 \rfloor})^{-1}(1/\delta) \sim n$ with $\delta = 1/T$. The width of the time-uniform confidence sequence of Corollary 4.17 satisfies the asymptotic expression:

$$\begin{aligned}\left|\widehat{\Theta}^{\delta}_{k,n,\rho_k,c}\right| &\sim \frac{2}{\rho_k}\sqrt{\frac{2\left(\sigma_k^2 + o(1)\right)}{n}(1+\frac{\alpha}{n})\log\left(\frac{3}{\delta}\sqrt{1+\frac{n}{\alpha}}\right)} \\ &\sim \frac{2\sigma_k}{\rho_k}\sqrt{\frac{2}{n}(1+\frac{\alpha}{n})\log\left(\frac{3}{\delta}\sqrt{1+\frac{n}{\alpha}}\right)}\,. \end{aligned} \tag{C.31}$$

From there, we simply apply the generic UCB analysis for $R$-sub-Gaussian distributions developed in Corollary 1.34 with $R = \sigma_k/\rho_k$ (with the caveat that we may only derive the first order term since the above asymptotic expansion for $\left|\widehat{\Theta}^{\delta}_{k,n,\rho_k,c}\right|$ is only at the first order). ■

# Appendix D

# Risk-aware linear bandits with convex loss

This chapter presents supplementary material for Chapter 5.

## Contents

## D.1 Summary and interpretation of elicitable risk measures

We report in Table D.1 an overview of common elicitable risk measures and their associated loss functions. We recall that for a distribution $\nu \in \mathcal{M}_+^1(\mathbb{R})$ and a loss function $\mathcal{L} \colon \mathbb{R} \times \mathbb{R}^p \to \mathbb{R}$, we defined the risk measure elicited by $\mathcal{L}$ as $\rho_{\mathcal{L}}(\nu) = \arg\min_{\xi \in \mathbb{R}^p} \mathbb{E}_{Y \sim \nu}\left[\mathcal{L}(Y, \xi)\right]$. Note that the pairs (mean, variance) and (VaR, CVaR) are second order elicitable but neither the variance nor the CVaR are first order elicitable. For these pairs, we report the generic form of elicitation losses, which depend on arbitrary convex functions $\psi_1$ and $\psi_2$, as well as instances of such losses obtained for the natural choice $\psi_1(\xi) = \psi_2(\xi) = \xi^2/2$ for all $\xi \in \mathbb{R}$.

We provide below some intuition about these commonly used measures in risk management.

**Mean-variance.** Assessing the risk-reward tradeoff of an underlying distribution $\nu$ by penalizing its mean by a higher order moment (typically the variance) is perhaps the most intuitive of risk measures. Following Markowitz (1952), the mean-variance risk measure at risk aversion level $\lambda \in \mathbb{R}$ is defined by $\rho_{\mathrm{MV}_1^\lambda}(\nu) = \mu - \lambda\sigma$, where $\mu$ and $\sigma$ denote the mean and standard deviation of $\nu$. Alternatively, it can also be defined as $\rho_{\mathrm{MV}_2^\lambda}(\nu) = \mu - \frac{\lambda}{2}\sigma^2$, using the variance rather than the standard deviation in the penalisation term. Both measures are especially well-suited for Gaussian distributions as $\mu$ and $\sigma$ fully characterise this family.

**Table D.1** – Example of elicitable risk measures.

| Name | $\rho_{\mathcal{L}}(\nu)$ | Associated loss $\mathcal{L}(y,\xi)$ | Domain |
|---|---|---|---|
| Mean | $\mathbb{E}_{Y\sim\nu}[Y]$ | $(y-\xi)^2$<br><br>Bregman divergence $\mathcal{B}_\psi(y,\xi)$<br>$\psi(y)-\psi(\xi)-\psi'(\xi)(y-\xi)$,<br>$\psi$ differentiable,<br>strictly convex. | $\xi \in \mathbb{R}$ |
| Derived from potential $\psi$ | $\underset{\xi\in\mathbb{R}}{\operatorname{argmin}}\, \mathbb{E}_{Y\sim\nu}[\psi(Y-\xi)]$ | $\psi(y-\xi)$ | $\xi \in \operatorname{dom}(\psi)$ |
| Generalised moment $T\colon \mathbb{R}\to\mathbb{R}$ | $\mathbb{E}_{Y\sim\nu}[T(Y)]$ | $\frac{1}{2}\xi^2 - \xi T(y)$ | $\xi \in \mathbb{R}$ |
| Entropic risk, $\gamma \neq 0$ | $\frac{1}{\gamma}\log \mathbb{E}_{Y\sim\nu}[e^{\gamma Y}]$ | $\xi + \frac{1}{\gamma}(e^{\gamma(y-\xi)}-1)$ | $\xi \in \mathbb{R}$ |
| Mean and variance | $\mu = \mathbb{E}_{Y\sim\nu}[Y]$<br>$\sigma^2 = \mathbb{E}_{Y\sim\nu}[Y^2]-\mu^2$ | $\frac{1}{2}\xi_1^2 + \frac{1}{2}(\xi_2+\xi_1^2)^2$<br>$-\xi_1 y - (\xi_2+\xi_1^2)y^2$<br><br>$-\psi_1(\xi_1) - \psi_1'(\xi_1)(y-\xi_1)$<br>$-\psi_2(\xi_2+\xi_1^2)$<br>$-\psi_2'(\xi_2+\xi_1^2)(y^2-\xi_2-\xi_1^2))$,<br>$\psi_1,\psi_2$ differentiable,<br>strictly convex. | $\xi_1 \in \mathbb{R}$<br>$\xi_2 \geqslant 0$ |
| $\mathrm{VaR}_\alpha$ and $\mathrm{CVaR}_\alpha$, $\alpha\in(0,1)$ | $\mathrm{VaR}_\alpha = \inf\{y\in\mathbb{R}, \int_{-\infty}^{y} d\nu \geqslant \alpha\}$<br>$\mathrm{CVaR}_\alpha = \frac{1}{\alpha}\int_0^\alpha \mathrm{VaR}_a da$ | $(\xi_1-y)_+ - \alpha\xi_1$<br>$+\xi_2(\frac{1}{\alpha}(\xi_1-y)_+ - \xi_1)$<br>$+\frac{1}{2}\xi_2^2$<br><br>$(\mathbb{1}_{y\leqslant\xi_1}-\alpha)\psi_1'(\xi_1)$<br>$-\mathbb{1}_{y\leqslant\xi_1}\psi_1'(y)$<br>$+\psi_2'(\xi_2)(\xi_2-\xi_1+\frac{1}{\alpha}\mathbb{1}_{y\leqslant\xi_1}(\xi_1-y))$<br>$-\psi_2(\xi_2)+c(y)$,<br>$\psi_1$ convex,<br>$\psi_2$ strictly convex and increasing,<br>$c\colon\mathbb{R}\to\mathbb{R}$. | $\xi_1 \geqslant \xi_2$ |

The pair (mean, variance) is jointly second order (but not first order) elicitable by a pair of convex functions; the formula recalled in Table D.1 can be found in Brehmer (2017, Example 1.23).

VaR **and** CVaR**.** For a distribution with continuous c.d.f. (i.e. atomless), the value at risk $\mathrm{VaR}_\alpha(\nu)$ at level $\alpha \in (0,1)$ is equivalent to the $\alpha$ quantile, and a simple change of variable reveals that the conditional value at risk $\mathrm{CVaR}_\alpha(\nu)$ is thus $\mathbb{E}\left[X \mid X \leqslant \mathrm{VaR}_\alpha(\nu)\right]$. Intuitively, a random variable with a high $\mathrm{CVaR}_\alpha$ distribution takes on average relatively high values in the "$\alpha\%$ worst case" scenario. For $\alpha \to 1^-$, $\mathrm{CVaR}_\alpha(\nu) \to \mathbb{E}_{Y\sim\nu}[Y]$ and thus the risk measure becomes oblivious to the tail risk; on the contrary, the case $\alpha \to 0^+$ emphasises only the worst outcomes.

In the Gaussian case $\nu \sim \mathcal{N}(\mu, \sigma)$, using the notations $\phi$ and $\Phi$ respectively for the p.d.f. and c.d.f. of the standard normal distribution, simple calculus shows that

$$\mathrm{VaR}_\alpha(\nu) = \mu + \sigma\Phi^{-1}(\alpha)\,, \tag{D.1}$$

$$\mathrm{CVaR}_\alpha(\nu) = \mu - \frac{\sigma}{\alpha\sqrt{2\pi}}\phi\left(\Phi^{-1}(\alpha)\right)\,, \tag{D.2}$$

i.e. $\mathrm{CVaR}_\alpha(\nu) = \rho_{\mathrm{MV}_1^\lambda}(\nu)$ with risk aversion level $\lambda = \frac{1}{\alpha\sqrt{2\pi}}\phi\left(\Phi^{-1}(\alpha)\right)$. In particular, increasing the variance $\sigma^2$ reduces $\mathrm{CVaR}_\alpha(\nu)$, corresponding to the intuition of higher volatility risk.

The pair (VaR, CVaR) is also jointly second order (but not first order) elicitable by a pair of convex functions; the formula recalled in Table D.1 can be found in Fissler and Ziegel (2016, Corollary 5.5).

**Entropic risk.** The non-elicitability of $\mathrm{CVaR}_\alpha$ motivated the use of the entropic risk as an alternative measure. For a given risk level $\gamma \neq 0$, this measures rewrites as (Brandtner et al., 2018)

$$\rho_\gamma(\nu) = \sup_{\nu' \in \mathcal{M}_1^+(\mathbb{R})} \left\{ \mathbb{E}_{Y\sim\nu'}[Y] - \frac{1}{\gamma}\mathrm{KL}(\nu' \parallel \nu) \right\}\,. \tag{D.3}$$

The intuition here is similar to the mean-variance measure, i.e. penalizing the expected value by a measure of uncertainty, but differs by the use of the Kullback-Leibler divergence $\mathrm{KL}(\nu' \parallel \nu) = \mathbb{E}_{Y\sim\nu'}[\log \frac{d\nu'}{d\nu}]$ instead of the variance. The entropic risk measure can be inter-preted as the largest expected value that a mispecified model $\nu'$ (in place of the true underlying distribution $\nu$) may have, where $\mathrm{KL}(\nu' \parallel \nu)$ controls the magnitude of the mispecification.

Again, in the Gaussian case, this measure reduces to $\rho_\gamma(\nu) = \mu + \frac{\gamma}{2}\sigma^2 = \rho_{\mathrm{MV}_{\frac{\lambda}{2}}}(\nu)$ at risk aversion level $\lambda = -\gamma$.

The entropic risk at level $\gamma \neq 0$ is first order elicitable; the formula recalled in Talbe D.1 can be found in Embrechts et al. (2021, Example 1).

**Expectile.** Beyond their interpretation as generalised, smooth quantiles, expectiles can also be understood in light of the financial risk management literature. Let $e_p(\nu)$ denote the $p$-expectile of $\nu$ for a given probability $p \in (0, 1)$. Then, simple calculus shows that

$$(1-p)\mathbb{E}_{Y\sim\nu}[(e_p(\nu) - Y)_+] = p\mathbb{E}_{Y\sim\nu}[(Y - e_p(\nu))_+] \,, \tag{D.4}$$

where $z_+ = \max(z, 0)$. If $\nu$ represents the distribution of a tradeable asset $Y$ at time $T$, then the $p$-expectile is the strike $K = e_p(\nu)$ such that call and put on $Y$ struck at $K$ at maturity $T$ are in proportion $\frac{1-p}{p}$ to each other, where we define the call and put prices (with zero time discounting) by respectively

$$C(\nu, K) = \mathbb{E}_{Y\sim\nu}[(Y - K)_+] \,, \tag{D.5}$$

$$P(\nu, K) = \mathbb{E}_{Y\sim\nu}[(K - Y)_+] \,. \tag{D.6}$$

Similarly, Keating and Shadwick (2002) introduced the notion of Omega ratio as a risk-return performance measures. It is defined at level $K$ by

$$\Omega(K) = \frac{\int_K^{+\infty} (1 - F(y))\, dy}{\int_{-\infty}^K F(y) dy} \,, \tag{D.7}$$

where $F$ is the c.d.f. of $\nu$. This ratio can also be viewed as a call-put ratio, hence another definition of the $p$-expectile is via the implicit equation $\Omega(K) = \frac{1-p}{p}$ for $K = e_p(\nu)$.

Contrary to the previous risk measures, it may not be clear from this definition alone that expectiles do encode a notion of aversion to risk. The next proposition shows that $p$-expectiles of many distributions, including normal and adjusted lognormal, are decreasing functions of their variances when $p < \frac{1}{2}$, thus penalizing more volatile distributions, making them suitable for risk management. We provide an elementary proof using the tools of the financial mathematics literature, where such risk measures were extensively studied.

**Proposition D.1.** *Let $\mathcal{I} \subseteq \mathbb{R}_+^*$ an open interval and $\{\nu_\sigma, \sigma \in \mathcal{I}\}$ a family of probability distributions such that*

*(i) the expectation mapping $\sigma \in \mathcal{I} \mapsto \mathbb{E}_{Y\sim\nu_\sigma}[Y]$ is constant,*

*(ii) for any $K \in \mathbb{R}$, both the call and put mappings*

$$C(\cdot, K) \colon \sigma \in \mathcal{I} \mapsto \mathbb{E}_{Y\sim\nu_\sigma}\left[(Y-K)_+\right], \tag{D.8}$$

$$P(\cdot, K) \colon \sigma \in \mathcal{I} \mapsto \mathbb{E}_{Y\sim\nu_\sigma}\left[(K-Y)_+\right], \tag{D.9}$$

*are differentiable and nondecreasing.*

*For any $p \in (0,1)$, the sign of the sensitivity of the $p$-expectile of $\nu_\sigma$ to $\sigma$ is given by*

$$\operatorname{sign} \frac{d}{d\sigma} e_p(\nu_\sigma) = \operatorname{sign}(p - \frac{1}{2}). \tag{D.10}$$

Before we proceed to the proof, let us note that two classical families of distributions satisfy these assumptions (see Merton (1973, Theorem 8) for a general result).

(i) **Normal**: for $\mu_0 \in \mathbb{R}$, $\{\nu_\sigma = \mathcal{N}(\mu_0, \sigma^2),\ \sigma \in \mathbb{R}_+^*\}$, for which $\mathbb{E}_{Y\sim\nu_\sigma}[Y] = \mu_0$.

(ii) **Adjusted lognormal**: for $\mu_0 \in \mathbb{R}$, $\{\nu_\sigma = \exp\left(\mathcal{N}(\mu_0, \sigma^2) - \frac{\sigma^2}{2}\right),\ \sigma \in \mathbb{R}_+^*\}$, for which $\mathbb{E}_{Y\sim\nu_\sigma}[Y] = e^{\mu_0}$.

In particular in the normal case, it follows from Lemma 5.4 that

$$e_p(\nu_\sigma) = \mu_0 + \sigma e_p(\mathcal{N}(0,1)), \tag{D.11}$$

and thus $e_p(\nu_\sigma) = \rho_{\mathrm{MV}_1^\lambda}(\nu_\sigma)$, where the risk aversion level is $\lambda = -e_p(\mathcal{N}(0,1))$ (which by Proposition D.1 is positive if $p < \frac{1}{2}$).

*Proof of Proposition D.1.* We first recall the call-put parity principle, which states that for any distribution $\nu_\sigma$ and strike $K \in \mathbb{R}$, the following equality holds:

$$C(\sigma, K) - P(\sigma, K) = \mathbb{E}_{Y\sim\nu_\sigma}[Y] - K, \tag{D.12}$$

where we write $C(\nu_\sigma, K) = C(\sigma, K)$ and $P(\nu_\sigma, K) = P(\sigma, K)$.

A straightforward consequence of the call-put parity principle and the assumption that $\frac{d}{d\sigma}\mathbb{E}_{Y\sim\nu_\sigma}[Y] = 0$ is that $\partial_\sigma C = \partial_\sigma P$. We denote this quantity by $V$. Now, note that the mapping $\sigma \in \mathcal{I} \mapsto e_p(\nu_\sigma)$ is differentiable (implicit function theorem). From the equation

$C(\sigma, e_p(\nu_\sigma)) = \frac{1-p}{p} P(\sigma, e_p(\nu_\sigma))$, we deduce that

$$\frac{d}{d\sigma} C(\sigma, e_p(\nu_\sigma)) = \partial_\sigma C(\sigma, e_p(\nu_\sigma)) + \partial_K C(\sigma, e_p(\nu_\sigma)) \frac{d}{d\sigma} e_p(\nu_\sigma) \,, \tag{D.13}$$

$$\frac{d}{d\sigma} P(\sigma, e_p(\nu_\sigma)) = \partial_\sigma P(\sigma, e_p(\nu_\sigma)) + \partial_K P(\sigma, e_p(\nu_\sigma)) \frac{d}{d\sigma} e_p(\nu_\sigma) \,, \tag{D.14}$$

and thus

$$\frac{1-2p}{p} V + \frac{d}{d\sigma} e_p(\nu_\sigma) \left( \frac{1-p}{p} \partial_K P(\sigma, e_p(\nu_\sigma)) - \partial_K C(\sigma, e_p(\nu_\sigma)) \right) = 0 \,. \tag{D.15}$$

Elementary option pricing principles show that $V \geqslant 0$, i.e. the call and put prices both increase with higher volatility, and that $\partial_K C \leqslant 0$ and $\partial_K P \geqslant 0$. Therefore, we deduce that $\frac{d}{d\sigma} e_p(\nu_\sigma) \leqslant 0$. ■

In particular for $p = 1/2$, the $p$-expectile corresponds to the strike $K$ at which call and put have equal prices, which by the call-put parity principle (with zero discounting) implies that $K = \mathbb{E}[Y]$, thus giving an alternative derivation of the equivalence between $1/2$-expectile and mean.

## D.2 Alternative analysis of LinUCB-CR

We detail here the steps of the alternative setting for the analysis of LinUCB-CR where actions sets are stochastically generated and satisfy a conditional covariance lower bound (Assumption 5.22).

### Impact of using the local Hessian metric $H_t$ versus the global metric $V_t$

To highlight the benefit of using local metrics, we detail here the regret bound obtained using the above proof with the natural global metric induced by $V^{\alpha/m}$ (independent of the local point $\theta$). Instantiating the positive definite matrix $P$ to $V_t^{\alpha/m}$ instead of $\bar{H}^\alpha(\theta^*, \bar{\theta}_t)$ in the bound on the prediction error of Section 5.2 yields

$$\begin{aligned}
\Delta(x, \bar{\theta}_t) &\leqslant \|\theta^* - \bar{\theta}_t\|_{V_t^{\alpha/m}} \|x\|_{(V_t^{\alpha/m})^{-1}} \\
&= \|F_t^\alpha(\theta^*) - F_t^\alpha(\bar{\theta}_t)\|_{\bar{H}_t^\alpha(\theta^*, \bar{\theta}_t)^{-1} V_t^{\alpha/m} \bar{H}_t^\alpha(\theta^*, \bar{\theta}_t)^{-1}} \|x\|_{(V_t^{\alpha/m})^{-1}} \\
&\leqslant \frac{1}{\sqrt{m}} \|F_t^\alpha(\theta^*) - F_t^\alpha(\bar{\theta}_t)\|_{\bar{H}_t^\alpha(\theta^*, \bar{\theta}_t)^{-1}} \|x\|_{(V_t^{\alpha/m})^{-1}} \\
&\leqslant 2\sqrt{\frac{\kappa}{m}} \left( \sigma \sqrt{2 \log \frac{1}{\delta} + d \log \frac{m}{\alpha} + \log \det V_t^{\frac{\alpha}{m}}} + \sqrt{\frac{\alpha}{\kappa}} S \right) \|x\|_{(V_t^{\alpha/m})^{-1}} \,.
\end{aligned} \tag{D.16}$$

Similarly to the above proof, this shows that

$$\gamma_t^{\text{global}}: x \in \mathcal{X}_t \mapsto 2\sqrt{\frac{\kappa}{m}}\left(\sigma\sqrt{2\log\frac{1}{\delta} + d\log\frac{m}{\alpha} + \log\det V_t^{\frac{\alpha}{m}}} + \sqrt{\frac{\alpha}{\kappa}}S\right)\|x\|_{(V_t^{\alpha/m})^{-1}} \tag{D.17}$$

is also a valid choice of exploration sequence. Finally, a straightforward application of Lemma 5.19 shows the regret of the corresponding LinUCB-CR strategy is upper bounded with probability at least $1-\delta$ by

$$4\sqrt{\frac{\kappa}{m}}\left(\sigma\sqrt{2\log\frac{1}{\delta} + d\log\frac{m}{\alpha} + \log\det V_t^{\frac{\alpha}{m}}} + \sqrt{\frac{\alpha}{\kappa}}S\right)\sqrt{2Td\log\left(1+\frac{mTL^2}{d\alpha}\right)}. \tag{D.18}$$

Compared to the local analysis, this improves the scaling of the regret in $\kappa$ by a factor $\sqrt{\kappa}$. However, it forces the use of the global metric $\|\cdot\|_{(V_t^{\alpha/m})^{-1}}$ instead of the local one $\|\cdot\|_{H_t^{\kappa\alpha}(\bar{\theta}_t)^{-1}}$, thus ignoring the precise shape of the loss function $\mathcal{L}$.

Looking at our proof, we see that $\kappa$ and $m$ fulfil two different roles. $\sqrt{\kappa}$ is the price to pay in order to transport local metrics $H_t^{\alpha}(\theta)$ between $\theta = \theta^*$ (true parameter) and $\theta = \bar{\theta}_t$ (estimate); it is paid once to bound the prediction error $\Delta(X_t, \theta)$ using the concentration bound of Proposition 5.12, and it is also paid a second time if local metrics are used in the exploration bonus when moving from $\bar{H}_t(\theta^*, \bar{\theta}_t)^{-1}$ to $\bar{H}_t(\bar{\theta}_t)^{-1}$ (the former cannot be used directly in the algorithm as it depends on the a priori unknown paramter $\theta^*$). On the other hand, $m^{-1/2}$ is the price paid in both the local and global analyses to move from $H_t(\bar{\theta}_t)^{-1}$ to $V_t^{-1}$ in order to apply the elliptic potential lemma, which is in general incompatible with local metrics. A similar phenomenon is observed in the analysis of Faury et al. (2020): the regret of their algorithm Logistic-UCB-1 (global) scales as $\sqrt{\kappa}$ while that of Logistic-UCB-2 (local) scales as $\kappa$. In addition to local metrics, Logistic-UCB-2 also makes use of an intricate projection step that allows for a new elliptic potential lemma compatible with local metrics, thus removing the factor $m^{-1}$ (at least from the first order contribution to the regret in $T$). We conjecture that a similar analysis could be unlocked in the present risk-aware setting and leave it open for future investigation.

We reiterate that in the logistic setting, $\kappa$ is derived from self-concordance properties of the link function and is in particular independent of the curvature lower bound represented by $m$ (it is in fact equal to $1+2S$ where $S$ is an upper bound on the parameter space $\Theta$, as in Assumption 5.16). By analogy with logistic bandits, we argue that the exact scaling in $\kappa$ is likely not too harmful for the practical performances of Algorithm 3 and we therefore recommend the use of local metrics instead.

## Proof of Theorem 5.23

In this section, we prove the regret bound of Theorem 5.23 in the stochastic i.i.d. actions setting of Assumption 5.22. The main difference with the proof of Theorem 5.18 is the use of an alternative stochastic elliptic potential lemma, mirroring the classical result of Lemma 5.19, that exploits the lower bound on the covariance of actions (Assumption 5.22). This proof technique is adapted from Kim et al. (2023), although we use a different, sharper concentration result (Proposition D.3 below).

**Lemma D.2** (Stochastic elliptic potential lemma). *Let $\delta \in (0,1)$, $\beta > 0$, $(\theta_t)_{t\in\mathbb{N}}$ a sequence of vectors in $\Theta \subseteq \mathbb{R}^d$ and $(X_t)_{t\in\mathbb{N}}$ a sequence of random variables in $\mathbb{R}^d$. Recall that for $t \in \mathbb{N}$,*

$$H_t^\beta(\theta_t) = \sum_{s=1}^{t-1} \partial^2 \mathcal{L}(Y_s, \langle X_s, \theta_t \rangle) X_s X_s^\top + \beta I_d \,, \tag{D.19}$$

$$V_t^\beta = \sum_{s=1}^{t-1} X_s X_s^\top + \beta I_d \,. \tag{D.20}$$

*Fix $t_0 = \lceil \frac{8}{\rho_{\mathcal{X}}^2} \log \frac{2}{\delta} - \frac{2\beta}{m\rho_{\mathcal{X}} L^2} \rceil$. Under Assumptions 5.17 and 5.22, for $T \geqslant t_0$, with probability at least $1-\delta$, it holds that*

$$\sum_{t=1}^{T} \|X_t\|_{H_t^\beta(\theta_t)^{-1}} \leqslant 2\sqrt{\frac{2T}{m\rho_{\mathcal{X}}}} \left(1 + \frac{C}{\sqrt{T}}\right) , \tag{D.21}$$

*where*

$$\begin{aligned} C = \frac{1}{L^2}\sqrt{\frac{\beta - 4\frac{mL^2}{\rho_{\mathcal{X}}}\log\frac{2}{\delta}}{2m\rho_{\mathcal{X}}}} - \frac{1}{2L}\sqrt{t_0 - 1 + \frac{2(\beta - \frac{4mL^2}{\rho_{\mathcal{X}}}\log\frac{2}{\delta})}{m\rho_{\mathcal{X}} L^2}} \\ + \frac{\sqrt{\rho_{\mathcal{X}}}}{2} \max\left(1, L\sqrt{\frac{m}{\beta}}\right) \sqrt{t_0 d \log\left(1 + \frac{mL^2 t_0}{d\beta}\right)} . \end{aligned} \tag{D.22}$$

The intuition about this result is the following: if the matrix norms induced by $H_t^\beta(\theta_t)$ grow at least linearly in $t$, then the left-hand side should scale like $\sum_{t=1}^T \frac{1}{\sqrt{t}} = \mathcal{O}(\sqrt{T})$, without the extra $\mathcal{O}(\sqrt{\log T})$ factor present in Lemma 5.19. The lower curvature bound of Assumption 5.8 shows that it is enough to look at the norms induced by $V_t^{\beta/m}$, at the cost of an extra $m^{-1/2}$ factor (in particular Lemma D.2 holds for *any* sequence $(\theta_t)_{t\in\mathbb{N}}$, not just the sequence of estimators used in the bandit algorithms). Because of the stochastic sampling of actions (Assumption 5.22), it is likely that the sequence $(X_t)_{t\in\mathbb{N}}$ spans all directions of $\mathbb{R}^d$ quite fast; in other words, each new $X_t X_t^\top$ will contribute at least a fixed amount to the sum that defines $V_t^{\beta/m}$, leading to the linear growth of the induced norms.

We formalise this intuition in Lemma D.4 below, which relies on the following concentration bound in Hilbert spaces.

**Proposition D.3** (Time-uniform line crossing inequality for martingales with bounded increments in a Hilbert space). *Let $(\mathcal{H}, \langle\cdot,\cdot\rangle)$ a Hilbert space and $(M_t)_{\in\mathbb{N}}$ a $\mathcal{H}$-valued martingale (with respect to a filtration $(\mathcal{G}_t)_{t\in\mathbb{N}}$) such that $M_0 = 0$. Assume that there exists a sequence of positive scalars $(c_t)_{t\in\mathbb{N}}$ such that $\|M_{t+1} - M_t\| \leqslant c_t$ for all $t \in \mathbb{N}$, where $\|\cdot\|$ denotes the norm induced by the scalar product. Then for any $\eta > 0$ and $\delta \in (0,1)$, it holds that*

$$\mathbb{P}\left(\forall t \in \mathbb{N},\ \|M_t\| \leqslant \frac{1}{2\eta}\log\frac{2}{\delta} + \eta\sum_{s=1}^{t} c_s^2\right) \geqslant 1-\delta\,. \tag{D.23}$$

Interestingly, this concentration bound does not depend on the dimension of $\mathcal{H}$, and in particular remains valid even if the ambient space is infinite-dimensional. Moreover, this bound controls the probability that *any* deviation occurs in the sequence $(\|M_t\|)_{t\in\mathbb{N}}$, which is much stronger than controlling the deviation probability individually at each time $t \in \mathbb{N}$. The proof relies on martingale arguments rather than a crude union bound over a finite set of individual deviation probabilities (as in the results invoked in Kim et al. (2023)), which yields anytime ($t \in \mathbb{N}$ rather than $t \leqslant T$ for some known horizon $T$) and typically tighter bounds.

*Proof of Proposition D.3.* This result is directly taken from Howard et al. (2020). More precisely, Howard et al. (2020, Theorem 1) shows a variety of equivalent time-uniform line crossing inequalities for martingales, and Howard et al. (2020, Corollary 10) applies this generic result to concentration of norm-like operators in Banach spaces. In order to get the most convenient form for our problem, we derive Proposition D.3 from the generic theorem rather than the specific corollary.

The proofs of Pinelis (1992, Theorem 3) and Pinelis (1994, Theorem 3) reveals that for any $\lambda \in \mathbb{R}$ and $t \in \mathbb{N}$, the exponential process defined by $L_t = \cosh(\lambda\|M_t\|)\exp(-\frac{\lambda^2}{2}\sum_{s=1}^t c_s)$ is a nonnegative $\mathcal{G}$-supermartingale. Therefore, Howard et al. (2020, Theorem 1, (a)) shows that for any $a, b > 0$,

$$\mathbb{P}\left(\exists t \in \mathbb{N},\ S_t \geqslant a + bC_t\right) \leqslant 2e^{-aD(b)}\,, \tag{D.24}$$

where $S_t = \|M_t\|$, $C_t = \sum_{s=1}^{t} c_s^2$ and $D(b) = 2b$. Equating the right-hand side to $\delta$ and letting $b = \eta$ concludes the proof, with $a = \frac{1}{2\eta}\log\frac{2}{\delta}$. ■

In the next lemma, we show that the smallest eigenvalue of $H_t^{\beta}(\theta_t)$, which provides a lower bound to the corresponding induced norm, does indeed grow linearly with $t$ on an event of high probability.

**Lemma D.4** (Smallest eigenvalue of $H_t^{\beta}(\theta_t)$ grows linearly with $t$ with high probability). *Under Assumptions 5.8, 5.17 and 5.22, we have that*

$$\mathbb{P}\left(\forall t \in \mathbb{N},\ \lambda_{\min}\left(H_{t+1}^{\beta}(\theta_{t+1})\right) \geqslant \beta - \frac{4mL^2}{\rho_{\mathcal{X}}}\log\frac{2}{\delta} + \frac{m\rho_{\mathcal{X}}L^2}{2}t\right) \geqslant 1-\delta\,. \tag{D.25}$$

*Proof of Lemma D.4.* Let $t \in \mathbb{N}$. First notice that Assumption 5.8 implies

$$\lambda_{\min}\left(H_{t+1}^{\beta}(\theta_{t+1})\right) \geqslant m\lambda_{\min}\left(V_{t+1}^{0}\right) + \beta\,. \tag{D.26}$$

The idea is to relate $\lambda_{\min}\left(V_{t+1}^{0}\right)$ to the norm of some martingale in order to apply Proposition D.3. In the stochastic actions setting (Assumption 5.22), the following sum of random matrices naturally defines a martingale (with respect to the adapted bandit filtration $(\bar{\mathcal{G}}_t)_{t\in\mathbb{N}}$):

$$M_t = \sum_{s=1}^{t} X_s X_s^\top - \mathbb{E}\left[X_s X_s^\top \mid \bar{\mathcal{G}}_{s-1}\right] = V_{t+1}^{0} - \bar{V}_{t+1}^{0}\,, \tag{D.27}$$

where we defined $\bar{V}_{t+1}^{0} = \sum_{s=1}^{t} \mathbb{E}\left[X_s X_s^\top \mid \bar{\mathcal{G}}_{s-1}\right]$. We recall that a consequence of Weyl's inequality on eigenvalues is that for any $A, B \in \mathcal{S}_d(\mathbb{R})$, the following inequality holds:

$$\lambda_{\min}(A) + \lambda_{\min}(B) \leqslant \lambda_{\min}(A+B)\,. \tag{D.28}$$

Applying this to $A = M_t$ and $B = \bar{V}_{t+1}^{0}$ yields $\lambda_{\min}(M_t) + \lambda_{\min}\left(\bar{V}_{t+1}^{0}\right) \leqslant \lambda_{\min}\left(V_{t+1}^{0}\right)$. Now, notice that $\lambda_{\min}(A) = -\lambda_{\max}(A) \geqslant -\|A\|$ where $\|\cdot\|$ is the matrix norm induced by the scalar product $\langle A, B\rangle = \mathrm{Tr}(A^\top B)$ (also known as the Frobenius norm). Moreover, the conditional covariance lower bound of Assumption 5.22 and another application of Weyl's inequality imply

that

$$\lambda_{\min}\left(\bar{V}_{t+1}^{0}\right) \geqslant \sum_{s=1}^{t} \lambda_{\min}\left(\mathbb{E}\left[X_s X_s^\top \mid \bar{\mathcal{G}}_{s-1}\right]\right) \geqslant \rho_{\mathcal{X}} L^2 t\,. \tag{D.29}$$

Combining these together, we obtain that for arbitrary $a \in \mathbb{R}$ and $b > 0$, the following inequality holds:

$$\mathbb{P}\left(\exists t \in \mathbb{N},\ \lambda_{\min}\left(H_{t+1}^{\beta}(\theta_{t+1})\right) \leqslant a + bt\right) \leqslant \mathbb{P}\left(\exists t \in \mathbb{N},\ \|M_t\| \geqslant \frac{\beta - a}{m} + \left(\rho_{\mathcal{X}} L^2 - \frac{b}{m}\right) t\right). \tag{D.30}$$

Notice that $\|M_{t+1} - M_t\| = \|X_t X_t^\top - \mathbb{E}[X_t X_t^\top \mid \bar{\mathcal{G}}_{t-1}]\| \leqslant 2L^2$ (Assumption 5.17). Now if we choose $b = \frac{1}{2} m \rho_{\mathcal{X}} L^2$ and $a = \beta - \frac{4mL^2}{\rho_{\mathcal{X}}} \log \frac{2}{\delta}$, the bound from Proposition D.3 holds with $\eta = \frac{\rho_{\mathcal{X}}}{8L^2}$ and $\delta \in (0,1)$ and $c_t = 2L^2$, thus proving the result. ■

We are now ready to prove the stochastic elliptic potential lemma.

*Proof of Lemma D.2.* We fix $t_0 \in \mathbb{N}$ arbitrarily for now and let $T \geqslant t_0$. We start by splitting the sum in two and by applying the deterministic elliptic potential lemma (Lemma 5.19) up to time $t_0$:

$$\begin{aligned}
\sum_{t=1}^{T} \|X_t\|_{H_t^{\beta}(\theta_t)^{-1}} &= \sum_{t=1}^{t_0} \|X_t\|_{H_t^{\beta}(\theta_t)^{-1}} + \sum_{t=t_0+1}^{T} \|X_t\|_{H_t^{\beta}(\theta_t)^{-1}} \\
&\leqslant \frac{1}{\sqrt{m}} \sum_{t=1}^{t_0} \|X_t\|_{\left(V_t^{\beta/m}\right)^{-1}} + \sum_{t=t_0+1}^{T} \|X_t\|_{H_t^{\beta}(\theta_t)^{-1}} && \text{(Assumption 5.8)} \\
&\leqslant \left(\frac{1}{\sqrt{m}}, \frac{L}{\sqrt{\beta}}\right) \sqrt{2 t_0 d \log\left(1 + \frac{mL^2 t_0}{d\beta}\right)} + \sum_{t=t_0}^{T-1} \|X_{t+1}\|_{H_{t+1}^{\beta}(\theta_{t+1})^{-1}} \\
& && \text{(Assumption 5.17)} \\
&.
\end{aligned} \tag{D.31}$$

Now let $\mathcal{E}_t^\delta = \left\{\forall t' \geqslant t,\ \lambda_{\min}\left(H_{t+1}^{\beta}(\theta_{t+1})\right) > \beta - \frac{4mL^2}{\rho_{\mathcal{X}}} \log \frac{2}{\delta} + \frac{m\rho_{\mathcal{X}} L^2}{2} t\right\}$. It is clear that $\mathcal{E}_{t_0}^\delta \subseteq \mathcal{E}_0^\delta$, and thus by Lemma D.4, $\mathbb{P}\left(\mathcal{E}_{t_0}^\delta\right) \geqslant 1 - \delta$. The choice $t_0 = \lceil \frac{8}{\rho_{\mathcal{X}}^2} \log \frac{2}{\delta} - \frac{2\beta}{m\rho_{\mathcal{X}} L^2} \rceil$ implies that the right-hand side in the definition of $\mathcal{E}_{t_0}$ is positive. On this event, we bound the second sum as follows:

$$\begin{aligned}
\sum_{t=t_0}^{T-1} \|X_{t+1}\|_{H_{t+1}^{\beta}(\theta_{t+1})^{-1}} &\leqslant L \sum_{t=t_0}^{T-1} \frac{1}{\sqrt{a + bt}} \\
&\leqslant \frac{2L}{b} \left(\sqrt{a + b(T-1)} - \sqrt{a + b(t_0 - 1)}\right)
\end{aligned}$$

$$\leqslant \frac{2L}{b}\left(\sqrt{bT}+\sqrt{a}-\sqrt{a+b(t_0-1)}\right), \tag{D.32}$$

where we use the shorthand $a = \beta - \frac{4mL^2}{\rho_{\mathcal{X}}}\log\frac{2}{\delta}$ and $b = \frac{1}{2}m\rho_{\mathcal{X}}L^2$ (the penultimate line comes from sum-integral comparison while the last one follows from the inequality $\sqrt{x+y} \leqslant \sqrt{x}+\sqrt{y}$). After collecting the dominating term in $\sqrt{T}$, the two sums give the following upper bound:

$$\sum_{t=1}^{T}\|X_t\|_{H_t^{\beta}(\theta_t)^{-1}} \leqslant 2L\sqrt{\frac{T}{b}}\left(1+\frac{C}{\sqrt{T}}\right) \tag{D.33}$$

with

$$C = \sqrt{\frac{a}{b}} - \sqrt{\frac{a+b(t_0-1)}{b}} + \frac{\sqrt{b}}{2L}\left(\frac{1}{\sqrt{m}}, \frac{L}{\sqrt{\beta}}\right)\sqrt{2t_0 d\log\left(1+\frac{mL^2t_0}{d\beta}\right)}. \tag{D.34}$$

Substituting $a$ and $b$ with their expressions yields the result. ■

We finally prove the regret bound in the stochastic i.i.d. actions setting.

*Proof of Theorem 5.23.* We follow the exact same steps as with Theorem 5.18 in order to bound the regret deviation probability by

$$\mathbb{P}\left(\forall T \geqslant t_0,\ \mathcal{R}_T \leqslant 2c_T^{\delta}\sum_{t=1}^{T}\|X_t\|_{H_t^{\kappa\alpha}(\bar{\theta}_t)^{-1}}\right) \geqslant \mathbb{P}\left(\forall T \in \mathbb{N},\ \mathcal{R}_T \leqslant 2c_T^{\delta}\sum_{t=1}^{T}\|X_t\|_{H_t^{\kappa\alpha}(\bar{\theta}_t)^{-1}}\right) \geqslant 1-\delta\,. \tag{D.35}$$

Moreover, by Lemma D.2, we have

$$\mathbb{P}\left(\forall T \geqslant t_0,\ \sum_{t=1}^{T}\|X_t\|_{H_t^{\kappa\alpha}(\bar{\theta}_t)^{-1}} \leqslant 2\sqrt{\frac{2T}{m\rho_{\mathcal{X}}}}\left(1+\frac{C}{\sqrt{T}}\right)\right) \geqslant 1-\delta\,. \tag{D.36}$$

A simple union argument over both time-uniform events concludes the proof, resulting in a regret upper bound with probability at least $1-2\delta$. ■

**Remark D.5** (Previous results about the growth of the smallest eigenvalue). *Kim et al. (2021, 2023) prove similar results on the linear growth of the smallest eigenvalues of a different sequence of Hessian matrices. More precisely, they consider a fixed number $K$ of arms, i.e. action sets of the form $\mathcal{X}_t = \{X_{k,t}, k \in [K]\}$ and Hessian matrices constructed from* all *actions in $\mathcal{X}_t$, i.e. $V_{t+1}^{[K]} = \sum_{k\in[K]} \sum_{s=1}^{t} X_{k,s} X_{k,s}^\top$, instead of using only the actions played at previous time steps. This is made possible in their analyses by resorting to a doubly robust imputation of unobserved rewards associated to unplayed actions, which is significantly different from our approach. One theoretical benefit of their method is that the sequence $(V_{t+1}^{[K]})_{t\in\mathbb{N}}$ can be more easily transformed into a $\mathcal{G}$-martingale using only the unconditional lower bound on the covariance, as opposed to the conditional one of Assumption 5.22. Of note, Li et al. (2017) also questions the feasibility of a linear lower bound on the smallest eigenvalue of $V_{t+1}^0$ but concludes that it requires more stringent assumption as Lai and Wei (1982, Example 1) seemingly provides a counterexample of sublinear growth in the context of a regression problem. However, this counterexample studies autoregressive actions instead of i.i.d. action sets, which leads to $\mathbb{E}\left[X_t X_t^\top \mid \bar{\mathcal{G}}_{t-1}\right] \to 0$ when $t \to +\infty$. Therefore, this is different from what we consider in Assumption 5.22 and does not invalidate our analysis.*

## D.3 Regret analysis of LinUCB-OGD-CR

We derive here the correction to the regret upper bound when using online gradient descent (OGD) instead of exactly solving the empirical risk minimisation Z-estimation problem.

### Proof of Proposition 5.25

*Proof of Proposition 5.25.* Let $j \leqslant n$. The uniform lower bound on the Hessian of $\ell_j$ makes it $a$-strongly convex, which implies

$$\ell_j(z_j) - \ell_j(\bar{z}) \leqslant \langle \nabla \ell_j(z_j), z_j - \bar{z}_n \rangle - \frac{a}{2} \|\bar{z} - z_j\|^2.$$

By definition of the OGD scheme, the following holds:

$$\begin{aligned} \|z_{j+1} - \bar{z}_n\|^2 &= \|\Pi\left(z_j - \varepsilon_j \nabla \ell_j(z_j)\right) - \bar{z}_n\|^2 \\ &\leqslant \|z_j - \varepsilon_j \nabla \ell_j(z_j) - \bar{z}_n\|^2 \quad \text{(projection onto a convex set)} \\ &\leqslant \|z_j - \bar{z}_n\|^2 + \varepsilon_j^2 \|\nabla \ell_j(z_j)\|^2 - 2\varepsilon_j \langle \nabla \ell_j(z_j), z_j - \bar{z}_n \rangle \,, \end{aligned} \tag{D.37}$$

from which we deduce

$$\langle \nabla \ell_j(z_j), z_j - \bar{z}_n \rangle \leqslant \frac{\|z_j - \bar{z}_n\|^2 - \|z_{j+1} - \bar{z}_n\|^2}{2\varepsilon_j} + \frac{\varepsilon_j}{2}\|\nabla \ell_j(z_j)\|^2 \,. \tag{D.38}$$

**Bounded Gradients.** This case is covered by Theorem 3.3 in Hazan et al. (2016). We reproduce the proof here for reference and as a first step toward the more general setting of sub-Gaussian gradients.

Let $G > 0$ be such that $\|\nabla \ell_j(z_j)\| \leqslant G$ for all $j = 1, \ldots, n$. This allows to upper bound the above equation, leading to

$$\langle \nabla \ell_j(z_j), z_j - \bar{z}_n \rangle \leqslant \frac{\|z_j - \bar{z}_n\|^2 - \|z_{j+1} - \bar{z}_n\|^2}{2\varepsilon_j} + \frac{\varepsilon_j}{2} G^2 \,. \tag{D.39}$$

The online regret of OGD is therefore

$$\sum_{j=1}^n \ell_j(z_j) - \ell_j(\bar{z}_n) \leqslant \frac{1}{2}\sum_{j=1}^n \frac{\|z_j - \bar{z}_n\|^2 - \|z_{j+1} - \bar{z}_n\|^2}{\varepsilon_j} - a\|z_j - \bar{z}_n\|^2 + \frac{G^2}{2}\sum_{j=1}^n \varepsilon_j \,. \tag{D.40}$$

The first sum can be rewritten after a simple index shift and the convention $1/\varepsilon_0 := 0$:

$$\begin{aligned}
&\frac{1}{2}\sum_{j=1}^n \frac{\|z_j - \bar{z}_n\|^2 - \|z_{j+1} - \bar{z}_n\|^2}{\varepsilon_j} - a\|z_j - \bar{z}_n\|^2 \\
&= \frac{1}{2}\sum_{j=1}^n \|z_j - \bar{z}_n\|^2 \left(\frac{1}{\varepsilon_j} - \frac{1}{\varepsilon_{j-1}} - a\right) - \frac{1}{\varepsilon_n}\|z_{n+1} - \bar{z}_n\|^2 \\
&\leqslant \frac{1}{2}\sum_{j=1}^n \|z_j - \bar{z}_n\|^2 \left(\frac{1}{\varepsilon_j} - \frac{1}{\varepsilon_{j-1}} - a\right) \\
&= 0
\end{aligned} \tag{D.41}$$

for the choice $\varepsilon_j = \frac{1}{aj}$. Consequently, the online regret can be simplified as

$$\begin{aligned}
\sum_{j=1}^n \ell_j(z_j) - \ell_j(\bar{z}_n) &\leqslant \frac{G^2}{2}\sum_{j=1}^n \varepsilon_j \\
&= \frac{G^2}{2a}\sum_{j=1}^n \frac{1}{j} \\
&\leqslant \frac{G^2}{2a}(1 + \log n) \,.
\end{aligned} \tag{D.42}$$

**Sub-Gaussian Gradients.** We do not assume here that $\nabla \ell_j(z_j)$ is uniformly bounded, but instead rely on the weaker assumption that $\nabla \ell_j(z^*)$ is sub-Gaussian. The strategy is to control the variation between $\nabla\ell_j(z_j)$ and $\nabla\ell_j(z^*)$ on the one hand, and bound in high probability $\nabla\ell_j(z^*)$ on the other hand.

Notice that $\nabla\ell_j(z_j) = g_j + \frac{\alpha}{n}z^* + \nabla\ell_j(z_j) - \nabla\ell_j(z^*)$ and that there exists $\bar{z}_n \in [z_j, z^*] \subset \mathcal{C}$ such that $\nabla\ell_j(z_j) - \nabla\ell_j(z^*) = \nabla^2\ell_j(\bar{z}_n)\,(z_j - z^*)$ thanks to the mean value theorem and the convexity of $\mathcal{C}$. This yields

$$\begin{aligned}\|\nabla\ell_j(\phi_j)\|^2 &\leqslant 3\|g_j\|^2 + \frac{3\alpha^2}{n^2}\|z^*\|^2 + 3\|\nabla\ell_j(z_j) - \nabla\ell_j(z^*)\|^2 \\ &\leqslant 3\|g_j\|^2 + \frac{3\alpha^2}{n^2}\|z^*\|^2 + 3A^2\|z_j - z^*\|^2\,, \end{aligned} \tag{D.43}$$

since $\nabla\ell_j$ is $A$-Lipschitz. Combining this with the above yields

$$\begin{aligned}\langle\nabla\ell_j(z_j), z_j - \bar{z}_n\rangle &\leqslant \frac{3}{2}\frac{\|z_j - \bar{z}_n\|^2 - \|z_{j+1} - \bar{z}_n\|^2}{\varepsilon_j} + \frac{3}{2}\varepsilon_j\|g_j\|^2 + \frac{3}{2}\varepsilon_j\frac{\alpha^2}{n^2}\|z^*\|^2 + \frac{3}{2}\varepsilon_j A^2\|z_j - z^*\|^2 \\ &\leqslant \frac{3}{2}\frac{\|z_j - \bar{z}_n\|^2 - \|z_{j+1} - \bar{z}_n\|^2}{\varepsilon_j} + \frac{3}{2}\varepsilon_j\left(\|g_j\|^2 + \frac{\alpha^2}{n^2}\|z^*\|^2\right) + \frac{3}{2}\varepsilon_j A^2\mathrm{diam}(\mathcal{C})^2\,. \end{aligned} \tag{D.44}$$

The online regret of OGD is therefore

$$\begin{aligned}\sum_{j=1}^n \ell_j(z_j) - \ell_j(\bar{z}_n) \leqslant& \frac{3}{2}\sum_{j=1}^n \frac{\|z_j - \bar{z}_n\|^2 - \|z_{j+1} - \bar{z}_n\|^2}{\varepsilon_j} - \frac{a}{3}\|z_j - \bar{z}_n\|^2 \\ &+ \frac{3}{2}A^2\mathrm{diam}(\mathcal{C})^2\sum_{j=1}^n\varepsilon_j + \frac{3}{2}\sum_{j=1}^n\varepsilon_j\left(\|g_j\|^2 + \frac{\alpha^2}{n^2}\|z^*\|^2\right)\,. \end{aligned} \tag{D.45}$$

As in the bounded case, the choice $\varepsilon_j = \frac{3}{aj}$ makes the first sum vanish. Moreover, a simple union argument over the Chernoff bound for the $\sigma$-sub-Gaussian random variables $(g_j)_{j=1,\ldots,n}$ reveals that

$$\mathcal{E}_n = \left\{\forall j = 1,\ldots,n,\ \|g_j\| \leqslant \sigma\sqrt{2d\log\frac{2dn}{\delta}}\right\} \tag{D.46}$$

holds with probability at least $1-\delta$ for $\delta \in (0,1)$. Therefore, the following holds with probability at least $1-\delta$:

$$\sum_{j=1}^n \varepsilon_j\|g_j\|^2 \leqslant \sum_{j=1}^n \varepsilon_j\|g_j\|^2\mathbb{1}_{\mathcal{E}_n} \leqslant 2d\sigma^2\log\frac{2dn}{\delta}\sum_{j=1}^n\varepsilon_j\,. \tag{D.47}$$

Therefore, with probability at least $1-\delta$, we obtain the following online regret:

$$\begin{aligned}\sum_{j=1}^{n} \ell_j(z_s) - \ell_j(\bar{z}_n) &\leqslant \frac{3}{2}\left(2d\sigma^2 \log\frac{2dn}{\delta} + A^2\mathrm{diam}(\mathcal{C})^2 + \frac{\alpha^2}{n^2}\|z^*\|^2\right)\sum_{j=1}^{n}\varepsilon_j \\ &= \frac{9}{2a}\left(2d\sigma^2 \log\frac{2dn}{\delta} + A^2\mathrm{diam}(\mathcal{C})^2 + \frac{\alpha^2}{n^2}\|z^*\|^2\right)\sum_{j=1}^{n}\frac{1}{j} \\ &\leqslant \frac{9}{2a}\left(2d\sigma^2 \log\frac{2dn}{\delta} + A^2\mathrm{diam}(\mathcal{C})^2 + \frac{\alpha^2}{n^2}\|z^*\|^2\right)(1+\log n)\,. \qquad \text{(D.48)}\end{aligned}$$

■

## Proof of Theorem 5.26

*Proof of Theorem 5.26.* Let $j, h, T \in \mathbb{N}$, we define the aggregated episodic loss mapping as

$$\begin{aligned}\ell_j \colon \mathbb{R} &\longrightarrow \mathbb{R}_+ \\ \theta &\longmapsto \sum_{k=1}^{h} \mathcal{L}(Y_{(j-1)h+k}, \langle \theta, X_{(j-1)h+k}\rangle) + \frac{\alpha}{2N}\|\theta\|_2^2\,, \qquad \text{(D.49)}\end{aligned}$$

where $N = \lceil \frac{T-1}{h} \rceil$ denotes the total number of episodes of length $h$. For simplicity, we assume that $\bar{\theta}_t = \widehat{\theta}_t$ for all $t \leqslant T$, i.e. the empirical risk minimiser is always in the stable set of the projection operator $\Pi$. We recall that $\sum_{j=1}^{n} \nabla \ell_j(\bar{\theta}_t) = 0$ for $n = \frac{t-1}{h}$ (i.e. after episode $n$, when $\widehat{\theta}_n^{\mathrm{OGD}}$ is updated). In the general case, replacing $\widehat{\theta}_t$ by $\bar{\theta}_t$ induces an extra correction factor in the inequalities below which is at most polylogarithmic in $T$ (a consequence of Corollary 5.13), and hence does not change the conclusion. Again, we point out that, similarly to the generalised linear bandit setting (Filippi et al., 2010; Faury et al., 2020), $\widehat{\theta}_t$ is often in the stable set of $\Pi$ in practice.

We use the notations of Proposion 5.25 and define:

$$z_j = \widehat{\theta}_j^{\mathrm{OGD}}, \quad \bar{z}_n = \bar{\theta}_t \quad \text{and } z^* = \theta^*\,. \qquad \text{(D.50)}$$

We also denote by $\tilde{z}_n = \bar{\theta}_n^{\mathrm{OGD}} = \frac{1}{n}\sum_{j=1}^{n} z_j$ the average of the past $n$ OGD updates.

**Bound on $\|\tilde{z}_n - \bar{z}_n\|_2$.** Without loss of generality, we assume here that $\partial\mathcal{L}^*$ is a $\sqrt{m}\sigma$ sub-Gaussian process (this follows in variety of settings from the discussion of Assumption 5.10 and Lemma 5.11; the conclusions are essentially unchanged when assuming only Assumption 5.10, at the cost of slightly heavier notations).

We first note that $\nabla \ell_j(\theta^*) = g_j(\theta^*) + \frac{\alpha}{N}\theta^*$, where $g_j(\theta^*) = \sum_{k=1}^{h} \partial\mathcal{L}^*_{(j-1)h+k}$ and $j \in [N]$, is $\sqrt{hm}\sigma$-sub-Gaussian (sum of $h$ random variables, each of them being drawn from a $\sqrt{m}\sigma$-sub-Gaussian distribution). Setting the episode length to $h = \lceil \frac{2\varepsilon_h}{\rho_{\mathcal{X}} L^2} + \frac{8}{\rho_{\mathcal{X}}^2} \log \frac{2}{\delta} \rceil$ makes the one-step losses $\ell_j$ $m\varepsilon_h$-strongly convex with high probability. Indeed, let us define for $h' \in \mathbb{N}$ the function $f(h') = \frac{\rho_{\mathcal{X}} L^2}{2} h' - \frac{4L^2}{\rho_{\mathcal{X}}} \log \frac{2}{\delta}$ and the event $\mathcal{E}_{j,h'} = \left\{ \lambda_{\min} \left( \sum_{k=1}^{h} X_{(j-1)h'+k} X_{(j-1)h'+k}^\top \right) > f(h') \right\}$. First, notice that $\mathcal{E}_{j,h} \supseteq \bigcap_{h' \in \mathbb{N}} \mathcal{E}_{j,h'}$ and that $\sum_{k=1}^{h} X_{(j-1)h'+k} X_{(j-1)h'+k}^\top$ has the same distribution as $V_{h'+1}^0$ by Assumption 5.22. We deduce from Lemma D.4 applied to $V_{h+1}^0$ (that is with $\beta = 0$ and $m = 1$), that

$$\mathbb{P}\left(\mathcal{E}_{j,h}\right) \geqslant \mathbb{P}(\bigcap_{h' \in \mathbb{N}} \mathcal{E}_{j,h'}) = \mathbb{P}\left( \forall h' \in \mathbb{N},\ \lambda_{\min}\left(V_{h'+1}^0\right) > -\frac{4L^2}{\rho_{\mathcal{X}}} \log \frac{2}{\delta} + \frac{\rho_{\mathcal{X}} L^2}{2} h' \right) \geqslant 1 - \delta\,. \tag{D.51}$$

In particular for the value of $h$ defined above, $\mathcal{E}_{j,h} \subseteq \left\{ \lambda_{\min} \left( \sum_{k=1}^{h} X_{(j-1)h'+k} X_{(j-1)h'+k}^\top \right) \geqslant \varepsilon_h \right\}$, which implies the $m\varepsilon_h$-strong convexity of $\ell_j$ by the usual minoration $\partial^2 \mathcal{L} \geqslant m$ (Assumption 5.8). In the rest of this proof, we assume to be on the event $\bigcap_{j \in [N]} \mathcal{E}_{j,h}$, the probability of which is at least $1 - N\delta$ by a simple union argument.

Now, we apply the bound on the OGD regret of Proposition 5.25 with $a = m\varepsilon_h$, $A = hML^2$, namely that the *good event*

$$\forall n \leqslant N,\ \sum_{j=1}^{n-1} \ell_j(z_j) - \ell_j(\bar{z}_n) \leqslant \frac{C' dh\sigma^2}{\varepsilon_h} \log\left(\frac{2dN}{\delta}\right) \log(n)\,, \tag{D.52}$$

holds with probability at least $1-\delta$, for some constant $C' > 0$ (in which we hide the dependency on $h, M, L, \alpha$ and $S$ to avoid further cluttering). We assume to be on this event in the rest of the proof, which we combine to the previous events with a union argument, leading to a probability ot at least $1 - (N+1)\delta$.

The crux of the argument is similar to the proof of Lemma 2 in (Ding et al., 2021) and exploits the strong convexity of the losses $\ell_j$ to relate the online regret to a control on the distance $\|\tilde{z}_n - \bar{z}_n\|$. By Jensen's inequality, we have

$$\sum_{j=1}^{n} \ell_j(\tilde{z}_n) - \ell_j(\bar{z}_n) \leqslant \frac{Cdh\sigma^2}{\varepsilon_h} \log\left(\frac{2dN}{\delta}\right) \log(n)\,. \tag{D.53}$$

Strong convexity also implies the following inequality:

$$\ell_j(\tilde{z}_n) - \ell_j(\bar{z}_n) \geqslant \langle \nabla \ell_j(\bar{z}_n), \tilde{z}_n - \bar{z}_n \rangle + \frac{m\varepsilon_h}{2} \|\tilde{z}_n - \bar{z}_n\|_2^2\,. \tag{D.54}$$

Summing over $j = 1, \ldots, n$ and exploiting the fact that the sum of gradients vanishes at $\bar{z}_n$, we obtain after some simple algebra:

$$\|\tilde{z}_n - \bar{z}_n\|_2^2 \leqslant \frac{2Cdh\sigma^2}{m\varepsilon_h^2 n} \log\left(\frac{2dN}{\delta}\right) \log(n)\,. \tag{D.55}$$

**Regret Analysis of LinUCB-OGD.** Mirroring the regret proof of LinUCB, we see that we need

$$\forall t \leqslant T,\ \Delta(X_t, \bar{\theta}_n^{\mathrm{OGD}}) \leqslant \gamma_t(X_t) \tag{D.56}$$

to hold with high probability, for a certain exploration sequence $(\gamma_t)_{t\in\mathbb{N}}$. This amount to controlling the following norm:

$$\begin{aligned}
\|\theta^* - \bar{\theta}_n^{\mathrm{OGD}}\|_{\bar{H}_t^\alpha(\theta^*, \bar{\theta}_n^{\mathrm{OGD}})} &\leqslant \|\theta^* - \bar{\theta}_t\|_{\bar{H}_t^\alpha(\bar{\theta}_t, \bar{\theta}_n^{\mathrm{OGD}})} + \|\bar{\theta}_n^{\mathrm{OGD}} - \bar{\theta}_t\|_{\bar{H}_t^\alpha(\bar{\theta}_t, \bar{\theta}_n^{\mathrm{OGD}})} \\
&\leqslant \|\theta^* - \bar{\theta}_t\|_{\bar{H}_t^\alpha(\theta^*, \bar{\theta}_n^{\mathrm{OGD}})} + \sqrt{M}\|\bar{\theta}_n^{\mathrm{OGD}} - \bar{\theta}_t\|_{V_t^{\frac{\alpha}{m}}} \\
&\leqslant \|\theta^* - \bar{\theta}_t\|_{\bar{H}_t^\alpha(\theta^*, \bar{\theta}_n^{\mathrm{OGD}})} + \sqrt{M\left(L^2 t + \frac{\alpha}{m}\right)}\|\bar{\theta}_n^{\mathrm{OGD}} - \bar{\theta}_t\|_2\,.
\end{aligned} \tag{D.57}$$

The first term can be controlled by transportation of local metrics in the same way as in the proof of Theorem 5.18, i.e.

$$\|\theta^* - \bar{\theta}_t\|_{\bar{H}_t^\alpha(\theta^*, \bar{\theta}_n^{\mathrm{OGD}})} \leqslant \sqrt{\kappa}\left(\|F_t^\alpha(\theta^*) - F_t^\alpha(\widehat{\theta}_t)\|_{H_t^\beta(\theta^*)^{-1}} + \|F_t^\alpha(\bar{\theta}_t) - F_t^\alpha(\widehat{\theta})\|_{H_t^\beta(\bar{\theta}_t)^{-1}}\right), \tag{D.58}$$

and thus this term adds the same contribution to the design of the exploration bonus sequence and to the regret bound.

For the second term, we apply the previous bound on $\|\tilde{z}_n - \bar{z}_n\|_2 = \|\bar{\theta}_n^{\mathrm{OGD}} - \bar{\theta}_t\|_2$. Combining these two inequalities results in the following control:

$$\|\theta^* - \bar{\theta}_n^{\mathrm{OGD}}\|_{\bar{H}_t^\alpha(\theta^*, \bar{\theta}_n^{\mathrm{OGD}})} \leqslant \underbrace{c_t^\delta}_{\substack{\text{same as in} \\ \text{the proof of} \\ \text{Theorem 5.18}}} + \underbrace{\sqrt{\left(L^2 + \frac{\alpha}{mMt}\right)\left(\frac{2\kappa C d h^2\sigma^2}{\varepsilon_h^2} \log\left(\frac{2dT}{h\delta}\right)\log\left(\frac{t}{h}\right)\right)}}_{c_{t,T}^{\mathrm{OGD},\delta} = \mathcal{O}\left(\sigma L\sqrt{\kappa d}\log T\right)}\,. \tag{D.59}$$

The rest of the proof is now identical to that of Theorem 5.23, i.e. we use the exploration bonus sequence

$$\begin{aligned} \gamma_{t,T}^{\mathrm{OGD}} \colon \mathcal{X}_t &\longrightarrow \mathbb{R}_+ \\ x &\longmapsto (c_t^\delta + c_{t,T}^{\mathrm{OGD},\delta}) \|x\|_{H_t^{\kappa\alpha}(\bar\theta^{\mathrm{OGD}}_{\lfloor \frac{t-1}{h} \rfloor})^{-1}}, \end{aligned} \tag{D.60}$$

and control $\sum_{t=1}^{T} \|X_t\|_{H_t^{\kappa\alpha}(\bar\theta^{\mathrm{OGD}}_{\lfloor \frac{t-1}{h} \rfloor})^{-1}}$ using an elliptical potential lemma (since we operate under Assumption 5.22 for the strong convexity of the episodic losses $\ell_j$, we use the strong regret guarantee provided by Lemma D.2; note that the high probability event on which this lemma applies is already included in the above events, for a total probability of at least $1-(N+1)\delta$). ■

**Remark D.6** (The importance of strong convexity). *The key argument behind the proof of Theorem 5.26 is that the aggregated episodic loss $\ell_n$ is $m\varepsilon_h$ strongly convex. With simple convexity only, the online regret guarantee of the OGD approximation scales like $\mathcal{O}(\sqrt{T})$ instead of logarithmically. This would only guarantee $\|\bar\theta_n^{OGD} - \bar\theta_t\|_2 = \mathcal{O}(\sqrt{t})$, resulting in linear $\mathcal{O}(T)$ bandit regret after multiplying this term with the contribution of the elliptic potential lemma. Moreover, although $\ell_n$ is always trivially at least $\frac{\alpha}{N}$-strongly convex, it is necessary to ensure non-vanishing strong convexity when $T \to +\infty$ (we recall that $N = \lceil \frac{T-1}{h} \rceil$). Indeed, substituting $\varepsilon_h$ with $\frac{\alpha}{N}$ in the regret bound above gives $\mathcal{R}_T \leqslant \mathcal{O}(\varepsilon_h^{-1}) = \mathcal{O}(T)$.*

**Scaling of episode length $h$.** As shown in the proof, non-vanishing strong convexity of $\ell_n$ can be deduced from a fixed lower bound on the smallest eigenvalue of the Hessian of $\ell_n$. By Lemma D.4, this holds with high probability provided the episode length $h$ is high enough, which translates to $h = \lceil \frac{2\varepsilon_h}{\rho_{\mathcal{X}} L^2} + \frac{8}{\rho_{\mathcal{X}}^2} \log \frac{2}{\delta} \rceil$. Using the typical bound $\rho_{\mathcal{X}} = \mathcal{O}(d^{-1})$, we see that $h$ scales like $\mathcal{O}(d^2)$ in the action dimension. By comparison, the only similar OGD scheme for generalised linear bandit scales like $\mathcal{O}(d^3)$ (Ding et al., 2021, Lemma 2 and Remark 2), thus suffering to a greater extent from the curse of dimensionality. Note the practical tradeoff on $h$ faced by the agent running Algorithm 4: the higher $h$, the more likely it is that the OGD estimator $\bar\theta^{\mathrm{OGD}}$ well approximates the true empirical risk minimiser $\bar\theta$ (because of stronger convexity of the episodic losses); however, it also means longer episodes and thus less frequent updates of $\bar\theta^{\mathrm{OGD}}$, i.e. less learning.

Note that the value of $h$ is derived from a concentration bound that is *uniform* in $h$ (Lemma D.4). However, since $h$ is kept constant throughout the run of the algorithm, a similar, non-uniform result would actually be sufficient (we chose to use Lemma D.4 mainly for the sake of conve-nience, since we already assumed to be on the corresponding good event in order to mirror the regret analysis of Theorem 5.23). It is actually possible to tighten the lower bound on $h$ using a finer, non-uniform concentration result, which we state below.

**Lemma D.7** (Tighter bound on the episode length $h$). *Under Assumptions 5.17 and 5.22, let $\varepsilon_h > 0$, $\delta \in (0,1)$ and define*

$$h = \left\lceil \frac{1}{4\rho_{\mathcal{X}}^2} \left( \sqrt{2(1+\gamma_\delta)\log\left(\frac{2}{\delta}\sqrt{1+\frac{1}{\gamma_\delta}}\right)} + \sqrt{\sqrt{2(1+\gamma_\delta)\log\left(\frac{2}{\delta}\sqrt{1+\frac{1}{\gamma_\delta}}\right)} + \frac{\rho_{\mathcal{X}}\varepsilon_h}{L^2}} \right)^2 \right\rceil , \tag{D.61}$$

*where $\gamma_\delta = \frac{-1}{1+W_{-1}(-\frac{\delta^2}{4e})}$ and $W_{-1}$ is the first lower branch of the Lambert W function, i.e. the smallest real solution for $z = [-\frac{1}{e}, 0)$ of the equation $W_{-1}(z)e^{W_{-1}(z)} = z$. Then we have that*

$$\mathbb{P}\left(\lambda_{\min}\left(\sum_{s=1}^{h} X_s X_s^\top\right) \geqslant \varepsilon_h\right) \geqslant 1-\delta\,. \tag{D.62}$$

*Proof of Lemma D.7.* The idea is similar to that of the proof of Lemma D.4, i.e. relate the deviation of the smallest eigenvalue to that of a matrix martingale. The difference lies in the choice of the concentration bound for this martingale. Fix $\eta > 0$, Howard et al. (2021, Corollary S1(a)) with a normal mixture bound shows that a martingale $(M_t)_{t\in\mathbb{N}}$ taking values in a Hilbert space $(\mathcal{H}, \langle\cdot,\cdot\rangle)$ with uniformly bounded increments $\|M_{t+1} - M_t\| \leqslant c$ for some $c > 0$ and all $t \in \mathbb{N}$ satisfies

$$\mathbb{P}\left(\exists t \in \mathbb{N},\ \|M_t\| \geqslant c\sqrt{2(t+\eta)\log\left(\frac{2}{\delta}\sqrt{1+\frac{t}{\eta}}\right)}\right) \leqslant \delta\,. \tag{D.63}$$

Now, we fix $h \in \mathbb{N}$. Although this bound is uniform in $t \in \mathbb{N}$, we can make use of the free parameter $\eta$ to optimise it for $t = h$. This procedure is standard (see e.g. Howard et al. (2021, Proposition 3)) and yields $\eta = \gamma_\delta h$ with $\gamma_\delta = \frac{-1}{1+W_{-1}(-\frac{\delta^2}{4e})}$. In particular, we have

$$\mathbb{P}\left(\|M_h\| \geqslant c\sqrt{2h(1+\gamma_\delta)\log\left(\frac{2}{\delta}\sqrt{1+\frac{1}{\gamma_\delta}}\right)}\right) \leqslant \delta\,. \tag{D.64}$$

Following the steps of Lemma D.4, we have that

$$\mathbb{P}\left(\lambda_{\min}\left(V_{h+1}^0\right) \leqslant \varepsilon_h\right) \leqslant \mathbb{P}\left(\|M_h\| \geqslant \rho_{\mathcal{X}} L^2 h - \varepsilon_h\right) \leqslant \delta\,, \tag{D.65}$$

with $M_h = V_{h+1}^0 - \mathbb{E}\left[V_{h+1}^0\right]$, which defines a martingale with increments bounded by $c = 2L^2$. Equating both bounds on $\|M_h\|$ yields

$$\rho_{\mathcal{X}} L^2 h - 2L^2 \sqrt{2h(1+\gamma_\delta)\log\left(\frac{2}{\delta}\sqrt{1+\frac{1}{\gamma_\delta}}\right)} - \varepsilon_h = 0\,. \tag{D.66}$$

This expression is a quadratic equation in $H = \sqrt{h}$. Let us define the shorthand $a = \rho_{\mathcal{X}} L^2$ and $b = L^2\sqrt{2h(1+\gamma_\delta)\log\left(\frac{2}{\delta}\sqrt{1+\frac{1}{\gamma_\delta}}\right)}$. Notice that both $a$ and $b$ are positive and that the discriminant of $aH^2 + 2bH - \varepsilon_h = 0$ is $4b^2 + 4a\varepsilon_h$, which is also positive. The only positive solution is thus given by $\sqrt{h} = H = \frac{b+2\sqrt{b^2+a\varepsilon_h}}{2a}$, which concludes the proof. ■

The value of $h$ recommended by Lemma D.7 scales with $\rho_{\mathcal{X}}^{-2}$ at the first order, and thus $h = \mathcal{O}(d^2)$, just like in Theorem 5.23. However, it is typically smaller, and thus more practical as it allows for more frequent OGD updates with the same theoretical guarantees. We report in Figure D.1 the numerical values for both expressions of $h$ for typical choices of the parameters $L$, $\varepsilon_h$ and $\delta$ as a function the dimension $d$. In practice, even smaller values of $h$ may ensure enough convexity of the episodic losses to observe sublinear regret, which we empirically witnessed in the experiments (see Section 5.4). For the practitioner, $h$ may be viewed as a hyperparameter to be tuned manually, potentially on an instance-dependent basis, with Theorem 5.23 and Lemma D.7 giving only worst case guarantees.

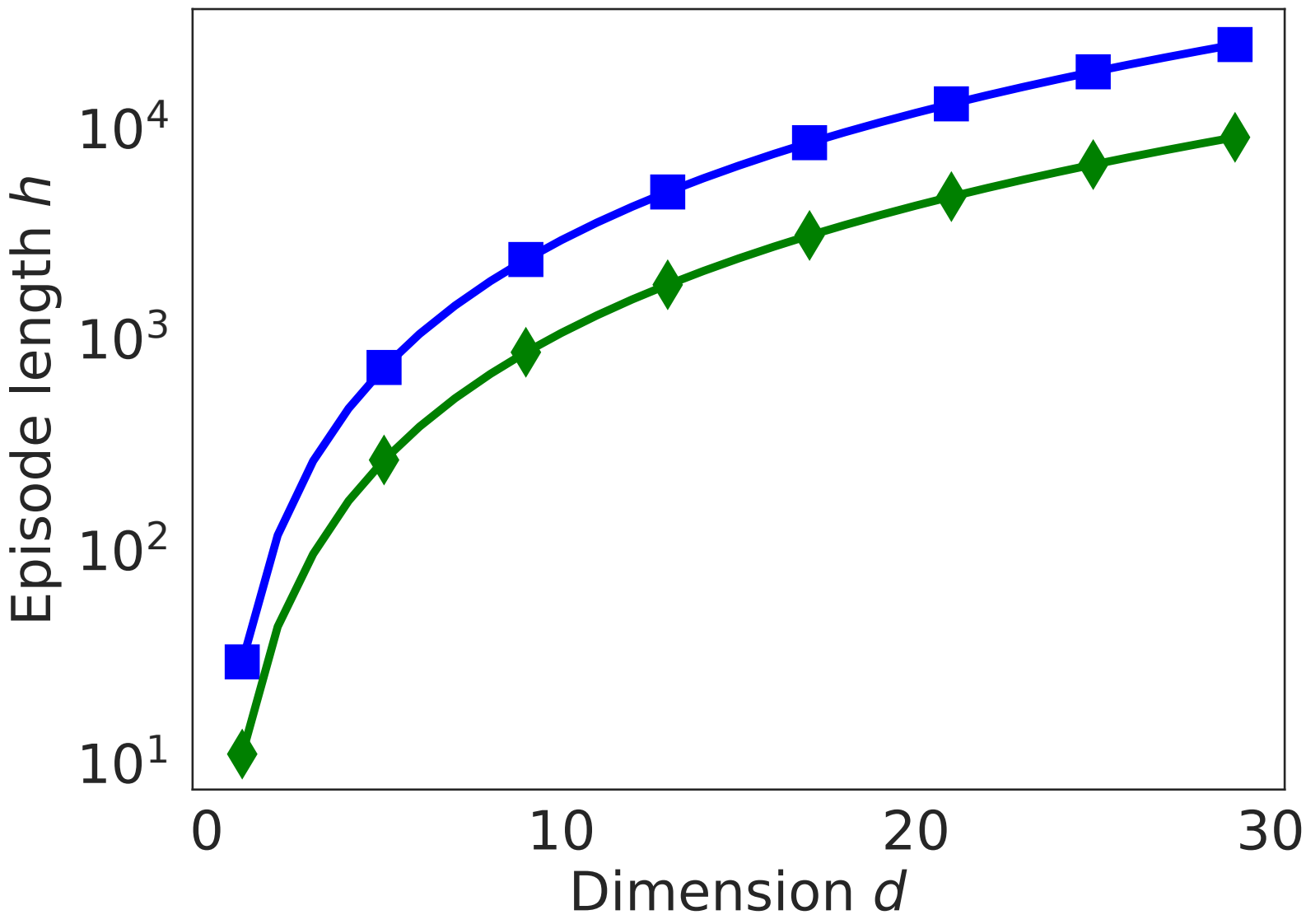

**Figure D.1 –** Recommended episode length $h = \lceil \frac{2\varepsilon_h}{\rho_\mathcal{X} L^2} + \frac{8}{\rho_\mathcal{X}^2} \log \frac{2}{\delta} \rceil$ for LinUCB-OGD-CR (Algorithm 4) according to Theorem 5.26 (blue) and Lemma D.7 (green) as a function of $d = \rho_\mathcal{X}^{-1}$, for typical values $L = 1$, $\varepsilon_h = 0.1$ and $\delta = 5\%$.

# Appendix E

# From optimality to robustness: Dirichlet sampling strategies in stochastic bandits

This chapter presents supplementary material for Chapter 6.

**Contents**

## E.1 Generic regret analysis for round-based randomised algorithms

In this section we derive the generic regret upper bound of Theorem 6.3, for any (randomised) index function $\widetilde{\mu}\colon \mathbb{R}^{(\mathbb{N})} \times \mathbb{R} \to \mathbb{R}$ (not necessarily a DS index in the sense of Definition 6.1). We exploit the round-based structure of Algorithm 5 to split the regret into two terms that depend on the DS index and control each of them separately. Note that this bound holds for *any* DS index, This analysis crucially rely on Assumption 6.2, which we recall ensures concentration of the empirical means around the true expectation for all arms $k \in [K]$ at an appropriate rate $I_k$.

*Proof of Theorem 6.3.* We recall that we denote by $k^\star \in [K]$ the optimal arm (assumed to be unique, without loss of generality), $k \in [K] \setminus \{k^\star\}$ a suboptimal arm, $\mu_k$ and $\mu^\star$ their respective mean rewards, $r \in \mathbb{N}$ a round of the algorithm, $\ell_r$ and $\mathcal{A}_{r+1}$ the leader and set of arms pulled after round $r$, $N_r^k$ and $N_r^{k^\star}$ the number of pulls to arm $k$ and $k^\star$ up to round $r$, and $\widehat{\mu}_r^k$ and $\widehat{\mu}_r^{k^\star}$ the empirical mean of the rewards observed for arm $k$ and $k^\star$ up to round $r$, respectively.

**Regret decomposition.** Thanks to the round-based structure, the fact that an arm is pulled or not depends of its status as a leader or a challenger. If an arm is a challenger, it can be pulled only if it wins its duel against the leader. For this reason, a natural first regret decomposition consists in considering the cases when (i) the optimal arm $k^\star$ is the leader but some suboptimal arms are pulled, and (ii) the optimal arm is not the leader. We upper bound the $T$-round cumulative expected pseudo regret $\mathbb{E}[\mathcal{R}_T]$ by controlling the expected number of pulls of each suboptimal arm $k$. Using the fact that all arms are pulled during the first round, we obtain the following inequality:

$$\begin{aligned}
\mathbb{E}[N_T^k] &= 1 + \mathbb{E}\left[\sum_{r=1}^{T-1} \mathbb{1}_{k\in\mathcal{A}_{r+1}}\right] \\
&= 1 + \sum_{r=1}^{T-1} \mathbb{E}\left[\mathbb{1}_{k\in\mathcal{A}_{r+1},\ell_r=k^\star} + \mathbb{1}_{k\in\mathcal{A}_{r+1},\ell_r\neq k^\star}\right] \\
&\leqslant 1 + \underbrace{\mathbb{E}\left[\sum_{r=1}^{T-1} \mathbb{1}_{k\in\mathcal{A}_{r+1},\ell_r=k^\star}\right]}_{n_k(T)} + \underbrace{\mathbb{E}\left[\sum_{r=1}^{T-1} \mathbb{1}_{\ell_r\neq k^\star}\right]}_{E_T} .
\end{aligned} \tag{E.1}$$

At this step, we have already extracted the term $n_k(T)$ of Theorem 6.3 and introduced a term $E_T$ that contains both $B_{T,\varepsilon}$ and $C_{\nu,\varepsilon}$. In the rest of this proof we work on the term $E_T$.

**Upper bound on $E_T^k$.** The following part of the proof is inspired by the proof of the SSMC algorithm in Chan (2020). Furthermore, we will see that we can further decompose $E_T^k$ into several events that will be handled using Assumption 6.2, showing the interest of the round-based structure. After that, we will finally exhibit the term that motivated Assumption 6.4.

Before that, we start by analising two alternatives that can cause the event $\{\ell_r \neq k^\star\}$, namely (i) arm $k^\star$ has already been leader and has lost the leadership at some point (leadership takeover), or (ii) arm $k^\star$ has never been leader. To formalise these alternatives we define the sequence $(a_r)_{r\in\mathbb{N}} = (\lceil r/4 \rceil)_{r\in\mathbb{N}}$ and the event

$$\mathcal{D}_r = \{\exists s \in [a_r, r] : \ell_s = k^\star\} . \tag{E.2}$$

**$\mathcal{D}_r$ is true: arm $k^\star$ has already been leader after round $a_r$.** We first justify the choice of $a_r$: starting after a number of rounds that is linear in $r$ ensures a number of observations for the leader during the whole segment $[a_r, r]$ that is also linear in $r$, that is for all $s \in [a_r, r]$,

$$N_s^{\ell^s} \geqslant \lceil a_r/K \rceil =: b_r \,. \tag{E.3}$$

Under $\mathcal{D}_r$ we study the probability of a leadership takeover by a suboptimal arm between $a_r$ and $r$. Such takeover can happen only if (i) a suboptimal arm obtains the same number of samples as arm $k^\star$, and (ii) its empirical mean is larger than the one of arm $k^\star$ at the round when it happens. We formalise the leadership takeover with the following events:

$$\begin{aligned}
\{\ell_r \neq k^\star\} \cap \mathcal{D}_r &\subset \bigcup_{s=a_r}^{r-1} \{\ell_s = k^\star, \ell_{s+1} \neq k^\star\} \\
&\subset \bigcup_{s=a_r}^{r-1} \bigcup_{k\in[K]\setminus\{k^\star\}} \left\{\ell_s = k^\star, k \in \mathcal{A}_{s+1}, N_{s+1}^k = N_{s+1}^{k^\star}, \widehat{\mu}_{s+1}^k \geqslant \widehat{\mu}_{s+1}^{k^\star}\right\} \\
&\subset \bigcup_{s=a_r}^{r-1} \bigcup_{k\in[K]\setminus\{k^\star\}} \left\{N_{s+1}^k = N_{s+1}^{k^\star}, \widehat{\mu}_{s+1}^k \geqslant \widehat{\mu}_{s+1}^{k^\star}\right\} \\
&= \bigcup_{s=a_r+1}^{r} \bigcup_{k\in[K]\setminus\{k^\star\}} \left\{N_s^k = N_s^{k^*}, \widehat{\mu}_s^k \geqslant \widehat{\mu}_s^{k^\star}\right\} \\
&=: \bigcup_{s=a_r+1}^{r} \bigcup_{k\in[K]\setminus\{k^\star\}} \mathcal{B}_s^k \,.
\end{aligned} \tag{E.4}$$

We denote by $\mathbb{Y}_{(n)}^k = (Y_i^k)_{i=1}^n \sim \nu_k^{\otimes n}$ (respectively $\mathbb{Y}_{(n)}^\star = (Y_i^\star)_{i=1}^n \sim \nu_{k^\star}^{\otimes n}$) the list of the first $n \in \mathbb{N}$ random rewards of arm $k$ (respectively of arm $k^\star$) and by $\widehat{\mu}_{(n)}^k$ (respectively $\widehat{\mu}_{(n)}^{k^\star}$) the corresponding empirical means. We first develop and express the sum of these terms as

$$\begin{aligned}
\mathbb{E}\left[\sum_{r=1}^{T-1} \sum_{s=a_r+1}^{t} \mathbb{1}_{\mathcal{B}_s^k}\right] &= \mathbb{E}\left[\sum_{r=1}^{T-1} \sum_{s=a_r+1}^{t} \mathbb{1}_{N_s^k = N_s^{k^\star}, \widehat{\mu}_s^k \geqslant \widehat{\mu}_s^{k^\star}}\right] \\
&\leqslant \mathbb{E}\left[\sum_{r=1}^{T-1} \sum_{s=a_r+1}^{r} \sum_{n=\lceil s/K \rceil}^{s} \mathbb{1}_{N_s^k = N_s^{k^\star} = n, \widehat{\mu}_{(n)}^k \geqslant \widehat{\mu}_{(n)}^{k^\star}}\right] \\
&\leqslant \mathbb{E}\left[\sum_{r=1}^{T-1} \sum_{s=a_r+1}^{r} \sum_{n=\lceil s/K \rceil}^{s} \mathbb{1}_{\widehat{\mu}_{(n)}^k \geqslant \widehat{\mu}_{(n)}^{k^\star}}\right] .
\end{aligned} \tag{E.5}$$

We now define $x_k = \frac{\mu_{k^\star} + \mu_k}{2}$. If $\widehat{\mu}_{(n)}^k \geqslant \widehat{\mu}_{(n)}^{k^\star}$, then either the arm $k^\star$ has underperformed or arm $k$ has overperformed, which means that $\widehat{\mu}_{(n)}^k \geqslant x_k$ or $\widehat{\mu}_{(n)}^{k^\star} \leqslant x_k$, which gives

$$\mathbb{E}\left[\sum_{r=1}^{T-1} \sum_{s=a_r+1}^{t} \mathbb{1}_{\mathcal{B}_s^k}\right] \leqslant \mathbb{E}\left[\sum_{r=1}^{T-1} \sum_{s=a_t+1}^{t} \sum_{n=\lceil s/K \rceil}^{s} \mathbb{1}_{\left\{\widehat{\mu}_{(n)}^k \geqslant x_k\right\} \bigcup \left\{\widehat{\mu}_{(n)}^k \leqslant x_k\right\}}\right]$$

$$\leqslant \sum_{r=1}^{T-1} \sum_{s=a_t+1}^{t} \sum_{n=\lceil s/K \rceil}^{s} \left( \mathbb{P}\left(\widehat{\mu}^k_{(n)} \geqslant x_k\right) + \mathbb{P}\left(\widehat{\mu}^{k^\star}_{(n)} \leqslant x_k\right)\right)$$
$$\leqslant \sum_{r=1}^{T-1} r^2 \left( \mathbb{P}\left(\widehat{\mu}^k_{(b_r)} \geqslant x_k\right) + \mathbb{P}\left(\widehat{\mu}^{k^\star}_{(b_r)} \leqslant x_k\right)\right)$$
$$\leqslant \sum_{r=1}^{T-1} r^2 \left( e^{-b_r I_{k^\star}(x_k)} + e^{-b_r I_k(x_k)} \right)$$
$$= O(1)\,, \tag{E.6}$$

where the two last lines come from Assumption 6.2. Finally, summing over all suboptimal arms, we obtain:

$$\sum_{r=1}^{T-1} \mathbb{P}\left(\ell_r \neq k^\star, \mathcal{D}_r\right) \leqslant \sum_{k \in [K]\setminus\{k^\star\}} \sum_{r=1}^{T-1} r^2 \left( e^{-b_r I_{k^\star}(x_k)} + e^{-b_r I_k(x_k)} \right) =: C_{\nu,\varepsilon} = \mathcal{O}(1)\,. \tag{E.7}$$

In other words, the simple condition on the concentration of empirical means is enough to ensure that the cost of leadership takeover by suboptimal arms only contributes to a constant (in $T$) in the regret bound.

This convergent series is the first component of the term $C_{\nu,\varepsilon}$ in Theorem 6.3. We now consider the case when arm $k^*$ has never been leader.

**Upper bound when $\mathcal{D}_r$ is not true.** We now consider the complement event $\bar{\mathcal{D}}_r$. The idea here is to leverage the fact that if the optimal arm is not leader between $\lceil r/4 \rceil$ and $r$, then it has necessarily lost a lot of duels. We introduce the count of the number of duels lost by arm $k^\star$,

$$\forall s \in \{a_r, \dots, r\},\ \mathcal{C}^s = \{\exists k \neq k^\star, \ell_s = k, k^\star \notin \mathcal{A}_{s+1}\} \quad \text{and} \quad L_r = \sum_{s=a_r}^{r} \mathbb{1}(\mathcal{C}_s)\,, \tag{E.8}$$

where $\mathcal{C}_s$represents the event that at round $s$, arm $k^\star$ was a challenger and lost its duel. A direct adaptation from Chan (2020, equation 7.12) provides the following upper bound:

$$\mathbb{P}(\ell_r \neq k^\star, \bar{\mathcal{D}}_r) \leqslant \mathbb{P}(L_r \geqslant r/4)\,. \tag{E.9}$$

We then use Markov's inequality to obtain

$$\mathbb{P}(L_r \geqslant r/4) \leqslant \frac{\mathbb{E}(L_r)}{r/4} = \frac{4}{r} \sum_{s=a_r}^{r} \mathbb{P}(\mathcal{C}_s)\,. \tag{E.10}$$

We then remove the double sum on $s$ and $t$ by simply counting the occurrences of each term:

$$\sum_{r=1}^{T-1} \mathbb{P}(L_r \geqslant r/4) \leqslant \mathbb{E}\left[\sum_{r=1}^{T-1} \frac{4}{r} \sum_{s=a_r}^{r} \mathbb{1}_{\mathcal{C}_s}\right] \leqslant \mathbb{E}\left[\sum_{s=1}^{T-1} \mathbb{1}_{\mathcal{C}_s} \sum_{r=1}^{T-1} \frac{4}{r} \mathbb{1}_{s\in[a_r,r]}\right]. \tag{E.11}$$

From this step we can control independently the sum in $r$,

$$\sum_{r=1}^{T-1}\frac{4}{r}\mathbb{1}_{s\in[a_r,r]} = \sum_{r=1}^{T-1}\frac{4}{r}\mathbb{1}_{s\leqslant r}\mathbb{1}_{a_r\leqslant s} \leqslant \frac{4}{s}\sum_{r=1}^{T-1}\mathbb{1}_{a_r\leqslant s} \leqslant \frac{4}{s}\sum_{r=1}^{T-1}\mathbb{1}_{\lceil r/4\rceil\leqslant s} \leqslant \frac{4}{s}\sum_{r=1}^{T-1}\mathbb{1}_{r/4\leqslant s+1} \leqslant \frac{16(s+1)}{s} \leqslant 32\,.$$

With this result we obtain the following inequality:

$$\sum_{r=1}^{T-1} \mathbb{P}(L_r \geqslant r/4) \leqslant 32 \sum_{r=1}^{T-1} \mathbb{P}(\mathcal{C}_r)\,. \tag{E.12}$$

We then decompose $\mathcal{C}_r$ as the union of all the terms corresponding to each possible suboptimal leader, i.e.

$$\mathcal{C}_r = \bigcup_{j\in[K]\setminus\{k^\star\}} \{\ell_r = j, k^\star \notin \mathcal{A}_{r+1}\} =: \bigcup_{j\in[K]\setminus\{k^\star\}} \mathcal{C}_r^j\,. \tag{E.13}$$

We now fix a suboptimal leader $j$ and work on the term $\mathcal{C}_r^j$. We recall that arm $k^\star$ has two chances to win the duel: first with its empirical mean, and then with the randomised index. We first handle the case when the suboptimal leader could be overperforming, by writing for any $\varepsilon > 0$

$$\begin{aligned}\mathcal{C}_r^j \subset & \left\{\left|\widehat{\mu}_r^j - \mu_j\right| \geqslant \varepsilon, \ell_r = j\right\} \\ & \bigcup \left\{\left|\widehat{\mu}_r^j - \mu_j\right| \leqslant \varepsilon, \ell_r = j, \widehat{\mu}_r^{k^\star} \leqslant \widehat{\mu}_r^j, \widetilde{\mu}\left(\mathbb{Y}_r^\star, \widehat{\mu}_r^j\right) \leqslant \widehat{\mu}_r^j\right\}.\end{aligned} \tag{E.14}$$

We then upper bound the left-hand term using again the concentration of the empirical mean for the leader, and obtain

$$\begin{aligned}\sum_{r=1}^{T-1} \mathbb{P}\left(\underbrace{\left|\widehat{\mu}_r^j - \mu_j\right| \geqslant \varepsilon, \ell_r = j}_{=:\widetilde{\mathcal{C}}_r^j}\right) &= \sum_{r=1}^{T-1} \sum_{n_j=\lceil r/K\rceil} \mathbb{P}\left(\left|\widehat{\mu}_r^j - \mu_j\right| \geqslant \varepsilon, N_r^j = n_j\right) \\ &= O(1)\,.\end{aligned} \tag{E.15}$$

We then continue the analysis of $\widetilde{\mathcal{C}}_r^j$ by considering the number of samples of arm $k^\star$, and in particular whether $N_r^{k^\star} \geqslant n_j^\star(T)$ or not, for some new quantity $n_j^\star(T)$. The idea is to choose $n_j^\star(T)$ such that $\widetilde{\mathcal{C}}_r^j$ is unlikely for $n \geqslant n_j^\star(T)$ thanks to the first step of the duel with empirical means. Writing $C_j = \sum_{r=1}^{T-1} \mathbb{P}\left(\widetilde{\mathcal{C}}_r^j, \ell_r = j\right)$, we have that:

$$\begin{aligned}
C_j &\leqslant \sum_{r=1}^{T-1} \mathbb{P}\left(\widetilde{\mathcal{C}}_r^j, N_r^{k^\star} \geqslant n_j^\star(T), \ell_r = j\right) + \sum_{r=1}^{T-1} \mathbb{P}\left(\widetilde{\mathcal{C}}_r^j, N_r^{k^\star} \leqslant n_j^\star(T), \ell_r = j\right) \\
&\leqslant \sum_{r=1}^{T-1} \sum_{n=n_j^\star(T)}^{T-1} \mathbb{P}\left(\widehat{\mu}_{(n)}^{k^\star} \leqslant \mu_j + \varepsilon\right) + \sum_{r=1}^{T-1} \mathbb{P}\left(\widetilde{\mathcal{C}}_r^j, N_r^{k^\star} \leqslant n_j^\star(T), \ell_r = j\right) \\
&\leqslant \sum_{r=1}^{T-1} \mathbb{P}\left(\widetilde{\mathcal{C}}_r^j, N_r^{k^\star} \leqslant n_j^\star(T), \ell_r = j\right) + \mathcal{O}(1)\,, \qquad \text{(E.16)}
\end{aligned}$$

as long as $n_j^\star(T) \geqslant \frac{\log T}{I_{k^\star}(\mu_j+\varepsilon)}$. Now, let $n \in \mathbb{N}$ and define the event

$$\mathcal{H}_{r,n}^{j,\varepsilon} = \left\{N_r^{k^\star} = n, \left|\widehat{\mu}_r^j - \mu_j\right| \leqslant \varepsilon, \widehat{\mu}_{(n)}^{k^\star} \leqslant \widehat{\mu}_r^j, \widetilde{\mu}\left(\mathbb{Y}_{(n)}^\star, \widehat{\mu}_r^j\right) \leqslant \widehat{\mu}_r^j\right\}\,, \qquad \text{(E.17)}$$

and use it to write the following upper bound on the remaining sum:

$$\sum_{r=1}^{T-1} \mathbb{P}\left(\widetilde{\mathcal{C}}_r^j, N_r^{k^\star} \leqslant n_j^\star(T), \ell_r = j\right) \leqslant \sum_{r=1}^{T-1} \sum_{n=1}^{n_j^\star(T)} \mathbb{P}\left(\mathcal{H}_{r,n}^{j,\varepsilon}\right)\,. \qquad \text{(E.18)}$$

Using the classical rewriting of a sum of indicators (see e.g. Riou and Honda (2020)), we observe that the following equality holds:

$$\sum_{r=1}^{T-1} \mathbb{1}_{\mathcal{H}_{r,n}^{j,\varepsilon}} = \sum_{m=1}^{T-1} \mathbb{1}_{\left(\sum_{r=1}^{T-1} \mathbb{1}_{\mathcal{H}_{r,n}^{j,\varepsilon}} \geqslant m\right)}\,. \qquad \text{(E.19)}$$

Now, define as $(\tau_i^n)_{i=1}^m$ the sequence of stopping times (with respect to the natural filtration defined by the round-based structure) corresponding to the first $m \in \mathbb{N}$ rounds $r$ at which $\mathcal{H}_{r,n}^{j,\varepsilon}$ holds. If $\left\{\sum_{r=1}^{T-1} \mathbb{1}_{\mathcal{H}_{r,n}^{j,\varepsilon}} \geqslant m\right\}$ holds, then $\mathcal{H}_{\tau_i^n,n}^{j,\varepsilon}$ holds for any $i \leqslant m$ and all these $\tau_i^n$ are finite, which means that

$$\mathbb{1}_{\left(\sum_{r=1}^{T-1} \mathbb{1}_{\mathcal{H}_{r,n}^{j,\varepsilon}} \geqslant m\right)} \leqslant \prod_{i=1}^{m} \mathbb{1}_{\mathcal{H}_{\tau_i^n,n}^{j,\varepsilon}}\,. \qquad \text{(E.20)}$$

Finally, we denote the last term to upper bound by

$$D_{T,\varepsilon}^j := \sum_{n=1}^{n_j^\star(T)} \sum_{m=1}^{T-1} \mathbb{E}\left[\prod_{i=1}^{m} \mathbb{1}_{\mathcal{H}_{\tau_i^n,n}^{j,\varepsilon}}\right]$$

$$= \sum_{n=1}^{n_j^\star(T)} \sum_{m=1}^{T-1} \mathbb{E}_{\mathbb{Y}^\star_{(n)}} \left[ \prod_{i=1}^{m} \mathbb{P}\left( \widetilde{\mu}\left( \mathbb{Y}^\star_{(n)}, \widehat{\mu}^j_{(N^j_{\tau_i^n})} \right) \leqslant \widehat{\mu}^j_{(N^j_{\tau_i^n})} \,\middle|\, \mathbb{Y}^\star_{(n)} \right) \mathbb{1}_{\mathcal{H}^{j,\varepsilon}_{\tau_i^n,n}} \right] . \tag{E.21}$$

At this step, we notice that $\widehat{\mu}^j_{(N^j_{\tau_i^n})}$ is concentrated in a neighbourhood of $\mu_j$, and thus we can simply bound the above term by replacing this empirical mean with the supremum over this neighbourhood, i.e.

$$\begin{aligned}
D^j_{T,\varepsilon} &\leqslant \sum_{n=1}^{n_j^\star(T)} \sum_{m=1}^{T-1} \sup_{\mu \in [\mu_j - \varepsilon, \mu_j + \varepsilon]} \mathbb{E}_{\mathbb{Y}^\star_{(n)}} \left[ \mathbb{P}\left( \widetilde{\mu}\left( \mathbb{Y}^\star_{(n)}, \mu \right) \leqslant \mu \,\middle|\, \mathbb{Y}^\star_{(n)} \right)^m \mathbb{1}_{\widehat{\mu}^{k^\star}_{(n)} \leqslant \mu} \right] \\
&\leqslant \sum_{n=1}^{n_j^\star(T)} \sup_{\mu \in [\mu_j - \varepsilon, \mu_j + \varepsilon]} \mathbb{E}_{\mathbb{Y}^\star_{(n)}} \left[ \frac{\mathbb{P}\left( \widetilde{\mu}\left( \mathbb{Y}^\star_{(n)}, \mu \right) \leqslant \mu \,\middle|\, \mathbb{Y}^\star_{(n)} \right)}{\mathbb{P}\left( \widetilde{\mu}\left( \mathbb{Y}^\star_{(n)}, \mu \right) \geqslant \mu \,\middle|\, \mathbb{Y}^\star_{(n)} \right)} \mathbb{1}_{\widehat{\mu}^{k^\star}_{(n)} \leqslant \mu} \right] \qquad \text{(geometric series)} \\
&\leqslant \sum_{n=1}^{n_j^\star(T)} \sup_{\mu \in [\mu_j - \varepsilon, \mu_j + \varepsilon]} \mathbb{E}_{\mathbb{Y}^\star_{(n)}} \left[ \frac{\mathbb{1}_{\widehat{\mu}^{k^\star}_{(n)} \leqslant \mu}}{\mathbb{P}\left( \widetilde{\mu}\left( \mathbb{Y}^\star_{(n)}, \mu \right) \geqslant \mu \,\middle|\, \mathbb{Y}^\star_{(n)} \right)} \right] .
\end{aligned} \tag{E.22}$$

We conclude the proof by letting $B_{T,\varepsilon} := \sum_{j \in [K] \setminus \{k^\star\}} D^j_{T,\varepsilon}$. ■

## E.2 Technical results on Dirichlet distributions and BCP

We recall basic results on Dirichlet distributions that are used in the analysis of DS algorithms.

**Elementary properties of Dirichlet distributions** We consider the Dirichlet distribution $\mathrm{Dir}(\alpha)$ for some parameter $\alpha = (\alpha_i)_{i=1}^{n+1} \in (\mathbb{R}_+^\star)^{n+1}$. Let $W = (W_i)_{i=1}^{n+1}$ be a random variable drawn from the distribution $\mathrm{Dir}(\alpha)$. We first recall that $W$ takes its values in the simplex $\mathfrak{D}^n = \{w \in [0,1]^{n+1} : \sum_{i=1}^{n+1} w_i = 1\}$. The distribution admits the following density (with respect to the Lebesgue measure on $\mathfrak{D}^n$):

$$\begin{aligned}
f \colon \mathfrak{D}^n &\longrightarrow \mathbb{R}_+ \\
w &\longmapsto \frac{\Gamma\left(\sum_{i=1}^{n+1} \alpha_i\right)}{\prod_{i=1}^{n+1} \Gamma(\alpha_i)} \prod_{i=1}^{n+1} w_i^{\alpha_i - 1} ,
\end{aligned} \tag{E.23}$$

where $\Gamma$ denotes the Gamma function. In this chapter, we only consider integer values for the coefficients $\alpha$, and in particular for any $n \in \mathbb{N}$, $\Gamma(n+1) = n!$. Letting $N = \sum_{i=1}^{n+1} \alpha_i$, we obtain the more convenient form

$$f \colon \mathfrak{D}^n \longrightarrow \mathbb{R}_+$$

$$w \longmapsto \frac{(N-1)!}{\prod_{i=1}^{n+1}(\alpha_i - 1)!} \prod_{i=1}^{n+1} w_i^{\alpha_i - 1}\,. \tag{E.24}$$

Furthermore, for most DS algorithms, we further restrict $\alpha$ to be $\alpha = (1, \ldots, 1)$, in which case the Dirichlet distribution corresponds to the uniform distribution over $\mathfrak{D}^n$, i.e. $f(w) = n!$ for all $w \in \mathfrak{D}^n$ (note that this recovers the standard result that the volume of $\mathfrak{D}^n$ is $1/n!$).

For any $i \in \{1, \ldots, n+1\}$, the marginal distribution of $W_i$ follows a Beta distribution $\mathrm{Beta}(\alpha_i, N - \alpha_i)$, and in particular we have that

$$\mathbb{E}[W_i] = \frac{\alpha_i}{N} \quad \text{and} \quad \mathbb{V}[w_i] = \frac{\alpha_i(N - \alpha_i)}{N^2(N+1)}\,. \tag{E.25}$$

In the uniform case $\alpha = (1, \ldots, 1)$, we can interpret $W$ as a random, unbiased *reweighting* of $n+1$ observations. Indeed, for a given sequence $\mathbb{Y}_{n+1} = (y_i)_{i=1}^{n+1} \in \mathbb{R}^{n+1}$, consider the probability measure $\widehat{\nu}_{\mathbb{Y}_{n+1},W} = \sum_{i=1}^{n+1} W_i \delta_{y_i}$ (note that this defines a random variables taking values in the space $\mathcal{M}_1^+(\mathbb{R})$, i.e. a random probability measure). For a given realisation $W = w \in \mathfrak{D}^n$, sampling from $\widehat{\nu}_{\mathbb{Y}_{n+1},w}$ results in different frequencies of observation for each element of $\mathbb{Y}_{n+1}$ compared to the empirical measure $\widehat{\nu}_{\mathbb{Y}_{n+1}} = 1/n \sum_{i=1}^{n+1} \delta_{y_i}$ (where each element is equally likely). However, sampling multiple times from $W$ and then from $\widehat{\nu}_{\mathbb{Y}_{n+1},W}$ conditional on $W$ does result in equal frequencies on average thanks to equation E.25. Therefore, reweighting empirical measures with Dirichlet weights is a way to add noise (increase variance) while preserving the empirical frequencies of observation, which is especially convenient to generalise the Thompson sampling scheme.

We also use the following two classical properties of Dirichlet distributions, both using the relation between the Dirichlet and exponential distributions. We refer to standard probability textbooks for the proofs.

**Lemma E.1** (Exponential representation). *Let* $(R_i)_{i=1}^{n+1} \sim \bigotimes_{i=1}^{n+1} \mathcal{E}(\alpha_i)$ *be* $n+1$ *independent exponential random variables. Then* $(R_i / \sum_{j=1}^{n+1} R_j)_{i=1}^{n+1} \sim \mathrm{Dir}(\alpha)$.

The second property is a direct consequence of the first and states that a Dirichlet random vector can be aggregated into another Dirichlet random vector.

**Lemma E.2** (Aggregation property). *For $i \neq j$ in $\{1, \ldots, n+1\}$, if $W \sim \mathrm{Dir}(\alpha)$ then the vector of size $n$ defined as $W^{i,j} = (W_1, \ldots, \underbrace{W_i + W_j}_{i\text{-th}}, \ldots, W_{j-1}, W_{j+1}, \ldots, W_{n+1})$ is distributed as $W^{i,j} \sim \mathrm{Dir}((\alpha_1, \ldots, \underbrace{\alpha_i + \alpha_j}_{i\text{-th}}, \ldots, \alpha_{j-1}, \alpha_{j+1}, \ldots, \alpha_{n+1}))$ (putting the sum in the $i$-th slot and removing the $j$-th slot without changing the other indices).*

In particular, we will make use of this property in the proofs of Theorem 6.15 and 6.17, where we resort to a *discretisation* of observed rewards, i.e. grouping observations from a continuous distribution into equally size bins. The aggregation property shows that the DS indices built from true reward histories are still Dirichlet reweight indices after discretisation, with parameters summed within each bin.

**Boundary crossing probability** We now provide an alternative lower bound on $[BCP]$, which we recall is defined as:

$$[BCP]\,(\mathbb{Y}_{n+1}; \mu) = \mathbb{P}_{W \sim \mathcal{D}_{n+1}}\left(\sum_{i=1}^{n+1} W_i y_i \geqslant \mu\right), \tag{E.26}$$

where $\mu \in \mathbb{R}$, $\mathbb{Y}_{n+1} = (y_i)_{i=1}^{n+1} \in \mathbb{R}^{n+1}$ and $\mathcal{D}_{n+1}$ is the Dirichlet distribution with parameter $(1, \ldots, 1)$ of size $n+1 \in \mathbb{N}$. The next two results are essentially a rewriting of Riou and Honda (2020, Lemma 14).

**Lemma E.3** (Second lower bound for the BCP). *Let $\alpha = (\alpha_i)_{i=1}^n \in (\mathbb{N}^\star)^n$ and $\mu < \max \mathbb{Y}_{n+1}$. Define $\tilde{\alpha} = \alpha \sqcup (1) \in \mathbb{N}^{n+1}$ and $N = \sum_{i=1}^n \alpha_i$. Then, for any $\underline{w} \in \mathfrak{D}^n$ such that $\sum_{i=1}^{n+1} \underline{w}_i y_i \geqslant \mu$, we have*

$$\mathbb{P}_{W \sim \mathrm{Dir}(\tilde{\alpha})}\left(\sum_{i=1}^{n+1} W_i y_i \geqslant \mu\right) \geqslant \tilde{\alpha}_{i^\star} N! \frac{\underline{w}_{n+1}}{\underline{w}_{i^\star}} \prod_{i=1}^{n} \frac{\underline{w}_i^{\alpha_i}}{\alpha_i!}, \tag{E.27}$$

*where $i^\star = \operatorname*{argmax}_{i=1,\ldots,n+1} \mathbb{Y}_{n+1}$.*

*Proof of Lemma E.3.* We use the expression of the density of the Dirichlet distribution defined above, and consider the set $\mathcal{S} = \{w \in \mathfrak{D}^n,\ \sum_{i=1}^{n+1} w_i y_i \geqslant \mu\}$. Then, for any given $\underline{w} \in \mathcal{S}$, we define the set $\mathcal{S}_{\underline{w}} = \{w \in \mathfrak{D}^n,\ \forall i \neq i^\star,\ w_i \in [0, \underline{w}_i]\}$. The inclusion $\mathcal{S}_{\underline{w}} \subseteq \mathcal{S}$ is direct since transferring weights to the maximum can only increase the value of the weighted sum. Hence, $\mathbb{P}_{w\sim \mathrm{Dir}(\widetilde{\alpha})}(w \in \mathcal{S}_{\underline{w}})$ is a lower bound on $\mathbb{P}_{W\sim \mathrm{Dir}(\widetilde{\alpha})}(W \in \mathcal{S})$. We then obtain

$$\begin{aligned}
\mathbb{P}_{W\sim \mathrm{Dir}(\widetilde{\alpha})}\left(\sum_{i=1}^{n+1} W_i y_i \geqslant \mu\right) &\geqslant \mathbb{P}_{W\sim \mathrm{Dir}(\widetilde{\alpha})}(W \in \mathcal{S}_{\underline{w}}) \\
&\geqslant \frac{N!}{\prod_{i=1}^n (\alpha_i - 1)!} \int_0^{\underline{w}_1} \cdots \int_0^{\underline{w}_{n+1}} \prod_{i=1}^n w_i^{\alpha_i - 1} \prod_{i=1, i\neq i^\star}^{n+1} dw_i \\
&= \frac{N!}{\prod_{i=1, i\neq i^\star}^n \alpha_i!} \prod_{i=1}^n (\underline{w}_i)^{\alpha_i} \left(\frac{\underline{w}_{n+1}}{\underline{w}_{i^\star}}\right) \\
&= \frac{N!}{\prod_{i=1}^n \alpha_i!} \prod_{i=1}^n (\underline{w}_i)^{\alpha_i} \left(\widetilde{\alpha}_{i^\star} \frac{\underline{w}_{n+1}}{\underline{w}_{i^\star}}\right), \qquad \text{(E.28)}
\end{aligned}$$

which concludes the proof. ∎

A direct corollary of this result is to choose the weights $\underline{w}$ that maximise the lower bound, which gives as in Lemma 6.7 an expression with $\mathcal{K}_{\inf}^{\max \mathbb{Y}_{n+1}}$.

**Corollary E.4.** *Let $\alpha = (\alpha_i)_{i=1}^n \in (\mathbb{N}^\star)^n$ and $\mu < \max \mathbb{Y}_{n+1}$. Assume that $\max \mathbb{Y}_{n+1}$ is unique. Define $\widetilde{\alpha} = \alpha \sqcup (1) \in \mathbb{N}^n$, $N = \sum_{i=1}^n \alpha_i$ and $\widehat{\nu}_{\mathbb{Y}_{n+1}^-, \alpha/N}$ the multinomial distribution over $n$ points $\mathbb{Y}_{n+1}^- = \mathbb{Y}_{n+1} \setminus \{\max \mathbb{Y}_{n+1}\}$ with probabilities $(\alpha_i/N)_{i=1}^n$. We have that*

$$\mathbb{P}_{W\sim \mathrm{Dir}(\widetilde{\alpha})}\left(\sum_{i=1}^{n+1} W_i y_i \geqslant \mu\right) \geqslant \frac{N!}{\prod_{i=1}^n \alpha_i!} \exp\left(-N\left(\mathcal{K}_{\inf}^{\max \mathbb{Y}_{n+1}}\left(\widehat{\nu}_{\mathbb{Y}_{n+1}^-, \frac{\alpha}{N}}, \mu\right) + \mathbb{H}(\widehat{\nu}_{\mathbb{Y}_{n+1}^-, \frac{\alpha}{N}})\right)\right), \qquad \text{(E.29)}$$

*where $\mathbb{H}$ is the entropy.*

*Proof of Corollary E.4.* We use the same notation as in Lemma E.3. Consider two multinomial distributions $\nu, \nu'$ supported on $\mathbb{Y}_{n+1}^-$ and $\mathbb{Y}_{n+1}$ respectively, and with probabilities $p = (p_i)_{i=1}^n$ and $q = (q_i)_{i=1}^{n+1}$. Since $\mathbb{Y}_{n+1}^- \subset \mathbb{Y}_{n+1}$, their Kullback-Leibler divergence is simply

$$\mathrm{KL}(\nu \parallel \nu') = \sum_{i=1}^n p_i \log \frac{p_i}{q_i} = -\sum_{i=1}^n p_i \log(q_i) - \mathbb{H}(\nu). \qquad \text{(E.30)}$$

We use the result of Lemma E.3 with $p = \alpha/N$ and $q = \underline{w}$, and choose

$$\underline{w} = \operatorname*{argmin}_{w \in \mathcal{S}} \mathrm{KL}(\widehat{\nu}_{\mathbb{Y}^-_{n+1}, \frac{\alpha}{N}} \parallel \widehat{\nu}_{\mathbb{Y}_{n+1}, w}) \,, \tag{E.31}$$

such that

$$\mathrm{KL}(\widehat{\nu}_{\mathbb{Y}^-_{n+1}, \frac{\alpha}{N}} \parallel \widehat{\nu}_{\mathbb{Y}_{n+1}, \underline{w}}) = \mathcal{K}_{\inf}^{\max \mathbb{Y}_{n+1}}(\widehat{\nu}_{\mathbb{Y}^-_{n+1}, \frac{\alpha}{N}}; \mu) \,, \tag{E.32}$$

Furthermore, we simplify the constants using the inequalities

$$\widetilde{\alpha}_{i^\star} \frac{\underline{w}_{n+1}}{\underline{w}_{i^\star}} \geqslant \frac{\underline{w}_{n+1}}{\underline{w}_{i^\star}} \geqslant 1 \,, \tag{E.33}$$

with equality only if $i^\star = n+1$. Indeed, the optimal allocation will necessarily put more weights on largest values, so $\underline{w}_{i^\star} \geqslant \underline{w}_{n+1}$. ■

Finally, we use Corollary E.4 to derive a bound on the ratio between the likelihood of an empirical distribution and its BCP in the case of multinomial distributions. This result will be useful in the proofs of Theorems 6.15 and 6.17 where we use it to analyse a discretisation of the random rewards; for this reason, we use the notation $\widetilde{\mathbb{Y}}_{S+1} = (\widetilde{y}_i)_{i=1}^{S+1} \in \mathbb{R}^{S+1}$ for $S \in \mathbb{N}$ instead of the usual $\mathbb{Y}_{n+1}$.

**Lemma E.5** (Balance between the likelihood and the BCP for multinomial distribution). *Assume that* $\max \widetilde{Y}_{S+1}$ *is unique. Let* $\mu < \max \widetilde{\mathbb{Y}}_{S+1}$ *and* $\widehat{\nu}_{\widetilde{\mathbb{Y}}^-_{S+1}, p_S}$ *the multinomial distribution supported on* $\widetilde{\mathbb{Y}}^-_{S+1}$ *with probabilities* $p_S \in \mathfrak{D}^{S-1}$. *We fix* $n \in \mathbb{N}$ *and denote by* $\beta_{n,S} \in \mathbb{N}^S$ *the random vector counting the occurrences of* $(\widetilde{Y}_i)_{i=1}^S$ *when drawing* $n$ *observations from* $\widehat{\nu}_{\widetilde{\mathbb{Y}}^-_{S+1}, p_S}$. *Then for any vector* $\beta \in \mathbb{N}^S$ *satisfying* $\sum_{i=1}^S \beta_i = n$ *and* $\sum_{i=1}^S \beta_i \widetilde{Y}_i \leqslant \mu$, *and letting* $\widetilde{\beta} = \beta \sqcup (1)$, *we have that*

$$\frac{\mathbb{P}(\beta_{n,S} = \beta)}{\mathbb{P}_{W \sim Dir(\widetilde{\beta})}\left(\sum_{i=1}^{S} W_i \widetilde{y}_i + W_{S+1}\widetilde{y}_{S+1} \geqslant \mu\right)} \leqslant \exp\left(-n\left[\mathrm{KL}(\widehat{\nu}_{\widetilde{\mathbb{Y}}^-_{S+1}, \frac{\beta}{n}} \parallel p_S) - \mathcal{K}_{\inf}^{\max \widetilde{\mathbb{Y}}_{S+1}}\left(\widehat{\nu}_{\widetilde{\mathbb{Y}}^-_{S+1}, \frac{\beta}{n}}; \mu\right)\right]\right) . \tag{E.34}$$

*Proof of Lemma E.5.* Dembo and Zeitouni (2009, Lemma 2.1.6) provides the following formula for multinomial distributions:

$$\mathbb{P}(\beta_{n,S} = \beta) = \frac{n!}{\prod_{i=1}^{S} \beta_i!} \prod_{i=1}^{S} p_{S,i}^{\beta_i} = \frac{n!}{\prod_{i=1}^{S} \beta_i!} \exp\left(-n\left[\mathrm{KL}(\widehat{\nu}_{\widetilde{\mathbb{Y}}^{-}_{S+1}, \frac{\beta}{n}} \parallel \widehat{\nu}_{\widetilde{\mathbb{Y}}^{-}_{S+1}, p_S}) + \mathbb{H}(\widehat{\nu}_{\widetilde{\mathbb{Y}}^{-}_{S+1}, \frac{\beta}{n}})\right]\right) , \tag{E.35}$$

where $\mathbb{H}$ denotes the entropy. Then, Corollary E.4 directly provides the result as all the constant terms are equal, and the entropy term can be simplified. ∎

## E.3 Regret analysis for DS algorithms ( )

In this section, we provide the complete proofs of Theorems 6.15, 6.19 and 6.17, presented in Section 6.3. We adopt the same proof strategy for each of them: starting from Theorem 6.3, which holds for all three of these DS instances, we then detail the terms $n_k(T)$ for all suboptimal arm $k \in [K] \setminus \{k^\star\}$ and $T \in \mathbb{N}$. which drives the first order terms of the regret upper bound. Then, we justify that Assumptions 6.2 and Assumption 6.4 hold, using respectively the light tailed properties of the distributions and the lower bound on $[BCP]$ (Lemma 6.8), avoiding underexploration of the best arm. We recall the definition of the main terms we have to analyse from Theorem 6.3:

$$\textbf{First order term} : \forall k \in [K] \setminus \{k^\star\},\ n_k(T) = \mathbb{E}\left[\sum_{r=1}^{T-1} \mathbb{1}(k \in \mathcal{A}_{r+1}, \ell_r = 1)\right] , \tag{E.36}$$

$$\textbf{Assumption 6.4} : \forall \varepsilon > 0, n^\star(T) = o(\log T),\ \sum_{n=1}^{n^\star(T)} \mathbb{E}\left[\frac{\mathbb{1}_{\widehat{\mu}^{k^\star}_{(n)} \leqslant \mu^\star - \varepsilon}}{\mathbb{P}\left(\widetilde{\mu}(\mathbb{Y}^\star_{(n)}; \mu^\star - \varepsilon) \geqslant \mu^\star - \varepsilon \mid \mathbb{Y}^\star_{(n)}\right)}\right] = o(\log T) . \tag{E.37}$$

The regret proofs of the three DS instances share common elements, so before instantiating the proof for each algorithm we further work on these two terms under general assumptions.

### General proof sketches

In this section we derive the parts of the proofs that are shared by all three instances of DS, namely the necessary conditions for Theorem 6.3 and Assumption 6.4.

**Further characterisation of $n_k(T)$.** We start by detailing further the term $n_k(T)$, which is known to drive the first order regret term in $T$ (Corollary 6.5).

**Lemma E.6.** *Let $k \in [K] \setminus \{k^\star\}$ a suboptimal arm. Assume that $\nu_k$ satisfies Assumption 6.2 and that for any $n \in \mathbb{N}$, there exists a subset $\mathcal{B}_n^k \subset \mathbb{R}^n$ and an increasing function $f_k$ such that for any $\mu < \mu^\star$, the following two conditions hold:*

$$\mathbb{Y}_{(n)}^k \subset \mathcal{B}_n^k \implies \mathbb{P}\left(\widetilde{\mu}(\mathbb{Y}_{(n)}^k, \mu) \geqslant \mu\right) \leqslant \exp\left(-f_k(n, \mathcal{B}_n^k, \mu)\right) , \tag{E.38}$$

$$\sum_{n=1}^{T-1} \mathbb{P}\left(\mathbb{Y}_{(n)}^k \notin \mathcal{B}_n^k\right) = \mathcal{O}(1) \, . \tag{E.39}$$

*Then we have*

$$n_k(T) = m_k(T) + \mathcal{O}(1) \, , \tag{E.40}$$

*where for any $\varepsilon > 0$ and $T \geqslant 1$, $m_k(T)$ satisfies $f_k\left(m_k(T), \mathcal{B}_n^k, \mu^\star - \varepsilon\right) = \log T$.*

*Proof of Lemma E.6.* We first split the sequence $(\mathbb{1}_{k \in \mathcal{A}_{r+1}, \ell_r = k^\star})_{r=1}^{T-1}$ into (i) a preconvergence phase, the size of which we control, and (ii) a postconvergence phase, for which a suboptimal arm $k \in [K] \setminus \{k^\star\}$ has been pulled enough times to be suitably concentrated according to Assumption 6.2. To this end, we define a (for now arbitrary) quantity $m_k(T) \in \mathbb{N}$ and write

$$\begin{aligned}
\mathbb{E}\left[\sum_{r=1}^{T-1} \mathbb{1}(k \in \mathcal{A}_{r+1}, \ell_r = k^\star)\right] &\leqslant \mathbb{E}\left[\sum_{r=1}^{T-1} \mathbb{1}_{k \in \mathcal{A}_{r+1}, \ell_r = k^\star, N_r^k < m_k(T)}\right] \\
&\quad + \mathbb{E}\left[\sum_{r=1}^{T-1} \mathbb{1}_{k \in \mathcal{A}_{r+1}, \ell_r = k^\star, N_r^k \geqslant m_k(T)}\right] \\
&\leqslant m_k(T) + \mathbb{E}\left[\sum_{r=1}^{T-1} \mathbb{1}_{k \in \mathcal{A}_{r+1}, \ell_r = k^\star, N_r^k \geqslant m_k(T)}\right] ,
\end{aligned} \tag{E.41}$$

where we used that for any $n \in \mathbb{N}$, the event $\{k \in \mathcal{A}_{r+1}, N_r^k = n\}$ may happen at most once during the run of the algorithm (if $k \in \mathcal{A}_{r+1}$, arm $k$ will be pulled, which we increase $N_r^k$). The next step is to further split the second term by defining a *good event* of large probability under which $\{k \in \mathcal{A}_{r+1}\}$ has a low probability. As we aim to keep some level of generality in this section, we simply define this event as

$$\mathcal{G}_r^k = \left\{\mathbb{Y}_r^k \in \mathcal{B}_{N_r^k}^k\right\} \bigcap \left\{\widehat{\mu}_r^{k^\star} \geqslant \mu^\star - \varepsilon\right\} , \tag{E.42}$$

where $\varepsilon > 0$. We further define the event

$$\mathcal{W}_r^k = \left\{k \in \mathcal{A}_{r+1}, \ell_r = k^\star, N_r^k \geqslant m_k(T)\right\} , \tag{E.43}$$

which allows use to write:

$$n_k(T) \leqslant m_k(T) + \mathbb{E}\left[\sum_{r=1}^{T-1} \mathbb{1}_{\mathcal{W}_r^k \cap \mathcal{G}_r^k} + \sum_{r=1}^{T-1} \mathbb{1}_{\mathcal{W}_r^k \cap \bar{\mathcal{G}}_r^k}\right] . \tag{E.44}$$

We use the first assumption in the lemma to upper bound the first term as

$$\begin{aligned}
\mathbb{E}\left[\sum_{r=1}^{T-1} \mathbb{1}_{\mathcal{W}_r^k, \mathcal{G}_r^k}\right] &= \mathbb{E}\left[\sum_{r=1}^{T-1} \sum_{n=m_k(T)}^{T-1} \mathbb{1}_{k \in \mathcal{A}_{r+1}, \ell_r = k^\star, N_r^k = n, \mathcal{G}_r^k}\right] \\
&\leqslant \mathbb{E}\left[\sum_{r=1}^{T-1} \sum_{n=m_k(T)}^{T-1} \mathbb{1}_{\widetilde{\mu}(\mathbb{Y}_r^k, \widehat{\mu}_r^{k^\star}) \geqslant \widehat{\mu}_r^{k^\star}, N_r^k = n, \mathcal{W}_r^k, \mathcal{G}_r^k}\right] \\
&\leqslant \mathbb{E}\left[\sum_{r=1}^{T-1} \sum_{n=m_k(T)}^{T-1} \mathbb{P}\left(\widetilde{\mu}\left(\mathbb{Y}_r^k, \widehat{\mu}_r^{k^\star}\right) \geqslant \widehat{\mu}_r^{k^\star} \;\middle|\; \mathbb{Y}_r^k, \mathbb{Y}_r^\star\right) \mathbb{1}_{N_r^k = n, \mathcal{W}_r^k, \mathcal{G}_r^k}\right] \\
&\leqslant \mathbb{E}\left[\sum_{r=1}^{T-1} \sum_{n=m_k(T)}^{T-1} \exp\left(-f\left(n, \mathcal{B}_n^k, \widehat{\mu}_r^{k^\star}\right)\right) \mathbb{1}_{k \in \mathcal{A}_{r+1}, \ell_r = k^\star, N_r^k = n, \mathcal{G}_r^k}\right] \\
&\leqslant \mathbb{E}\left[\sum_{r=1}^{T-1} \sum_{n=m_k(T)}^{T-1} \exp\left(-f\left(n, \mathcal{B}_n^k, \mu^\star - \varepsilon\right)\right) \mathbb{1}_{k \in \mathcal{A}_{r+1}, N_r^k = n}\right] ,
\end{aligned} \tag{E.45}$$

We complete this step of the proof by further using the monotonicity of $f$ in $n$,

$$\begin{aligned}
\mathbb{E}\left[\sum_{r=1}^{T-1} \mathbb{1}_{\mathcal{W}_r^k, \mathcal{G}_r^k}\right] &\leqslant \exp\left(-f(m_k(T), \mathcal{B}_{m_k(T)}^k, \mu^\star - \varepsilon)\right) \mathbb{E}\left[\sum_{r=1}^{T-1} \sum_{n=m_k(T)}^{T-1} \mathbb{1}_{k \in \mathcal{A}_{r+1}, N_r^k = n}\right] \\
&\leqslant \exp\left(-f(m_k(T), \mathcal{B}_{m_k(T)}^k, \mu^\star - \varepsilon)\right) \mathbb{E}\left[\sum_{n=m_k(T)}^{T-1} 1\right] \\
&\leqslant T \exp\left(-f(m_k(T), \mathcal{B}_{m_k(T)}^k, \mu^\star - \varepsilon)\right) .
\end{aligned} \tag{E.46}$$

For the second term, note that

$$\bar{\mathcal{G}}_r^k = \left\{\mathbb{Y}_r^k \notin \mathcal{B}_{N_k^r}^k\right\} \bigcup \left\{\widehat{\mu}_r^{k^\star} < \mu^\star - \varepsilon\right\} , \tag{E.47}$$

which leads to

$$\begin{aligned}
\mathbb{E}\left[\sum_{r=1}^{T-1} \mathbb{1}_{\mathcal{W}_r^k, \bar{\mathcal{G}}_r^k}\right] &= \mathbb{E}\left[\sum_{r=1}^{T-1} \mathbb{1}_{k \in \mathcal{A}_{r+1}, \ell_r = k^\star, N_r^k \geqslant m_k(T), \bar{\mathcal{G}}_r^k}\right] \\
&\leqslant \underbrace{\mathbb{E}\left[\sum_{r=1}^{T-1} \mathbb{1}_{\ell_r = k^\star, \widehat{\mu}_{k^\star, r} < \mu^\star - \varepsilon}\right]}_{A_1} + \underbrace{\mathbb{E}\left[\sum_{r=1}^{T-1} \mathbb{1}_{k \in \mathcal{A}_{r+1}, N_r^k \geqslant m_k(T), \mathbb{Y}_r^k \notin \mathcal{B}_{N_r^k}^k}\right]}_{A_2} ,
\end{aligned} \tag{E.48}$$

where the two terms $A_1$ and $A_2$ depend respectively only of arm $k^\star$ and arm $k$. The first term can be handled thanks to Assumption 6.2 on arm $k^\star$, and using that the leader has necessarily a linear number of samples, we obtain

$$\begin{aligned} A_1 \leqslant \sum_{r=1}^{T-1} \mathbb{E}\left[\mathbb{1}_{N_r^{k^\star} \geqslant \lceil r/K \rceil, \widehat{\mu}_r^{k^\star} < \mu^\star - \varepsilon}\right] \leqslant \sum_{r=1}^{T-1} \sum_{n=\lceil r/K \rceil}^{r} \mathbb{P}\left(\widehat{\mu}_r^{k^\star} < \mu^\star - \varepsilon\right) &\leqslant \sum_{r=1}^{T-1} \sum_{n=\lceil r/K \rceil}^{r} e^{-nI_{k^\star}(\mu^\star - \varepsilon)} \\ &\leqslant \sum_{r=1}^{T-1} r e^{-\lceil r/K \rceil I_{k^\star}(\mu^\star - \varepsilon)} \\ &= \mathcal{O}(1)\,. \end{aligned} \tag{E.49}$$

We now upper bound $A_2$, using again that $\sum_{r=1}^{T-1} \mathbb{1}_{k \in \mathcal{A}_{r+1}, N_r^k = n} \leqslant 1$ for any $n \in \mathbb{N}$:

$$\begin{aligned} A_2 &\leqslant \sum_{r=1}^{T-1} \sum_{n=m_k(T)}^{T-1} \mathbb{E}\left[\mathbb{1}_{k \in \mathcal{A}_{r+1}, N_r^k = n, \mathbb{Y}_r^k \notin \mathcal{B}_{N_r^k}^k}\right] \\ &\leqslant \sum_{n=m_k(T)}^{T-1} \mathbb{P}\left(\mathbb{Y}_{(n)}^k \notin \mathcal{B}_n^k\right)\,. \end{aligned} \tag{E.50}$$

Combining these results, we obtain a bound on $n_k(T)$ for arm $k$ as

$$n_k(T) \leqslant m_k(T) + T e^{-f(m_k(T), \mathcal{B}_{m_k(T)}^k, \mu^\star - \varepsilon)} + \sum_{n=m_k(T)}^{T-1} \mathbb{P}\left(\mathbb{Y}_{(n)}^k \notin \mathcal{B}_n^k\right) + \mathcal{O}(1)\,. \tag{E.51}$$

We see that if $\mathcal{B}_n^k$ is designed to make the series convergent, and $(m_k(T))_{T \in \mathbb{N}^\star}$ is chosen as a sequence satisfying $f(m_k(T), \mathcal{B}_{m_k(T)}^k, \mu^\star - \varepsilon) = \log T$ for all $T \in \mathbb{N}^\star$, then we finally obtain

$$n_k(T) \leqslant m_k(T) + \mathcal{O}(1)\,. \tag{E.52}$$

■

Thanks to this result, when we will adapt the proof for each algorithm we will be able to combine Lemma E.6 with Lemma 6.7 to directly look for a proper choice of the set $\mathcal{B}_n^k$.

### Regret bound for BDS

We recall that two settings are considered for BDS: (B1) distributions are supported in $[\underline{B}, \overline{B}]$ and the upper bound of the support is known, and (B2) the upper bound is not known but for each distribution $\nu_k$ it holds that $\mathbb{P}_{\nu_k}([\overline{B} - \gamma, \overline{B}]) \geqslant p$ for some known $\gamma, p$, which we denoted by $\nu_k \in \mathcal{F}_{\underline{B}, \overline{B}}^{\gamma, p}$ (to avoid cluttering, we assume all arms $k \in [K]$ have same parameters $\overline{B}$, $p$ and $\gamma$, without loss of generality). The first setting being essentially equivalent to that of Riou and Honda (2020), we refer to this paper for the proof and focus on the second case.

*Proof of Theorem 6.15.* We recall that for $\rho > 0$,

$$B^{\text{BDS}}(\mathbb{Y}^k_{(n)}; \mu, \rho) = (\max \mathbb{Y}^k_{(n)} + \gamma) \vee (\mu + \frac{\rho}{n} \sum_{i=1}^{n} (\mu - Y_i)_+) \,. \tag{E.53}$$

First of all, the bounded support hypothesis ensures that Assumption 6.2 holds thanks to Hoeffding's inequality, with a rate function $I_k \colon x \in \mathbb{R} \mapsto \frac{2(x-\mu_k)^2}{\overline{B}^2}$. We can now focus on the expression of $n_k(T)$ and on proving Assumption 6.4 holds.

**First order term.** We use Lemma E.6 for this. Fix $k \in [K] \setminus \{k^\star\}$ and $n \in \mathbb{N}$. To define the high probability event $\mathcal{B}^k_n$, we consider the Lévy metric

$$\begin{aligned} d \colon \mathcal{M}_1^+(\mathbb{R}) \times \mathcal{M}_1^+(\mathbb{R}) &\longrightarrow \mathbb{R}_+ \\ (\nu_F, \nu_G) &\longmapsto \inf \left\{ \varepsilon > 0 : G(x - \varepsilon) - \varepsilon \leqslant F(x) \leqslant G(x + \varepsilon) + \varepsilon \right\} \,, \end{aligned} \tag{E.54}$$

where $F$ and $G$ denote the c.d.f. of $\nu_F$ and $\nu_G$ respectively. We recall that this distance metrises the topology of weak convergence on $\mathcal{M}_1^+(\mathbb{R})$.

In this section, we fix some $\varepsilon > 0$ and we choose $\mathcal{B}^k_n$ to be the intersection of the subspace of empirical distributions supported by $n$ points and a Lévy ball of radius $\varepsilon$ around the true distribution, which we embed in $\mathbb{R}^n$ as follows:

$$\mathcal{B}^k_n = \left\{ \mathbb{Y} \in \mathbb{R}^n : d(\widehat{\nu}_{\mathbb{Y}}, \nu_k) \leqslant \varepsilon \right\} \,. \tag{E.55}$$

The objective is to use the continuity of the $\mathcal{K}_{\inf}^{\overline{B}}$ operator in its first argument with respect to the Lévy metric (Honda and Takemura, 2010). For the rest of this proof, we identify distributions and their c.d.f. and let $F^k$ be the c.d.f. of $\nu_k$, $\widehat{F}^k_n$ the c.d.f. of $\widehat{\nu}^k_{(n)}$ the empirical distribution associated with the set $\mathbb{Y}^k_{(n)}$ and $\widetilde{F}^k_n$ the c.d.f. of the *biased* empirical distribution to which the bonus of the BDS algorithm has been added, which we denote by $\widetilde{\nu}^k_{(n)}$. We first prove that $\widehat{F}^k_n$ belongs to the Lévy ball with high probability, using the relation between the Lévy metric and the supremum norm

$$d(\widehat{F}^k_n, F^k) \leqslant \left\| \widehat{F}^k_n - F^k \right\|_\infty \,, \tag{E.56}$$

which essentially states that the uniform convergence implies the weak convergence. We also use the Dvoretzky-Kiefer-Wolfowitz (DKW) inequality (see e.g. Massart (1990)), i.e.

$$\mathbb{P}\left( \left\| \widehat{F}^k_n - F^k \right\|_\infty \geqslant \varepsilon \right) \leqslant 2e^{-2n\varepsilon^2} \,. \tag{E.57}$$

Hence, the series $\sum_{n\in\mathbb{N}^\star} \mathbb{P}\left(d(\widehat{F}_n^k, F^k) \geqslant \varepsilon\right)$ converges. Assuming that the event $\|\widehat{F}_n^k - F^k\|_\infty \geqslant \varepsilon$ holds, we prove that the biased distribution $\widetilde{F}_n^k$ is also close to $F^k$ in the sense of the supremum norm. First, the triangular inequality provides $\|\widetilde{F}_n^k - F^k\|_\infty \leqslant \|\widetilde{F}_n^k - \widehat{F}_n^k\|_\infty + \|\widehat{F}_n^k - F^k\|_\infty$. Then, we notice that for any $x \in \mathbb{R}$, we have

$$|\widetilde{F}_n^k(x) - \widehat{F}_n^k(x)| \leqslant \frac{1}{n+1}\,, \tag{E.58}$$

and thus if $\|\widehat{F}_n^k - F^k\|_\infty \leqslant \varepsilon$, then $\|\widetilde{F}_n^k - F^k\|_\infty \leqslant \varepsilon + \frac{1}{n+1}$, and finally

$$\left\|\widehat{F}_n^k - F^k\right\|_\infty \leqslant \varepsilon \implies d(\widetilde{F}_n^k, F^k) \leqslant \varepsilon + \frac{1}{n+1}\,. \tag{E.59}$$

Hence, if we combine these results we obtain that for $n$ large enough $\widetilde{F}_n^k$ is also in a Levy ball around $F^k$, of size $\varepsilon'$ slightly larger than $\varepsilon$, with large probability.

Now that the event $\mathcal{B}_n^k$ is defined and we stated its properties, we can find the function $f$ in Lemma E.6 in the case of BDS. We denote by $Y_{n+1}^k = B^{\text{BDS}}(\mathbb{Y}_{(n+1)}^K; \mu^\star, \rho)$ the bonus of BDS and use Lemma 6.7 to obtain

$$[BCP](\mathbb{Y}_{(n+1)}^k; \mu^\star) \leqslant \exp\left(-(n+1)\mathcal{K}_{\text{inf}}^{B^{\text{BDS}}(\mathbb{Y}_{(n+1)}^k;\mu^\star,\rho)}\left(\widetilde{\nu}_{(n)}^k, \mu^\star\right)\right)\,. \tag{E.60}$$

Furthermore, under the event $\mathcal{B}_n^k$ and the fact that the mean of the leader is concentrated around its true mean, the bonus of the BDS index is upper bounded by $\mathfrak{B} + \varepsilon'$, for some $\varepsilon' > 0$ and

$$\mathfrak{B} = \left(\overline{B} + \gamma\right) \vee (\mu^\star + \rho\mathbb{E}_{Y\sim\nu_k}[(\mu^\star - Y)_+])\,. \tag{E.61}$$

We use the continuity of $\mathcal{K}_{\text{inf}}^{\mathfrak{B}}$ with respect to (i) the first argument in terms of the L'evy distance, (ii) the second argument (e.g. with respect to the Euclidian norm), (iii) the upper bound (Agrawal et al., 2021): for any $\varepsilon_0 > 0$, we can calibrate the $\varepsilon$ in the Lévy ball to obtain

$$[BCP](\mathbb{Y}_{(n+1)}^k; \mu^\star) \leqslant e^{-(n+1)(\mathcal{K}_{\text{inf}}^{\mathfrak{B}}(\nu_k,\mu^\star)-\varepsilon_0)}\,, \tag{E.62}$$

hence we conclude this part by setting exactly $m_k(T) = \frac{\log(T)}{\mathcal{K}_{\text{inf}}^{\mathfrak{B}}(\nu_k,\mu^\star)-\varepsilon_0}$.

**Assumption 6.4.** We now study the quantity

$$E_n = \mathbb{E}_{\mathbb{Y}_{(n)}^\star \sim \nu_{k^\star}^{\otimes n}}\left[\frac{\mathbb{1}_{\widehat{\mu}_{(n)}^{k\star} \leqslant \mu}}{\mathbb{P}\left(\widetilde{\mu}(\mathbb{Y}_{(n)}^\star; \mu) \geqslant \mu \;\middle|\; \mathbb{Y}_{(n)}^\star\right)}\right]\,, \tag{E.63}$$

for $\mu < \mu^\star$. We first use Lemma 6.8 to obtain the following lower bound:

$$[BCP](\mathbb{Y}^\star_{(n)}; \mu) \geqslant e^{-\frac{n}{\rho}}\,, \tag{E.64}$$

where $\rho$ is the leverage parameter. Using this results and Assumption 6.2, we obtain a first bound

$$E_n \leqslant e^{-n\left(I_{k^\star}(\mu) - \frac{1}{\rho}\right)}\,, \tag{E.65}$$

which is sufficient to ensure Assumption 6.4 if $I_1(\mu) \geqslant 1/\rho$. In the opposite case, we use the assumption of setting $B2$, i.e. $\mathbb{P}([B-\gamma, B]) \geqslant p$ and the second component of the exploration bonus, $\max \mathbb{Y}^\star_{(n)} + \gamma$ to obtain

$$\begin{aligned} E_n &\leqslant \mathbb{E}_{\mathbb{Y}^\star_{(n)} \sim \nu_{k^\star}^{\otimes n}}\left[\frac{\mathbb{1}_{\widehat{\mu}^{k^\star}_{(n)} \leqslant \mu}\left(\mathbb{1}_{\max \mathbb{Y}^\star_{(n)} \leqslant B-\gamma} + \mathbb{1}_{\max \mathbb{Y}^\star_{(n)} \geqslant B-\gamma}\right)}{\mathbb{P}\left(\widetilde{\mu}(\mathbb{Y}^\star_{(n)};\mu) \geqslant \mu \,\middle|\, \mathbb{Y}^\star_{(n)}\right)}\right] \\ &\leqslant \underbrace{(1-p)^n e^{\frac{n}{\rho}}}_{E_{n,1}} + \underbrace{\mathbb{E}_{\mathbb{Y}^\star_{(n)} \sim \nu_{k^\star}^{\otimes n}}\left[\frac{\mathbb{1}_{\widehat{\mu}^{k^\star}_{(n)} \leqslant \mu}\mathbb{1}_{\max \mathbb{Y}^\star_{(n)} + \gamma \geqslant B}}{\mathbb{P}\left(\widetilde{\mu}(\mathbb{Y}^\star_{(n)};\mu) \geqslant \mu \,\middle|\, \mathbb{Y}^\star_{(n)}\right)}\right]}_{E_{n,2}}\,. \end{aligned} \tag{E.66}$$

The two terms correspond to the two possible expressions for the bonus. The term $E_{n,1}$ gives the sufficient condition for the tuning of $\rho$ in Theorem 6.15 since the following implication holds:

$$\rho > \frac{-1}{\log(1-p)} \implies \sum_{n=1}^{+\infty} E_{n,1} = O(1)\,. \tag{E.67}$$

In the second term, the exploration bonus is larger than $\overline{B}$, so we can use the same proof scheme as in Riou and Honda (2020). For simplicity, we assume that $\underline{B} = 0$ and $\overline{B} = 1$ and observe that the general case may be recovered by a simple translation and scaling argument. First, we discretise the interval $[0,1]$ in $S = \lceil 1/\eta \rceil$ equally sized bins of size $\eta > 0$, and consider the truncated variables $\widetilde{\mathbb{Y}}_S = (\widetilde{Y}_i)_{i=1}^S = (\eta \lfloor Y_i/\eta \rfloor)_{i=1}^n$ (counting only the unique occurrences) and $\widetilde{Y}_{S+1} = \max \widetilde{\mathbb{Y}}_S + \gamma$. The parameter $\eta$ is chosen small enough to ensure that $\mu^\star - \eta > \mu$, i.e. the truncated distribution still has a mean larger than $\mu$. An upper bound of $E_{n,2}$ is obtained by replacing the sequence $(Y_i)_{i=1}^n$ by $(\widetilde{Y}_i)_{i=1}^S$. Now, we define $\beta_{n,S} \in \mathbb{N}^S$ the vector which counts the number of occurrences of the sequence $(Y_i)_{i=1}^n$ falling in each bin. We also fix $\beta \in \mathbb{N}^S$ and $\widetilde{\beta} = \beta \sqcup (1)$. The number of possible values for $\beta_{n,S}$ is upper bounded by $n^S$, and we use

Lemma E.5 to obtain

$$\frac{\mathbb{P}(\beta_{n,S}=\beta)}{\mathbb{P}_{W\sim \mathrm{Dir}(\widetilde{\beta})}\left(\sum_{i=1}^{S} W_i\widetilde{Y}_i + W_{S+1}\widetilde{Y}_{S+1}\geqslant \mu\right)} \leqslant \exp\left(-n\left[\mathrm{KL}(\widehat{\nu}_{\widetilde{\mathbb{Y}}_S,\frac{\beta}{n}} \parallel p_S) - \mathcal{K}_{\inf}^{\max \widetilde{\mathbb{Y}}_{S+1}}\left(\widehat{\nu}_{\widetilde{\mathbb{Y}}_S,\frac{\beta}{n}};\mu\right)\right]\right)$$
$$\leqslant \exp\left(-n\left[\mathrm{KL}(\widehat{\nu}_{\widetilde{\mathbb{Y}}_S,\frac{\beta}{n}} \parallel p_S) - \mathcal{K}_{\inf}^{1}\left(\widehat{\nu}_{\widetilde{\mathbb{Y}}_S,\frac{\beta}{n}};\mu\right)\right]\right)$$
$$\leqslant \exp\left(-n\left[\mathcal{K}_{\inf}^{1}\left(\widehat{\nu}_{\widetilde{\mathbb{Y}}_S,\frac{\beta}{n}};\mu^\star-\eta\right) - \mathcal{K}_{\inf}^{1}\left(\widehat{\nu}_{\widetilde{\mathbb{Y}}_S,\frac{\beta}{n}};\mu\right)\right]\right). \tag{E.68}$$

As $\mu < \mu^\star - \eta$, there exists $\kappa > 0$ such that for all $\beta \in \mathbb{N}^S$ satisfying $\sum_{i=1}^S \beta_i \widetilde{Y}_i < \mu$, we have

$$\mathcal{K}_{\inf}^{1}\left(\widehat{\nu}_{\widetilde{\mathbb{Y}}_S,\frac{\beta}{n}};\mu^\star-\eta\right) - \mathcal{K}_{\inf}^{1}\left(\widehat{\nu}_{\widetilde{\mathbb{Y}}_S,\frac{\beta}{n}};\mu\right) > \kappa\,. \tag{E.69}$$

Finally, denoting by $\mathcal{B}_S$ the set of the possible count vectors $\beta$, we have that

$$E_{n,2} \leqslant \sum_{\beta\in\mathcal{B}_S} e^{-n\left[\mathcal{K}_{\inf}^{1}\left(\widehat{\nu}_{\widetilde{\mathbb{Y}}_S,\frac{\beta}{n}};\mu^\star-\eta\right) - \mathcal{K}_{\inf}^{1}\left(\widehat{\nu}_{\widetilde{\mathbb{Y}}_S,\frac{\beta}{n}};\mu\right)\right]} \leqslant n^S e^{-n\kappa}\,. \tag{E.70}$$

The two components of the bonus ensures that Assumption 6.4 is satisfied if $\nu_k \in \mathcal{F}_{0,1}^{\gamma,p}$. By translation and scaling, this immediately extends to the full bandit model (B2) of Theorem 6.15, which concludes the proof. ■

### E.3.1 Regret bound for QDS

We recall that QDS is stated in the bandit model (Q): each distribution $\nu_k$ is lower bounded by $\underline{B} \in \mathbb{R}$ and satisfies, for $\alpha \in (0,1)$, $\rho \geqslant 0$:

$$\forall \mu > \mu_k,\ \mathcal{K}_{\inf}^{\mathcal{F}_{[\underline{B},+\infty)}}(\nu_k;\mu) \geqslant \mathcal{K}_{\inf}^{\mathfrak{M}_q^k}(\mathcal{T}_\alpha(\nu_k);\mu)\,, \tag{E.71}$$
$$\tag{E.72}$$

where $\mathcal{T}_\alpha$ is the truncation operator that replaces the upper $1-\alpha$-quantile by its $\mathrm{CVaR}_\alpha$, denoted by the $C_\alpha$ operator, and $\mathfrak{M}_q^k = (q_{1-\alpha}(\nu_k)) \vee (\mu^\star + \rho\mathbb{E}_{Y\sim\nu_k}[(\mu^\star - Y)_+])$.

*Proof of Theorem 6.17.* First, Assumption 6.2 holds since distributions are assumed to be light tailed. The rest of the proof is similar to the proof of Theorem 6.15 for BDS. We recall that

$$B^{\mathrm{QDS}}(\mathbb{Y}_{(n)}^k;\mu,\rho) = \mu + \frac{\rho}{n_\alpha - 1}\sum_{i=1}^{n_\alpha-1}\left(\mu - Y_{(i)}^k\right)_+ + \rho\left(\mu - C_\alpha(\widehat{\nu}_{(n)}^k)\right)_+\,, \tag{E.73}$$

where $n_\alpha = \lceil \alpha n \rceil / n$, $\widehat{\nu}^k_{(n)}$ is the empirical distribution of $\mathbb{Y}^k_{(n)}$, and $Y^k_{(1)} \leqslant \ldots \leqslant Y^k_{(n)}$.

**Upper bounding $n_k(T)$.** We again want to use Lemma E.6, and formulate a high-probability event on the empirical distributions. Fist, observe that the variable $(\mu - Y)_+$ for $Y \sim \nu_k$ is also light tailed and hence admits a good rate function $I_k$ thanks to Cramér's theorem (Dembo and Zeitouni, 2009, Theorem 2.2.3). Then we build a Lévy ball around the true distribution $\nu_k$ with additional control on the $\mathrm{CVaR}_\alpha$:

$$\mathcal{B}^k_n = \left\{ \mathbb{Y} \in \mathbb{R}^n : d(\widehat{\nu}_{\mathbb{Y}}, \nu_k) \leqslant \varepsilon, |B^{\mathrm{QDS}}(\mathbb{Y}; \rho, \mu^\star) - B^k_{\rho,\mu^\star}| \leqslant \varepsilon, C_\alpha(\widehat{\nu}_{\mathbb{Y}}) \leqslant C_\alpha(\nu_k) + \varepsilon \right\} , \quad \text{(E.74)}$$

for some $\varepsilon > 0$ and $B^k_{\rho,\mu^\star} = \mu^\star + \rho \times \mathbb{E}_{Y\sim\nu_k}[(\mu^\star - Y)_+]$.

As in the proof of Theorem 6.15, we need to show that the series $\sum_{n\in\mathbb{N}^\star} \mathbb{P}\left(\mathbb{Y}^k_{(i)} \notin \mathcal{B}^k_n\right)$ converges. To handle the additional conditions introduced in the definition of $\mathcal{B}^k_n$, we use the Wasserstein metric, defined for $p \geqslant 1$ as

$$\begin{aligned} W_p \colon \mathbb{L}^p(\mathbb{R}) \times \mathbb{L}^p(\mathbb{R}) &\longrightarrow \mathbb{R}_+ \\ (\nu, \nu') &\longmapsto \left( \inf_{\gamma\in\Gamma(\nu,\nu')} \int_{\mathbb{R}\times\mathbb{R}} |y - y'|^p \gamma(dy, dy') \right)^{\frac{1}{p}} , \end{aligned} \quad \text{(E.75)}$$

where $\Gamma(\nu, \nu')$ is the set of couplings of measures $\nu$ and $\nu'$, i.e. the set of $\gamma \in \mathcal{M}^+_1(\mathbb{R}^2)$ such that $\nu$ and $\nu'$ are the marginal distributions of $\gamma$. Compared to the Lévy metric, this one metrises simultaneously the weak convergence and the convergence of all moments of order up to $p$ (Villani, 2008, Theorem 6.9). In addition, Bhat and Prashanth (2019, Lemma 2) shows that for two distributions $\nu$ and $\nu'$:

$$|C_\alpha(\nu) - C_\alpha(\nu')| \leqslant \frac{1}{1-\alpha} W_1(\nu, \nu') , \quad \text{(E.76)}$$

and then Fournier and Guillin (2015, Theorem 2) provides a concentration result on this term, similarly to the use of DWK inequality in the proof of Theorem 6.15. Combining all these arguments together shows that the above series is convergent, and therefore $\mathcal{B}^k_n$ hold with high probability.

We now assume that the event $\mathbb{Y}^k_{(n)} \in \mathcal{B}^k_n$ holds. The difference compared to BDS is that the BCP is considered for the truncated distribution $\mathcal{T}(\widehat{\nu}^k_{(n)})$. However, this is not a problem as the upper bound of Lemma 6.7 still holds. Thanks to the aggregation properties of the Dirichlet distribution (see Appendix E.2), the BCP with parameter $(1, \ldots, 1, n_\alpha)$ (of size $n - n_\alpha$) is the same as the BCP with parameters $(1, \ldots, 1)$ (of size $n$) with $n_\alpha$ copies of the last term. Hence,

the QDS index satisfies

$$\mathbb{P}\left(\widetilde{\mu}(\mathbb{Y}^k_{(n)}, \mu^\star) \geqslant \mu^\star \;\middle|\; \mathbb{Y}^k_{(n)}\right) \leqslant \exp\left(-(n+1)\mathcal{K}_{\inf}^{\widehat{\mathfrak{M}}^k_{C,n}}(\mathcal{T}(\widehat{\nu}^k_{(n)}); \mu^\star)\right), \tag{E.77}$$

where $\widehat{\mathfrak{M}}^k_{C,n} = C_\alpha(\widehat{\nu}^k_{(n)}) \vee B^{\mathrm{QDS}}(\mathbb{Y}^k_{(n)}; \rho, \mu^\star)$. Under $\mathcal{B}^k_n$, letting $\mathfrak{M}^k_C = C_\alpha(\nu_k) \vee B^k_{\rho,\mu^\star}$, we have:

$$\widehat{\mathfrak{M}}^k_{C,n} \leqslant \mathfrak{M}^k_C + \varepsilon\,. \tag{E.78}$$

Finally, the definition of the Levy distance ensures that

$$d\left(\widehat{\nu}^k_{(n)}, \nu_k\right) \leqslant \varepsilon \implies d\left(\mathcal{T}_\alpha(\widehat{\nu}^k_{(n)}), \mathcal{T}_\alpha(\nu_k)\right) \leqslant \varepsilon\,. \tag{E.79}$$

Hence, we can use the continuity of $\mathcal{K}_{\inf}^{\widehat{\mathfrak{M}}^k_{C,n}}$ in all arguments (including $\widehat{\mathfrak{M}}^k_{C,n}$, see e.g. Honda and Takemura (2015)), and thus, for any $\varepsilon_0 > 0$, we can calibrate $\varepsilon$ so that

$$\mathbb{P}\left(\widetilde{\mu}(\mathbb{Y}^k_{(n)}, \mu^\star) \geqslant \mu^\star \;\middle|\; \mathbb{Y}^k_{(n)}\right) \leqslant \exp\left(-(n+1)\left(\mathcal{K}_{\inf}^{\mathfrak{M}^k_C}(\mathcal{T}(\nu_k); \mu^\star) - \varepsilon_0\right)\right), \tag{E.80}$$

which gives the first order regret term by setting $m_k(T) = \frac{\log T}{\mathcal{K}_{\inf}^{\mathfrak{M}^k_C}(\mathcal{T}(\nu_k);\mu^\star) - \varepsilon_0}$ in Lemma E.6.

**Assumption 6.4.** We now resort to the assumption that rewards are lower bounded by $\underline{B}$, which we assume to be $\underline{B} = 0$ for simplicity (again, the general case follows from a simple translation). Indeed, we can then find a value $\underline{y}$ and a discretisation step $\eta$ such that (i) truncating the sequence $(Y_i^k)_{i=1}^n$ to $(Y_i^k \wedge \underline{y})_{i=1}^n$ and (ii) projecting each of the $Y_i$ that are less than $\underline{y}$ to $\widetilde{Y}_i = \eta \lfloor Y_i/\eta \rfloor$ preserves the order of $\mu < \widetilde{\mu}^\star$. Note that this value $\underline{y}$ does not have to be known by the algorithm and is purely an artifact for the proof. This discretisation is similar to the proof of Theorem 6.15. We denote by $S$ the number of items created by the discretisation and $\beta_{n,S} \in \mathbb{N}^S$ the vector of counts. However, in contrast to the previous proof, we directly use Lemma E.5. To this end, for any $\beta \in \mathbb{N}^S$ such that $\sum_{i=1}^S \beta_i = n$ and $\widetilde{\beta} = \beta \sqcup (1)$, we define:

$$K_\beta = \mathrm{KL}(\widehat{\nu}_{\widetilde{\mathbb{Y}}_{S,\frac{\beta}{n}}} \,\|\, \widetilde{\nu}_{k^\star}) - \mathcal{K}_{\inf}^{m_\beta}(\widetilde{\mathbb{Y}}_{S,\frac{\beta}{n}}; \mu_k)\,, \tag{E.81}$$

where $\widetilde{\nu}_{k^\star}$ denote the discretised and truncated version of $\nu_{k^\star}$ and $m_\beta$ denotes the largest item with a nonzero coefficient in $\widetilde{\beta}$. We recall that QDS summarises the information larger than the empirical $(1-\alpha)$-quantile by their mean (i.e. the $\mathrm{CVaR}_\alpha$ of the empirical distribution). The truncation in $\underline{y}$ does not change that, and will simply makes this quantity smaller which will itself makes the BCP smaller (although not so much with well chosen $\eta, y$). We use the result from Honda and Takemura (2010, proof of Theorem 7) stating that

$$\mathcal{K}_{\inf}^{m_\beta}(\widetilde{\mathbb{Y}}_{S,\frac{\beta}{n}}; \mu_k) \leqslant \frac{\bar{\Delta}_n}{m_\beta - \mu}, \tag{E.82}$$

where $\bar{\Delta}_n = \mu^\star - \widehat{\mu}_{\mathbb{Y}}$. As we know that $m_\beta$ is at least as large as the exploration bonus, we furthermore have, with $\bar{\Delta}_n^+ = 1/n \sum_{i=1}^n (\mu^\star - Y_i)_+$,

$$\mathcal{K}_{\inf}^{m_\beta}(\widetilde{\mathbb{Y}}_{S,\frac{\beta}{n}}; \mu_k) \leqslant \frac{\bar{\Delta}_n}{\rho \bar{\Delta}_n^+} \leqslant 1/\rho. \tag{E.83}$$

Consequently, for any $\xi > 0$, $K_\beta \geqslant \xi$ on all the subspace of empirical distributions satisfying $\mathrm{KL}(\widehat{\nu}_{\widetilde{\mathbb{Y}}_{S,\frac{\beta}{n}}} \parallel \widetilde{\nu}_{k^\star}) \geqslant (1+\xi)/\rho$. We now use Pinsker's inequality to relate the KL divergence and the total variation distance $\delta_{\mathrm{TV}}$, in this subspace, i.e.

$$\delta_{\mathrm{TV}}(\widehat{\nu}_{\widetilde{\mathbb{Y}}_{S,\frac{\beta}{n}}}, \widetilde{\nu}_{k^\star}) \leqslant \sqrt{\frac{1+\xi}{2\rho}}. \tag{E.84}$$

If this quantity is small, we can control the probability of each event of interest. In particular, we want the quantile *used in the algorithm* to be strictly larger than the $(1-\alpha)$-quantile *of the bandit model* (Q) of Theorem 6.17. If the latter is $\alpha$, and we run the algorithm with a parameter $\alpha' < \alpha$, then it is possible to tune $\rho$ such that $\widehat{F}_n^k(q_{1-\alpha}(\nu_k)) < 1-\alpha$. This means that the *true quantile $q_{1-\alpha}(\nu_k)$ is present in the set $\mathbb{Y}_{(n)}^k$ and is not truncated by the algorithm*. In particular, this holds if $\rho \geqslant \frac{1+\alpha'}{\alpha'^2}$, in which case we have

$$\begin{aligned}
\mathrm{KL}(\widehat{\nu}_{\widetilde{\mathbb{Y}}_{S,\frac{\beta}{n}}} \parallel \widetilde{\nu}_{k^\star}) - \mathcal{K}_{\inf}^{m_\beta}(\widetilde{\mathbb{Y}}_{S,\frac{\beta}{n}}, \mu_k) &\geqslant \mathcal{K}_{\inf}^{\mathcal{F}}(\widetilde{\mathbb{Y}}_{S,\frac{\beta}{n}}, \mu^\star - \eta) - \mathcal{K}_{\inf}^{q_{1-\alpha'}}(\widetilde{\mathbb{Y}}_{S,\frac{\beta}{n}}, \mu_k) \\
&\geqslant \mathcal{K}_{\inf}^{q_{1-\alpha'}}(\widetilde{\mathbb{Y}}_{S,\frac{\beta}{n}}, \mu^\star - \eta) - \mathcal{K}_{\inf}^{q_{1-\alpha'}}(\widetilde{\mathbb{Y}}_{S,\frac{\beta}{n}}, \mu_k) \\
&\geqslant \kappa,
\end{aligned} \tag{E.85}$$

for some $\kappa > 0$ and thanks to the definition of the family $\mathcal{F}_{[\underline{B},+\infty]}^{\alpha}$. This concludes the proof as it ensures that Assumption 6.4 is satisfied by the QDS algorithm on $\mathcal{F}_{[\underline{B},+\infty]}^{\alpha}$. ■

### Regret bound for RDS

The bandit model (R) for this third instance of DS algorithms is simply that of general light tailed distributions, i.e. $\nu_k$ is such that $\lambda \mapsto \mathbb{E}_{Y\sim\nu_k}[e^{\lambda Y}]$ is defined in a neighbourhood of zero.

*Proof of Theorem 6.19.* We recall that

$$B^{\mathrm{RDS}}(\mathbb{Y}^k_{(n)}; \rho_n, \mu) = \mu + \frac{\rho_n}{n}\sum_{i=1}^{n}(\mu - Y_i^k)_+ \,, \tag{E.86}$$

for a sequence $(\rho_n)_{n\in\mathbb{N}}$ satisfying $\rho_n \to +\infty$ and $\rho_n = o(n)$. First, once again, Cramér's theorem (Dembo and Zeitouni, 2009, Theorem 2.2.3) provides the existence of good rate function $I_k$ satisfying Assumption 6.2. We now show that this bonus implies Assumption 6.4.

**Upper bounding $n_k(T)$.** We again aim to find a suitable set $\mathcal{B}_n^k$ for rewards $\mathbb{Y}^k_{(n)} = (Y_i^k)_{i=1}^n$ that would allow to use Lemma E.6. In the bandit model (R), we show that this can be achieved by simply controlling the mean, the excess gap used in the exploration bonus and a range on the maximum observed reward. Hence, we fix some $\varepsilon > 0$ and consider

$$\mathcal{B}_n^k = \left\{\mathbb{Y} \in \mathbb{R}^n,\ \widehat{\mu}_{\mathbb{Y}} \leqslant \mu_k + \varepsilon, \widehat{\Delta}(\mathbb{Y};\mu^\star) \leqslant \Delta_k^+(\mu^\star) + \varepsilon, \widehat{\sigma}_{\mathbb{Y}}(\mu^\star) \leqslant \sigma^k(\mu^\star 0) + \varepsilon, \max \mathbb{Y} \in [m_n^k, M_n^k]\right\} , \tag{E.87}$$

where if $\mathbb{Y} = (Y_i)_{i=1}^n$, we defined $\widehat{\Delta}(\mathbb{Y};\mu^\star) = \frac{1}{n}\sum_{i=1}^n(\mu^\star - Y_i)_+$ and $\Delta_k^+(\mu^\star) = \mathbb{E}_{Y\sim\nu_k}[(\mu^\star - Y)_+]$, $\widehat{\sigma}^2_{\mathbb{Y}}(\mu^\star) = \frac{1}{n}\sum_{i=1}^n (Y_i - \mu^\star)^2$ and $\sigma^k(\mu^\star) = \mathbb{E}_{Y\sim\nu_k}[(Y-\mu^\star)^2]$, as well as $(m_n^k)_{n\in\mathbb{N}}$, $(M_n^k)_{n\in\mathbb{N}}$ two fixed sequences to be tuned later.

We start with the two conditions $\{\widehat{\mu}_{\mathbb{Y}} \leqslant \mu_k + \varepsilon\}$ and $\{\widehat{\Delta}(\mathbb{Y};\mu^\star) \leqslant \Delta_k^+(\mu^\star) + \varepsilon\}$. The concentration of the empirical mean derives from Assumption 6.2, and since $(\mu^\star - Y)_+$ is also light tailed if $Y$ is, there also exists a rate function $I_k^+$ satisfying $\mathbb{P}\left(\widehat{\Delta}(\mathbb{Y}^k_{(n)}, \mu^\star) \geqslant \Delta_k^+(\mu^\star) + \varepsilon\right) \leqslant \exp\left(-nI_k^+(\Delta_k^+(\mu^\star)+\varepsilon)\right)$ if $\mathbb{Y}^k_{(n)}$ is a sequence of $n$ i.i.d. random variables drawn from $\nu_k$. Therefore we have

$$\begin{aligned}
&\sum_{n=1}^{+\infty}\left(\mathbb{P}\left(\widehat{\mu}^k_{(n)} \geqslant \mu_k + \varepsilon\right) + \mathbb{P}\left(\widehat{\Delta}(\mathbb{Y}^k_{(n)};\mu^\star) \geqslant \Delta_k^+(\mu^\star) + \varepsilon\right)\right) \\
&\qquad \leqslant \sum_{n=1}^{+\infty}\left(\exp(-nI_k(\mu_k+\varepsilon)) + \exp(-nI_k^+(\Delta_k^+(\mu^\star)+\varepsilon))\right) \\
&\qquad \leqslant \frac{1}{1-e^{-I_k(\mu_k+\varepsilon)}} + \frac{1}{1-e^{-I_k^+(\Delta_k^+(\mu^\star)+\varepsilon)}} \,.
\end{aligned} \tag{E.88}$$

We now turn to the event with the variance terms. To this end, we consider the Wasserstein metric $W_2$ between the empirical distribution of $\mathbb{Y}^k_{(n)}$ and the true distribution $\nu_k$. Following Fournier and Guillin (2015, Theorem 2), there exists constants $c > 0$ and $C > 0$ such that for any $x \in (0,1]$

$$\mathbb{P}\left(W_2(\widehat{\nu}^k_{(n)}, \nu_k) \geqslant \sqrt{x}\right) \leqslant C\left(e^{-cnx^2} + e^{-c(nx)^{\frac{1}{3}}}\right) \,. \tag{E.89}$$

The coefficient $1/3$ comes from choosing $\varepsilon$ as $(1-\varepsilon)/2 = 1/3$ in the statement of Theorem 2 in the above reference (which is different from the $\varepsilon$ in this proof). This inequality is dominated by the second term, thus for any $\varepsilon > 0$, there exists $\varepsilon_1 > 0$ such that if $W_2(\widehat{\nu}^k_{(n)}, \nu_k) \leqslant \varepsilon_1$ then $\widehat{\sigma}_{\mathbb{Y}^k_{(n)}}(\mu^\star) = \widehat{\sigma}^k_{(n)}(\mu^\star) \leqslant \sigma^k(\mu^\star) + \varepsilon$. Furthermore, this shows that the series $\sum_{n\in\mathbb{N}^\star} \mathbb{P}(W_2(\widehat{\nu}^k_{(n)}, \nu_k) \geqslant \varepsilon_1)$ converges, which concludes the part of the proof corresponding to this term.

We now investigate possible values for the sequence $(m^k_n)_{n\in\mathbb{N}}$ and $(M^k_n)_{\mathbb{N}}$ that would allow $\mathcal{B}^k_n$ to happen with high probability. The maximum $\max \mathbb{Y}^k_{(n)}$ of a set of $n$ i.i.d. random variables has an explicit distribution, with c.d.f. given by, for any $x \in \mathbb{R}$,

$$\mathbb{P}_{\mathbb{Y}^k_{(n)}\sim\nu_k^{\otimes n}}(\max \mathbb{Y}^k_{(n)} \leqslant x) = F^k(x)^n \,, \tag{E.90}$$

where $F^k$ is the c.d.f. of $\nu_k$. We first calibrate $M^k_n$ to be a high probability upper bound on the maximum of $n$ rewards for arm $k$, e.g. $\mathbb{P}(\max \mathbb{Y}^k_{(n)} \leqslant M^k_n) \geqslant 1 - \frac{1}{n\log(n)^2}$ for all $n \in \mathbb{N}^\star$, which is achieved by setting:

$$M^k_n := F_k^{-1}\left(\exp\left(-\frac{1}{n^2(\log n)^2}\right)\right) \geqslant F_k^{-1}\left(\left(1-\frac{1}{n(\log n)^2}\right)^{\frac{1}{n}}\right) \,, \tag{E.91}$$

where $F_k^{-1}$ is the (pseuso)inverse of the c.d.f. of $\nu_k$. Note that $\lim_{n\to+\infty}(1-\frac{1}{n(\log n)^2})^{1/n} = 1$ and thus $\lim_{n\to+\infty} M^k_n = +\infty$. This way, the series $\sum_{n\in\mathbb{N}^\star} \mathbb{P}(\max \mathbb{Y}^k_{(n)} > M^k_n) \leqslant \frac{1}{n\log(n)^2}$ converges. On the other hand, we want $m^k_n$ to be a lower bound on the maximum observed reward, e.g. $\mathbb{P}(\max \mathbb{Y}^k_{(n)} \leqslant m^k_n) \leqslant \frac{1}{n\log(n)^2}$, hence we define

$$m^k_n := F_k^{-1}\left(\frac{1}{n(\log n)^2}^{\frac{1}{n}}\right) = F_k^{-1}\left(\exp\left(-\frac{\log n + 2\log\log n}{n}\right)\right) \,. \tag{E.92}$$

Again, $\lim_{n\to+\infty} \frac{1}{n(\log n)^2}^{1/n} = 1$ and thus $\lim_{n\to+\infty} m^k_n = +\infty$. Combining all these results, we obtain

$$\sum_{n=1}^{T-1} \mathbb{P}_{\mathbb{Y}^k_{(n)}\sim\nu_k^{\otimes n}}(\mathbb{Y}^k_{(n)} \notin \mathcal{B}^k_n) = \mathcal{O}(1) \,. \tag{E.93}$$

We now use the first part of Lemma 6.7 and the fact that for any $\eta \in [0,1)$ and $x \in (-\infty, \eta]$, we have $-\log(1-x) \leqslant x + \frac{1}{1-\eta}\frac{x^2}{2}$. Letting $\mathfrak{M}^k_n = \max \mathbb{Y}^k_{(n)} \vee B^{\mathrm{RDS}}(\mathbb{Y}^k_{(n)}, \rho_n, \mu^\star)$ and in addition $Y_{n+1} = B^{\mathrm{RDS}}(\mathbb{Y}^k_{(n)}, \rho_n, \mu^\star)$, we use the representation of Dirichlet samples as normalised exponential variables to obtain, by a similar argument as in the Chernoff method,

$$[BCP]\left(\mathbb{Y}^k_{(n+1)}; \mu^\star\right) = \mathbb{P}_{R_1,\dots,R_{n+1}\sim\mathcal{E}(1)}\left(\sum_{i=1}^{n+1} R_i(Y^k_i - \mu^\star) \geqslant 0 \,\middle|\, \mathbb{Y}^k_{(n)}\right)$$

$$
\begin{aligned}
&\leqslant \inf_{\lambda\in[0,\frac{\eta}{\mathfrak{M}_n^k-\mu^\star})} \prod_{i=1}^{n+1} \mathbb{E}_{R_i\sim\mathcal{E}(1)}\left[e^{\lambda R_i\left(Y_i^k-\mu^\star\right)} \;\middle|\; Y_i^k\right] \\
&\leqslant \exp\left(-\sum_{i=1}^{n+1}\log\left(1-\eta\frac{Y_i^k-\mu^\star}{\mathfrak{M}_n^k-\mu^\star}\right)\right) \\
&\leqslant \frac{1}{1-\eta}\exp\left(-\sum_{i=1}^{n}\log\left(1-\eta\frac{Y_i^k-\mu^\star}{\mathfrak{M}_n^k-\mu^\star}\right)\right) \\
&\leqslant \frac{1}{1-\eta}\exp\left(\sum_{i=1}^{n}\left(\eta\frac{Y_i^k-\mu^\star}{\mathfrak{M}_n^k-\mu^\star}+\frac{\eta^2}{2(1-\eta)}\left(\frac{Y_i^k-\mu^\star}{\mathfrak{M}_n^k-\mu^\star}\right)^2\right)\right) \\
&= \frac{1}{1-\eta}\exp\left(-n\eta\frac{\mu^\star-\widehat{\mu}_{(n)}^k}{\mathfrak{M}_n^k-\mu^\star}+n\frac{\eta^2}{2(1-\eta)}\frac{\widehat{\sigma}^2_{\mathbb{Y}^k_{(n)}}(\mu^\star)^2}{(\mathfrak{M}_n^k-\mu^\star)}\right). \qquad \text{(E.94)}
\end{aligned}
$$

We recall that we consider this upper bound under the event $\mathbb{Y}^k_{(n)} \in \mathcal{B}_n^k$, which ensures that (i) $\max \mathbb{Y}^k_{(n)} \in [m_n^k, M_n^k]$, (ii) $\mu^\star - \widehat{\mu}^k_{(n)} \geqslant \mu^\star - \mu_k + \varepsilon$, (iii) the bonus is upper bounded by $\mu^\star + \rho_n \times (\Delta_k^+ + \varepsilon)$, and (iv) the quadratic deviation satisfies $\widehat{\sigma}^k_{(n)}(\mu^\star) \leqslant \sigma^k(\mu^\star) + \varepsilon$. For any $\varepsilon_0 > 0$, if we further assume that $M_n^k = o((m_n^k)^2)$, for any $n$ large enough these results finally provide

$$
\mathbb{P}\left(\widetilde{\mu}(\mathbb{Y}^k_{(n)}, \mu^\star) \geqslant \mu^\star\right) \leqslant \frac{1}{1-\eta}\exp\left(-n\eta\frac{\Delta_k-\varepsilon_0}{M_n^k \vee B^k_{n,\mu^\star,\varepsilon}-\mu^\star}\right), \qquad \text{(E.95)}
$$

where $B^k_{n,\mu^\star,\varepsilon} = \mu + \rho_n\left(\mathbb{E}_{\nu_k}\left[(\mu - X)_+\right] + \varepsilon\right)$. The condition $M_n^k = o\left((m_n^k)^2\right)$ is satisfied for light tailed distributions, as they generally have at most a polylogarithmic growth of the maximum (e.g. $\log(n)$ for exponential tails and $\sqrt{\log n}$ for Gaussian tails) and so $M_n^k$ and $m_n^k$ are actually of the same order of magnitude. We then recover all the terms of Theorem 6.19 by matching the exponent of the upper bound with $-\log T$.

To conclude this part, the light tailed hypothesis provides an asymptotic upper bound on the expected number of pulls of each suboptimal arm for the RDS index. Then, the choice of $m_k(T)$ in Lemma E.6 can be $m_k(T) = O(\log(T) M^k_{\log(T)})$ if $M_n^k$ is polylogarithmic in $n$ when $n \to +\infty$.

**Remark E.7**. *The concentration bound we use on the Dirichlet weighted average requires the control of the second empirical decentred moment* $\widehat{\sigma}^2_{\mathbb{Y}^k_{(n)}}(\mu^\star)$ *since we use* $-\log(1-x) \leqslant x + \frac{1}{1-\eta}\frac{x^2}{2}$. *This control follows from the existence of exponential moments, i.e.* $\nu_k \in \mathcal{F}_\ell$. *A tighter analysis is possible, indeed for any* $q > 0$ *and* $\eta \in (0,1)$ *there exists* $C_{q,\eta} > 0$ *such that for all* $x \leqslant \eta$, $-\log(1-x) \leqslant x + C_{q,\eta}|x|^{1+q}$ ($C_{1,\eta} = \frac{1}{1-\eta}$), *relaxing the requirement to a mere control of the moment of order* $1+q$ *in the topology of* $W_{1+q}$ *(for which concentration results are similar to the one we provide, for the families we consider). Furthermore, it would relax the condition relating* $m^k_n$ *and* $M^k_n$ *to* $M^k_n = o\left((m^k_n)^{1+q}\right)$.

**Assumption 6.4.** We use the left-hand term of Lemma 6.8 and obtain a lower bound on $[BCP]$ in $e^{-\frac{n}{\rho_n}}$. Combining this result with Assumption 6.2 we obtain

$$E_n \leqslant e^{-n(I_{k^\star}(\mu_k)-1/\rho_n)}\,, \tag{E.96}$$

and for $n$ large enough $\rho_n > 1/I_{k^\star}(\mu_k)$, which is sufficient to obtain the convergence of $\sum_{n=1}^{+\infty} E_n$.
■

## E.4 Examples of distributions fitting the family of QDS

We first show that virtually any lower bounded distribution satisfies the quantile condition of QDS (Theorem 6.17) provided the exploration bonus $\rho$ is large enough.

**Lemma E.8**. *Let* $\alpha \in (0,1)$, $\underline{B} \in \mathbb{R}$ *and* $\mathcal{F} \subset \mathcal{F}_{[\underline{B},+\infty)}$. *For any atomless distribution* $\nu \in \mathcal{F}$ *and* $\mu > \mathbb{E}_{Y\sim\nu}[Y]$, *there exists* $\rho > 0$ *and* $\mathfrak{M}_{\rho,\alpha} > 0$ *such that*

$$\mathcal{K}_{\inf}^{\mathfrak{M}_\rho}\left(\mathcal{T}_\alpha(\nu);\mu\right) \leqslant \mathcal{K}_{\inf}^{\mathcal{F}}(\nu;\mu)\,. \tag{E.97}$$

*Proof of Lemma E.8.* Let $\mathfrak{M}_{\rho,\alpha} = \max\left\{C_\alpha(\nu), \mu + \rho\mathbb{E}_{Y\sim\nu}\left[(\mu - Y)_+\right]\right\}$. By construction, the support of $\mathcal{T}_\alpha(\nu)$ is upper bounded by $\mathfrak{M}_{\rho,\alpha}$ and $\mu < \mathfrak{M}_{\rho,\alpha}$, therefore it follows from Honda and

Takemura (2010, Theorem 8) that

$$\mathcal{K}_{\inf}^{\mathfrak{M}_{\rho,\alpha}}\left(\mathcal{T}_\alpha(\nu),\mu\right) = \max_{\lambda\in[0,\frac{1}{\mathfrak{M}_{\rho,\alpha}-\mu}]} \mathbb{E}_{Y\sim\mathcal{T}_\alpha(\nu)}\left[\log\left(1-\lambda(Y-\mu)\right)\right] . \tag{E.98}$$

Then the concavity of $\log$ shows that, for $\lambda \in [0, \frac{1}{\mathfrak{M}_{\rho,\alpha}-\mu}]$,

$$\begin{aligned}
\mathbb{E}_{Y\sim\mathcal{T}_\alpha(\nu)}\left[\log\left(1-\lambda(Y-\mu)\right)\right] &\leqslant -\mathbb{E}_{Y\sim\mathcal{T}_\alpha(\nu)}\left[\lambda(Y-\mu)\right] \\
&= \lambda\left(\mu - \mathbb{E}_{Y\sim\mathcal{T}_\alpha(\nu)}\left[Y\right]\right) \\
&= \lambda\left(\mu - \mathbb{E}_{Y\sim\nu}\left[Y\mathbb{1}_{Y\leqslant q_{1-\alpha}(\nu)}\right] - \mathbb{E}_{Y\sim\mathcal{T}_\alpha(\nu)}\left[Y\mathbb{1}_{Y>q_{1-\alpha}(\nu)}\right]\right) \\
&= \lambda\left(\mu - \mathbb{E}_{Y\sim\nu}\left[Y\mathbb{1}_{Y\leqslant q_{1-\alpha}(\nu)}\right] - \alpha C_\alpha(\nu)\right) \\
&= \lambda\left(\mu - \mathbb{E}_{Y\sim\nu}\left[Y\mathbb{1}_{Y\leqslant q_{1-\alpha}(\nu)}\right] - \alpha\mathbb{E}_{Y\sim\nu}\left[Y \mid Y > q_{1-\alpha}(\nu)\right]\right) \\
&= \lambda\left(\mu - \mathbb{E}_{Y\sim\nu}\left[Y\mathbb{1}_{Y\leqslant q_{1-\alpha}(\nu)}\right] - \mathbb{E}_{Y\sim\nu}\left[Y\mathbb{1}_{Y>q_{1-\alpha}(\nu)}\right]\right) \\
&= \lambda\left(\mu - \mathbb{E}_{Y\sim\nu}\left[Y\right]\right) ,
\end{aligned} \tag{E.99}$$

by definition of $C_\alpha(\nu)$ and the conditional expectation, as well as the absence of atom at $q_{1-\alpha}(\nu)$. Since $\mu > \mathbb{E}_{Y\sim\nu}[Y]$, the maximum of the right-hand side in $\lambda$ is attained at the rightmost point, which yields

$$\mathcal{K}_{\inf}^{\mathfrak{M}_{\rho,\alpha}}\left(\mathcal{T}_\alpha(\nu);\mu\right) \leqslant \frac{\mu - \mathbb{E}_{Y\sim\nu}[Y]}{\mathfrak{M}_{\rho,\alpha}-\mu} . \tag{E.100}$$

For $\rho > 0$ large enough, we have $\mathfrak{M}_{\rho,\alpha} = \mu + \rho\mathbb{E}_{Y\sim\nu}\left[(\mu - Y)_+\right]$ which further simplifies as

$$\mathcal{K}_{\inf}^{\mathfrak{M}_{\rho,\alpha}}\left(\mathcal{T}_\alpha(\nu);\mu\right) \leqslant \frac{\mu - \mathbb{E}_{Y\sim\nu}[Y]}{\rho\mathbb{E}_{Y\sim\nu}\left[(\mu-Y)_+\right]} . \tag{E.101}$$

Therefore for $\rho > 0$ large enough, in particular $\rho \geqslant \frac{\mu - \mathbb{E}_{Y\sim\nu}[Y]}{\mathbb{E}_{Y\sim\nu}\left[(\mu-Y)_+\right]\mathcal{K}_{\inf}^{\mathcal{F}}(\nu;\mu)}$, we have

$$\mathcal{K}_{\inf}^{\mathfrak{M}_{\rho,\alpha}}\left(\mathcal{T}_\alpha(\nu);\mu\right) \leqslant \mathcal{K}_{\inf}^{\mathcal{F}}\left(\nu;\mu\right) . \tag{E.102}$$

■

The bound on $\rho$ given in the above lemma can be rather loose because of the crude concave inequality we use. It also comes at the price of increasing $\rho$, which may hurt the performances of QDS due to overexploration. We now show that this quantile condition can be calculated almost in closed-form and is naturally satisfied by some classical families of distributions.

**Exponential.** Let $\mathcal{F} = (\nu_\theta)_{\theta\in\mathbb{R}_+^*}$ with density $p_\theta(x) = \theta e^{-\theta x}\mathbb{1}_{x\geqslant 0}$. We summarise in the below lemma a number of explicit formulas for the $\mathcal{K}_{\inf}$ operators and quantiles of $\nu_\theta$.

**Lemma E.9** (Some statistics of exponential distributions). *Let $0 < \phi < \theta$ and $\alpha \in (0,1)$.*

*(i)* $\mathbb{E}_\theta[X] = \frac{1}{\theta}$.

*(ii)* $q_{1-\alpha}(\nu_\theta) = -\frac{\log\alpha}{\theta}$.

*(iii)* $\mathbb{E}_\theta\left[Y \mid Y \geqslant q_{1-\alpha}(\nu_\theta)\right] = \frac{1}{\theta} + q_{1-\alpha}(\nu_\theta)$.

*(iv)* $\mathbb{E}_\theta\left[\left(\frac{1}{\phi} - Y\right)_+\right] = \frac{1}{\phi} - \frac{1}{\theta}\left(1 - e^{-\theta/\phi}\right)$.

*(v)* $\mathcal{K}_{\inf}^{\mathcal{F}}(\nu_\theta; \frac{1}{\phi}) = \frac{\phi}{\theta} - \log\frac{\phi}{\theta} - 1$.

*Proof.* $(i)$-$(iv)$ result from straightforward integral calculations. $(v)$ is a direct consequence of $\mathcal{F}$ being a SPEF, which implies $\mathcal{K}_{\inf}^{\mathcal{F}}(\nu_\theta; \frac{1}{\phi}) = \mathrm{KL}(\nu_\theta \parallel \nu_\phi)$, which has the stated closed-form for exponential distributions. ∎

Using these formulas, we numerically compute $\mathcal{K}_{\inf}^{\mathfrak{M}_\rho}$ as a function of $\rho$ by solving the dual optimisation problem (Honda and Takemura, 2010) and compare it to $\mathcal{K}_{\inf}^{\mathcal{F}}$. Conversely, for a fixed exploration bonus $\rho$, we compute the $\mathcal{K}_{\inf}$ of the truncated distribution $\mathcal{T}_\alpha(\nu_\theta)$ for a range of $\alpha$. As per intuition, smaller values of $\alpha$, corresponding to smaller truncations of the support, help satisfy the quantile condition. Results are reported in Figure E.1.

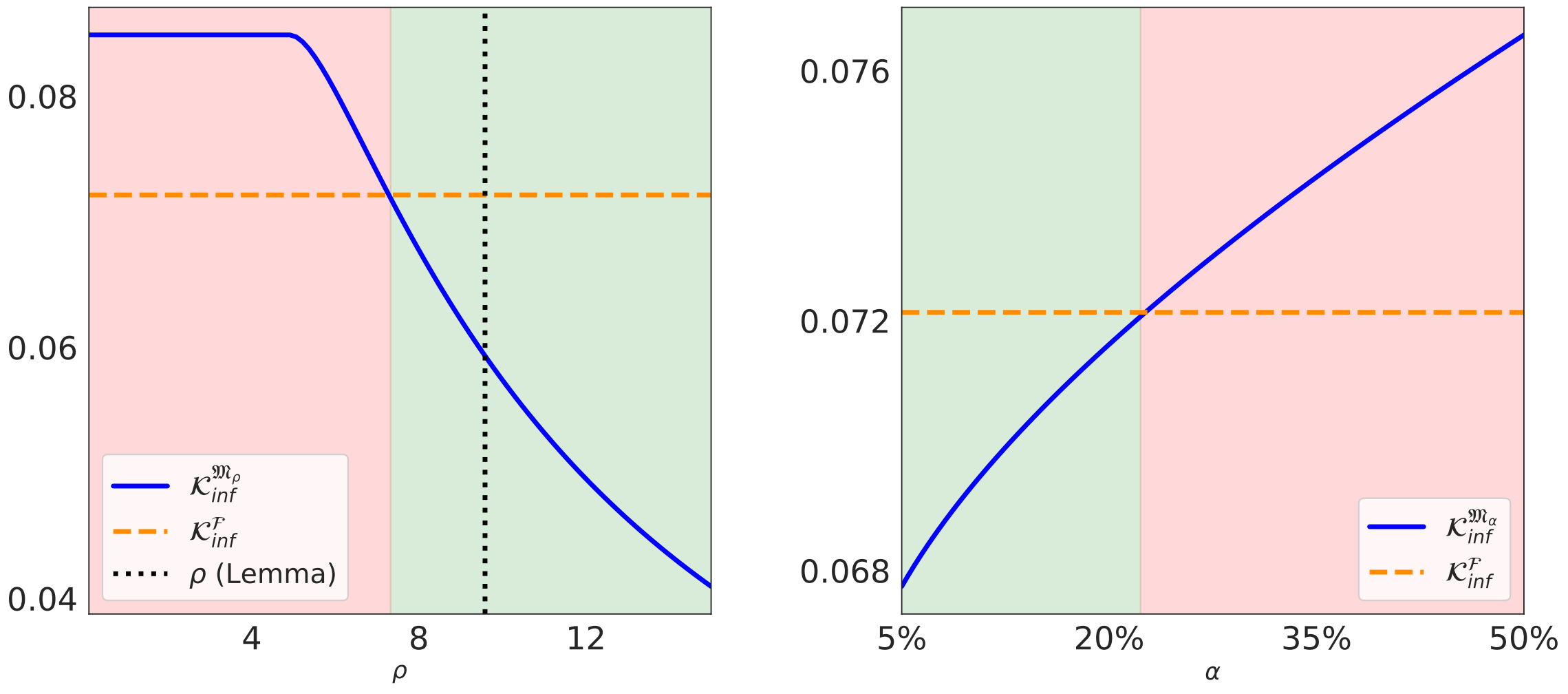

(a) $\alpha = 5\%$.

(b) $\rho = 8$.

**Figure E.1 –** Comparison of $\mathcal{K}_{\inf}^{\mathfrak{M}_{\rho,\alpha}}(\mathcal{T}_\alpha(\nu_\theta); \frac{1}{\phi})$ and $\mathcal{K}_{\inf}^{\mathcal{F}}(\nu_\theta; \frac{1}{\phi})$ for the exponential distribution $\mathcal{E}(\theta)$ with $\theta = \frac{1}{2}$, $\phi = \frac{1}{3}$. Admissible regions for $\rho$ and $\alpha$ are shaded in green, the rest is shaded in red. The dotted line corresponds to the minimum value of $\rho$ recommended by Lemma E.8.

**Gaussian.** Let $\sigma > 0$ and $\mathcal{F}_\sigma = (\nu_\theta)_{\theta \in \mathbb{R}}$ with density $p_\theta(x) = \frac{1}{\sqrt{2\pi}\sigma} e^{-\frac{(x-\theta)^2}{2\sigma^2}}$. We recall some useful statistics of the SPEF of fixed variance Gaussian distributions.

**Lemma E.10** (Some statistics of fixed variance Gaussian distributions). *Let $\theta < \phi$ and $\alpha \in (0,1)$. We denote by $\Phi \colon x \in \mathbb{R} \mapsto \frac{1}{\sqrt{2\pi}} \int_{-\infty}^{x} e^{-\frac{y^2}{2}} dy$ the standard Gaussian cdf.*

*(i)* $\mathbb{E}_\theta[Y] = \theta$.

*(ii)* $q_{1-\alpha}(\nu_\theta) = \theta + \sigma \Phi^{-1}(1-\alpha)$.

*(iii)* $\mathbb{E}_\theta\left[Y \mid Y \geqslant q_{1-\alpha}(\nu_\theta)\right] = \theta + \frac{\sigma}{\alpha\sqrt{2\pi}} e^{-\frac{\Phi^{-1}(1-\alpha)^2}{2}}$.

*(iv)* $\mathbb{E}_\theta\left[(\phi - Y)_+\right] = (\phi - \theta)\,\Phi\left(\frac{\phi-\theta}{\sigma}\right) + \frac{\sigma}{\sqrt{2\pi}} e^{-\frac{(\phi-\theta)^2}{2\sigma^2}}$.

*(v)* $\mathcal{K}_{\inf}^{\mathcal{F}_\sigma}(\nu_\theta; \phi) = \frac{(\phi-\theta)^2}{2\sigma^2}$.

The proof is similar to that of the previous lemma; in particular $(v)$ uses the fact that $\mathcal{F}_\sigma$ forms a SPEF. Results are reported in Figure E.2. The lighter right tail of Gaussian distributions, compared to that of exponential distributions, results in much less stringent conditions on $\alpha$ and $\rho$; in other words, Gaussian distributions are "easier" to summarise with the truncation and conditional value at risk operator than the exponential distributions.

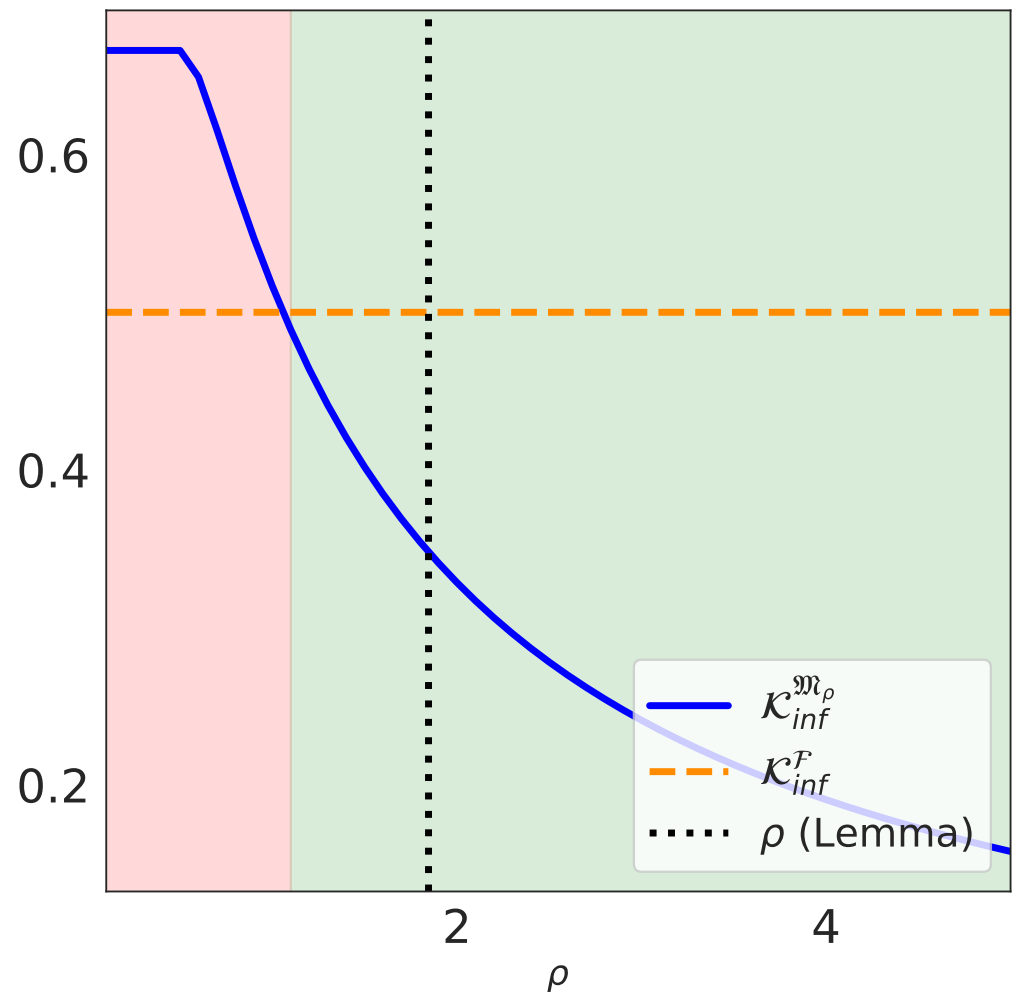

**(a)** $\alpha = 20\%$.

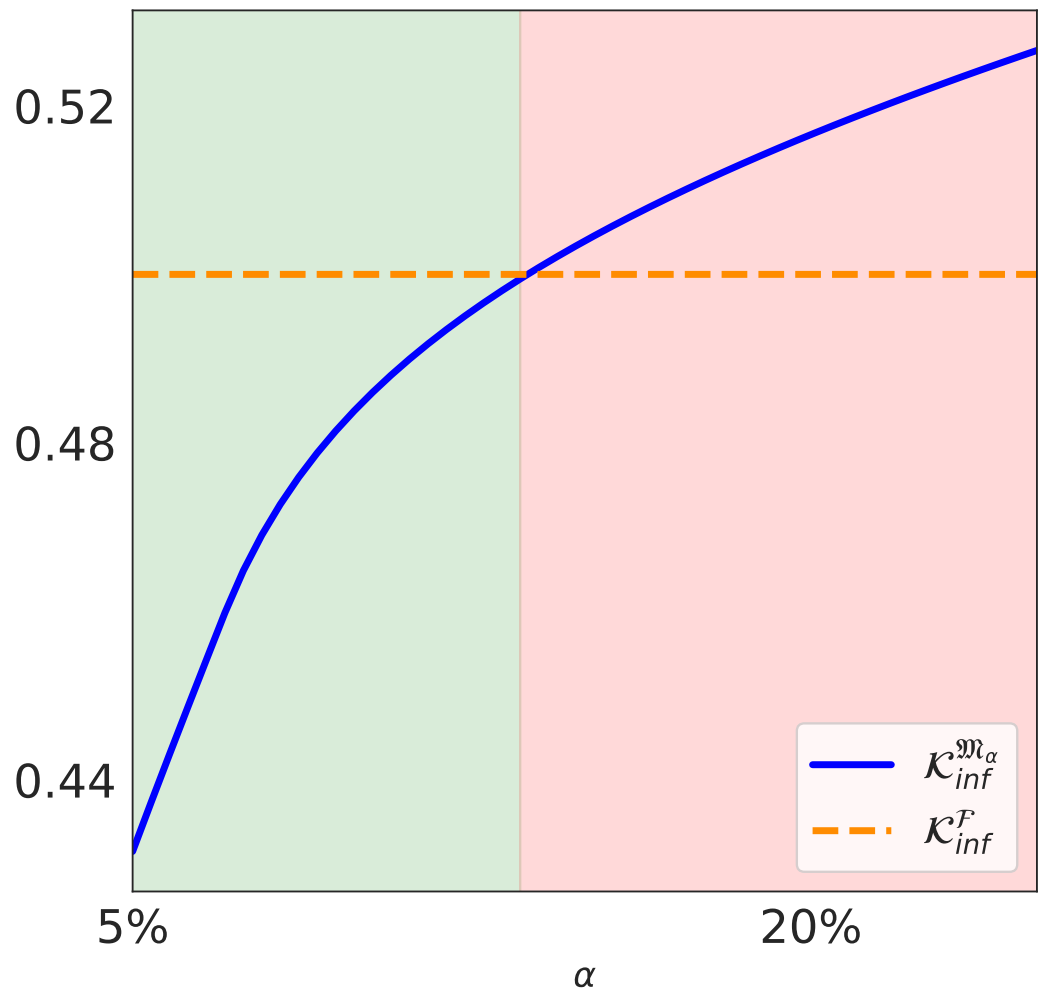

**(b)** $\rho = 1$.

**Figure E.2 –** Comparison of $\mathcal{K}_{\text{inf}}^{\mathfrak{M}_{\rho,\alpha}}(\mathcal{T}_\alpha(\nu_\theta); \phi)$ and $\mathcal{K}_{\text{inf}}^{\mathcal{F}}(\nu_\theta; \phi)$ for the Gaussian distribution $\mathcal{N}(\theta, \sigma^2)$ with $\theta = 0$, $\sigma = 1$, $\phi = 1$. Admissible regions for $\rho$ and $\alpha$ are shaded in green, the rest is shaded in red. The dotted line corresponds to the minimum value of $\rho$ recommended by Lemma E.8.

# Appendix F

# Perspectives on Dirichlet sampling for contextual bandits

## Contents

## F.1 Comparison of deterministic and weighted bootstrap estimators

We present here a simple experiment to show the empirical effect of randomised estimators. We generated samples $(Y_t^m)_{t=1}^T$ for $T, M \in \mathbb{N}$ and $m \in \{1, \ldots, M\}$, drawn from the standard Gaussian distribution $\mathcal{N}(0, 1)$, and then calculated three estimators of the mean:

(i) deterministic estimators: $\widehat{\mu}_t^m = \frac{1}{t}\sum\limits_{s=1}^{t} Y_s^m$;

(ii) randomised estimators: $\widetilde{\mu}_t^m = \sum\limits_{s=1}^{t} W_{s,t}^m Y_s^m$, where $W_{s,t}^m = \frac{Z_s^{m,t}}{\sum\limits_{s=1}^{t} Z_s^{m,t}}$ and $(Z_s^{m,t})_{s=1}^t \sim \mathcal{E}(1)$.

(iii) randomised estimators with exploration: $\widetilde{\mu}_t^{E,m} = \sum\limits_{s=1}^{t} W'^m_{s,t+1} Y_s^m + W'^m_{t+1,t+1}\widetilde{Y}_t$, where $\widetilde{Y}_t = \sqrt{\log(1+t)}$, $W'^m_{s,t} = \frac{Z'^{m,t}_s}{\sum\limits_{s=1}^{t} Z'^{m,t}_s}$ and $(Z'^{m,t}_s)_{s=1}^t \sim \mathcal{E}(1)$.

In particular, this last estimators correspond to the Dirichlet sampling scheme (Chapter 6), and also the weighted bootstrap regression (Chapter 7). Note that the first two estimators are unbiased, i.e. $\mathbb{E}[\widehat{\mu}_t] = \mathbb{E}[\widetilde{\mu}_t] = \mathbb{E}_{Y\sim\mathcal{N}(0,1)}[Y_1] = 0$, while the last exhibits a vanishing bias $\mathbb{E}[\widetilde{\mu}^E] = \sqrt{\log(1+t)}/(t+1)$. However, their variances are different, i.e.

$$\mathbb{V}[\widehat{\mu}_t] = \frac{1}{t}\,, \tag{F.1}$$

$$\mathbb{V}[\widetilde{\mu}_t] = \sum_{s=1}^{t} \mathbb{E}[W_{s,t}^2]\mathbb{E}[Y_s] = \sum_{s=1}^{t} \mathbb{E}[W_{s,t}^2] = \sum_{s=1}^{t} \frac{2}{t(t+1)} = \frac{2}{t+1} > \mathbb{V}[\widehat{\mu}_t]\,, \tag{F.2}$$

$$\mathbb{V}[\widetilde{\mu}_t^E] = \sum_{s=1}^{t} \mathbb{E}[W_{s,t+1}^2] + \widetilde{Y}_t \mathbb{E}[W_{t+1,t+1}^2] = \frac{2t+\widetilde{Y}_t}{(t+1)(t+2)}\,. \tag{F.3}$$

In particular, both randomised estimators have about twice the variance of their deterministic counterpart. In the context of sequential decision-making, this translates into more exploration when using $\widetilde{\mu}_t$ and $\widetilde{\mu}_t^E$. Furthermore, the correlation between successive estimators is given by

$$\rho(\widehat{\mu}_{t-1}, \widehat{\mu}_t) = \frac{1/t}{\sqrt{\mathbb{V}[\widehat{\mu}_{t-1}]\mathbb{V}[\widehat{\mu}_t]}} = \sqrt{1-\frac{1}{t}}\,, \tag{F.4}$$

$$\rho(\widetilde{\mu}_{t-1}, \widetilde{\mu}_t) = \frac{1/t}{\sqrt{\mathbb{V}[\widetilde{\mu}_{t-1}]\mathbb{V}[\widetilde{\mu}_t]}} = \frac{1}{2}\sqrt{1+\frac{1}{t}}\,, \tag{F.5}$$

$$\rho(\widetilde{\mu}_{t-1}^E, \widetilde{\mu}_t^E) = \frac{(t-1)/(t(t+1))}{\sqrt{\mathbb{V}[\widetilde{\mu}_{t-1}^E]\mathbb{V}[\widetilde{\mu}_t^E]}} = \frac{(t-1)\sqrt{1+2/t}}{\sqrt{(2(t-1)+\widetilde{Y}_{t-1})(2t+\widetilde{Y}_t)}} \approx \frac{1}{2}\sqrt{(1-\frac{1}{t})(1+\frac{2}{t})}\,. \tag{F.6}$$

We observe that the deterministic estimators $\widehat{\mu}_t$ are becoming increasingly correlated when $t \to +\infty$, which is at the heart of time-uniform concentration: the deviation of $\widehat{\mu}_{t-1}$ around zero provides a lot of information on the deviation of the next estimator $\widehat{\mu}_t$. This is in stark contrast with the randomised estimators, for which the correlation converges to $1/2$: because of the independent randomisation, $\widetilde{\mu}_{t-1}$ and $\widetilde{\mu}_{t-1}^E$ share little knowledge with $\widetilde{\mu}_t$ and $\widetilde{\mu}_t^E$ respectively.

In Figure F.1, we report trajectories of empirical quantiles (over the $M$ independent replicates) for $t \mapsto (\widehat{\mu}_t^m)_{m=1}^M$, $t \mapsto (\widetilde{\mu}_t^m)_{m=1}^M$ and $t \mapsto (\widetilde{\mu}_t^{E,m})_{m=1}^M$. We observe the variance effect of randomisation, resulting in enlarged quantiles; in addition, the positive exploration term $\widetilde{Y}_t$ contributes to shifting the estimators and their quantiles upwards, especially for small $t$. We also report empirical correlations between consecutive estimators, which illustrates the decorrelation effect of the independent randomisation.

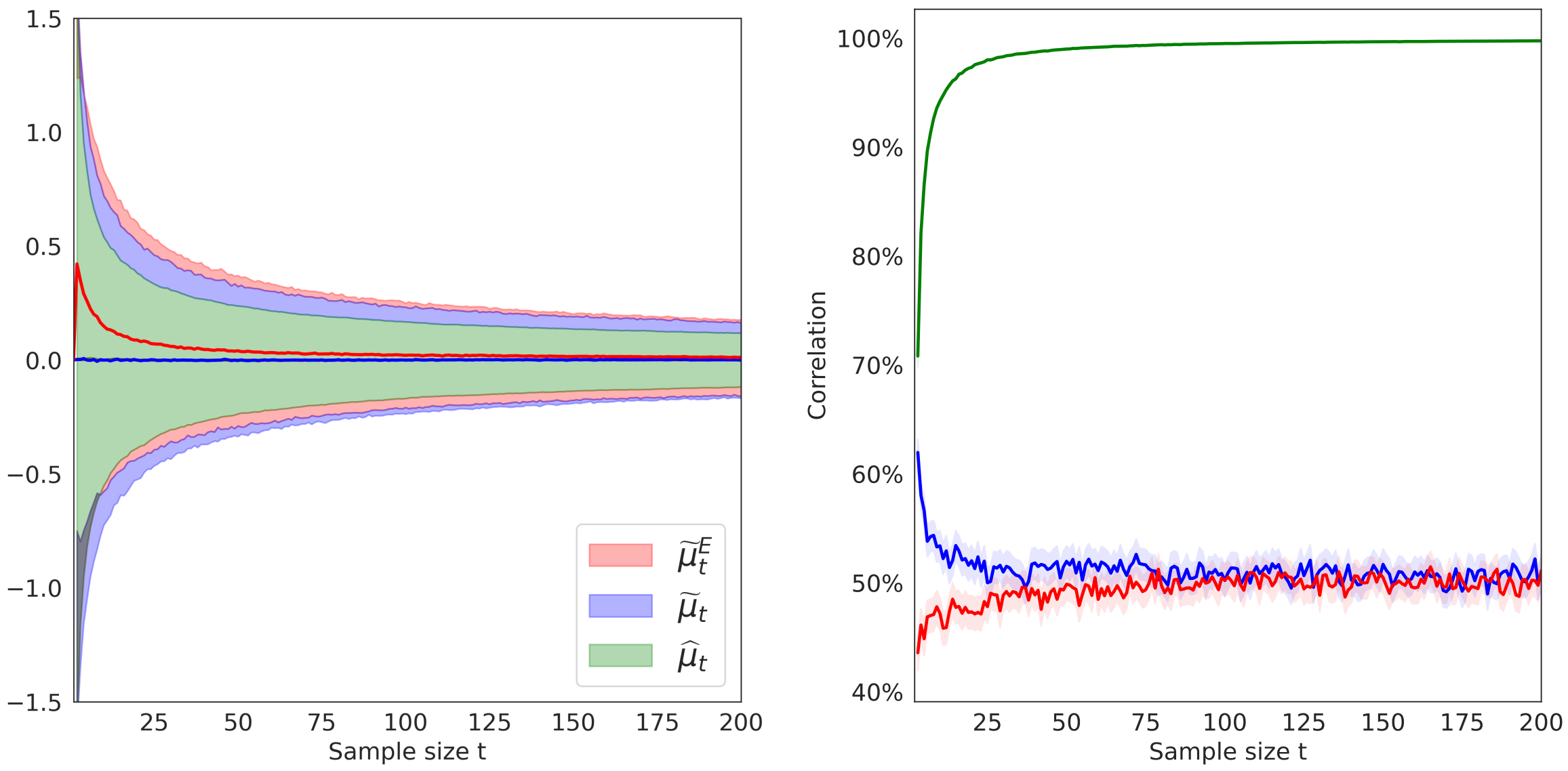

**Figure F.1** – Green: deterministic estimator $\widehat{\mu}_t$. Blue: randomised estimator $\widetilde{\mu}_t$. Red: randomied estimator with exploration $\widetilde{\mu}_t^E$. Results are averaged over $10^4$ independent replicates. Left: trajectories of $\widehat{\mu}_t$, $\widetilde{\mu}_t$ and $\widetilde{\mu}_t^E$. Thick lines are the means of estimators across replicates, shaded areas are the 5th and 95th percentiles. Right: autocorrelation of $\widehat{\mu}_t$, $\widetilde{\mu}_t$ and $\widetilde{\mu}_t^E$.

# Appendix G

# Development and validation of an easy-to-use and interpretable machine learning based calculator predicting five year-weight trajectories after bariatric surgery: a multiple international cohort SOPHIA study

## Contents

## G.1 Preprocessing and visit date adjustment

We first performed a preprocessing of all patients' features: those containing more than 50% of missing values were excluded, and so were categorical features with high-class imbalance (i.e.

one class accounting for less than 2% of all patients) and features consisting of free text input. Categorical variables were converted into binary variables.

Participants with a previous history of bariatric surgery were removed (n=286). Moreover, patients with large delays between scheduled and actual visits related to postoperative complications were also removed (n=59). We defined significant delays as more than 10 days early or 1 month late for M1; 25 days early or 2 months late for M3; 2 months early or 6 months late for M12; 3 months early or 6 months late for M24; 6 months early or 24 months late for M60. In case of missing follow-up visits, patients were kept in the analysis but censored at the corresponding dates. Patients not expected at a given time (recent interventions) were also censored after the last completed visit.

Regarding the visit dates, we performed a time adjustment to improve data quality and downstream prediction accuracy. Indeed, weight predictions are made at conventional dates corresponding to fixed numbers of months after surgery. However, the actual visit dates in the training cohort (when weights are measured) differ from the conventional dates and vary across patients due to early or late follow-up visits. This introduces a bias between the prediction targets and the training data, a priori unrelated to the baseline characteristics. For instance, if a patient had an M3 visit on schedule (i.e. exactly 3 months after surgery) and an M12 visit delayed by a month (i.e. 13 months after surgery rather than 12 months), we would expect the true weight 12 months after surgery to have been between the reported M3 and M12 weights, and larger than the latter since the patient was likely on a downward weight trajectory.

In order to motivate this adjustment, we introduce the following elementary result on conditioned Gaussian processes, which extend the classical Brownian bridge.

**Lemma G.1** (Conditional expectation of real Gaussian processes). *Let $(B_t)_{t\in\mathbb{R}_+}$ a Brownian motion on a filtered probability space, $t_0, T \in \mathbb{R}_+$ such that $T > t_0$ and two mappings $f\colon \mathbb{R}_+ \to \mathbb{R}$ and $\sigma\colon \mathbb{R}_+ \to \mathbb{R}_+^\star$ such that there exists a strong solution $(W_t)_{t\in[t_0,T]}$ to the following SDE with initial (random) condition $W_{t_0}$ (with respect to the augment filtration):*

$$dW_t = f(t)dt + \sigma(t)dB_t\,. \tag{G.1}$$

*Then for any $t \in [t_0, T]$, letting $F_{t_0}(t) = \int_{t_0}^t f(s)ds$ and $\Sigma_{t_0}^2(t) = \int_{t_0}^t \sigma^2(s)ds$, we have*

$$\mathbb{E}\left[W_t \mid W_{t_0}, W_T\right] = W_{t_0} + F_{t_0}(t) + \frac{\Sigma_{t_0}^2(t)}{\Sigma_{t_0}^2(T)}\left(W_T - W_{t_0} - F_{t_0}(T)\right)\,. \tag{G.2}$$

*Proof of Lemma G.1.* First note that $(W_t)_{t\in[t_0,T]}$ is a Gaussian process, and hence for $t \in [t_0, T]$, $W_t$ is a Gaussian random variable $\mathcal{N}(F_{t_0,T}(t), \Sigma^2_{t_0,T}(t))$ (the variance calculation follows from Itô isometry). We write $W_t = \alpha W_{t_0} + \beta W_T + U$ and tune $\alpha, \beta \in \mathbb{R}$ such that the random variable $U$ is independent on both $W_{t_0}$ and $W_T$. A simple covariance calculation shows that

$$\mathbb{V}\left[U, W_{t_0}\right] = \mathbb{V}\left[W_t - \alpha W_{t_0} - \beta W_T, W_{t_0}\right] = (1-\alpha-\beta)\int_0^{t_0} \sigma^2(s)ds\,, \tag{G.3}$$

$$\mathbb{V}\left[U, W_T\right] = \mathbb{V}\left[W_t - \alpha W_{t_0} - \beta W_T, W_T\right] = \int_0^t \sigma^2(s)ds - \alpha\int_0^{t_0} \sigma^2(s)ds - \beta\int_0^T \sigma^2(s)ds\,. \tag{G.4}$$

Since $(W_t)_{t\in[t_0,T]}$ is a Gaussian process, $(U, W_{t_0}, W_T)$ is a Gaussian vector and thus setting the covariances above to zero is equivalent to the independence condition, which suggests to set $\beta = \Sigma^2_{t_0}(t)/\Sigma^2_{t_0}(T)$ and $\alpha = 1 - \beta$. Therefore, we have

$$\begin{aligned}
\mathbb{E}\left[W_t \mid W_{t_0}, W_T\right] &= \mathbb{E}\left[\alpha W_{t_0} + \beta W_T + U \mid W_{t_0}, W_T\right] \\
&= \alpha W_{t_0} + \beta W_T + \mathbb{E}\left[U\right] \\
&= \alpha W_{t_0} + \beta W_T + \mathbb{E}\left[W_t - \alpha W_{t_0} - \beta W_T\right] \\
&= \alpha W_{t_0} + \beta W_T + \int_0^t f(s)ds - \alpha\int_0^{t_0} f(s)ds - \beta\int_0^T f(s)ds \\
&= \alpha W_{t_0} + \beta W_T + F_{t_0}(t) - \beta F_{t_0}(T) \\
&= W_{t_0} + F_{t_0}(t) + \beta\left(W_T - W_{t_0} - F_{t_0}(T)\right)\,.
\end{aligned} \tag{G.5}$$

■

Let us first interpret this result. The SDE $dW_t = \frac{\mathrm{d}F}{\mathrm{d}t}(t)dt + \sigma(t)dB_t$ models random cumulative fluctuations around the backbone curve $t \mapsto F(t)$ (by cumulative, we mean that variance, here $\int_{t_0}^t \sigma^2(s)ds$ increases with time $t$). The conditional expectation $\mathbb{E}\left[W_t \mid W_{t_0}, W_T\right]$ is the optimal predictor of $W_t$ (in the $\mathbb{L}^2$ sense) given the observations $W_{t_0}$ and $W_T$, and hence the formula above reads:

$$\underbrace{\mathbb{E}\left[W_t \mid W_{t_0}, W_T\right]}_{\text{predictor of } W_t} = \underbrace{W_{t_0} + F_{t_0}(t)}_{\substack{\text{deterministic prediction at } t \\ \text{along the backbone from } t_0}} + \underbrace{\frac{\Sigma^2_{t_0}(t)}{\Sigma^2_{t_0}(T)}}_{\text{variance ratio}} \underbrace{\left(W_T - W_{t_0} - F_{t_0}(T)\right)}_{\substack{\text{error between the observation at } T \\ \text{and the deterministic backbone from } t_0}}\,. \tag{G.6}$$

Without additional prior assumption, we may simply model the random fluctuations as a Brownian motion, i.e. $\sigma$ is a constant mapping, in which case the variance ratio is simply $(t - t_0)/(T - t_0)$ (interestingly, it does not depend on the value of $\sigma$ itself).

In the use case of weight prediction, $t_0$ and $T$ correspond to two consecutive actual follow-up visit dates by a study participant, and the process $(W_t)_{t\in[t_0,T]}$ to their weight curve between these dates. If $F$ corresponds to a typical postoperative weight curve, we may use it to interpolate between these actual dates and predict the weight $W_t$ at the conventional visit date $t$. In practice, we used a simple linear backbone $F\colon t \in [t_0, T] \mapsto \gamma t$ for some $\gamma \in \mathbb{R}$, such that

$$\mathbb{E}\left[W_t \mid W_{t_0}, W_T\right] = W_{t_0} + \frac{t - t_0}{T - t_0}\left(W_T - W_{t_0}\right) , \tag{G.7}$$

which corresponds to the linear interpolation between $(t_0, W_{t_0})$ and $(T, W_T)$ (again, independently of the value of $\gamma$ itself). We chose this simple linear backbone as it was sufficient to increase downstream accuracy (compared to training on nonadjusted weights), and more involved backbones did not significantly improve it further (at the cost of increased model complexity). Moreover, we believe this local linear assumption is justified as we excluded participants with large delays between actual and conventional visits (therefore $t$ is close to either $t_0$ or $T$ for participants in the training set, which justifies replacing the backbone $F$ by its first order expansion around either $t_0$ or $T$).

Circling back to the example above, a patient with actual visits at $t_0 = 3$ months and $T = 13$ months would have an adjusted weight at $t = 12$ month (M12) of $9/10 \times W_{13} + 1/10 \times W_3$.

## G.2 Smoothing of weight trajectories

In the online tool shown in Figures 8.5 and 8.6, predicted trajectories of BMI and TWL are represented by smooth, continuous curves. However, the predicted outcomes of the tree-based model correspond to a set of postoperative dates, namely months M1, M3, M12, M24 and M60 after surgery. In order to smoothly generalise to arbitrary dates, we use the following nonlinear parametric model, inspired by the Morse potential in molecular physics.

**Definition G.2** (Morse-like model). *Let* $\theta^\star = (\alpha, \beta, \gamma, \Delta, t^\star, \tau) \in \mathbb{R}^4 \times (\mathbb{R}_+^\star)^2$. *We define the Morse-like model as the mapping*

$$\begin{aligned} F_{\theta^\star} \colon \mathbb{R}_+ &\longrightarrow \mathbb{R} \\ t &\longmapsto \frac{\alpha}{(1+t)^2} + \beta\left(\gamma - e^{-\frac{t-t^\star}{\tau}}\right)^2 + \Delta\,. \end{aligned} \tag{G.8}$$

The original Morse potential was defined as $t \in \mathbb{R}_+ \mapsto \beta(1 - e^{-\frac{t-t^\star}{\tau}})^2$ and exhibits a steep decrease for $t < t^\star$, followed by an increase and an eventual flattening for $t > t^\star$ (in particular $t^\star$ parametrises the nadir in this model), which is precisely the expected shape of weight loss curves. The additional parameters introduced in this definition allow to precisely fit the weights, BMI or TWL points at M0, M1, M3, M12, M24 and M60 (note that there are 6 points and 6 parameters). In particular, $\Delta$ induces a parallel shift of the curve to match the initial point at M0, and $\alpha$ makes for a steeper decrease in the first few months, which are typically associated with a fast weight loss. For $\alpha = 0$, the nadir (minimum weight) is attained at $t_{\text{nadir}} = t^\star + \tau \log 1/\gamma$; for $\alpha \neq 0$, the exact expression of the nadir is not explicit but is approximately the same since the contribution of the term $\alpha/(1+t)^2$ affects mostly the front end of the curve.

We used the Morse-like model for the two main operations, namely RYGB and SG. For AGB, which does not exhibit such a pronounced weight regain (bands are typically tightened or converted into a different operation when too much weight regain is observed), we used instead the linear regression model defined for $\theta^\star = (\alpha, \beta, \gamma, \Delta) \in \mathbb{R}^4$ as

$$\begin{aligned} F_{\theta^\star} : \mathbb{R}_+ &\longrightarrow \mathbb{R} \\ t &\longmapsto \frac{\alpha}{(1+t)^2} + \beta \log\left(1 + t^2\right) + \gamma t + \Delta \,. \end{aligned} \tag{G.9}$$

**Calibration.** This parametric curve model is nonlinear, and we first calibrated it on the weights by solving the following (nonconvex) nonlinear least squares regression problem

$$\theta^\star \in \underset{\theta \in \mathbb{R}^4 \times (\mathbb{R}_+^\star)^2}{\operatorname{argmin}} \sum_{t \in \{0,1,3,12,24,60\}} (W_t - F_\theta(t))^2 + \lambda \|\theta\|_2^2 \,, \tag{G.10}$$

where $\lambda \in \mathbb{R}_+^\star$ is a regularisation parameter. We used the Levenberg-Marquardt algorithm (Press, 2007) initialised at an arbitrary vector $\theta_0$.[1] Of note, this model is close to being linear; indeed, if we fix $\tau \in \mathbb{R}_+^\star$, we remark that $F_\theta(t) = \widetilde{\theta}^\top X_t$ where $\widetilde{\theta} = (\beta\gamma^2 + \Delta, \alpha, -2\beta\gamma e^{t^\star/\tau}, \beta e^{2t^\star/\tau})$ and $X_t = (1, 1/(1+t)^2, e^{-t/\tau}, e^{-2t/\tau})$. Being a nonconvex problem, the output of the minimisation algorithm may be sensitive to the initialisation $\theta_0$. In practice, we manually tuned a reasonable $\theta_0$ and monitored the resulting fitted curves, which we found to be stable.

**Alternative models.** We also considered cubic splines, in particular natural splines, to interpolate between predicted weights. We rejected this approach for two reasons. First, the main argument for splines is that they are able to perfectly fit the target data (interpolation), however weights are typically measured and predicted with uncertainty, so we do not want to fit the main trend but not the noise (regression). Second, all tested types of splines produced undesired curvatures between visit dates when interpolating between predicted weights for at least one combination of the 7 input variables (intervention, preoperative weight, height, age, smoking status, diabetes status and duration). In contrast, the Morse-like model of Definition G.2 allows to precisely control the shape of the resulting curves (note that we could also add constraints on the parameters of the Morse-like model, e.g. $\beta > 0$ to enforce the shape constraint, although we did not need this in practice). We report in Figure G.1 both methods for the smoothing of the predicted weights of sleeve gastrectomy patient.

[1] This is implemented in the scipy.optimize library in Python.

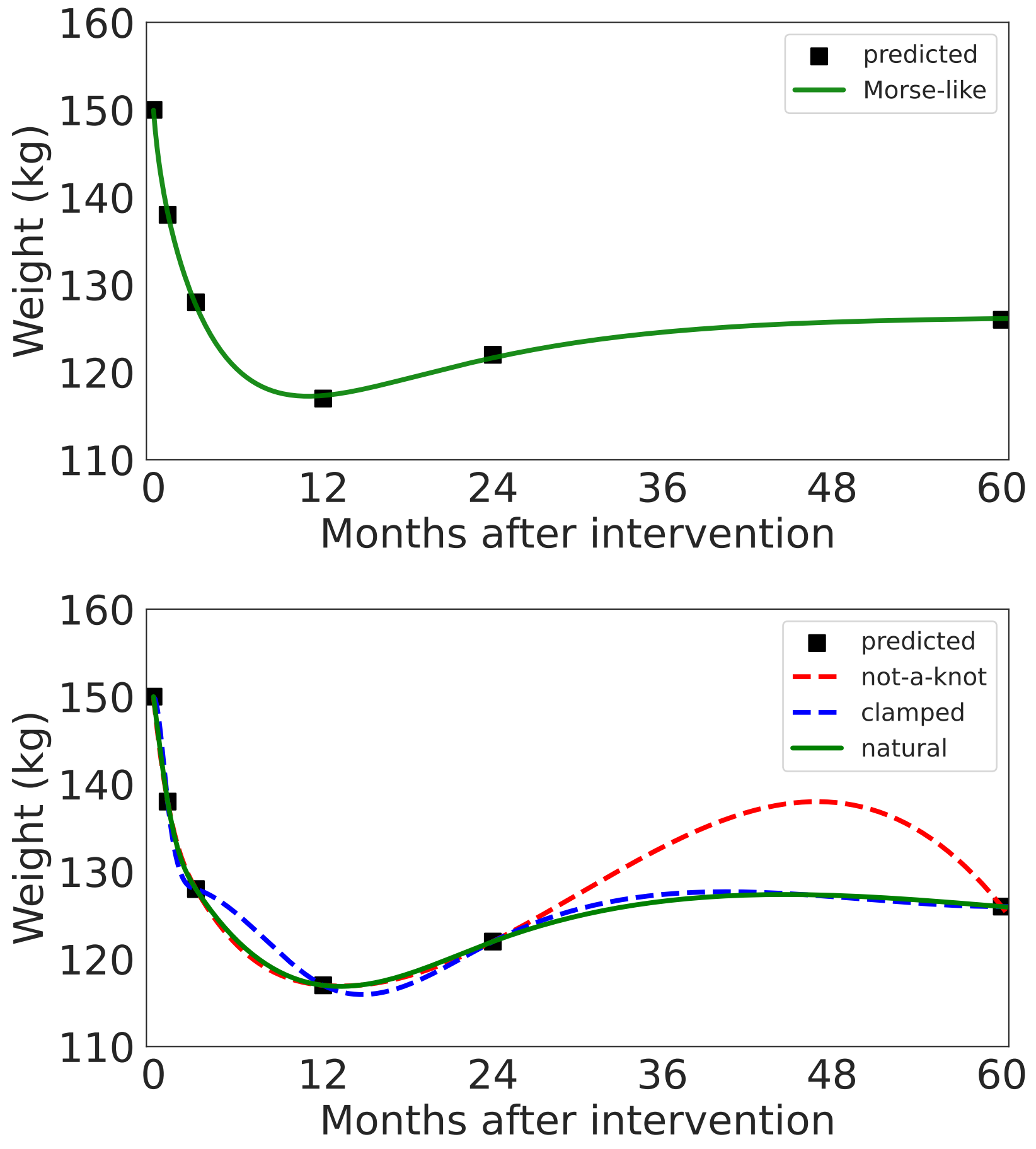

**Figure G.1** – Smoothed trajectories for the predicted weight after a sleeve gastrectomy for a non-smoker 65 year old patient with preoperative 150 kg, height 170 cm and type 2 diabetes for 20 years.

**Smoothing of other metrics.** For now, we have introduced the Morse-like model for the smoothing of weights. Another desirable feature of this model is that it is stable under the transformation that maps weights to the other metrics of interest (BMI, TWL, EWL). In particular, if $F_{\theta^\star}$ is a fitted Morse-like model for weights, we define

$$\theta^\star_{\mathrm{TWL}} = (-\frac{\alpha}{W_0}, -\frac{\beta}{W_0}, \gamma, 1-\Delta, t^\star, \tau)\,, \tag{G.11}$$

$$\theta^\star_{\text{BMI}} = (\frac{\alpha}{h_0^2}, \frac{\beta}{h_0^2}, \gamma, \frac{\Delta}{h_0^2}, t^\star, \tau), \tag{G.12}$$

$$\theta^\star_{\text{EWL}} = (-\frac{\alpha}{h_0^2(\text{BMI}_0 - 25)}, -\frac{\beta}{h_0^2(\text{BMI}_0 - 25)}, \gamma, \frac{\text{BMI}_0}{\text{BMI}_0 - 25} - \frac{\Delta}{h_0^2(\text{BMI}_0 - 25)}, t^\star, \tau). \tag{G.13}$$

Then $F_{\theta^\star_{\text{TWL}}}$ is a Morse-like model for TWL and, assuming the height is constant through time ($h_t = h_0$), $F_{\theta^\star_{\text{BMI}}}$ and $F_{\theta^\star_{\text{EWL}}}$ are also Morse-like models for BMI and EWL respectively. We acknowledge that height may be affected by surgery (significant weight loss may increase height by a few centimetres, for instance due to improved posture), however this (rather small) effect is often neglected.

## G.3 Cohorts

**Training cohorts.** The ABOS cohort (*Atlas Biologique de l'Obésite Sévère*; NCT01129297), is a prospective study conducted in Lille, France, aiming to identify the determinants of the outcome of bariatric surgery. All ABOS participants who underwent their first bariatric surgery at Lille University Hospital were enrolled in the present study (Figure G.2). Patients with gastric bypass, sleeve gastrectomy and gastric banding recruited from 2006 through 2020 corresponding to a follow-up data duration of 60 months were included in this analysis. Informed consent was obtained from all participants. Demographic characteristics, anthropomorphic measurements, medical history, medication use, and clinical laboratory tests were prospectively collected before surgery. A 75 g oral glucose tolerance test (OGTT) was performed after overnight fasting at baseline. Diabetes status was defined at baseline, based upon the previous history of diabetes, use of antidiabetic medication, and/or fasting plasma glucose (FBG) $\geqslant$ 126 mg/dL (7.0 mmol/L) or a 2 hour plasma glucose $\geqslant$ 200 mg/dL (11.1 mmol/L) during the OGTT or HbA1c $\geqslant$ 6.5% (48 mmol/L) (on the Diagnosis and of Diabetes Mellitus, 2003). Homeostasis model assessment (HOMA) 2 estimates of $\beta$-cell function (HOMA2-B) and insulin resistance (HOMA2-IR) based on C-peptide concentrations (which performs better than insulin in patients with diabetes) was calculated with the HOMA calculator (University of Oxford, Oxford, UK).

Since ABOS enrolled a majority of patients with RYGB and AGB, we combined the data with those from the BAREVAL cohort from Montpellier (NCT02310178), a sleeve specific cohort to constitute the training set. The BAREVAL cohort is a prospective cohort designed to evaluate the long-term effectiveness of sleeve gastrectomy on weight loss. Demographic characteristics, anthropomorphic measurements, medical history, medication use, and clinical laboratory tests were prospectively collected before surgery. Diabetes status was defined at baseline, based upon the previous history of diabetes, use of antidiabetic medication, and/or fasting plasma glucose (FBG) $\geqslant$ 126 mg/dL (7.0 mmol/L) or HbA1c $\geqslant$ 6.5% (48 mmol/L).1 Data was available at baseline and at 3, 12, 24, and 60 months follow-up. All patients that underwent primary bariatric surgery between 2014 and 2020 were selected.

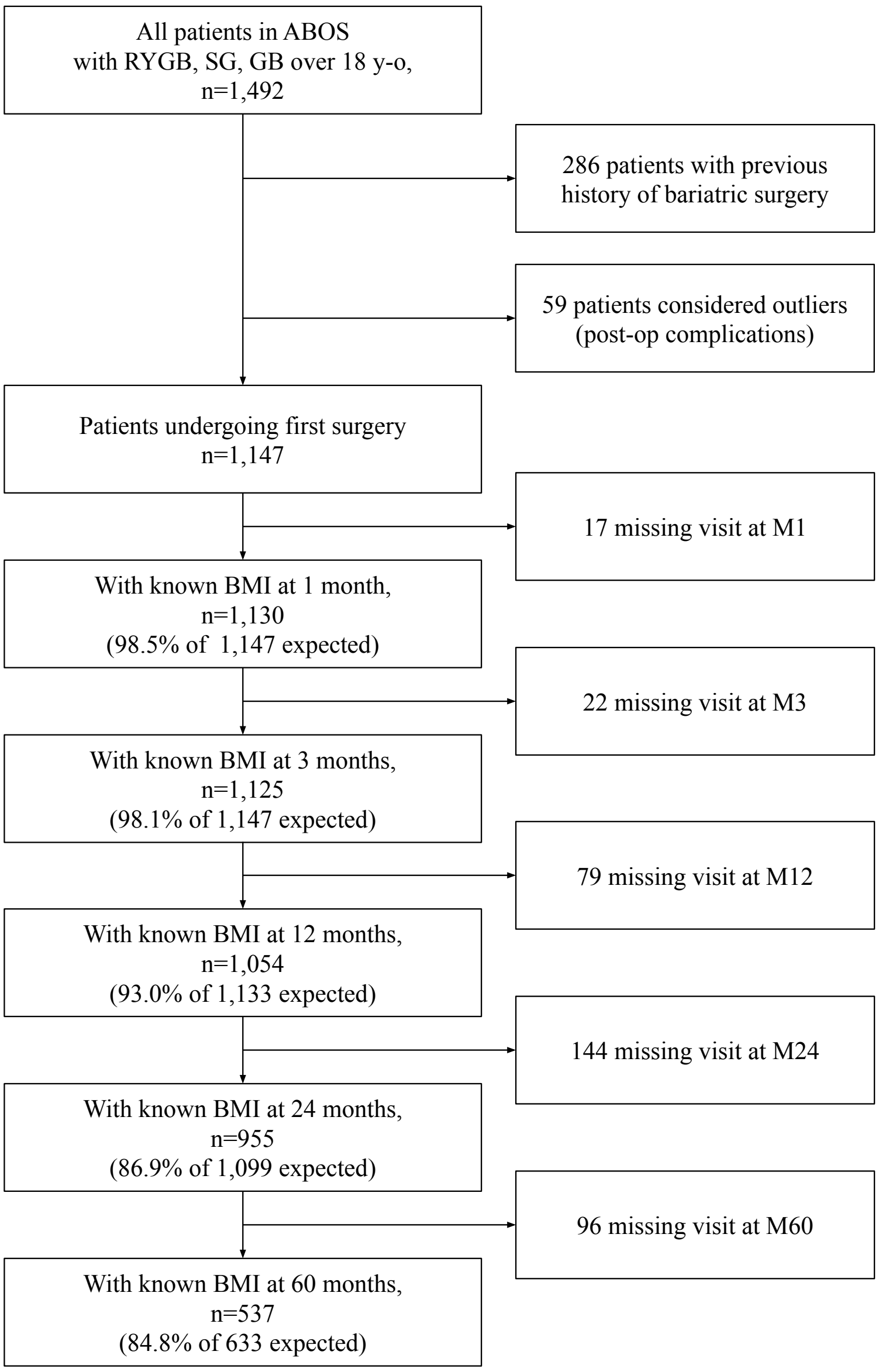

**Figure G.2** – Flowchart of the ABOS study. Expected: number of patients for which a visit is expected to be recorded in ABOS. For instance, patients who underwent surgery six months prior to the ABOS extraction date cannot have a M12 visit recorded in the database and thus do not count as expected at M12. We used the following cutoff durations after intervention: 20 days at M1, 65 days at M3, 300 days at M12, 630 days at M24. For M60, we extended the cutoff to 6.5 years to reflect the postponed visits during the COVID-19 pandemic.

**External testing cohorts.** Datasets from the validation centers were prospectively collected as part of registered prospective cohort studies, and stored in the centers' databases in compliance with local and national regulatory requirements. Analyses were done locally (NOK and SOS study cohorts) and aggregated results were sent to Lille University Hospital. For other cohorts, pseudonymised data were received and analysed in Lille, in accordance with the General Data Protection Regulation (GDPR) recommendations. The validation cohorts followed the legal and ethical rules applied in each country. Informed consent was obtained from all participants.

The Dutch Obesity Clinic (Nederlandse Obesitas Kliniek [NOK]) is the largest outpatient clinic for the surgery of obesity in the Netherlands with currently ten locations over the country and is treating circa 5000 patients yearly. All patients attend an extensive pre- and postoperative counselling programme, which is led by a multidisciplinary team, consisting of a physician, dietician, psychologist and physiotherapist. Current cohort is a retrospective cohort with prospectively collected data (Tettero et al., 2022). In order to provide the five year follow-up data, all patients that underwent primary bariatric surgery in 2015 and 2016 were selected. Patients with interventions other than gastric bypass, gastric sleeve or gastric band were excluded, as well as patients with missing type of intervention and missing baseline weight. Demographic characteristics, anthropometric measurements and medical history were collected at baseline. Diabetes was scored as present or absent by the treating physician, based on the medical history and use of antidiabetic drugs. Smoking at baseline was also scored as either present or absent. Weight was collected before surgery and at 3, 6, 12, 18, 24, 36, 48, and 60 months at regular postoperative follow-up visits.

The *Swedish Obesity Subjects* (SOS) study is a non-randomised, controlled intervention study investigating the long-term effects of bariatric surgery. The study has been described in detail elsewhere (Sjöström et al., 1992). In brief, 2007 participants who underwent bariatric surgery and 2040 matched controls were recruited between September 1, 1987 and January 31, 2001. The inclusion criteria were age 37 to 60 years and body mass index (BMI) of 34 or more for men and 38 or more for women. Exclusion criteria were identical in the surgery and control groups and were selected to ensure all participants were eligible for surgery. Oral or written informed consent was obtained from all study participants. Participants in the surgery group underwent gastric banding, vertical banded gastroplasty or gastric bypass. Participants in the control group received conventional obesity treatment. Participants in the surgery group underwent gastric banding, vertical banded gastroplasty or gastric bypass. Participants in the control group received conventional obesity treatment at their primary health care centers. The current analysis included only patients which underwent gastric banding (n=376) or gastric bypass (n=266). Participants that had a change in treatment type (i.e. band removal or conversion to a different surgery type) were censored on the corresponding date. Data was available at baseline and at 3, 12, 24, 48, and 72 months follow-up. Since no information about weight at 60 months (5 years) after surgery was available, we used the mean of the weights at 48 and 72 months as a proxy. Type 2 diabetes status at baseline was defined based on previous diagnosis of diabetes and/or use of antidiabetic medication and/or fasting plasma glucose $\geqslant 7$ mmol/L and/or HbA1c $\geqslant 48$ mmol/L.

The *Projet régional de Recherche en Obésité Sévère* (PRECOS; NCT03517072) study is a multi-centre prospective cohort study, conducted in Arras, Boulogne sur/ Mer, Lille and Valenciennes to explore the determinants of 5 year-weight loss following bariatric surgery. Demographic characteristics, anthropomorphic measurements, medical history, medication use, and clinical laboratory tests were prospectively collected before surgery. Diabetes status was defined at baseline, based upon the previous history of diabetes, use of antidiabetic medication, and/or fasting plasma glucose (FBG) $\geqslant$ 126 mg/dL (7.0 mmol/L) and/or HbA1c $\geqslant$ 6.5% (48 mmol/L).1 Data was available at baseline and at 3, 12, 24, and 60 months follow-up.

The Roma cohort (Mingrone et al., 2021) was collected from *Randomised clinical study comparing the effect of Roux-en-Y gastric bypass and Sleeve Gastrectomy on reactive hypoglycemia* (NCT01581801) and from *Diet and Medical Therapy Versus Bariatric Surgery in Type 2 Diabetes* (DIBASY; NCT00888836). All subjects from these cohorts were enrolled for metabolic-bariatric surgery between 2009 and 2016 at Fondazione Policlinico Universitario A. Gemelli IRCCS - Catholic University of the Sacred Heart in Rome. The studies were approved by the Ethical Committee of the Fondazione Policlinico Universitario A. Gemelli IRCCS in Rome. Written informed consent was obtained from all participants. The weight data were obtained from 2009 to 2020 corresponding to a follow-up duration of 60 months. Demographic characteristics, anthropometric measurements, medical history, medication use and clinical laboratory tests were prospectively collected before and after surgery. A 75 g oral glucose tolerance test (OGTT) was performed after overnight fasting at baseline in all subjects and also one year after surgery in participants at NCT01581801. Body composition was assessed by dual energy X-ray absorptiometry (Lunar-iDXA). Diagnosis of Type 2 Diabetes (T2D) was made at baseline taking into account the previous history of diabetes, use of antidiabetic medication, and/or fasting plasma glucose (FPG) $\geqslant$ 126 mg/dL (7.0 mmol/L) or a 2-hour plasma glucose $\geqslant$ 200 mg/dL (11.1 mmol/L) during the OGTT and/or HbA1c $\geqslant$ 6.5% (48 mmol/L). Remission of T2D was defined as a FPG $\leqslant$ 100 mg/dL (5.6 mmol/L) and a HbA1c $\leqslant$ 6.5% (48 mmol/L) for at least 1 year without active pharmacologic therapy.

The Center for the treatment of Obesity and Diabetes (COD) cohort comprised 121 patients with T2D and obesity who received RYGB at the Oswaldo Cruz German Hospital, in Sao Paulo, Brazil between 2008 and 2016, and attended clinical and biological assessment at baseline and after one year of follow-up, as previously reported (Cohen et al., 2020, 2022). All patients provided written informed consent prior to inclusion.

The Mexico cohort was collected from *Long versus short biliopancreatic limb in Roux-en-Y gastric bypass: short-term results of a randomised clinical trial* (NTC04609449), a randomised study with patients undergoing Roux-en-Y gastric bypass at a single academic center in Mexico city from 2016 to 2018 (Zerrweck et al., 2021). The study was approved by National and Institutional ethical Committees.

**Randomised control trials.** In 2018, two RCT comparing laparoscopic SG and RYGB with a long-term follow-up rate of more than 80 per cent at 5 years were published. The Finnish SleevePass (Sleeve versus Bypass) equivalence study 10 enrolled 240 patients with percentage excess weight loss as the primary endpoint (Salminen et al., 2018, 2022). The Swiss SM-BOSS (Swiss Multicentre Bypass or Sleeve Study) trial 11 included 217 patients with percentage excess BMI loss (%EBMIL) as the primary endpoint (Peterli et al., 2018). Both trials showed better weight loss for RYGB. In the SleevePass trial the difference was not statistically significant based on the prespecified equivalence margins. In the SM-BOSS trial, the difference was also not statistically significant after adjustments for multiple comparisons. The trials were designed as two-group, randomised, controlled, multicentre studies comparing SG with RYGB, involving 240 patients with severe obesity from Finland and 225 from Switzerland (Wölnerhanssen et al., 2021). Study protocols were reviewed and approved by the local ethical committees of each participating hospital. Both RCT were conducted in accordance with the principles of the Declaration of Helsinki and registered at the clinical trials registry of the National Institutes of Health (NCT00356213, NCT00793143). Additional patient-level data were included, either prospectively collected unpublished 5-year data or information retrieved retrospectively from patient files. To ensure standardisation of outcomes between the studies, all parameters and variables were translated into English to ensure identical definitions in both trials. T2D remission was defined according to the American Diabetes Association criteria 14: complete remission by haemoglobin (Hb) A1c level below 6.0 per cent and fasting glucose concentration less than 100 mg/dL (below 5.6 mmol/l) for at least 1 year in the absence of active pharmacological therapy, and partial remission by HbA1c level below 6.5 per cent and fasting glucose concentration 100–125 mg/dl (5.6–6.9 mmol/l) for at least 1 year in the absence of active pharmacological therapy.

## G.4 Model development

To derive the model, the training cohort was divided into two subsets: a training subset consisting of 80% of randomly selected patients, and an internal testing subset comprising the remaining 20% of patients.

We first performed a preprocessing of all patients' features (Appendix G.1). As the ABOS cohort had a large number of preoperative attributes per patient (Appendix Table 4), we ran a feature selection algorithm on this patient subgroup to extract the most statistically relevant ones concerning outcome prediction. We used the Least Absolute Shrinkage and Selection Operator (LASSO) with a cross-validated regularisation parameter to limit overfitting, as well as the hierarchical group LASSO to account for interactions. The LASSO were trained on each outcome visit separately (months M1, M3, M12, M24 and M60).

For outcome prediction, we further leveraged a class of machine learning algorithms called *decision trees* to learn meaningful subgroups of patients that share statistical similarities in their baseline characteristics, and second, to fit a TWL prediction model for each subgroup. In contrast with classical regression analysis, such algorithms are valuable for learning hierarchical stratification of patients from baseline characteristics and observed outcomes without supervision. For instance, decision trees can predict weight loss when they are trained on a heterogeneous cohort of different bariatric interventions such as RYGB, SG and AGB, according to the type of intervention as well as using other variables such as the age at intervention, BMI and other clinical features.

For calibration, we used the LASSO-extracted features as input for the *classification and regression trees* (CART) algorithm (Breiman et al., 1984). CART is a nonparametric statistical method used for prediction analysis of categorical (classification) and continuous (regression) response variables and for explanatory variables which may consist of nominal, ordinal, or continuous features. It is designed to create a sequence of binary choices according to attribute categories for categorical features, or based on a calibrated threshold value for numerical features. The CART algorithm recursively decides which attribute, category, or threshold value to use, and in which order, by maximising the remaining statistical dispersion (variance) between the two subtrees that emerge from this choice. CART identifies groups of patients who share the same binary choices, and fits a single outcome prediction for each cluster. As weight loss outcomes are continuous, CART produced here a regression tree. A workflow diagram of the machine learning process is displayed in Figure G.3.

The algorithm was calibrated on the training subset of the training cohort. We further compared the predicted TWL to the observed outcomes of patients in the testing subset of the training cohort (internal validation).

The external validation consisted of all remaining eight validation cohorts (by decreasing sample size: NOK, SGH, SOS, PRECOS, Roma, Lyon, COD and Mexico) and the two RCT cohorts (SleevePass and SM-BOSS). In addition to computing the performance metrics (MAD and RMSE), we report Bland-Altman plots (Altman and Bland, 1983; Bland and Altman, 1986) in Figure G.4. These plots allow for a quick visual inspection of the calibration of a model by comparing the average of predicted and observed outcome to their difference: the "flatter" the point cloud, the better the calibration. This analysis is further completed with the mean difference (zero indicates the absence of systematic bias) and confidence bands. In our case, these plots do not reveal any major bias at M12, M24 and M60.

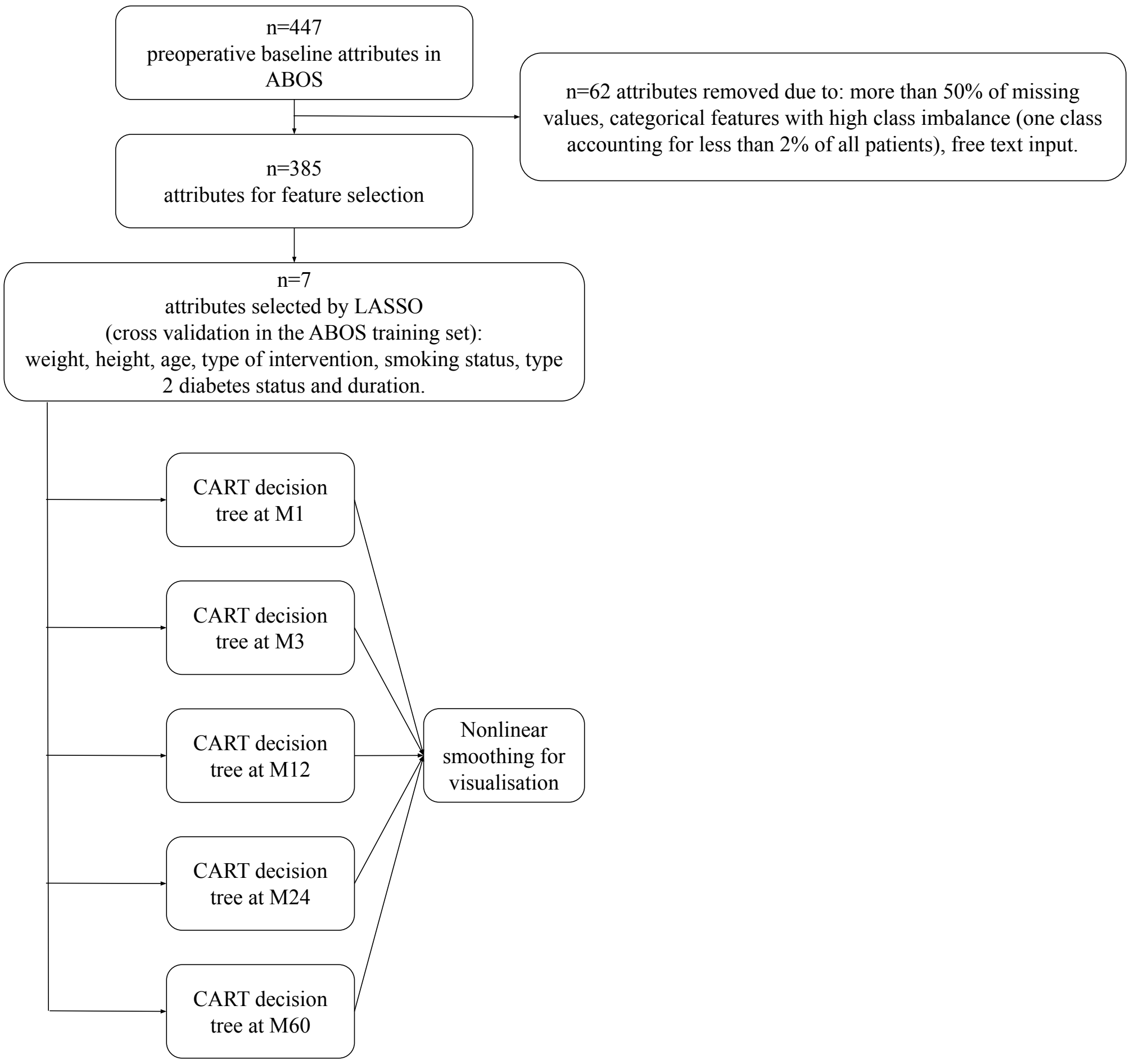

**Figure G.3 –** Machine learning pipeline for the prediction of postoperative weight trajectories.

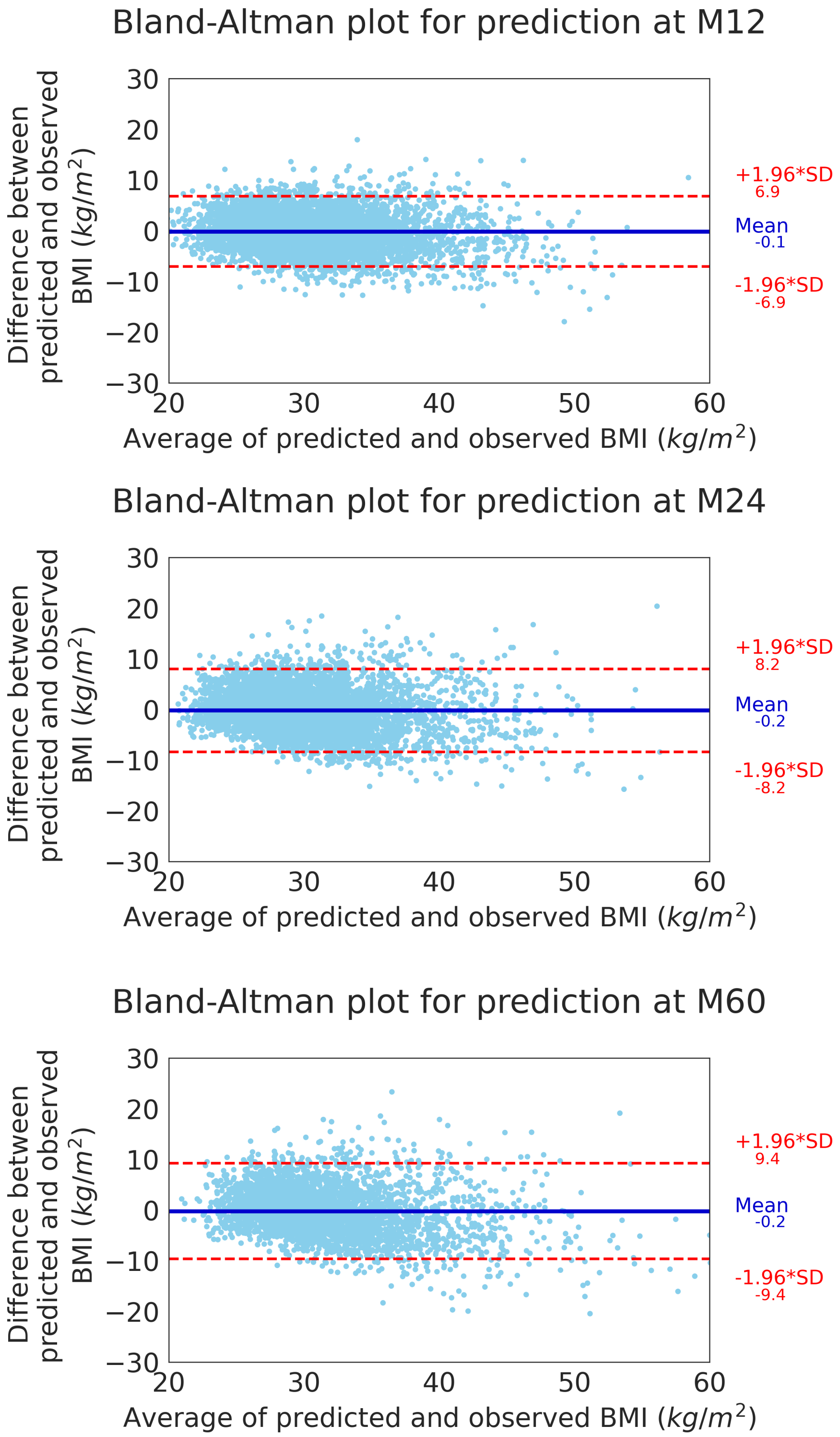

**Figure G.4 –** Bland-Altman plots for the BMI predictions at M12, M24 and M60.

## G.5 Comparison with other methods

In parallel with our LASSO+CART model, we also explored other methods for prediction. First, we tried simple linear regression, based on the 7 selected attributes. Second, we tried to take into account the longitudinal nature of the data by using mixed effect model, with different kernel functions for the time component (power functions, restricted cubic splines), as well as (fixed or random) intercept and slopes, to model the predicted weight loss after bariatric surgery from the preoperative data. The best mixed effect model uses the seven selected attributes, a restricted cubic splines to model time after surgery, as well as a random intercept and slope.

Furthermore, CART-trained regression trees are able to achieve variable selection by themselves using a pruning mechanism, such as cost complexity pruning. Hence, we also applied CART with all baseline variables and then pruning the resulting trees; it produced similar trees to the LASSO+CART approach (in particular all trees started by splitting on the type of intervention) and selected a total of six variables: weight, height, age, T2D status and duration and type of intervention. In terms of selected features, the only difference between LASSO and CART was the smoking history. Last, because decision trees are typically weak predictors and that bagging (random forests [RF]) is a natural step to try and improve accuracy, at the cost of less interpretability, we also tried RF on our training set. [2]. Hyperparameters, optimised with 3-fold cross validation, included the maximum depth, maximum number of features, minimum number of samples at a leaf node, number of tree predictors and the cost-complexity pruning parameter. Figure G.5 displays the results of these comparisons.

In conclusion, linear models, mixed effect models and CART with pruning are underperforming compared to CART decision trees, in terms of MAD error, whereas RFs resulted in only slightly smaller MAD than CART (in the sense that confidence intervals largely overlapped), which we believe does not justify the loss of interpretability.

[2]We used the RandomForestRegressor implementation from the emphscikit-learn library in Python.

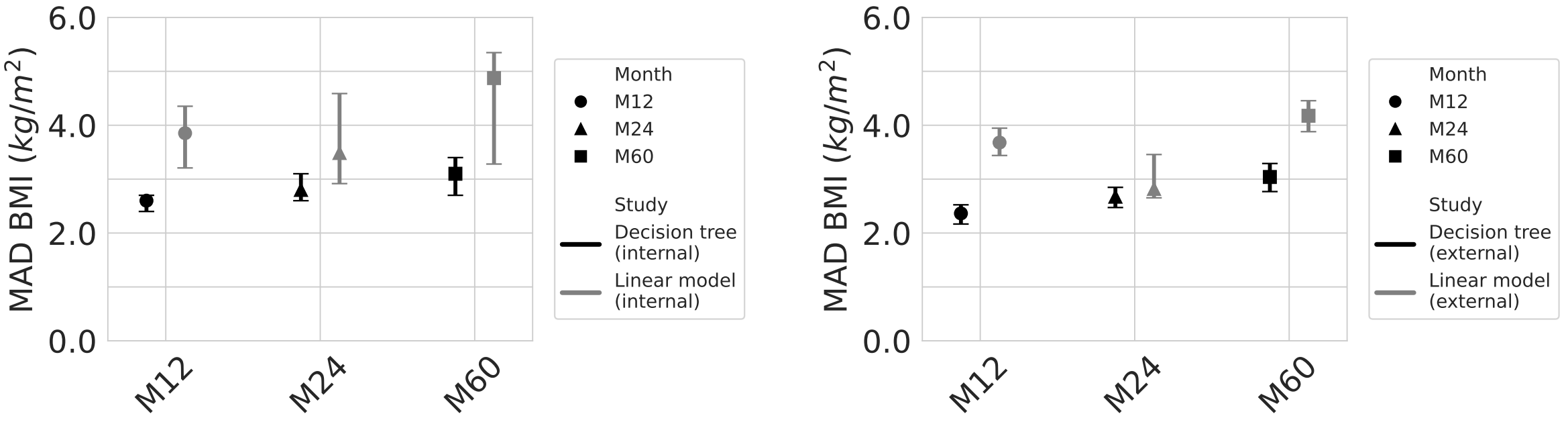

(**a**) Linear regression.

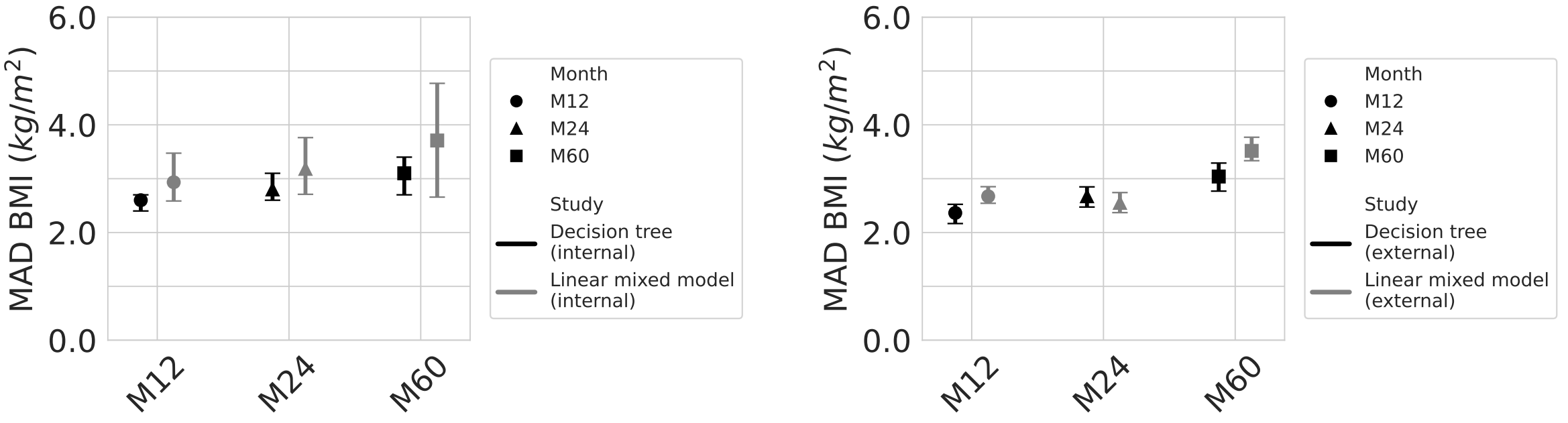

(**b**) Linear mixed effect model.

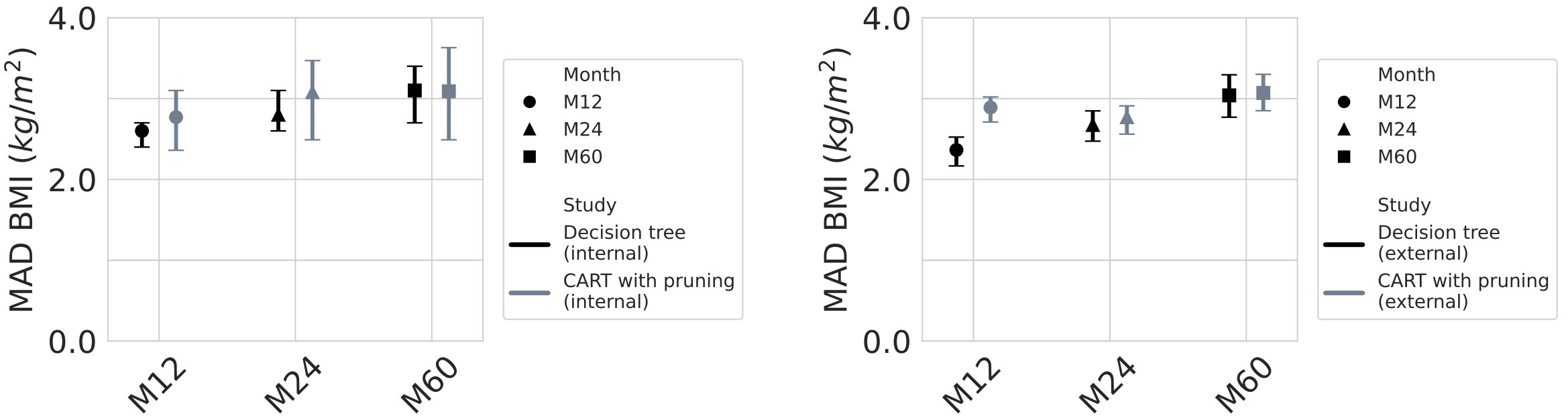

(**c**) CART with pruning on all baseline variables (no LASSO).

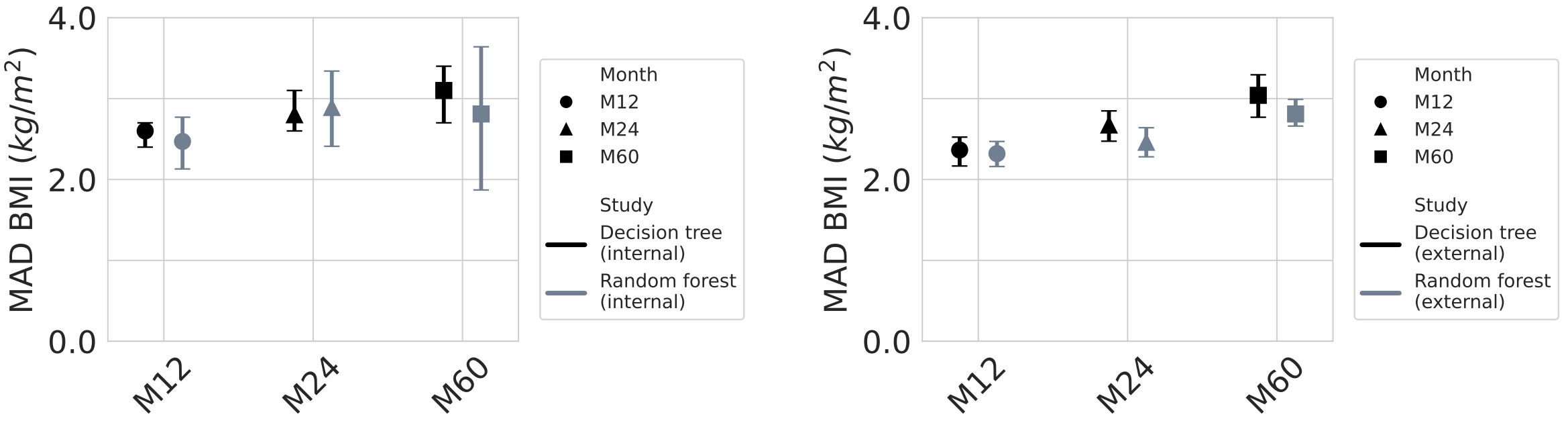

(**d**) Random forests.

**Figure G.5 –** MAD of predicted BMI outcomes of our model versus various benchmarks on the internal test set (left) and the external validation cohorts (right).

## G.6 Comparison with previously published models

We have replicated in ABOS the results of previously published models (Cottam et al., 2018; Goulart et al., 2016; Janik et al., 2019; Velázquez-Fernández et al., 2019; Wise et al., 2016; Seyssel et al., 2018; Baltasar et al., 2011) between January 2000 and December 2021, identified using the following search terms: bariatric surgery, postoperative weight loss, weight loss prediction, and prediction model. We included English language studies that investigated RYGB, SG, and AGB and used a prospective or retrospective design. The identified models were implemented at the appropriate postoperative times to predict the TWL of corresponding patients from the training cohort, allowing the calculation of MAD and RMSE as previously described. In contrast with our tree-based model, which can simultaneously predict weight loss at different dates and across multiple types of bariatric surgery, the existing models were restricted to a single outcome date and a single type of intervention. For a fair comparison, these existing models were compared with corresponding subgroup of ABOS cohort according to type of surgery and outcome date. Results are reported in Table G.1.

To compare two given models, their respective RMSE were submitted to the Diebold-Mariano test (Harvey et al., 1997), the p-value (for the null hypothesis that both RMSE are equal) of which are reported in Table G.2. Six out of twelve models had significantly higher RMSE (at level $0.05$), and most of the others had higher RMSE, albeit not significantly. We emphasise here that the comparisons are performed on small subgroups of the internal test set of ABOS in order to match the operation and month of prediction of each of the benchmark model, resulting in small sample size and correspondingly small statistical power.

In conclusion, not only is our model not limited to a single date nor operation, it is also more accurate than previously published prediction models.

**Table G.1** – Validation of existing prediction models on the ABOS cohort for corresponding type of intervention and date of prediction.

| Model | Type of operation | Month of prediction | Sample size | Mean observed BMI (SD) | Mean predicted BMI (SD) | RMSE in $kg/m^2$ (95% CI) | Normalised RMSE in % (95% CI) |
|---|---|---|---|---|---|---|---|
| Cottam<br>Cottam et al. (2018) | SG | 12 | 193 | 35.2<br>(7.9) | 33.2<br>(5.9) | 4.7<br>(4.3; 5.2) | 13.3<br>(11.6; 14.9) |
| Goulart<br>Goulart et al. (2016) | SG | 12 | 193 | 35.2<br>(7.9) | 31.5<br>(5.6) | 5.6<br>(5.2; 6.2) | 16.0<br>(14.3; 18.0) |
| Janik<br>Janik et al. (2019) | SG | 12 | 193 | 35.2<br>(7.9) | 31.0<br>(7.7) | 6.1<br>(5.6; 6.6) | 17.4<br>(15.7; 19.0) |
| Velazquez-1<br>Velázquez-Fernández et al. (2019) | RYGB | 12 | 652 | 32.5<br>(6.1) | 31.9<br>(4.5) | 3.8<br>(3.6; 4.0) | 11.8<br>(11.0; 12.6) |
| Wise<br>Wise et al. (2016) | RYGB | 12 | 652 | 31.4<br>(5.9) | 30.2<br>(3.9) | 3.9<br>(3.5; 4.3) | 12.4<br>(11.3; 13.6) |
| Seyssel-1<br>Seyssel et al. (2018) | RYGB | 12 | 652 | 32.5<br>(6.1) | 31.5<br>(4.3) | 3.8<br>(3.5; 4.0) | 11.7<br>(10.8; 12.5) |
| Baltasar-1<br>Baltasar et al. (2011) | SG | 24 | 162 | 35.1<br>(7.9) | 31.1<br>(3.8) | 7.0<br>(6.3; 7.6) | 20.0<br>(17.7; 22.2) |
| Baltasar-2<br>Baltasar et al. (2011) | Any | 24 | 959 | 33.5<br>(7.3) | 30.5<br>(2.9) | 6.8<br>(6.4; 7.0) | 20.2<br>(19.0; 21.3) |
| Baltasar-3<br>Baltasar et al. (2011) | RYGB | 24 | 606 | 31.8<br>(6.5) | 30.7<br>(3.0) | 5.2<br>(4.8; 5.5) | 16.3<br>(15.1; 17.5) |
| Velazquez-2<br>Velázquez-Fernández et al. (2019) | RYGB | 24 | 606 | 31.8<br>(6.5) | 31.3<br>(3.9) | 4.9<br>(4.6; 5.2) | 15.5<br>(14.5; 16.5) |
| Seyssel-2<br>Seyssel et al. (2018) | RYGB | 24 | 606 | 31.8<br>(6.5) | 31.0<br>(3.6) | 4.9<br>(4.7; 5.2) | 15.5<br>(14.3; 16.5) |
| Seyssel-3<br>Seyssel et al. (2018) | RYGB | 60 | 348 | 34.1<br>(6.9) | 33.3<br>(3.7) | 5.4<br>(5.0; 5.7) | 15.8<br>(14.3; 17.3) |

**Table G.2 –** Comparison of existing prediction models versus our model on the ABOS testing subset.

| Model | Type of operation | Month of prediction | Sample size | RMSE in kg/m$^2$ (95% CI) of literature model | RMSE in kg/m$^2$ (95% CI) of our model | p-value |
|---|---|---|---|---|---|---|
| Cottam<br>Cottam et al. (2018) | SG | 12 | 41 | 5.3<br>(3.8; 6.7) | 5.1<br>(4.1; 6.1) | 0.22 |
| Goulart<br>Goulart et al. (2016) | SG | 12 | 41 | 6.3<br>(4.8; 7.8) | 5.1<br>(4.0; 6.0) | 0.011 |
| Janik<br>Janik et al. (2019) | SG | 12 | 41 | 6.3<br>(5.6; 7.2) | 5.1<br>(4.1; 6.1) | 0.011 |
| Velazquez-1<br>Velázquez-Fernández et al. (2019) | RYGB | 12 | 124 | 4.0<br>(3.5; 4.5) | 3.8<br>(3.3; 4.2) | 0.13 |
| Wise<br>Wise et al. (2016) | RYGB | 12 | 60 | 3.5<br>(2.9; 3.9) | 3.6<br>(3.1; 4.1) | 0.64 |
| Seyssel-1<br>Seyssel et al. (2018) | RYGB | 12 | 124 | 4.1<br>(3.5; 4.8) | 3.8<br>(3.3; 4.2) | 0.11 |
| Baltasar-1<br>Baltasar et al. (2011) | SG | 24 | 33 | 6.6<br>(5.6; 7.6) | 5.5<br>(4.3; 6.7) | 0.099 |
| Baltasar-2<br>Baltasar et al. (2011) | Any | 24 | 187 | 6.8<br>(6.0; 7.4) | 4.7<br>(4.0; 5.2) | $<$0.0001 |
| Baltasar-3<br>Baltasar et al. (2011) | RYGB | 24 | 111 | 5.1<br>(4.3; 5.9) | 4.3<br>(3.4; 5.1) | 0.009 |
| Velazquez-2<br>Velázquez-Fernández et al. (2019) | RYGB | 24 | 111 | 4.8<br>(4.1; 5.6) | 4.2<br>(3.7; 5.0) | 0.002 |
| Seyssel-2<br>Seyssel et al. (2018) | RYGB | 24 | 111 | 4.9<br>(4.1; 5.7) | 4.3<br>(3.6; 4.9) | 0.007 |
| Seyssel-3<br>Seyssel et al. (2018) | RYGB | 60 | 55 | 4.3<br>(3.7; 4.8) | 4.0<br>(3.0; 4.7) | 0.12 |

# List of Figures

# List of Algorithms

# List of Tables

*Enfoui dans son fauteuil, il ruminait maintenant sur cette expresse observance qui bouleversait ses plans, rompait les attaches de sa vie présente, enterrait ses projets futurs. Ainsi, sa béatitude était finie ! ce havre qui l'abritait, il fallait l'abandonner, rentrer en plein dans cette intempérie de bêtise qui l'avait autrefois battu ! Les médecins parlaient d'amusement, de distraction ; et avec qui, et, avec quoi, voulaient-ils donc qu'il s'égayât et qu'il se plût ?*

— Joris-Karl Huysmans, *À rebours*

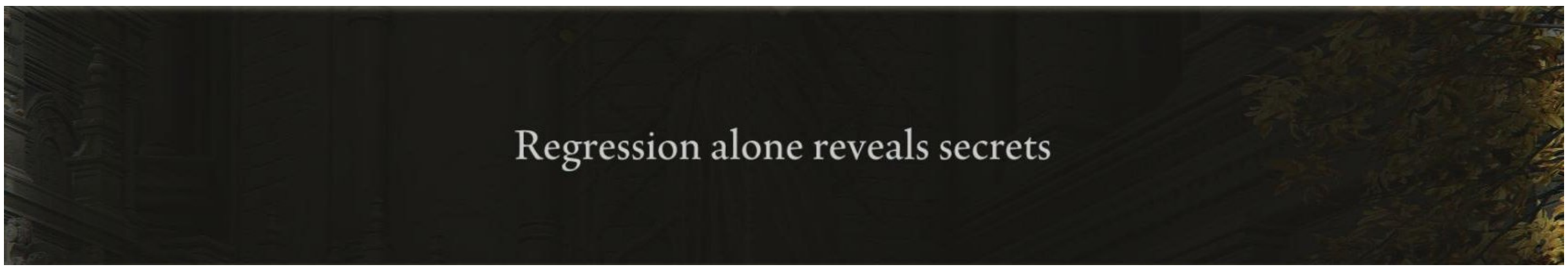